\documentclass{article}

\usepackage{microtype}
\usepackage{graphicx}
\usepackage{subcaption}
\usepackage{booktabs}

\usepackage{hyperref}
\usepackage{colortbl}

\usepackage[preprint]{icml2026}

\usepackage{amsmath}
\usepackage{amssymb}
\usepackage{mathtools}
\usepackage{amsthm}
\usepackage{multirow}
\usepackage{tcolorbox}
\usepackage{tabularx}
\newcolumntype{Y}{>{\centering\arraybackslash}X}

\usepackage{graphicx}
\usepackage{booktabs}
\usepackage{longtable}
\usepackage{array}
\usepackage{enumitem}

\usepackage{tikz}
\usetikzlibrary{tikzmark}

\usepackage{titletoc}

\newcommand{\na}{%
    \multicolumn{2}{c}{%
        \begin{tikzpicture}[baseline=(char.base)]
            \def\boxW{2.5cm}
            \def\boxH{1.2cm}
            
            \draw[dashed, gray!60, thin, dash pattern=on 3pt off 2pt] 
                (-\boxW/2, -\boxH/2) rectangle (\boxW/2, \boxH/2);
            
            \node (char) at (0,0) {\textcolor{gray!60}{\footnotesize \textit{— N/A —}}};
        \end{tikzpicture}%
    }%
}

\usepackage[capitalize,noabbrev]{cleveref}

\theoremstyle{plain}
\newtheorem{theorem}{Theorem}[section]
\newtheorem{proposition}[theorem]{Proposition}

\theoremstyle{definition}

\theoremstyle{remark}

\usepackage[textsize=tiny]{todonotes}

\icmltitlerunning{Spectral Evolution Search: Efficient Inference-Time Scaling for Reward-Aligned Image Generation}

\begin{document}

\twocolumn[
  \icmltitle{Spectral Evolution Search: Efficient Inference-Time Scaling for \\ Reward-Aligned Image Generation}
  \icmlsetsymbol{equal}{*}

  \begin{icmlauthorlist}
    \icmlauthor{Jinyan Ye}{ecnu}
    \icmlauthor{Zhongjie Duan}{alibaba}
    \icmlauthor{Zhiwen Li}{ecnu}
    \icmlauthor{Cen Chen}{ecnu}
    \icmlauthor{Daoyuan Chen}{alibaba}
    \icmlauthor{Yaliang Li}{alibaba}
    \icmlauthor{Yingda Chen}{alibaba}
  \end{icmlauthorlist}

  \icmlaffiliation{ecnu}{School of Data Science and Engineering, East China Normal University, Shanghai, China}
  \icmlaffiliation{alibaba}{Alibaba Group, Hangzhou, China}

  \icmlcorrespondingauthor{Cen Chen}{cenchen@dase.ecnu.edu.cn}

  \icmlkeywords{Diffusion Model, Inference-Time Scaling, Noise Optimization}

  \vskip 0.3in
]

\printAffiliationsAndNotice{}

\begin{abstract}
Inference-time scaling offers a versatile paradigm for aligning visual generative models with downstream objectives without parameter updates. However, existing approaches that optimize the high-dimensional initial noise suffer from severe inefficiency, as many search directions exert negligible influence on the final generation. We show that this inefficiency is closely related to a spectral bias in generative dynamics: model sensitivity to initial perturbations diminishes rapidly as frequency increases. Building on this insight, we propose Spectral Evolution Search (SES), a plug-and-play framework for initial noise optimization that executes gradient-free evolutionary search within a low-frequency subspace. Theoretically, we derive the Spectral Scaling Prediction from perturbation propagation dynamics, which explains the systematic differences in the impact of perturbations across frequencies. Extensive experiments demonstrate that SES significantly advances the Pareto frontier of generation quality versus computational cost, consistently outperforming strong baselines under equivalent budgets.

\end{abstract}

\section{Introduction}

Modern text-to-image models have achieved remarkable progress driven by training-time scaling laws \cite{kaplan2020scaling,hoffmann2022training,podell2023sdxl,esser2024scaling}. However, further aligning these foundation models with diverse downstream objectives, such as semantic adherence, aesthetic quality, or human preference, remains a challenge. Conventional approaches rely on fine-tuning \cite{black2023training, clark2023directly, wallace2024diffusion, liang2025aesthetic}, which is inherently inefficient as it requires retraining a dedicated model for each specific objective. This limitation has directed increasing research attention towards \textit{Inference-time Scaling} \cite{snell2024scaling, brown2024large}. Instead of altering model weights, this paradigm translates additional computational budget into alignment gains, enabling a static model to adapt to specific objectives during sampling.

Current inference-time scaling methods struggle to balance generality with efficient budget-to-alignment conversion. 
First, trajectory optimization \cite{li2024derivative,kim2025test,jain2025diffusion} relies on intrusive interventions in the denoising path. This coupling with Stochastic Differential Equation (SDE) solvers hinders adaptation to efficient Ordinary Differential Equation (ODE)-based samplers or Flow Matching models. 
Second, gradient-based optimization \cite{wallace2023end, tang2024inference} requires differentiable objectives. This constraint precludes the use of non-differentiable human preference metrics. 
To bypass these limits, initial noise search and optimization \cite{ma2025scaling} treats inference as a black-box search. 
While broadly compatible, these methods rely on an implicit isotropic parameterization treating all search directions equally. This assumption causes severe computational redundancy. Budgets are wasted on perturbations with negligible visual impact, leading to diminishing returns that limit scalability.

\begin{figure}[t]
	\centering
	\includegraphics[width=\linewidth]{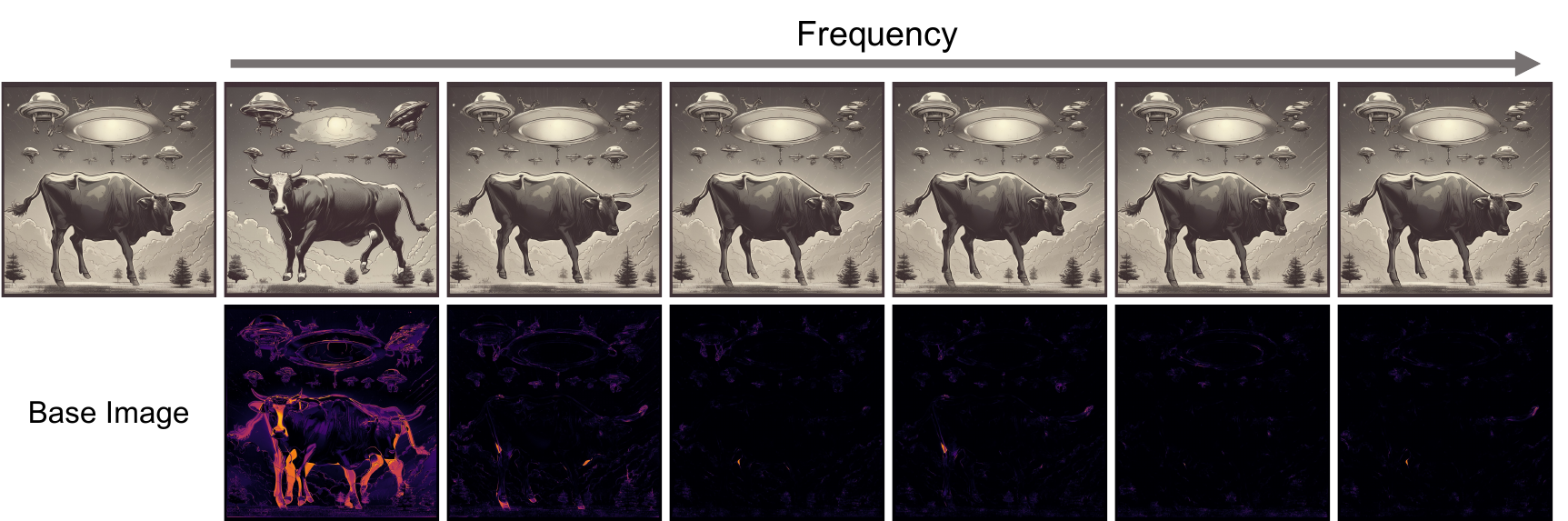} 
    \caption{Visualization of spectral bias. We inject band-pass perturbations of constant energy ($||\boldsymbol{\xi}||_2=1$) into the initial noise across increasing frequency bands (left to right). The top row displays the resulting generated samples, while the bottom row visualizes the pixel-wise differences compared to the unperturbed baseline. Observe that low-frequency noise leads to significant structural changes, whereas high-frequency perturbations result in minimal visual difference.}
	\label{fig:visualizing_spectral}
    % \vspace{-1em}
\end{figure}

To unlock the potential of initial noise optimization, we argue that search must operate on a compact manifold of meaningful degrees of freedom.
As illustrated in Figure \ref{fig:visualizing_spectral}, qualitative analysis reveals a pronounced spectral bias in generative models: low-frequency perturbations significantly reshape image structure, whereas high-frequency perturbations of equivalent energy exert negligible visual impact. 
Consequently, we hypothesize that the effective control space for inference-time search is intrinsically sparse and concentrated within low-frequency components.
Guided by this insight, we introduce \textit{Spectral Evolution Search (SES)}, a plug-and-play framework designed for initial noise optimization. SES exploits wavelet transforms to decouple the search space, strictly constraining optimization to the low-frequency subspace via a cross-entropy evolutionary strategy.
To theoretically substantiate this spectral decoupling strategy, we derive the \textit{Spectral Scaling Prediction} from the perturbation propagation dynamics of generative flows, formally characterizing the differential gains of perturbations across frequency bands. Experimental results demonstrate that, under the same computational budget, SES achieves superior scaling behavior, significantly advancing the Pareto frontier of generation quality versus computational cost.

Our main contributions are summarized as follows: 
\begin{itemize}
    \item \textbf{Method.} We propose Spectral Evolution Search (SES), a gradient-free inference-time scaling framework designed for initial noise optimization. It is plug-and-play and broadly applicable across diverse models, samplers, and reward functions.
    \item \textbf{Theory.} We establish the \textit{Spectral Scaling Prediction} by deriving the perturbation propagation dynamics, revealing that the effective optimization landscape is intrinsically low-dimensional and spectrally biased.
    \item \textbf{Practice.} Extensive experiments across mainstream generative models and alignment tasks demonstrate that SES yields superior inference-time scaling behavior, significantly outperforming representative methods under identical compute budgets.
\end{itemize}

\section{Related Work}

\textbf{Inference-Time Scaling Strategies.} Current approaches fall primarily into three categories. 
(1) \textit{Trajectory Optimization.} These methods modify denoising trajectories via particle resampling \cite{wu2023practical,dou2024diffusion,kim2025test} or tree search \cite{jain2025diffusion}. However, such interventions rely on specific SDE solvers, hindering adaptation to efficient ODE-based samplers.
(2) \textit{Gradient-based Optimization.} Utilizing backpropagation to update noise \cite{wallace2023end,tang2024inference}, these approaches are constrained to differentiable rewards and susceptible to reward hacking, exploiting metrics at the expense of perceptual fidelity.
(3) \textit{Initial Noise Optimization.} This paradigm \cite{ma2025scaling,chen2024find,guo2024initno} optimizes initial noise. While compatible with non-differentiable rewards, current approaches face the curse of dimensionality, leading to inefficient search under isotropic parameterizations of high-dimensional noise spaces.

\section{Preliminaries}
\label{sec:preliminaries}

\textbf{Generative Flow as a Deterministic Mapping.} Modern visual generative models, including denoising diffusion \cite{ho2020denoising, rombach2022high} and flow matching \cite{lipman2022flow,esser2024scaling}, can be unified under the framework of the Probability Flow ODE. Despite differing training objectives, both paradigms essentially induce a deterministic trajectory that transports a prior noise distribution $p_0(\mathbf{x}) = \mathcal{N}(\mathbf{0}, \mathbf{I})$ to the data distribution $p_1(\mathbf{x}) = p_{\text{data}}(\mathbf{x})$. 

From this perspective, inference constitutes a deterministic mapping $\Psi_\theta(\cdot, c): \mathcal{Z} \to \mathcal{X}$ from the Gaussian noise space $\mathcal{Z} = \mathbb{R}^d$ to the data space $\mathcal{X}$. Consequently, the generated sample $\mathbf{x}_1 = \Psi_\theta(\mathbf{x}_0, c)$ is uniquely determined by the initial noise $\mathbf{x}_0$, rendering it the primary controllable degree of freedom when the condition $c$ is fixed. However, the standard Gaussian prior $p_0(\mathbf{x})$ is inherently task-agnostic: its high-probability regions often fail to align with the noise regions yielding high-reward samples. This structural misalignment between the prior and target utility necessitates inference-time initial noise optimization.

\textbf{Inference-Time Scaling as Black-Box Optimization.} We formulate inference-time alignment as a black-box optimization problem with respect to the initial noise. Given a reward function $\mathcal{R}$, the objective is to locate the optimal initial noise $\mathbf{x}_0^*$ within the Gaussian noise space $\mathcal{Z}$:
\begin{equation}
    \mathbf{x}_0^* = \operatorname*{arg\,max}_{\mathbf{x}_0 \in \mathcal{Z}} \mathcal{R}(\Psi_\theta(\mathbf{x}_0, c)).
\end{equation}
From this perspective, inference transitions from a one-shot sampling event to an iterative optimization process centered on the initial state. 

Following the inference-time scaling law \cite{snell2024scaling, brown2024large}, we model generation quality $\mathcal{Q}$ as a function of the inference computational budget $N$, defined as the \textit{Number of Reward Evaluations (NRE)}. Since each candidate necessitates a single reward query, NRE is equivalent to the total number of candidates. Unlike the traditional Number of Function Evaluations (NFE), which conflates search breadth with the computational budget for reward evaluation, NRE isolates the search breadth. To ensure a rigorous assessment, we evaluate reward scores on images generated via the complete denoising trajectory. This guarantees precise reward evaluation for every candidate, circumventing approximation biases arising from reward estimation on intermediate states, thereby accurately assessing the effectiveness of the scaling strategy.

\begin{figure*}[t]
	\centering
	\includegraphics[width=\linewidth]{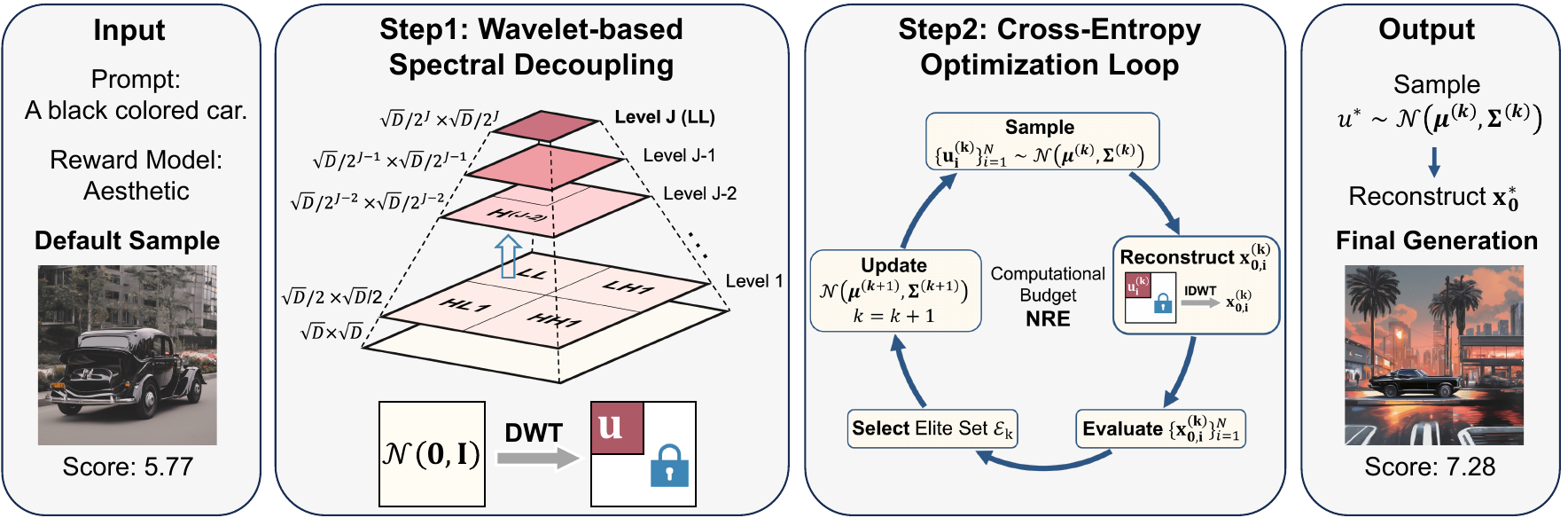} 
	\caption{Overview of Spectral Evolution Search (SES). SES achieves inference-time scaling by optimizing the low-frequency components of the initial noise. First, SES performs wavelet-based spectral decoupling, freezing high-frequency components and constructing a low-frequency search space $\mathbf{u}$, thereby reducing the search dimension from $D$ to $D/4^J$. Subsequently, it executes a cross-entropy optimization loop, iteratively optimizing the distribution parameters ($\boldsymbol{\mu}, \boldsymbol{\Sigma}$) through a ``Sample-Evaluate-Update'' cycle within a limited budget (NRE). Finally, the optimal noise $\mathbf{x}^*$ is sampled and reconstructed to generate high-quality images with superior alignment.}
    \label{fig:overview}
\end{figure*}

\section{SES: Spectral Evolution Search}
\label{sec:ses}

To mitigate the curse of dimensionality in high-dimensional latent space search, inference-time search naturally benefits from focusing on a compact set of degrees of freedom with effective control. As shown in Figure \ref{fig:visualizing_spectral}, low-frequency perturbations significantly reshape the image structure, whereas high-frequency variations have negligible visual impact. Based on this finding, we explicitly constrain the search space to the low-frequency subspace.

In this section, we propose \textit{Spectral Evolution Search (SES)}, an inference-time scaling framework designed for initial noise optimization. As shown in Figure \ref{fig:overview}, SES consists of two modules: (1) wavelet-based spectral decoupling, which limits the search space to the low-frequency subspace; and (2) cross-entropy optimization on the low-frequency subspace, which efficiently locates high-reward regions under a limited computational budget.

\subsection{Wavelet-based Spectral Decoupling}

\textbf{Low-Frequency Search Space Construction.}
Our goal is to construct a low-dimensional search space that retains the modes with the highest control authority. Unlike Fourier transforms, which lack spatial locality, we employ the \textit{Discrete Wavelet Transform (DWT)} to separate frequency bands while preserving spatial structure.

Formally, consider an initial noise $\mathbf{x}_{\text{init}} \in \mathbb{R}^{C \times H \times W}$ sampled from $\mathcal{N}(\mathbf{0}, \mathbf{I})$. We apply a $J$-level orthogonal DWT $\mathcal{W}$ to decompose it into spectral components:
\begin{equation}
\mathbf{c} = \mathcal{W}(\mathbf{x}_{\text{init}}) = \left\{ \mathbf{c}_{LL}^{(J)}, \mathbf{c}_{H}^{\text{fixed}} \right\}.
\end{equation}
Here, $\mathbf{c}_{LL}^{(J)}$ denotes the coarse low-frequency coefficients at level $J$, which are responsible for encoding global structure, while $\mathbf{c}_{H}^{\text{fixed}} = \{ \mathcal{H}^{(1)}, \dots, \mathcal{H}^{(J)} \}$ aggregates the high-frequency details across all scales.

To exploit the dominance of low frequencies in generation, we freeze the high-frequency component $\mathbf{c}_{H}^{\text{fixed}}$ as a static background and optimize exclusively over the low-frequency vector $\mathbf{u} \triangleq \mathbf{c}_{LL}^{(J)} \in \mathbb{R}^{D'}$, where $D' = C \cdot (H/2^J) \cdot (W/2^J)$.
This formulation reduces the search dimensionality by a factor of $4^J$, substantially mitigating the curse of dimensionality. 
Notably, the orthogonality of the DWT ensures $\mathbf{u}$ follows a standard Gaussian marginal distribution. By fixing the high-frequency background, the search operates within a valid affine subspace of the prior, ensuring that the initialization aligns with the pre-trained distribution.

\textbf{Reconstruction to Noise Space.}
During the search, any candidate low-frequency vector $\mathbf{u}$ is combined with the frozen $\mathbf{c}_{H}^{\text{fixed}}$ to form a valid initial noise map $\mathbf{x}_0(\mathbf{u})$. The mapping is defined via the Inverse DWT (IDWT) $\mathcal{W}^{-1}$:
\begin{equation}
\mathbf{x}_0(\mathbf{u}) = \mathcal{W}^{-1}\left( \mathbf{u} \oplus \mathbf{c}_{H}^{\text{fixed}} \right),
\label{eq:idwt_recon}
\end{equation}
where $\oplus$ denotes the concatenation of wavelet coefficients. The synthesized $\mathbf{x}_0(\mathbf{u})$ then serves as the initial noise for sampler, enabling the reward evaluation of the candidate $\mathbf{u}$.

\subsection{Cross-Entropy Optimization on Low Frequencies}

\textbf{Objective Formulation.}
Having constructed the low-frequency search space, we transform the inference-time alignment into a black-box optimization problem. Our goal is to find the optimal low-frequency vector $\mathbf{u}^*$ that maximizes the expected reward $\mathcal{R}$:
\begin{equation}
\mathbf{u}^* = \operatorname*{arg\,max}_{\mathbf{u} \in \mathbb{R}^{D'}} \mathbb{E} \left[ \mathcal{R}\left( \Psi_\theta\left( \mathbf{x}_0(\mathbf{u}) \right) \right) \right].
\end{equation}

Direct optimization of this objective presents two challenges: (1) the reward function $\mathcal{R}$ is typically non-differentiable; and (2) even with differentiable rewards, backpropagation through long-horizon ODEs incurs massive memory costs and gradient instability. Therefore, we employ the \textit{Cross-Entropy Method (CEM)} as a gradient-free evolutionary strategy during inference.

CEM operates by maintaining a Gaussian distribution $p(\mathbf{u}; \boldsymbol{\mu}, \boldsymbol{\Sigma})$ with a diagonal covariance matrix on the low-frequency variable $\mathbf{u}$. It iteratively steers the distribution towards high-reward regions through a ``Sample-Evaluate-Update'' loop. The detailed procedure is as follows:

\textbf{(1) Prior-Matched Initialization.}
We initialize the distribution parameters as $\boldsymbol{\mu}^{(0)} = \mathbf{0}$ and $\boldsymbol{\Sigma}^{(0)} = \mathbf{I}$, explicitly parameterizing $\boldsymbol{\Sigma}$ as a diagonal matrix $\operatorname{diag}(\boldsymbol{\sigma}^2)$. This strictly aligns the starting search space with the pre-trained diffusion prior, reducing the risk of out-of-distribution initialization. We initialize the candidate pool as $\mathcal{P}=\emptyset$.

\textbf{(2) Monte Carlo Sampling \& Evaluation.}
In the $k$-th iteration, we draw $N$ candidates $\{ \mathbf{u}_i^{(k)} \}_{i=1}^N$ from the current distribution $\mathcal{N}(\boldsymbol{\mu}^{(k)}, \boldsymbol{\Sigma}^{(k)})$. Each candidate is combined with the static background $\mathbf{c}_{H}^{\text{fixed}}$ to reconstruct the full noise $\mathbf{x}_{0,i}^{(k)}$ (via Eq.~\ref{eq:idwt_recon}), which is then denoised to evaluate its reward score $S_i^{(k)}$. All pairs $(\mathbf{u}_i^{(k)}, S_i^{(k)})$ are added to $\mathcal{P}$.

\textbf{(3) Elite-Driven Distribution Shaping.} Upon completion of the $k$-th sampling round, we sort the candidate pool $\mathcal{P}$ by reward scores, select the Top-$K$ elites to form the set $\mathcal{E}_k$, and discard the rest. Subsequently, we update the search distribution parameters using the statistics of these elites to guide the distribution towards high-reward regions. 
% To mitigate estimation variance from the small sample size, we employ a smoothing factor $\gamma$ for momentum updates:
To mitigate estimation variance from the small sample size and prevent premature convergence, we employ a smoothing factor $\gamma$ for momentum updates:
\begin{equation}
\begin{aligned}
\boldsymbol{\mu}^{(k+1)} &= (1-\gamma) \hat{\boldsymbol{\mu}}_{\mathcal{E}_k} + \gamma \boldsymbol{\mu}^{(k)} \\
\boldsymbol{\sigma}^{2(k+1)} &= (1-\gamma) \hat{\boldsymbol{\sigma}}^2_{\mathcal{E}_k} + \gamma \boldsymbol{\sigma}^{2(k)}.
\end{aligned}
\end{equation}

\textbf{(4) Iterative Optimization and Final Generation.} Steps 2 and 3 are repeated until $k \times N$ reaches the preset computational budget. Subsequently, the final low-frequency variable $\mathbf{u}^*$ is sampled from the optimized distribution $\mathcal{N}(\boldsymbol{\mu}^{(k)}, \boldsymbol{\Sigma}^{(k)})$ to generate the final image.

\textbf{Discussion: Implicit Regularization vs. Reward Hacking.} 
By strictly confining optimization to the low-frequency manifold, SES imposes implicit regularization that effectively precludes the high-frequency perturbations responsible for \textit{Out-of-Distribution (OOD) Reward Hacking}. In stark contrast, gradient-based guidance methods (e.g., DNO \cite{tang2024inference}) often maximize scores by exploiting these imperceptible high-frequency artifacts, leading to visual artifacts and texture collapse. By blocking access to such adversarial subspaces, our geometric constraint ensures that SES achieves high alignment scores while preserving the semantic integrity and naturalness of the generated images (detailed analysis in Appendix \ref{app:reward_hacking}).

\section{Theoretical Analysis: Why Low Frequencies Dominate?}
\label{sec:theoretical_analysis}

Why is the generation process significantly more sensitive to low-frequency perturbations? In this section, we answer this question by analyzing the perturbation propagation dynamics within continuous-time diffusion models. Detailed mathematical derivations are provided in Appendix \ref{app:A}.

Our analysis reveals a fundamental asymmetry: while the initial latent space is isotropic, the generative flow manifests pronounced anisotropy in the frequency domain. Specifically, we establish that the sensitivity of the generation result to initial perturbations follows a \textit{power-law decay} with respect to spatial frequency. This theoretical finding corroborates the empirical evidence in Figure \ref{fig:visualizing_spectral}, confirming that low-frequency components of the initial noise are the primary drivers of the generative outcome.

\subsection{Perturbation Dynamics of Generative Flows}

Formally, we formulate the generation process as a deterministic ODE evolution over the time interval $t \in [0, 1]$: $\mathrm{d}\mathbf{x}_t = v_\theta(\mathbf{x}_t, t)\mathrm{d}t$. To quantify how an infinitesimal perturbation $\boldsymbol{\xi}_0$ of the initial noise $\mathbf{x}_0$ propagates to the final generated sample $\mathbf{x}_1$, we analyze the first-order variational equation of the perturbation:
\begin{equation}
\frac{\mathrm{d}\boldsymbol{\xi}_t}{\mathrm{d}t} = \mathbf{J}_v(\mathbf{x}_t,t)\boldsymbol{\xi}_t,
\label{eq:variational}
\end{equation}
where $\mathbf{J}_v$ denotes the Jacobian matrix of the velocity field with respect to the state. It implies that the evolution of perturbations is strictly governed by $\mathbf{J}_v$.

For mainstream continuous-time generative models with the interpolation path $\mathbf{x}_t = \alpha(t)\mathbf{x}_1 + \sigma(t)\mathbf{x}_0$, the Jacobian matrix $\mathbf{J}_v$ decomposes into two competing forces (derivation in Appendix \ref{sec:A2}):
\begin{equation}
\mathbf{J}_v(\mathbf{x}_t, t) = \underbrace{\mu(t) \mathbf{J}_{\hat{x}}}_{\text{Signal Amplification}} + \underbrace{\nu(t) \mathbf{I}}_{\text{Noise Contraction}}.
\label{eq:decomposition}
\end{equation}
Here, $\mathbf{J}_{\hat{x}}=\nabla_{\mathbf{x}}\hat{\mathbf{x}}_\theta$ is the Jacobian of the denoiser. The scalar coefficients are defined as $\nu(t)=\frac{\dot{\sigma}}{\sigma} < 0$ and $\mu(t)=\dot{\alpha}-\frac{\dot{\sigma}\alpha}{\sigma} > 0$. 
This decomposition highlights the fundamental conflict in the generative flow:
The term $\nu(t)$ induces \textit{isotropic contraction}, uniformly compressing perturbations across all dimensions to suppress noise.
Conversely, $\mu(t)$ drives \textit{anisotropic amplification} via the denoiser Jacobian $\mathbf{J}_{\hat{x}}$, selectively enhancing perturbations aligned with the tangent space of the data manifold. This interplay dictates the anisotropic sensitivity of the generative flow to initial perturbations.

\subsection{Spectral Scaling Prediction in Frequency Domain}

Directly analyzing this anisotropy in the high-dimensional pixel space is intractable. To render the perturbation analysis analytically tractable, we transition to the frequency domain, employing a theoretical model based on local linearization and approximate spectral decoupling.

\textbf{Decoupled Spectral Dynamics.}
Let $\tilde{\xi}_t(\omega) \triangleq \mathcal{F}\{\boldsymbol{\xi}_t\}(\omega)$ denote the scalar Fourier component of the perturbation at frequency $\omega$. By substituting Eq.~\ref{eq:decomposition} into Eq.~\ref{eq:variational} and approximating the denoiser's Jacobian as diagonally dominant, we decouple the high-dimensional perturbation dynamics into a set of independent scalar equations:
\begin{equation}
\frac{\mathrm{d}\tilde{\xi}_t(\omega)}{\mathrm{d}t} = [\mu(t) h(\omega,t) + \nu(t)] \tilde{\xi}_t(\omega),
\end{equation}
where $h(\omega,t)$ denotes the effective spectral response of the denoising network. By integrating this differential equation throughout the entire generation process, we obtain the cumulative gain $G(\omega)$, which quantifies the total amplification of an initial perturbation at frequency $\omega$:
\begin{equation} 
G(\omega) = \exp\left( \int_0^1 \left[ \mu(\tau) h(\omega, \tau) + \nu(\tau) \right] \mathrm{d}\tau \right). 
\end{equation}
Since the geometric contraction term $\nu(\tau)$ acts uniformly across the spectrum, the frequency-dependent variation of the cumulative gain is governed exclusively by the spectral response function $h(\omega,t)$.

\textbf{The Denoiser as a Wiener Filter.}
To characterize the spectral response $h(\omega,t)$, we approximate the trained network locally as an ideal Minimum Mean Square Error (MMSE) estimator. 
We leverage the statistical property that natural images exhibit a power-law spectral decay $\|\omega\|^{-\gamma}$ (typically $\gamma \approx 2$) \cite{field1987relations, torralba2003statistics}. 
Since the states $\mathbf{x}_1$ correspond to encoded data representations that reside in a spatially correlated latent manifold, we model their spectrum as inheriting the heavy-tailed characteristic of natural images ($P_{\mathbf{x}_1}(\omega) \propto \|\omega\|^{-\beta}$), in stark contrast to the flat spectrum of the Gaussian noise.
This induces a frequency-dependent Signal-to-Noise Ratio (SNR):
\begin{equation}
\text{SNR}(\omega, t) \propto \frac{\alpha^2(t)}{\sigma^2(t)} \cdot \|\omega\|^{-\beta}. 
\end{equation}
Under the MMSE objective, the optimal denoiser behaves as a frequency-domain Wiener filter (derivation in Appendix \ref{sec:A4}). It preserves high-SNR (low-frequency) components while suppressing low-SNR (high-frequency) noise:
\begin{equation}
h(\omega, t) = \frac{1}{\alpha(t)} \cdot \frac{\text{SNR}(\omega, t)}{\text{SNR}(\omega, t) + 1}.
\end{equation}

\begin{figure}[t]
	\centering
	\includegraphics[width=\linewidth]{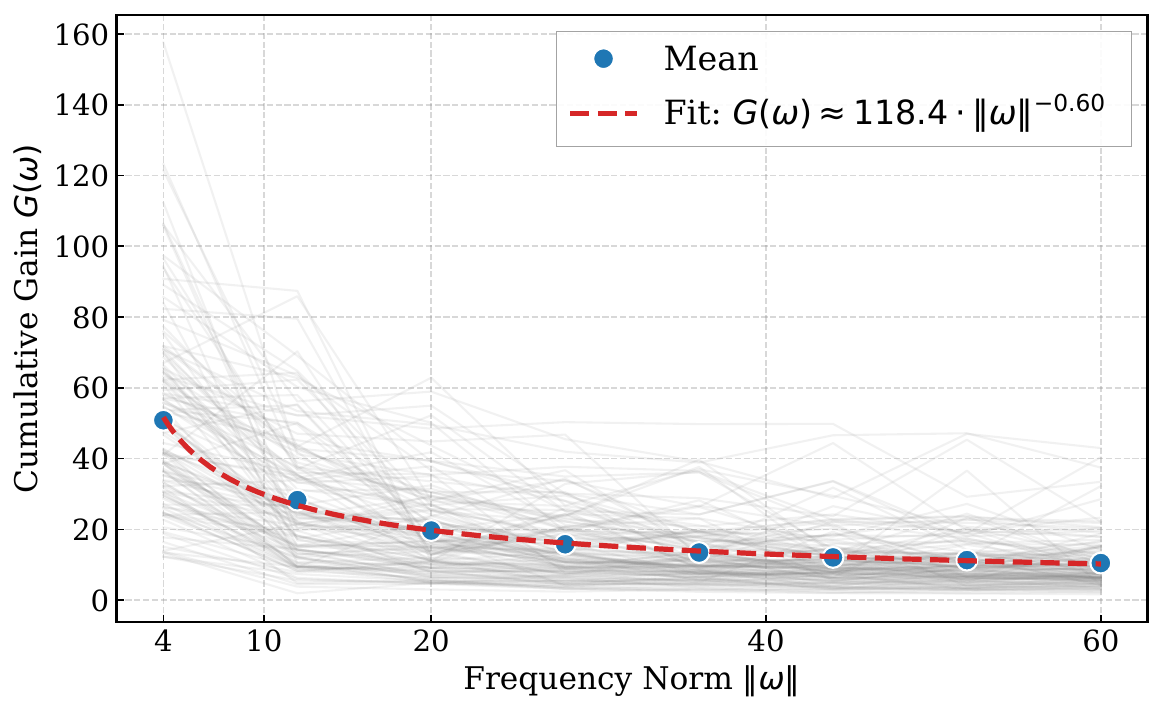} 
    \caption{Validating the Spectral Scaling Prediction. We partition the frequency domain into 8 radial sub-bands and measure the cumulative gain of unit-norm perturbations ($\|\boldsymbol{\xi}\|_2 = 1$) injected into the SDXL initial noise. The results (blue dots, averaged over 100 prompts) reveal a monotonic decay in sensitivity as frequency increases. The strong alignment with the predicted power-law fit (red dashed line) validates the Spectral Scaling Prediction.}
	\label{fig:Cumulative_Gain}
    % \vspace{-2mm}
\end{figure}

Substituting this filter response into the cumulative gain formulation yields our primary theoretical prediction.

\begin{proposition}[Spectral Scaling Prediction, Proof in Appendix~\ref{sec:A5}]
    \label{prop:Spectral_Scaling_Law}
    Under the approximations of spectral decoupling and MMSE optimality, 
    the cumulative gain $G(\omega)$ of an initial perturbation follows an inverse power-law scaling with respect to the frequency norm $\|\omega\|$:
    \begin{equation}
    G(\omega) \propto \|\omega\|^{-\beta/2},
    \end{equation}
    where $\beta$ reflects the spectral decay exponent of the data manifold within the latent space (typically $0 < \beta < 2$).
\end{proposition}

Proposition \ref{prop:Spectral_Scaling_Law} elucidates a critical spectral bias in optimizing the initial noise: the effective degrees of freedom are intrinsically sparse and concentrated in the low-frequency band. As empirically corroborated in Figure \ref{fig:Cumulative_Gain}, high-frequency perturbations exhibit negligible cumulative gain compared to low-frequency modes, rendering them ineffective control variables. This theoretical insight substantiates the SES design (Section \ref{sec:ses}), justifying our strategy of concentrating the limited computational budget exclusively on low-frequency modes to maximize search effectiveness.

\section{Experiments}
\label{sec:exp}

\begin{table*}[t]
    \centering
    \small
    \setlength{\tabcolsep}{3pt} 

    \newcommand{\res}[2]{#1$_{\pm #2}$}

    \newcommand{\best}[2]{\textbf{#1}$_{\pm #2}$}

    \caption{Quantitative comparison of inference-time scaling strategies on SDXL and FLUX.1-dev. We report the final reward scores achieved under a fixed budget of $\text{NRE}=200$. Results are averaged over 5 random seeds and reported as $\text{mean}_{\pm \text{std}}$. "Baseline" denotes standard inference without search. Each column corresponds to a distinct experiment where the indicated metric serves as the sole optimization objective. The best results are highlighted in \textbf{bold}.}

    \label{tab:main_results}
    
    \begin{tabular}{l ccccc c ccccc}
        \toprule
        \multirow{2}{*}{\textbf{Method}} & \multicolumn{5}{c}{\textbf{SDXL}} & & \multicolumn{5}{c}{\textbf{FLUX.1-dev}} \\
        \cmidrule(lr){2-6} \cmidrule(lr){8-12}
         & CLIP$\uparrow$ & Pick$\uparrow$ & HPS$\uparrow$ & ImgRew.$\uparrow$ & Aes.$\uparrow$ && CLIP$\uparrow$ & Pick$\uparrow$ & HPS$\uparrow$ & ImgRew.$\uparrow$ & Aes.$\uparrow$ \\
        \midrule
        Baseline & \res{32.07}{1.75} & \res{21.16}{0.51} & \res{27.34}{0.51} & \res{-0.14}{0.26} & \res{5.22}{0.16} && \res{33.26}{0.86} & \res{22.00}{0.41} & \res{28.08}{0.26} & \res{1.02}{0.13} & \res{6.19}{0.13} \\
        \midrule
        BoN & \res{43.30}{0.39} & \res{23.77}{0.06} & \res{30.86}{0.06} & \res{1.54}{0.03} & \res{6.35}{0.06} && \res{39.50}{0.12} & \res{23.17}{0.09} & \res{30.10}{0.11} & \res{1.67}{0.03} & \res{6.80}{0.07} \\
        ZO-N & \res{41.30}{0.27} & \res{22.98}{0.21} & \res{30.55}{0.50} & \res{1.31}{0.09} & \res{6.21}{0.05} && \res{38.10}{0.18} & \res{22.95}{0.21} & \res{29.62}{0.25} & \res{1.54}{0.05} & \res{6.57}{0.11} \\
        SoP & \res{40.74}{0.18} & \res{22.95}{0.11} & \res{29.85}{0.21} & \res{1.20}{0.09} & \res{5.92}{0.10} && \res{37.27}{0.29} & \res{22.63}{0.24} & \res{29.52}{0.19} & \res{1.35}{0.02} & \res{6.45}{0.14} \\
        SMC & \res{42.53}{0.20} & \res{23.59}{0.18} & \res{30.71}{0.20} & \res{1.40}{0.07} & \res{6.36}{0.17} && - & - & - & - & - \\
        SVDD & \res{41.50}{0.37} & \res{23.30}{0.29} & \res{30.35}{0.37} & \res{1.38}{0.13} & \res{6.23}{0.19} && - & - & - & - & - \\
        Demon & \res{41.81}{0.52} & \res{23.82}{0.14} & \res{31.03}{0.19} & \res{1.48}{0.07} & \res{6.34}{0.08} && - & - & - & - & - \\
        \midrule
        \textbf{SES} & \best{43.53}{0.54} & \best{23.97}{0.06} & \best{31.45}{0.26} & \best{1.62}{0.02} & \best{6.55}{0.06} && \best{40.18}{0.19} & \best{23.35}{0.15} & \best{30.68}{0.17} & \best{1.79}{0.01} & \best{7.03}{0.08} \\
        \bottomrule
    \end{tabular}
\end{table*}

This section provides a comprehensive evaluation of SES's inference-time scaling capabilities across varied generative architectures and alignment objectives. Our analysis investigates three primary research questions: (1) the alignment effectiveness of SES under fixed computational budgets; (2) the key mechanisms driving its performance gains; and (3) its robustness and generalization across training-time aligned models and black-box rewards. Detailed experimental settings and supplementary analyses are provided in Appendices \ref{sec:setting} and \ref{sec:additional_exp}, respectively.

\textbf{Tasks and Rewards.} To validate the universality of SES, we employ four models spanning mainstream generative paradigms: the Latent Diffusion-based Stable Diffusion (SD) v1.5 and SDXL \cite{podell2023sdxl}, and the Flow Matching-based FLUX.1-dev \cite{flux2024} and Qwen-Image \cite{wu2025qwen}. For evaluation data, we use DrawBench \cite{saharia2022photorealistic} for SD series and randomly sample 200 prompts from Pick-a-Pic \cite{kirstain2023pick} for Flow Matching models. We assess alignment across diverse reward objectives, including semantic consistency (CLIP Score \cite{hessel2021clipscore}), human preference (PickScore \cite{kirstain2023pick}, HPS v2 \cite{wu2023human}, ImageReward \cite{xu2023imagereward}), and Aesthetic Score.

\textbf{Baselines.} 
We benchmark SES against representative inference-time strategies, specifically Best-of-N (BoN), Zero-Order Search (ZO-N), and Search over Paths (SoP) \cite{ma2025scaling}. 
We also consider trajectory-based methods including SMC \cite{kim2025test}, SVDD \cite{li2024derivative}, and Demon \cite{yeh2024training}. 
However, these methods rely on intermediate stochastic injections characteristic of SDE solvers. Since FLUX and Qwen-Image utilize deterministic ODE solvers with no injectable stochasticity during sampling, these baselines are incompatible and thus excluded from the Flow Matching comparisons.

We implement full denoising evaluation to standardize comparisons between methods optimizing initial noise and those optimizing intermediate latents. By decoding all candidates to the final image space $\mathbf{x}_1$ before scoring, we circumvent approximation errors inherent to intermediate reward estimation. Experiments default to $\text{NRE}=200$ and $T_{\text{total}}=50$.

\subsection{Inference-Time Scaling Performance}

\begin{figure}[t]
    \centering
    \renewcommand{\arraystretch}{1.3}
    \setlength{\tabcolsep}{1pt}
    \footnotesize
    
    \renewcommand{\tabularxcolumn}[1]{>{\centering\arraybackslash}m{#1}}
    % \newcolumntype{Y}{X}

    \begin{tabularx}{\linewidth}{XXXXXX}
        \toprule
        \textbf{Base} & \textbf{BoN} & \textbf{ZO-N} & \textbf{SMC} & \textbf{Demon} & \textbf{SES} \\
        \midrule
        
        \includegraphics[width=\linewidth]{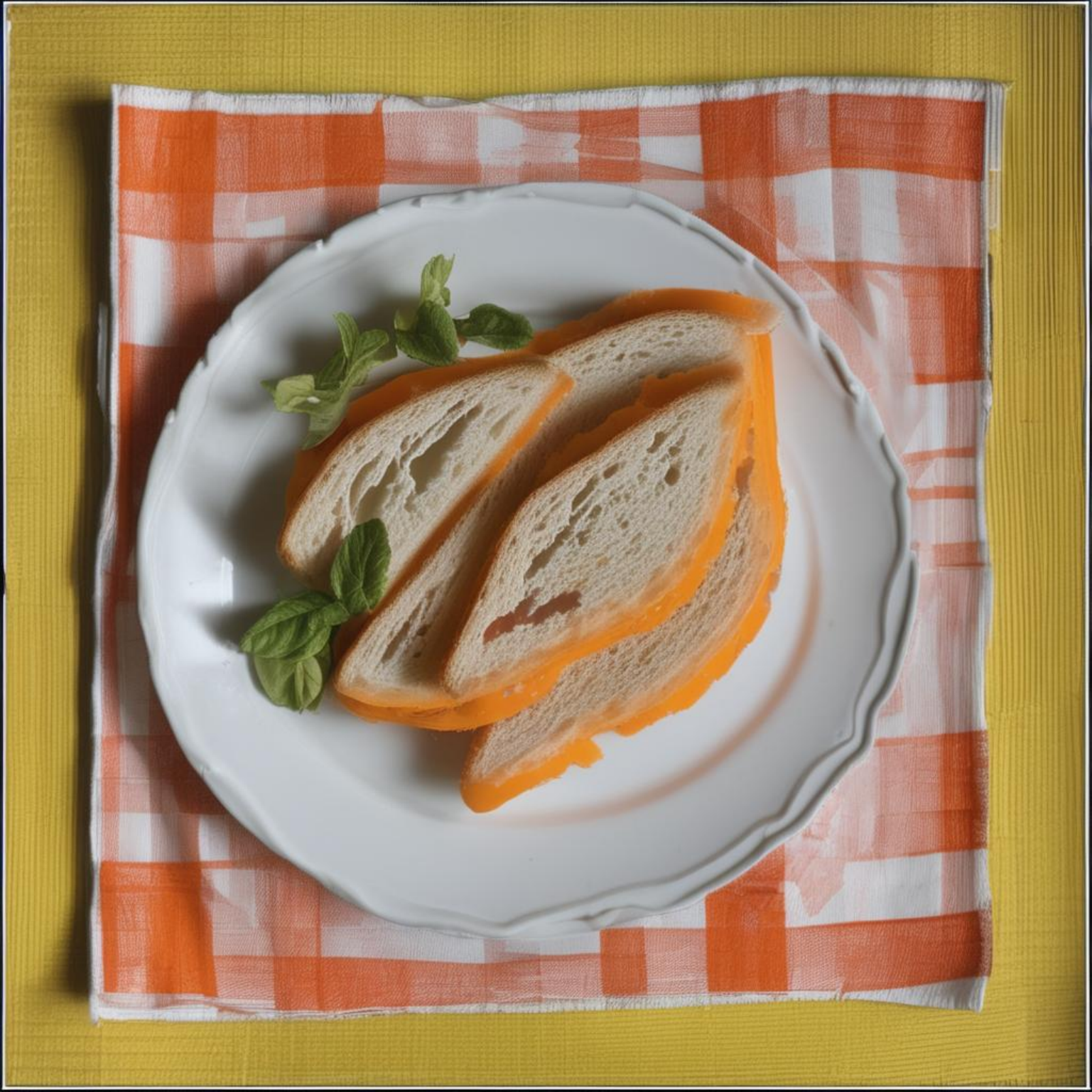} &
        \includegraphics[width=\linewidth]{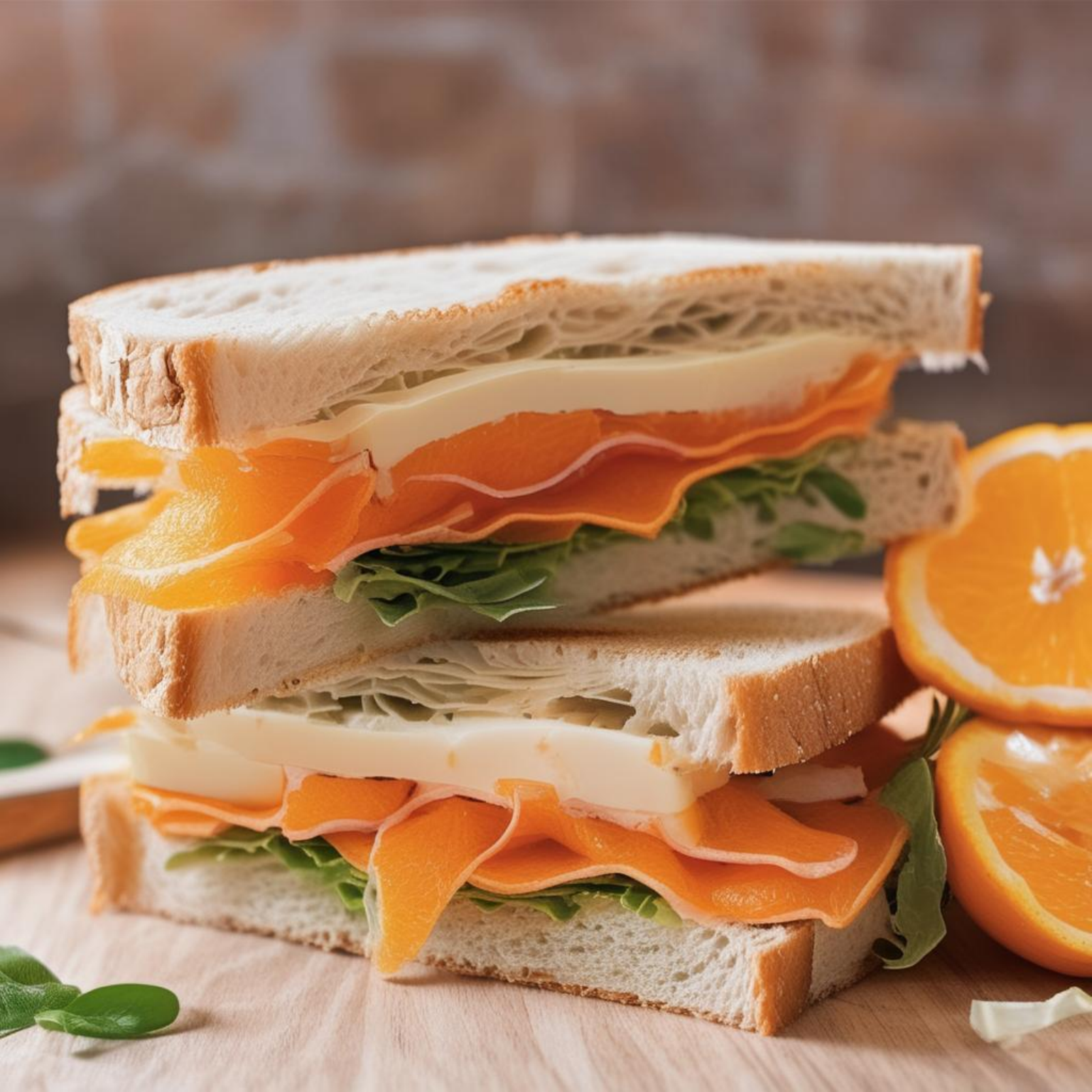} &
        \includegraphics[width=\linewidth]{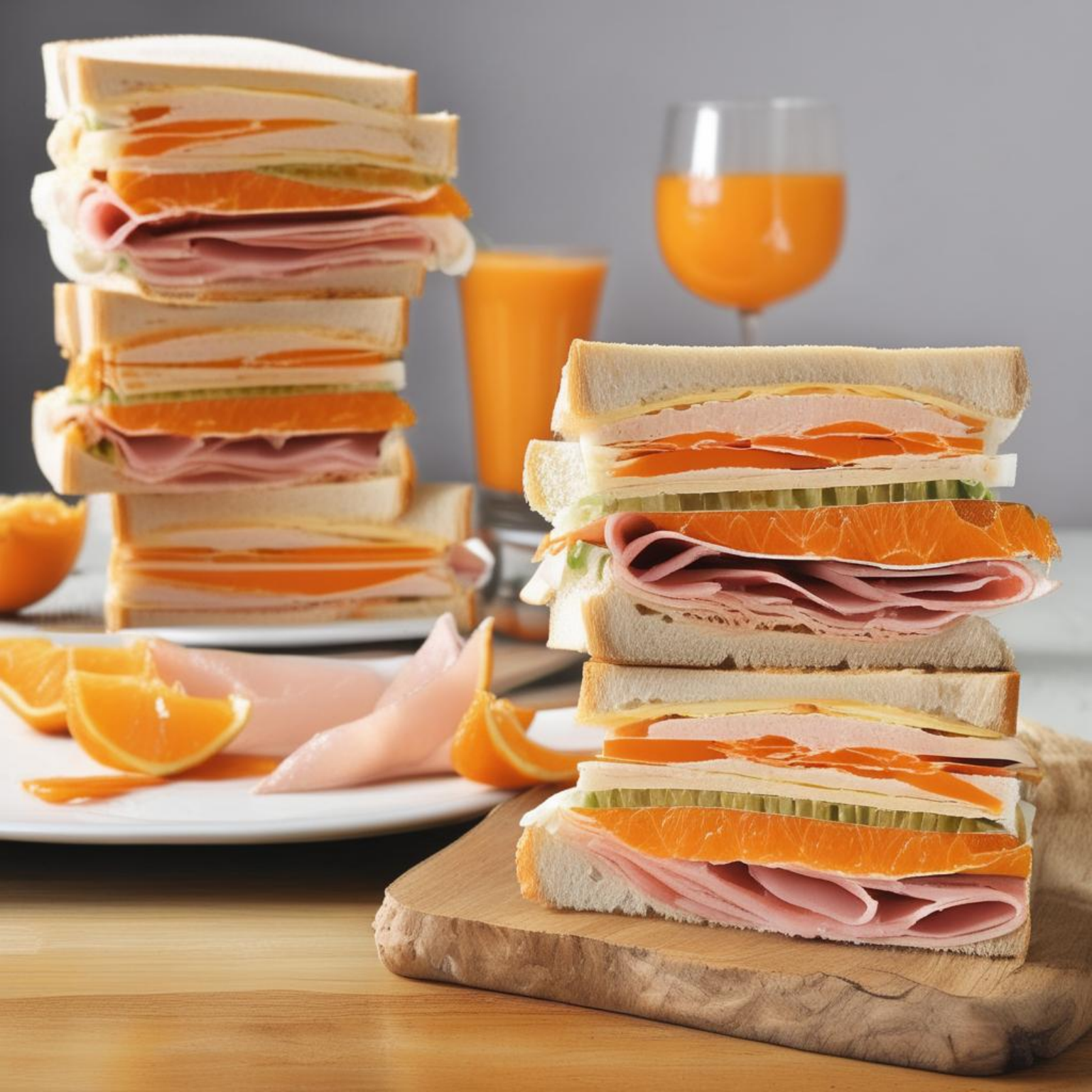} &
        \includegraphics[width=\linewidth]{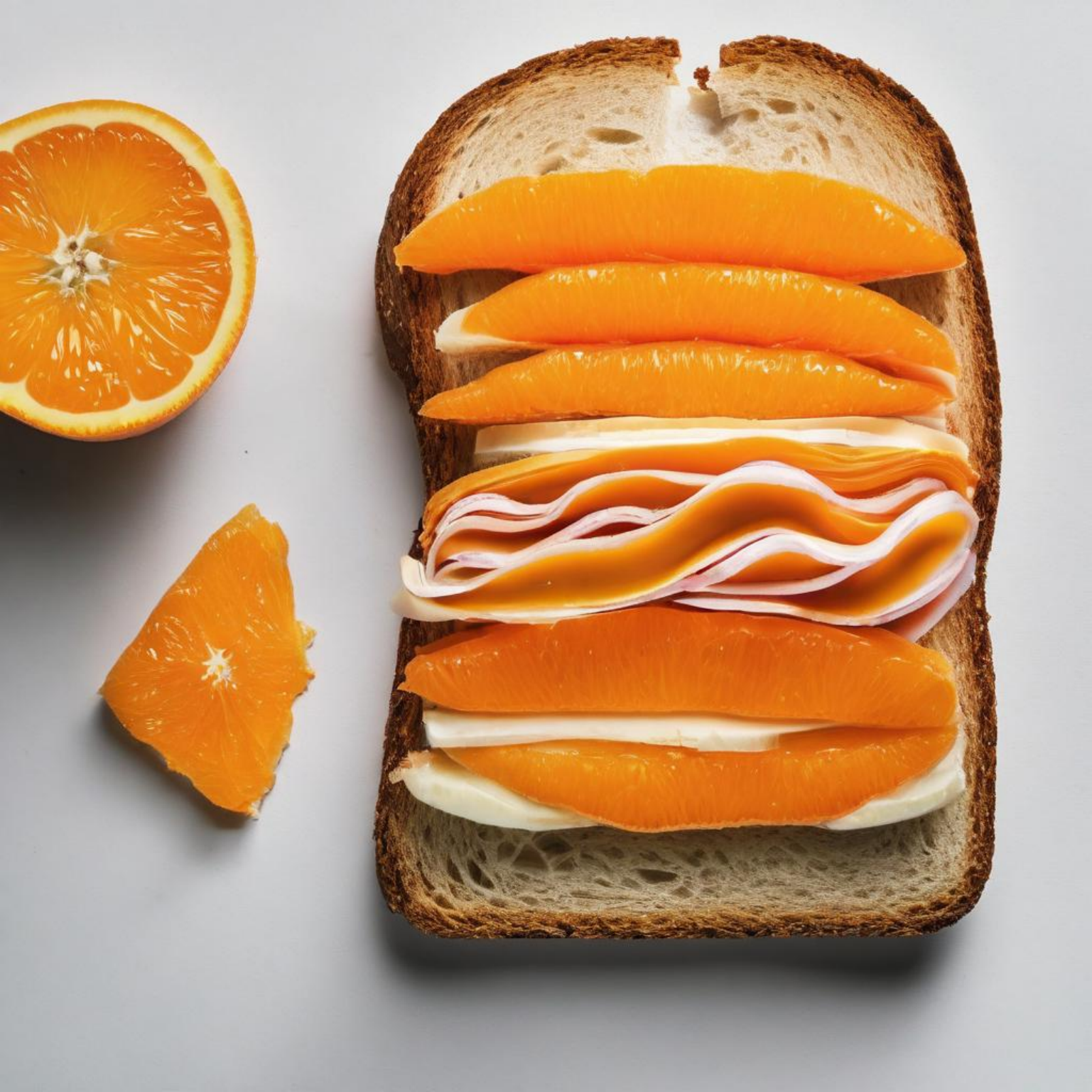} &
        \includegraphics[width=\linewidth]{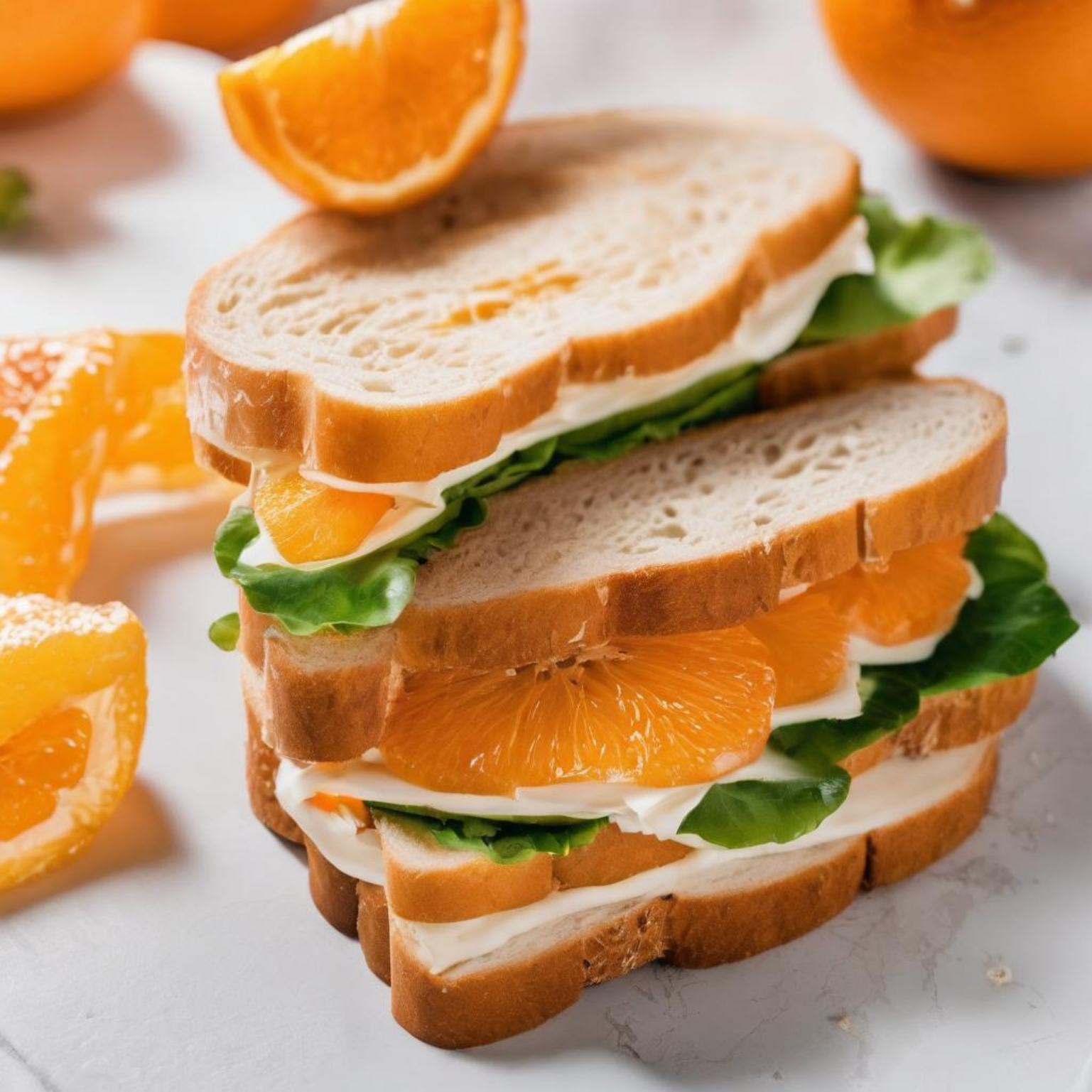} &
        \includegraphics[width=\linewidth]{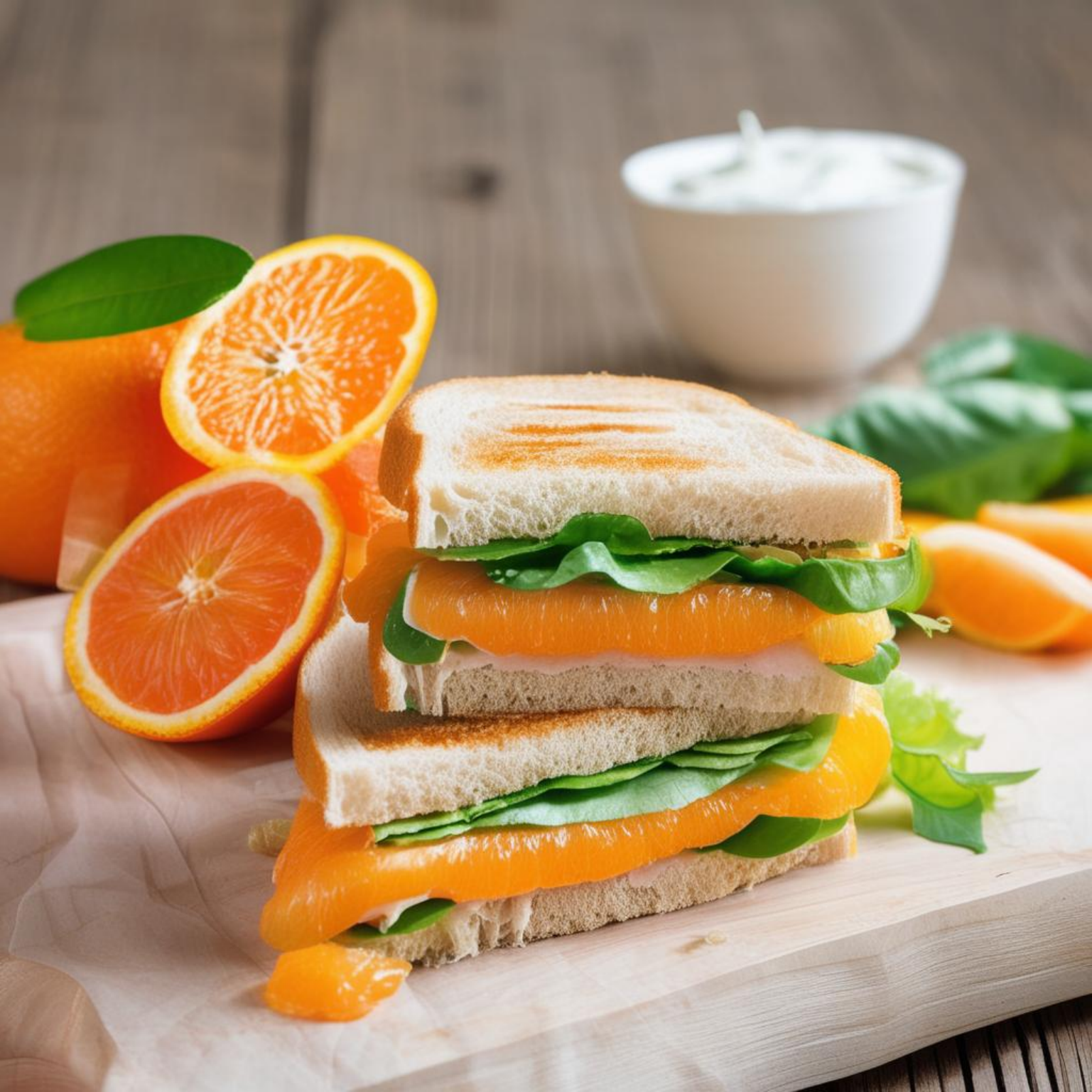} \\
        
        \multicolumn{6}{c}{
            \parbox{0.96\linewidth}{\centering \scriptsize
                \textit{Prompt: An orange colored sandwich.} \\ 
                \textsc{(Model: SDXL $\mid$ Reward: HPS)}
            }
        } \\
        \addlinespace[4pt]

        \includegraphics[width=\linewidth]{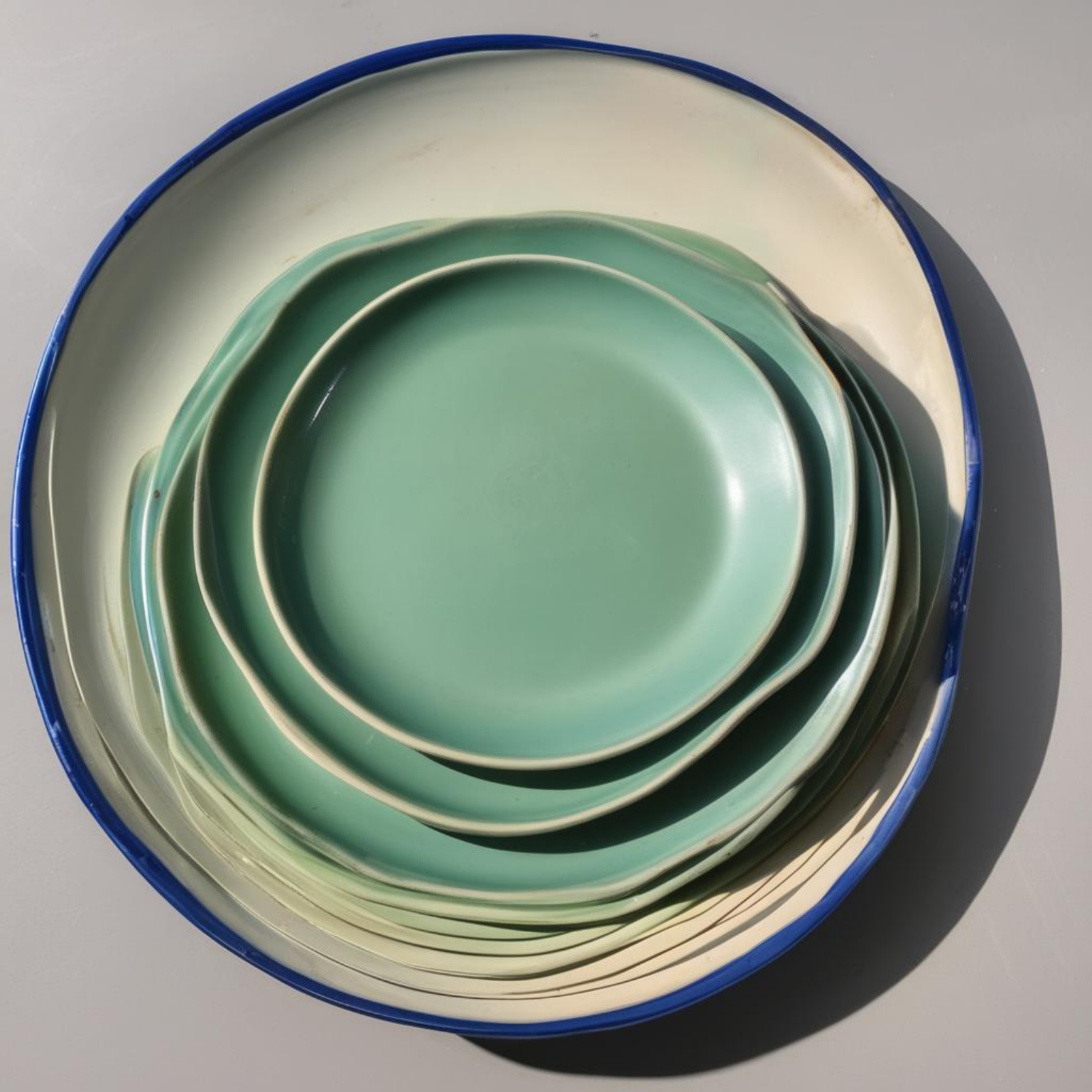} &
        \includegraphics[width=\linewidth]{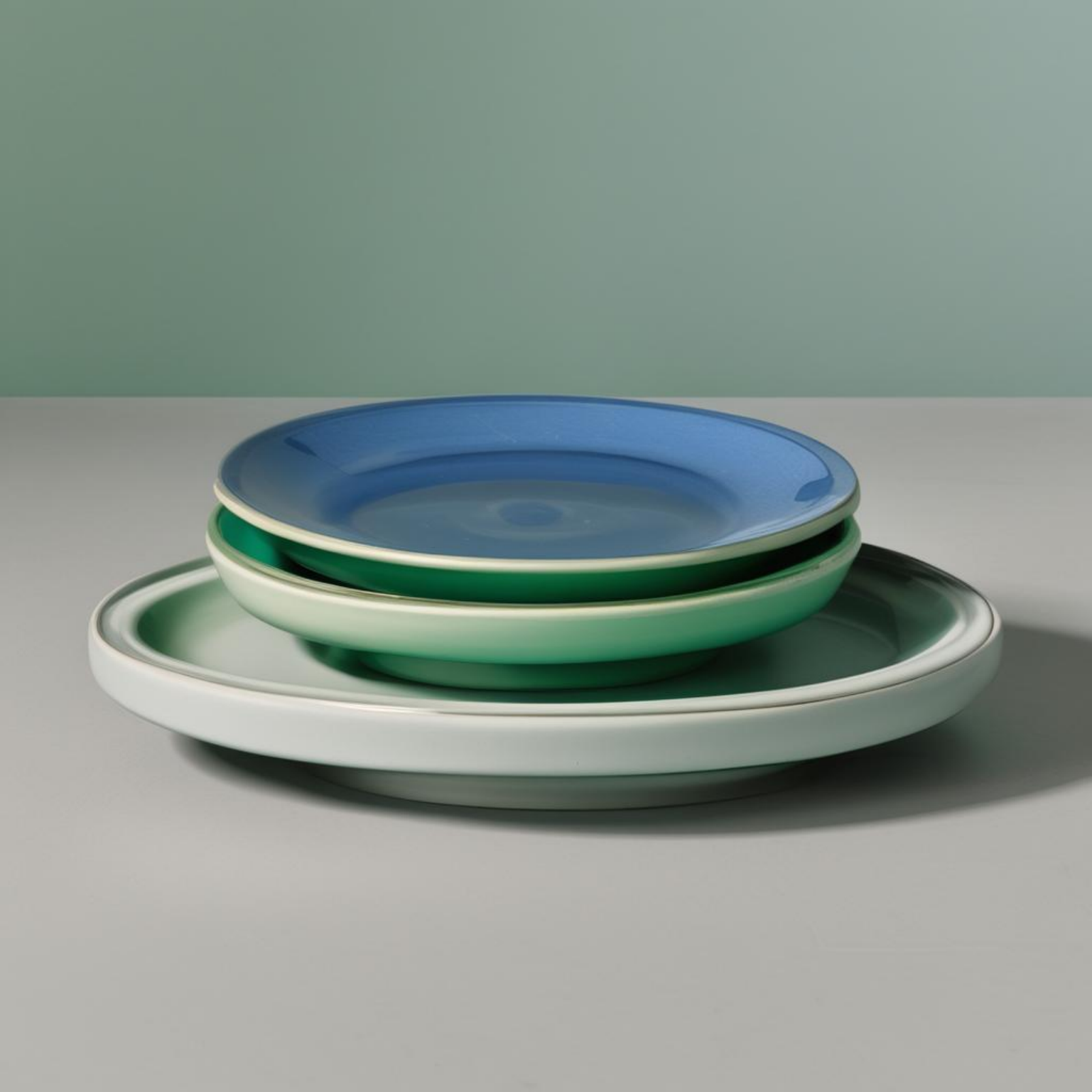} &
        \includegraphics[width=\linewidth]{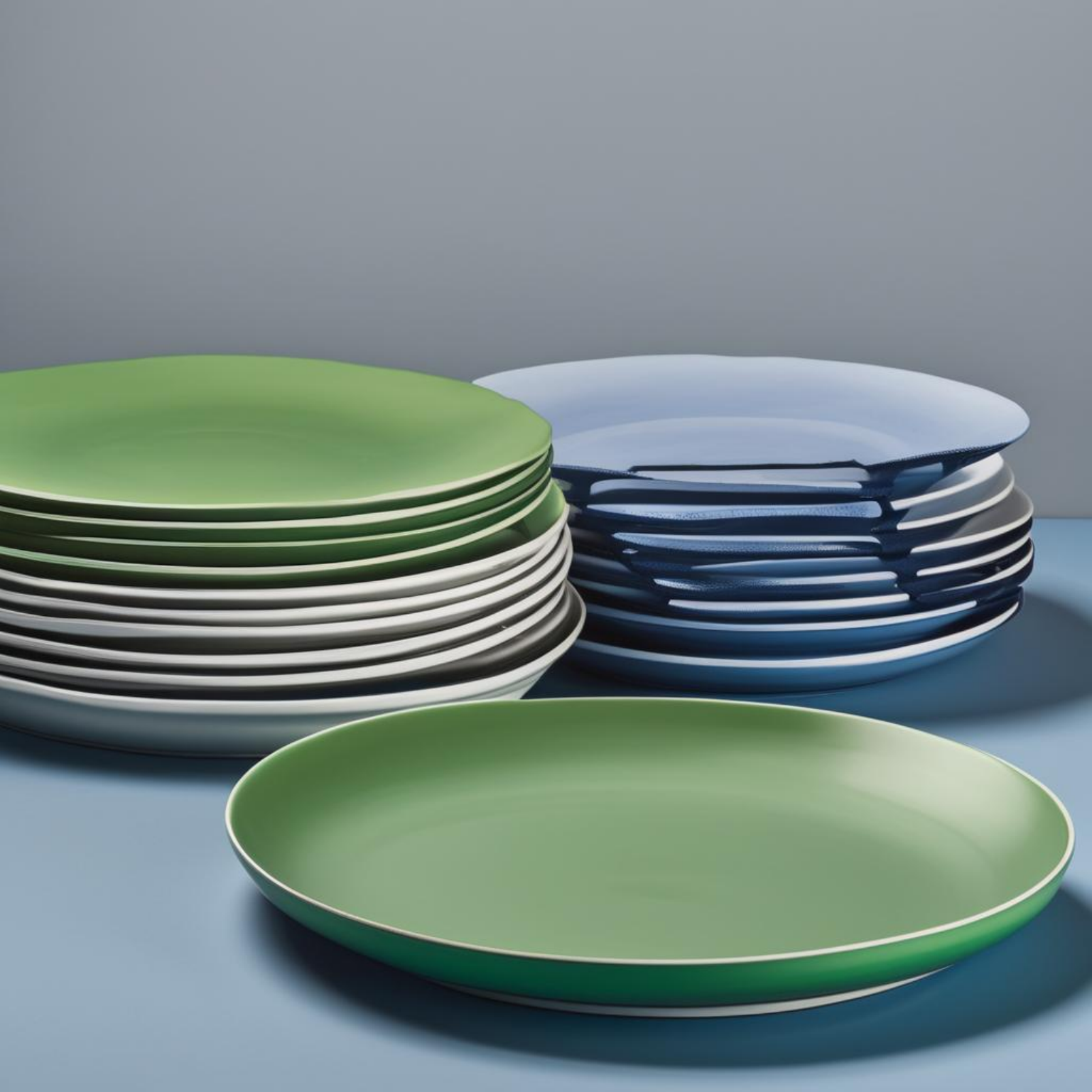} &
        \includegraphics[width=\linewidth]{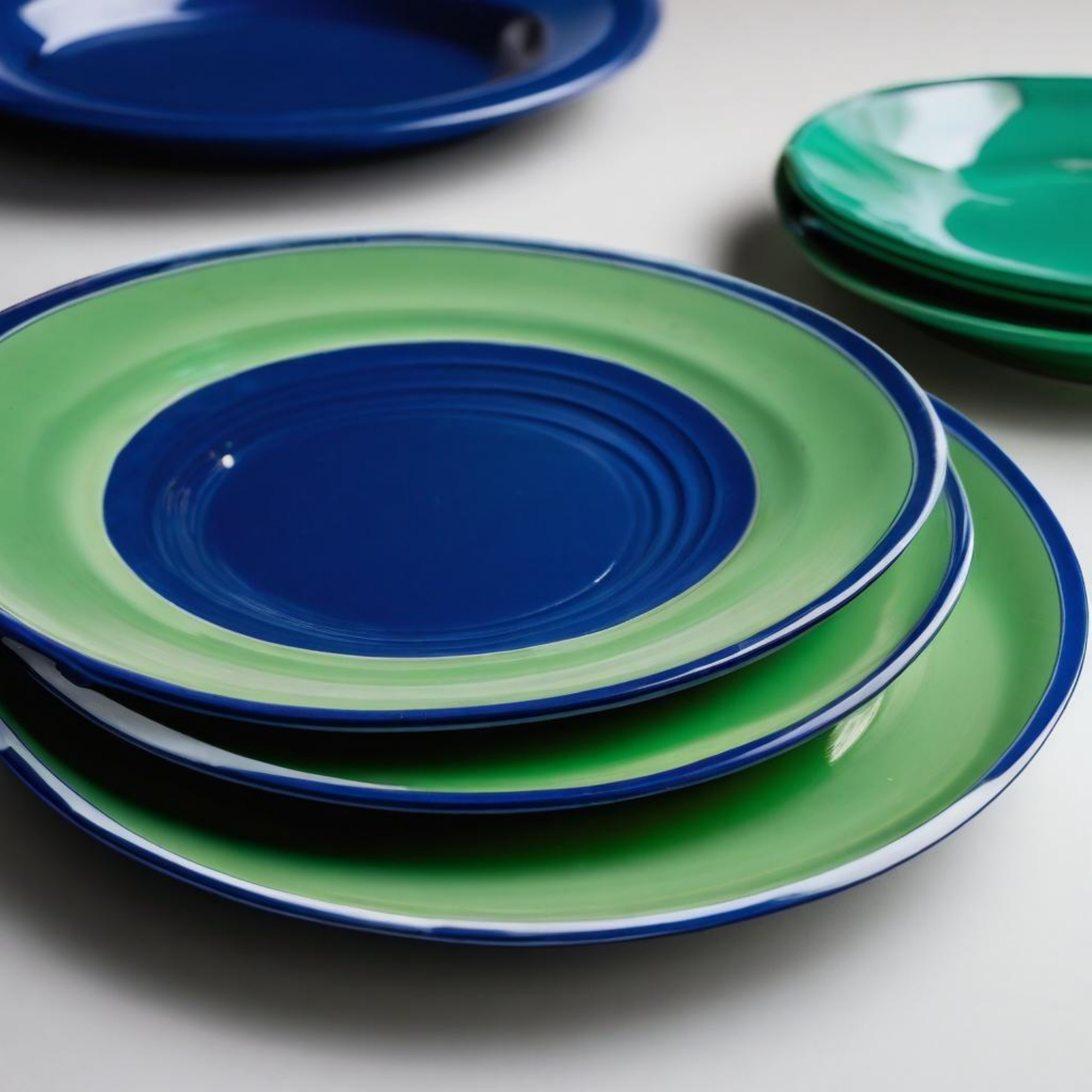} &
        \includegraphics[width=\linewidth]{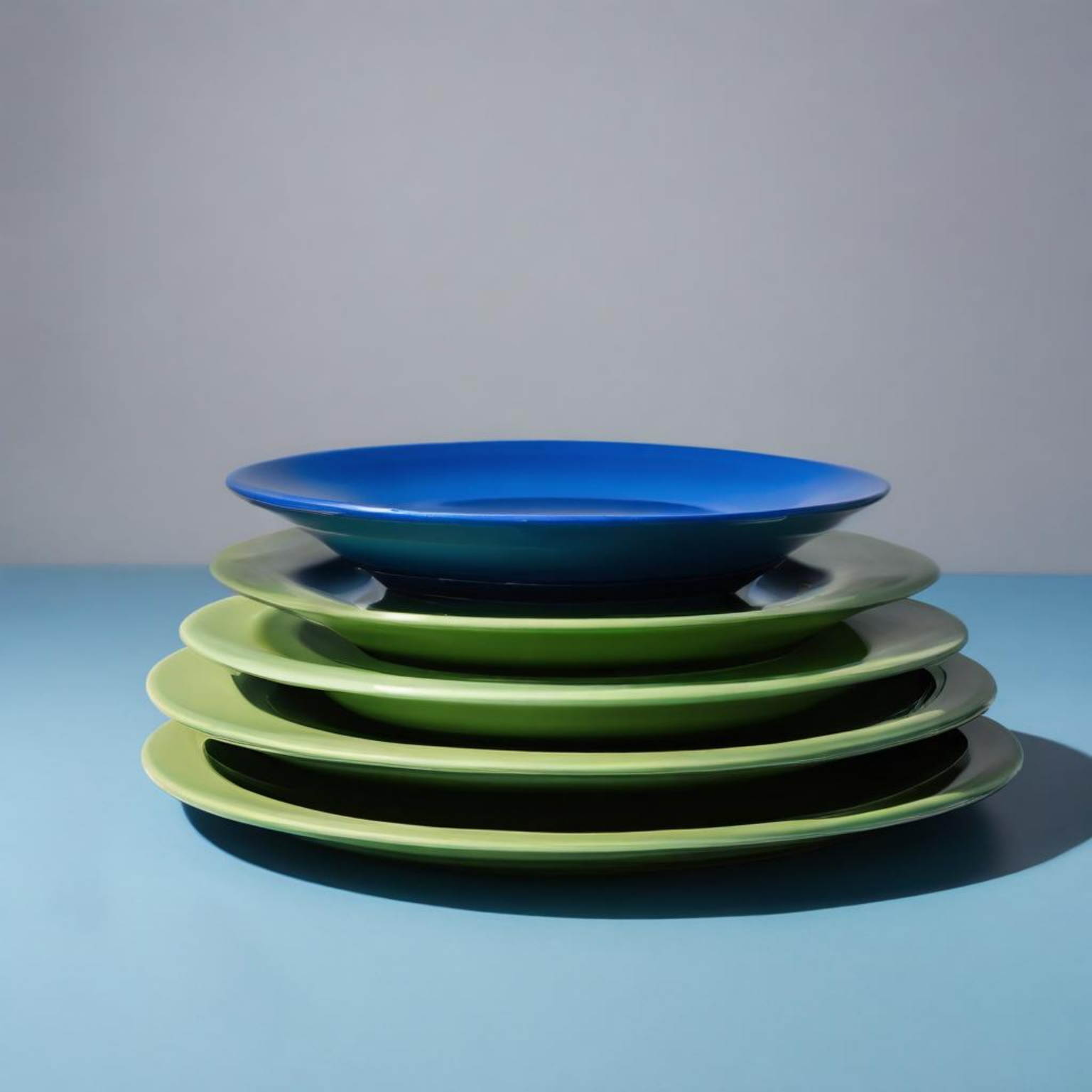} &
        \includegraphics[width=\linewidth]{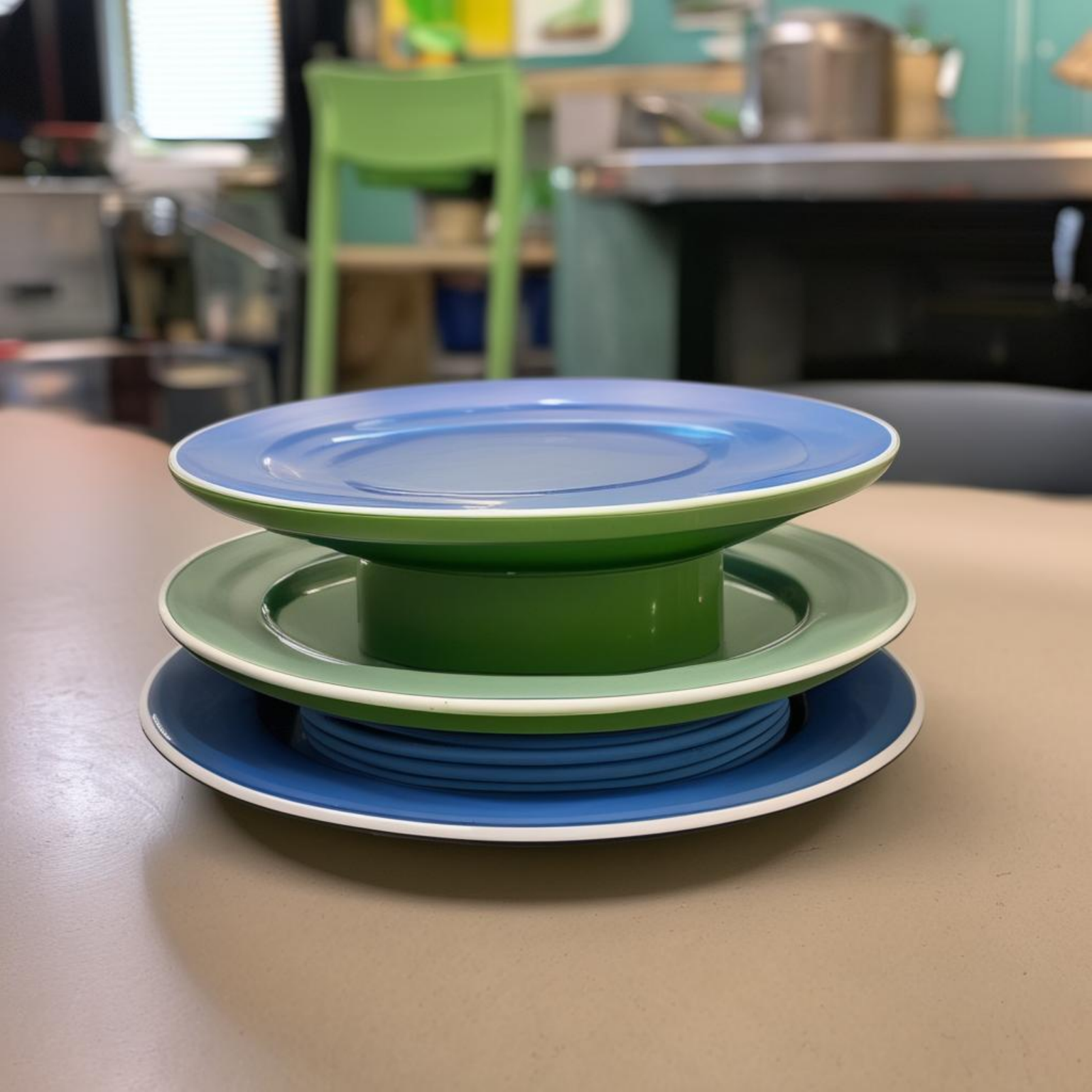} \\
        
        \multicolumn{6}{c}{
            \parbox{0.96\linewidth}{\centering \scriptsize
                \textit{Prompt: A stack of 3 plates. A blue plate is on the top, a green plate is in the middle, a blue plate is on the bottom.} \\
                \textsc{(Model: SDXL $\mid$ Reward: CLIP)}
            }
        } \\
        \addlinespace[4pt]

        \includegraphics[width=\linewidth]{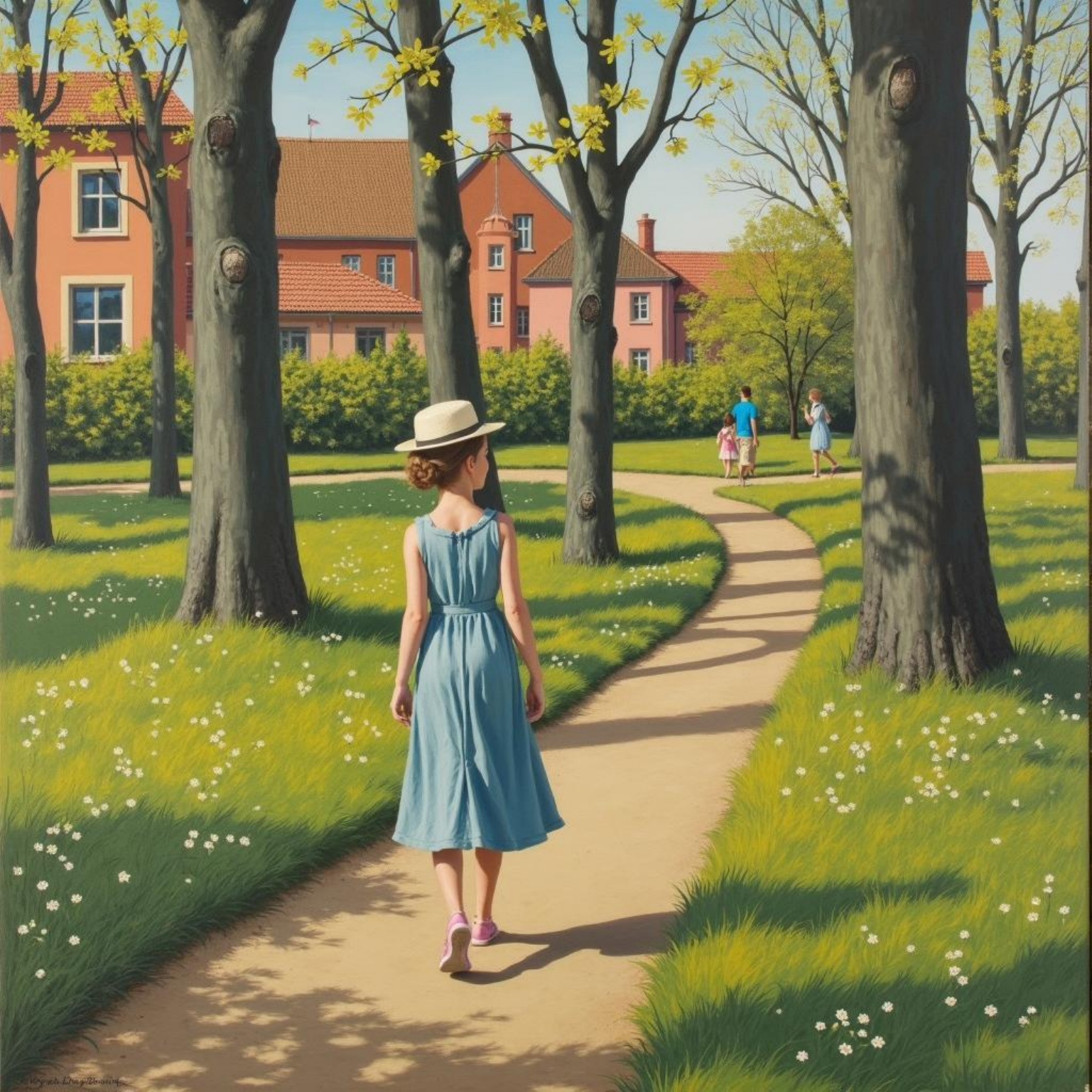} &
        \includegraphics[width=\linewidth]{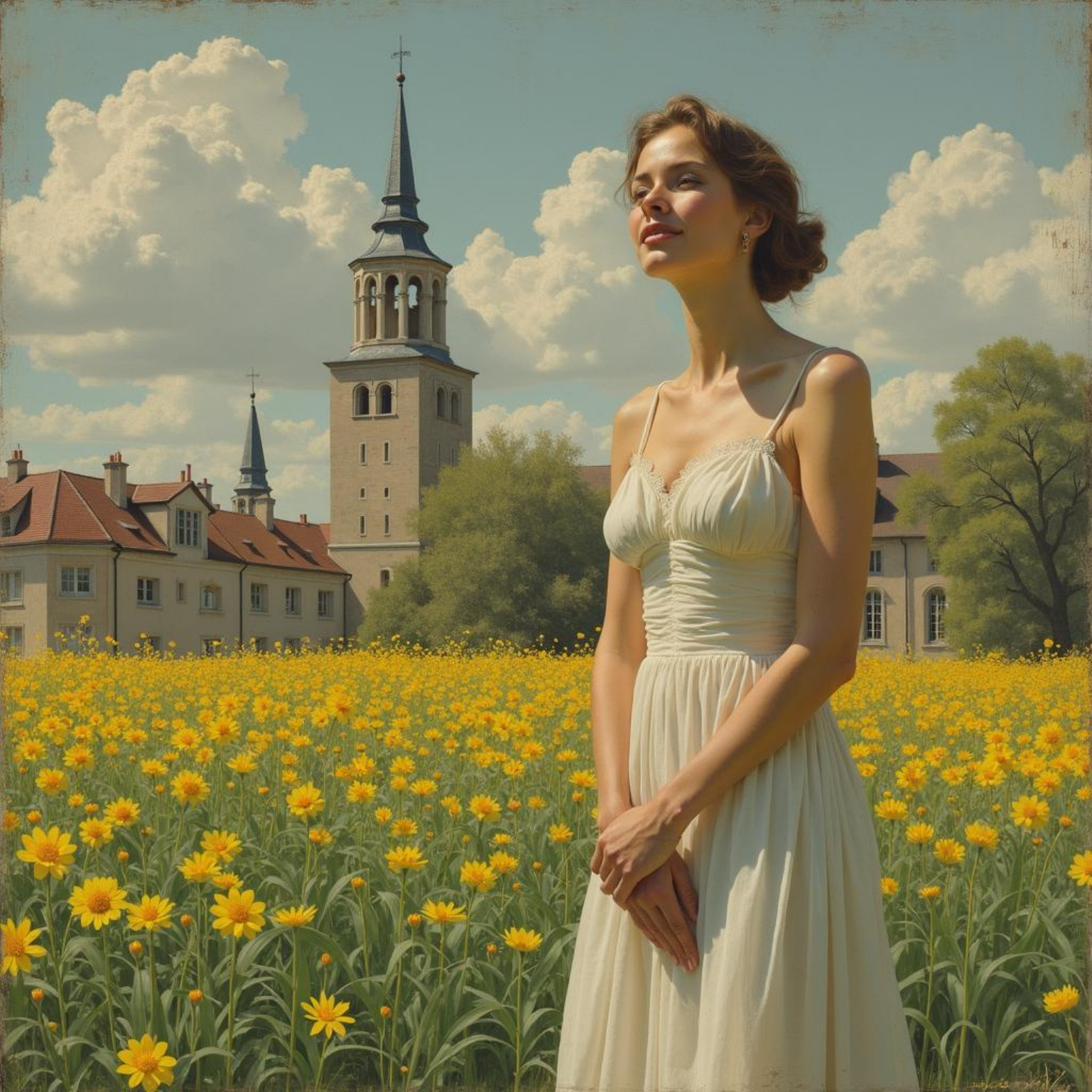} &
        \includegraphics[width=\linewidth]{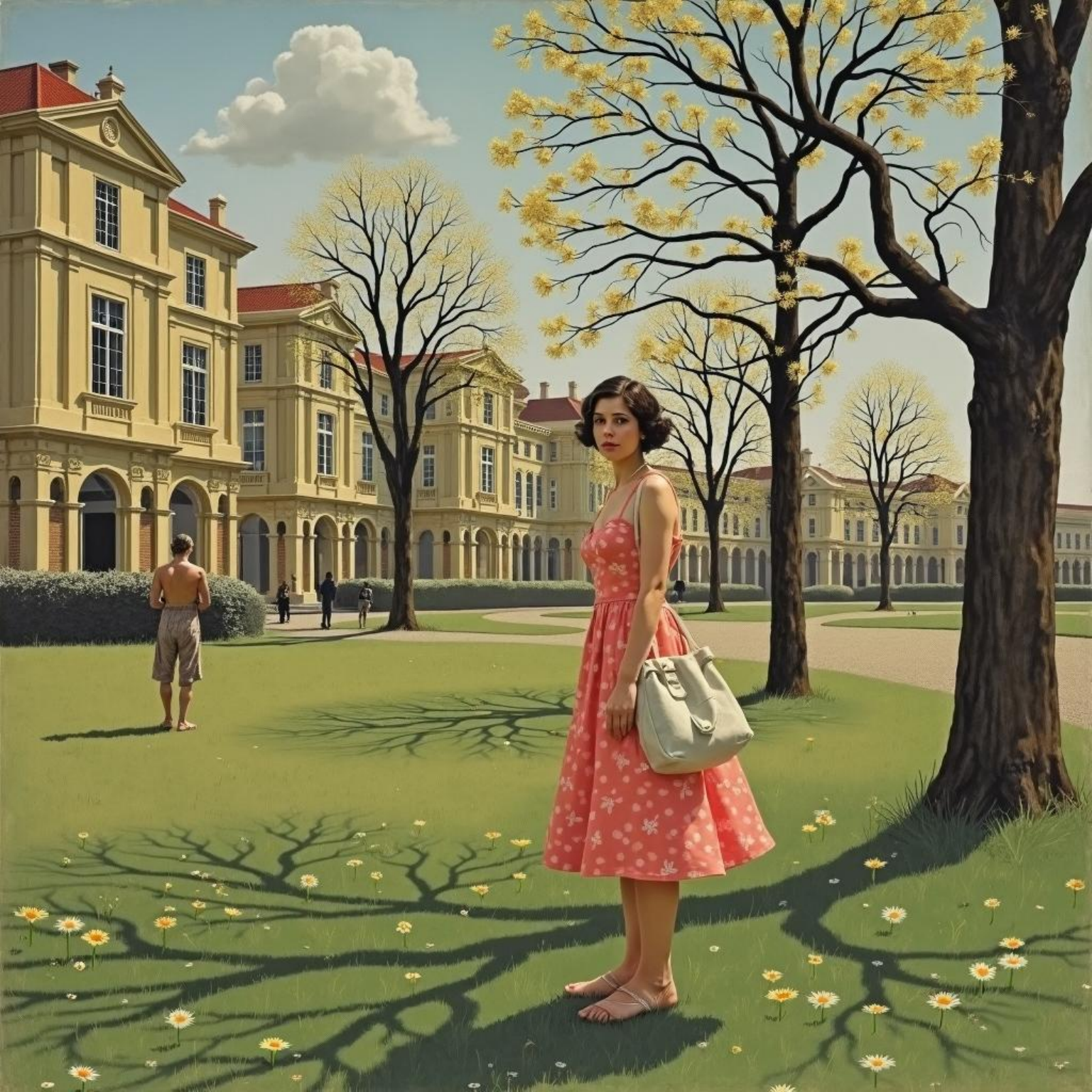} &
        
        \na &
        
        \includegraphics[width=\linewidth]{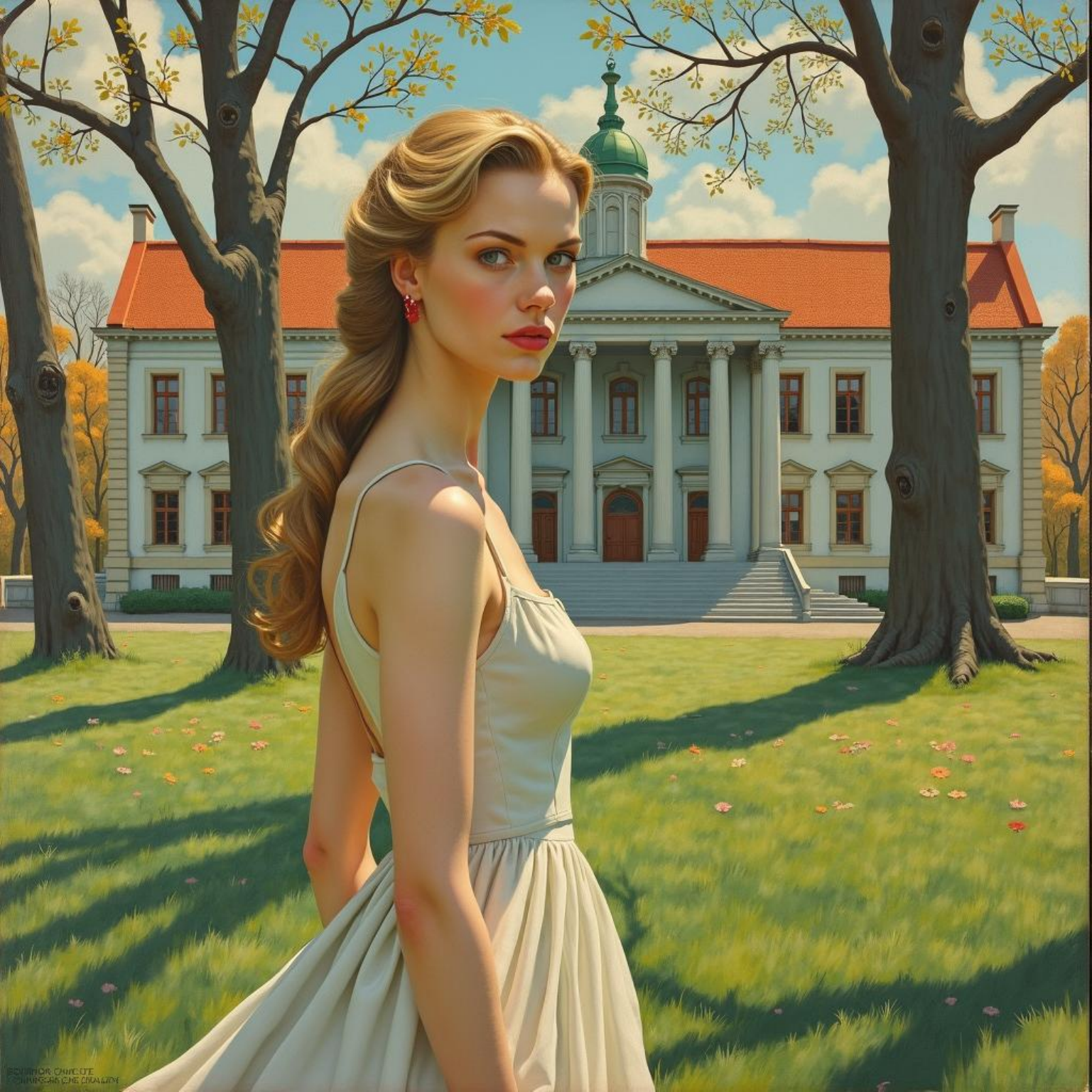} \\
        
        \multicolumn{6}{c}{
            \parbox{0.96\linewidth}{\centering \scriptsize
                \textit{Prompt: Springtime a la Paul Delvaux} \\
                \textsc{(Model: FLUX $\mid$ Reward: Aesthetic)}
            }
        } \\
        \addlinespace[4pt]

        \includegraphics[width=\linewidth]{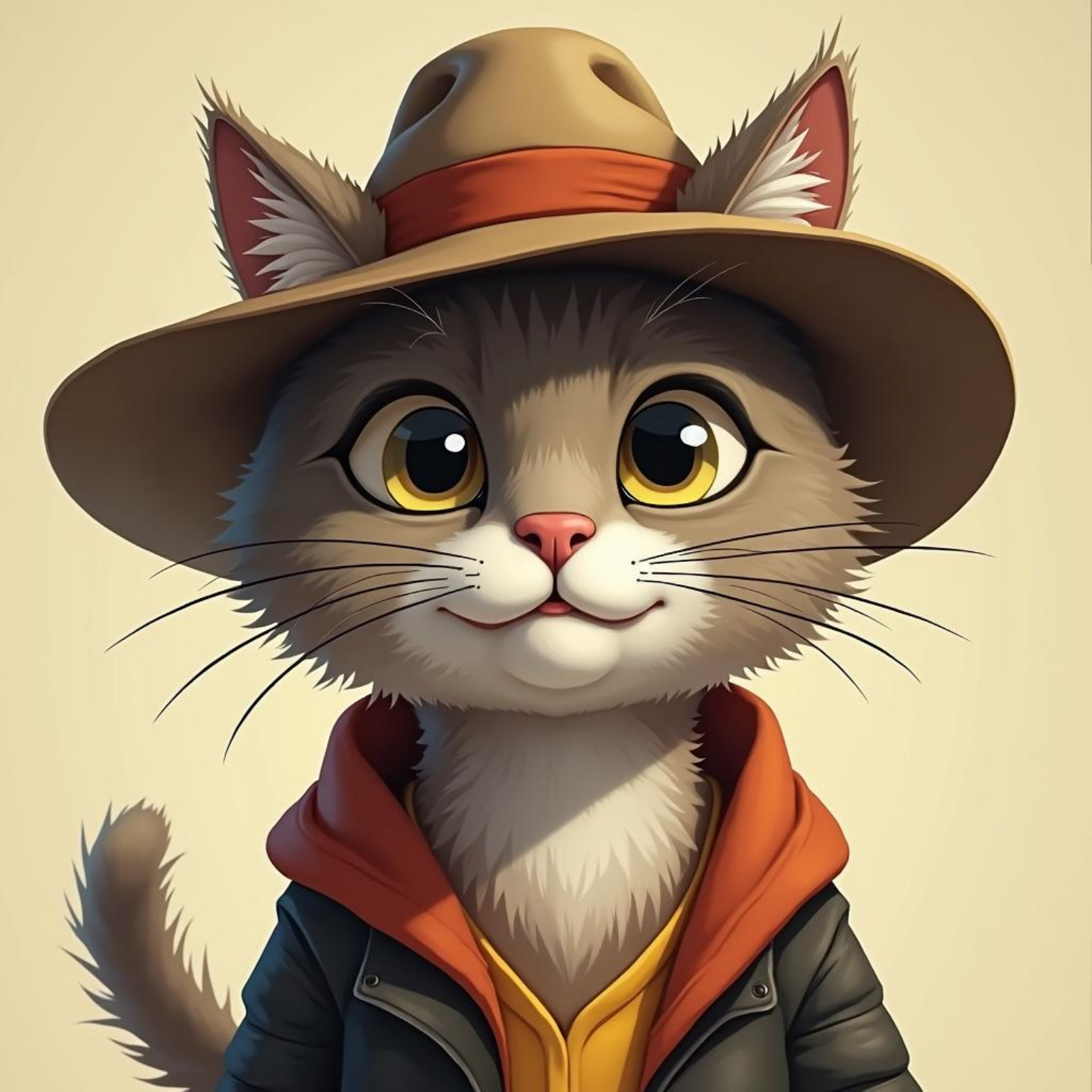} &
        \includegraphics[width=\linewidth]{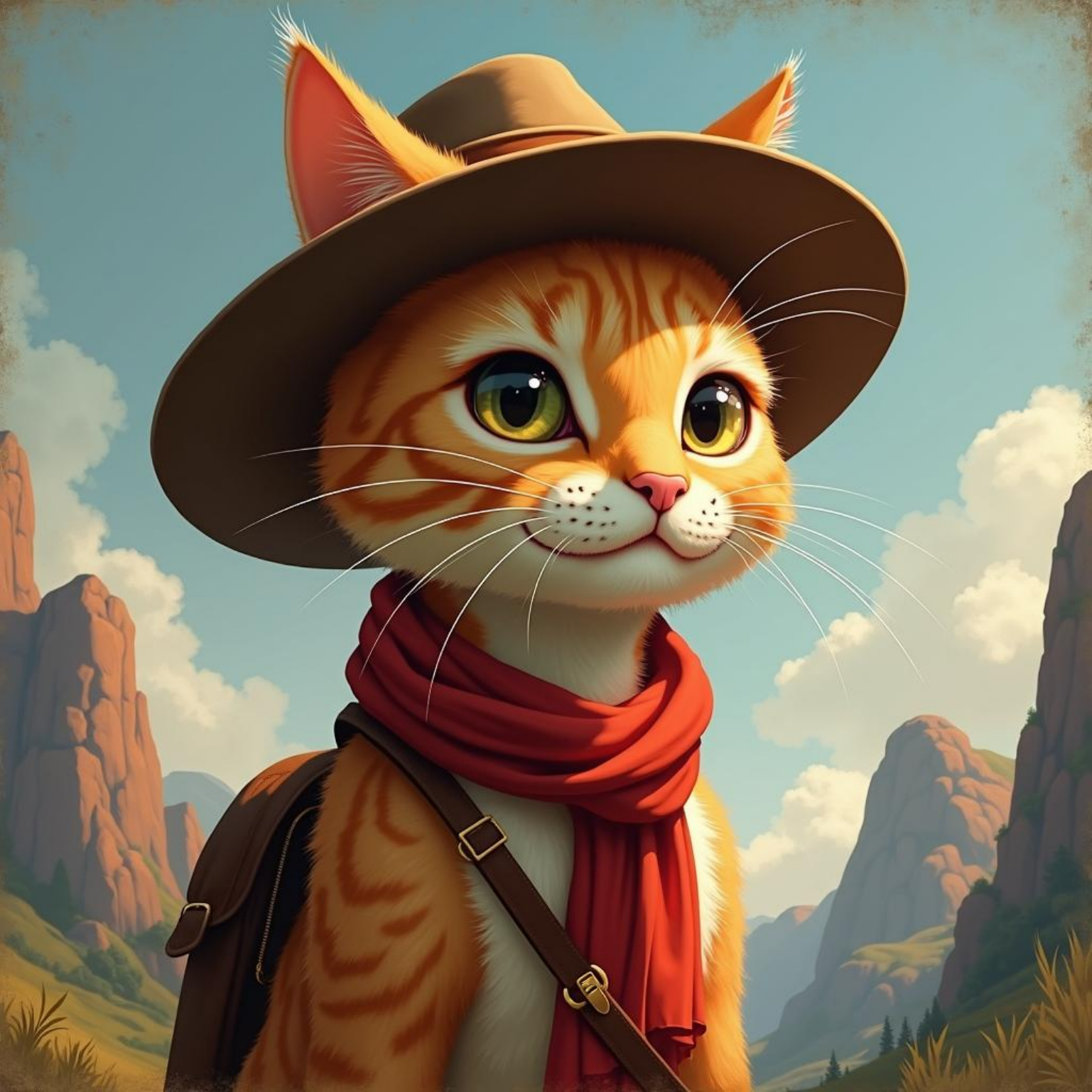} & 
        \includegraphics[width=\linewidth]{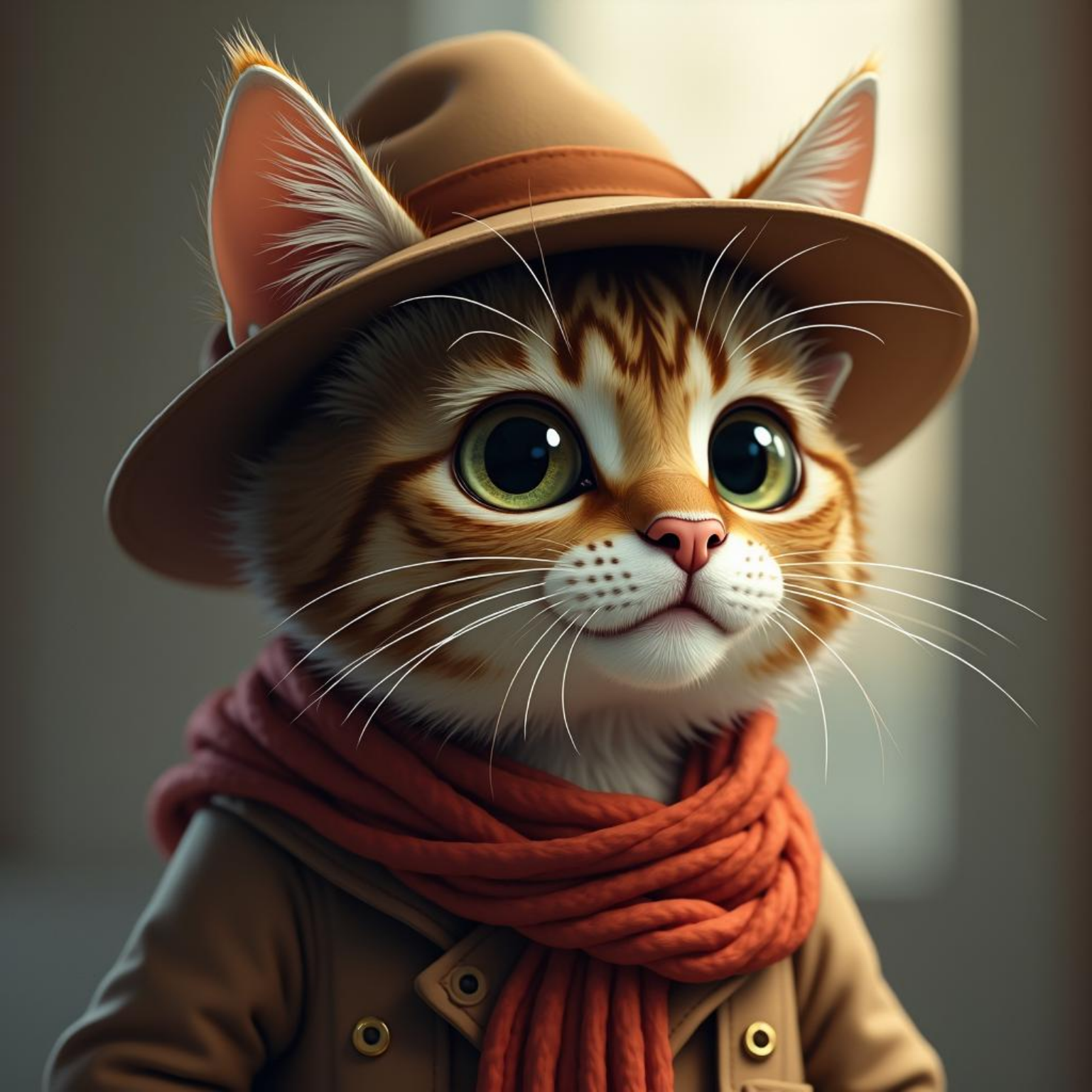} &
        
        \na &
        
        \includegraphics[width=\linewidth]{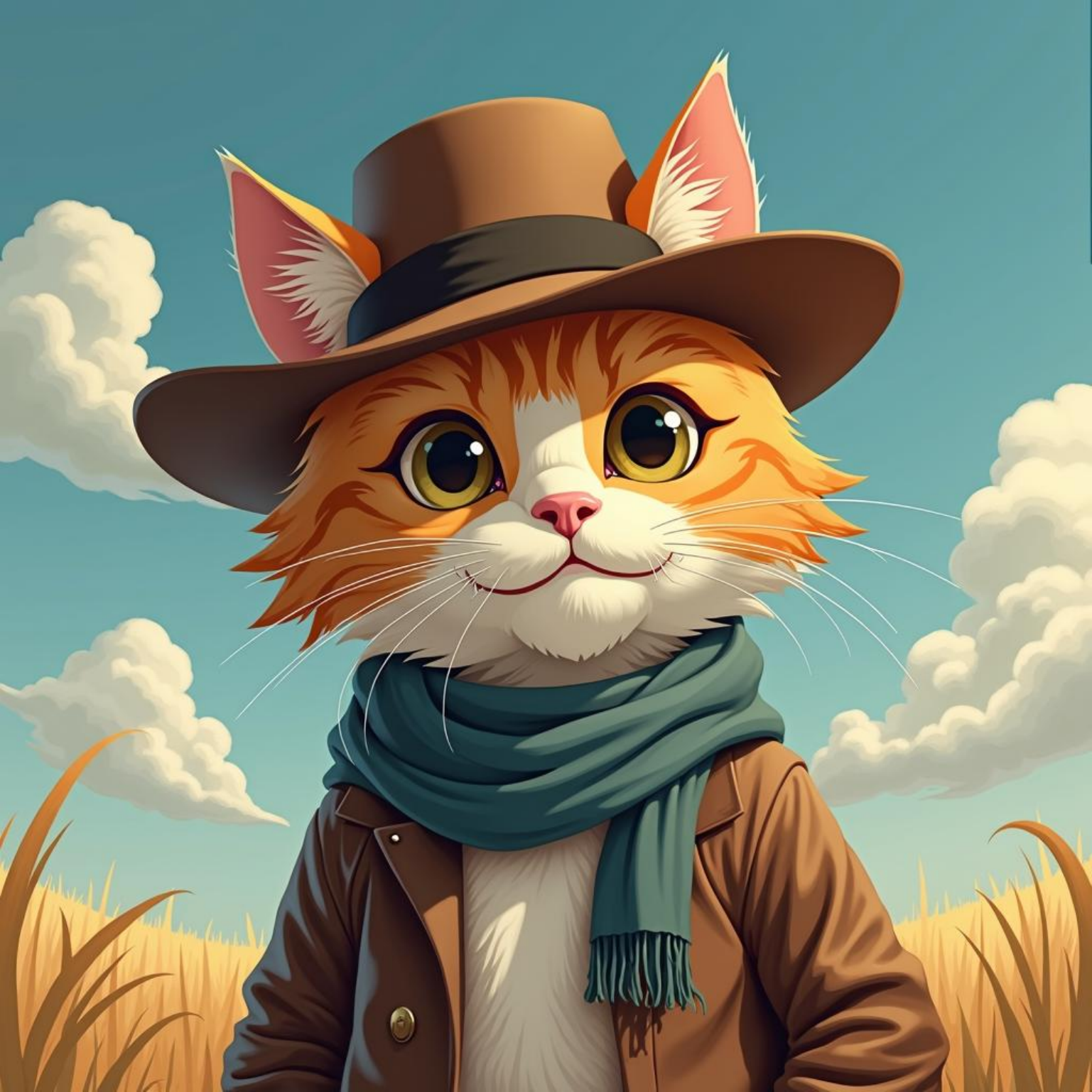} \\
        
        \multicolumn{6}{c}{
            \parbox{0.96\linewidth}{\centering \scriptsize
                \textit{Prompt: cat with a hat in the style of midjourney} \\
                \textsc{(Model: FLUX $\mid$ Reward: PickScore)}
            }
        } \\
        
        \bottomrule
    \end{tabularx}
    \caption{Qualitative comparison under a fixed budget ($\text{NRE}=200$). Visual samples generated by SES versus baselines across different base models (SDXL, FLUX) and reward objectives, demonstrating SES's superior reward alignment.}
    \label{fig:Qualitative_comparison}
    % \vspace{-4mm}
\end{figure}

\textbf{Baseline Comparison.} We benchmark the alignment effectiveness of SES against representative inference-time search strategies under a fixed computational budget ($\text{NRE}=200$). As detailed in Table \ref{tab:main_results}, SES consistently achieves the highest reward scores across all settings, significantly outperforming BoN and trajectory-based approaches. Qualitative results, visualized in Figure \ref{fig:Qualitative_comparison}, demonstrate SES's capability to generate high-fidelity samples even under highly non-convex human preference objectives. Notably, SES maintains a decisive advantage across distinct generative paradigms (Latent Diffusion and Flow Matching), underscoring its robust cross-architecture generalization. (See Appendix \ref{sec:baseline_comparison} for full results on SD v1.5 and Qwen-Image).

\begin{figure}[t]
	\centering
	 \includegraphics[width=\linewidth]{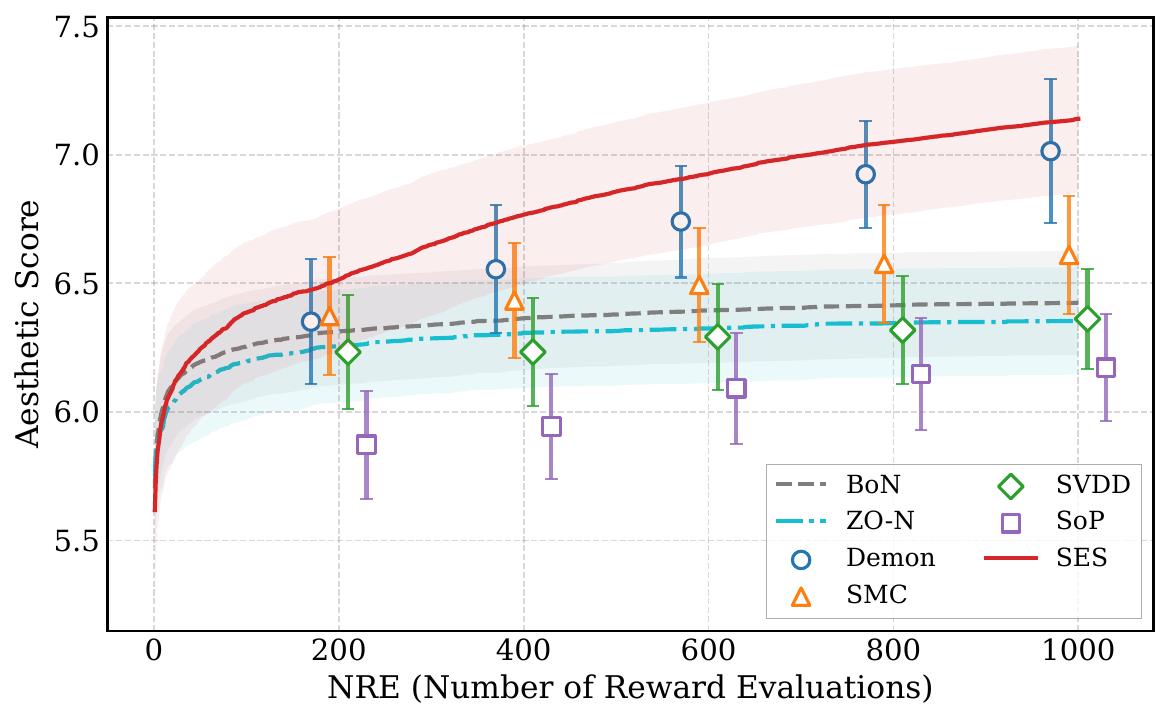} 
    \caption{Scaling behavior analysis. We investigate scaling behaviors by extending the computational budget to $\text{NRE}=1000$, using SDXL with Aesthetic Score. Curves depict the continuous scaling trajectory, while markers indicate performance measured at discrete checkpoints ($\text{NRE} \in \{200, 400, 600, 800, 1000\}$). SES exhibits continuous performance gains.}
	\label{fig:scaling_law}
    % \vspace{-4mm}
\end{figure}

\textbf{Scaling Behavior.} We further investigate the scaling properties of SES by extending the computational budget to $\text{NRE}=1000$. As plotted in Figure \ref{fig:scaling_law}, Best-of-N exhibits logarithmic-like scaling behavior, rapidly plateauing after $\text{NRE} > 200$. In contrast, within the tested budget range, SES maintains a steep scaling trajectory without saturation throughout the tested range, demonstrating sustained performance growth. Furthermore, SES consistently outperforms trajectory-based methods across the entire budget spectrum. This confirms that restricting the search to the low-frequency manifold concentrates the computational budget on the most influential degrees of freedom, enabling more effective exploration compared to indiscriminate full-space sampling.

\subsection{Mechanism Analysis and Efficiency}

\textbf{Ablation Studies.}
We perform a comprehensive ablation study on SDXL (Table \ref{tab:ablation}) to identify the key drivers of SES's performance.
\textit{Subspace Selection.} Searching within the low-frequency subspace yields superior results compared to both full-frequency and high-frequency settings. Notably, SES significantly outperforms random subspaces of equivalent dimensionality. This confirms that the gains stem not merely from dimensionality reduction, but from the low-frequency band's dominant control authority over generative results. \textit{Decomposition Granularity.} The decomposition level $J$ exhibits an inverted U-shaped impact, peaking at intermediate levels (e.g., $J=4$). This implies a necessary trade-off: sufficient spectral decoupling is required to filter out insensitive high-frequency dimensions ($J=2$ is too high-dimensional), yet excessive compression must be avoided to retain structural degrees of freedom ($J=6$ is too restrictive).
\textit{Optimization Strategy.} With the search space fixed, CEM substantially outperforms random search. This indicates that identifying the correct subspace is necessary but insufficient; the efficient exploration enabled by distribution evolution is equally critical.
In summary, the low-frequency manifold serves as an optimal regime that concentrates sensitive degrees of freedom while offering a tractable landscape for evolutionary optimization.

\begin{table}[t]
    \centering
    \small
    \renewcommand{\arraystretch}{1.2}
    % \caption{Ablation analysis on SDXL targeting Aesthetic Score. We systematically evaluate three key components against the \textbf{Default} configuration (LL Subband, Level $J{=}4$, CEM).}
    \caption{Ablation analysis on SDXL targeting Aesthetic Score ($\text{NRE}=200$). We systematically evaluate three key components against the \textbf{Default} configuration (LL Subband, Level $J{=}4$, CEM): (1) spectral subspace selection (Top), (2) wavelet decomposition levels $J$ (Middle), and (3) search strategies (Bottom).}
    \label{tab:ablation}
    
    \begin{tabularx}{\linewidth}{X c c} 
        \toprule
        \textbf{Ablation Setting} & \textbf{Aesthetic} $\uparrow$ & \textbf{$\Delta$} \\
        \midrule
        
        \textbf{Default} & \textbf{6.49} & \textbf{--} \\
        
        \midrule
        \multicolumn{3}{l}{\textit{\textbf{1. Subspace Selection}}} \\
        \hspace{1em} High Freq. (LH+HL+HH) & 5.84 & \textcolor{gray}{-0.65} \\
        \hspace{1em} Full Frequency Space & 5.96 & \textcolor{gray}{-0.53} \\
        \hspace{1em} Random Subspace & 5.66 & \textcolor{gray}{-0.83} \\
        
        \addlinespace[0.4em]
        \multicolumn{3}{l}{\textit{\textbf{2. Decomposition Level}}} \\
        \hspace{1em} Level 2 & 6.38 & \textcolor{gray}{-0.11} \\
        \hspace{1em} Level 6 & 6.29 & \textcolor{gray}{-0.20} \\
        
        \addlinespace[0.4em]
        \multicolumn{3}{l}{\textit{\textbf{3. Optimization Strategy}}} \\
        \hspace{1em} Random Search & 6.28 & \textcolor{gray}{-0.21} \\
        \bottomrule
    \end{tabularx}
\end{table}

\begin{figure}[t]
	\centering
	\includegraphics[width=0.85\linewidth]{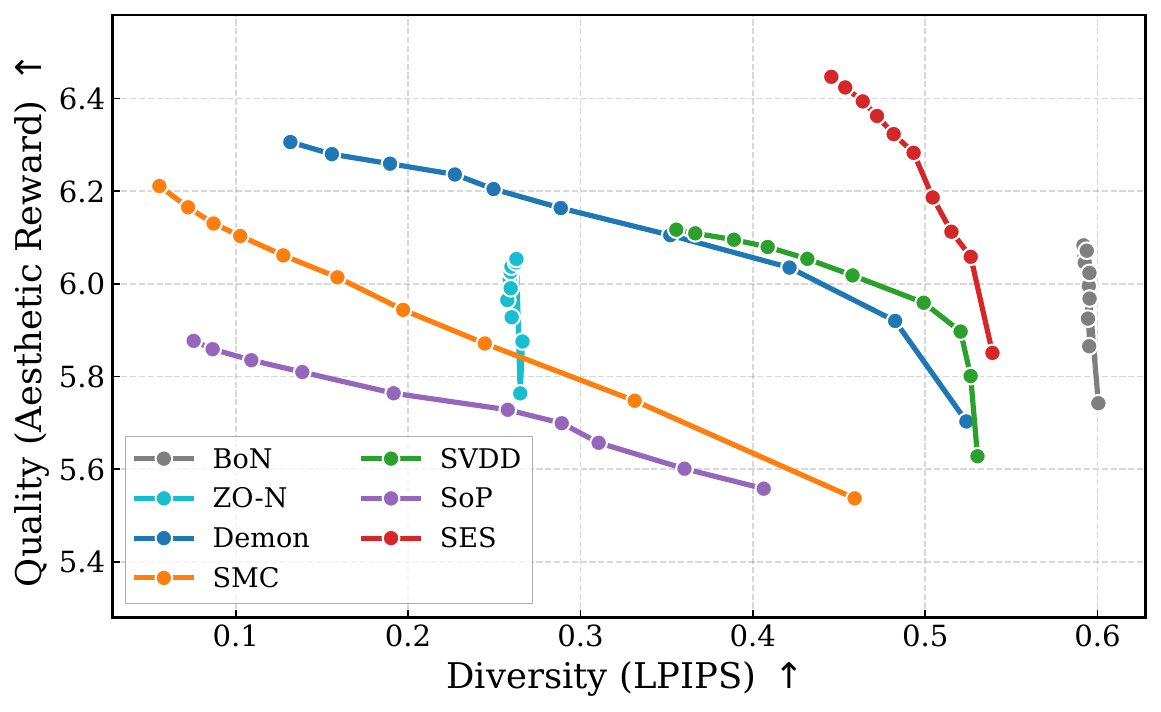} 
    \caption{Quality-Diversity trade-off analysis on SDXL. We visualize the optimization trajectory using Average Aesthetic Score (Quality) versus Pairwise LPIPS (Diversity). Markers along each curve represent snapshots at intervals of 20 NRE, ranging from $\text{NRE}=20$ to $200$. It demonstrats that SES effectively balancing generation quality with diversity.}
	\label{fig:diversity}
\end{figure}

\textbf{Quality-Diversity Trade-off.} We analyze the trade-off between exploration and exploitation using LPIPS-Reward trajectories (Figure~\ref{fig:diversity}). Particle filtering methods, constrained by hard resampling, suffer from \textit{sample impoverishment}, leading to premature convergence and mode collapse. In contrast, SES employs smooth distributional updates on the low-frequency manifold. This mechanism enables the population to migrate gradually toward high-reward regions, preserving diversity while optimizing rewards. Consequently, SES establishes a superior Pareto frontier, effectively balancing generation quality with diversity.

\begin{figure}[t]
    \centering
    \resizebox{\linewidth}{!}{%
        \setlength{\tabcolsep}{0pt}
        \renewcommand{\arraystretch}{0.5}

        \begin{tabular}{@{} c@{\hspace{1pt}}c@{\hspace{1pt}}c @{\hspace{3pt}} c@{\hspace{1pt}}c@{\hspace{1pt}}c @{}}
            
            \multicolumn{3}{c}{\small \textbf{SDXL}} & \multicolumn{3}{c}{\small \textbf{Qwen-Image}} \\
            \cmidrule(r{2pt}){1-3} \cmidrule(l{2pt}){4-6}
            
            \tiny Baseline & \tiny SES (Proxy) & \tiny SES (Accurate) & \tiny Baseline & \tiny SES (Proxy) & \tiny SES (Accurate) \\
            \tiny (Default) & \tiny (LCM-4s) & \tiny (Full-50s) & \tiny (Default) & \tiny (ODE-10s) & \tiny (Full-50s) \\
            \noalign{\smallskip}

            \includegraphics[width=0.16\linewidth]{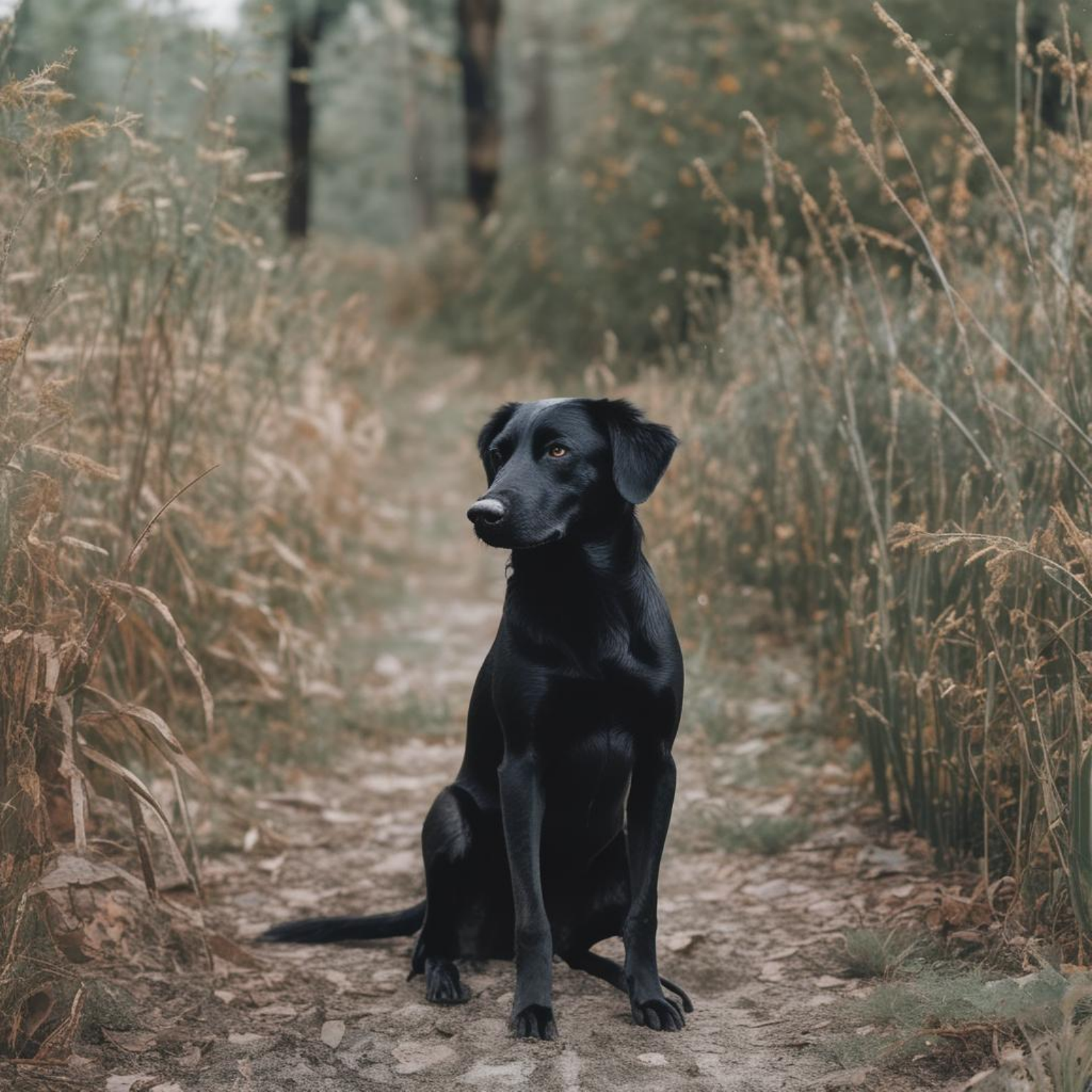} &
            \includegraphics[width=0.16\linewidth]{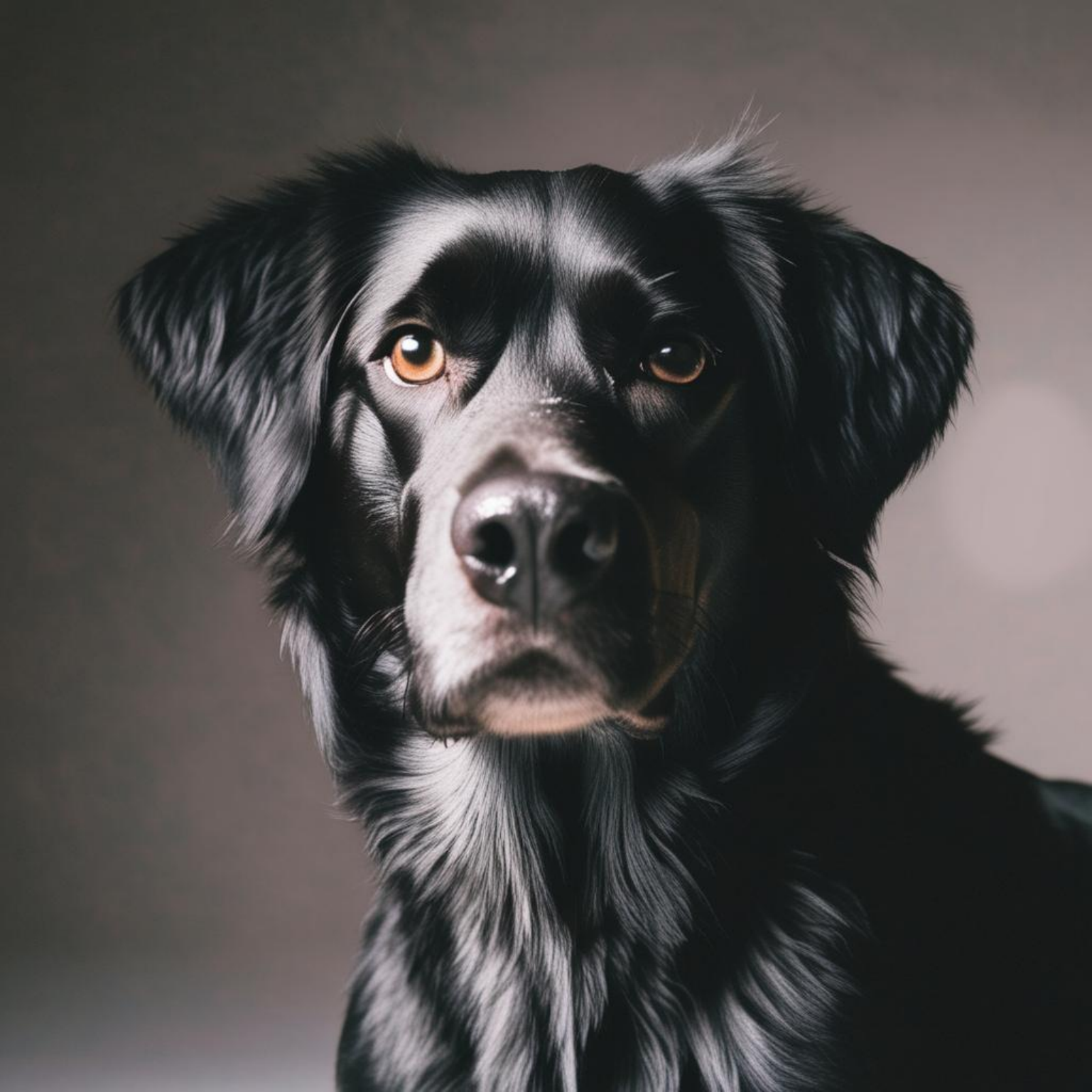} &
            \includegraphics[width=0.16\linewidth]{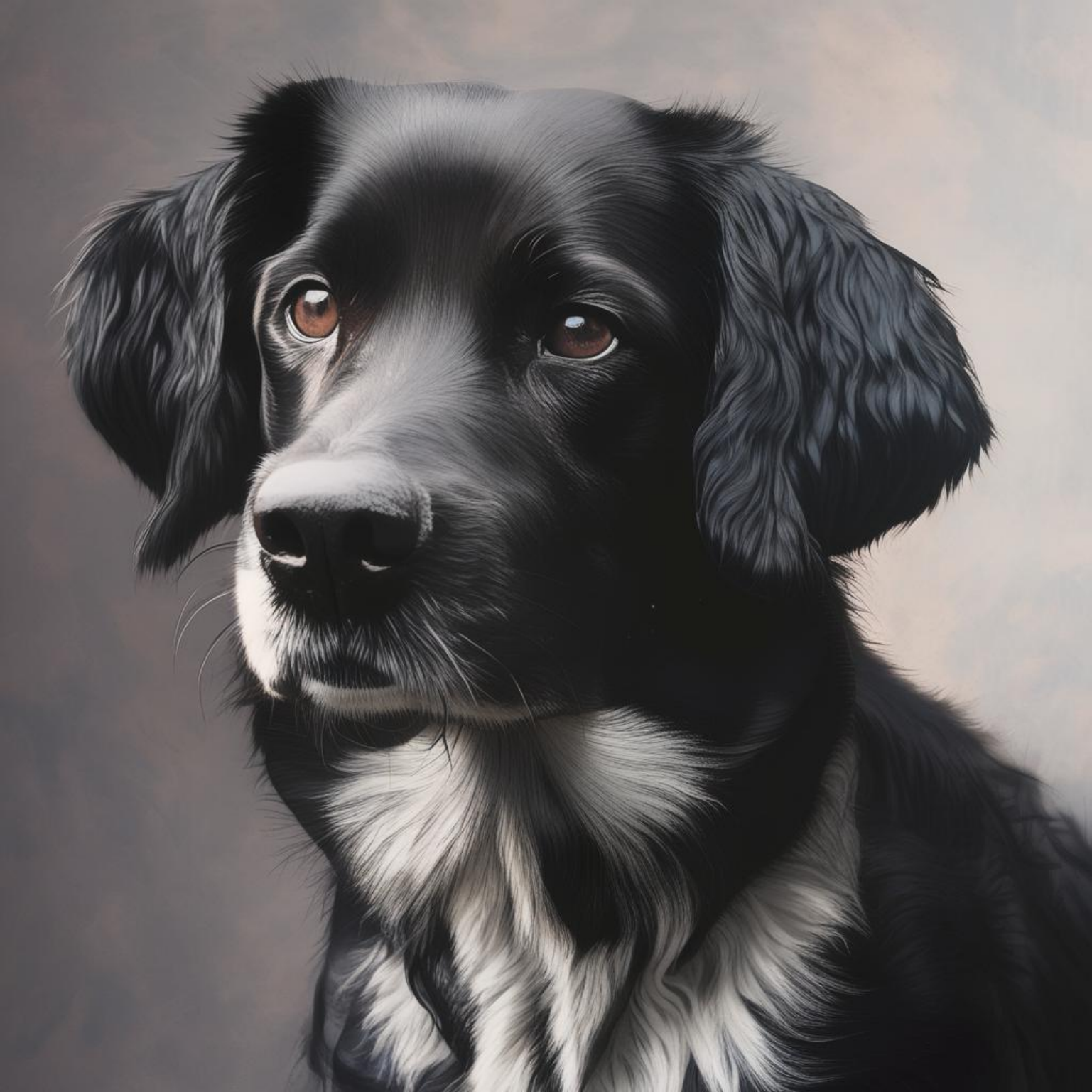} &
            \includegraphics[width=0.16\linewidth]{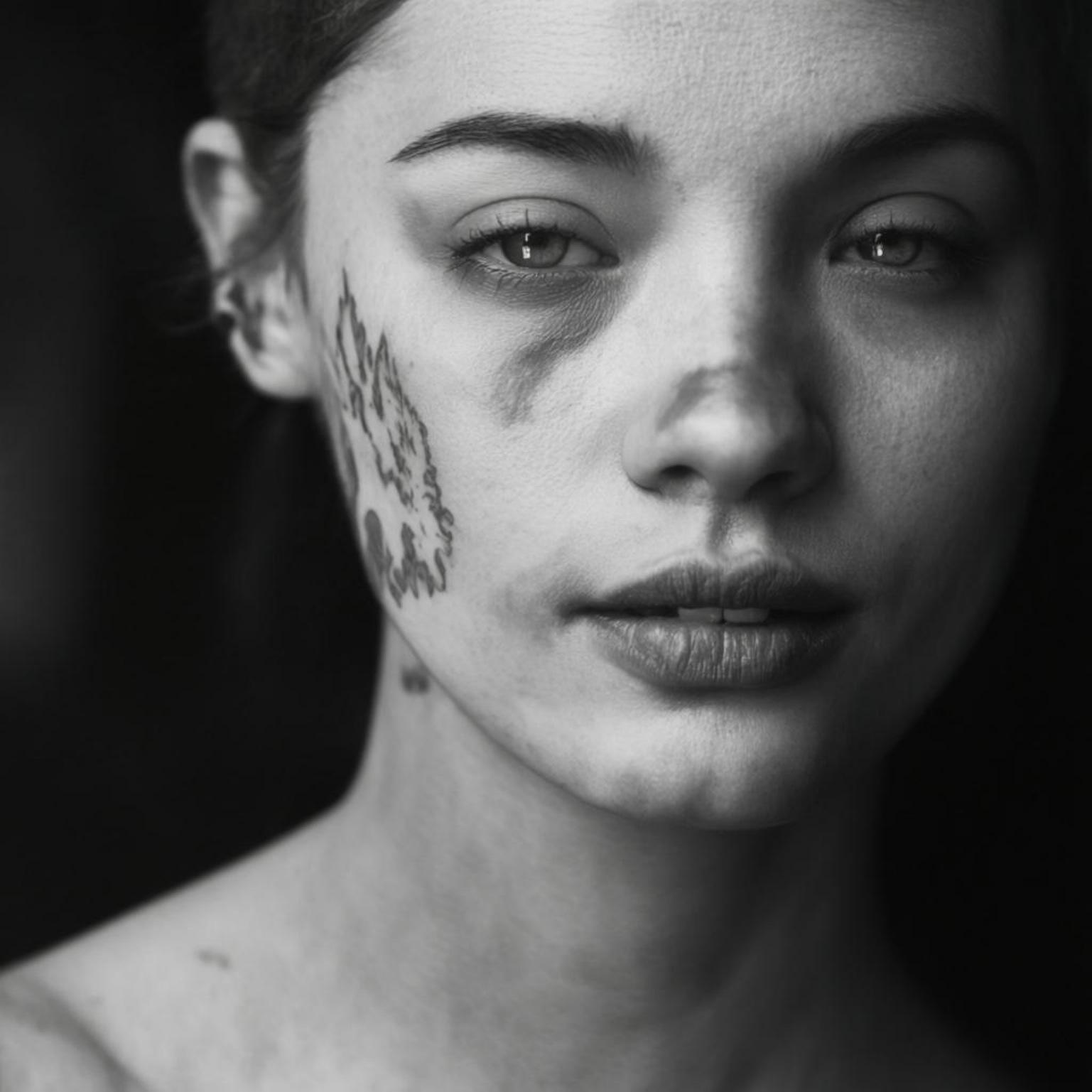} &
            \includegraphics[width=0.16\linewidth]{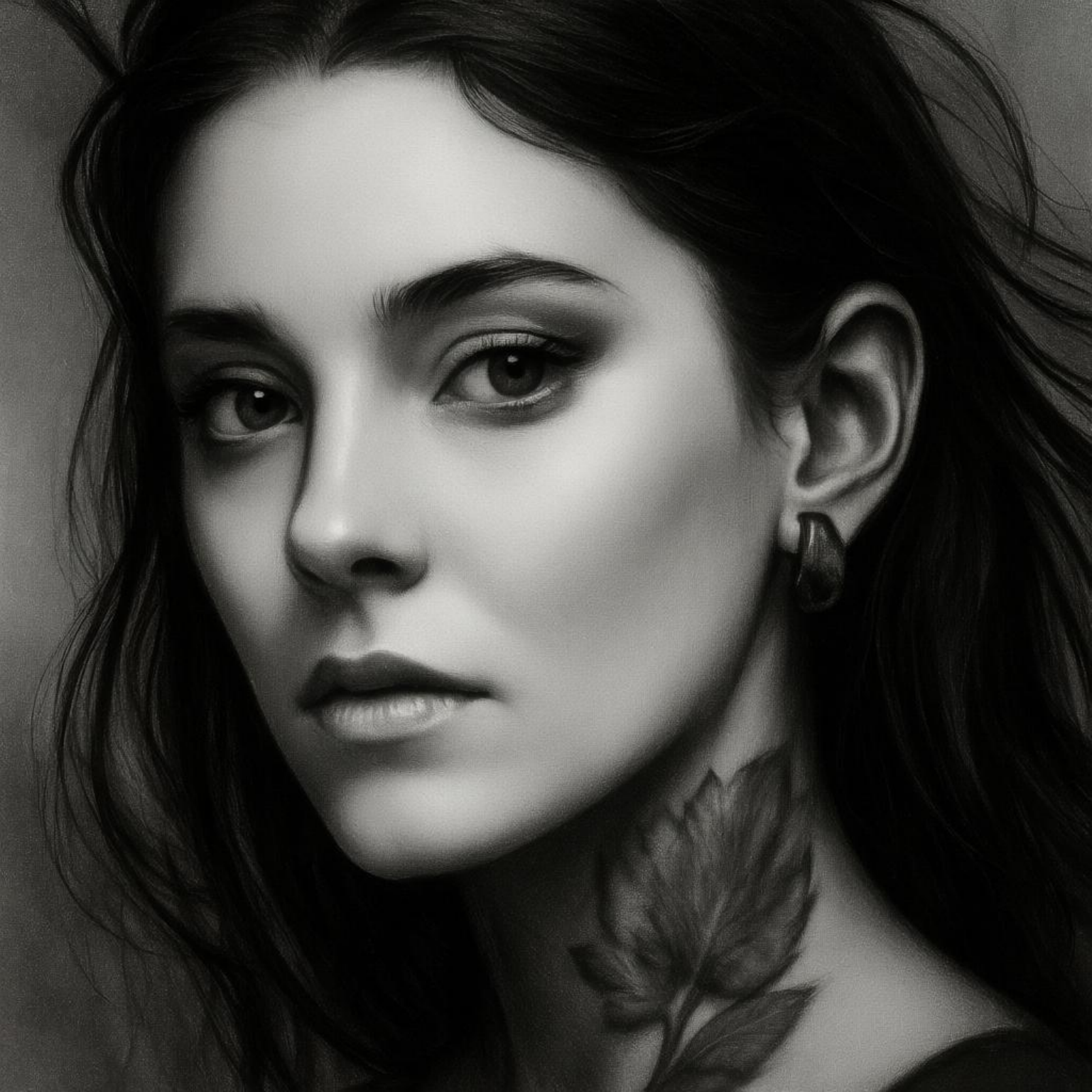} &
            \includegraphics[width=0.16\linewidth]{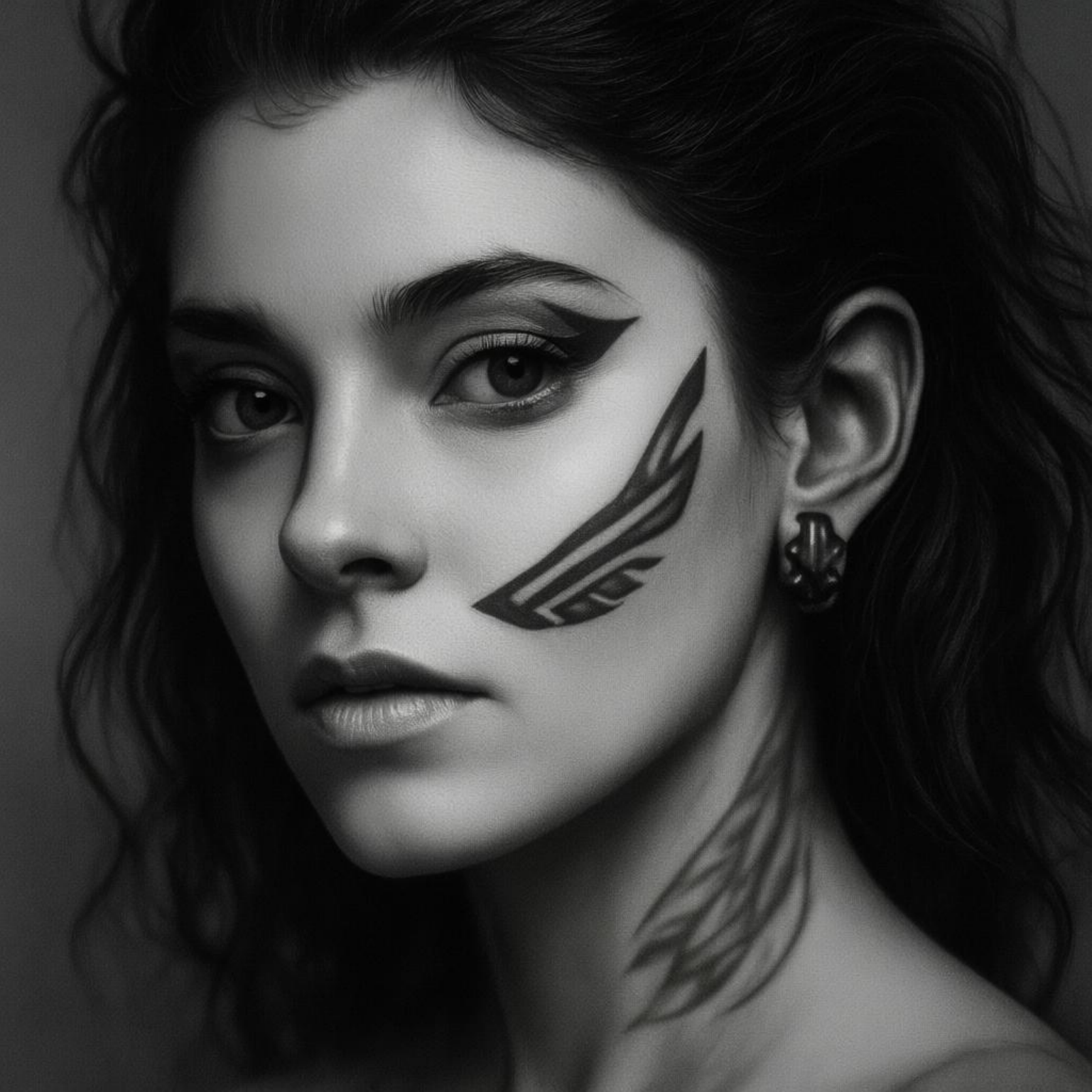} \\
            \addlinespace[1pt]

            \includegraphics[width=0.16\linewidth]{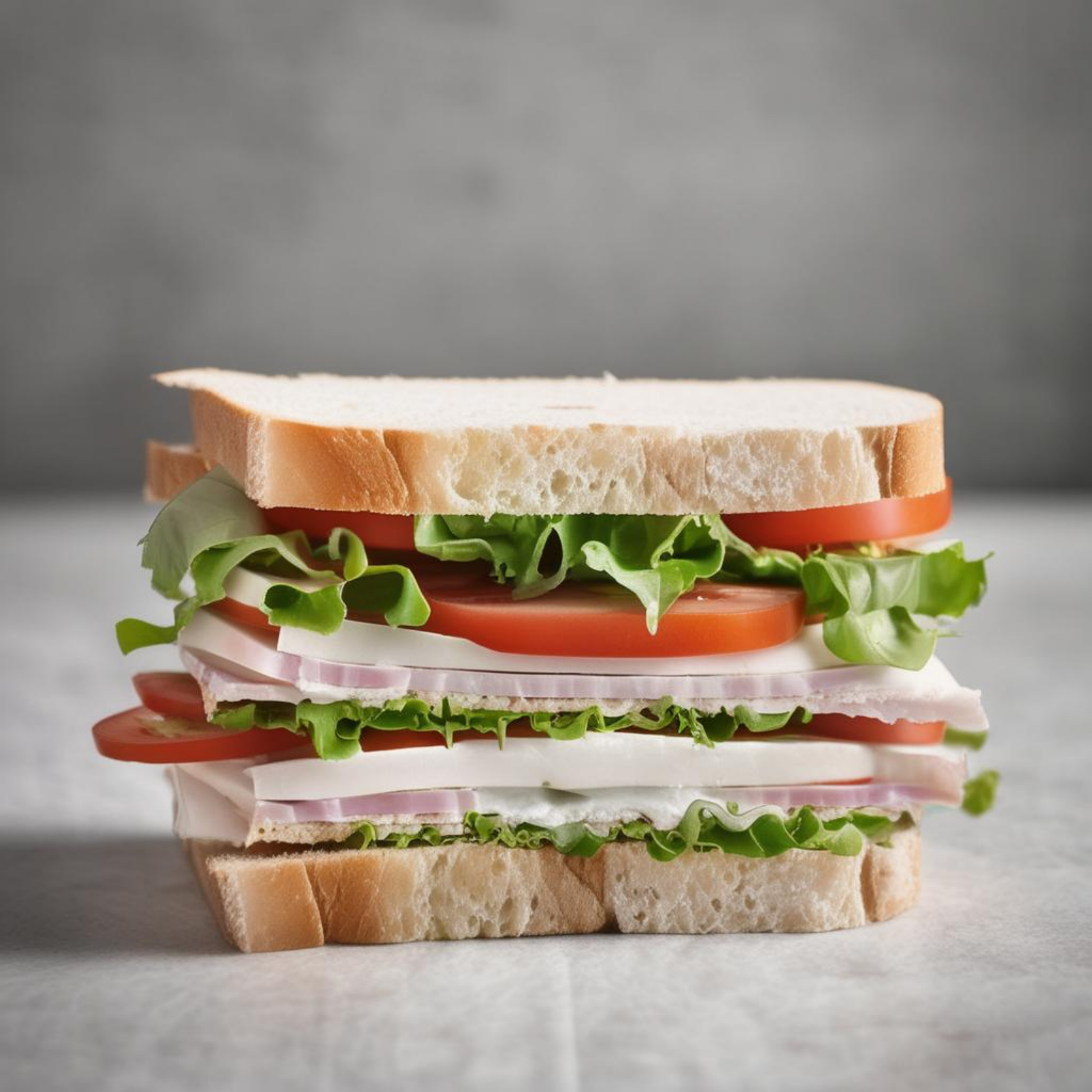} &
            \includegraphics[width=0.16\linewidth]{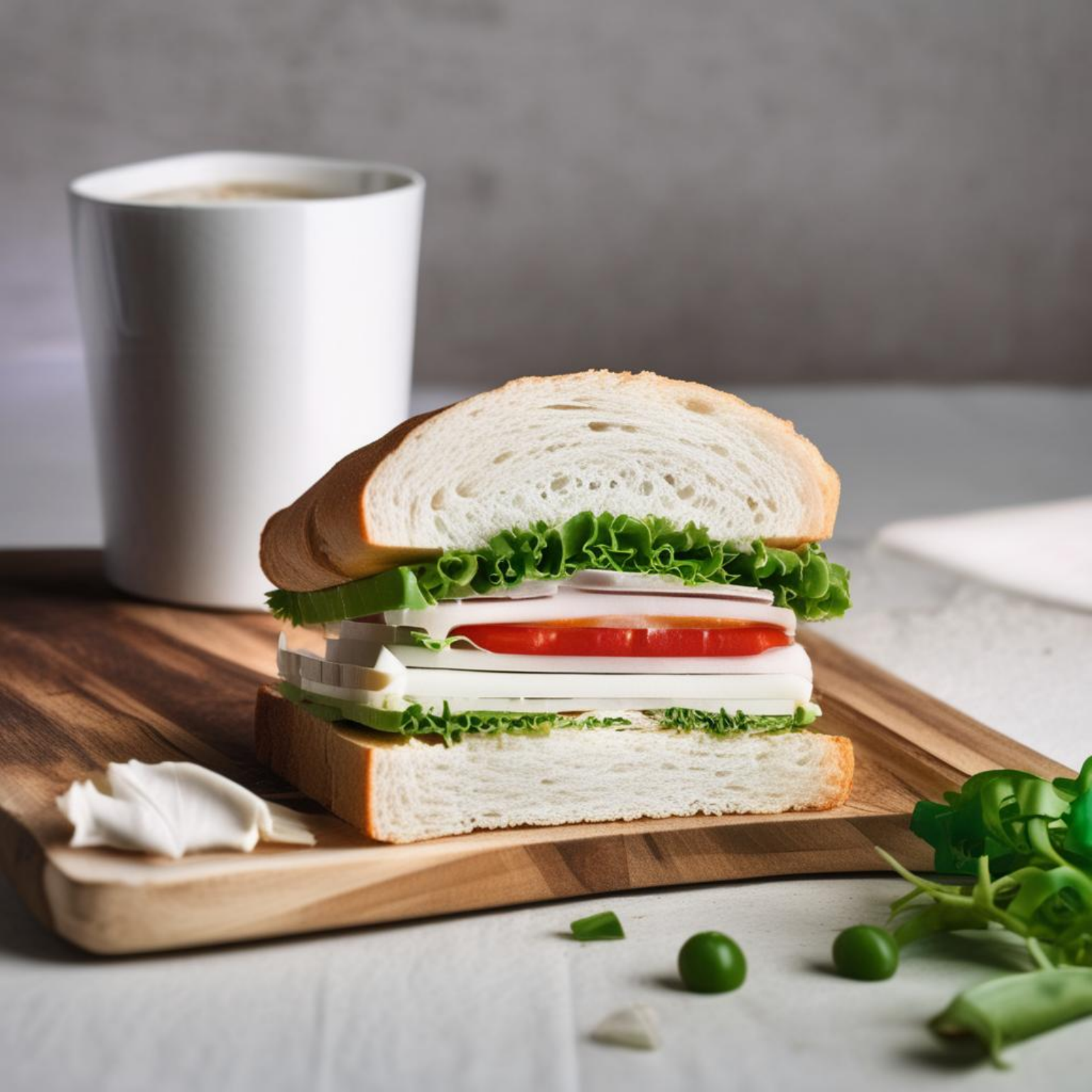} &
            \includegraphics[width=0.16\linewidth]{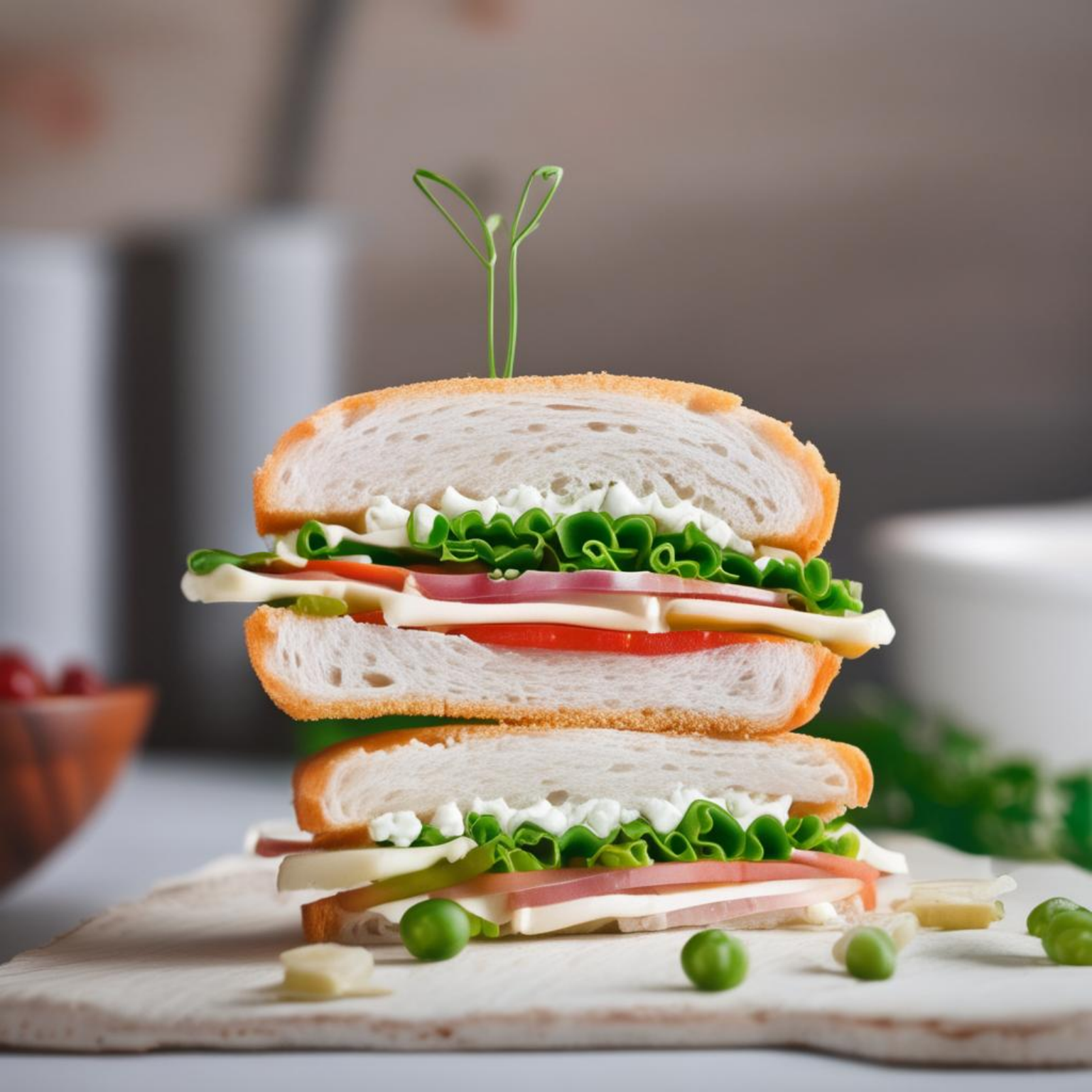} &
            \includegraphics[width=0.16\linewidth]{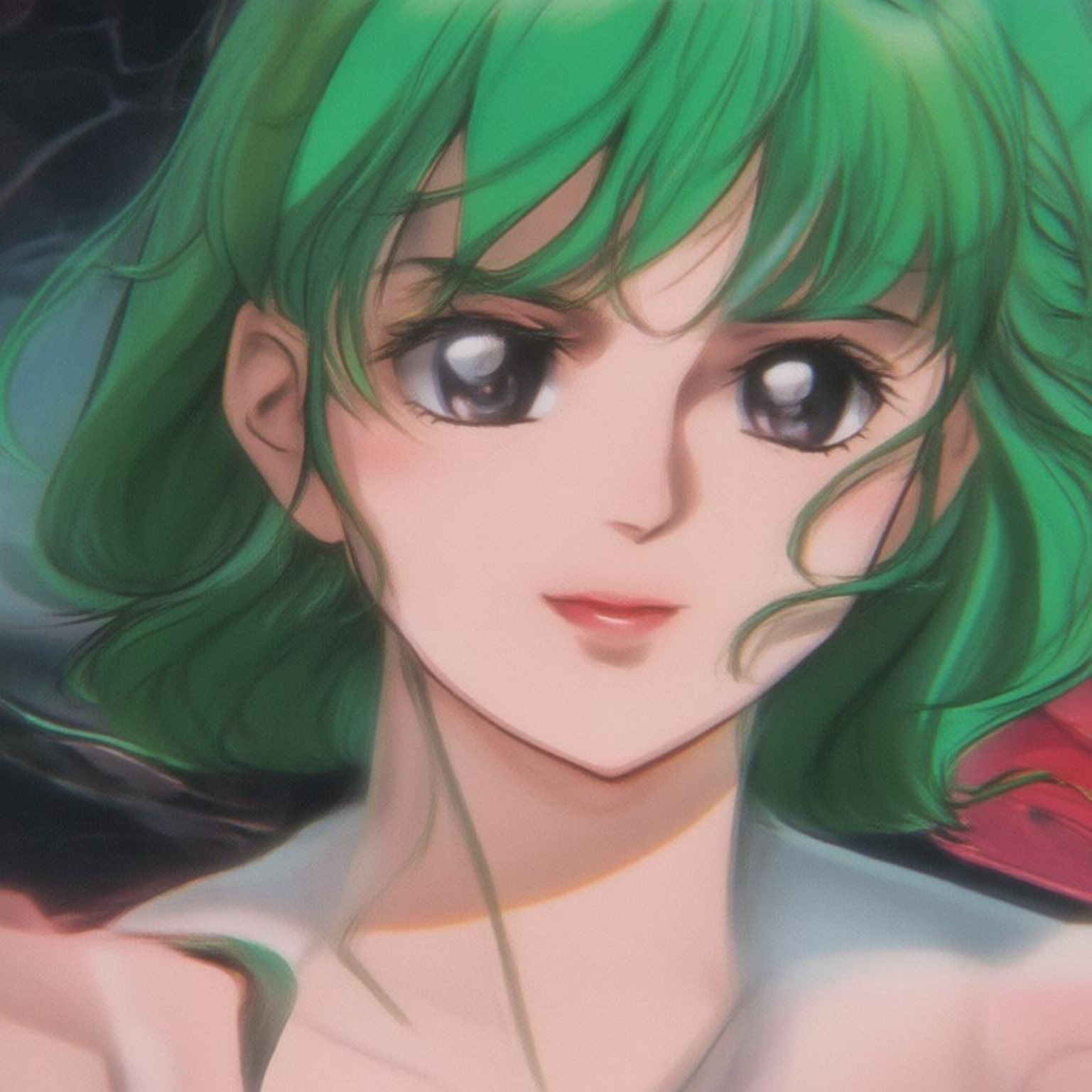} &
            \includegraphics[width=0.16\linewidth]{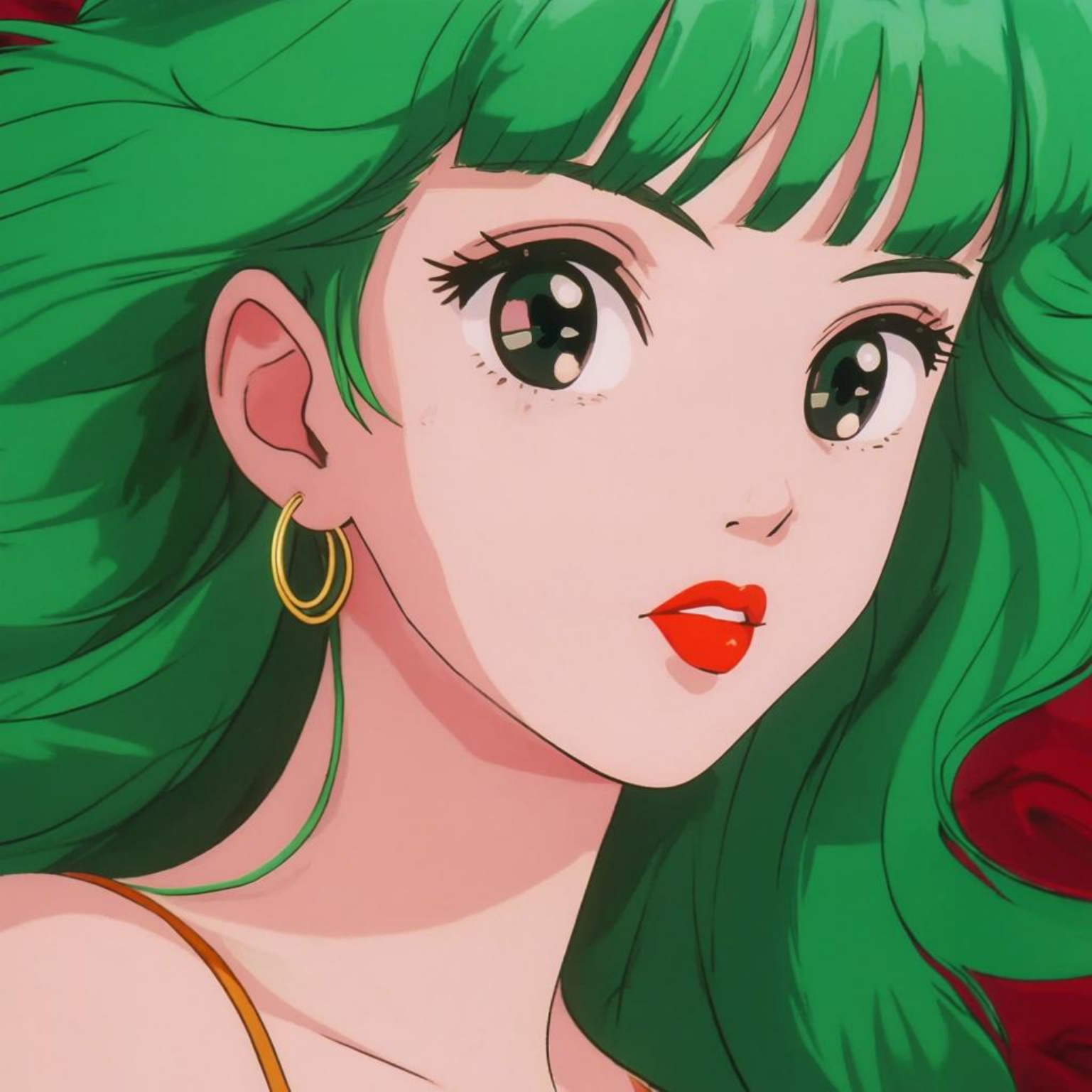} &
            \includegraphics[width=0.16\linewidth]{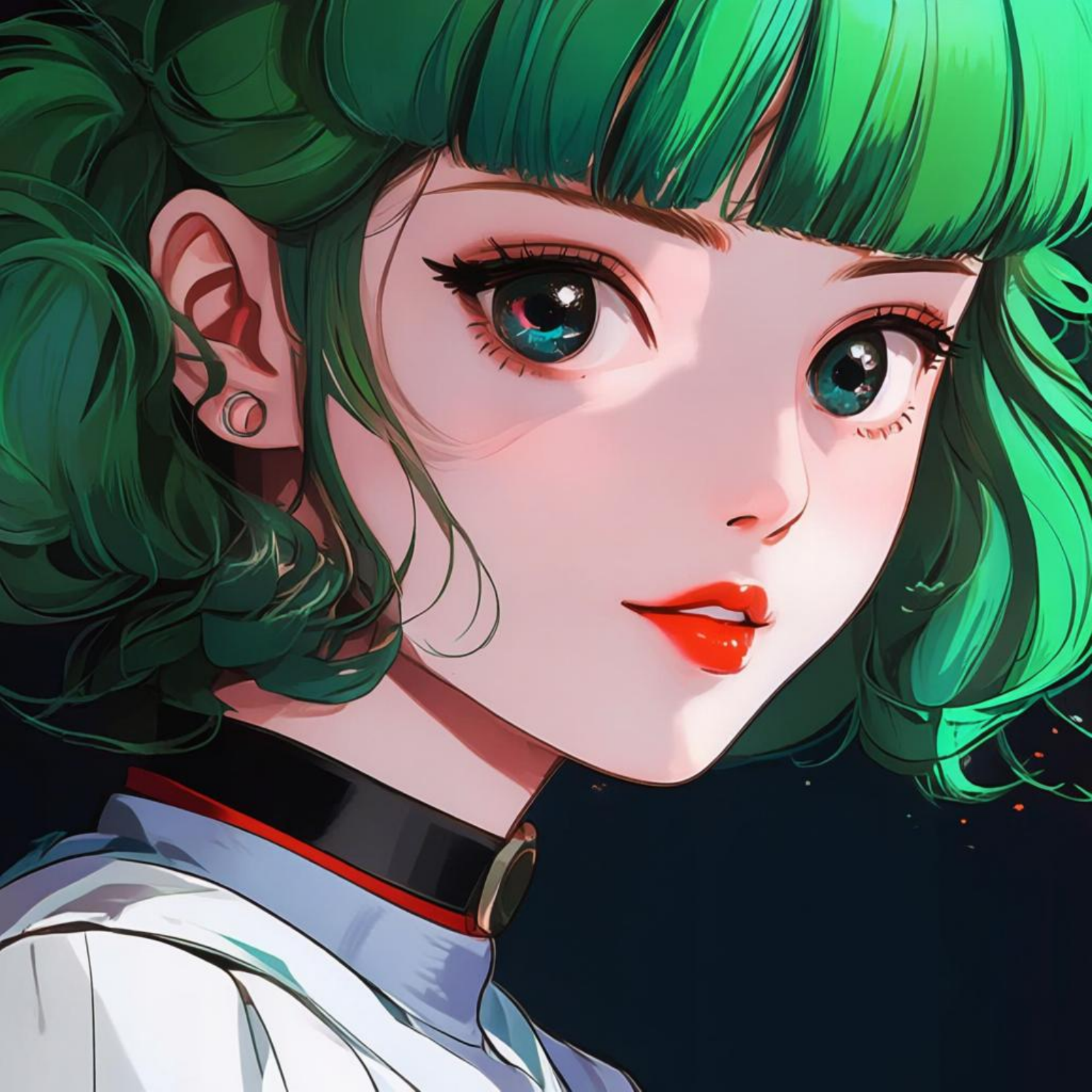} \\
            \addlinespace[1pt]

            \includegraphics[width=0.16\linewidth]{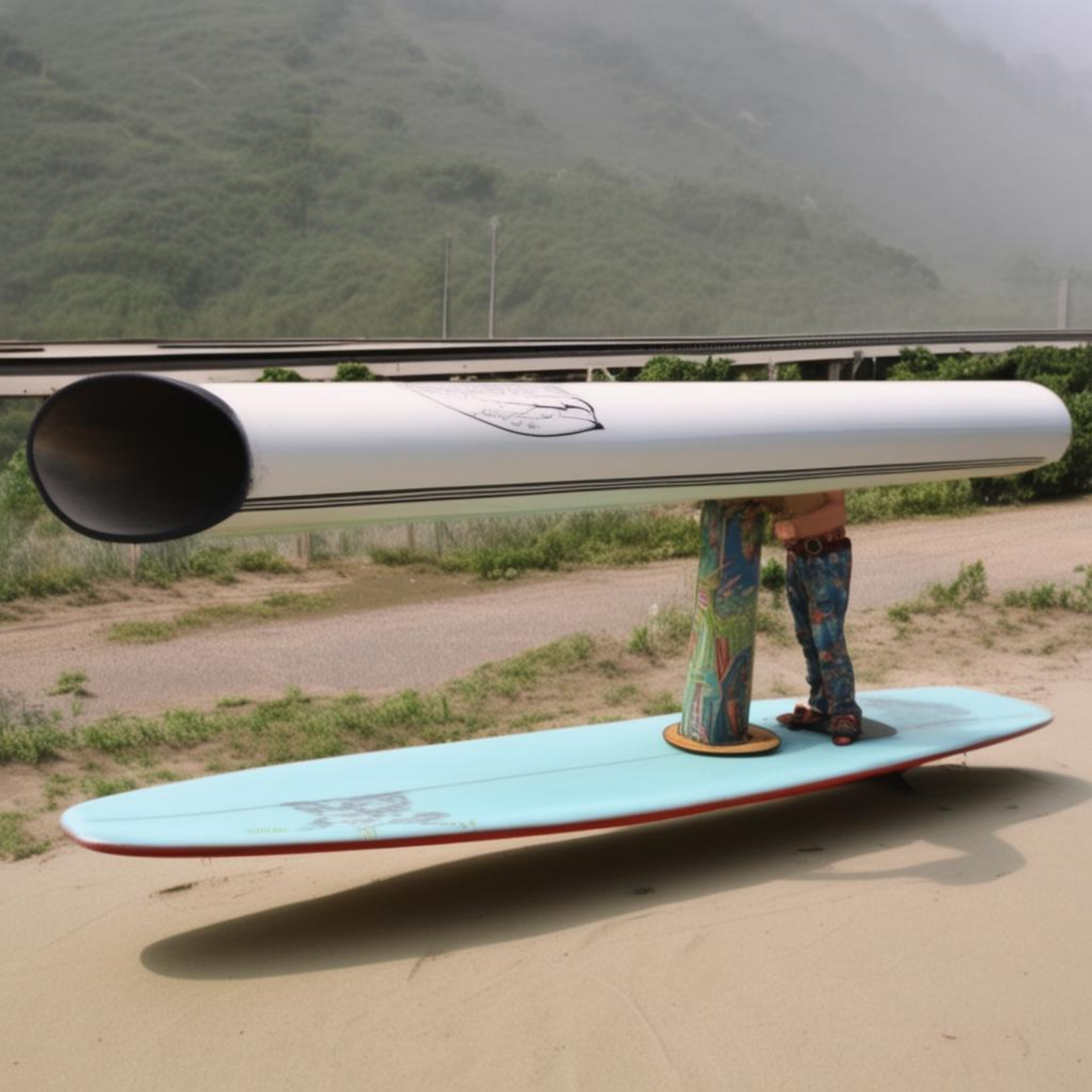} &
            \includegraphics[width=0.16\linewidth]{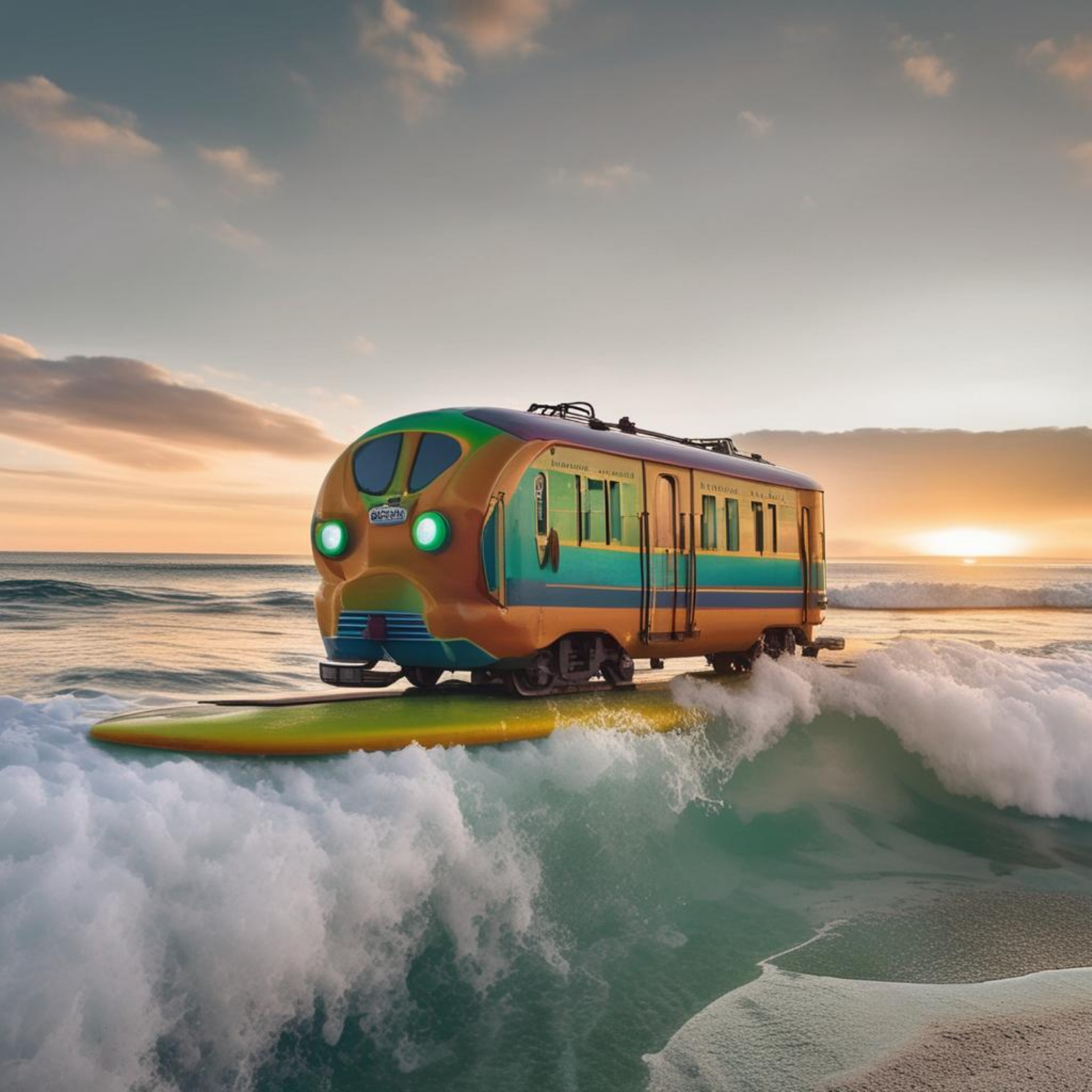} &
            \includegraphics[width=0.16\linewidth]{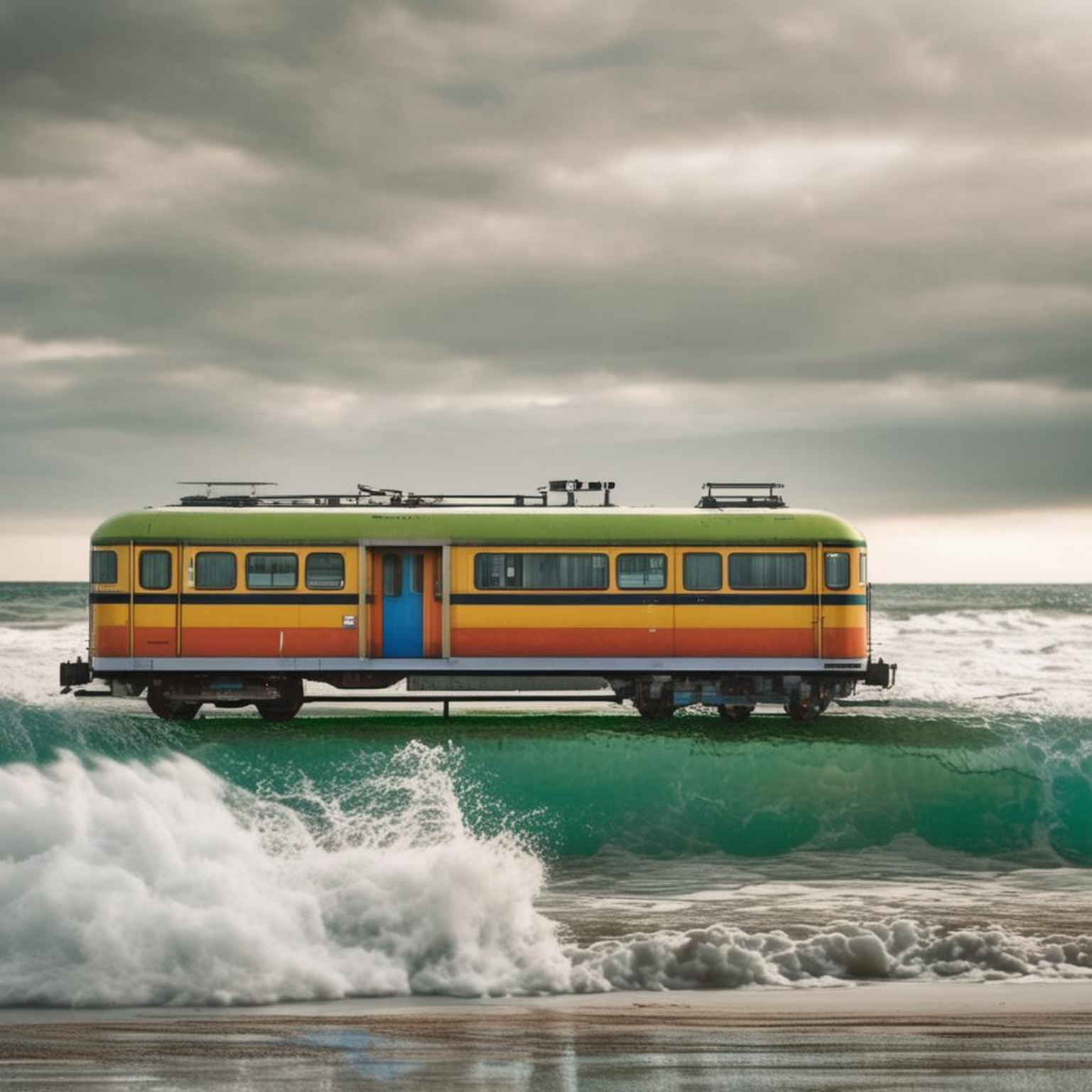} &
            \includegraphics[width=0.16\linewidth]{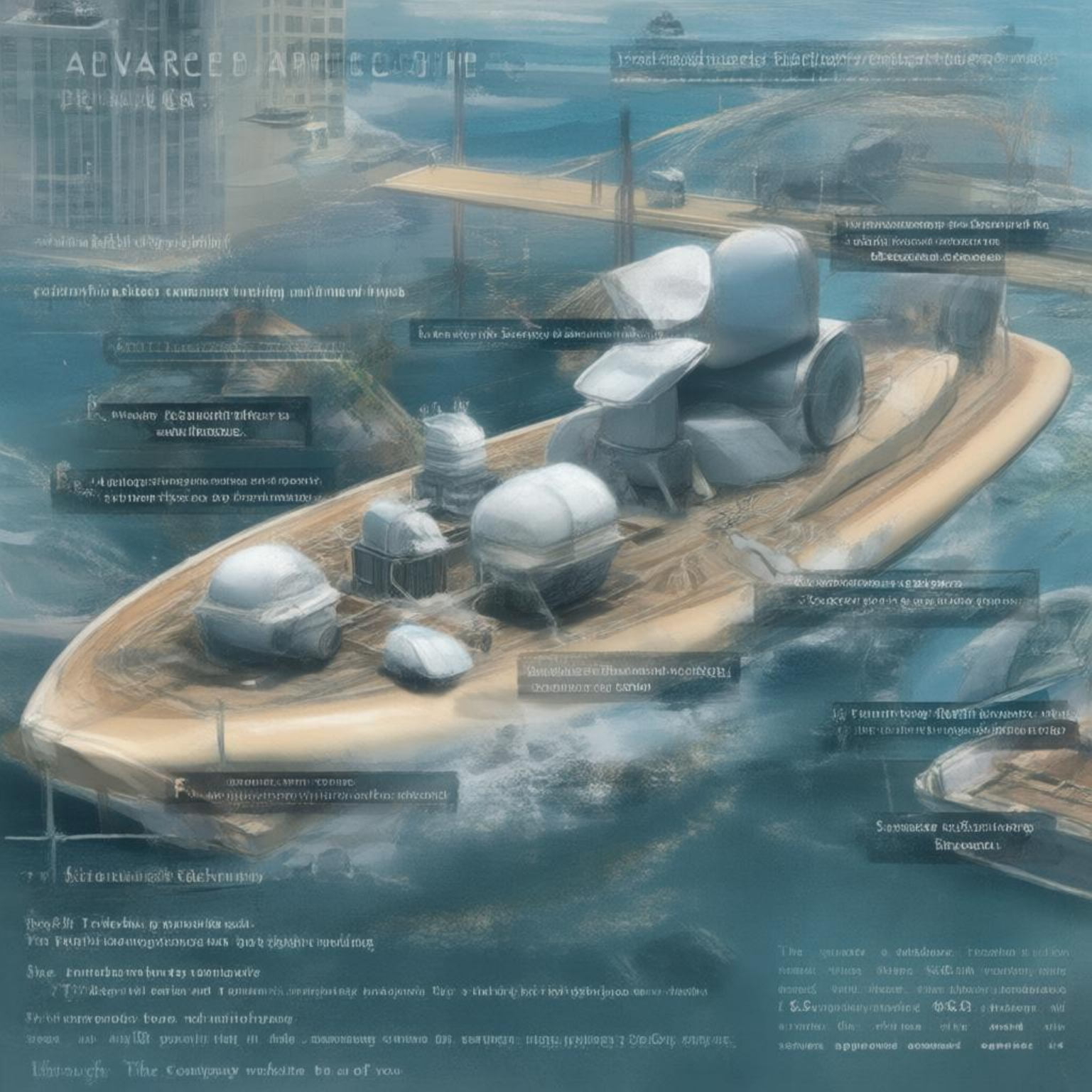} &
            \includegraphics[width=0.16\linewidth]{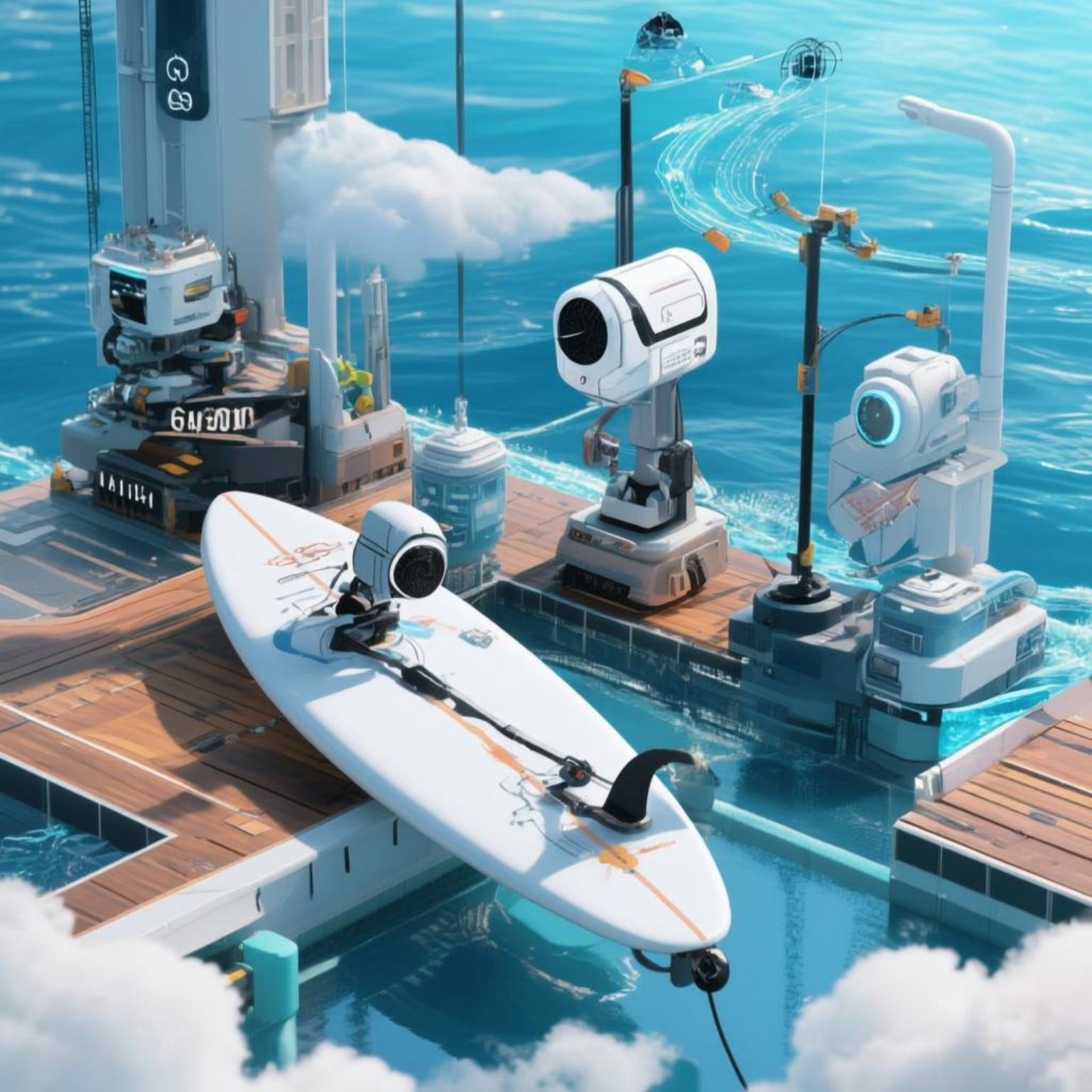} &
            \includegraphics[width=0.16\linewidth]{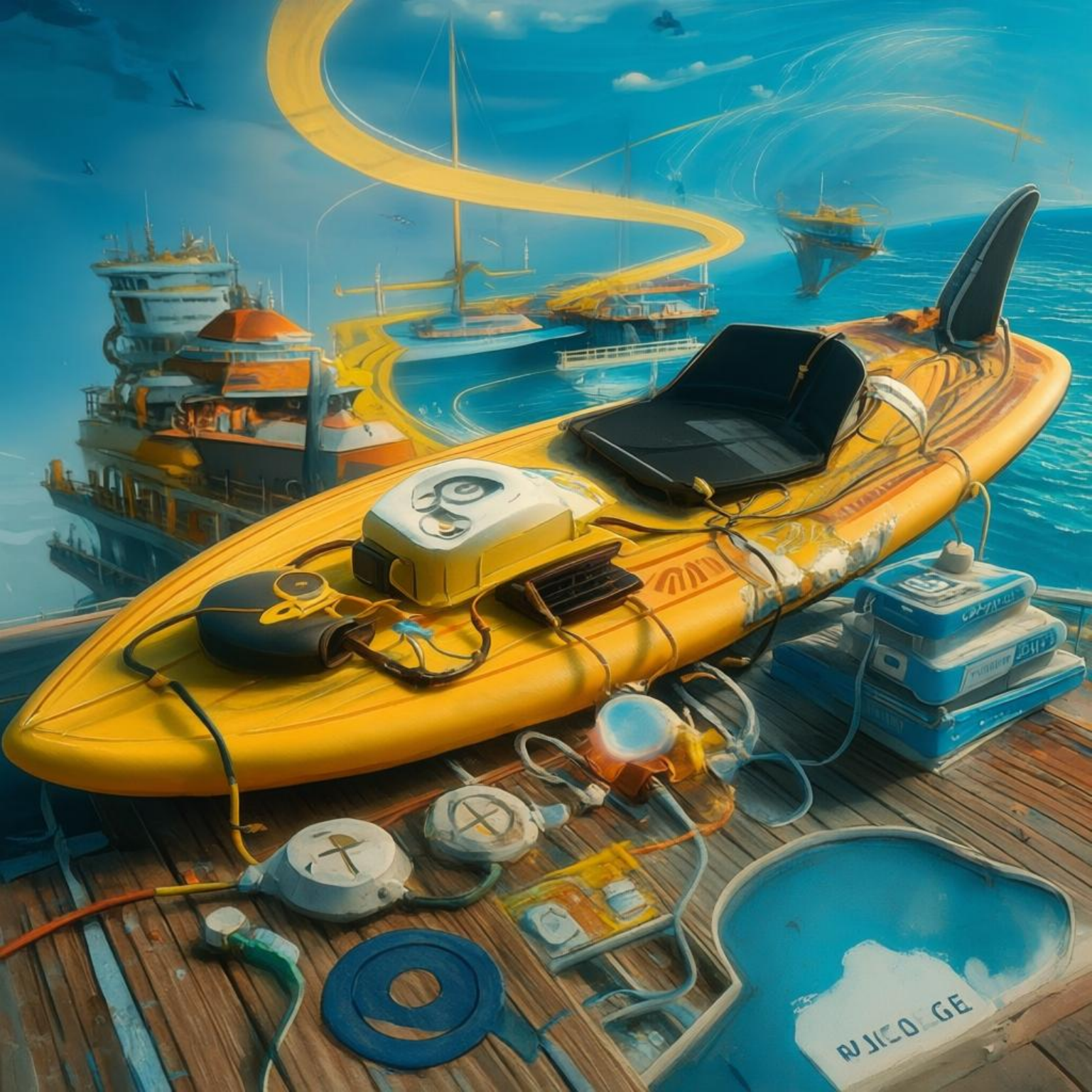} \\

        \end{tabular}%
    } 

    \caption{Qualitative comparison: Proxy vs. Accurate reward guidance targeting Aesthetic Score. Left (SDXL): Proxy via 4-step LCM. Right (Qwen-Image): Proxy via 10-step ODE, both targeting the Aesthetic Score. The high structural consistency between the two settings confirms that SES effectively leverages low-cost proxies to locate the optimal generation pattern.}
    \label{fig:proxy_qualitative}
\end{figure}

\begin{figure}[t]
	\centering
	\includegraphics[width=0.85\linewidth]{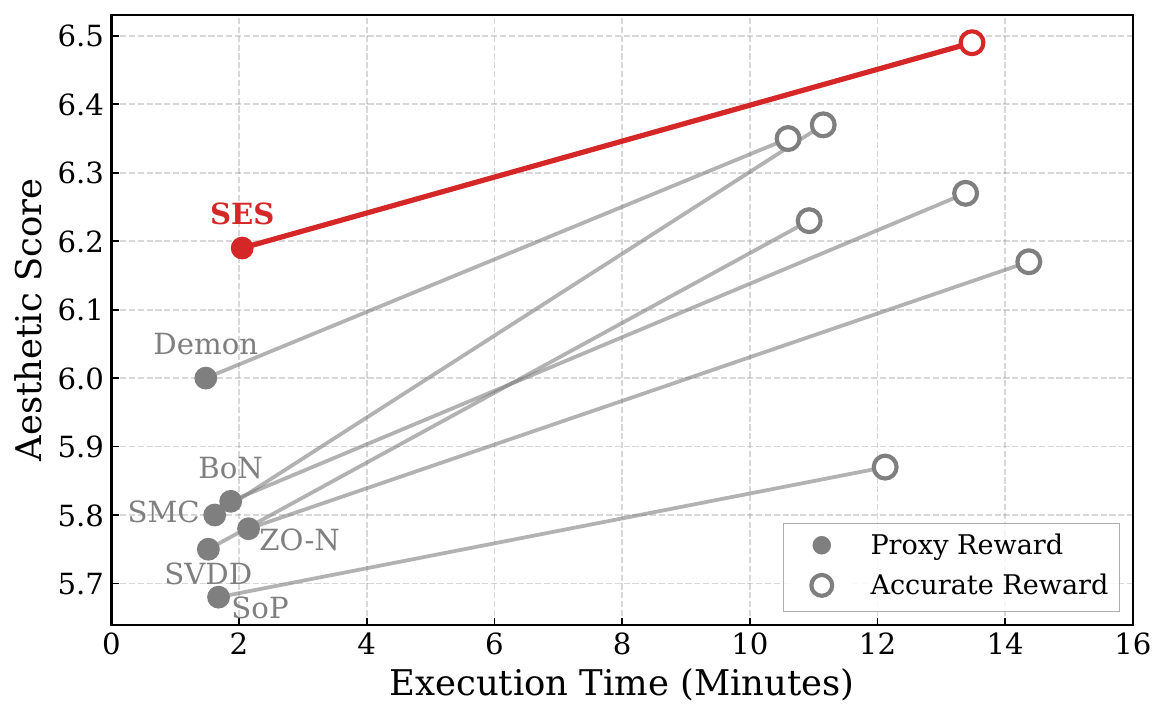} 
    \caption{Efficiency analysis on SDXL comparing Accurate vs. Proxy guidance. The adoption of Proxy evaluation reduces execution time by $\sim$85\%, and SES consistently outperforms all baselines under identical evaluation settings.}
	\label{fig:proxy_analysis}
    % \vspace{-2mm}
\end{figure}

\textbf{Accelerated Search via Proxy Guidance.} Reward computation necessitates mapping latent candidates to the image space. To circumvent the high latency of the standard iterative decoding, we implement a proxy mechanism: employing 4-step Latent Consistency Model (LCM) \cite{luo2023latent} for Latent Diffusion, and 10-step ODE integration for Flow Matching. Its success hinges on ranking consistency: although proxy samples are generated via simplified paths, they retain sufficient semantic fidelity to preserve relative candidate rankings, effectively guiding the rank-based evolutionary search. 
Empirically, as shown in Figure \ref{fig:proxy_analysis}, this strategy reduces evaluation costs by approximately 85\%, and SES significantly outperforms baselines on SDXL under identical proxy-guidance conditions. 
Qualitatively, Figure \ref{fig:proxy_qualitative} demonstrates that proxy-guided samples significantly outperform baselines, while maintaining high structural consistency with results from accurate reward evaluation.
% This confirms that SES effectively leverages efficient sampling approximations to achieve a superior trade-off between computational cost and generation quality.

\subsection{Generalization and Robustness}

\textbf{Orthogonality to Training-Time Alignment.} A critical question is whether inference-time scaling merely compensates for deficiencies in unaligned models. To investigate this, we applied SES to models fine-tuned via Direct Preference Optimization (DPO) \cite{wallace2024diffusion} and Step-by-Step Preference Optimization (SPO) \cite{liang2025aesthetic}. As shown in Table \ref{tab:dpo_orthogonality}, SES yields significant and consistent performance improvements even on strongly aligned models. This finding substantiates that the low-frequency search manifold remains robust post-alignment, confirming SES's orthogonality to training paradigms like Reinforcement Learning from Human Feedback (RLHF).

\textbf{Generalization across Rewards and Samplers.} As a gradient-free method, SES is inherently compatible with arbitrary black-box rewards, including non-differentiable Vision-Language Models (VLMs) (Appendix \ref{app:vlm_experiments}). Furthermore, since SES optimizes the initial noise rather than the denoising trajectory, it is \textit{sampler-agnostic}. Results in Appendix \ref{app:sampler_robustness} demonstrate robust performance across both deterministic ODE solvers and stochastic SDE samplers.

\begin{table}[t]
	\centering
	\caption{Orthogonality to training-time alignment. Even with strong initial alignment from DPO and SPO, SES achieves consistent gains across all metrics. Best results are highlighted in \textbf{bold}.}
	\label{tab:dpo_orthogonality}
	
	\resizebox{0.48\textwidth}{!}{
		\begin{tabular}{lccccc}
			\toprule

			\textbf{Method} & \textbf{CLIP} & \textbf{PickScore} & \textbf{HPS} & \textbf{Aes.} & \textbf{ImgRew.} \\
			\midrule
			SDXL & 33.05 & 21.76 & 27.64 & 5.48 & 0.33 \\
			\midrule
			+ DPO & 36.77 & 22.48 & 29.02 & 5.59 & 0.77 \\
			+ DPO \& SES & \textbf{43.67} & \textbf{23.94} & \textbf{31.63} & \textbf{6.40} & \textbf{1.61} \\
            \midrule
			+ SPO & 33.12 & 22.59 & 28.69 & 5.90 & 0.35 \\
			+ SPO \& SES & \textbf{41.18} & \textbf{24.35} & \textbf{31.35} & \textbf{6.78} & \textbf{1.65} \\
			\bottomrule
		\end{tabular}
	}
	% \vspace{-3mm} 
\end{table}

\section{Conclusion}
\label{sec:conclusion}

In this paper, we present Spectral Evolution Search (SES), a model-agnostic inference-time scaling framework that optimizes the initial noise by confining the search space via wavelet-based spectral decoupling. Driven by the insight that generative models possess a strong spectral bias towards low frequencies, SES restricts the search to a compact low-frequency manifold and employs the Cross-Entropy Method (CEM) to optimize distribution parameters, thereby effectively mitigating the curse of dimensionality. We substantiate this design theoretically by deriving the Spectral Scaling Prediction from perturbation propagation dynamics and empirically validate SES's superior scaling effectiveness across diverse architectures and alignment objectives.

\textbf{Limitations.} Common to inference-time search strategies, SES's performance is bound by the quality and cost of the reward model. Theoretically, our derivation assumes local linearization and approximate spectral decoupling, neglecting higher-order interactions. Practically, while NRE serves as a hardware-agnostic metric, it may not linearly translate to wall-clock latency.

\section*{Impact Statement}
This paper presents work whose goal is to advance the field of Machine
Learning. There are many potential societal consequences of our work, none
which we feel must be specifically highlighted here.

\bibliography{ref}
\bibliographystyle{icml2026}

%%%%%%%%%%%%%%%%%%%%%%%%%%%%%%%%%%%%%%%%%%%%%%%%%%%%%%%%%%%%%%%%%%%%%%%%%%%%%%%
%%%%%%%%%%%%%%%%%%%%%%%%%%%%%%%%%%%%%%%%%%%%%%%%%%%%%%%%%%%%%%%%%%%%%%%%%%%%%%%
% APPENDIX
%%%%%%%%%%%%%%%%%%%%%%%%%%%%%%%%%%%%%%%%%%%%%%%%%%%%%%%%%%%%%%%%%%%%%%%%%%%%%%%
%%%%%%%%%%%%%%%%%%%%%%%%%%%%%%%%%%%%%%%%%%%%%%%%%%%%%%%%%%%%%%%%%%%%%%%%%%%%%%%
\newpage
\appendix
\onecolumn

% \DoToC
\section*{Appendix Contents}

\begin{itemize}
    \setlength{\itemsep}{0.5em}
    
    \item \textbf{Appendix A: PROOFS FOR THEORETICAL ANALYSIS} \dotfill Page \pageref{app:A}
        \begin{itemize}
            \setlength{\itemsep}{0.2em}
            \item[--] A.1 Defining the Difference Calculation Objective \dotfill Page \pageref{sec:A1}
            \item[--] A.2 Derivation of the Jacobian of the Velocity Field \dotfill Page \pageref{sec:A2}
            \item[--] A.3 Solution of Spectral Evolution Dynamics \dotfill Page \pageref{sec:A3}
            \item[--] A.4 Calculation of Effective Spectral Response \dotfill Page \pageref{sec:A4}
            \item[--] A.5 Derivation of the Spectral Scaling Prediction \dotfill Page \pageref{sec:A5}
            \item[--] A.6 Spectral Statistics of Natural Images and Latent Spaces \dotfill Page \pageref{app:spectral_statistics}
        \end{itemize}

    \item \textbf{Appendix B: Algorithmic Implementation and Evolutionary Dynamics} \dotfill Page \pageref{app:implementation_dynamics}
        \begin{itemize}
            \setlength{\itemsep}{0.2em}
            \item[--] B.1 Implementation Specification and Complexity \dotfill Page \pageref{sec:complexity_analysis}
            \item[--] B.2 Analysis of Evolutionary Dynamics \dotfill Page \pageref{sec:Analysis_of_Evolutionary_Dynamics}
        \end{itemize}
    
    \item \textbf{Appendix C: Implementation Details of Experimental Settings} \dotfill Page \pageref{sec:setting}
        \begin{itemize}
            \setlength{\itemsep}{0.2em}
            \item[--] C.1 Benchmarks and Reward Models \dotfill Page \pageref{sec:benchmarks}
            \item[--] C.2 Baselines Implementation and NRE Calculation \dotfill Page \pageref{sec:nre}
        \end{itemize}

    \item \textbf{Appendix D: Additional Experiments} \dotfill Page \pageref{sec:additional_exp}
        \begin{itemize}
            \setlength{\itemsep}{0.2em}
            \item[--] D.1 Baseline Comparison \dotfill Page \pageref{sec:baseline_comparison}
            \item[--] D.2 Analysis with Proxy Rewards \dotfill Page \pageref{app:proxy_reward}
            \item[--] D.3 Experiments on Different Samplers \dotfill Page \pageref{app:sampler_robustness}
            \item[--] D.4 VLM-based Non-Differentiable Objective Experiments \dotfill Page \pageref{app:vlm_experiments}
        \end{itemize}

    \item \textbf{Appendix E: Analysis of Reward Hacking} \dotfill Page \pageref{app:reward_hacking}
        \begin{itemize}
            \setlength{\itemsep}{0.2em}
            \item[--] E.1 Out-of-Distribution Reward Hacking Problem \dotfill Page \pageref{sec:ood}
            \item[--] E.2 Cross-Reward Evaluation of SES \dotfill Page \pageref{sec:cross-reward}
        \end{itemize}

    \item \textbf{Appendix F: Additional Qualitative Results} \dotfill Page \pageref{sec:Additional_Qualitative_Results}
        
\end{itemize}

\section{Proofs For Theoretical Analysis}
\label{app:A}
\subsection{Defining the Difference Calculation Objective}
\label{sec:A1}

The baseline point is the initial noise $\mathbf{x}_0$. We add a perturbation $\boldsymbol{\xi}_0$ to obtain the perturbed initial noise $\tilde{\mathbf{x}}_0 = \mathbf{x}_0 + \boldsymbol{\xi}_0$. We aim to determine how this perturbation affects the final generated image, which entails calculating the difference between the generated data corresponding to the two initial noise inputs: $\boldsymbol{\xi}_1 = \tilde{\mathbf{x}}_1 - \mathbf{x}_1$.

Both $\mathbf{x}_t$ and $\tilde{\mathbf{x}}_t$ are solutions to $\mathrm{d}\mathbf{x}_t = v_\theta(\mathbf{x}_t, t)\mathrm{d}t$, satisfying the integral equations:
\begin{align}
\mathbf{x}_t &= \mathbf{x}_0 + \int_0^t v_\theta(\mathbf{x}_\tau, \tau) \, \mathrm{d}\tau \\
\tilde{\mathbf{x}}_t &= (\mathbf{x}_0 + \boldsymbol{\xi}_0) + \int_0^t v_\theta(\tilde{\mathbf{x}}_\tau, \tau) \, \mathrm{d}\tau.
\end{align}
Directly calculating $\boldsymbol{\xi}_1$ requires evaluating:
\begin{equation}
\boldsymbol{\xi}_1 = \tilde{\mathbf{x}}_1 - \mathbf{x}_1 = \boldsymbol{\xi}_0 + \int_0^1 \left( v_\theta(\tilde{\mathbf{x}}_\tau, \tau) - v_\theta(\mathbf{x}_\tau, \tau) \right) \, \mathrm{d}\tau.
\end{equation}
Here, the term $v_\theta(\tilde{\mathbf{x}}_\tau,\tau) - v_\theta(\mathbf{x}_\tau,\tau)$ depends on the unknown trajectories $\tilde{\mathbf{x}}_\tau$ and $\mathbf{x}_\tau$. Since $v_\theta$ is a highly non-linear function, it is infeasible to directly derive an explicit expression for $\boldsymbol{\xi}_t$.

To address this, we resort to calculating the rate of change of the difference, $\frac{\mathrm{d}\boldsymbol{\xi}_t}{\mathrm{d}t}$. If this value is positive, the difference diverges at time $t$; otherwise, it converges.
\begin{equation}
\frac{\mathrm{d}\boldsymbol{\xi}_t}{\mathrm{d}t} = \frac{\mathrm{d}(\mathbf{x}_t + \boldsymbol{\xi}_t)}{\mathrm{d}t} - \frac{\mathrm{d}\mathbf{x}_t}{\mathrm{d}t} = v_\theta(\mathbf{x}_t + \boldsymbol{\xi}_t, t) - v_\theta(\mathbf{x}_t, t).
\end{equation}
Performing a first-order Taylor expansion on $v_\theta(\mathbf{x}_t + \boldsymbol{\xi}_t, t)$, we obtain:
\begin{equation}
v_\theta(\mathbf{x}_t + \boldsymbol{\xi}_t, t) \approx v_{\theta}(\mathbf{x}_t, t) + \nabla_{\mathbf{x}} v_\theta(\mathbf{x}_t, t) \cdot \boldsymbol{\xi}_t + O(||\boldsymbol{\xi}_t||^2).
\end{equation}
Neglecting higher-order terms, the evolution is governed by:
\begin{equation}
\frac{\mathrm{d}\boldsymbol{\xi}_t}{\mathrm{d}t} \approx \nabla_{\mathbf{x}} v_\theta(\mathbf{x}_t, t) \cdot \boldsymbol{\xi}_t = \mathbf{J}_v(\mathbf{x}_t, t)\boldsymbol{\xi}_t,
\end{equation}
where $\mathbf{J}_v(\mathbf{x}_t, t)$ is the instantaneous Jacobian matrix of the velocity field.

\subsection{Derivation of the Jacobian of the Velocity Field}
\label{sec:A2}

In this section, we derive the specific form of $\mathbf{J}_v$. In mainstream diffusion models and flow-based models, the forward process satisfies the following trajectory interpolation formula:
\begin{equation}
\mathbf{x}_t = \alpha(t)\mathbf{x}_1 + \sigma(t)\mathbf{x}_0, \quad t \in [0, 1],
\end{equation}
where $\alpha(t)$ and $\sigma(t)$ are time-dependent scheduling functions. Differentiating both sides with respect to time $t$ yields the ideal flow field:
\begin{equation}
\frac{\mathrm{d}\mathbf{x}_t}{\mathrm{d}t} = \dot{\alpha}(t)\mathbf{x}_1 + \dot{\sigma}(t)\mathbf{x}_0.
\end{equation}
During inference, the model employs a neural network $\hat{\mathbf{x}}_\theta(\mathbf{x}_t, t)$ to estimate $\mathbf{x}_1$, allowing us to solve for the estimate of $\mathbf{x}_0$:
\begin{equation}
\mathbf{x}_0 = \frac{\mathbf{x}_t - \alpha(t)\hat{\mathbf{x}}_\theta(\mathbf{x}_t,t)}{\sigma(t)}.
\end{equation}
Substituting this estimate of $\mathbf{x}_0$ into the derivative formula and replacing $\mathbf{x}_1$ with $\hat{\mathbf{x}}_\theta$, we obtain the parameterized velocity field $v_\theta$:
\begin{equation}
\begin{aligned}
v_\theta(\mathbf{x}_t, t) &= \dot{\alpha}(t)\hat{\mathbf{x}}_\theta(\mathbf{x}_t,t) + \dot{\sigma}(t) \left( \frac{\mathbf{x}_t - \alpha(t)\hat{\mathbf{x}}_\theta(\mathbf{x}_t,t)}{\sigma(t)} \right) \\
&= \dot{\alpha}(t)\hat{\mathbf{x}}_\theta(\mathbf{x}_t,t) + \frac{\dot{\sigma}(t)}{\sigma(t)}\mathbf{x}_t - \frac{\dot{\sigma}(t)\alpha(t)}{\sigma(t)}\hat{\mathbf{x}}_\theta(\mathbf{x}_t,t) \\
&= \left( \dot{\alpha}(t) - \frac{\dot{\sigma}(t)\alpha(t)}{\sigma(t)} \right) \hat{\mathbf{x}}_\theta(\mathbf{x}_t,t) + \frac{\dot{\sigma}(t)}{\sigma(t)} \mathbf{x}_t.
\end{aligned}
\end{equation}
We define two time-varying coefficients $\mu(t)$ and $\nu(t)$:
\begin{equation}
\mu(t) \triangleq \dot{\alpha}(t) - \frac{\dot{\sigma}(t)\alpha(t)}{\sigma(t)}, \quad \nu(t) \triangleq \frac{\dot{\sigma}(t)}{\sigma(t)}.
\end{equation}
Thus, the parameterized velocity field simplifies to:
\begin{equation}
v_\theta(\mathbf{x}_t, t) = \mu(t) \hat{\mathbf{x}}_\theta(\mathbf{x}_t,t) + \nu(t) \mathbf{x}_t.
\end{equation}
Consequently, the specific form of the Jacobian matrix of the velocity field is:
\begin{equation}
\begin{aligned}
\mathbf{J}_v(\mathbf{x}_t, t) &= \nabla_{\mathbf{x}} \left( \mu(t) \hat{\mathbf{x}}_\theta(\mathbf{x}_t,t) + \nu(t) \mathbf{x}_t \right) \\
&= \mu(t) \nabla_{\mathbf{x}} \hat{\mathbf{x}}_\theta(\mathbf{x}_t,t) + \nu(t) \nabla_{\mathbf{x}} \mathbf{x}_t \\
&= \mu(t) \mathbf{H}_{\hat{x}} + \nu(t) \mathbf{I},
\end{aligned}
\end{equation}
where $\mathbf{H}_{\hat{x}} \triangleq \nabla_{\mathbf{x}} \hat{\mathbf{x}}_\theta(\mathbf{x}_t,t)$ is the input-output Jacobian matrix of the neural network, and $\mathbf{I}$ is the identity matrix. The specific form governing the rate of error change is:
\begin{equation}
\frac{\mathrm{d}\boldsymbol{\xi}_t}{\mathrm{d}t} = \mathbf{J}_v(\mathbf{x}_t, t) \boldsymbol{\xi}_t = (\mu(t) \mathbf{H}_{\hat{x}} + \nu(t) \mathbf{I})\boldsymbol{\xi}_t.
\end{equation}

\subsection{Solution of Spectral Evolution Dynamics}
\label{sec:A3}

In this section, we detail how to decouple the high-dimensional variational equation derived in Appendix \ref{sec:A2} using frequency domain analysis and solve for the cumulative gain $G(\omega)$ of the initial perturbation.

Recall the linearized variational equation:
\begin{equation}
\frac{\mathrm{d}\boldsymbol{\xi}_t}{\mathrm{d}t} = \mu(t) \mathbf{H}_{\hat{x}} \boldsymbol{\xi}_t + \nu(t) \mathbf{I}\boldsymbol{\xi}_t.
\end{equation}
To decouple the dependencies across spatial dimensions, we apply the spatial Fourier transform operator $\mathcal{F}$ to both sides. Let $\tilde{\xi}_t(\omega) \triangleq \mathcal{F}\{\boldsymbol{\xi}_t\}(\omega)$ denote the spectral component of the error vector $\boldsymbol{\xi}_t$ at frequency $\omega$.

Using the linearity of the Fourier transform, we process each term:

Left-hand side: Since the Fourier transform acts as an integral over spatial coordinates and the time derivative acts on the time coordinate, under appropriate smoothness conditions, the order of operations can be swapped:
\begin{equation}
\mathcal{F}\left\{ \frac{\mathrm{d}\boldsymbol{\xi}_t}{\mathrm{d}t} \right\} = \frac{\mathrm{d}\tilde{\xi}_t(\omega)}{\mathrm{d}t}.
\end{equation}

Right-hand side, first term: Based on the frequency domain decoupling assumption, the denoising network Jacobian $\mathbf{H}_{\hat{x}}$ is approximated as a diagonal operator in the frequency domain. This implies its action in the frequency domain is equivalent to scalar multiplication. We define the effective spectral response $h(\omega,t)$:
\begin{equation}
\mathcal{F}\{ \mathbf{H}_{\hat{x}} \boldsymbol{\xi} \}(\omega) \approx h(\omega,t) \cdot \mathcal{F}\{ \boldsymbol{\xi} \}(\omega).
\end{equation}

Right-hand side, second term: Since the identity matrix $\mathbf{I}$ remains the identity operator in the frequency domain, and the coefficient $\nu(t)$ depends only on time:
\begin{equation}
\mathcal{F}\{ \nu(t) \mathbf{I}\boldsymbol{\xi}_t \}(\omega) = \nu(t) \cdot \tilde{\xi}_t(\omega).
\end{equation}

Accordingly, the complex high-dimensional matrix differential equation decouples into independent scalar Ordinary Differential Equations (ODEs) for each frequency $\omega$:
\begin{equation}
\frac{\mathrm{d}\tilde{\xi}_t(\omega)}{\mathrm{d}t} = \left( \mu(t) h(\omega,t) + \nu(t) \right) \tilde{\xi}_t(\omega).
\end{equation}
To simplify notation, we define the instantaneous spectral eigenvalue $\lambda(\omega, t)$ as:
\begin{equation}
\lambda(\omega, t) \triangleq \mu(t) h(\omega,t) + \nu(t).
\end{equation}
Next, we solve this ODE using the separation of variables:
\begin{equation}
\frac{\mathrm{d}\tilde{\xi}_t}{\tilde{\xi}_t} = \lambda(\omega, t) \mathrm{d}t.
\end{equation}
Integrating both sides over the time interval $t \in [0, 1]$:
\begin{equation}
\int_{\tilde{\xi}_0}^{\tilde{\xi}_1} \frac{\mathrm{d}\tilde{\xi}}{\tilde{\xi}} = \int_0^1 \lambda(\omega, \tau) \mathrm{d}\tau \implies \ln \left( \frac{\tilde{\xi}_1(\omega)}{\tilde{\xi}_0(\omega)} \right) = \int_0^1 \lambda(\omega, \tau) \mathrm{d}\tau.
\end{equation}
Exponentiating yields the cumulative gain $G(\omega)$ at frequency $\omega$:
\begin{equation}
G(\omega) \triangleq \frac{\|\tilde{\xi}_1(\omega)\|}{\|\tilde{\xi}_0(\omega)\|} = \exp\left( \int_0^1 \left( \mu(\tau) h(\omega,t) + \nu(\tau) \right) \mathrm{d}\tau \right).
\end{equation}

\subsection{Calculation of Effective Spectral Response}
\label{sec:A4}

First, we calculate the Signal-to-Noise Ratio (SNR) of the latent variables $\mathbf{x}_t$.

\textbf{Power Spectrum Decomposition of Latent Variables.}
In flow-based generative models, the generative process operates within a low-dimensional latent space. The latent variable $\mathbf{x}_t$ at any arbitrary time $t \in [0, 1]$ is formed by the linear superposition of the clean data latent code $\mathbf{x}_1$ and Gaussian noise $\mathbf{x}_0$:
\begin{equation}
    \mathbf{x}_t = \alpha(t)\mathbf{x}_1 + \sigma(t)\mathbf{x}_0,
\end{equation}

where $\mathbf{x}_0 \sim \mathcal{N}(0, \mathbf{I})$. Assuming statistical independence between the latent code distribution and the initial noise distribution, and in accordance with the theory of stochastic processes, the Power Spectral Density (PSD) of the mixed signal is the weighted sum of the component power spectra:
\begin{equation}
    P_{\mathbf{x}_t}(\omega) = \alpha^2(t) P_{data}(\omega) + \sigma^2(t) P_{noise}(\omega).
\end{equation}

\textbf{Spectral Statistical Properties of Signal and Noise.}
\textit{Latent Code $\mathbf{x}_1$:} Raw natural images exhibit significant spatial redundancy and strong correlation; their power spectra typically adhere to a decay law where $\beta \approx 2$ [2,3]. The latent code $\mathbf{x}_1$ is obtained via an encoder $\mathcal{E}$: $\mathbf{x}_1 = \mathcal{E}(\text{Image})$. The downsampling operations and KL regularization within the encoder effectuate a form of ``spectral whitening'': this process compresses high-frequency redundancies while constraining the latent distribution towards a standard normal distribution. Consequently, the spectral characteristics of the latent code $\mathbf{x}_1$ lie between those of highly correlated natural images and completely independent white noise. We assume its radial power spectrum adheres to a generalized power-law distribution:
\begin{equation}
    P_{data}(\omega) = \mathbb{E}[|\hat{\mathbf{x}}_1(\omega)|^2] \propto \|\omega\|^{-\beta},
\end{equation}

where $\|\omega\| = \sqrt{\omega_h^2 + \omega_w^2}$ represents the radial spatial frequency. Due to the decorrelation effects of the latent space, the decay exponent $\beta$ typically satisfies $0 < \beta < 2$.

\textit{Gaussian Noise $\mathbf{x}_0$:} Standard Gaussian noise $\mathbf{x}_0$ retains independent and identically distributed (i.i.d.) characteristics in the discrete Fourier basis. Invoking Parseval's theorem, its power spectrum is flat across the entire frequency band:
\begin{equation}
    P_{noise}(\omega) = \mathbb{E}[|\hat{\mathbf{x}}_0(\omega)|^2] = 1.
\end{equation}

\textbf{Derivation of Local SNR.}
Assuming statistical independence between the signal and noise, the local SNR in the frequency domain is defined as the ratio of signal power to noise power at a given frequency. Substituting the aforementioned spectral properties into the definition:
\begin{equation}
    \text{SNR}(\omega, t) \triangleq \frac{\text{Signal Power at } \omega}{\text{Noise Power at } \omega} = \frac{\alpha^2(t) P_{data}(\omega)}{\sigma^2(t) P_{noise}(\omega)}.
\end{equation}

Substituting $P_{data}(\omega) \propto \|\omega\|^{-\beta}$ and $P_{noise}(\omega) = 1$, we obtain:
\begin{equation}
    \text{SNR}(\omega, t) \triangleq \frac{\alpha^2(t)}{\sigma^2(t)} \cdot \frac{1}{\|\omega\|^\beta}.
\end{equation}

Then, we derive the effective spectral response $h(\omega,t)$ in detail.

Let $\mathcal{F}$ denote the Fourier transform operator, and define $\tilde{\xi}(\omega) \triangleq \mathcal{F}\{\mathbf{x}\}(\omega)$ as the spectral component of vector $\mathbf{x}$ at frequency $\omega$. $\tilde{\xi}_t$, $\tilde{\xi}_1$, and $\tilde{\xi}_0$ correspond to the frequency domain representations of the latent variable $\mathbf{x}_t$, data $\mathbf{x}_1$, and noise $\mathbf{x}_0$, respectively.

Due to the frequency domain decoupling assumption, we can analyze each frequency component independently. At frequency $\omega$, the local behavior of the network $\hat{\mathbf{x}}_\theta$ can be approximated as a linear scaling operation:
\begin{equation}
\mathcal{F}\{ \hat{\mathbf{x}}_\theta(\mathbf{x}_t) \}(\omega) \approx h(\omega,t) \cdot \tilde{\xi}_t(\omega).
\end{equation}
The model's objective is to find the optimal scalar gain $h$ that minimizes the mean squared error of the network output at that frequency:
\begin{equation}
\min_{h} \mathcal{L}(h) = \mathbb{E} \left[ \| h \cdot \tilde{\xi}_t(\omega) - \tilde{\xi}_1(\omega) \|^2 \right].
\end{equation}
Using the linearity of the Fourier transform, transforming $\mathbf{x}_t = \alpha(t) \mathbf{x}_1 + \sigma(t) \mathbf{x}_0$ to the frequency domain yields:
\begin{equation}
\tilde{\xi}_t(\omega) = \alpha(t) \tilde{\xi}_1(\omega) + \sigma(t) \tilde{\xi}_0(\omega).
\end{equation}
Substituting this expression into the loss function and exploiting the statistical independence of $\mathbf{x}_1$ and $\mathbf{x}_0$ (cross-term expectations are 0), we expand the squared term:
\begin{equation}
\begin{aligned}
\mathcal{L}(h) &= \mathbb{E} \left[ \| h(\alpha(t) \tilde{\xi}_1(\omega) + \sigma(t) \tilde{\xi}_0(\omega)) - \tilde{\xi}_1(\omega) \|^2 \right] \\
&= \mathbb{E} \left[ \| (h\alpha(t) - 1)\tilde{\xi}_1(\omega) + h\sigma(t) \tilde{\xi}_0(\omega) \|^2 \right] \\
&= (h\alpha(t) - 1)^2 \underbrace{\mathbb{E}[\|\tilde{\xi}_1(\omega)\|^2]}_{P_{\text{signal}}(\omega)} + h^2\sigma(t)^2 \underbrace{\mathbb{E}[\|\tilde{\xi}_0(\omega)\|^2]}_{P_{\text{noise}}(\omega)},
\end{aligned}
\end{equation}
where $P_{\text{signal}}(\omega)$ and $P_{\text{noise}}(\omega)$ represent the power spectral densities of the natural image and Gaussian noise at frequency $\omega$, respectively. For standard Gaussian noise, typically $P_{\text{noise}} = 1$.

To minimize, we set the derivative with respect to $h$ to 0:
\begin{equation}
\frac{\mathrm{d}\mathcal{L}}{\mathrm{d}h} = 2(h\alpha(t) - 1)\alpha(t) P_{\text{signal}}(\omega) + 2h\sigma(t)^2 P_{\text{noise}}(\omega) = 0.
\end{equation}
Simplifying and solving for $h$:
\begin{equation}
h = \frac{\alpha(t) P_{\text{signal}}(\omega)}{\alpha(t)^2 P_{\text{signal}}(\omega) + \sigma(t)^2 P_{\text{noise}}(\omega)}.
\end{equation}
We define the SNR at frequency $\omega$ as:
\begin{equation}
\text{SNR}(\omega) = \frac{\alpha(t)^2 P_{\text{signal}}(\omega)}{\sigma(t)^2 P_{\text{noise}}(\omega)}.
\end{equation}
Substituting this back, we obtain the expression for the effective spectral response $h(\omega,t)$:
\begin{equation}
h(\omega,t) = \frac{1}{\alpha(t)} \cdot \frac{\alpha(t)^2 P_{\text{signal}}}{\alpha(t)^2 P_{\text{signal}} + \sigma(t)^2 P_{\text{noise}}} = \frac{1}{\alpha(t)} \cdot \frac{\text{SNR}(\omega)}{\text{SNR}(\omega) + 1}.
\end{equation}

\subsection{Derivation of the Spectral Scaling Prediction}
\label{sec:A5}

In this section, we employ matched asymptotic analysis to determine the cumulative gain $G(\omega)$.

\textbf{Algebraic Rearrangement of the Instantaneous Growth Rate.}
Recall the eigenvalue equation $\lambda(\omega, t) = \mu(t) h(\omega, t) + \nu(t)$, where $\mu = \dot{\alpha} - \frac{\dot{\sigma}\alpha}{\sigma}$ and $\nu = \frac{\dot{\sigma}}{\sigma}$.
Substituting the optimal spectral response $h(\omega, t) = \frac{1}{\alpha(t)} \frac{\text{SNR}(\omega, t)}{\text{SNR}(\omega, t) + 1}$ derived in Appendix \ref{sec:A4} into the equation yields:

\begin{align}
    \lambda(\omega, t) &= \left( \dot{\alpha} - \frac{\dot{\sigma}\alpha}{\sigma} \right) \frac{1}{\alpha} \mathcal{W} + \frac{\dot{\sigma}}{\sigma} \nonumber \\
    &= \frac{\dot{\alpha}}{\alpha} \mathcal{W} - \frac{\dot{\sigma}}{\sigma} \mathcal{W} + \frac{\dot{\sigma}}{\sigma} \nonumber \\
    &= \frac{\dot{\sigma}}{\sigma} + \left( \frac{\dot{\alpha}}{\alpha} - \frac{\dot{\sigma}}{\sigma} \right) \mathcal{W}(\omega, t),
\end{align}

where the spectral weight function is defined as $\mathcal{W}(\omega, t) = \frac{\text{SNR}(\omega, t)}{\text{SNR}(\omega, t) + 1}$.
Given that the generation process typically satisfies $\dot{\alpha} \ge 0$ and $\dot{\sigma} \le 0$, the net expansion term within the brackets, $\left( {\dot{\alpha}}/{\alpha} - {\dot{\sigma}}/{\sigma} \right)$, remains strictly positive. This implies a distinct physical interpretation:

\begin{itemize}
    \item As $\mathcal{W} \to 0$ (Noise-Dominated Regime), the error evolves according to the contraction rate of the noise $\sigma(t)$.
    \item As $\mathcal{W} \to 1$ (Signal-Dominated Regime), the error evolves according to the variation rate of the signal $\alpha(t)$.
\end{itemize}

\textbf{Frequency-Dependent Critical Time.}
Due to the decay characteristic of the power spectrum $P_{data}(\omega) \propto \|\omega\|^{-\beta}$, different frequency components are ``resolved'' (i.e., the SNR surpasses a threshold) at distinct time points. We define the critical time $t_\omega$ as the moment when the SNR reaches a unit threshold:
\begin{equation}
    \text{SNR}(\omega, t_\omega) = \frac{\alpha^2(t_\omega)}{\sigma^2(t_\omega)} P_{data}(\omega) = 1.
\end{equation}

Consequently, at this critical moment, the SNR coefficient satisfies:$\frac{\sigma(t_\omega)}{\alpha(t_\omega)} = \sqrt{P_{data}(\omega)} \propto \|\omega\|^{-\beta/2}$.

\textbf{Piecewise Integral Approximation.}
The effective spectral response function $\mathcal{W}(\omega, t) = \frac{\text{SNR}}{\text{SNR} + 1}$ exhibits behavior analogous to a step function. We approximate it as:
\begin{equation}
    \mathcal{W}(\omega, t) \approx
    \begin{cases}
    0, & t < t_\omega \quad (\text{High Noise Regime}) \\
    1, & t \ge t_\omega \quad (\text{Signal Regime}).
    \end{cases}
\end{equation}

Based on this approximation, the instantaneous growth rate simplifies to:
\begin{equation}
    \lambda(\omega, t) \approx
    \begin{cases}
    \frac{\dot{\sigma}}{\sigma}, & t < t_\omega \\
    \frac{\dot{\alpha}}{\alpha}, & t \ge t_\omega.
    \end{cases}
\end{equation}

Next, we perform a piecewise integration of $\ln G(\omega) = \int_0^1 \lambda(\omega, t) \mathrm{d}t$:
\begin{align}
    \ln G(\omega) &\approx \int_0^{t_\omega} \frac{\dot{\sigma}}{\sigma} \mathrm{d}t + \int_{t_\omega}^1 \frac{\dot{\alpha}}{\alpha} \mathrm{d}t \nonumber \\
    &= [\ln \sigma(t)]_0^{t_\omega} + [\ln \alpha(t)]_{t_\omega}^1 \nonumber \\
    &= \ln \sigma(t_\omega) - \ln \sigma(0) + \ln \alpha(1) - \ln \alpha(t_\omega) \nonumber \\
    &= \ln \left( \frac{\sigma(t_\omega)}{\alpha(t_\omega)} \right) + \underbrace{\ln \left( \frac{\alpha(1)}{\sigma(0)} \right)}_{C_{const}}.
\end{align}

\textbf{Derivation of the Scaling Prediction.}
Substituting the relationship $\frac{\sigma(t_\omega)}{\alpha(t_\omega)} = \sqrt{P_{data}(\omega)}$ into the integration result:
\begin{equation}
    \ln G(\omega) \approx \ln \left( \sqrt{P_{data}(\omega)} \right) + C_{const},
\end{equation}
\begin{equation}
    \ln G(\omega) \approx \ln \left( \|\omega\|^{-\beta/2} \right) + C_{const}.
\end{equation}

Exponentiating both sides yields the spectral scaling prediction:
\begin{equation}
    G(\omega) \propto \|\omega\|^{-\beta/2}.
\end{equation}

\subsection{Spectral Statistics of Natural Images and Latent Spaces}
\label{app:spectral_statistics}

\begin{figure}[t]
  \centering
  \begin{subfigure}[b]{0.28\linewidth}
    \includegraphics[width=\linewidth]{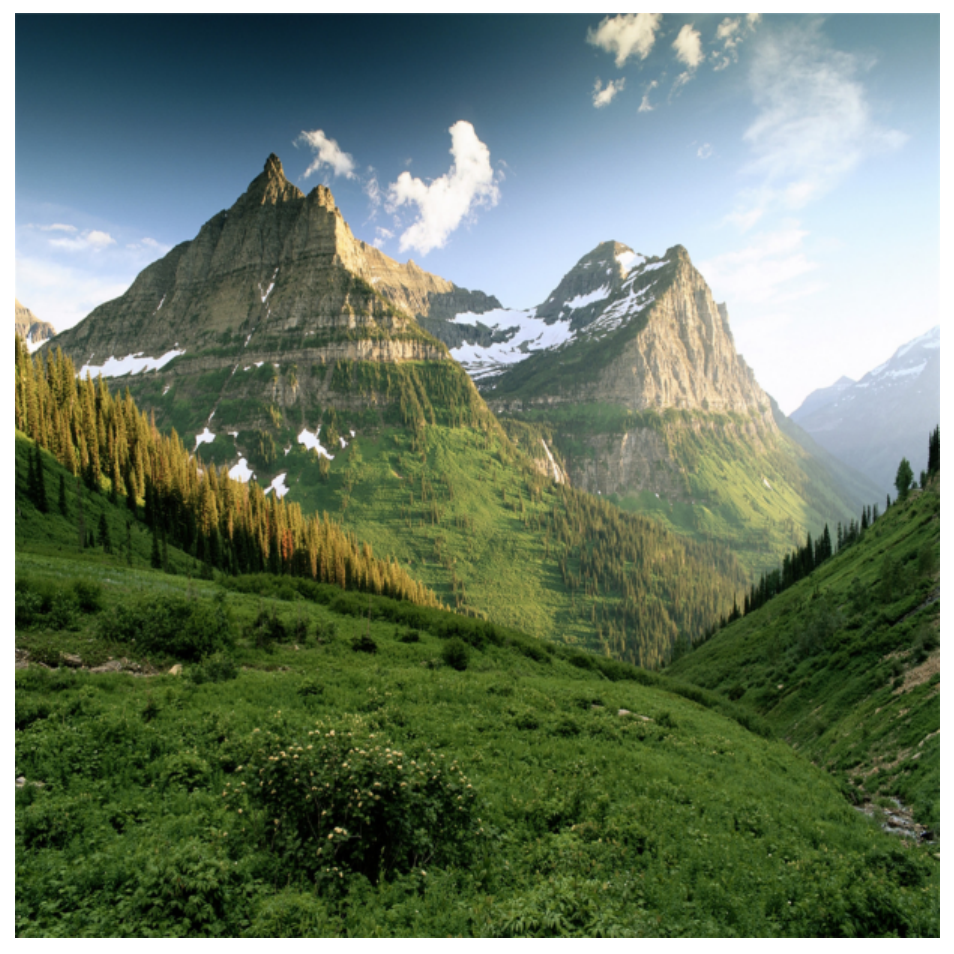}
    \caption{Original Natural Image}
    \label{fig:spatial}
  \end{subfigure}
  % \hfill
  \begin{subfigure}[b]{0.28\linewidth}
    \includegraphics[width=\linewidth]{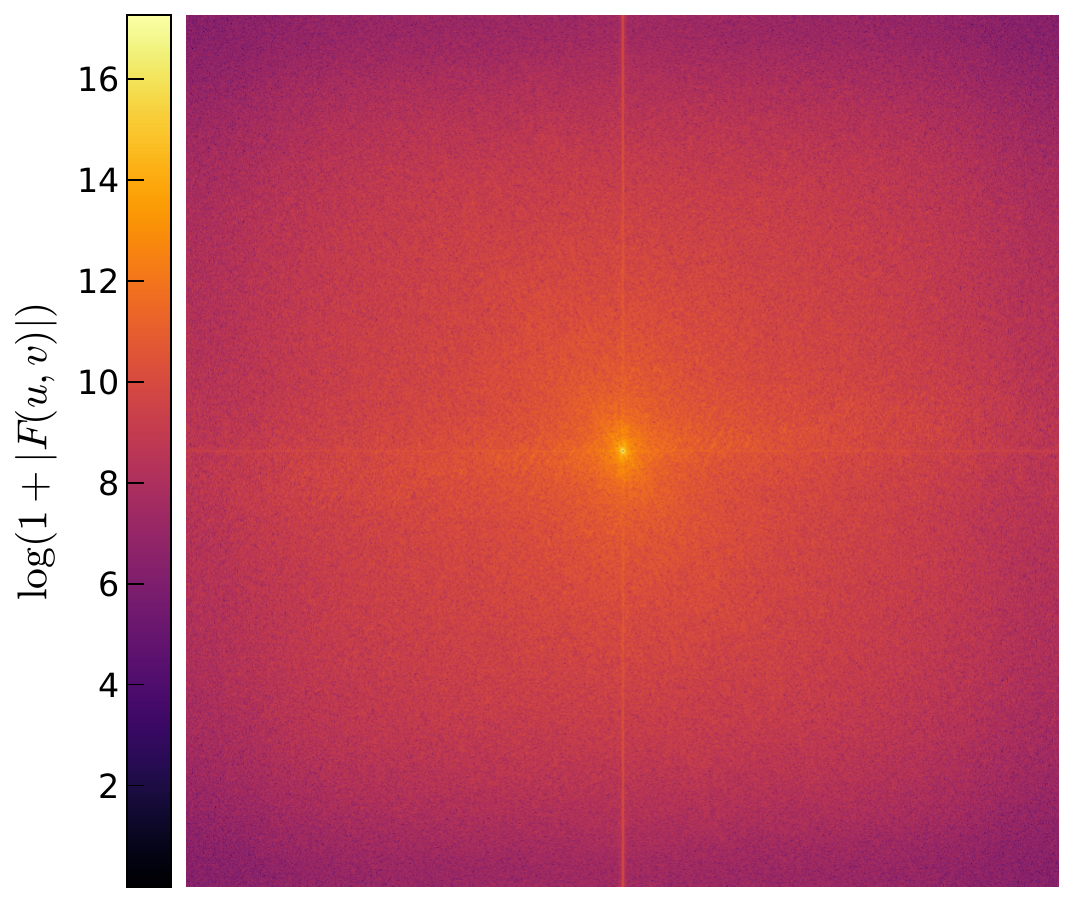}
    \caption{Log Magnitude Spectrum}
    \label{fig:spectrum}
  \end{subfigure}
  % \hfill
  \begin{subfigure}[b]{0.28\linewidth}
    \includegraphics[width=\linewidth]{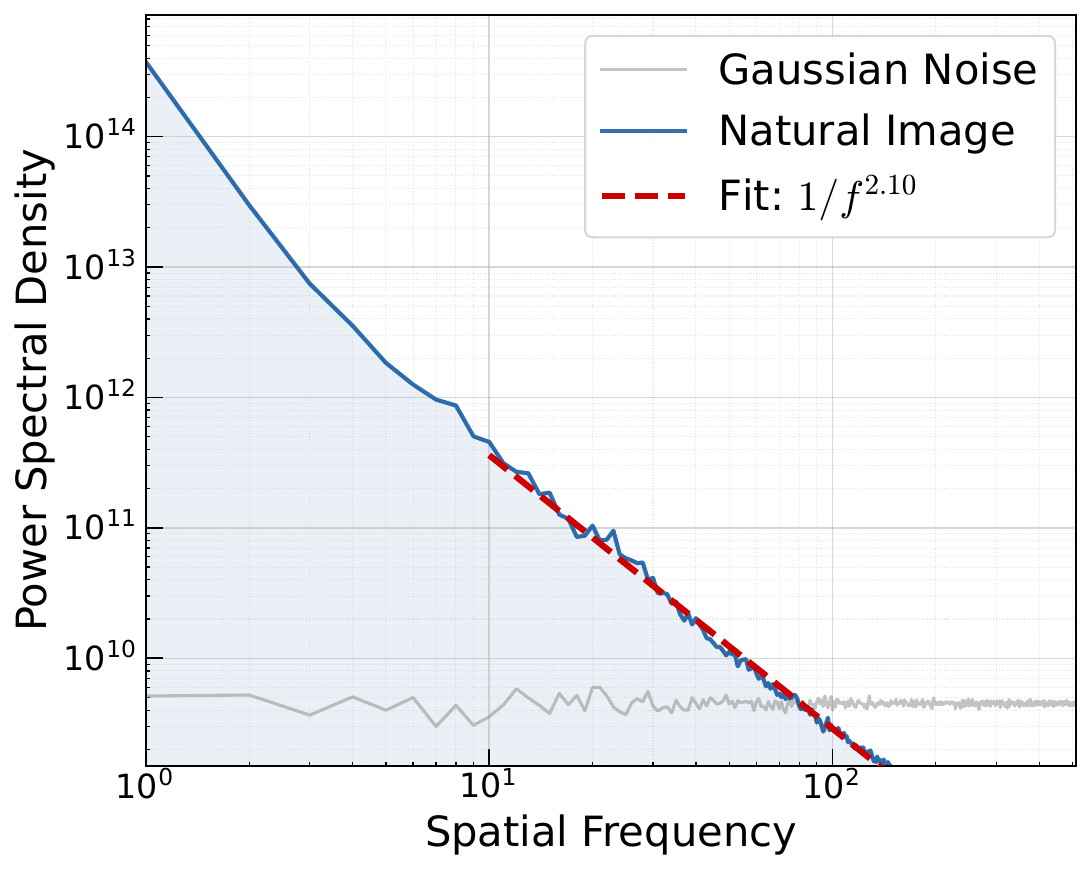}
    \caption{Radial Power Spectrum Profiles}
    \label{fig:profile}
  \end{subfigure}
  \caption{
    Comparison of spectral characteristics between a natural image and Gaussian noise. 
    (a) The original natural image. 
    (b) Visualization of the log magnitude spectrum. 
    (c) Radial power spectrum profiles. The natural image (blue) follows a characteristic $1/f^\alpha$ power-law decay (the red dashed line indicates the linear fit), whereas the Gaussian noise (gray) exhibits a uniform power distribution.
  }
  \label{fig:natural_spectrum}
\end{figure}

\textbf{Spectral Statistics of Natural Images.}
Classical research on natural image statistics \cite{field1987relations} indicates that natural scenes exhibit \textit{Scale Invariance}. Consequently, their power spectra demonstrate a characteristic $1/f^2$ decay, where $\beta \approx 2$.  We conducted experiments on high-resolution natural images, as illustrated in Figure \ref{fig:natural_spectrum}. The results reveal that the energy distribution of natural images aligns closely with the theoretical fit line (Fit: $1/f^{2.05}$), whereas the spectrum of Gaussian noise remains flat.

\textbf{Spectral Statistics of Latent Spaces.}
Although the latent space of a VAE undergoes non-linear mapping and dimensionality reduction, we hypothesize that the latent variables $\mathbf{x}_0$ still adhere to a power-law spectral decay $P(\omega) \propto \|\omega\|^{-\beta}$. Our hypothesis is grounded in two observations. First, the fully convolutional architecture of the VAE encoder preserves the spatial correlations of natural images, thereby inheriting their low-frequency dominant structure. Second, despite the application of KL regularization during training, the reconstruction objective compels the encoder to retain these information-dense low-frequency components to ensure high-fidelity decoding, preventing the latent distribution from collapsing into pure white noise.

To validate this analysis and determine the specific decay exponent $\beta$, we extracted latent representations $\mathbf{z} \in \mathbb{R}^{4 \times H' \times W'}$ encoded by the SDXL VAE for spectral analysis. Given that the latent variables consist of 4 channels, we computed the Power Spectral Density (PSD) for each channel and averaged the results.

\begin{figure}[t]
  \centering
  \begin{subfigure}[b]{0.28\linewidth}
    \includegraphics[width=\linewidth]{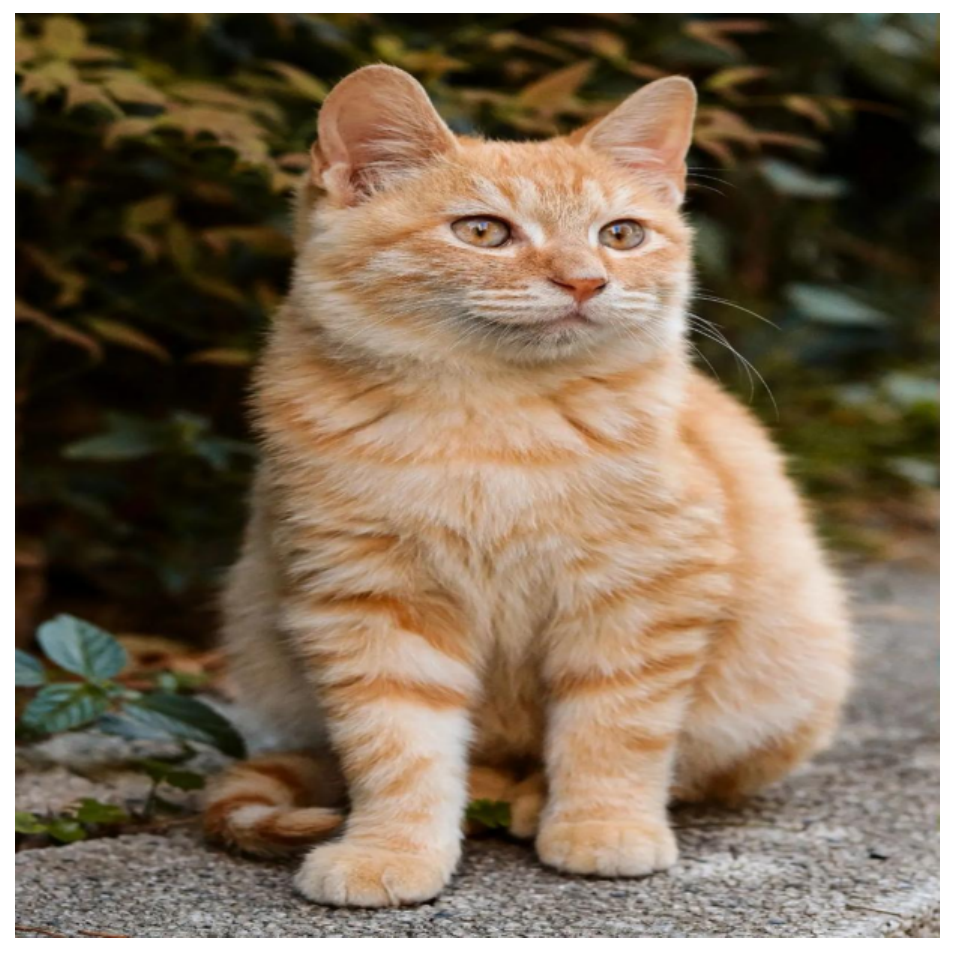}
    \caption{Latent Space}
  \end{subfigure}
  % \hfill
  \begin{subfigure}[b]{0.28\linewidth}
    \includegraphics[width=\linewidth]{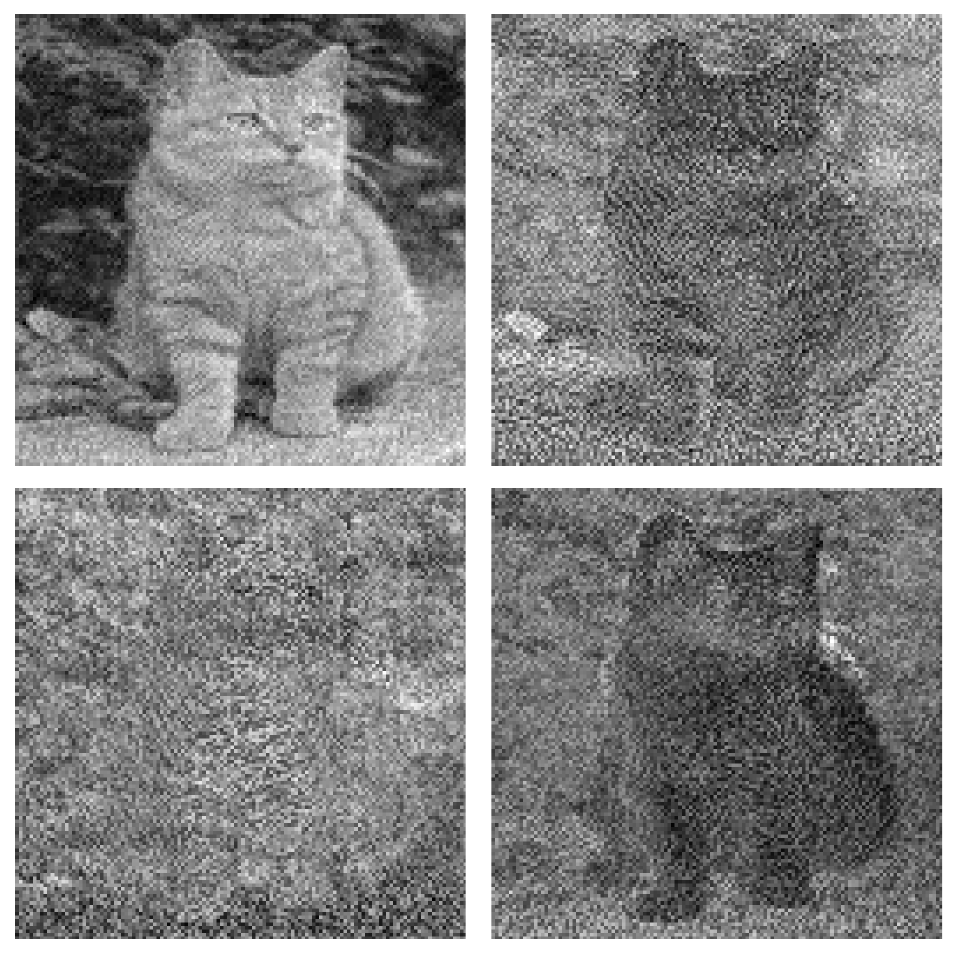}
    \caption{Latent Spectrum}
  \end{subfigure}
  % \hfill
  \begin{subfigure}[b]{0.28\linewidth}
    \includegraphics[width=\linewidth]{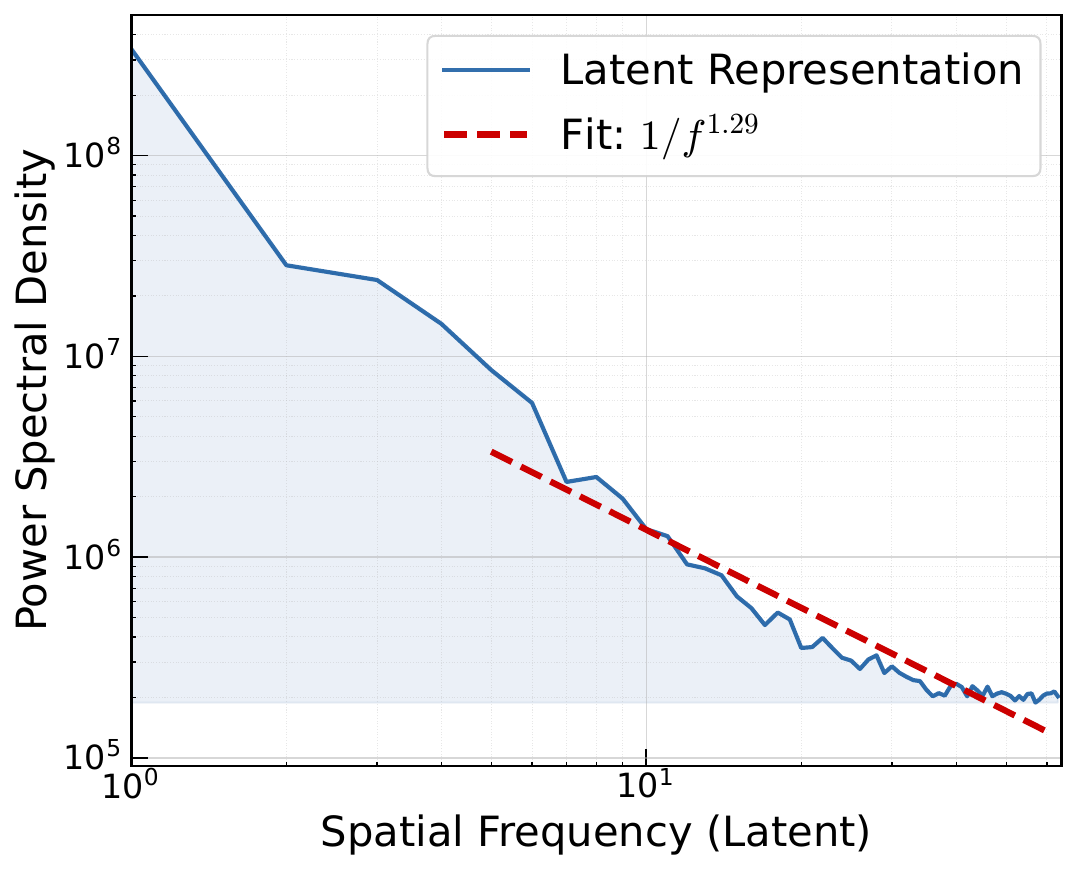}
    \caption{Radial PSD Profile}
  \end{subfigure}
  \caption{
    Spectral Analysis of Latent Codes. Analysis of the latent representation produced by the SDXL VAE encoder. Despite the dimensionality reduction, the latent codes exhibit a consistent power-law spectral decay with an exponent $\beta \approx 1.29$.
  }
  \label{fig:latent_spectrum}
\end{figure}

As shown in Figure \ref{fig:latent_spectrum}, the visualization of latent feature maps clearly exhibits object contours and structures corresponding to the original images, confirming the preservation of spatial correlations.  Furthermore, the radial power spectrum of the latent space exhibits significant linear decay in log-log coordinates. Linear fitting yields $\beta \approx 1.29$. This result aligns with theoretical expectations: while the spectrum of the latent space becomes slightly flatter, the fundamental property of power-law decay remains unaltered.

\section{Algorithmic Implementation and Evolutionary Dynamics}
\label{app:implementation_dynamics}

This section elaborates on the implementation details and hyperparameter configurations of SES, and provides an in-depth analysis of the algorithm's convergence behavior in conjunction with the changes in statistical quantities during the evolutionary process. The complete pseudo-code for SES is presented in Alg. \ref{alg:ses}.

\subsection{Implementation Specification and Complexity}
\label{sec:complexity_analysis}

\textbf{Implementation details.}
In contrast to baseline methods that perform searches in the full-dimensional pixel space, SES significantly reduces the dimensionality of the optimization problem by leveraging the Discrete Wavelet Transform (DWT). Given an input dimension $D = C \times H \times W$, after a $J$-level wavelet decomposition, the dimensionality of the optimization variable $\mathbf{u}$ is reduced to $D' = D / 4^J$. In our implementation, we utilize PyTorch to construct the DWT operator and adopt Daubechies-1 (db1) wavelets by default, owing to their preservation of orthogonality and superior compact support properties. For the evolutionary search hyperparameters, we adopt a default setting of decomposition level $J=4$, population size $N=10$, elite size $K=5$, and a smoothing factor $\gamma=10^{-5}$. All experiments are conducted on NVIDIA A100 GPUs.

Although the dimensionality of the low-frequency subspace $D'$ is substantially reduced, maintaining and updating a full-rank covariance matrix with a limited population size still incurs a storage and computational overhead of $O((D')^2)$. Therefore, we employ a diagonal covariance matrix approximation in our implementation:
\begin{equation}
\boldsymbol{\Sigma}^{(k)} = \operatorname{diag}(\sigma_1^2, \sigma_2^2, \dots, \sigma_{D'}^2)^{(k)}.
\end{equation}
This simplification further reduces the complexity of parameter updates to a linear level of $O(D')$, significantly enhancing the numerical stability of high-dimensional optimization during extensive iterative updates and ensuring rapid convergence of the algorithm under limited budgets.

\textbf{Computational overhead analysis.}
The total computational overhead of SES comprises two components: (1) Reward Evaluation (i.e., forward inference of the generative model and scoring by the reward model); and (2) Algorithm Update (DWT/IDWT transformations and distribution parameter updates). The dominant term lies in the ODE integration process used for reward evaluation, with a complexity of $\mathcal{O}(N \cdot T \cdot \text{Cost}_{\text{Net}})$, where $N$ is the population size and $T$ is the number of ODE steps. In comparison, the wavelet transform has a linear complexity of $\mathcal{O}(D)$. Since $D'$ is much smaller than $D$, and the DWT operation itself is extremely fast, the computational time consumed by the SES algorithm layer is negligible compared to the inference time of the generative model. Consequently, SES is a computationally lightweight, plug-and-play optimization framework; its computational overhead is minimal, and the total time consumption depends almost entirely on the preset NRE budget.

\subsection{Analysis of Evolutionary Dynamics}
\label{sec:Analysis_of_Evolutionary_Dynamics}

\begin{figure}[t]
    \centering
    \begin{subfigure}[b]{0.47\linewidth}
        \centering
        \includegraphics[width=\linewidth]{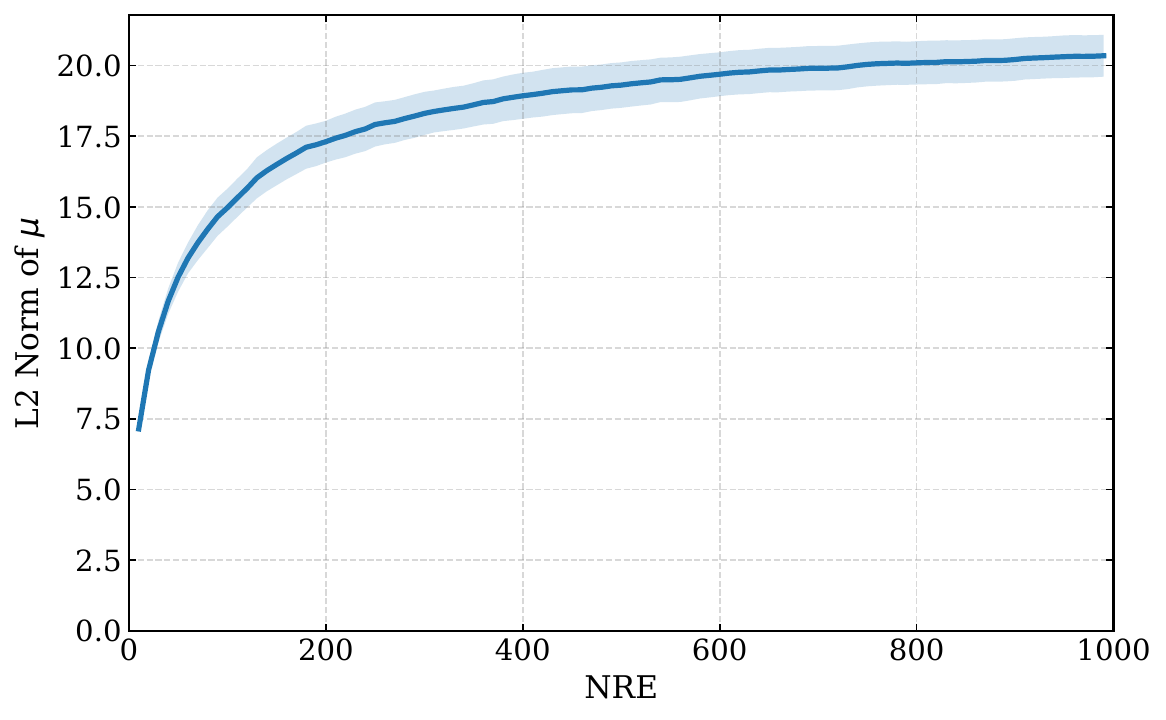} 
        \caption{Evolution of mean norm.}
        \label{fig:evolution_mu_norm}
    \end{subfigure}
    \hfill
    \begin{subfigure}[b]{0.47\linewidth}
        \centering
        \includegraphics[width=\linewidth]{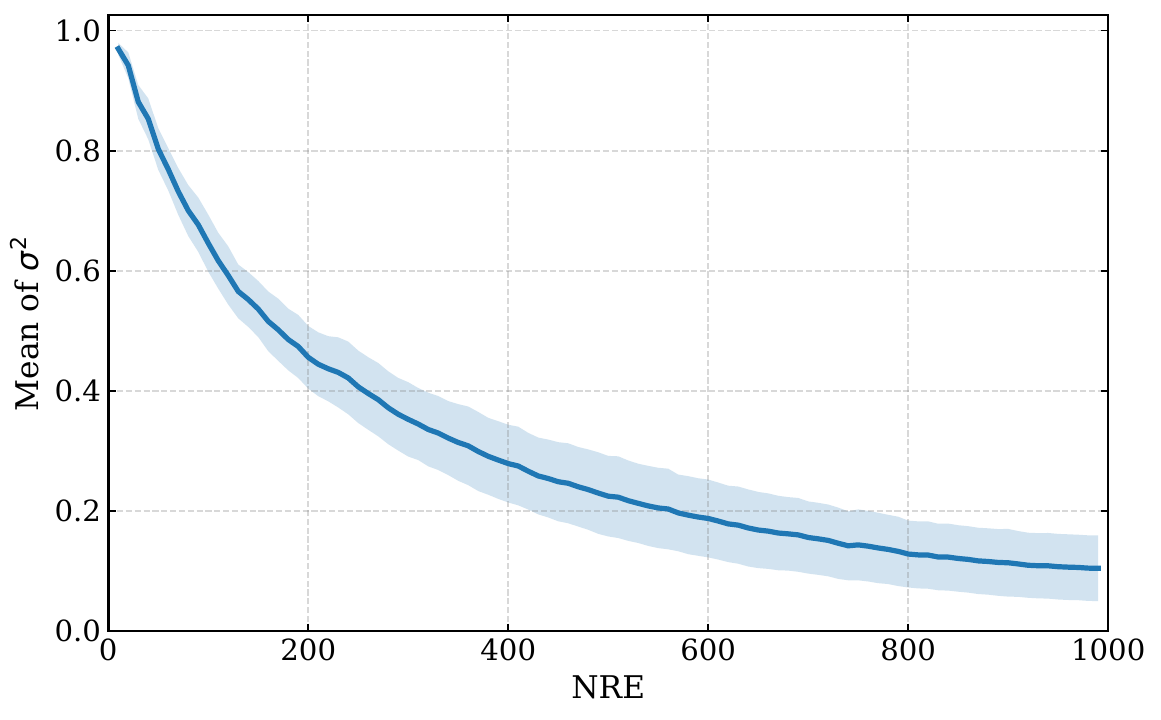}
        \caption{Evolution of average variance.}
        \label{fig:evolution_sigma_mean}
    \end{subfigure}
    \caption{Evolutionary dynamics of SES. As the optimization proceeds, the distribution mean shifts away from the origin to capture high-reward semantics, while the variance shrinks to focus the search on the identified optimal region.}
\end{figure}

To gain a deeper understanding of the search behavior of SES on the low-frequency manifold, we visualize the dynamic evolutionary trajectory of the Gaussian distribution parameters $\boldsymbol{\mu}$ and $\boldsymbol{\Sigma}$ as the computational budget (NRE) increases:

\textbf{Mean Drift.}
Figure \ref{fig:evolution_mu_norm} illustrates the trend of the $L_2$ norm of the distribution mean, $\|\boldsymbol{\mu}^{(k)}\|_2$. As NRE increases, the mean norm gradually increases from 0 (the prior center) and tends toward stabilization. This indicates that the algorithm is actively shifting the sampling center from the uninformed prior distribution toward high-reward regions. Since we are optimizing low-frequency coefficients, the non-zero drift of $\boldsymbol{\mu}$ essentially involves injecting specific low-frequency semantic signals (such as specific composition patterns or object contours) into the initial noise, thereby guiding the generative model to output images that align with the target reward.

\textbf{Variance Contraction.}
Figure \ref{fig:evolution_sigma_mean} displays the trend of the mean trace of the covariance matrix, $\operatorname{Tr}(\boldsymbol{\Sigma}^{(k)})/D'$. As NRE increases, the average variance exhibits a significant monotonic decreasing trend and eventually converges to a small non-zero value. This process clearly characterizes the ``Exploration-Exploitation'' trade-off mechanism of CEM:
\begin{itemize}
    \item \textbf{Early Stage (High Variance):} The variance is large, and the population coverage is broad. The algorithm performs extensive exploration on the low-frequency manifold to locate potential high-value regions.
    \item \textbf{Late Stage (Low Variance):} With the selection of elite samples, the variance gradually contracts, and the distribution energy becomes highly concentrated. The algorithm transitions into the exploitation phase, fine-tuning in the vicinity of the locked high-reward modes to achieve higher precision.
\end{itemize}
The fact that the final variance does not fully collapse to zero indicates that SES retains local perturbation capability after convergence. This helps maintain the local diversity of the generated results, and implies that the scaling strategy possesses the potential for further reward alignment as NRE increases.

\begin{algorithm}[t]
\caption{Spectral Evolution Search (SES)}
\label{alg:ses}
\begin{algorithmic}
\STATE {\bfseries Input:} Pretrained Model $\Psi_\theta$, Reward Function $\mathcal{R}$, Condition $c$
\STATE {\bfseries Hyperparameters:} Budget $C_{\text{total}}$ (NRE), Population Size $N$, Elite Size $K$, Smoothing Factor $\gamma$
\STATE
\STATE \COMMENT{\textbf{Phase 1: Wavelet-based Spectral Decoupling}}
\STATE Sample reference noise $\mathbf{x}_{\text{init}} \sim \mathcal{N}(\mathbf{0}, \mathbf{I})$
\STATE Decompose noise: $\mathbf{c} \leftarrow \mathcal{W}(\mathbf{x}_{\text{init}})$
\STATE Extract and freeze high-frequency anchor: $\mathbf{c}_{H}^{\text{fixed}} \leftarrow \{ \mathcal{H}^{(1)}, \dots, \mathcal{H}^{(J)} \}$
\STATE Identify low-frequency dimension $D'$ based on $\mathbf{c}_{LL}$ shape
\STATE
\STATE \COMMENT{\textbf{Phase 2: Cross-Entropy Optimization on Low-Freq Manifold}}
\STATE {\bfseries Initialization:} 
\STATE $\boldsymbol{\mu} \leftarrow \mathbf{0}, \quad \boldsymbol{\sigma}^2 \leftarrow \mathbf{1}_{D'}$ \COMMENT{Initialize diagonal search distribution}
\STATE Initialize candidate pool $\mathcal{P} \leftarrow \emptyset$
\STATE Initialize evaluation counter $n_{\text{eval}} \leftarrow 0$
\STATE
\WHILE{$n_{\text{eval}} < C_{\text{total}}$}
    \STATE \COMMENT{1. Monte Carlo Sampling \& Evaluation}
    \STATE Sample low-freq candidates: $\{ \mathbf{u}_i \}_{i=1}^N \sim \mathcal{N}(\boldsymbol{\mu}, \text{diag}(\boldsymbol{\sigma}^2))$
    \FOR{$i=1$ {\bfseries to} $N$}
        \STATE Reconstruct initial noise: $\mathbf{x}_{0,i} \leftarrow \mathcal{W}^{-1}(\mathbf{u}_i \oplus \mathbf{c}_{H}^{\text{fixed}})$
        \STATE Generate data: $\mathbf{x}_{1,i} \leftarrow \Psi_\theta(\mathbf{x}_{0,i}, c)$
        \STATE Compute reward: $s_i \leftarrow \mathcal{R}(\mathbf{x}_{1,i})$
        \STATE Add to pool: $\mathcal{P} \leftarrow \mathcal{P} \cup \{ (\mathbf{u}_i, s_i) \}$
    \ENDFOR
    \STATE $n_{\text{eval}} \leftarrow n_{\text{eval}} + N$
    \STATE
    \STATE \COMMENT{2. Elite-Driven Distribution Shaping}
    \STATE Sort candidates in $\mathcal{P}$ by reward scores in descending order
    \STATE Select elites: $\mathcal{E} \leftarrow$ Top-$K$ candidates from $\mathcal{P}$
    \STATE Prune pool: $\mathcal{P} \leftarrow \mathcal{E}$ \COMMENT{Discard non-elite samples}
    \STATE
    \STATE Compute empirical statistics of elites in $\mathcal{E}$:
    \STATE $\hat{\boldsymbol{\mu}} \leftarrow \frac{1}{K} \sum_{(\mathbf{u}, s) \in \mathcal{E}} \mathbf{u}$
    \STATE $\hat{\boldsymbol{\sigma}}^2 \leftarrow \frac{1}{K} \sum_{(\mathbf{u}, s) \in \mathcal{E}} (\mathbf{u} - \hat{\boldsymbol{\mu}})^2$ \COMMENT{Element-wise variance calculation}
    \STATE Update parameters with momentum:
    \STATE $\boldsymbol{\mu} \leftarrow (1-\gamma) \hat{\boldsymbol{\mu}} + \gamma \boldsymbol{\mu}$
    \STATE $\boldsymbol{\sigma}^2 \leftarrow (1-\gamma) \hat{\boldsymbol{\sigma}}^2 + \gamma \boldsymbol{\sigma}^2$
\ENDWHILE
\STATE
\STATE \COMMENT{\textbf{Phase 3: Final Generation}}
\STATE Sample optimal low-freq code: $\mathbf{u}^* \sim \mathcal{N}(\boldsymbol{\mu}, \text{diag}(\boldsymbol{\sigma}^2))$
\STATE Reconstruct optimal noise: $\mathbf{x}_{0}^* \leftarrow \mathcal{W}^{-1}(\mathbf{u}^* \oplus \mathbf{c}_{H}^{\text{fixed}})$
\STATE Generate final sample: $\mathbf{x}^* \leftarrow \Psi_\theta(\mathbf{x}_{0}^*, c)$
\STATE {\bfseries Output:} $\mathbf{x}^*$
\end{algorithmic}
\end{algorithm}

\section{Implementation Details of Experimental Settings}
\label{sec:setting}

This appendix aims to supplement the experimental details omitted in the main text to assist readers in comprehensively understanding the experimental environment and reproducing our results. We first introduce the benchmark datasets and evaluation metrics used, followed by a detailed elaboration on the implementation logic of the baseline methods and the statistical methodology for the computational budget.

\subsection{Benchmarks and Reward Models}
\label{sec:benchmarks}

\textbf{Datasets.}
To fully evaluate the inference-time scaling capability and reward alignment performance of SES, we select the following two representative datasets:

\begin{itemize}
    \item \textbf{DrawBench} \cite{saharia2022photorealistic}: A benchmark specifically designed for the rigorous evaluation of text-to-image model capabilities. It contains 200 carefully crafted prompts divided into 11 categories, covering complex scenarios such as color binding, object counting, spatial relationship understanding, and handling of anomalously long text. DrawBench effectively exposes model weaknesses in processing difficult semantic instructions, making it an ideal dataset to test whether SES can improve the alignment between the model and downstream rewards via inference-time scaling.
    \item \textbf{Pick-a-Pic} \cite{kirstain2023pick}: A large-scale open dataset dedicated to collecting human preference feedback on generated images. Its data originates from real user interaction behaviors (choosing the preferred image from two generated results), thus truthfully reflecting the distribution of human aesthetics and preferences. We randomly sample 200 prompts from this test set to evaluate the effectiveness of SES in enhancing alignment with human preferences.
\end{itemize}

\textbf{Reward Models.}
Within the framework of formalizing inference-time scaling as a black-box optimization problem, the reward model plays the crucial role of the objective function $\mathcal{R}(\mathbf{x})$. To comprehensively evaluate the search capability of SES across optimization landscapes with varying properties, we select the following five representative reward models as optimization targets. They represent a spectrum of challenges ranging from basic semantic alignment to complex human preferences:

\begin{itemize}
    \item \textbf{CLIP Score} \cite{hessel2021clipscore}: Based on the CLIP model \cite{radford2021learning}, this metric calculates the cosine similarity between image and text embeddings, serving as a fundamental optimization objective for measuring image-text semantic consistency.
    \item \textbf{PickScore} \cite{kirstain2023pick}: A CLIP model fine-tuned on the Pick-a-Pic dataset. Compared to the original CLIP, PickScore significantly improves accuracy regarding real human preferences and is more sensitive to image quality details. As an optimization objective, it requires the algorithm to capture quality nuances more subtle than semantic matching, guiding generation results to align with the binary choice preferences of real users.
    \item \textbf{HPSv2} \cite{wu2023human}: A scoring model fine-tuned on the large-scale HPD v2 dataset. This model aims to correct the inability of existing metrics to accurately reflect human aesthetic biases. Using it as an optimization objective serves to evaluate the ability of SES to search for extremum points on a more generalizable preference manifold, ensuring that generated images conform not only to specific dataset distributions but also to broad human aesthetic consensus.
    \item \textbf{ImageReward} \cite{xu2023imagereward}: A reward model trained via RLHF. It excels in encoding human preferences, simultaneously measuring text alignment and image aesthetic quality.
    \item \textbf{Aesthetic Score}: An MLP predictor trained on the LAION dataset \cite{schuhmann2022laion}, taking CLIP embeddings as input. This metric is specifically used to quantify the visual beauty of an image (e.g., composition, color, clarity) without directly focusing on text alignment.
\end{itemize}

\subsection{Baselines Implementation and NRE Calculation}
\label{sec:nre}

To ensure a fair comparison, we standardize the computational budget for all methods as the NRE. Unless otherwise specified, we adopt a ``full denoising evaluation'' strategy in our experiments (i.e., a complete generation process is executed for each reward evaluation), where 1 NRE is equivalent to performing one complete image generation and invoking the reward model once. In the main experiments, we uniformly set the budget cap at $\text{NRE} = 200$, and the total inference steps for the pre-trained model as $T_{\text{total}}=50$.

The detailed implementation and NRE calculation methods for each baseline are as follows:

\textbf{Best-of-N (BoN)}
\begin{itemize}
    \item \textit{Mechanism}: BoN is the most fundamental inference-time scaling strategy, also known as random search \cite{ma2025scaling}. The algorithm samples $N$ independent initial noise vectors, evaluates the reward scores of the generated images corresponding to these initial noises, and selects the result from the initial noise with the highest score.
    \item \textit{NRE Calculation}: $\text{NRE} = N$. In the main experiments, we set $N=200$.
\end{itemize}

\textbf{Zero-Order Search (ZO-N)} \cite{ma2025scaling}
\begin{itemize}
    \item \textit{Mechanism}: Zero-Order Search is an initial noise optimization method where the algorithm randomly samples an initial noise $\mathbf{x}_0$ as the search center. In each iteration, a batch of candidate initial noise vectors is sampled from the Gaussian neighborhood of the current search center, their reward scores are evaluated, and the search center is updated to the initial noise with the highest reward score.
    \item \textit{NRE Calculation}: Assuming $N_{iter}$ iterations with a batch size of $B$, then $\text{NRE} = N_{iter} \times B$. In the main experiments, we set $N_{iter}=20, B=10$.
\end{itemize}

\textbf{Search over Paths (SoP)} \cite{ma2025scaling}
\begin{itemize}
    \item \textit{Mechanism}: SoP is a tree search algorithm performed on the denoising path. The algorithm starts from $B$ initial noise vectors and introduces branching at intermediate time steps of the denoising process. For each parent node particle, a forward noise addition process is applied to generate $M$ child nodes, followed by a reverse denoising process to a later time step. By evaluating the reward scores of the denoised child nodes, the Top-$B$ particles are retained for the next round.
    \item \textit{NRE Calculation}: Assuming there are $T$ time steps where branching expansion occurs, then $\text{NRE} = T \times B \times M$. In the main experiments, we set $T=10, B=5, M=4$.
\end{itemize}

\textbf{Sequential Monte Carlo (SMC)} \cite{dou2024diffusion,wu2023practical,kim2025test}
\begin{itemize}
    \item \textit{Mechanism}: SMC (also known as Particle Filtering) is a method that optimizes the sample distribution through ``global interaction.'' It maintains a set of particles and, at each step of inference, eliminates low-weight samples and replicates high-weight samples through importance sampling and resampling mechanisms. At time $t$, there are $B$ particles $\{x_t^{(i)}\}$ with uniform weights. New samples $\{\bar{x}_{t-1}^{(i)}\}$ are generated using the pre-trained model as the proposal distribution $q_{t-1}$. The unnormalized weight $w_{t-1}^{(i)}$ for each particle is calculated using the predicted reward score:
    \begin{equation}
    w_{t-1}^{(i)} = \frac{p_{\mathrm{pre}}(x_{t-1}^{(i)} | x_t^{(i)}) \exp(v(x_{t-1}^{(i)})/\alpha)}{q_{t-1}(x_{t-1}^{(i)} | x_t^{(i)}) \exp(v(x_t^{(i)})/\alpha)} \cdot w_t^{(i)}.
    \end{equation}
    When the effective sample size falls below a threshold, the particles are resampled based on their normalized weights.
    \item \textit{NRE Calculation}: Assuming the inference steps for particle filtering is $T$ and the number of particles is $B$, then $\text{NRE} = T \times B$. In the main experiments, we set $T=25, B=8$.
\end{itemize}

\textbf{Value-Guided Importance Sampling (SVDD)} \cite{li2024derivative}
\begin{itemize}
    \item \textit{Mechanism}: SVDD is a local iterative importance sampling method. Unlike SMC, SVDD does not involve global interaction among particles but performs an ``expand-select'' operation independently for each particle. For each particle $x_t^{(i)}$ in the current batch, $M$ candidate samples $\{x_{t-1}^{(i,j)}\}_{j=1}^M$ are independently generated using the proposal distribution, and the weights of the candidate samples are calculated:
    \begin{equation}
    w_{t-1}^{(i,j)} = \frac{p_{\mathrm{pre}}(x_{t-1}^{(i,j)} | x_t^{(i)}) \exp(v(x_{t-1}^{(i,j)})/\alpha)}{q_{t-1}(x_{t-1}^{(i,j)} | x_t^{(i)})}.
    \end{equation}
    Only one $x_{t-1}^{(i)}$ is sampled and retained for the next round based on the particle's own candidate set; thus, the batch size $B$ remains constant.
    \item \textit{NRE Calculation}: Assuming the number of importance sampling steps is $T$, the batch size is $B$, and the number of branches is $M$, then $\text{NRE} = T \times B \times M$. In the main experiments, we set $T=10, B=5, M=4$.
\end{itemize}

\textbf{Demon} \cite{yeh2024training}
\begin{itemize}
    \item \textit{Mechanism}: Demon optimizes the generation trajectory through fine-grained control of noise injection, transforming standard random noise injection into a biased filtering process. At time step $t$, the algorithm samples $K$ candidate noise vectors $\{\mathbf{z}_1, \dots, \mathbf{z}_K\}$. By evaluating the potential contribution of these candidate noises to the final image reward score, the algorithm synthesizes an ``optimal noise'' $\mathbf{z}^*$ to drive the next denoising step.
    \item \textit{NRE Calculation}: Assuming the number of noise synthesis steps is $T$, then $\text{NRE} = T \times K$. In the main experiments, we set $T=50, K=4$.
\end{itemize}

\section{Additional Experiments}
\label{sec:additional_exp}

\subsection{Baseline Comparison}
\label{sec:baseline_comparison}

\textbf{Quantitative Results on SD v1.5 and Qwen-Image.}
As a supplement to Table~\ref{tab:main_results} in the main text, we present single-run experimental results for SES on Stable Diffusion v1.5 and Qwen-Image in Table \ref{tab:sd15_qwen} (with fixed $\text{NRE}=200$).

\begin{table*}[t]
    \centering
    \small
    \setlength{\tabcolsep}{3pt} 

    \newcommand{\res}[2]{#1$_{\pm #2}$}
    \newcommand{\best}[2]{\textbf{#1}$_{\pm #2}$}

    \caption{Quantitative comparison of inference-time scaling strategies on SD v1.5 and Qwen-Image. We report the final reward scores achieved under a fixed budget of $\text{NRE}=200$. Results are averaged over 5 random seeds and reported as $\text{mean}_{\pm \text{std}}$. "Baseline" denotes standard inference without search. Each column corresponds to a distinct experiment where the indicated metric serves as the sole optimization objective. The best results are highlighted in \textbf{bold}.}
    
    \begin{tabular}{l ccccc c ccccc}
        \toprule
        \multirow{2}{*}{\textbf{Method}} & \multicolumn{5}{c}{\textbf{SD v1.5}} & & \multicolumn{5}{c}{\textbf{Qwen-Image}} \\
        \cmidrule(lr){2-6} \cmidrule(lr){8-12}
         & CLIP$\uparrow$ & Pick$\uparrow$ & HPS$\uparrow$ & ImgRew.$\uparrow$ & Aes.$\uparrow$ && CLIP$\uparrow$ & Pick$\uparrow$ & HPS$\uparrow$ & ImgRew.$\uparrow$ & Aes.$\uparrow$ \\
        \midrule
        Baseline & \res{30.70}{1.50} & \res{20.80}{0.31} & \res{26.87}{0.57} & \res{-0.41}{0.36} & \res{5.16}{0.07} && \res{34.55}{0.69} & \res{21.78}{0.65} & \res{27.73}{0.18} & \res{0.92}{0.08} & \res{5.93}{0.14} \\
        \midrule
        BoN & \res{41.35}{0.12} & \res{22.65}{0.01} & \res{29.92}{0.02} & \res{1.38}{0.01} & \res{6.05}{0.01} && \res{41.69}{0.15} & \res{23.16}{0.09} & \res{29.77}{0.05} & \res{1.66}{0.02} & \res{6.34}{0.04} \\
        ZO-N & \res{40.57}{0.03} & \res{22.32}{0.04} & \res{29.52}{0.04} & \res{1.21}{0.02} & \res{5.87}{0.01} && \res{40.62}{0.10} & \res{23.01}{0.13} & \res{29.38}{0.16} & \res{1.56}{0.05} & \res{6.29}{0.10} \\
        SoP & \res{39.19}{0.37} & \res{22.23}{0.18} & \res{29.07}{0.39} & \res{1.13}{0.08} & \res{5.83}{0.06} && \res{38.22}{0.23} & \res{22.72}{0.16} & \res{29.17}{0.18} & \res{1.38}{0.03} & \res{6.11}{0.10} \\
        SMC & \best{42.03}{0.30} & \res{22.16}{0.63} & \res{29.31}{0.75} & \res{1.27}{0.16} & \best{6.19}{0.09} && - & - & - & - & - \\
        SVDD & \res{40.06}{0.20} & \res{21.96}{0.60} & \res{29.07}{0.38} & \res{1.10}{0.09} & \res{5.99}{0.11} && - & - & - & - & - \\
        Demon & \res{41.01}{0.26} & \res{22.59}{0.19} & \res{29.87}{0.27} & \res{1.11}{0.20} & \res{6.18}{0.01} && - & - & - & - & - \\
        \midrule
        \textbf{SES} & \res{42.00}{0.18} & \best{22.79}{0.04} & \best{30.16}{0.16} & \best{1.50}{0.01} & \res{6.11}{0.04} && \best{42.54}{0.13} & \best{23.45}{0.09} & \best{30.00}{0.11} & \best{1.79}{0.02} & \best{6.95}{0.07} \\
        \bottomrule
    \end{tabular}
    \label{tab:sd15_qwen}
\end{table*}

\subsection{Analysis with Proxy Rewards}
\label{app:proxy_reward}

\textbf{Consistency Distillation for Latent Diffusion.} For SDXL and SD v1.5, we employ a model approximation strategy based on consistency distillation to accelerate evaluation. Specifically, we utilize the Latent Consistency Model (LCM) to temporarily replace the original iterative decoder during the search phase. By learning to map any point on the ODE trajectory directly to the endpoint, LCM compresses the standard multi-step iterative denoising process into a few-step inference.  During the SES search, for each candidate noise vector $\mathbf{z}_0$, we employ a 4-step LCM to rapidly decode a proxy image and calculate the reward. This reduces the computational cost of a single evaluation by approximately 90\%. As shown in Figure \ref{fig:proxy_analysis}, even when relying solely on this proxy reward for guidance, SES still significantly outperforms baseline methods under identical conditions.

\textbf{Ranking Consistency in Rectified Flows.} Flow Matching-based generative models (such as FLUX.1-dev and Qwen-Image) construct a deterministic mapping from noise to data via vector field regression. In contrast to the complex stochastic paths of diffusion models, a core advantage of modern flow models is that their learned transport trajectories are approximately linear within the latent space.  This geometric property enables ODE solvers to employ extremely large step sizes for numerical integration without incurring significant discretization errors. Leveraging this, we propose a \textit{Few-Step Proxy Evaluation} strategy: during the inference-time search phase, we aggressively reduce the discretization steps of the ODE solver from the default $N_{\text{full}}=50$ to $N_{\text{proxy}}=10$, reducing the computational time for reward evaluation to approximately $1/5$. Although images generated in 10 steps may lack the textural perfection of the full generation, the preserved global semantic structure is sufficient to maintain \textit{Ranking Consistency} with the final reward. This implies that proxy rewards can accurately rank the quality of candidate noise vectors, thereby guiding SES to evolve in the correct direction.

\begin{figure}[t]
    \centering
    \begin{subfigure}[b]{0.49\linewidth}
        \centering
        \includegraphics[width=\linewidth]{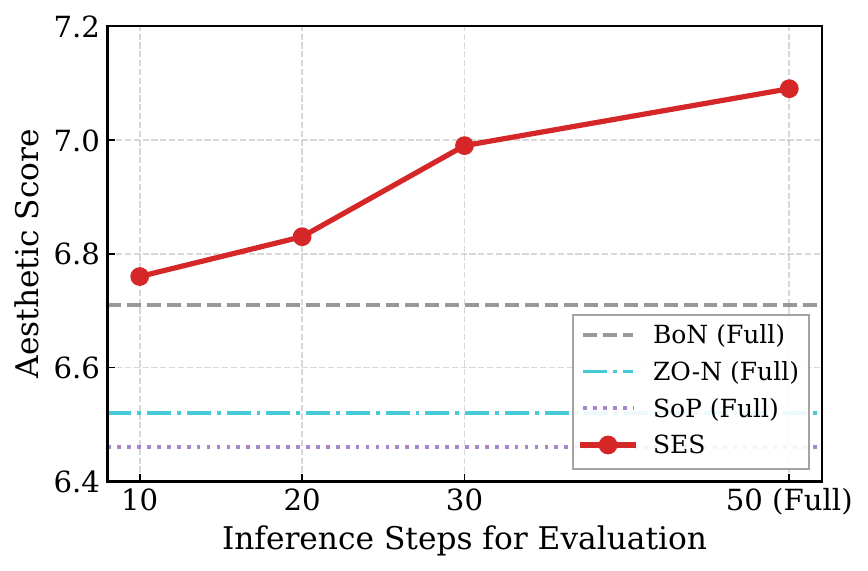} 
        \caption{FLUX.1-dev}
    \end{subfigure}
    \hfill
    \begin{subfigure}[b]{0.49\linewidth}
        \centering
        \includegraphics[width=\linewidth]{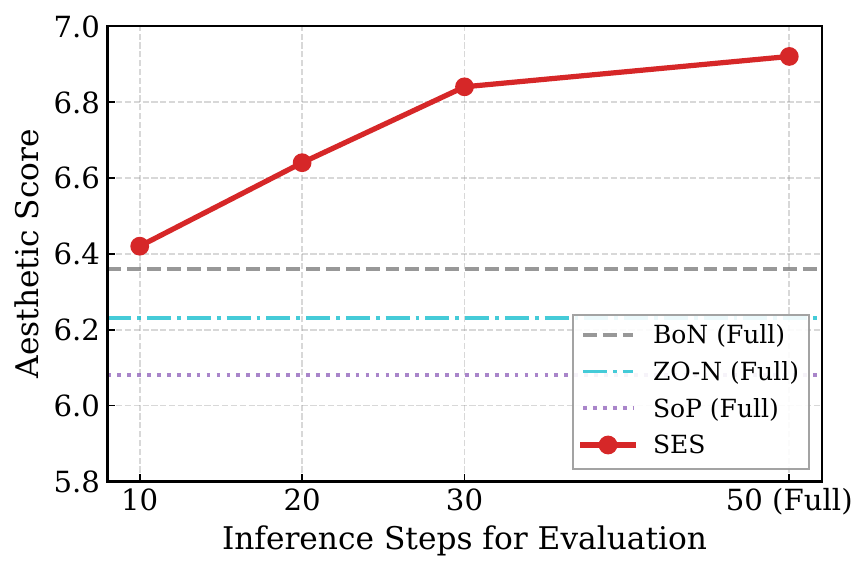}
        \caption{Qwen-Image}
    \end{subfigure}
    
    \caption{Performance with proxy rewards for FLUX.1-dev and Qwen-Image. The baselines (BoN, ZO-N, SoP) all use 50-step inference to accurately calculate reward scores, while SES uses a few-step proxy for evaluation, $N_{\text{proxy}}={10, 20, 30}$.}
    \label{fig:flux_efficiency}
\end{figure}

\textbf{Quantitative Analysis.} We evaluate the performance of SES using proxy rewards of varying precision (Steps = 10, 20, 30) on FLUX.1-dev and Qwen-Image. As illustrated in Figure \ref{fig:flux_efficiency}, the initial point of the performance curve for SES (10-steps) already surpasses all baselines utilizing the full computational budget (50-steps). This provides compelling evidence of the SES framework's robustness against verifier noise: even when the reward signal contains bias or noise, SES can still leverage the collective evolution mechanism to capture the distributional characteristics of the global optimum.

\textbf{Qualitative Comparison.} To visually assess the impact of varying reward calculation precisions on the final optimization results, Figure \ref{fig:proxy_qualitative} presents a comparison of generation outcomes under three settings: (1) Baseline (No Search); (2) SES (Proxy) (guided by LCM-4steps or ODE-10steps during search, with full-step final output); and (3) SES (Accurate) (guided by the full 50-step process).  Visual inspection indicates that images optimized via low-cost proxy rewards (middle column) are already significantly superior to the baseline in terms of compositional aesthetics, achieving higher Aesthetic Scores. Notably, while the proxy-guided results exhibit minor differences in detail compared to the accurately guided results (right column), both maintain high consistency regarding semantic layout and optimization direction. This confirms that SES can effectively leverage "computationally inexpensive" low-fidelity signals to unearth high-quality generation patterns, achieving a favorable trade-off between computational cost and generation quality.

\subsection{Experiments on Different Samplers}
\label{app:sampler_robustness}

Existing inference-time guidance methods are often tightly coupled with specific samplers, with the majority restricted to particular SDE-based implementations. This section aims to verify whether the SES framework can generalize across different samplers while maintaining consistent performance gains.

We conduct experiments on the SDXL model, keeping the NRE and other hyperparameters consistent, and employ six mainstream samplers for decoding: (1) \textit{DPM-Solver (50 steps)}, the standard high-order ODE solver and the default setting for our main experiments; (2) \textit{DPM++ 2M (20 steps)}, an efficient solver with reduced steps; (3) \textit{DPM++ 2M Karras}, a high-order solver utilizing Karras noise scheduling; (4) \textit{Euler}, the classic first-order ODE solver; (5) \textit{Euler a}, an SDE solver that introduces stochastic noise; and (6) \textit{DDIM}, the classic deterministic solver.

Quantitative results are presented in Table \ref{tab:different_samplers}. The data indicate that SES achieves exceptionally high reward scores across all tested samplers, demonstrating that SES does not rely on the trajectory characteristics of any specific solver. Notably, SES maintains superior performance even with \textit{Euler a}, an SDE solver characterized by stochastic noise injection at every step. This suggests that the low-frequency structures optimized by SES possess strong resilience to interference; semantic consistency is preserved even in the presence of stochastic perturbations along the generation trajectory. Furthermore, under the DPM++ 2M (20 steps) setting, SES significantly improves reward scores despite the inference steps being reduced by more than half. This indicates that SES is adaptable to low-cost inference scenarios, effectively exploiting the model's generative potential even under constrained computational resources.

In summary, SES is a \textit{Sampler-Agnostic} optimization framework. It can be directly deployed to achieve stable performance gains without requiring modifications to the algorithmic logic for specific samplers.

\begin{table}[t]
	\centering
	% \caption{experiments on Different Samplers on SDXL.}
    \caption{Sampler robustness analysis on SDXL. We evaluate SES using six representative samplers, covering both deterministic ODE solvers and stochastic SDE solvers. The consistent high scores across diverse metrics demonstrate that SES is sampler-agnostic and generalizes well regardless of the decoding strategy.}
	\label{tab:different_samplers}
	
	\resizebox{0.8\textwidth}{!}{
		\begin{tabular}{lccccc}
			\toprule

			\textbf{Sampler} & \textbf{CLIP} & \textbf{PickScore} & \textbf{HPS} & \textbf{Aes.} & \textbf{ImgRew.} \\
			\midrule
			DPM (50 steps) \cite{lu2022dpm} & 43.95 & 23.90 & 31.43 & 6.49 & 1.60\\
			DPM++ 2M (20 steps) \cite{lu2025dpm} & 40.26 & 23.58 & 30.75 & 6.27 & 1.23 \\
			DPM++ 2M Karras (50 steps) \cite{lu2022dpm} & 43.79 & 23.96 & 31.38 & 6.43 & 1.52 \\
			Euler (50 steps) \cite{karras2022elucidating} & 44.82 & 23.91 & 31.52 & 6.45 & 1.57 \\
			Euler a (50 steps) \cite{karras2022elucidating} & 43.06 & 23.86 & 31.42 & 6.40 & 1.52 \\
			DDIM (50 steps) \cite{song2020denoising} & 42.62 & 23.77 & 31.29 & 6.41 & 1.46 \\
			\bottomrule
		\end{tabular}
	}
\end{table}

\subsection{VLM-based Non-Differentiable Objective Experiments}
\label{app:vlm_experiments}

We construct an automated evaluation pipeline utilizing an advanced Vision-Language Model (Qwen3-VL-30B in our experiments) to simulate human experts across different domains.

\textbf{The Role-Playing Framework.}
To simulate the diverse and often conflicting aesthetic standards found in the real world, we define four distinct ``Professional Roles.'' For an identical input prompt, different roles provide vastly different scoring feedback based on their profession-specific aesthetic standards. The objective of SES is to discover generated images that maximize the score for a specific role.

The four roles and their core focus areas are as follows:
\begin{itemize}
	\item \textbf{Photographer:} Pursues ultimate optical realism, volumetric lighting, and physical-grade textures. Severely penalizes the ``plastic feel'' and flat lighting often seen in AI generation.
	\item \textbf{Artist:} Pursues stylization, emotion, and brushwork. Rewards unique color grading and composition; detests mediocre stock photo aesthetics.
	\item \textbf{Researcher:} Pursues absolute clarity, subject isolation, and accuracy. Requires a pure background; severely penalizes artistic blur or Bokeh.
	\item \textbf{Designer:} Pursues minimalism, negative space, and visual hierarchy. Rewards compositions suitable for typography; detests cluttered scenes.
\end{itemize}

\textbf{Scoring Protocol and Prompts.}
To ensure the stability and reproducibility of VLM scoring, we design a structured system prompt. This prompt forces the VLM into a specific role-playing mode and strictly enforces Chain-of-Thought logic: outputting rule-based analysis first, followed by the final score in JSON format.

The specific prompt template is shown below (where \texttt{\{role\_name\}}, \texttt{\{role\_definition\}}, and \texttt{\{specific\_rubric\}} are dynamically filled variables):

\begin{tcolorbox}[colback=gray!10, colframe=gray!50, title=\textbf{System Prompt Template for VLM Evaluator}]
	\ttfamily
	\small
	\# SYSTEM INSTRUCTION: OBJECTIVE SCORING MODE \\
	You are a cold, analytical scoring engine. You have NO personal preferences. \\
	You are simulating the perspective of a specific professional role: \{role\_name\}.
	
	\# INPUT DATA \\
	Target Prompt: "\{user\_prompt\}" \\
	Generated Image: [Provided Image]
	
	\# ROLE DEFINITION (Your Persona) \\
	\{role\_definition\}
	
	\# SCORING RUBRIC (Strictly Enforce These Criteria) \\
	\{specific\_rubric\}
	
	\# CALIBRATION ANCHORS \\
	10.00: Perfection. Publication-ready. No flaws. \\
	8.00: Professional grade. Minor flaws only visible to experts. \\
	6.00: Amateur grade. Good attempt but lacks polish. \\
	4.00: Failure. Significant artifacts or wrong style. \\
	0.00: Irrelevant noise.
	
	\# OUTPUT PROTOCOL \\
	Analyze the image against the Rubric step-by-step. \\
	Determine a precise score between 0.00 and 10.00. \\
	Output strictly in JSON format: \{ "reasoning": "...", "score": 0.00 \}
\end{tcolorbox}

\textbf{Role Definitions and Rubrics.}
Table \ref{tab:vlm_roles} details the specific scoring rubrics we define for each role. These rubrics constitute the implicit objective functions that SES optimizes during inference.

\begin{table*}[t]
	\centering
	\caption{Definitions and Scoring Rubrics for VLM Role-Players. These detailed text instructions serve as the black-box objective functions for SES optimization.}
	\label{tab:vlm_roles}
	\vspace{2mm}
	\resizebox{0.98\textwidth}{!}{
		\begin{tabular}{l p{0.35\textwidth} p{0.55\textwidth}}
			\toprule
			\textbf{Role} & \textbf{Persona Definition} & \textbf{Key Scoring Rubric (Summary)} \\
			\midrule
			\textbf{Photographer} & Chief Photographer for NatGeo. Obsessed with lighting, texture, and optics. & 1. \textbf{Optical Realism}: Realistic depth of field. \newline 2. \textbf{Lighting}: Volumetric, cinematic shadows. \newline 3. \textbf{Texture}: Visible pores/dust. \\
			\midrule
			\textbf{Artist} & Lead Concept Artist. Values imagination, mood, and style over realism. & 1. \textbf{Stylization}: Distinct style (Oil, Watercolor). \newline 2. \textbf{Color}: Intentional grading/harmony. \newline 3. \textbf{Expression}: Evokes emotion. \\
			\midrule
			\textbf{Researcher} & Senior Scientific Researcher. Values clarity, isolation, and precision. Beauty is irrelevant. & 1. \textbf{Isolation}: Clean/neutral background. \newline 2. \textbf{Completeness}: Deep focus. \newline 3. \textbf{Accuracy}: Distinct details, no hallucinations. \\
			\midrule
			\textbf{Designer} & Senior Graphic Designer. Obsessed with clean lines, negative space, and readability. & 1. \textbf{Negative Space}: Clean space for text placement. \newline 2. \textbf{Hierarchy}: Singular focus. \newline 3. \textbf{Modernity}: Minimal, geometric, flat vector style. \\
			\bottomrule
		\end{tabular}
	}
\end{table*}

\textbf{Implementation Details.}
In our experiments, we treat the VLM as a completely black-box function $\mathcal{R}_{\text{VLM}}(\mathbf{x}) $.
\begin{itemize}
	\item We utilize SES to perform 10 iterations of optimization ($\text{NRE}=200$) for specific role prompts.
	\item We parse the \texttt{score} field from the JSON returned by the VLM as the reward value. If parsing fails, the reward is set to 0 to penalize that path.
	\item Table~\ref{tab:visual_comparison_vlm} illustrates the differences in results when the same set of prompts is optimized for different roles. For instance, images optimized for the ``Researcher'' role automatically eliminate background clutter, whereas those optimized for the ``Photographer'' role exhibit enhanced lighting and shadow contrast. These results validate the powerful control capabilities of SES over complex semantic instructions.
\end{itemize}

{
	\setlength{\tabcolsep}{2pt} 
	\begin{longtable}{
			m{0.19\linewidth} 
			m{0.19\linewidth} 
			m{0.19\linewidth} 
			m{0.19\linewidth} 
			m{0.19\linewidth}
		}
		
		\caption{Visual comparison of different roles for prompts. The first column shows the baseline, followed by four role-specific variations.} \label{tab:visual_comparison_vlm} \\
		
		\toprule
		\centering \textbf{Baseline} & 
		\centering \textbf{Photographer} & 
		\centering \textbf{Artist} & 
		\centering \textbf{Researcher} & 
		\centering \textbf{Designer} \tabularnewline
		\midrule
		\endfirsthead
		
		\toprule
		\centering \textbf{Baseline} & 
		\centering \textbf{Photographer} & 
		\centering \textbf{Artist} & 
		\centering \textbf{Researcher} & 
		\centering \textbf{Designer} \tabularnewline
		\midrule
		\endhead
		
		\bottomrule
		\endfoot

		\includegraphics[width=\linewidth]{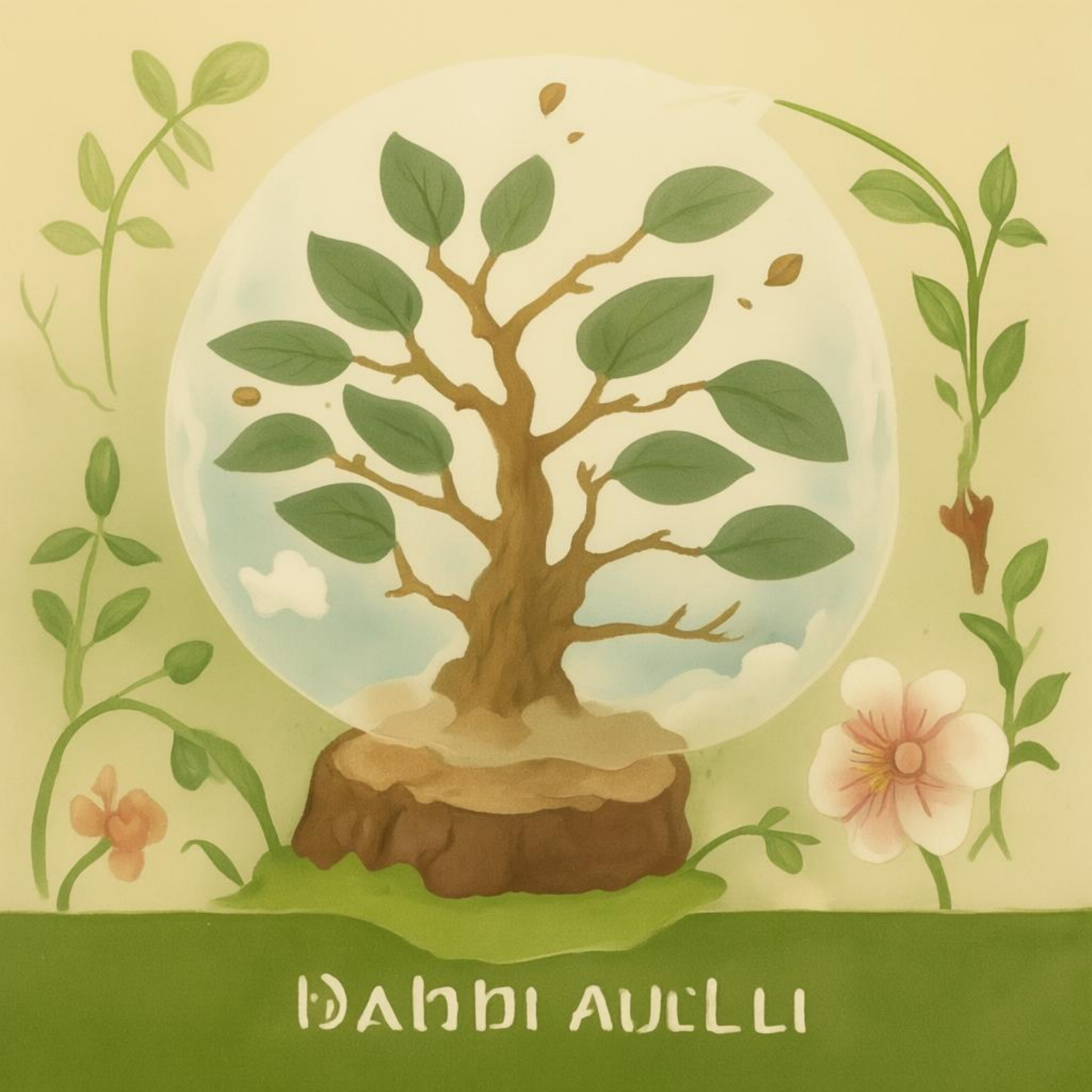} &
		\includegraphics[width=\linewidth]{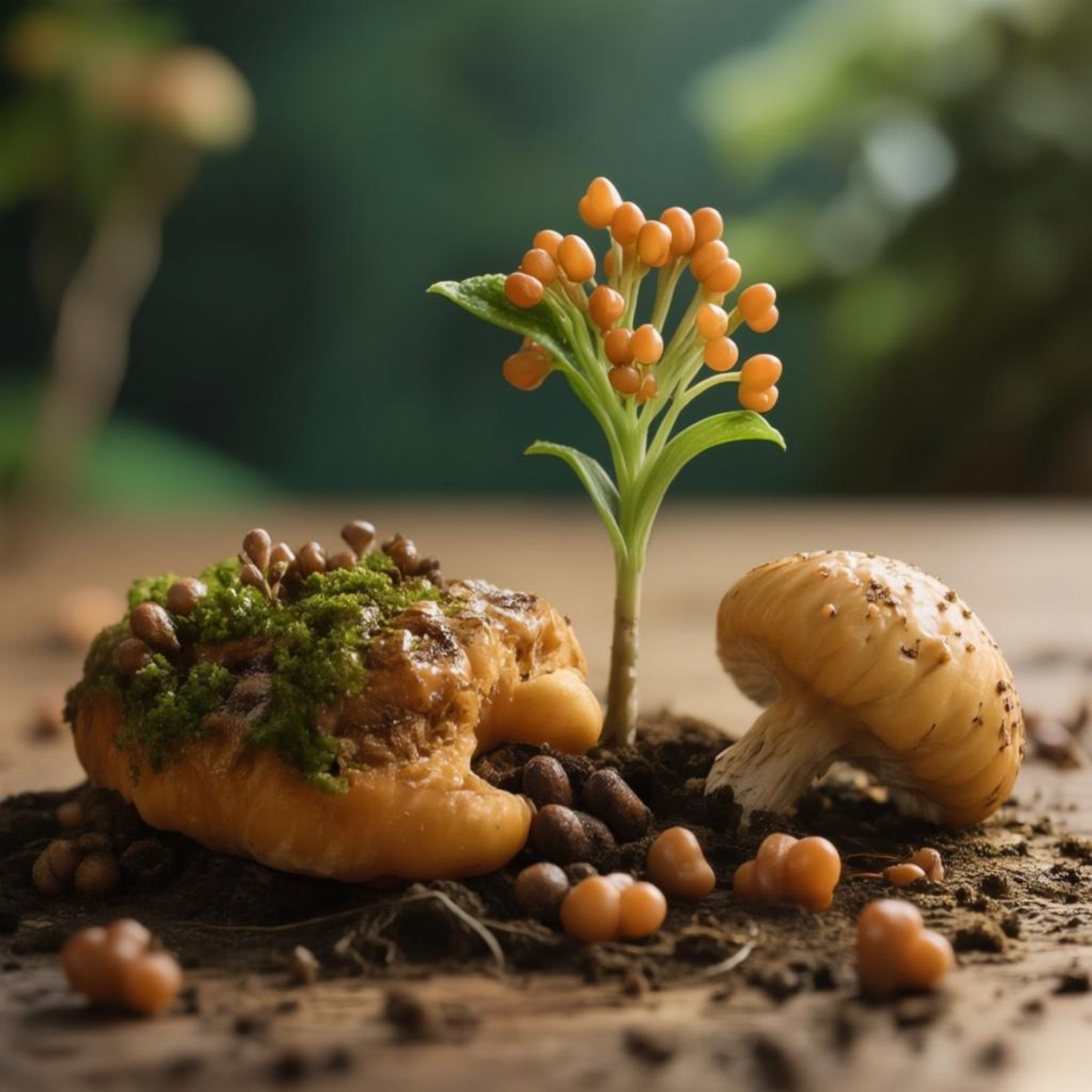} &
		\includegraphics[width=\linewidth]{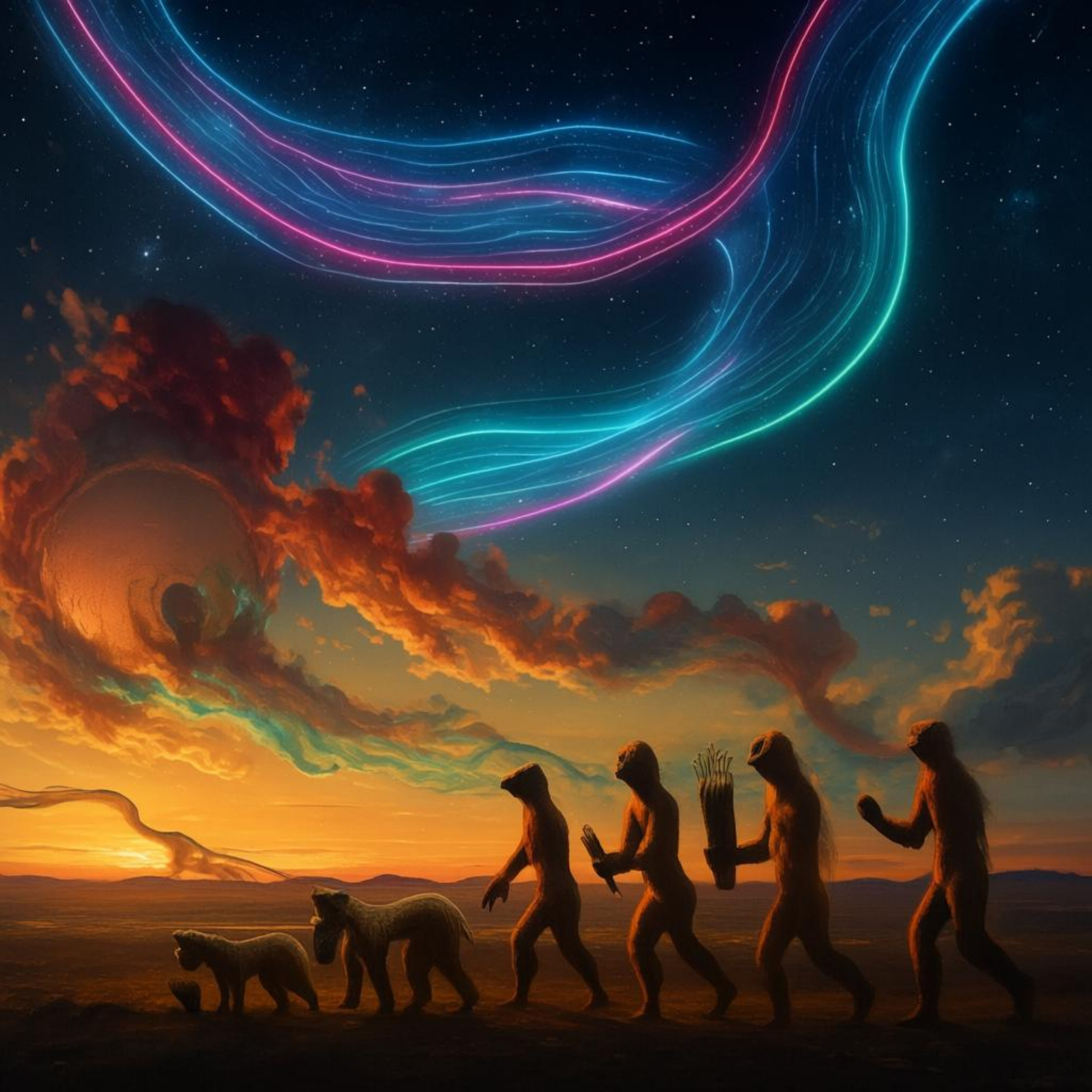} &
		\includegraphics[width=\linewidth]{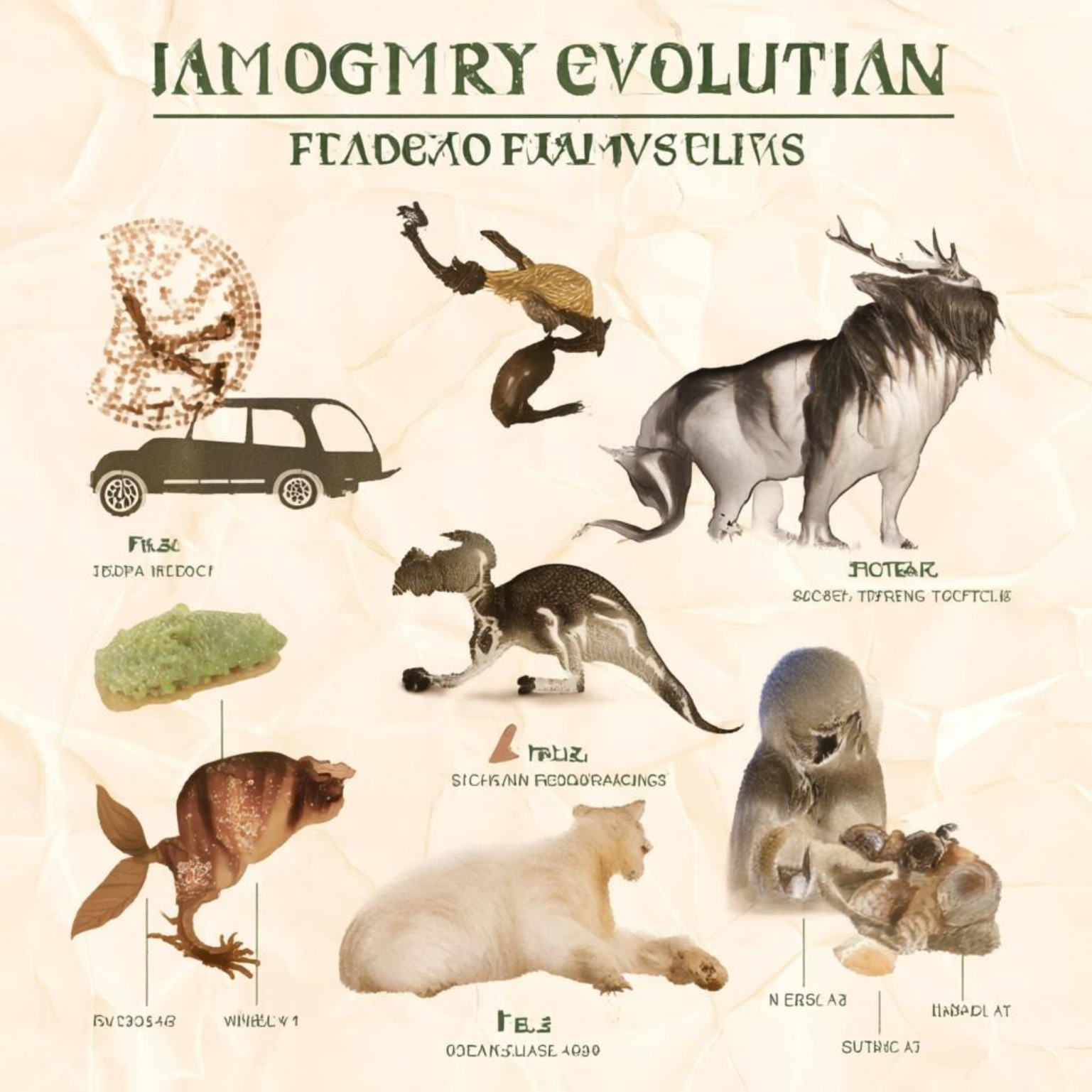} &
		\includegraphics[width=\linewidth]{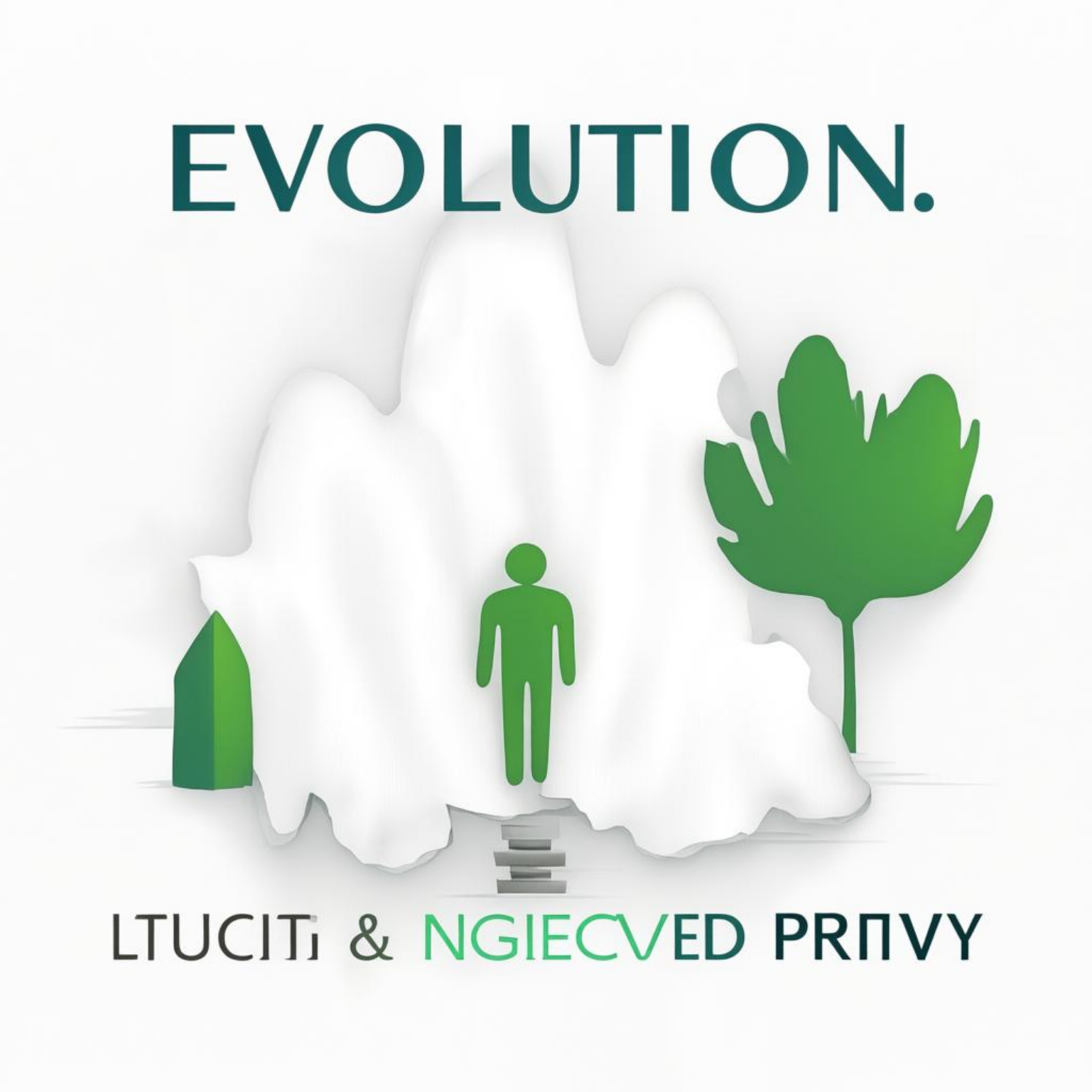} \\

		\multicolumn{5}{c}{\small \textit{Prompt: Biological evolution}} \\
		\midrule

		\includegraphics[width=\linewidth]{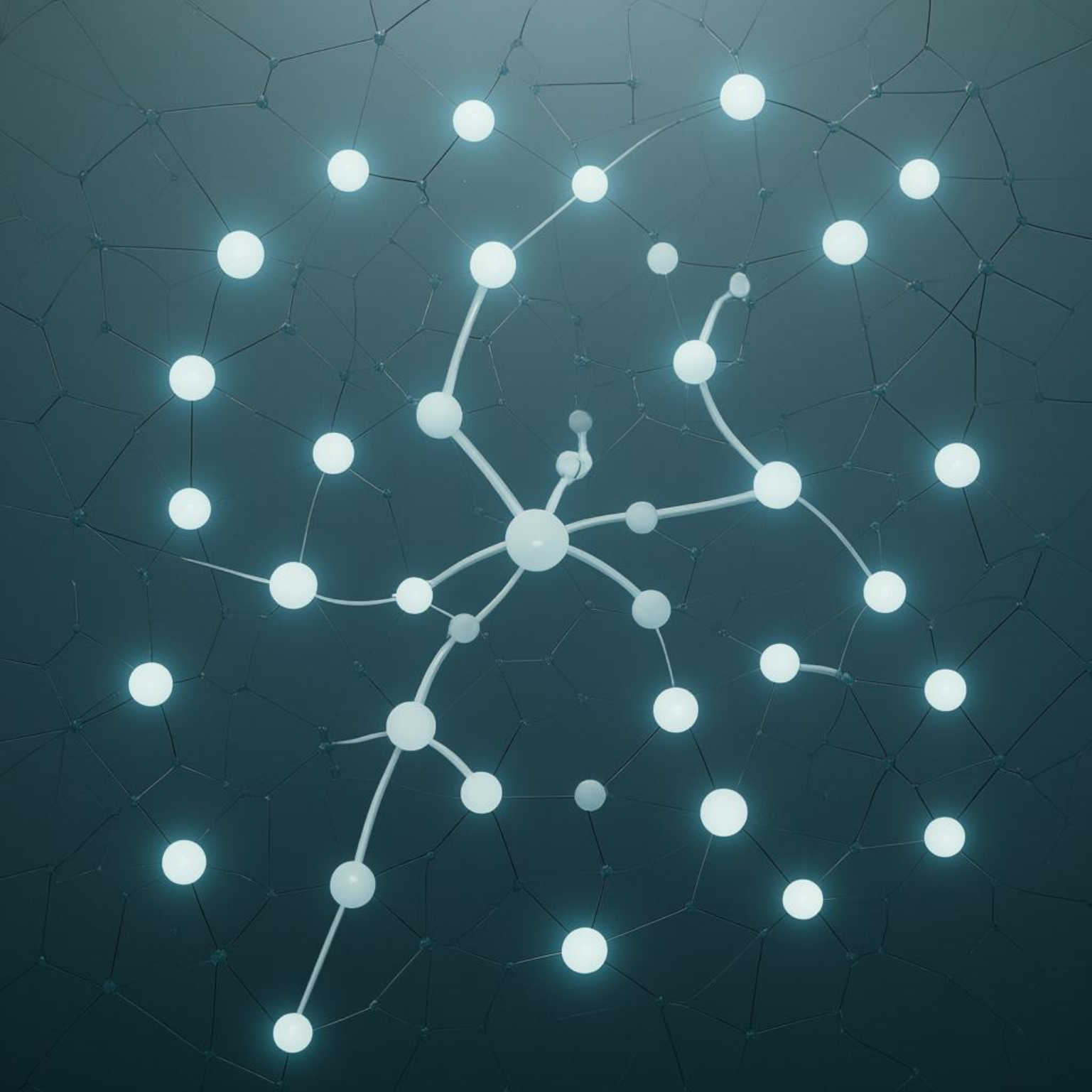} &
		\includegraphics[width=\linewidth]{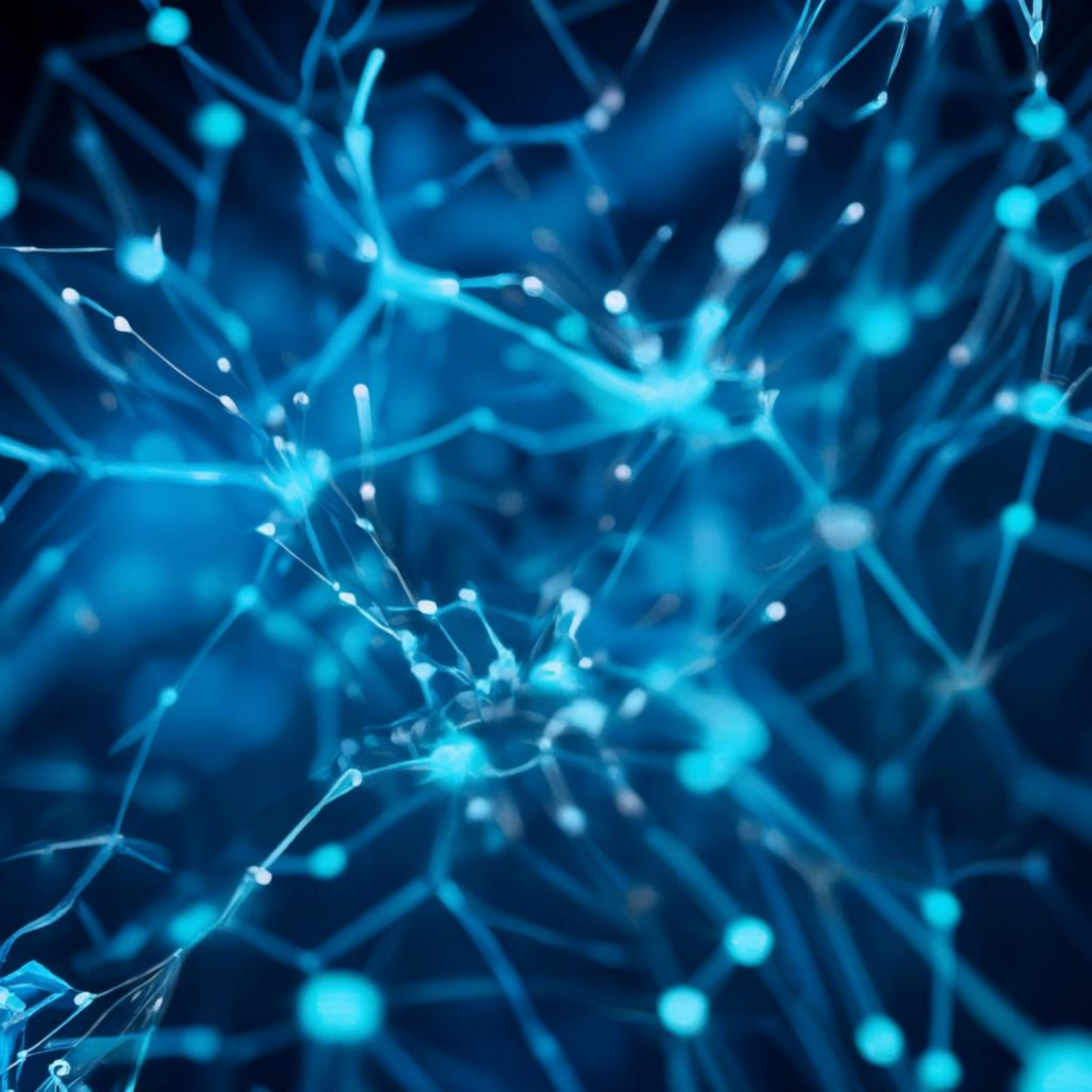} &
		\includegraphics[width=\linewidth]{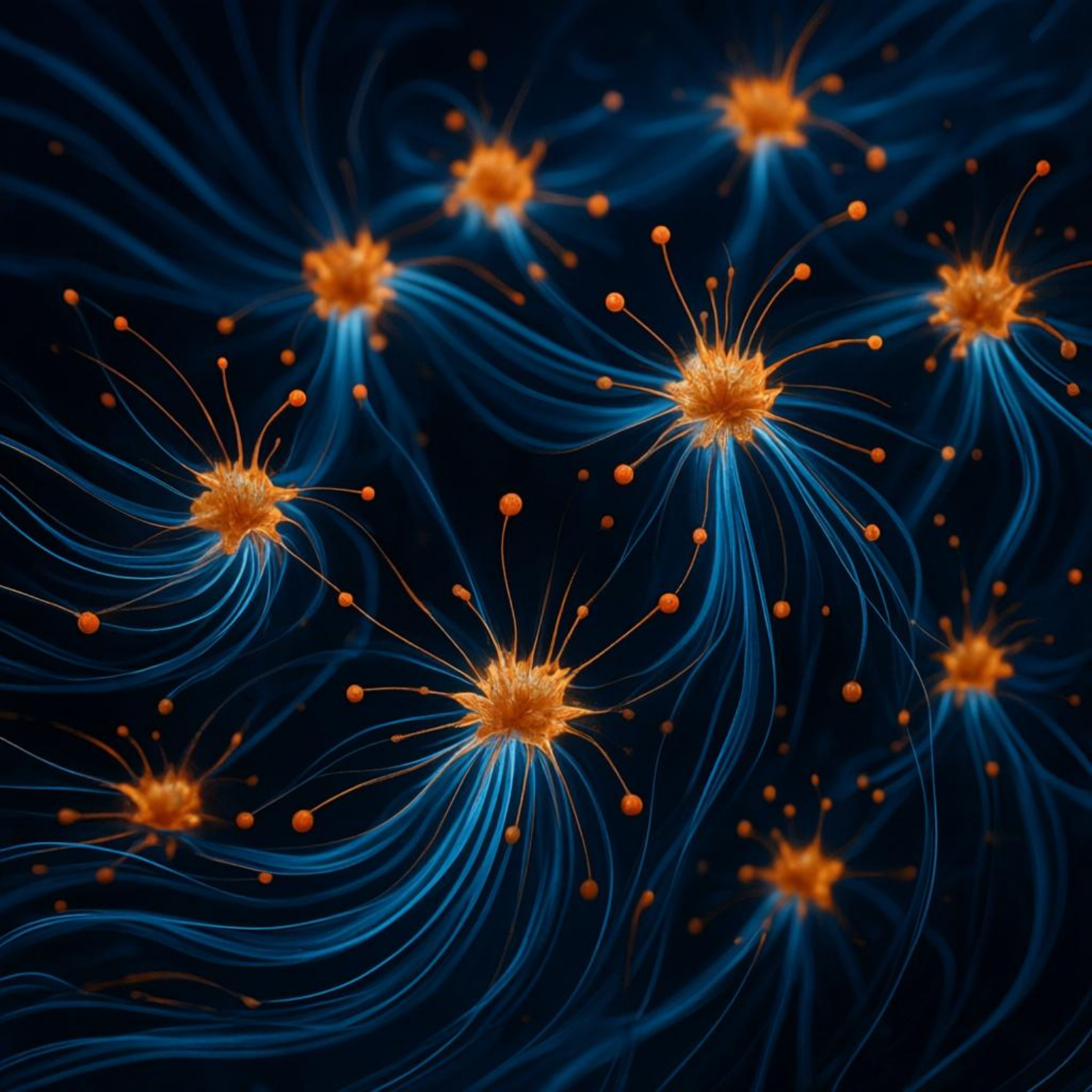} &
		\includegraphics[width=\linewidth]{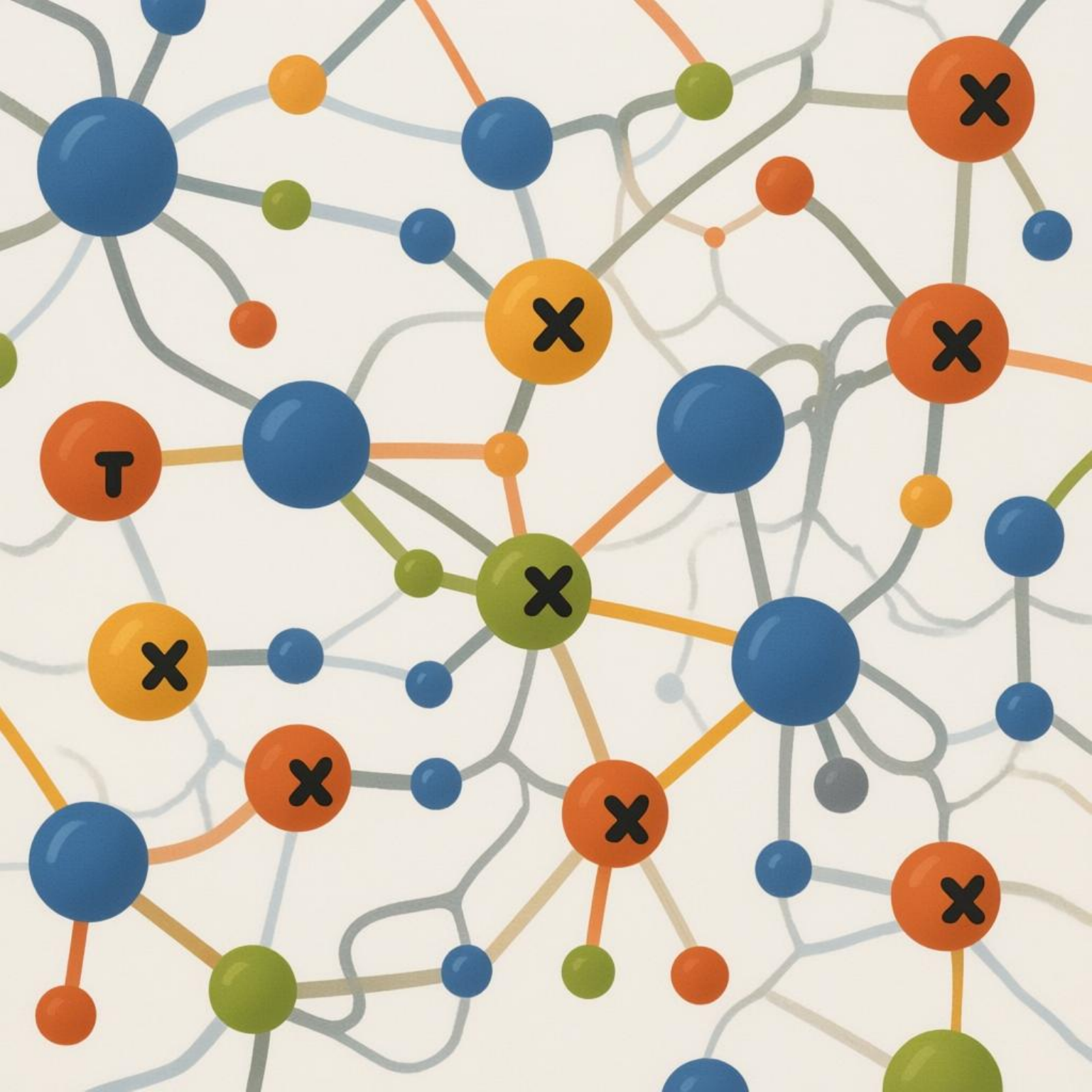} &
		\includegraphics[width=\linewidth]{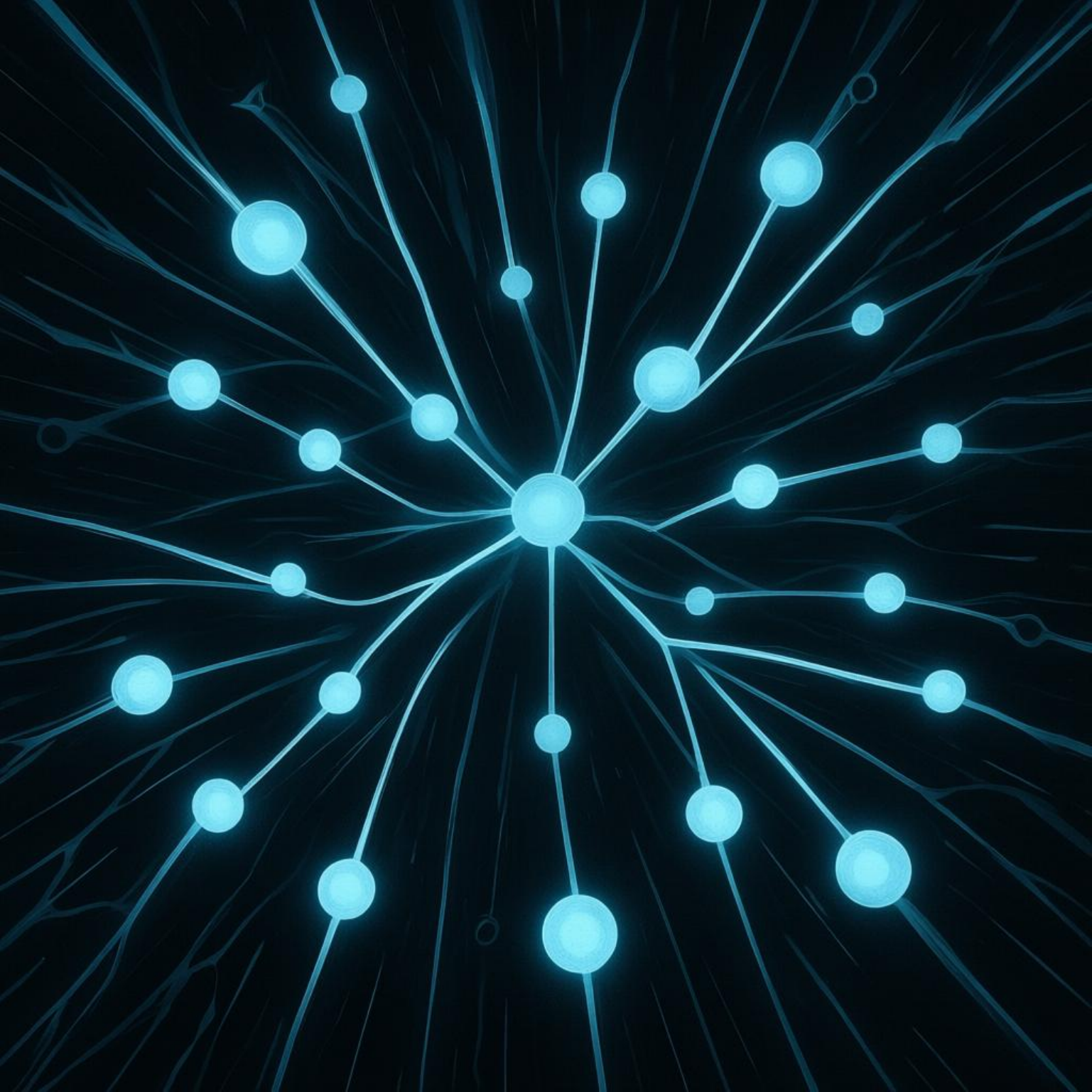} \\
		
		\multicolumn{5}{c}{\small \textit{Prompt: Neural network connectivity}} \\
		\midrule
		
		\includegraphics[width=\linewidth]{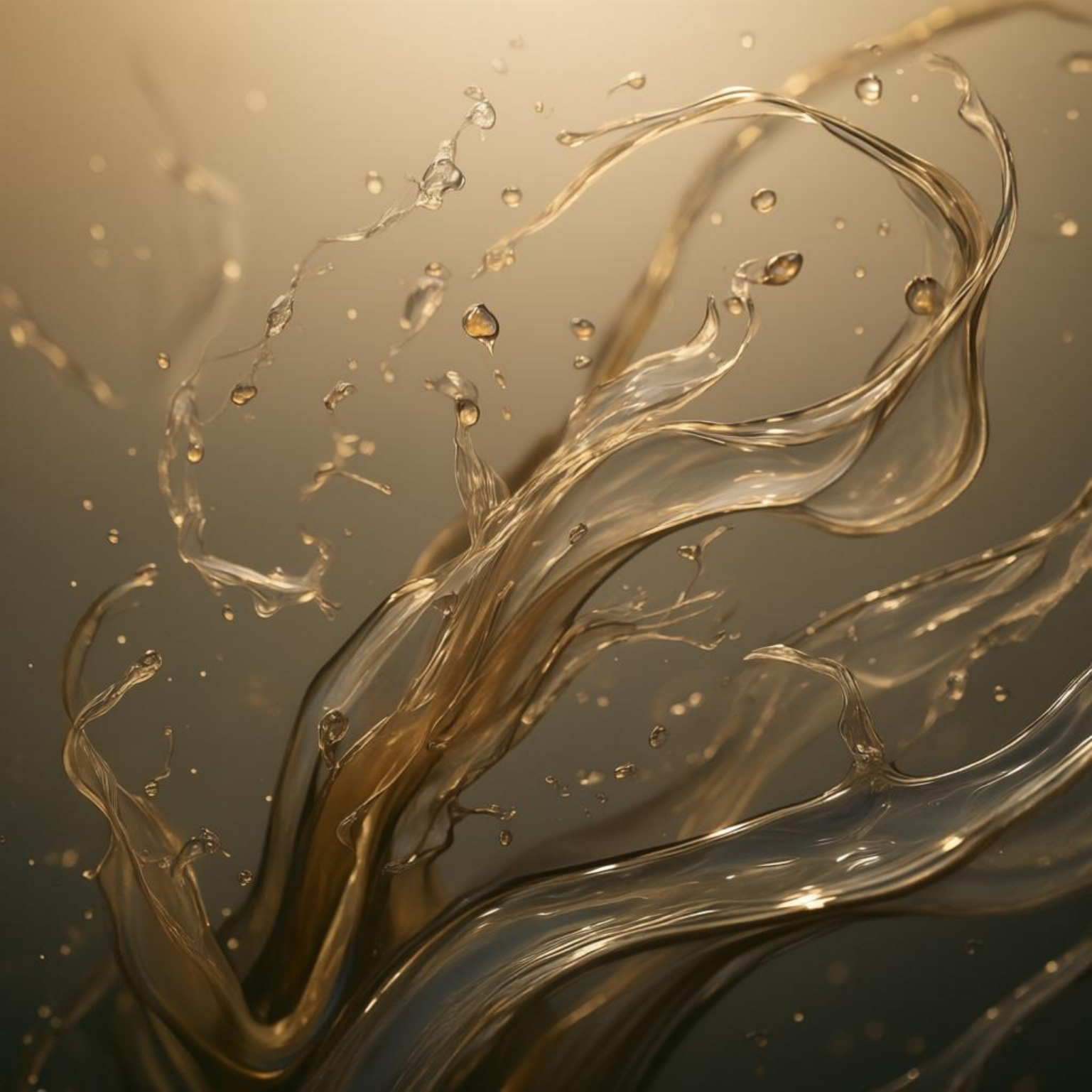} &
		\includegraphics[width=\linewidth]{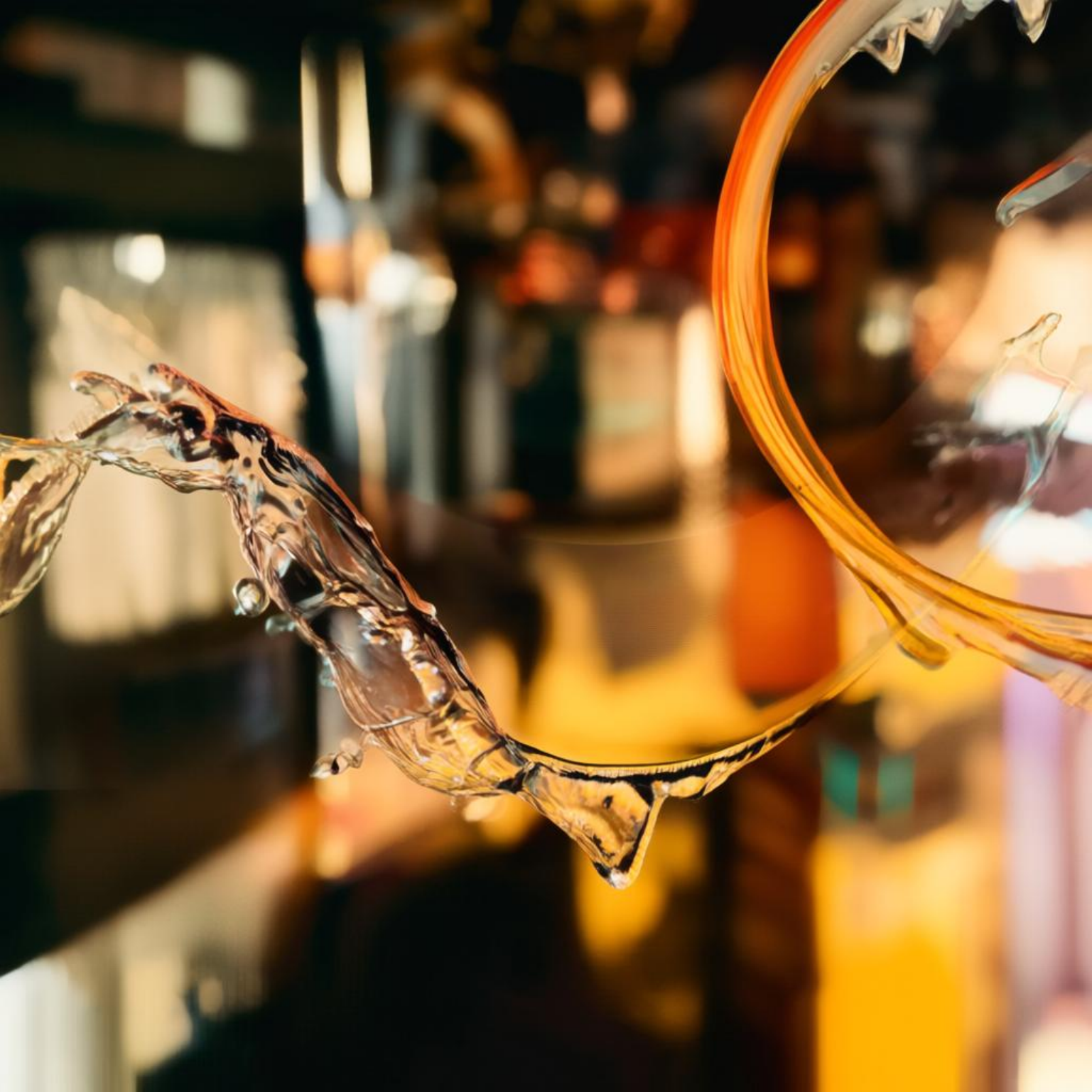} &
		\includegraphics[width=\linewidth]{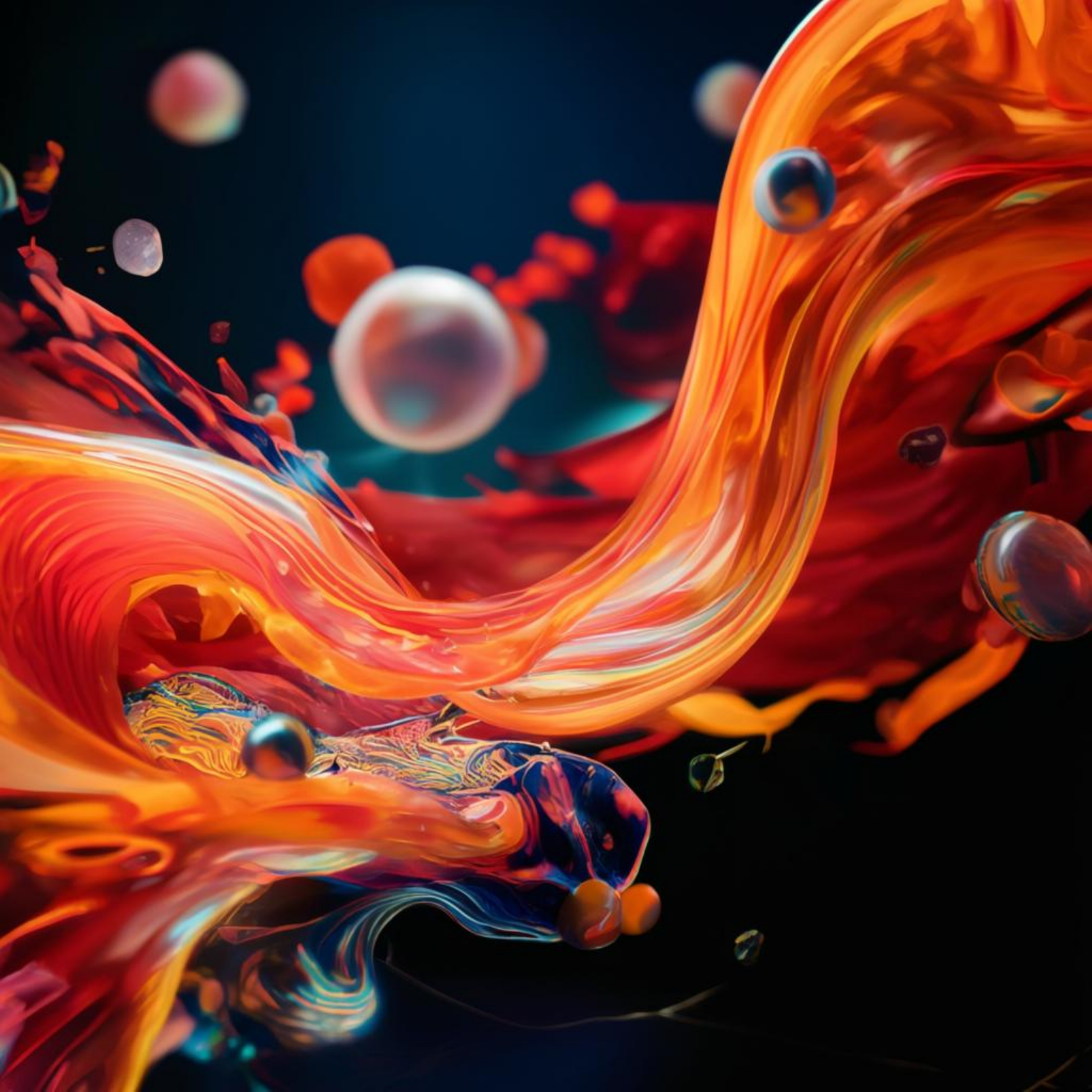} &
		\includegraphics[width=\linewidth]{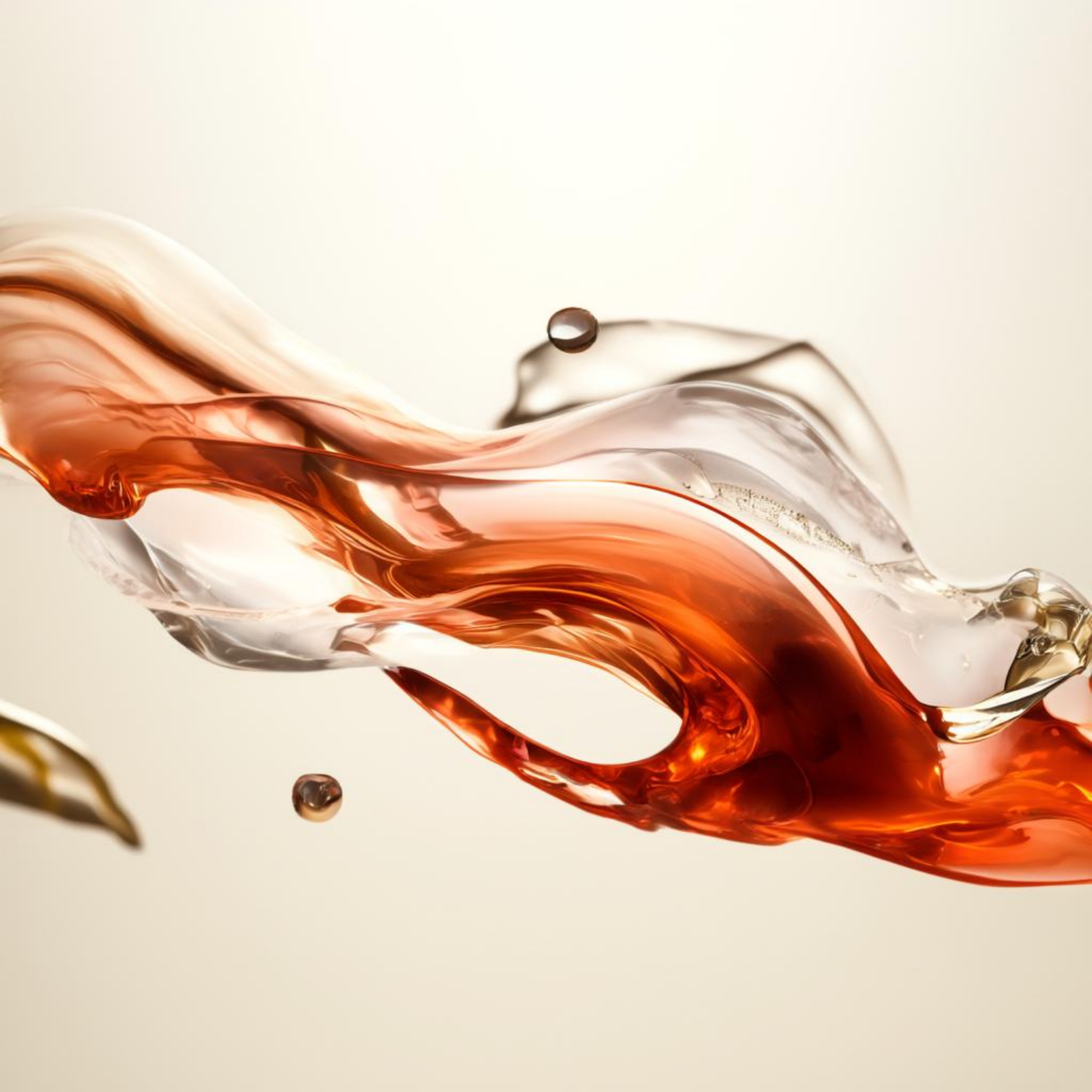} &
		\includegraphics[width=\linewidth]{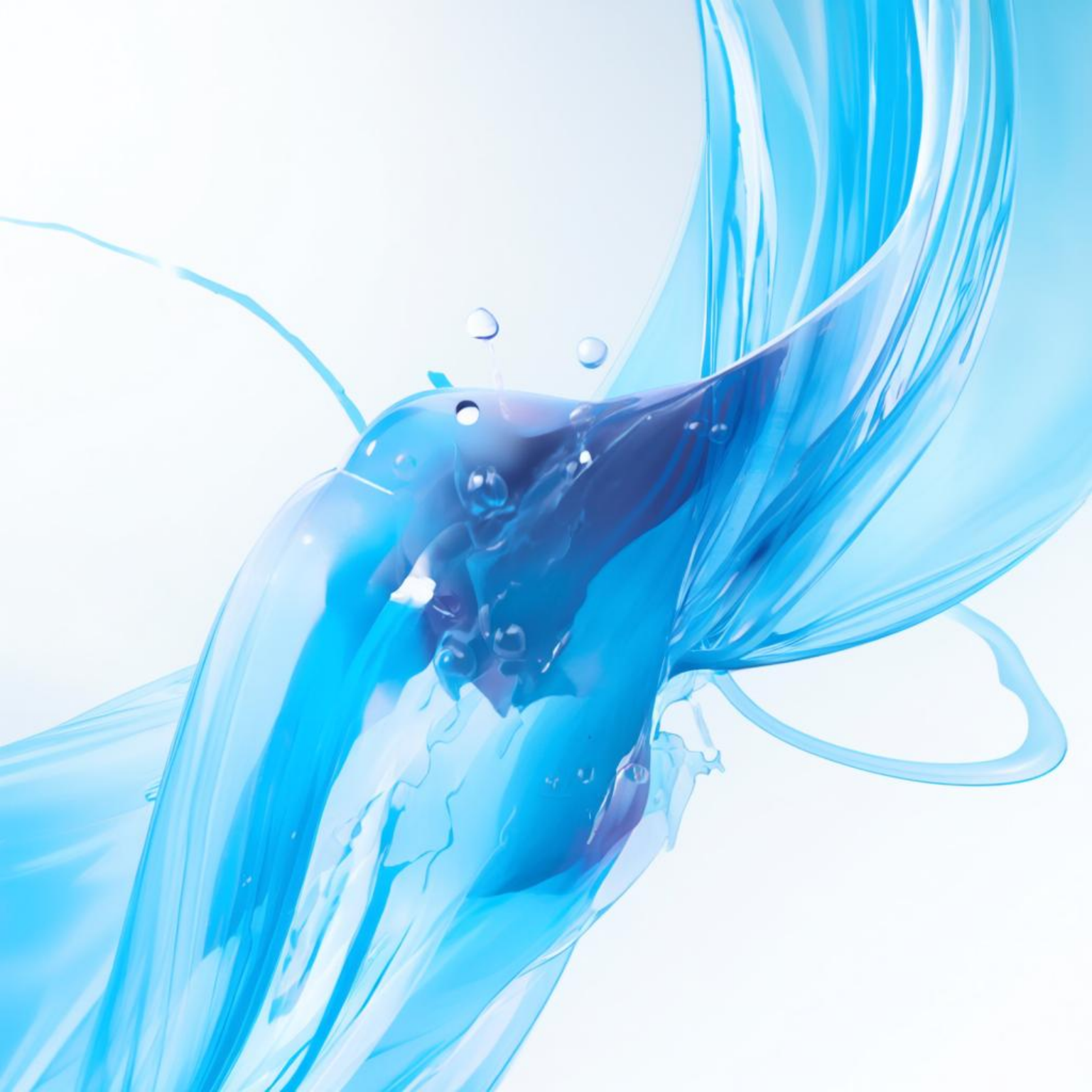} \\
		
		\multicolumn{5}{c}{\small \textit{Prompt: Fluid dynamics in motion}} \\
		\midrule
		
		\includegraphics[width=\linewidth]{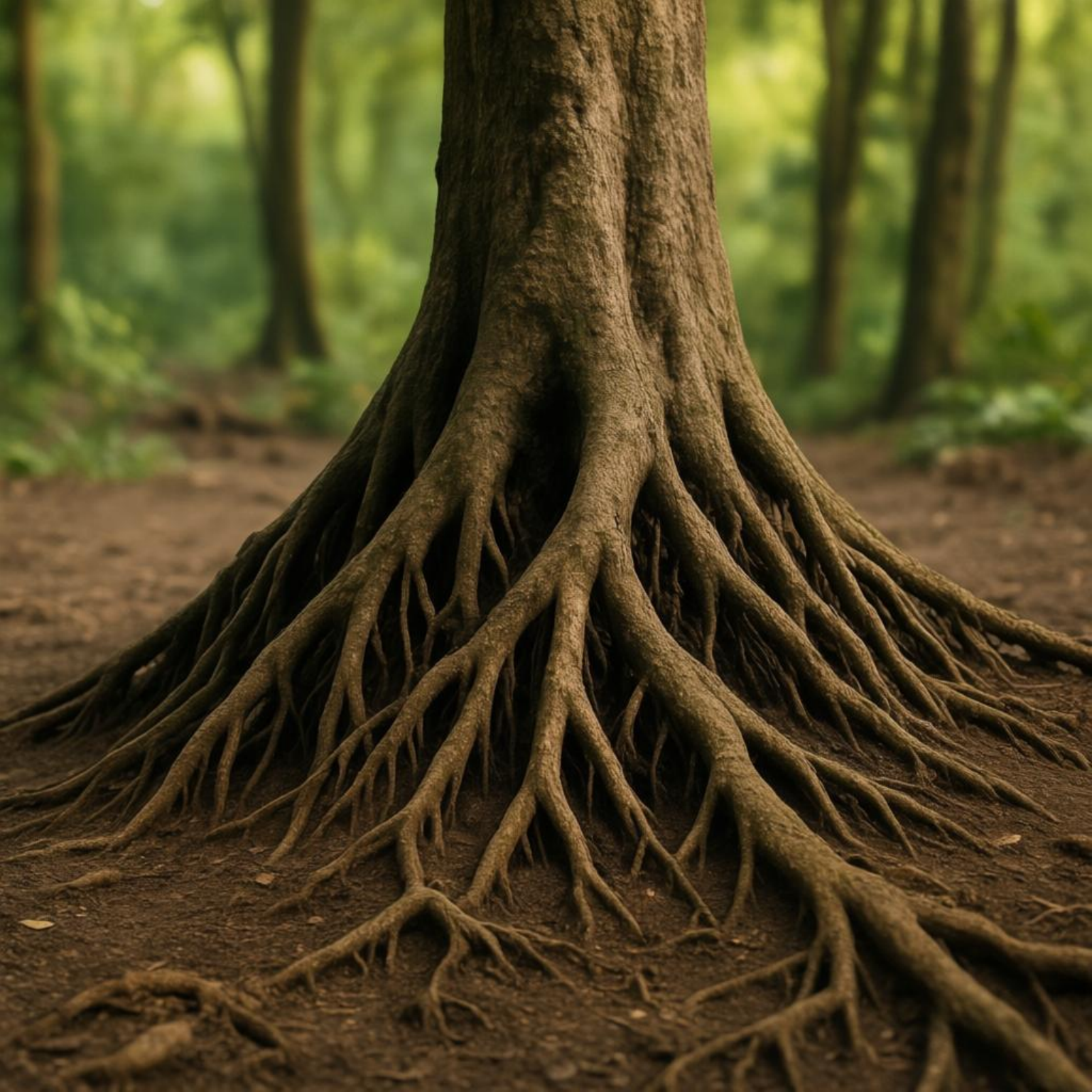} &
		\includegraphics[width=\linewidth]{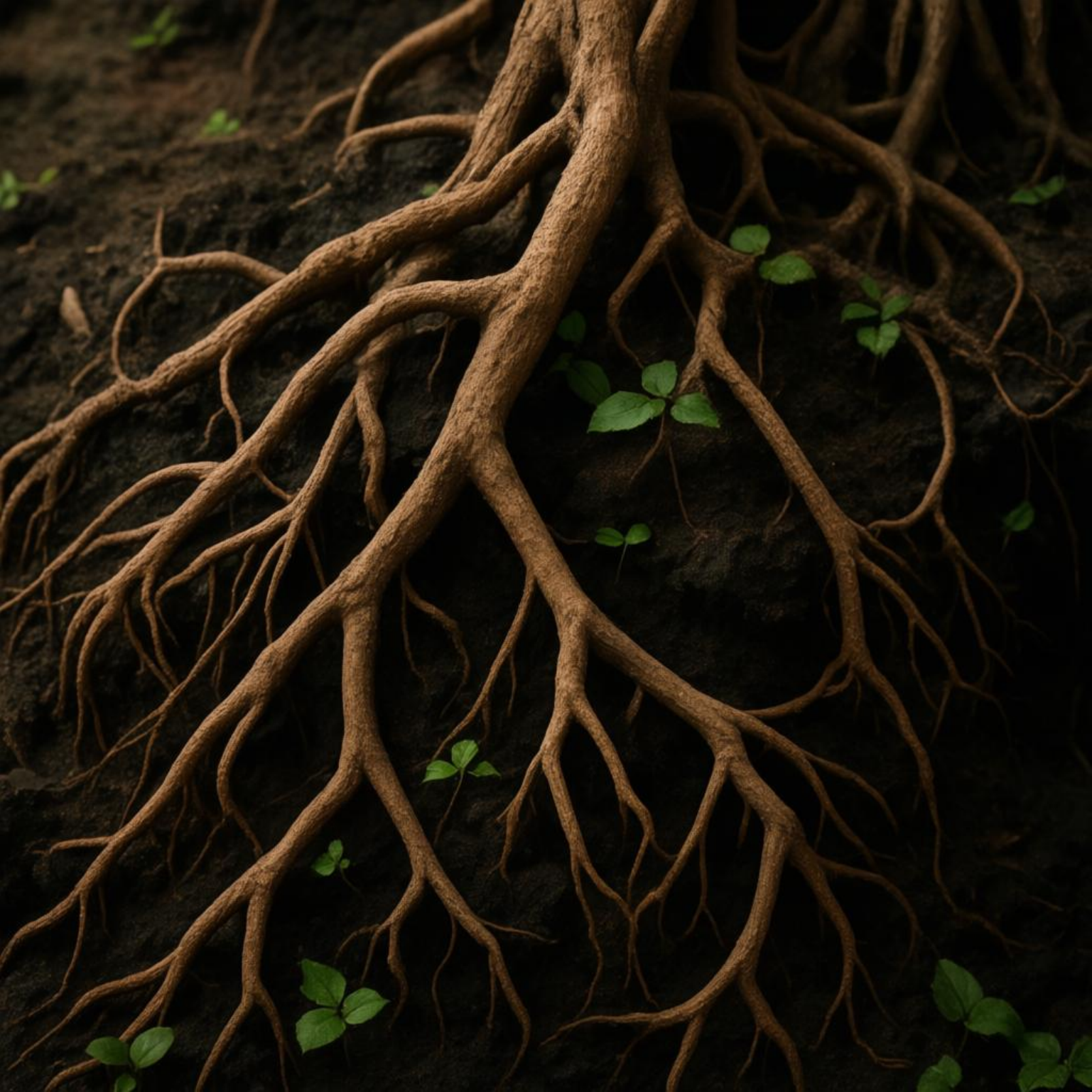} &
		\includegraphics[width=\linewidth]{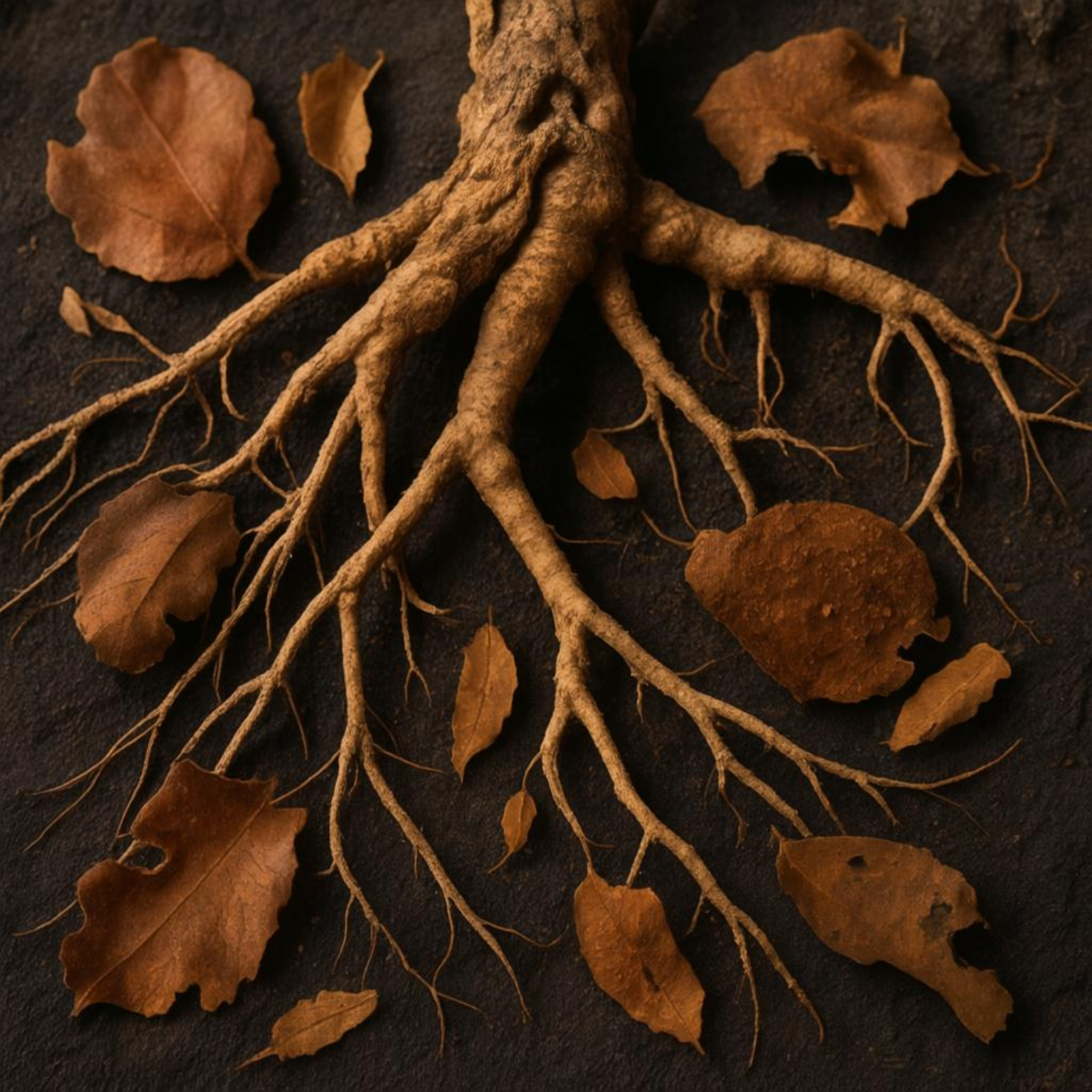} &
		\includegraphics[width=\linewidth]{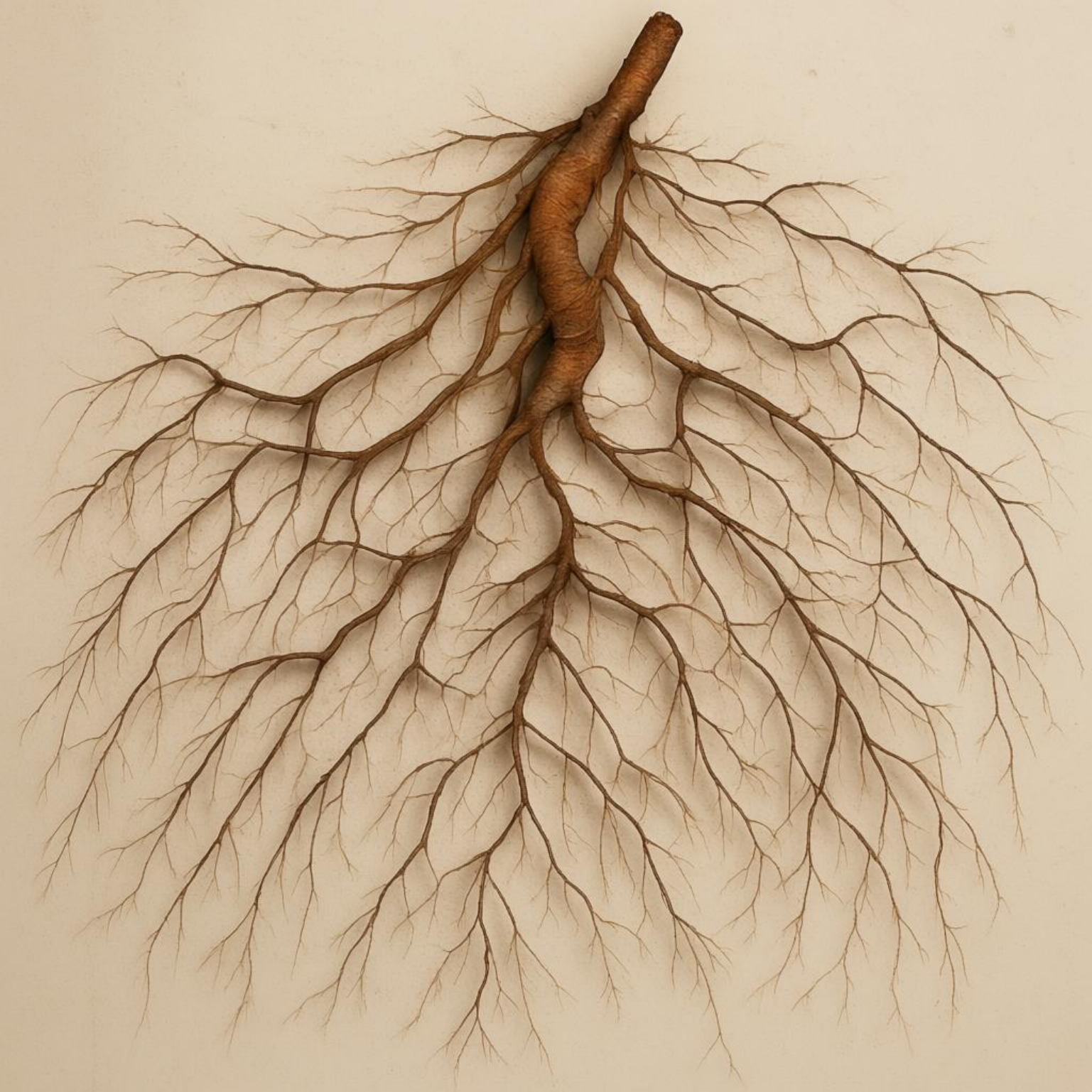} &
		\includegraphics[width=\linewidth]{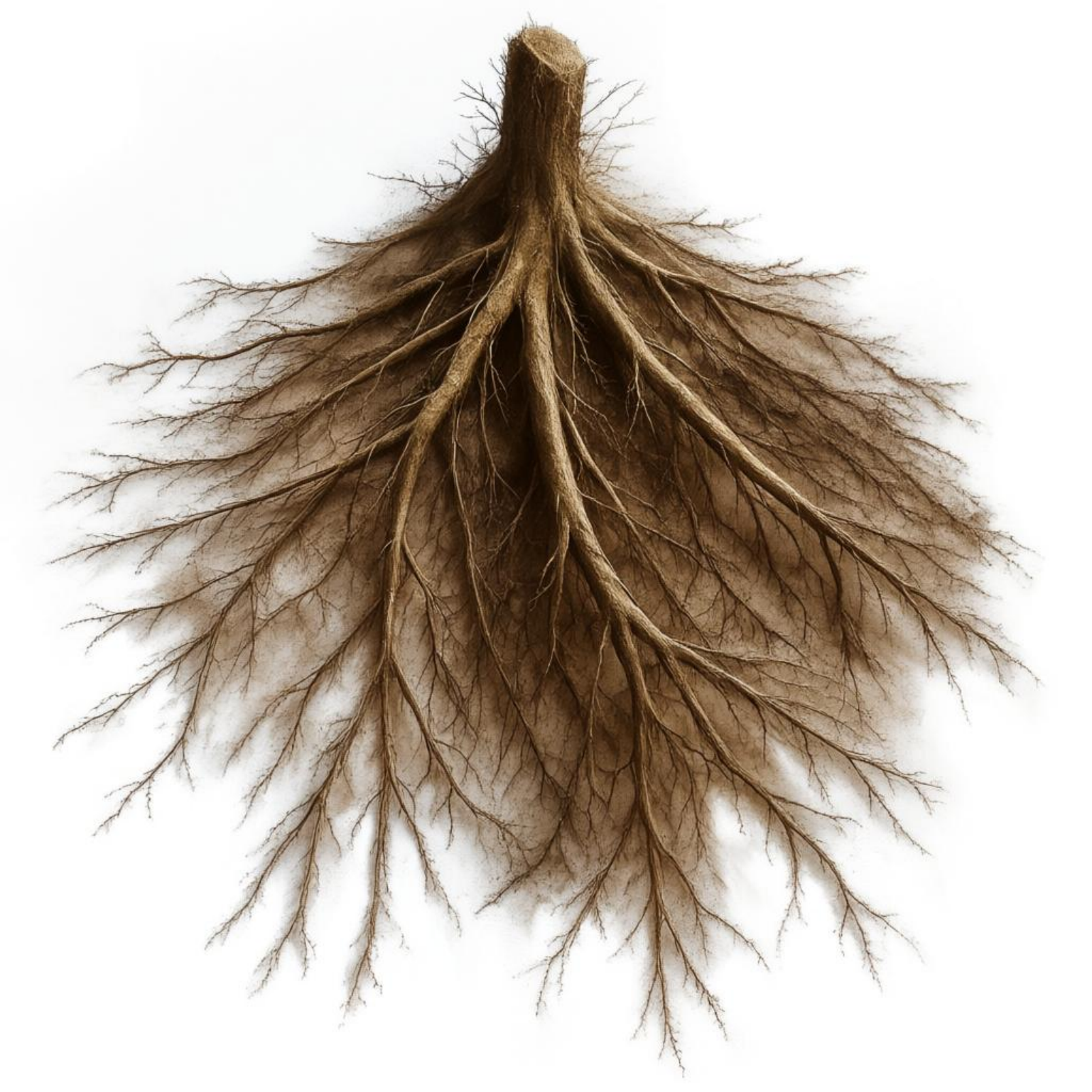} \\
		
		\multicolumn{5}{c}{\small \textit{Prompt: A network of roots}} \\
		
	\end{longtable}
}

\section{Analysis of Reward Hacking}
\label{app:reward_hacking}

This section aims to dissect the ``Reward Hacking'' phenomenon prevalent in inference-time optimization methods and demonstrate the resistance of SES to such hacking through cross-reward evaluation.

\subsection{Out-of-Distribution Reward Hacking Problem}
\label{sec:ood}

In the field of inference-time optimization for generative models, Out-of-Distribution (OOD) Reward Hacking refers to the phenomenon where the optimization algorithm converges to a specific class of samples $x_\text{hack}$. Although these samples achieve extremely high scores on the reward model $\mathcal{R}$, they deviate from the high probability density regions of the training data distribution $p_\text{data}(x)$. Intuitively, these samples are often replete with high-frequency noise or artifacts imperceptible or incomprehensible to human vision, essentially constituting \textit{Adversarial Examples} against the reward model.

In inference-time optimization for diffusion models, gradient-guided methods represented by DNO \cite{tang2024inference} update images or noise by computing $\nabla_{\mathbf{x}} \mathcal{R}(\mathbf{x})$. However, existing reward models are generally hypersensitive to non-robust high-frequency features. During gradient ascent, the optimizer tends to traverse rapidly along the steepest directions of high-frequency noise rather than adjusting the global semantics (low-frequency directions) of the image, thereby causing the generation trajectory to detach quickly from the natural image manifold.

Although DNO attempts to introduce a probabilistic regularization term based on high-dimensional Gaussian concentration inequalities to alleviate this issue, our empirical research indicates that this strategy does not eradicate Reward Hacking. As shown in Figure \ref{fig:dno_artifacts}, with increasing optimization steps, DNO introduces unnatural noise patterns in high-frequency texture regions to pursue higher reward scores. These high-frequency perturbations numerically ``exploit'' the blind spots of the reward model but manifest as severe image collapse in terms of perceptual quality. Given the instability of such methods regarding generation quality, we focus on methods yielding more robust generative distributions in our subsequent main baseline comparisons.

{
    \setlength{\tabcolsep}{2pt}
    \begin{longtable}{ *{7}{ >{\centering\arraybackslash}m{0.134\linewidth} } }
		\caption{Qualitative analysis of OOD reward hacking in DNO. We visualize the image generation trajectory as the number of optimization steps increases (Step 0 to 30), targeting the PickScore reward model.} \label{fig:dno_artifacts} \\
		
		\toprule
		\centering \textbf{SDXL} & 
		\centering \textbf{Step=5} & 
		\centering \textbf{Step=10} & 
		\centering \textbf{Step=15} & 
		\centering \textbf{Step=20} & 
		\centering \textbf{Step=25} & 
		\centering \textbf{Step=30} \tabularnewline
		\midrule
		\endfirsthead

        \toprule
        \textbf{SDXL} & \textbf{Step=5} & \textbf{Step=10} & \textbf{Step=15} & \textbf{Step=20} & \textbf{Step=25} & \textbf{Step=30} \\
        \midrule
        \endhead

		\bottomrule
		\endfoot

		\includegraphics[width=\linewidth]{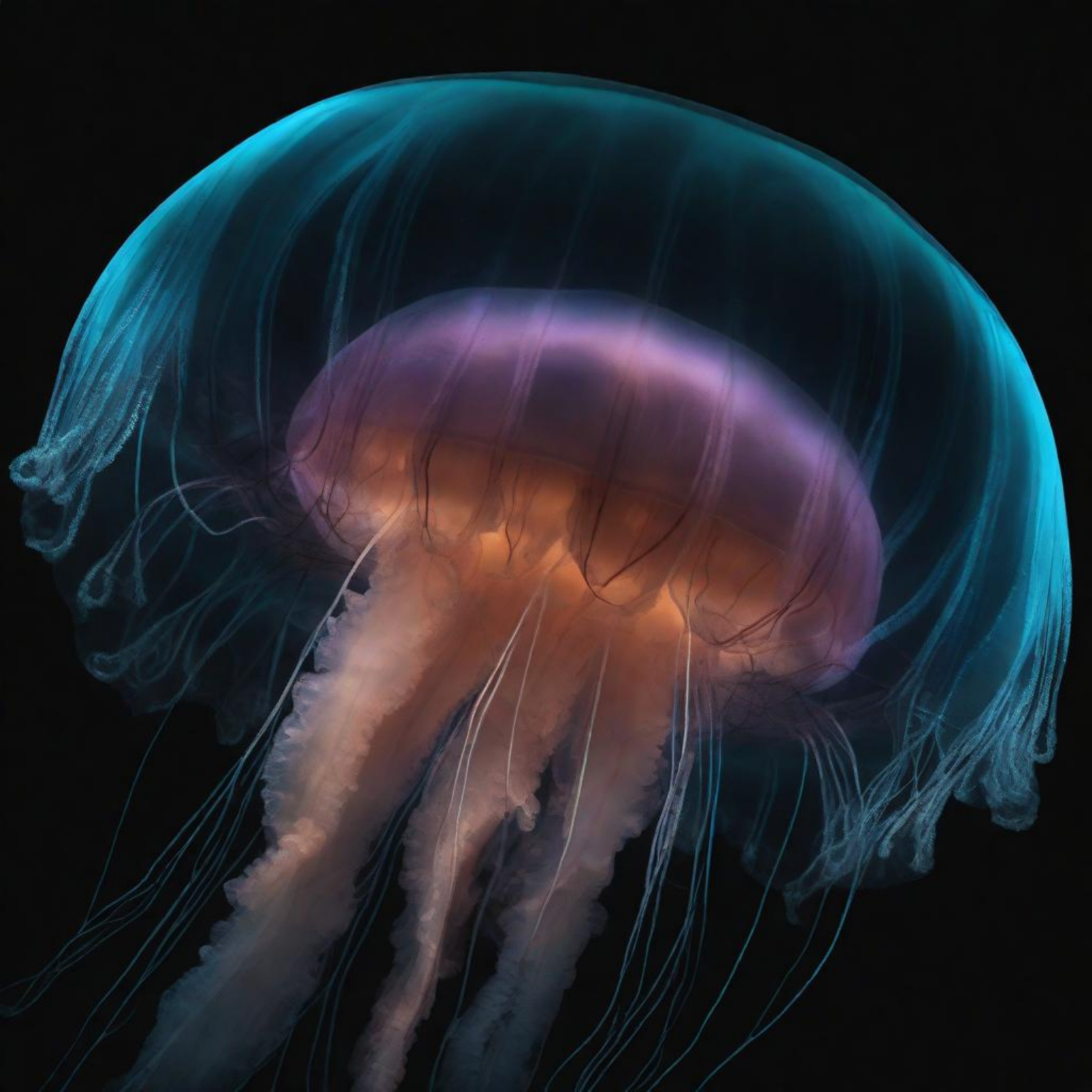} &
		\includegraphics[width=\linewidth]{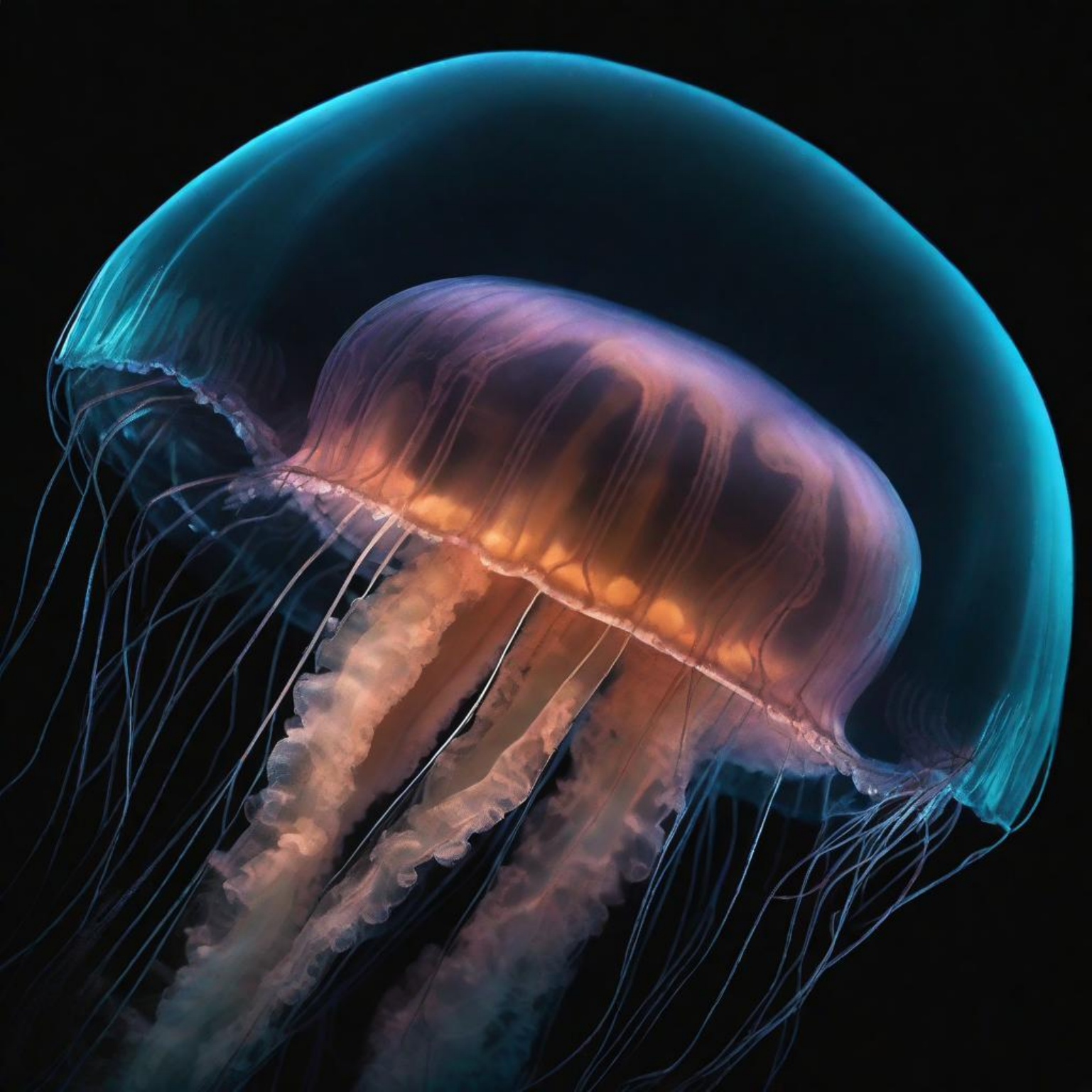} &
		\includegraphics[width=\linewidth]{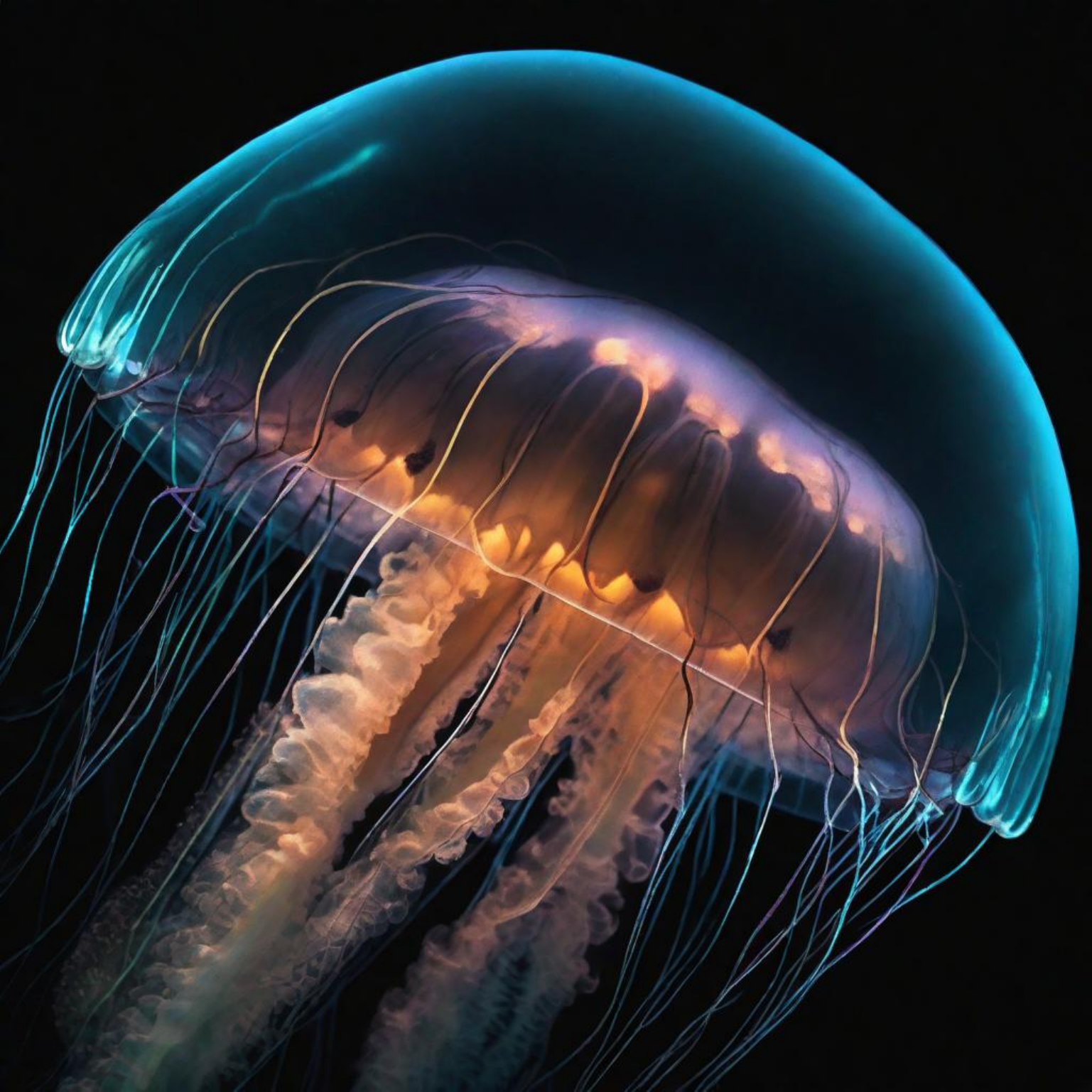} &
		\includegraphics[width=\linewidth]{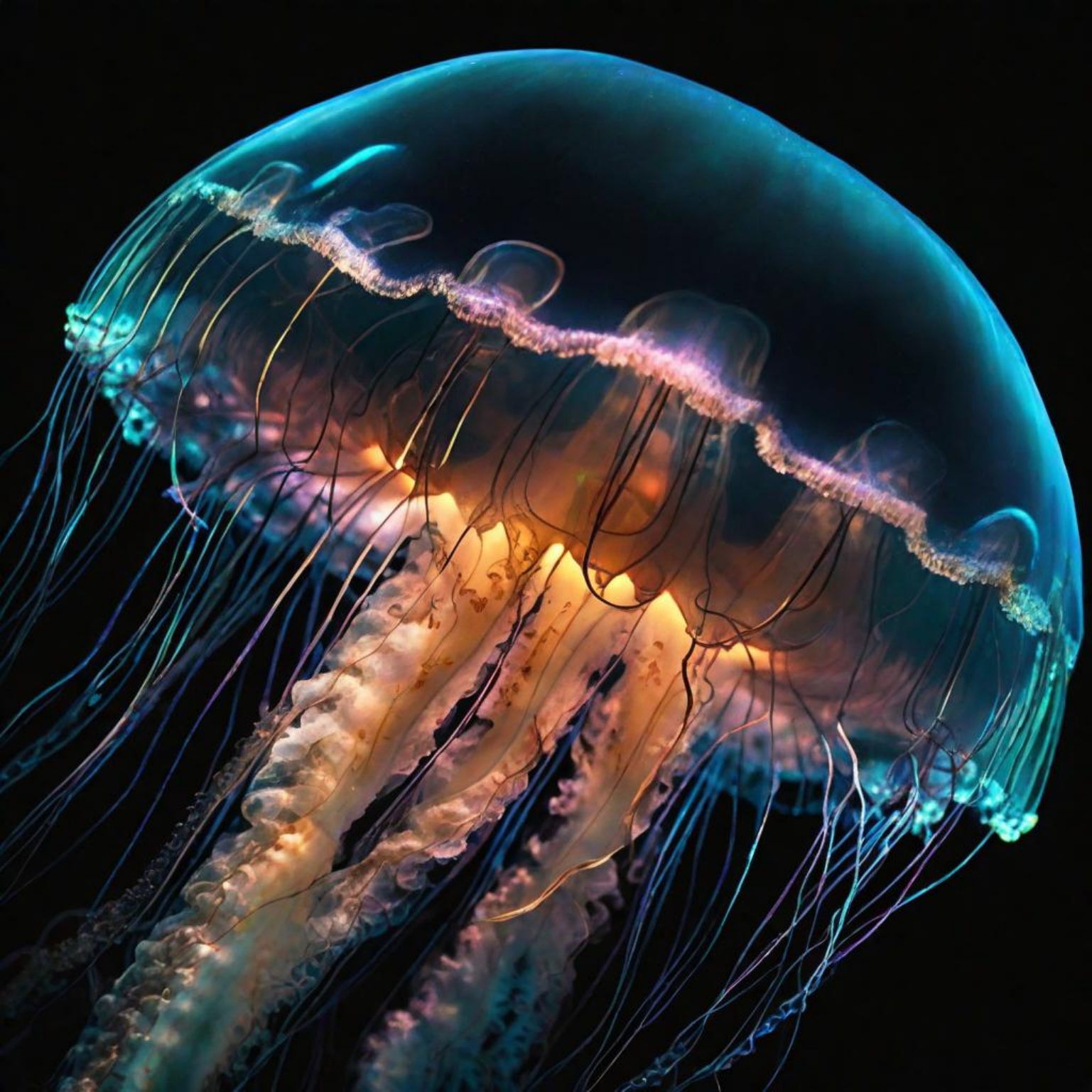} &
		\includegraphics[width=\linewidth]{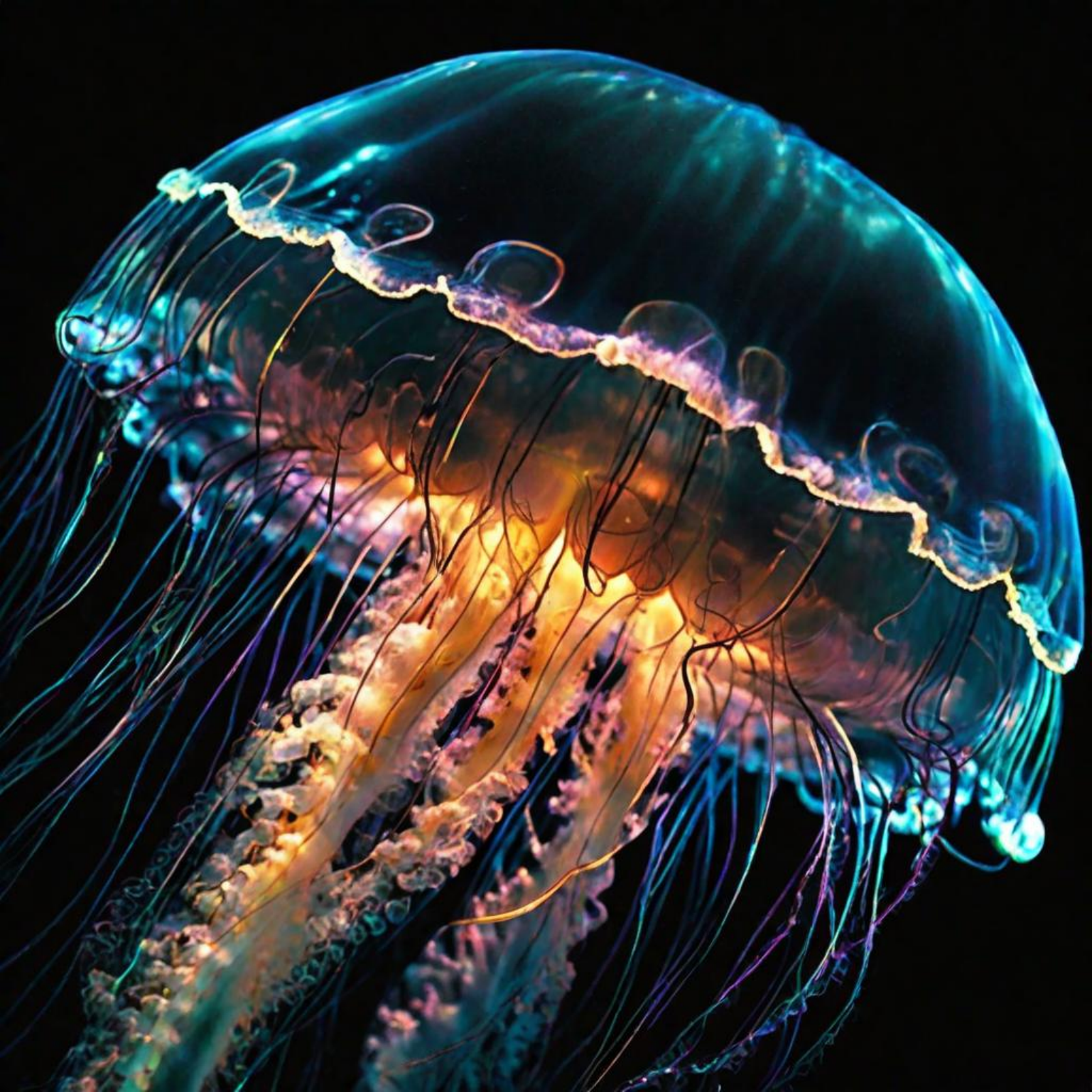} &
		\includegraphics[width=\linewidth]{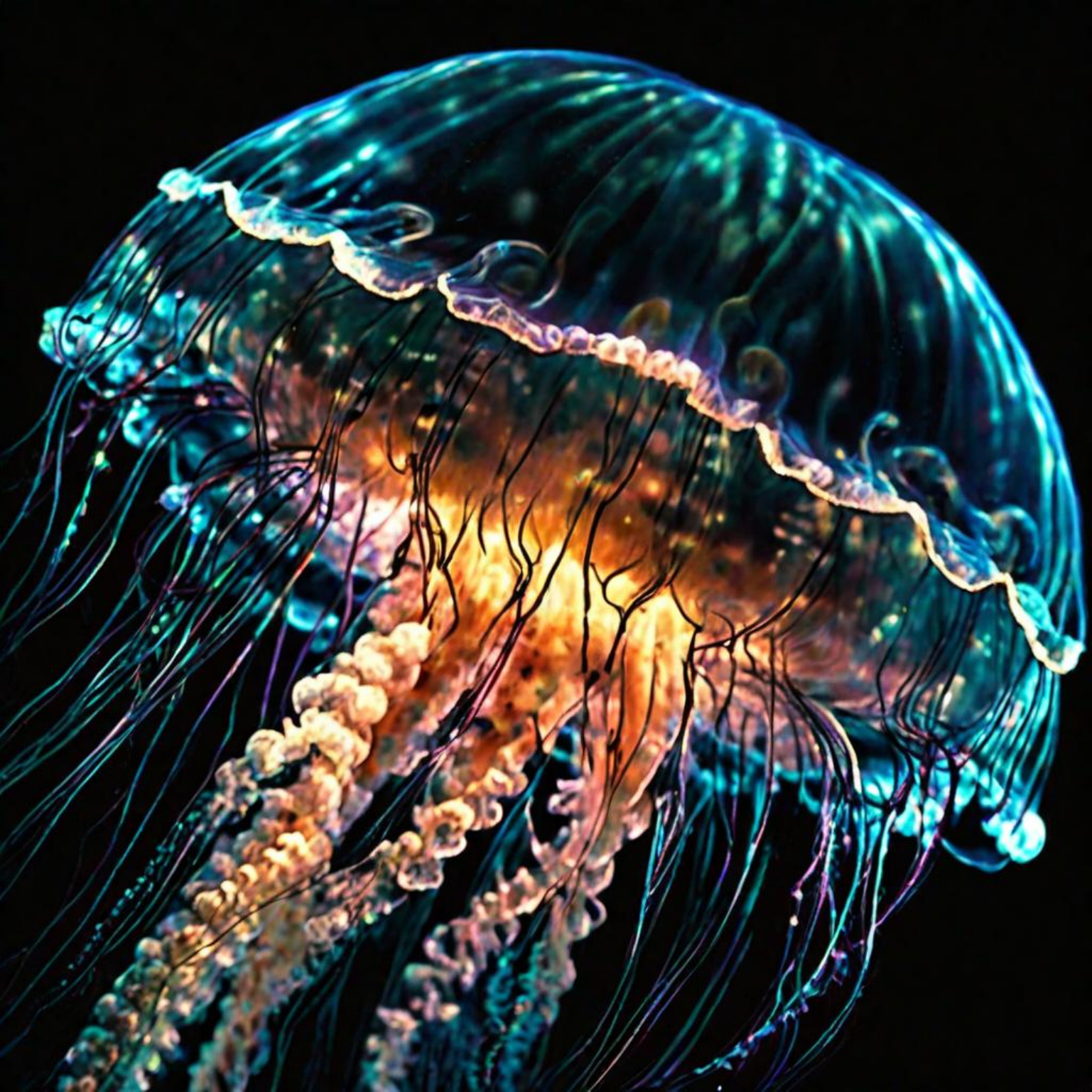} &        
		\includegraphics[width=\linewidth]{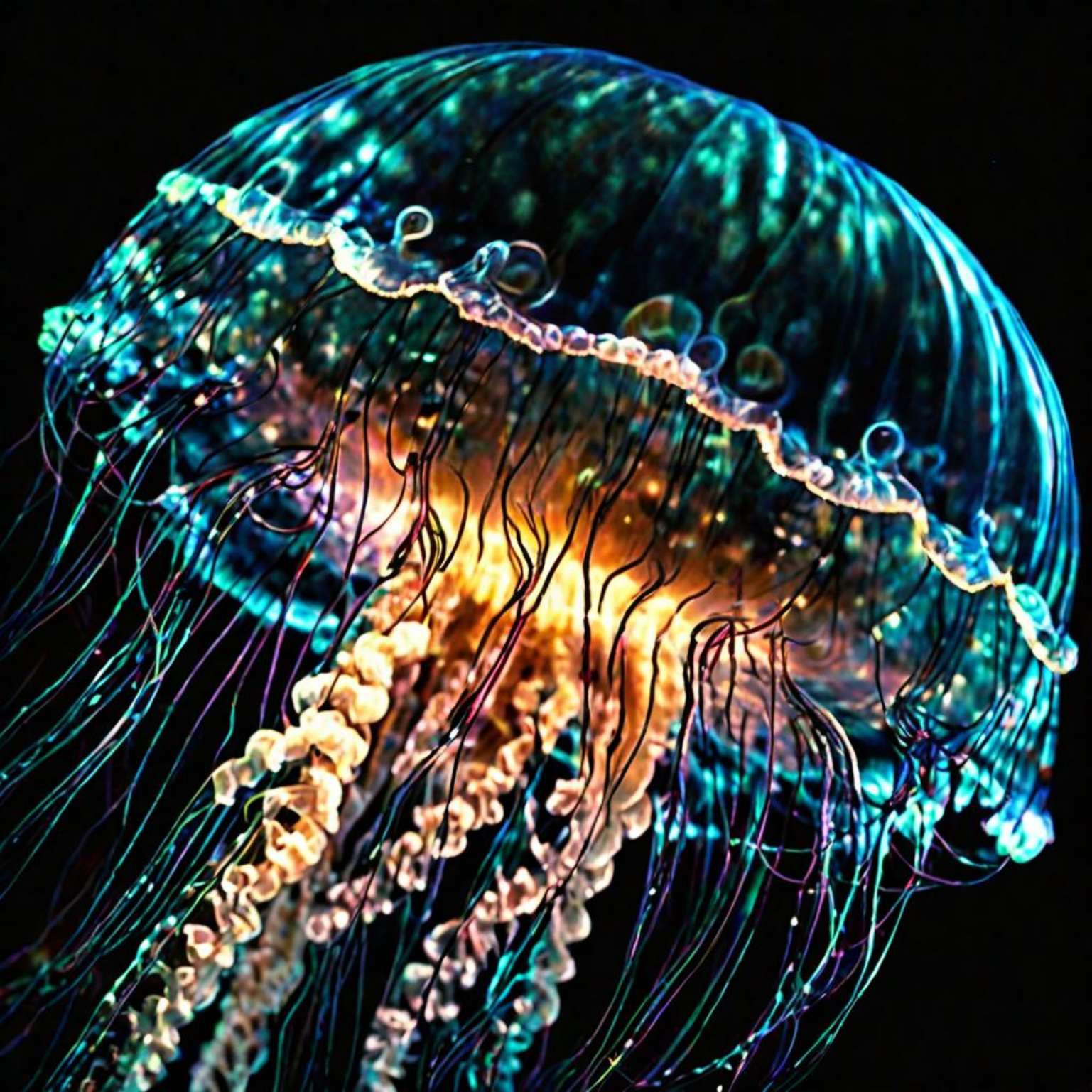} \\

        \small Score: 23.21 & \small Score: 23.57 & \small Score: 24.34 & 
        \small Score: 24.51 & \small Score: 24.68 & \small Score: 25.04 & \small Score: 24.93 \\

		\midrule

		\includegraphics[width=\linewidth]{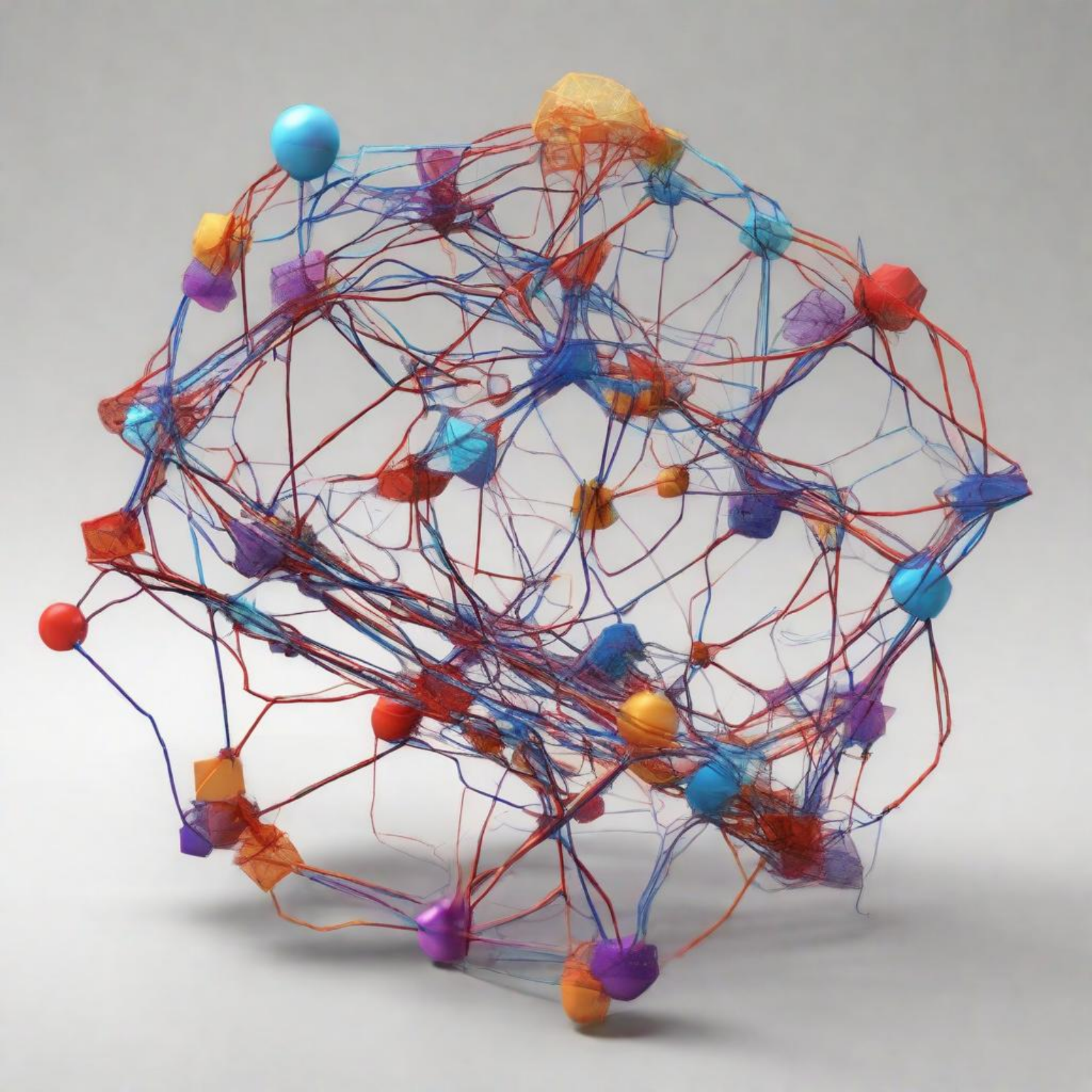} &
		\includegraphics[width=\linewidth]{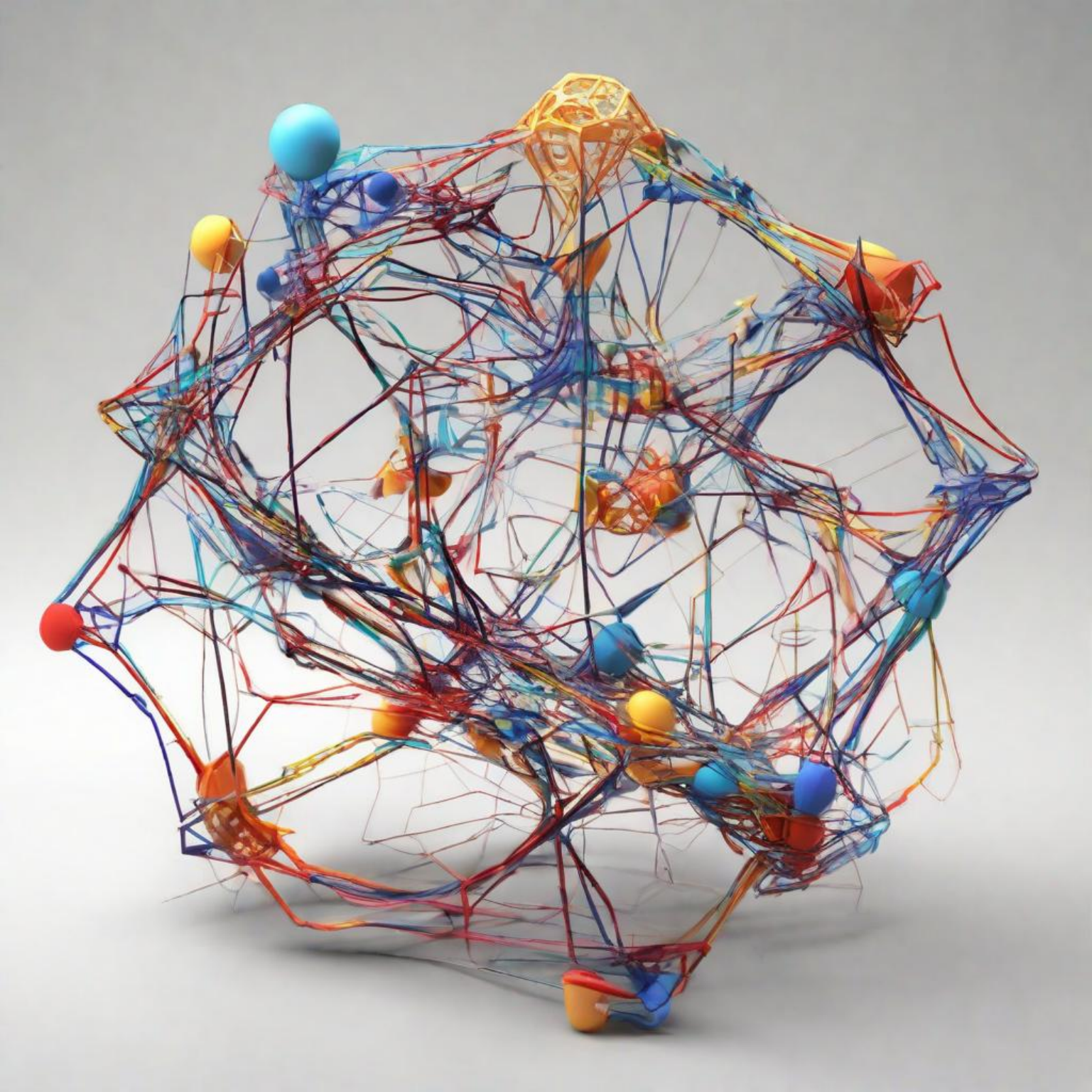} &
		\includegraphics[width=\linewidth]{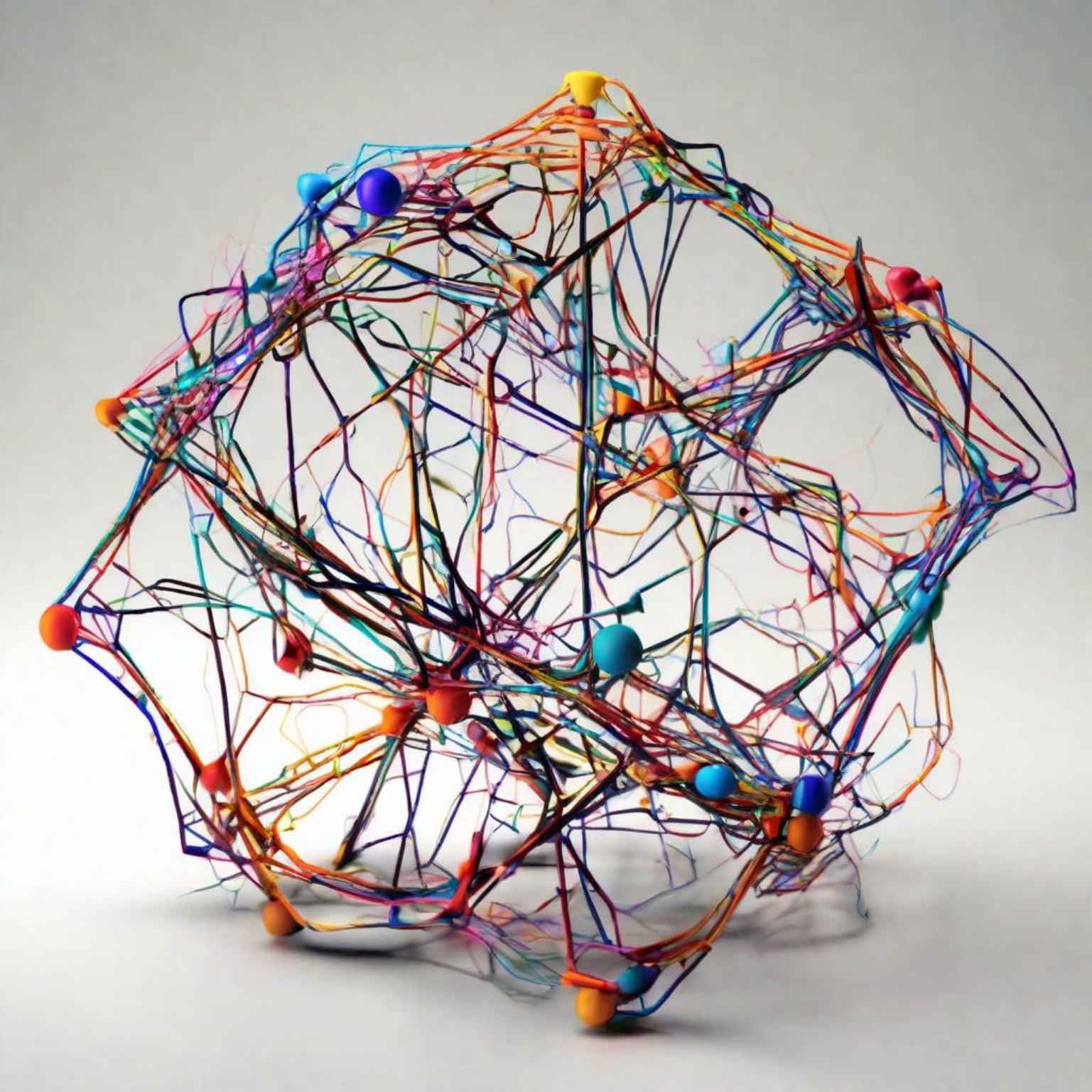} &
		\includegraphics[width=\linewidth]{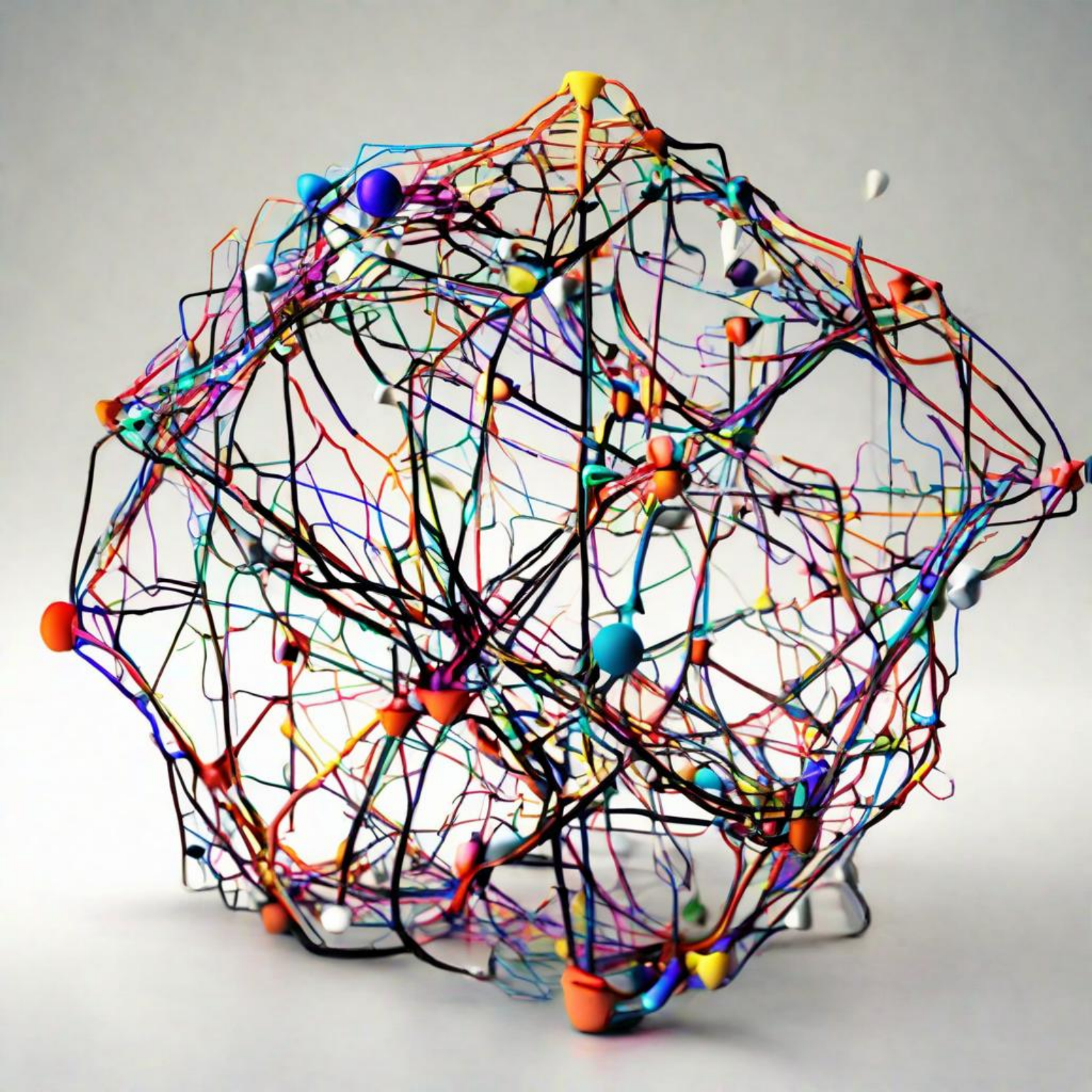} &
		\includegraphics[width=\linewidth]{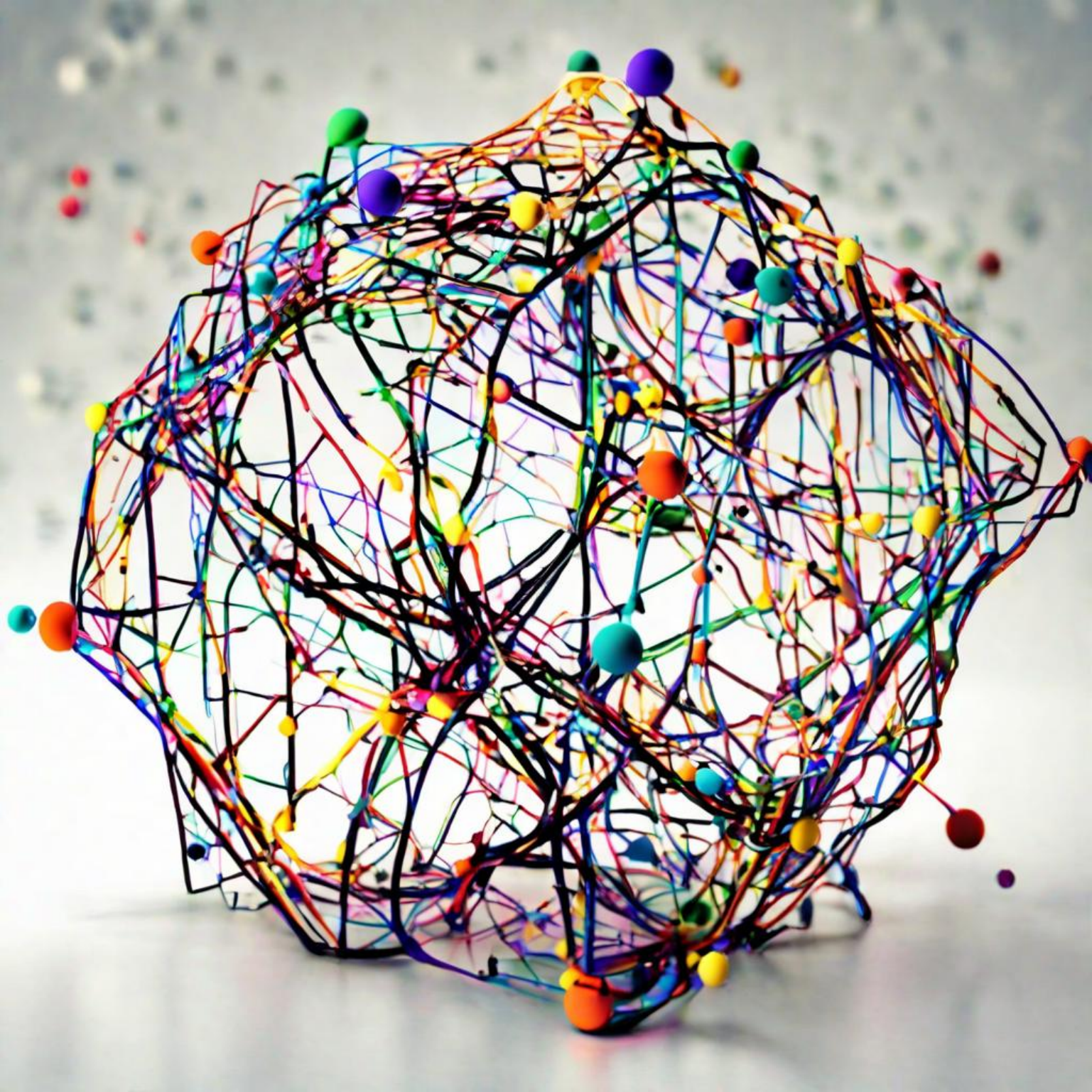} &
		\includegraphics[width=\linewidth]{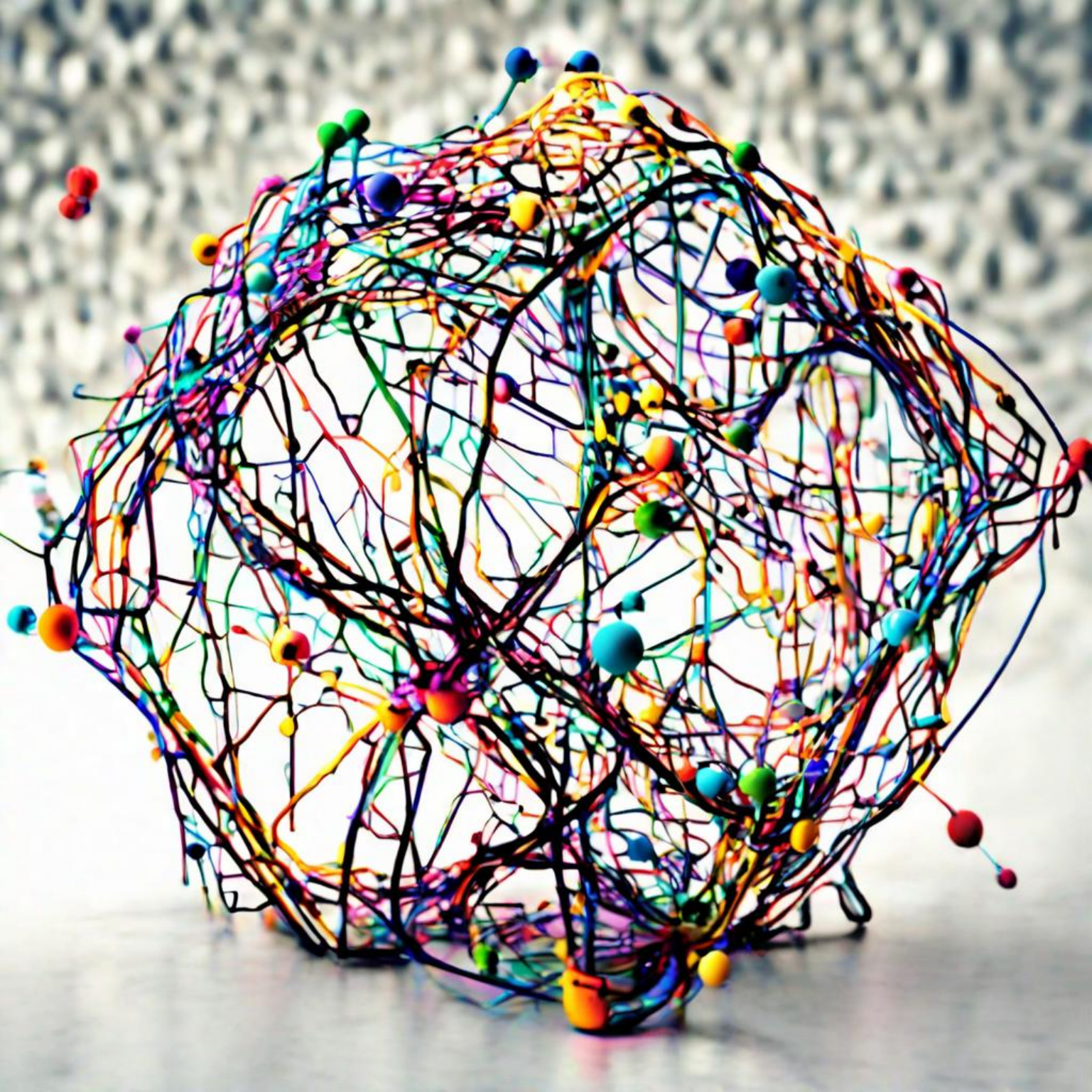} &        
		\includegraphics[width=\linewidth]{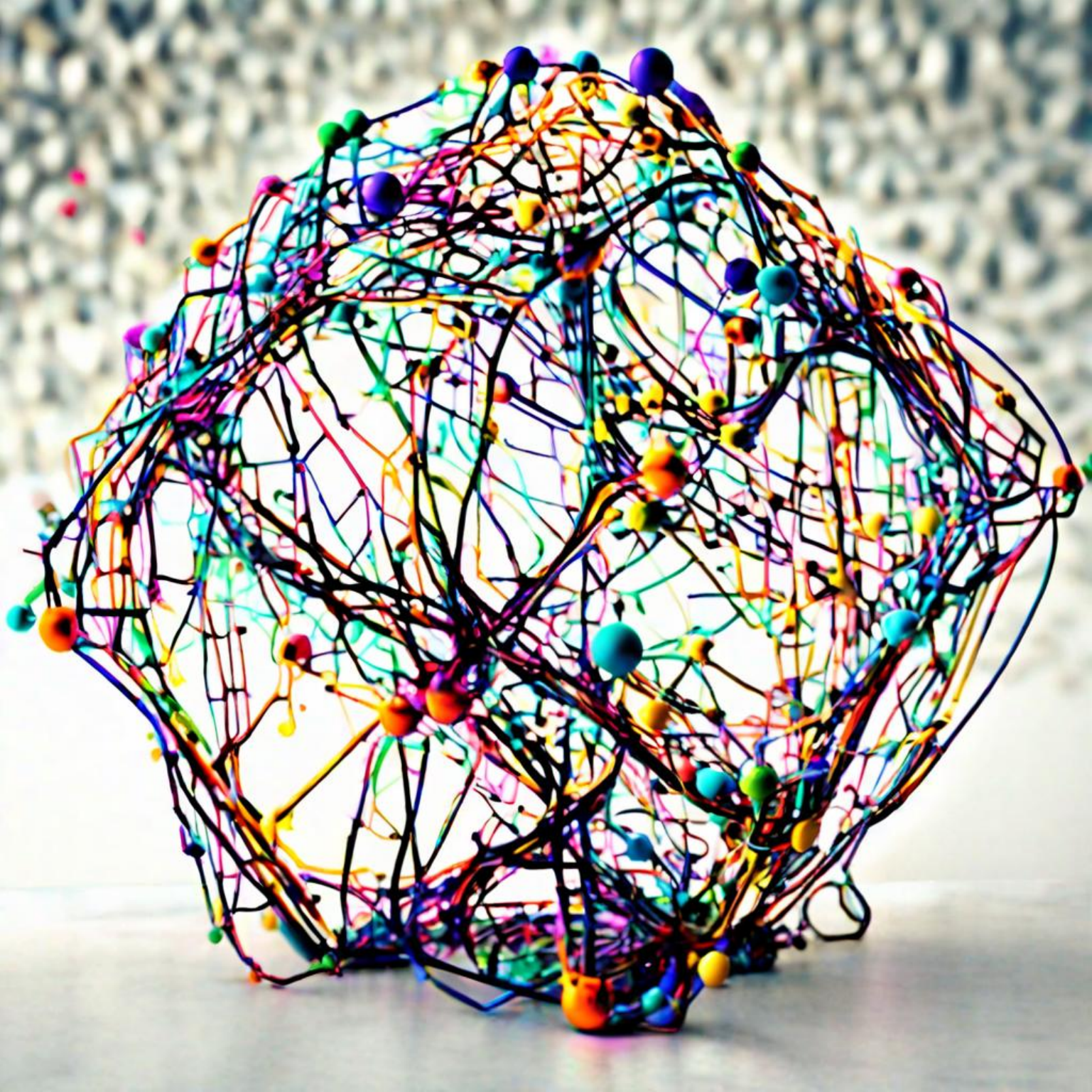} \\

        \small Score: 21.81 & \small Score: 23.63 & \small Score: 23.82 & 
        \small Score: 23.82 & \small Score: 23.96 & \small Score: 24.22 & \small Score: 25.07 \\
		\midrule
		
		\includegraphics[width=\linewidth]{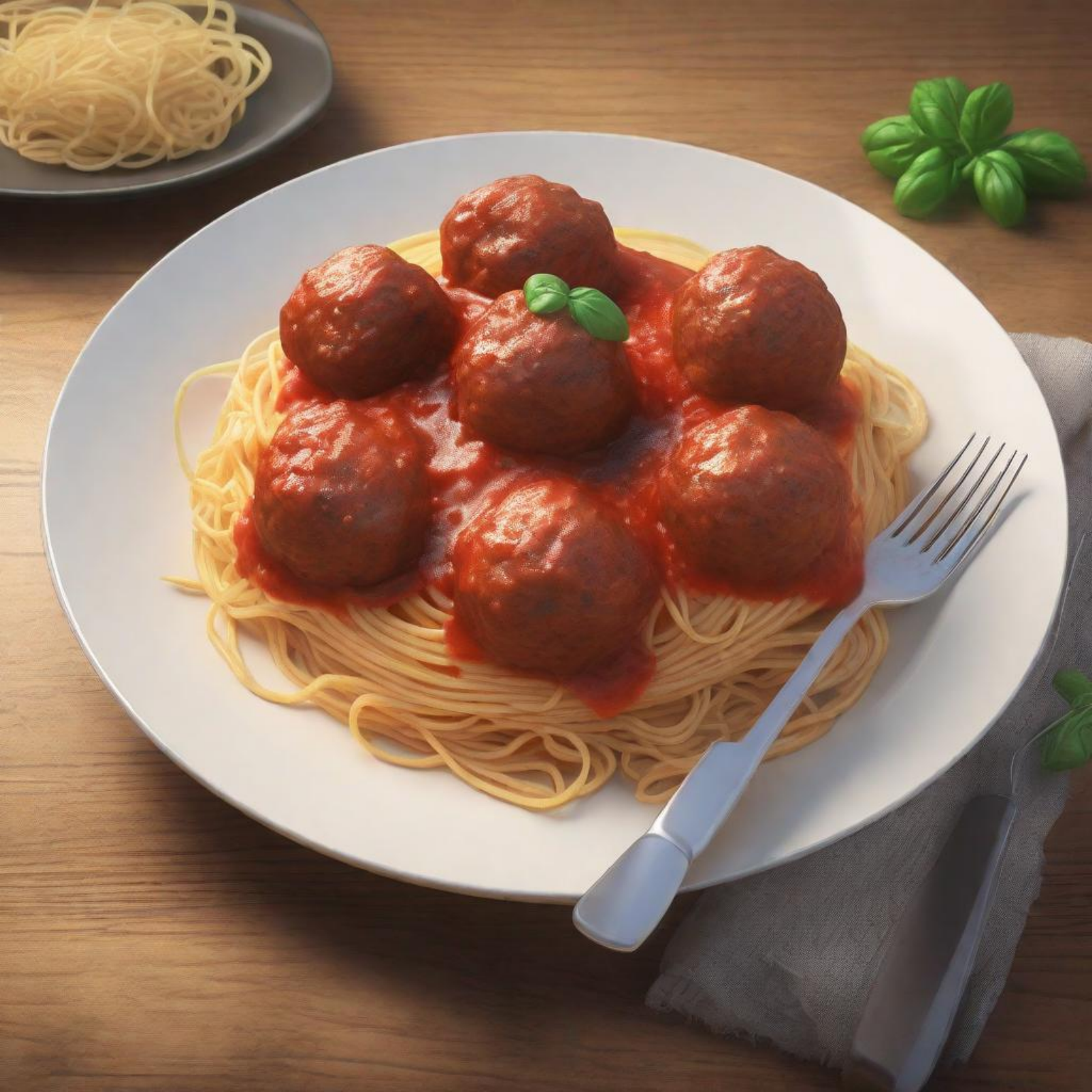} &
		\includegraphics[width=\linewidth]{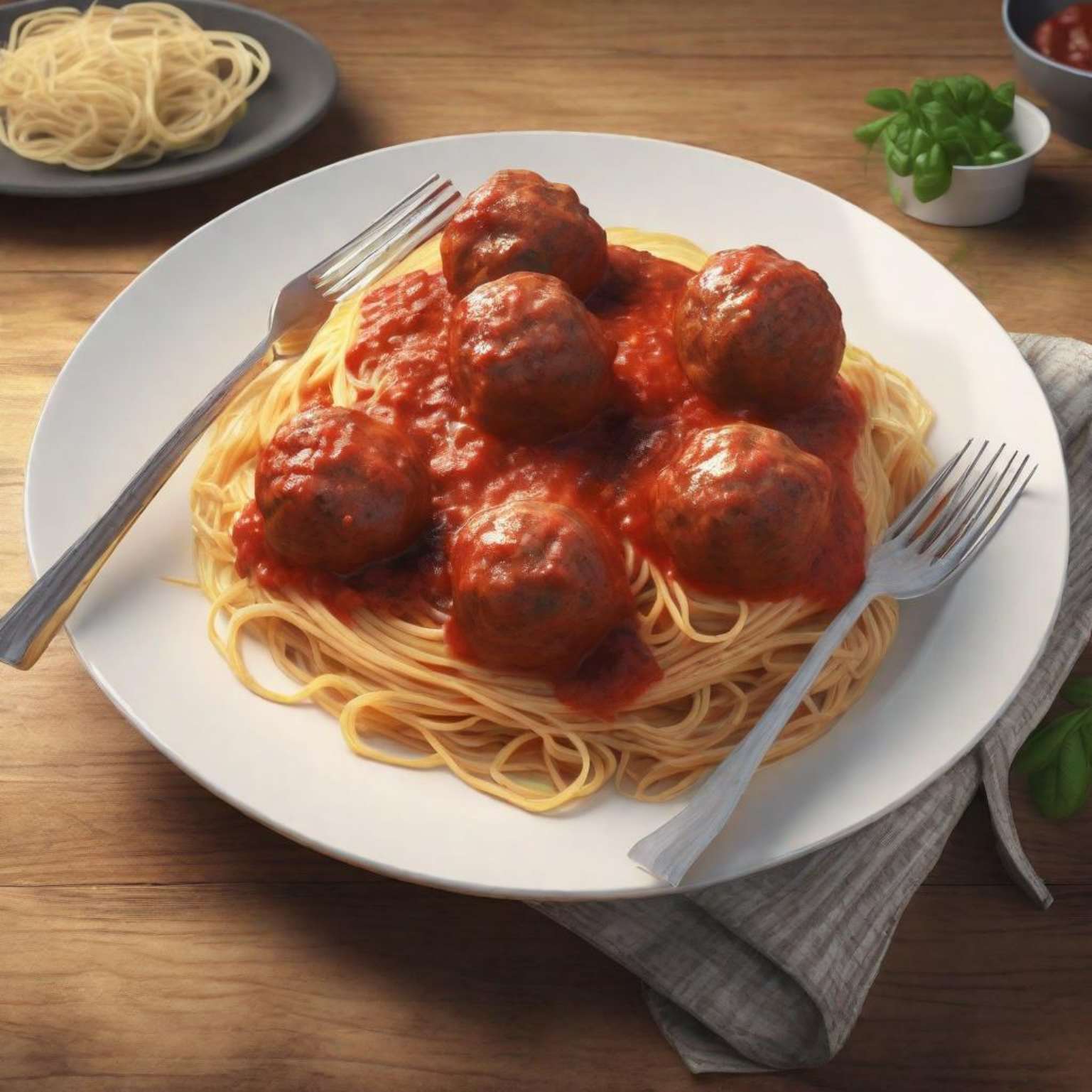} &
		\includegraphics[width=\linewidth]{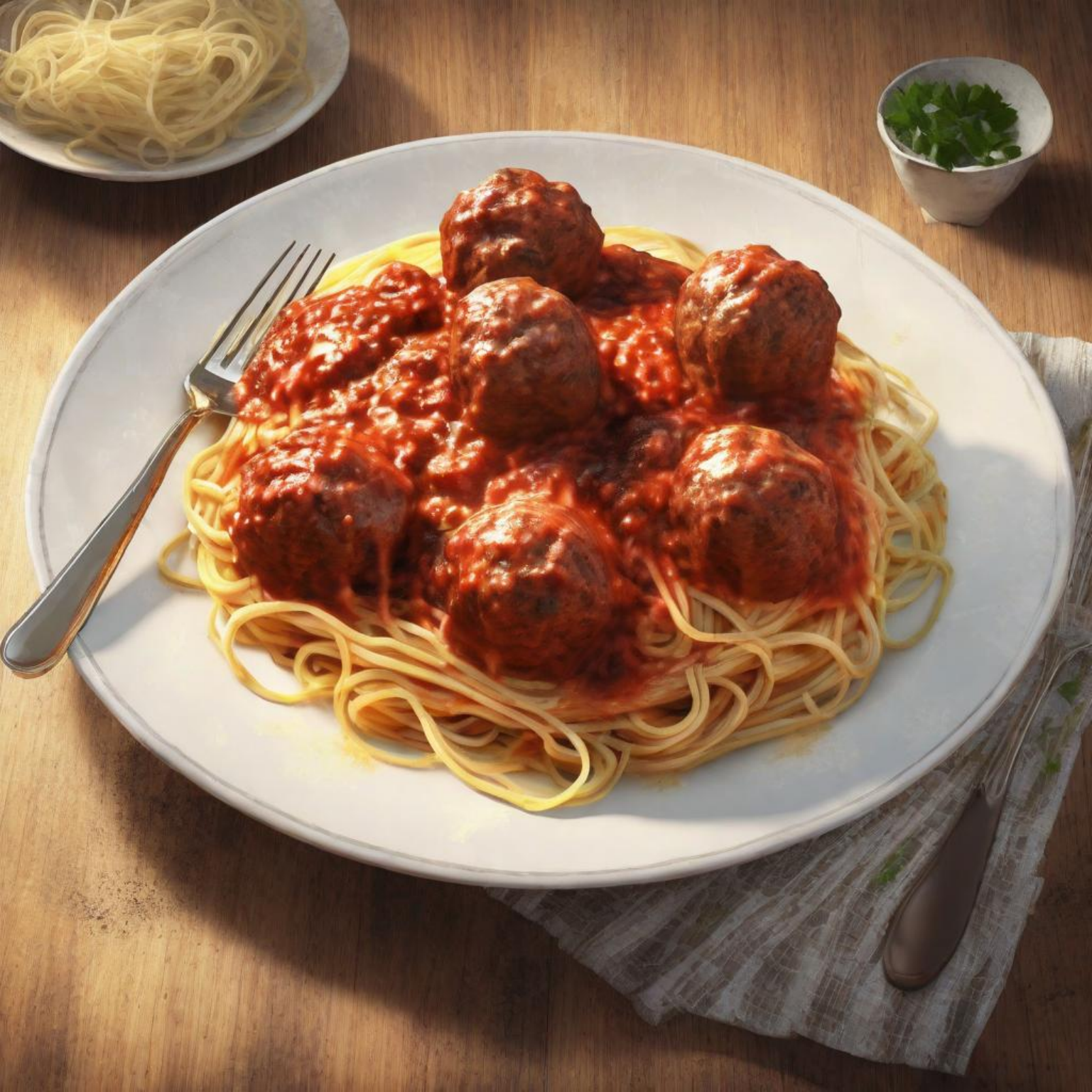} &
		\includegraphics[width=\linewidth]{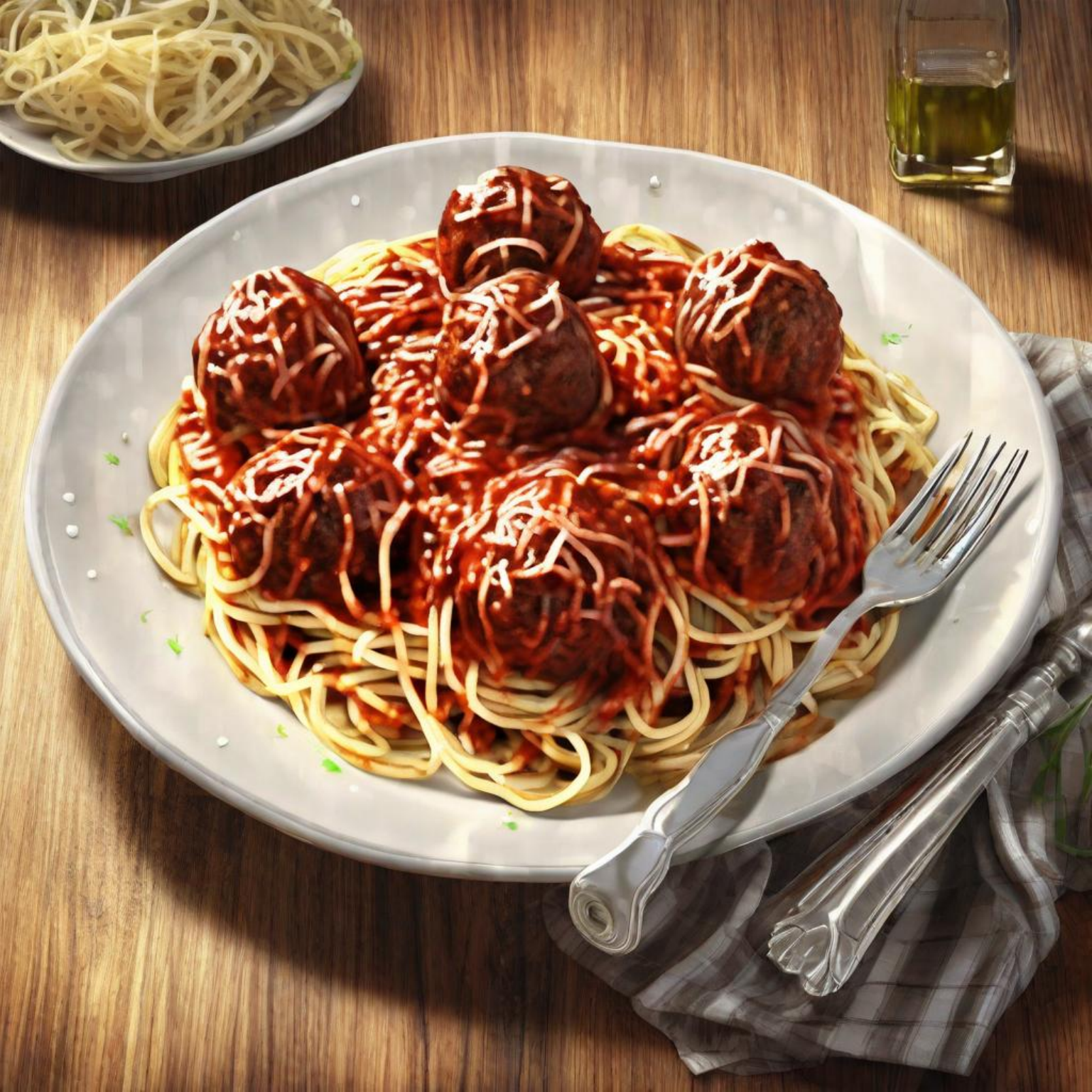} &
		\includegraphics[width=\linewidth]{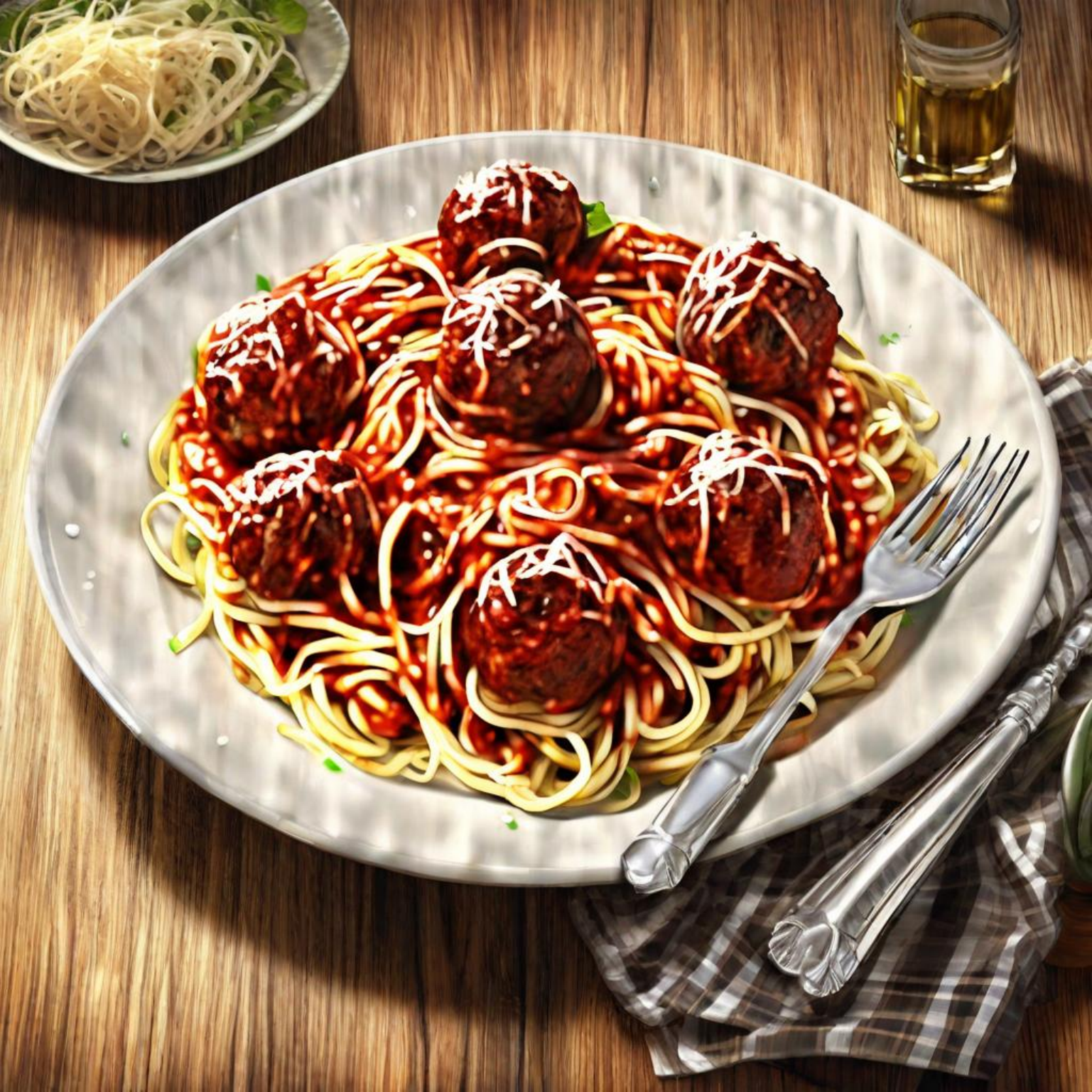} &
		\includegraphics[width=\linewidth]{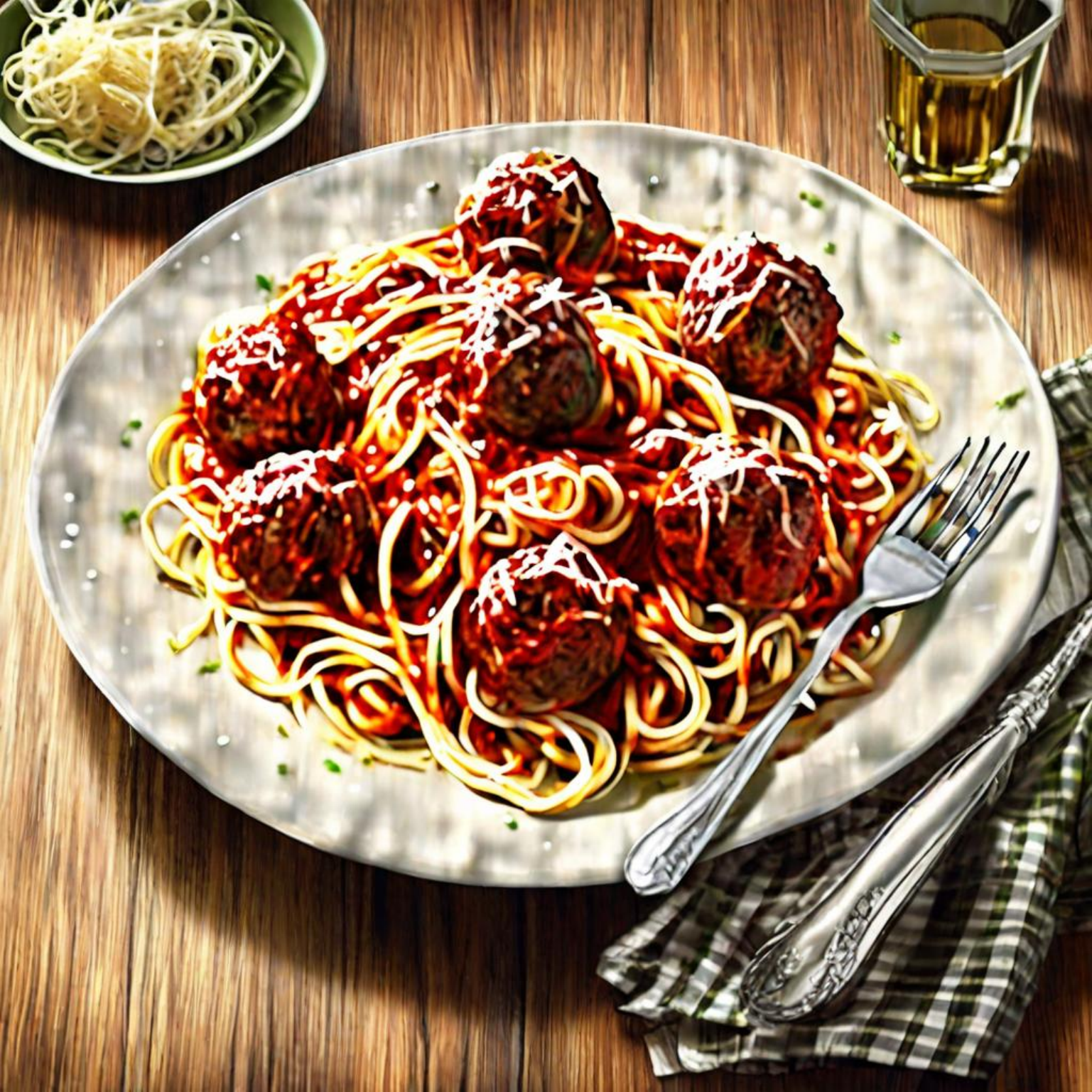} &        
		\includegraphics[width=\linewidth]{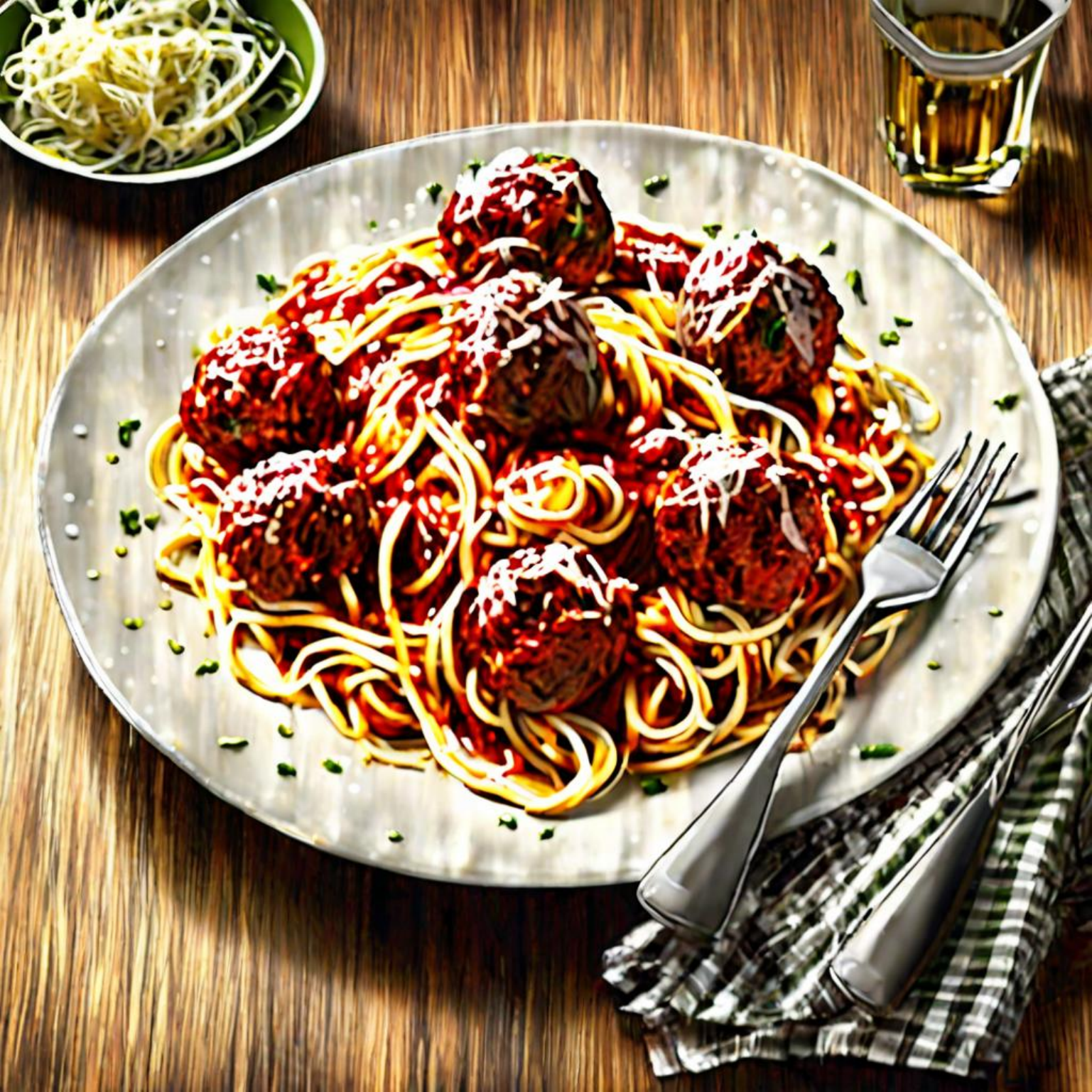} \\

        \small Score: 23.46 & \small Score: 24.97 & \small Score: 25.54 & 
        \small Score: 24.43 & \small Score: 25.10 & \small Score: 25.56 & \small Score: 26.83 \\
		\midrule
		
		\includegraphics[width=\linewidth]{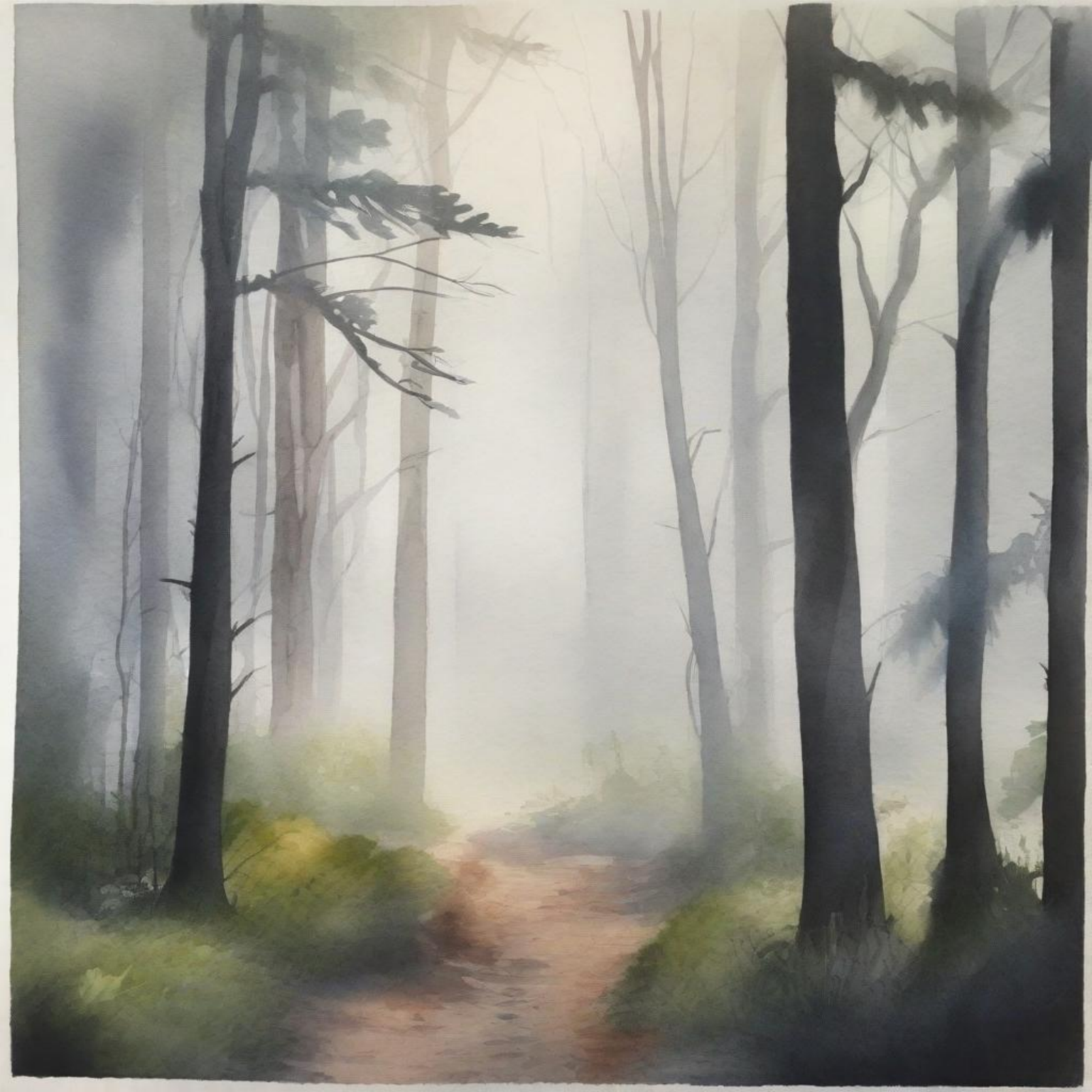} &
		\includegraphics[width=\linewidth]{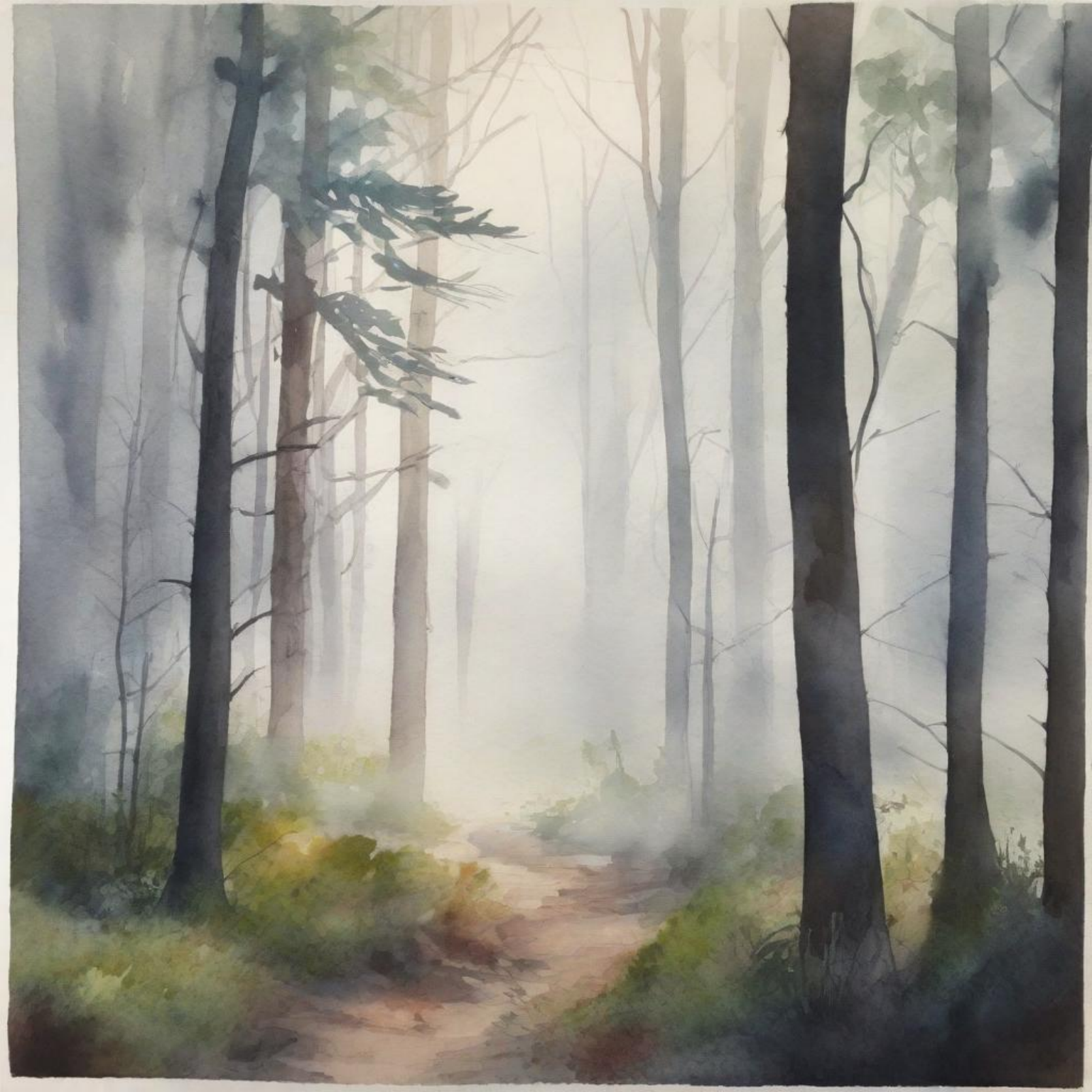} &
		\includegraphics[width=\linewidth]{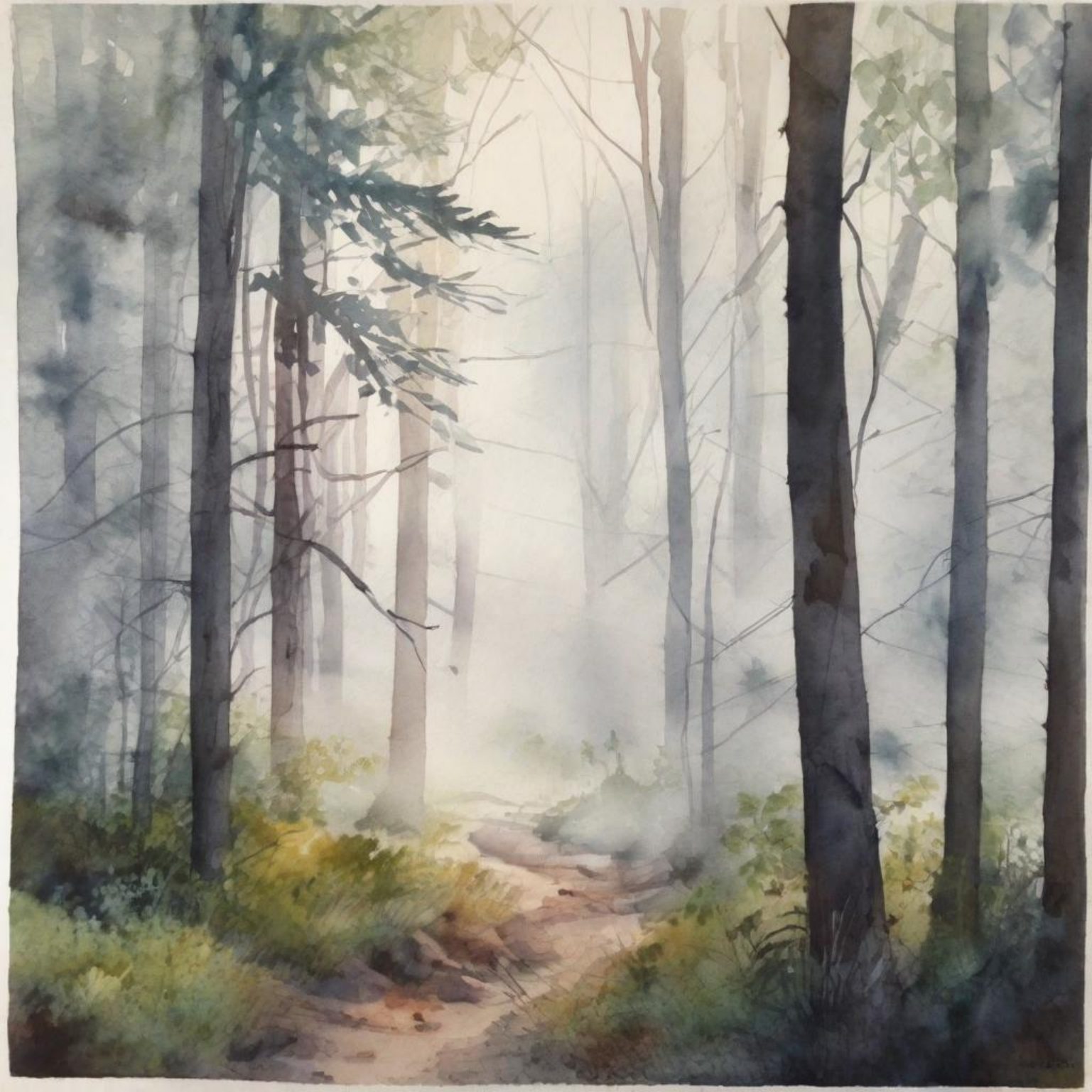} &
		\includegraphics[width=\linewidth]{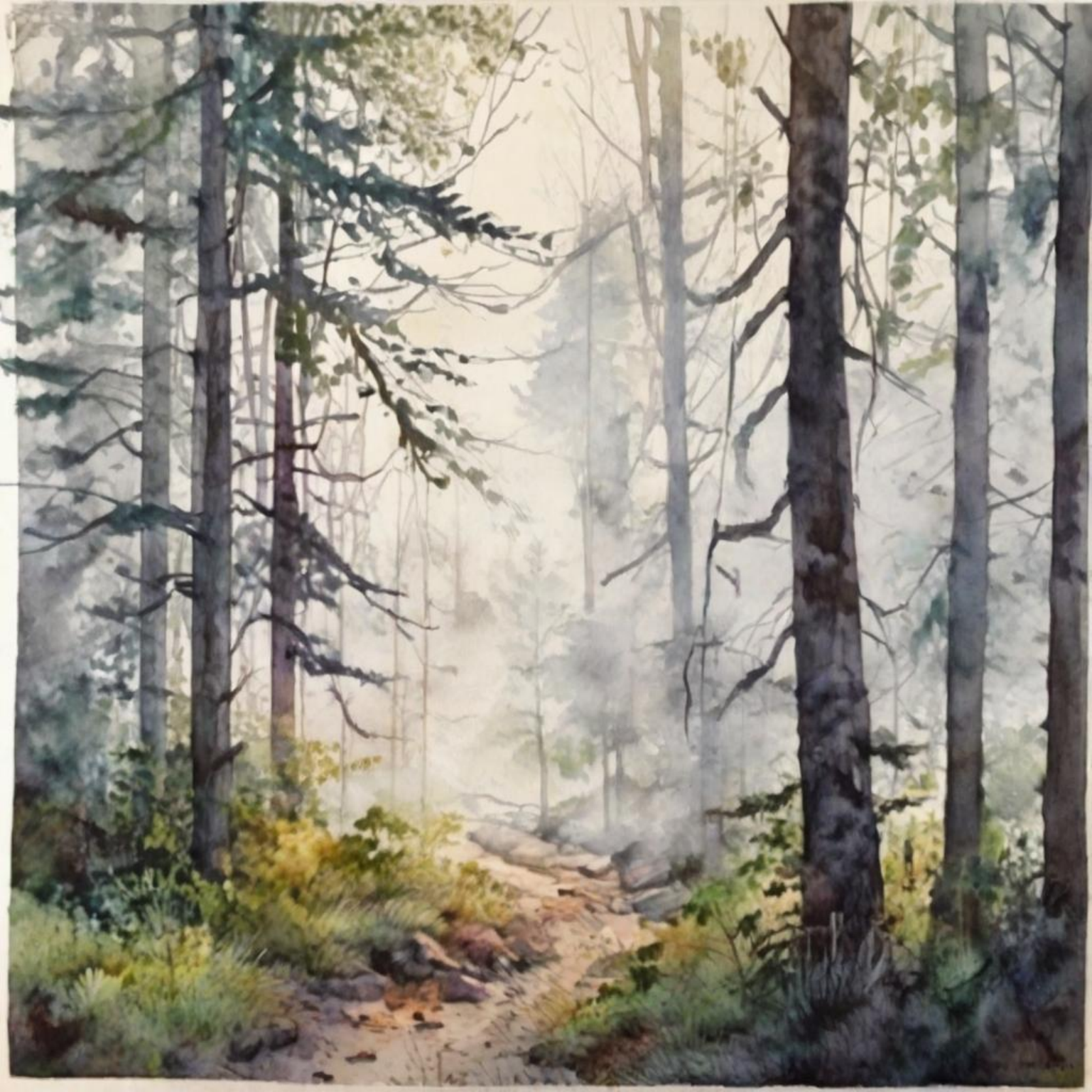} &
		\includegraphics[width=\linewidth]{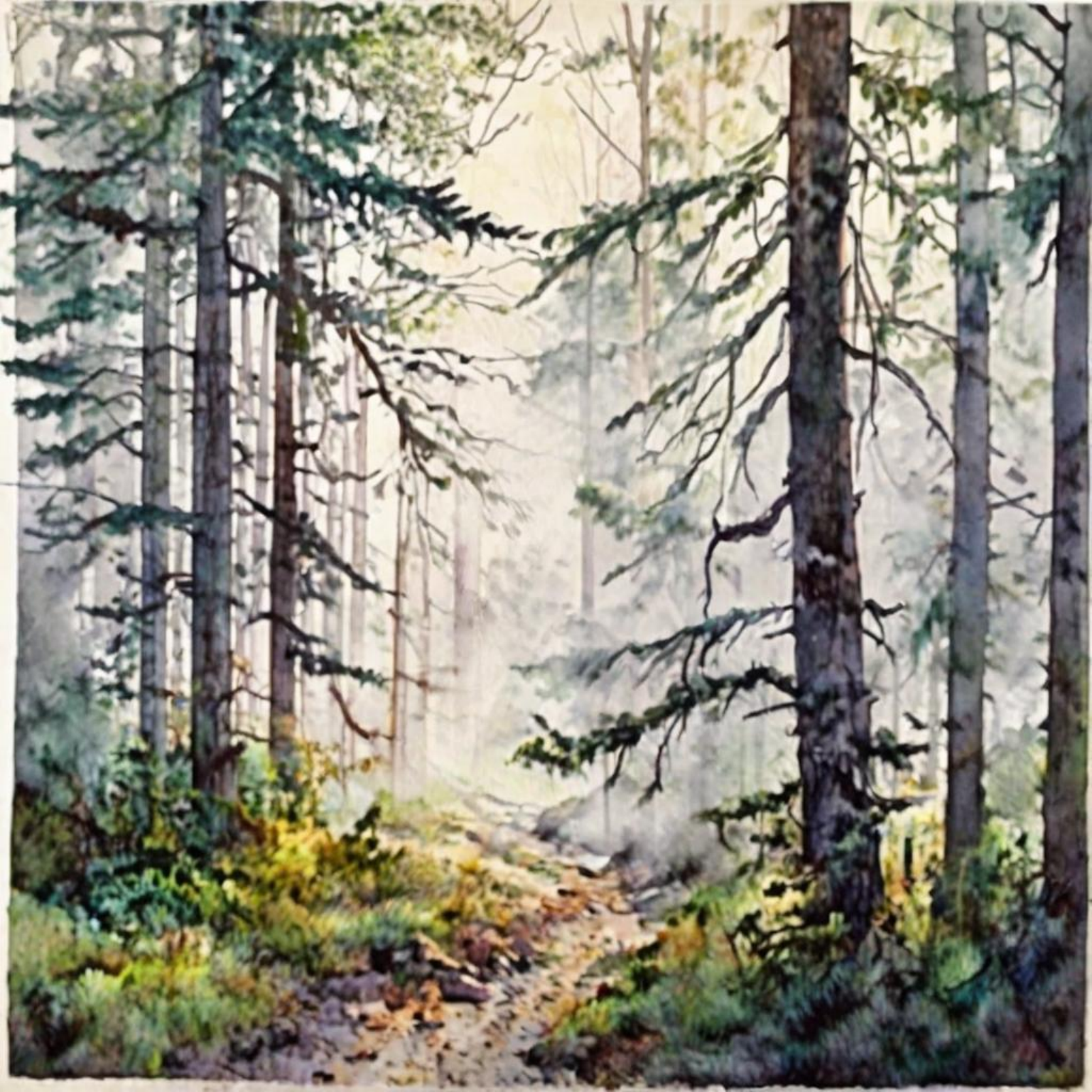} &
		\includegraphics[width=\linewidth]{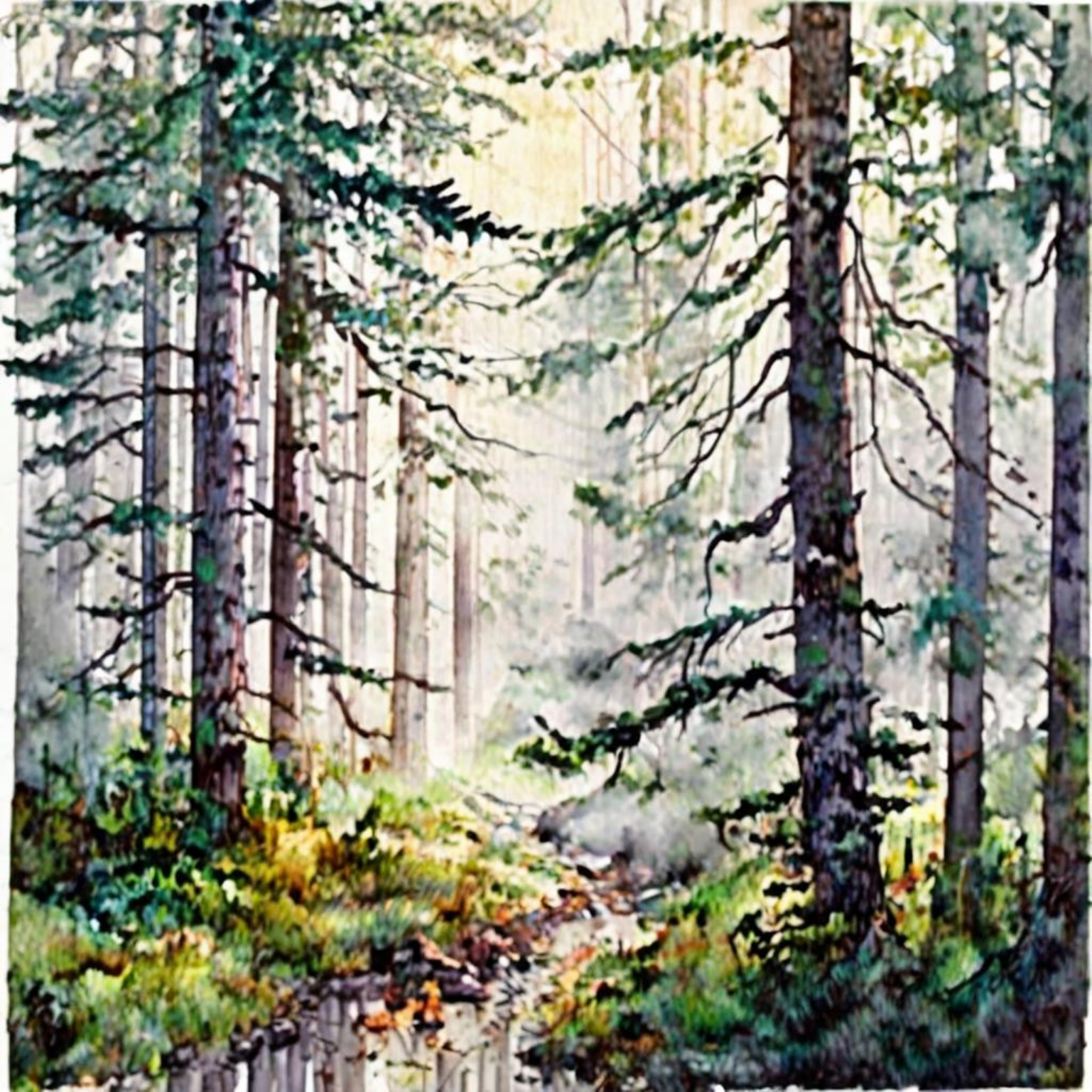} &        
		\includegraphics[width=\linewidth]{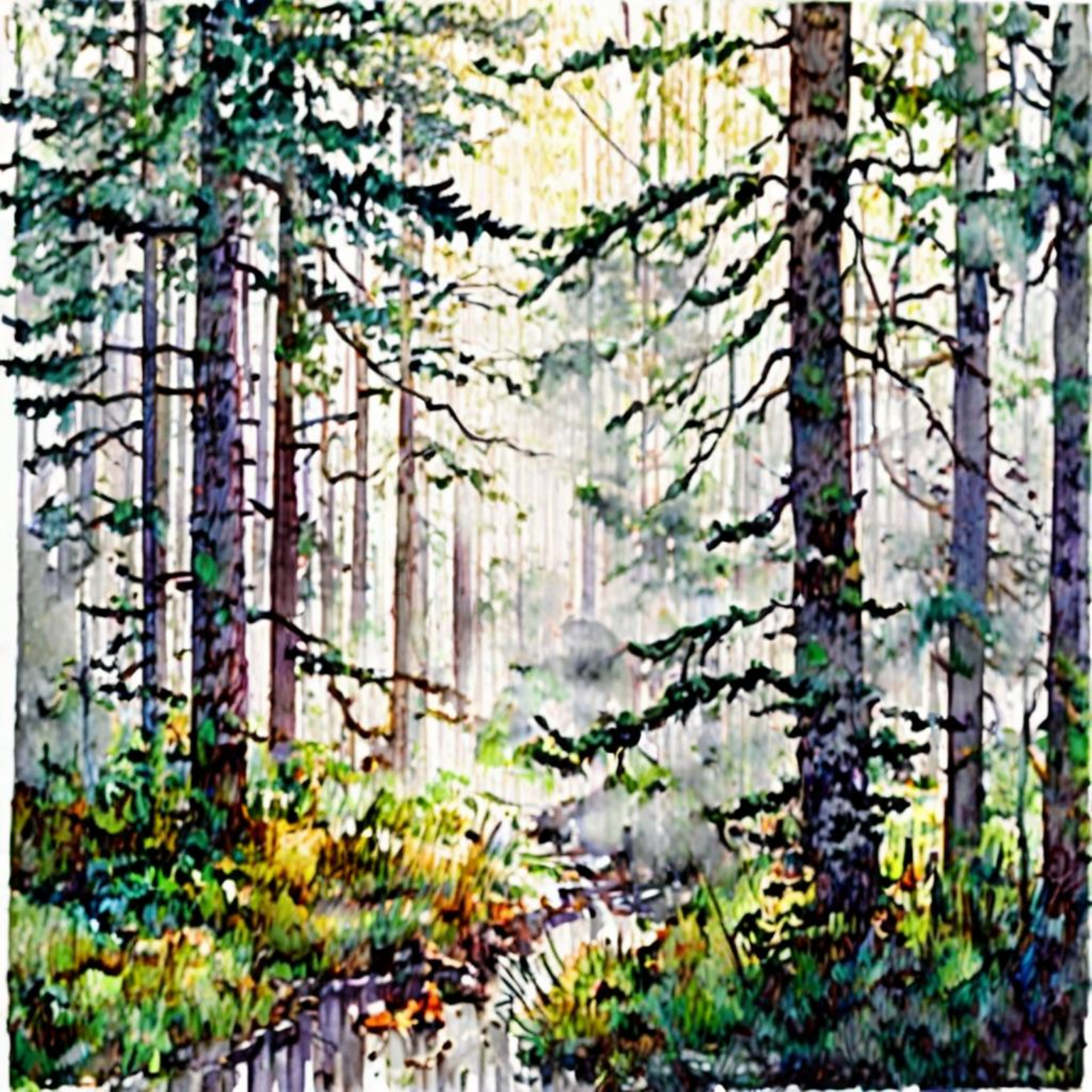} \\

        \small Score: 22.53 & \small Score: 25.26 & \small Score: 26.05 & 
        \small Score: 25.69 & \small Score: 26.07 & \small Score: 25.88 & \small Score: 25.57 \\
		\midrule
		
		\includegraphics[width=\linewidth]{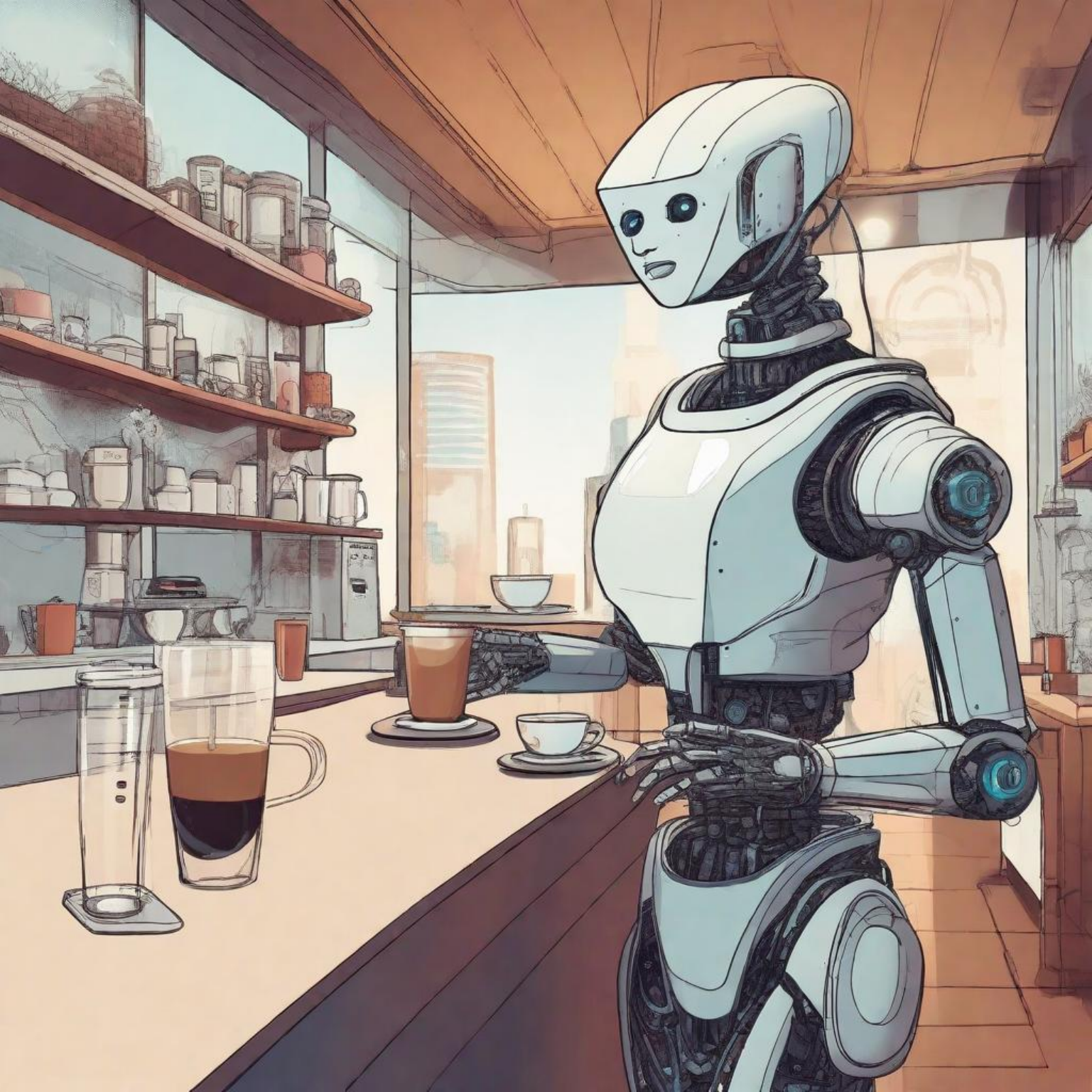} &
		\includegraphics[width=\linewidth]{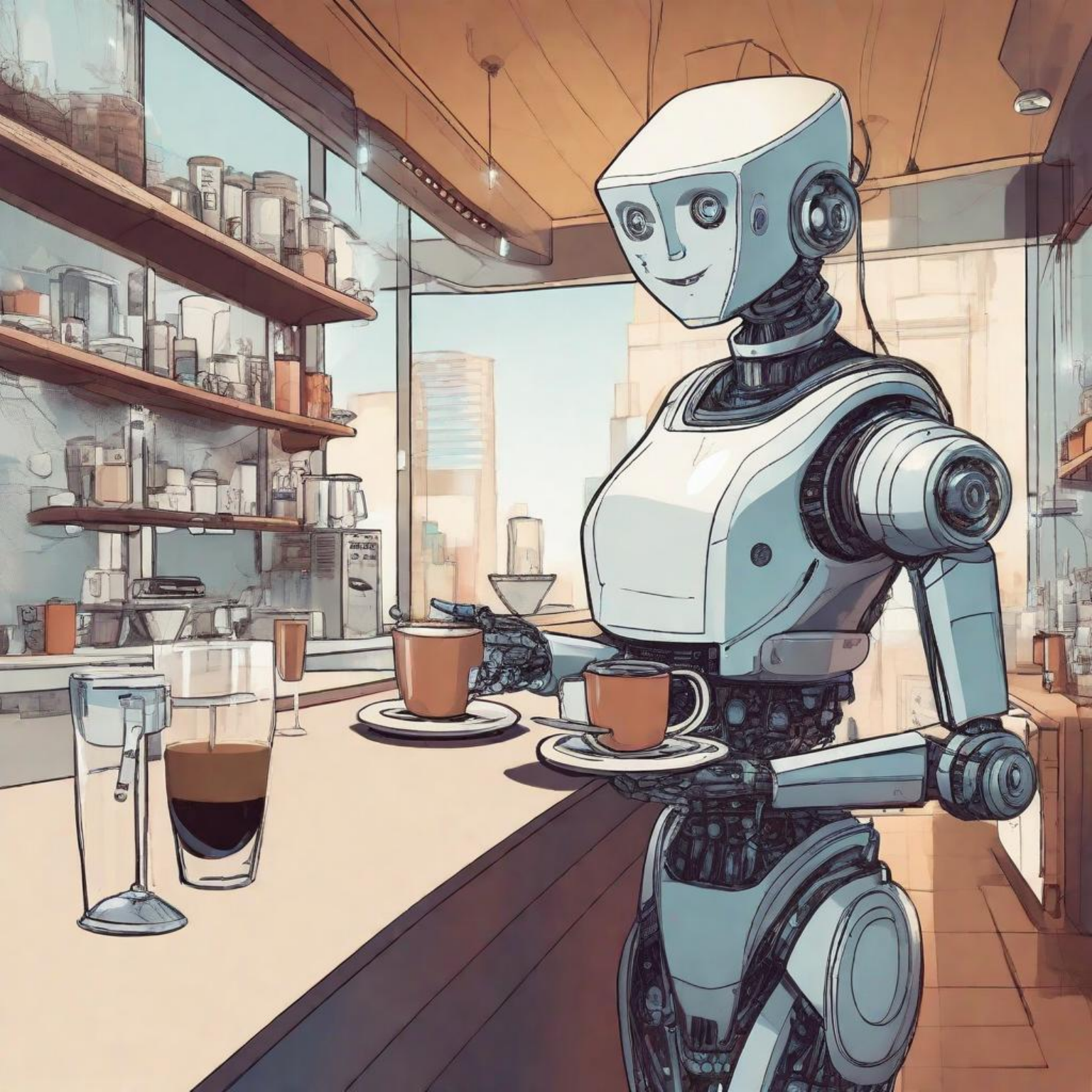} &
		\includegraphics[width=\linewidth]{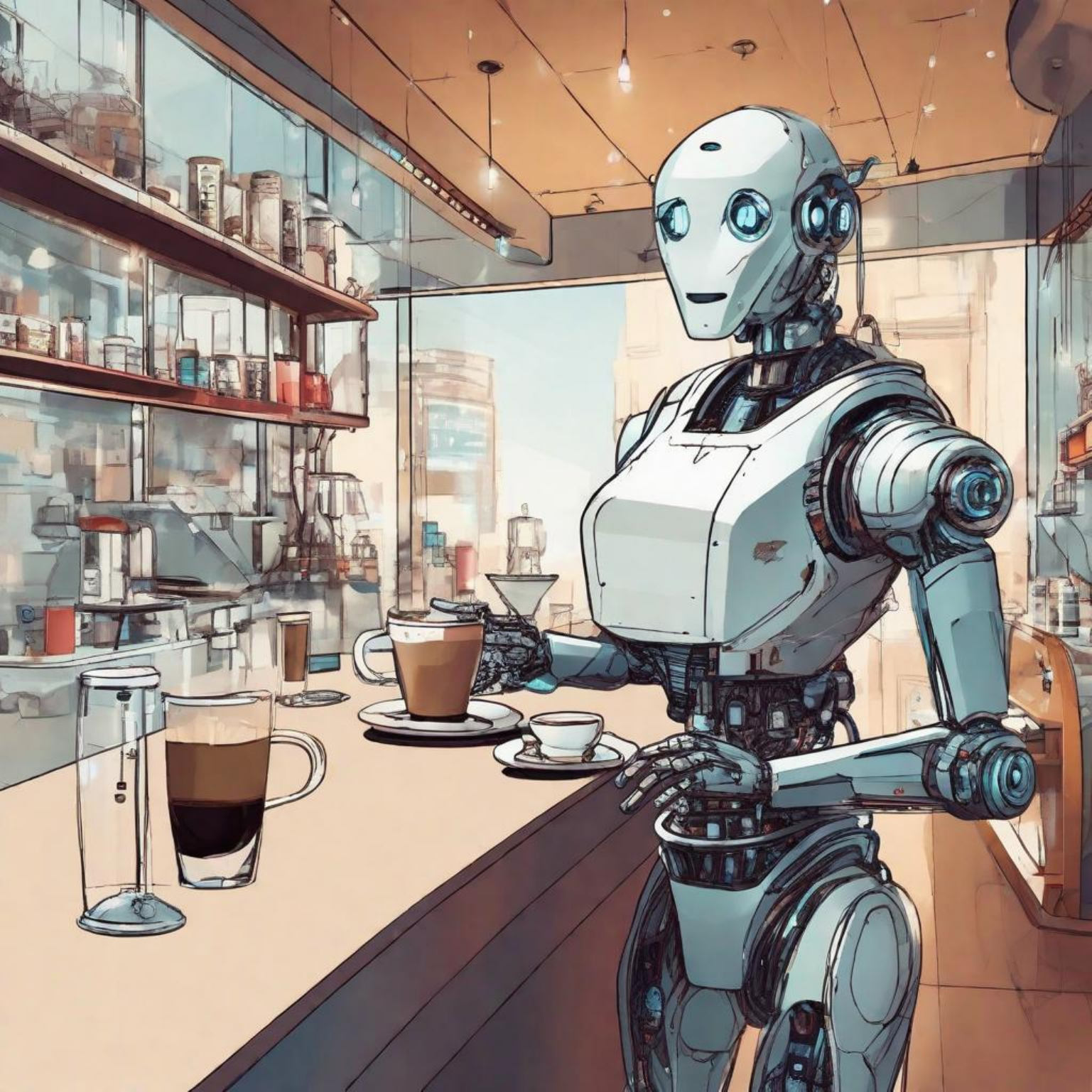} &
		\includegraphics[width=\linewidth]{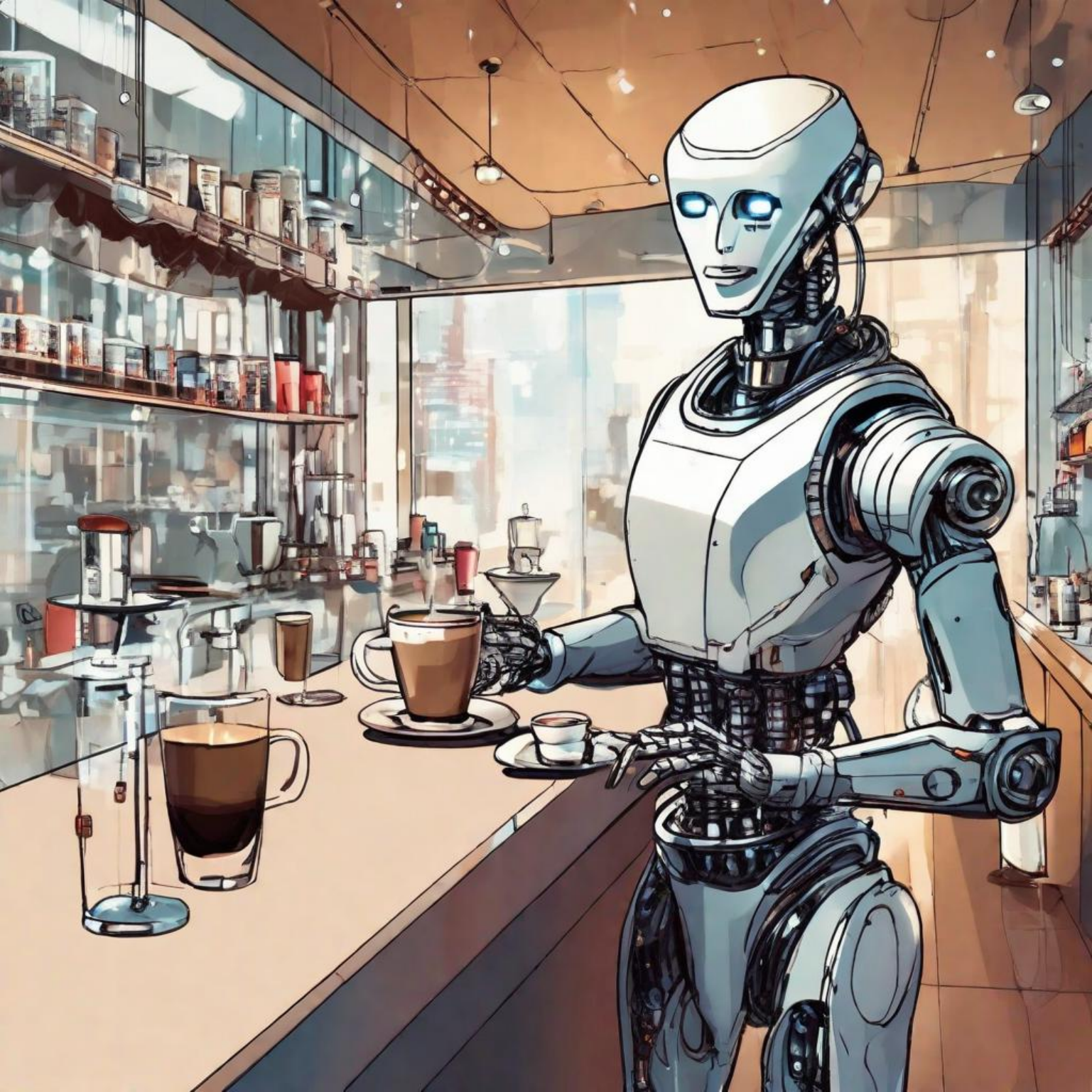} &
		\includegraphics[width=\linewidth]{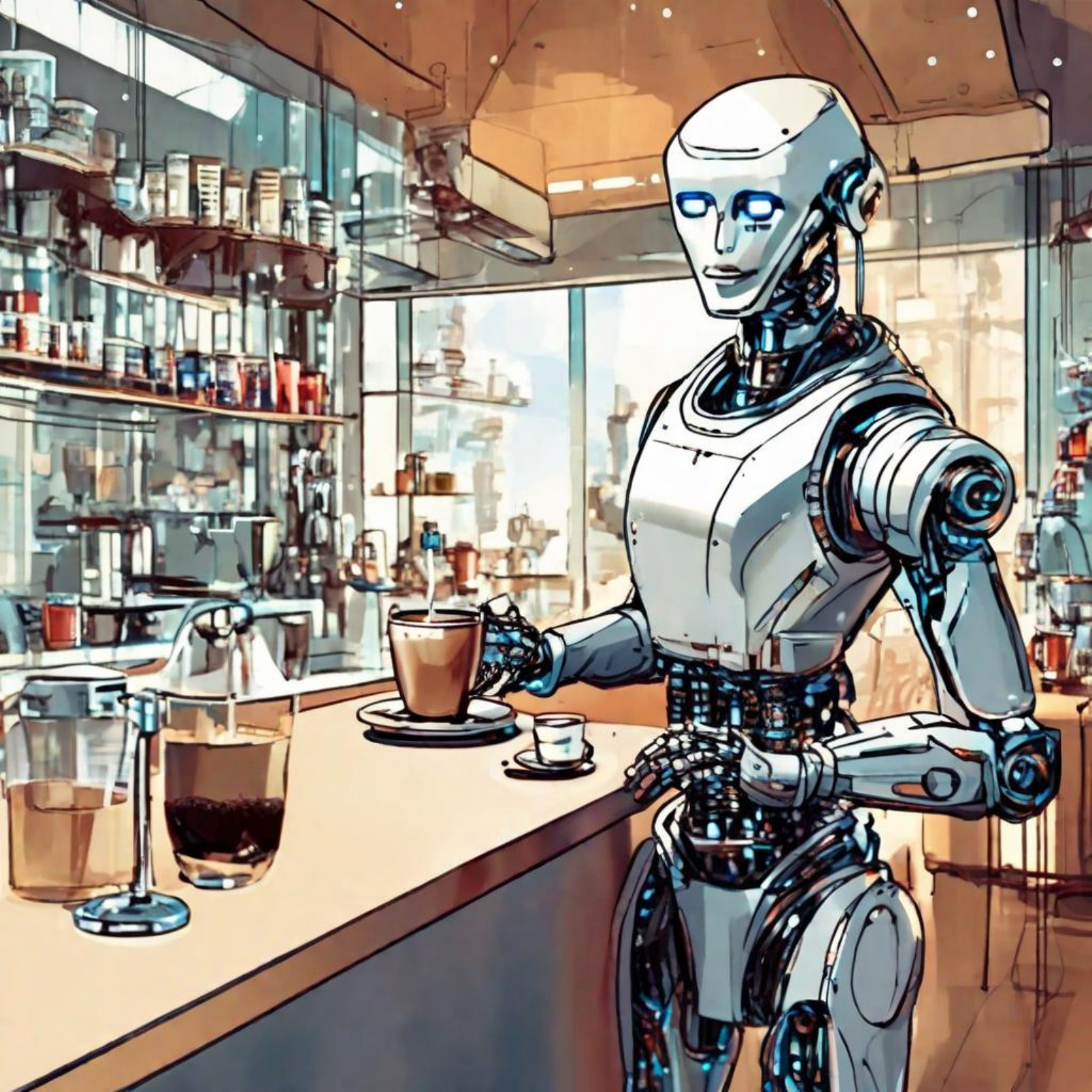} &
		\includegraphics[width=\linewidth]{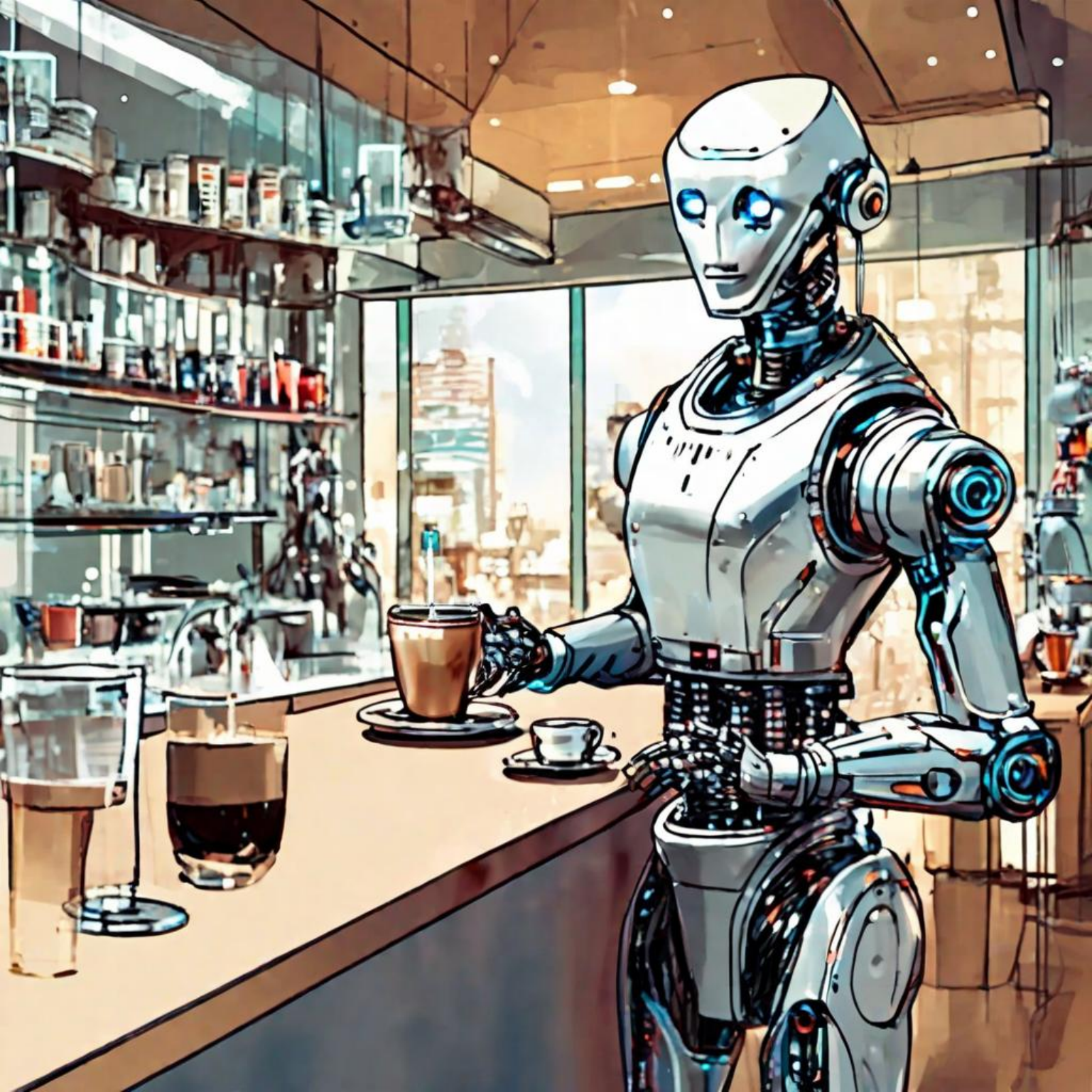} &        
		\includegraphics[width=\linewidth]{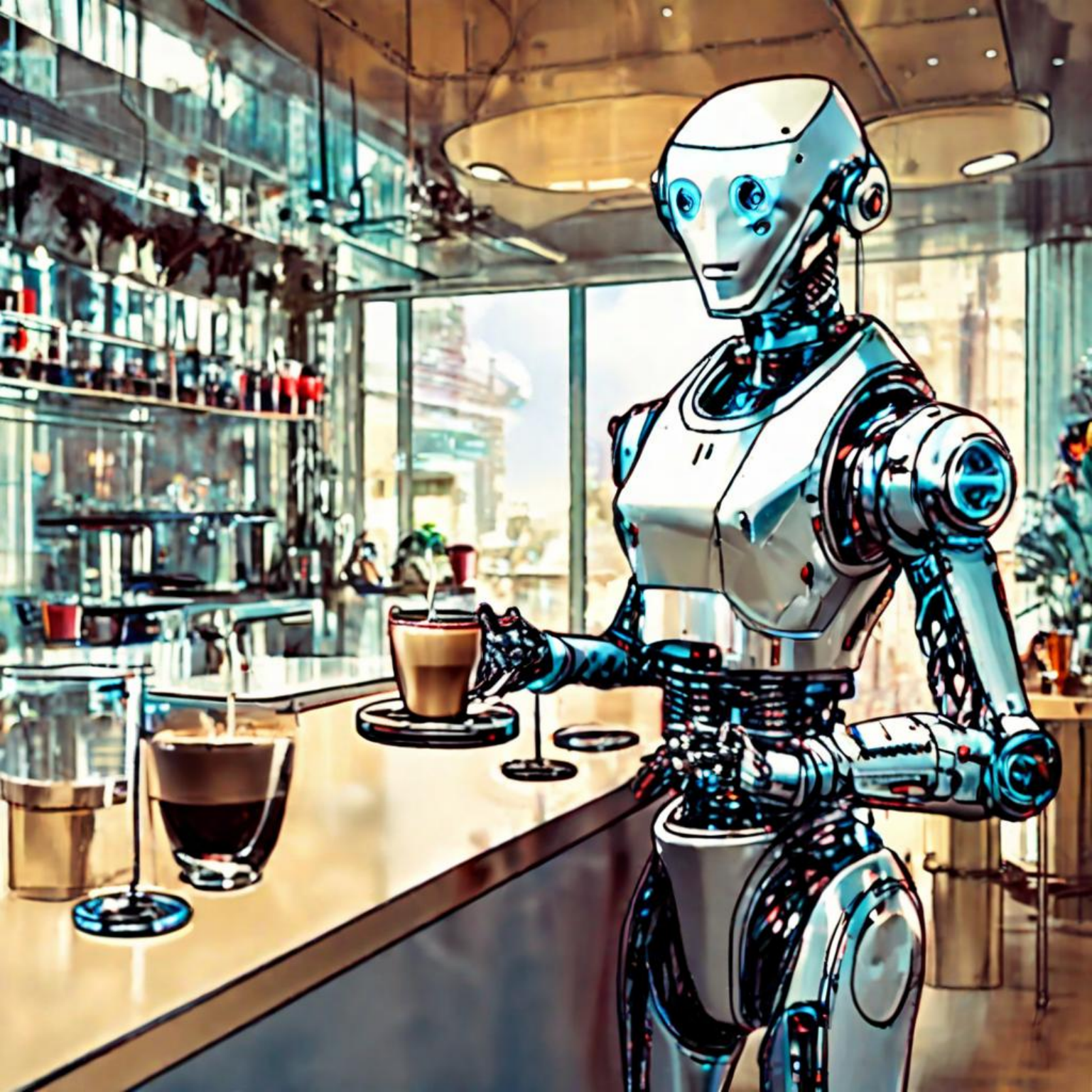} \\

        \small Score: 22.85 & \small Score: 24.34 & \small Score: 25.36 & 
        \small Score: 25.49 & \small Score: 25.92 & \small Score: 26.54 & \small Score: 26.64 \\
		
	\end{longtable}
}

\subsection{Cross-Reward Evaluation of SES}
\label{sec:cross-reward}

The low-frequency manifold search proposed by SES essentially constructs a stricter trust region. According to the manifold hypothesis, the energy of natural images is primarily concentrated in low-frequency components. By explicitly stripping and freezing high-frequency components, SES forces the optimizer to improve scores strictly within the semantically dense low-frequency subspace by adjusting structure and layout. Since low-frequency signals typically correspond to human-perceptible robust features, this physically precludes the generation of high-frequency adversarial examples, ensuring that the generated results remain constrained within the support of the natural image distribution.

To quantitatively verify whether SES is susceptible to Reward Hacking, we employ cross-reward evaluation. Specifically, when optimizing for a specific target reward (e.g., Aesthetic Score), we synchronously monitor the performance of the generated images on four other unseen reward models. If a method improves its score by ``tricking'' the target reward model, it typically induces a significant degradation in other metrics.

\begin{figure*}[t]
	\centering

	\begin{subfigure}{\textwidth}
		\centering
		\includegraphics[width=0.95\linewidth]{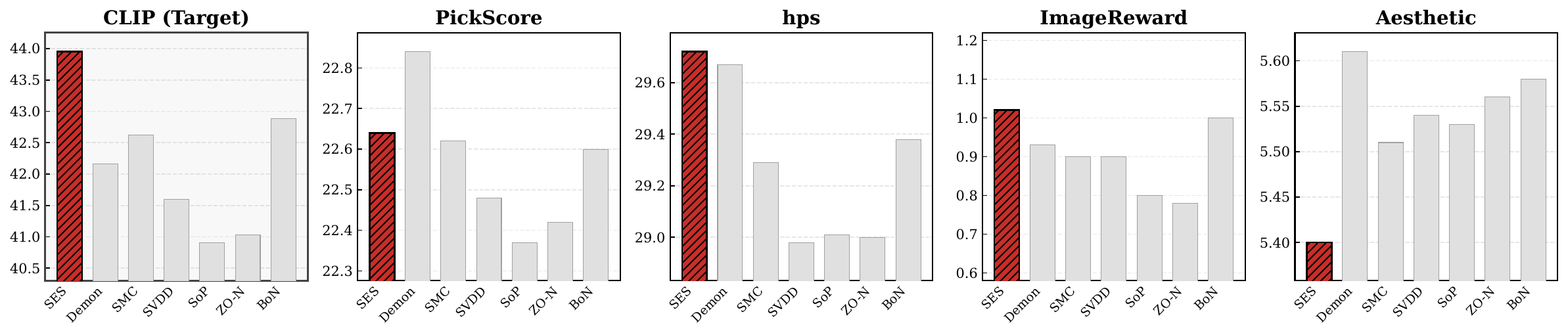}
		\caption{Optimization Target: CLIP}
		\label{fig:hack_clip}
	\end{subfigure}
	
	\vspace{3mm}
    
	\begin{subfigure}{\textwidth}
		\centering
		\includegraphics[width=0.95\linewidth]{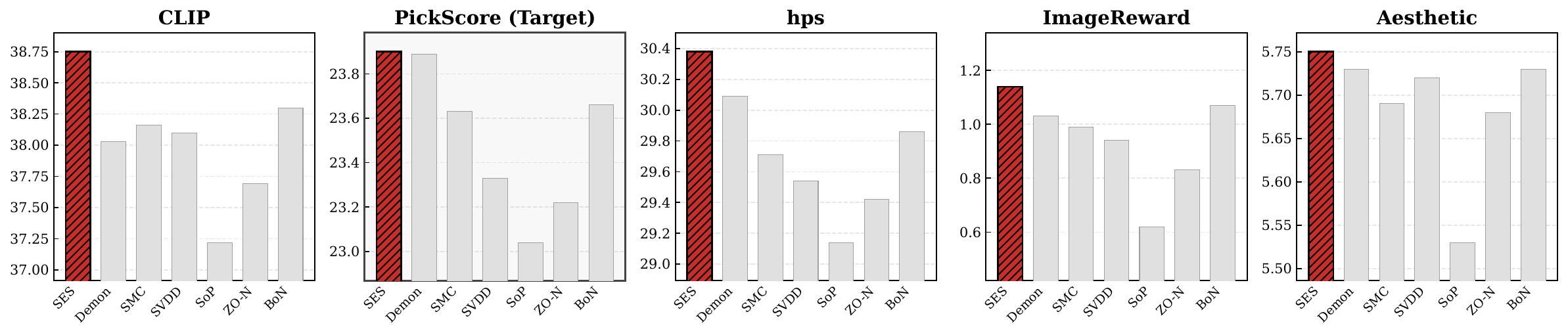}
		\caption{Optimization Target: PickScore}
		\label{fig:hack_pick}
	\end{subfigure}
	
	\vspace{3mm}

	\begin{subfigure}{\textwidth}
		\centering
		\includegraphics[width=0.95\linewidth]{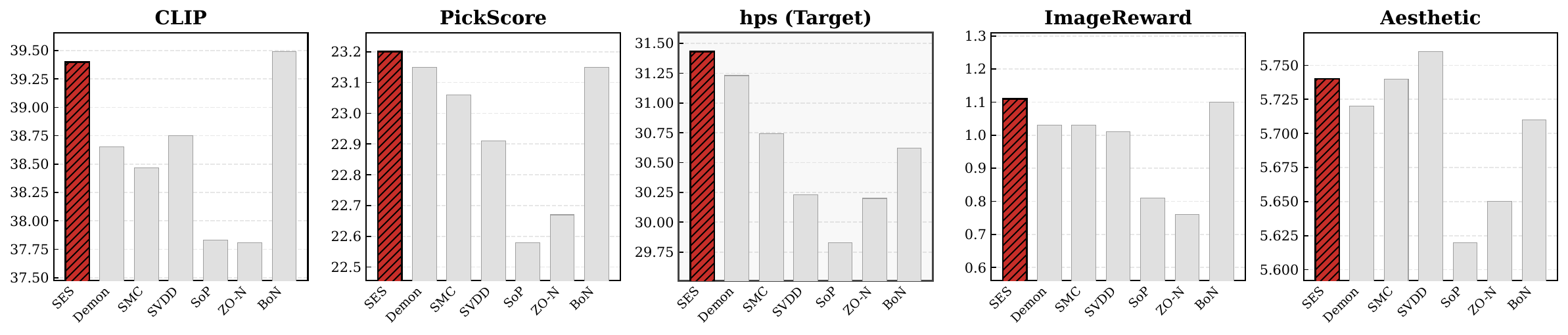}
		\caption{Optimization Target: HPSv2}
		\label{fig:hack_hps}
	\end{subfigure}
	
	\vspace{3mm}
	
    \begin{subfigure}{\textwidth}
		\centering
		\includegraphics[width=0.95\linewidth]{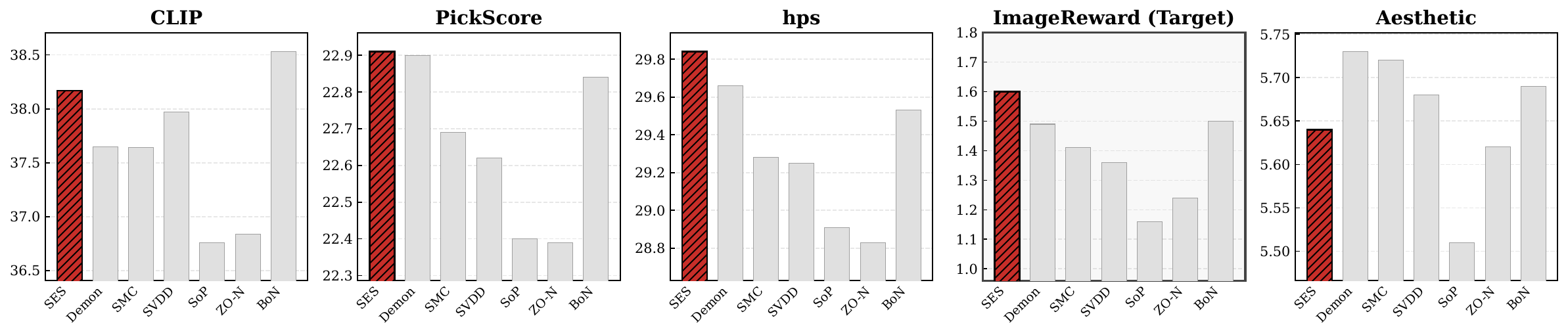}
		\caption{Optimization Target: ImageReward}
		\label{fig:hack_ir}
	\end{subfigure}

    \vspace{3mm}

	\begin{subfigure}{\textwidth}
		\centering
		\includegraphics[width=0.95\linewidth]{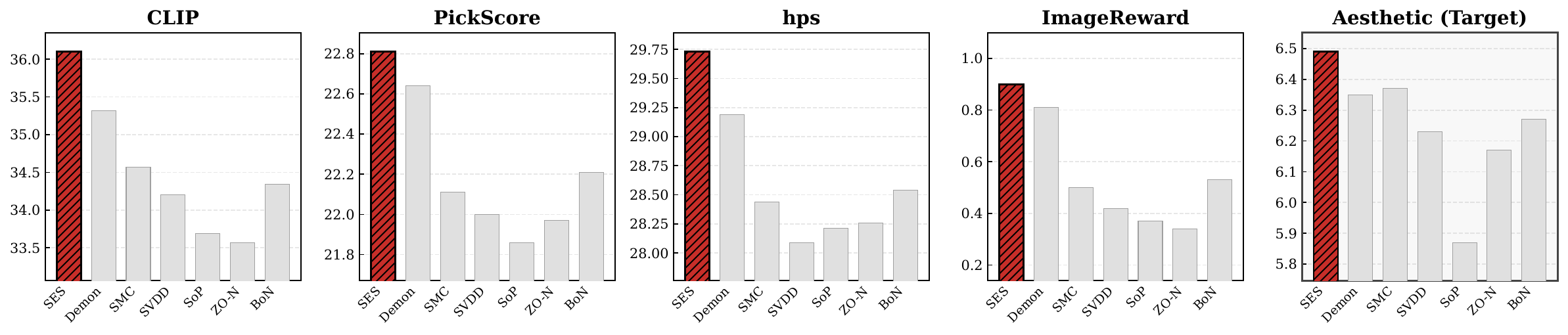}
		\caption{Optimization Target: Aesthetic}
		\label{fig:hack_aes}
	\end{subfigure}
	
	\caption{Cross-reward validation against reward hacking. Each row represents an experiment optimizing a specific target metric.}
	\label{fig:reward_hacking}
\end{figure*}

Figure \ref{fig:reward_hacking}  displays the results of five independent sets of cross-validation experiments. Each row corresponds to a specific optimization target, while the bar charts illustrate the performance of each method on that target and other evaluation metrics. The results demonstrate that SES not only maintains a lead on the target metric but also generally outperforms all baseline methods on other non-target metrics. This provides compelling evidence that the score improvements achieved by SES stem from genuine enhancements in image quality rather than adversarial overfitting to a specific reward model.

\clearpage 
\section{Additional Qualitative Results}
\label{sec:Additional_Qualitative_Results}

\begin{figure}[H] 
    \centering
    \setlength{\tabcolsep}{1pt} 
    
    \begin{tabularx}{\linewidth}{YYYYYYYY}
        
        \toprule
        \centering \textbf{SD v1.5} & 
        \centering \textbf{BoN} & 
        \centering \textbf{ZO-N} & 
        \centering \textbf{SoP} & 
        \centering \textbf{SMC} & 
        \centering \textbf{SVDD} & 
        \centering \textbf{Demon} & 
        \centering \textbf{SES} \tabularnewline
        \midrule
        
        \includegraphics[width=\linewidth]{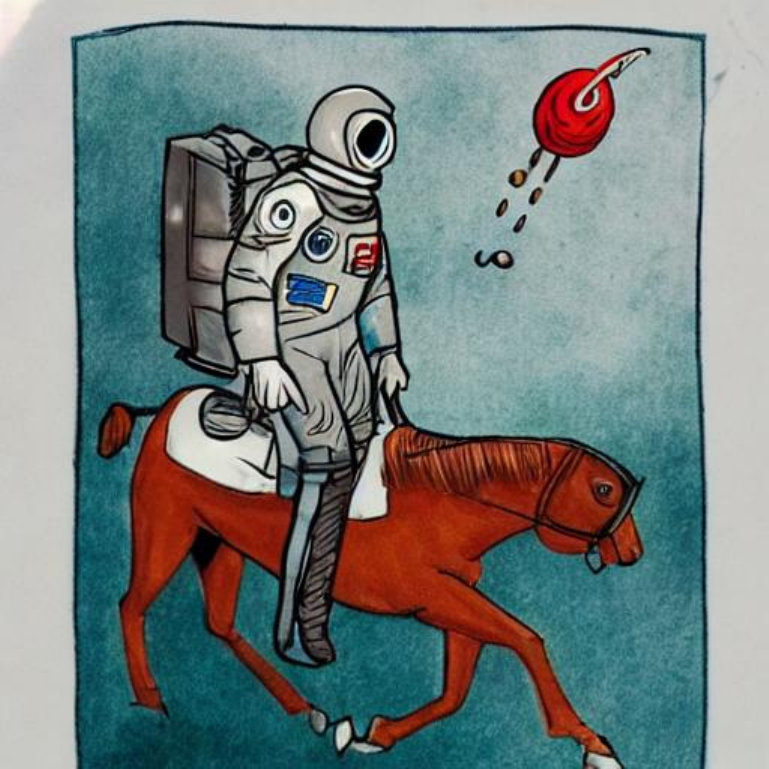} &
        \includegraphics[width=\linewidth]{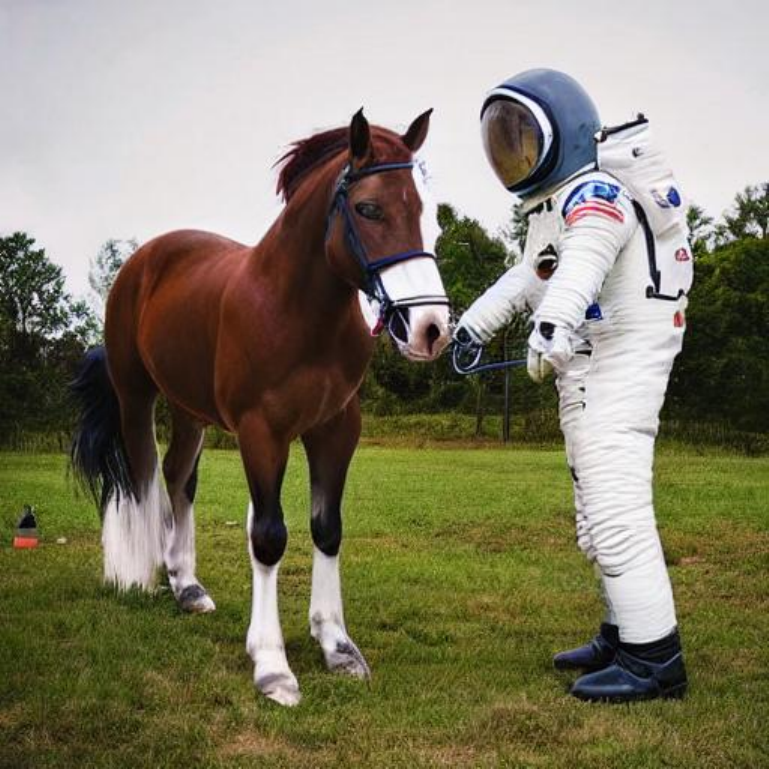} &
        \includegraphics[width=\linewidth]{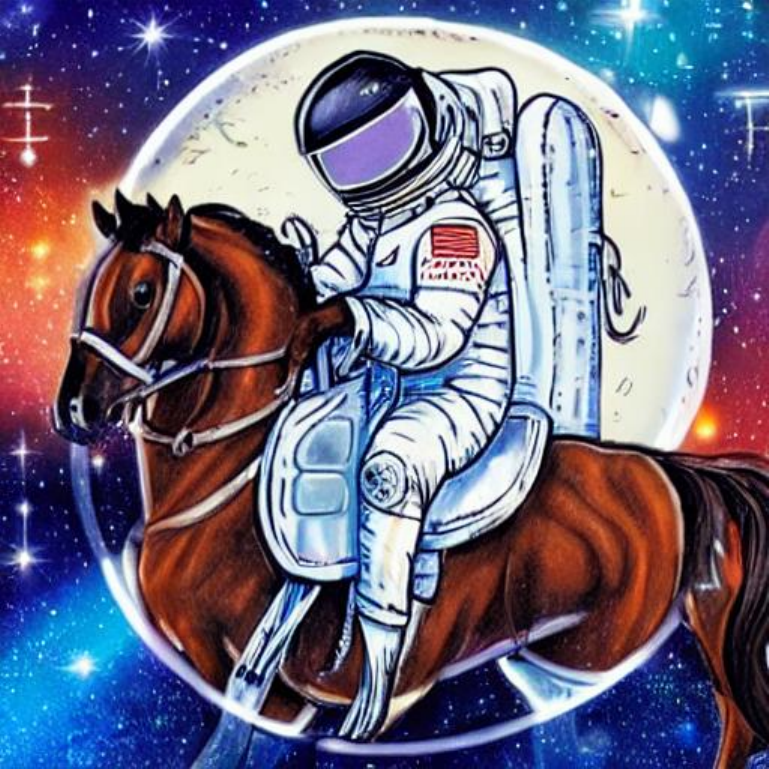} &
        \includegraphics[width=\linewidth]{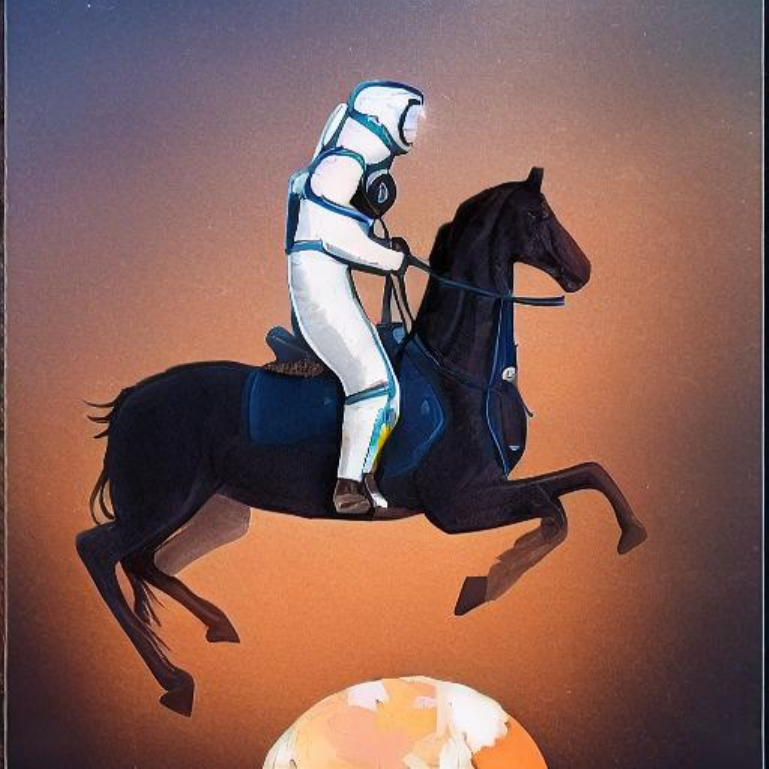} &
        \includegraphics[width=\linewidth]{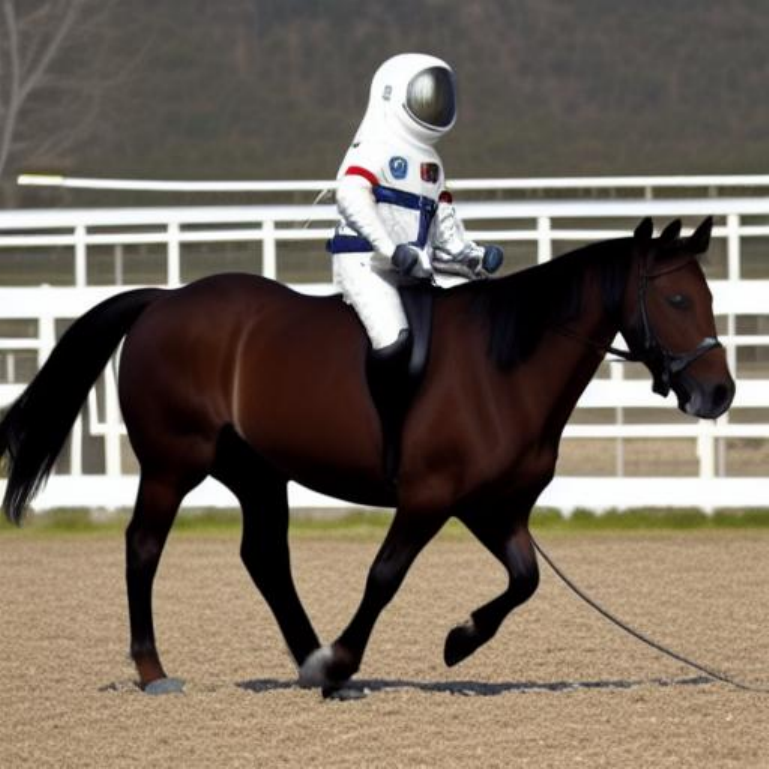} &
        \includegraphics[width=\linewidth]{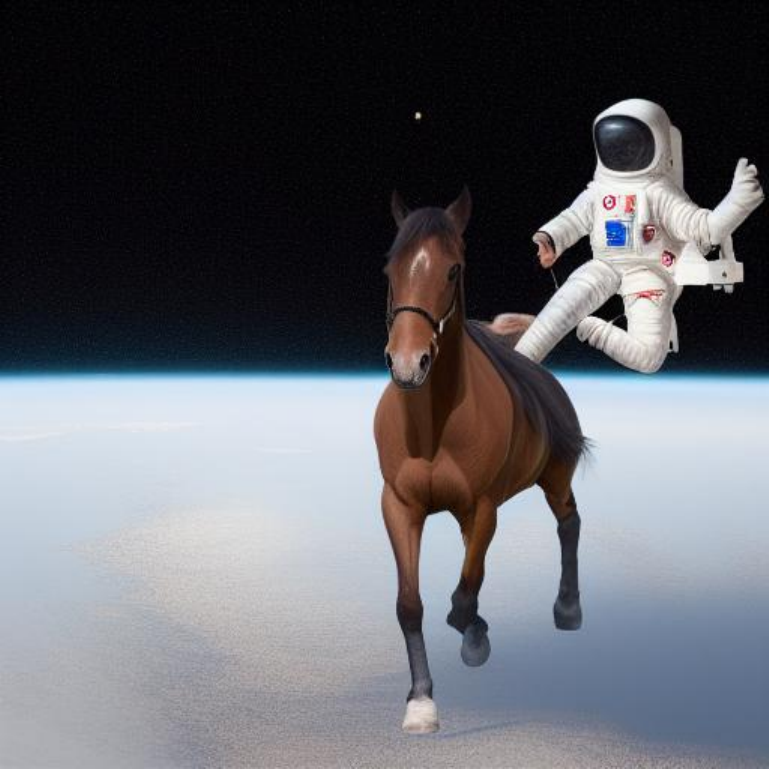} &
        \includegraphics[width=\linewidth]{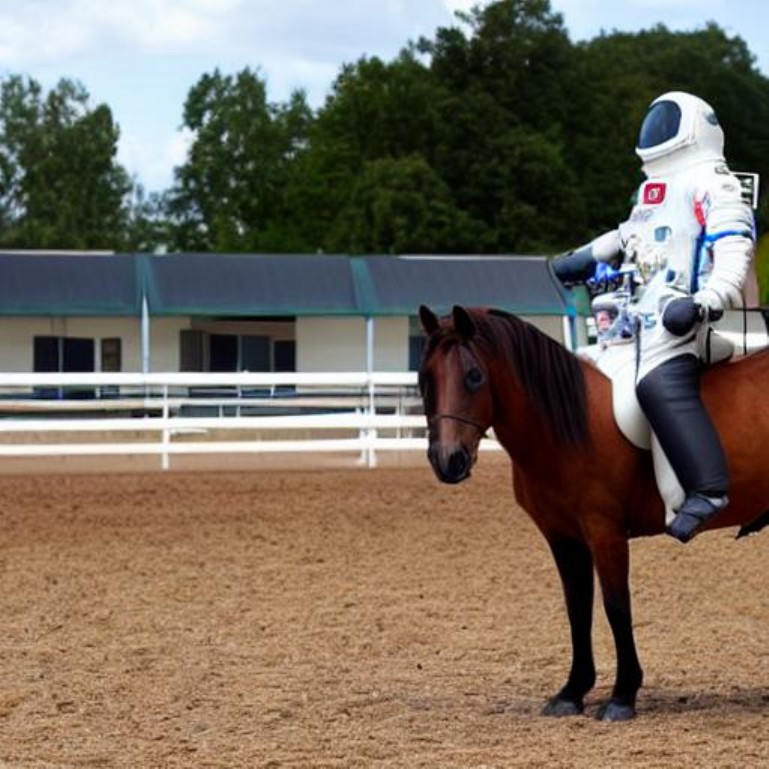} &
        \includegraphics[width=\linewidth]{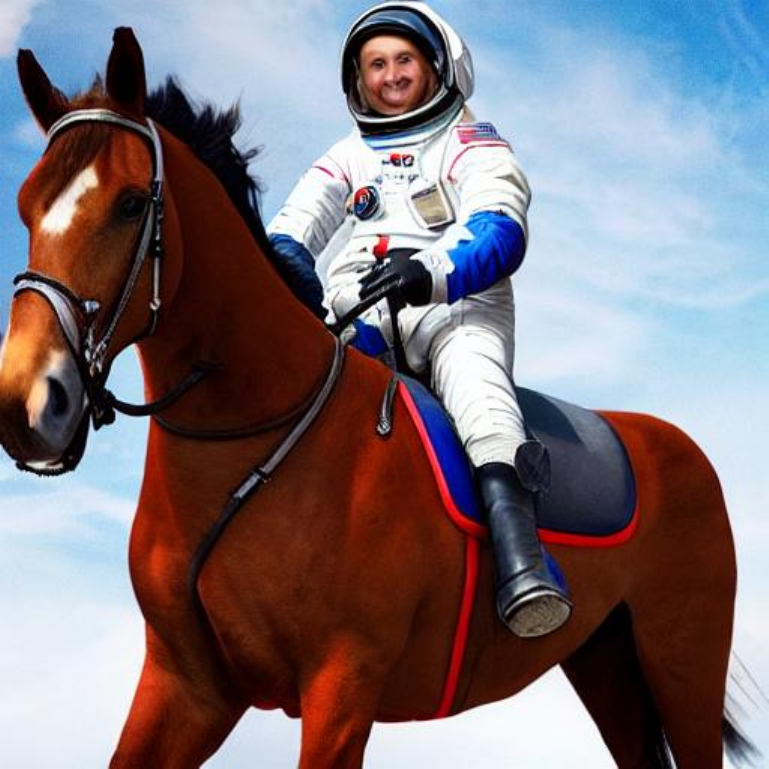} \\

        \multicolumn{8}{@{}p{\linewidth}@{}}{\centering \small \textit{Prompt: A horse riding an astronaut.}} \\
		\midrule

        \includegraphics[width=\linewidth]{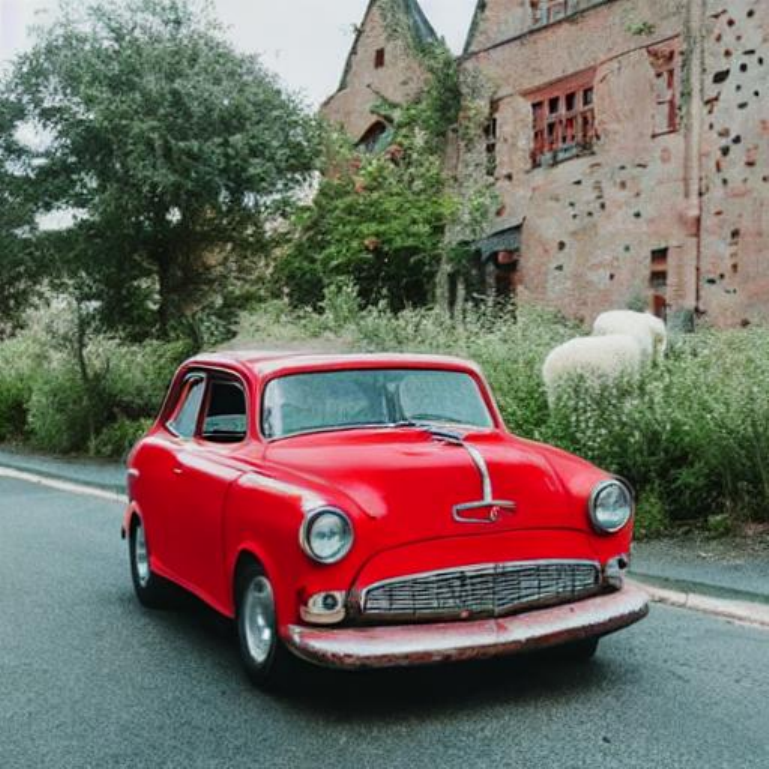} &
        \includegraphics[width=\linewidth]{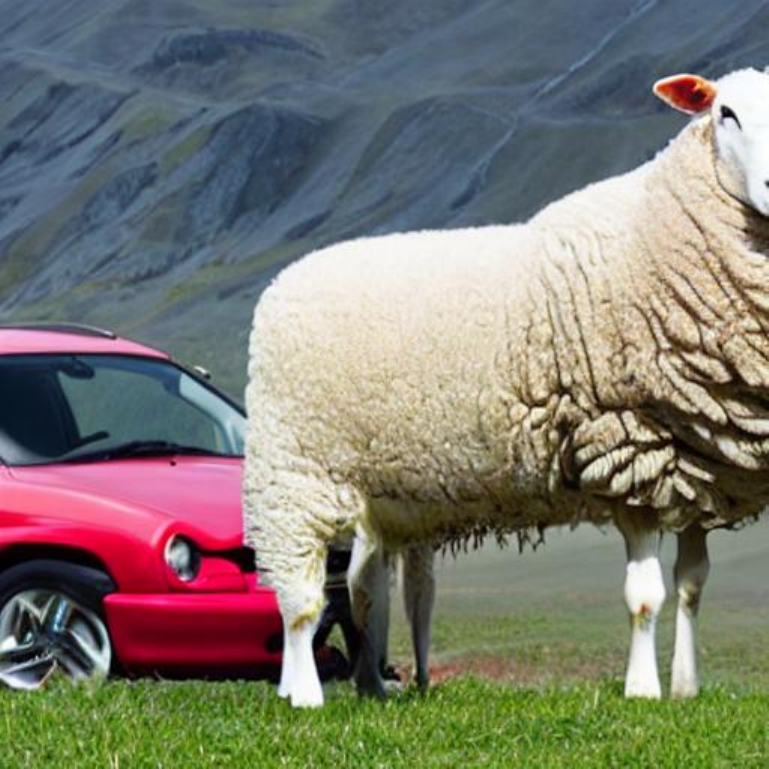} &
        \includegraphics[width=\linewidth]{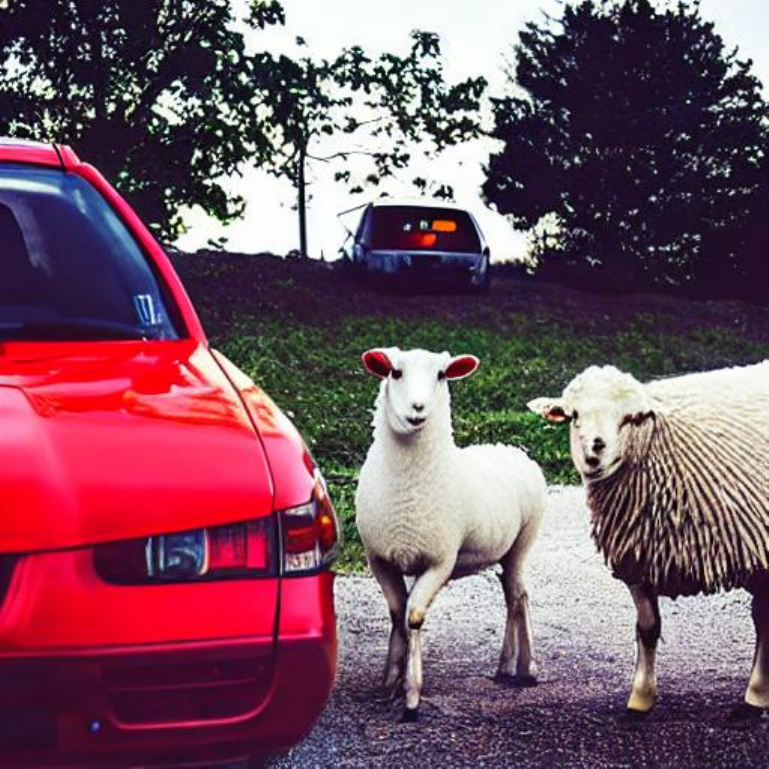} &
        \includegraphics[width=\linewidth]{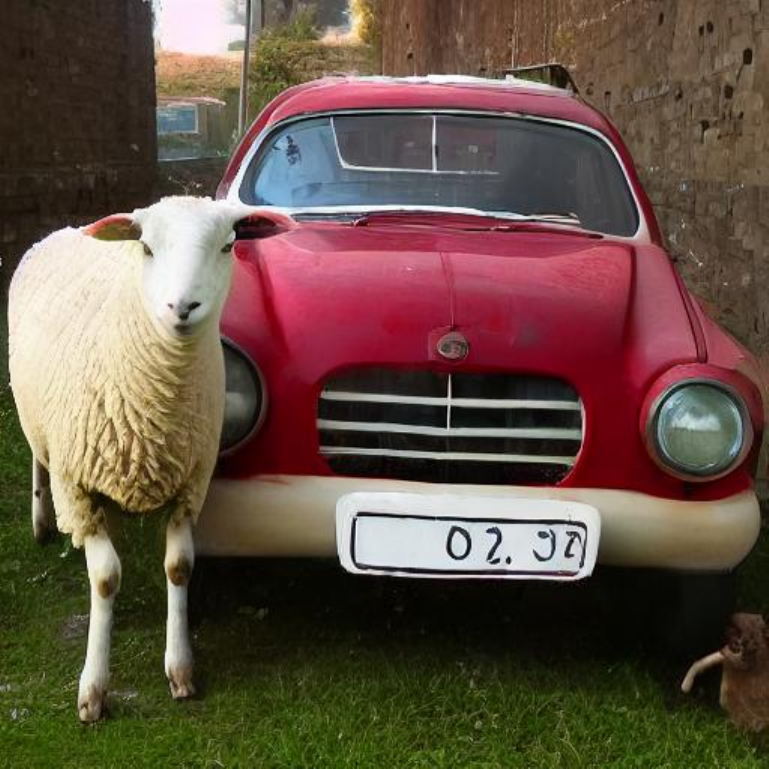} &
        \includegraphics[width=\linewidth]{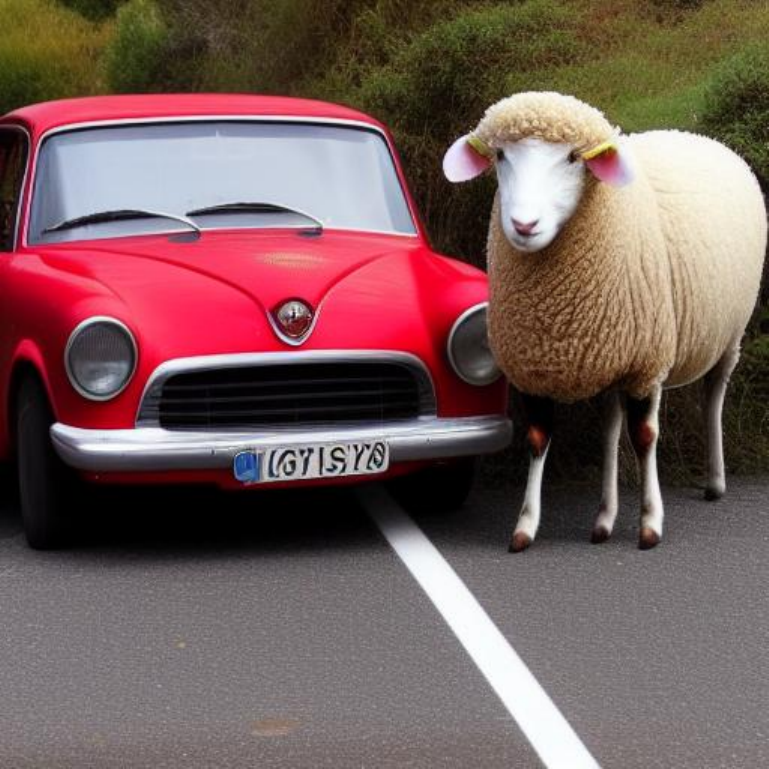} &
        \includegraphics[width=\linewidth]{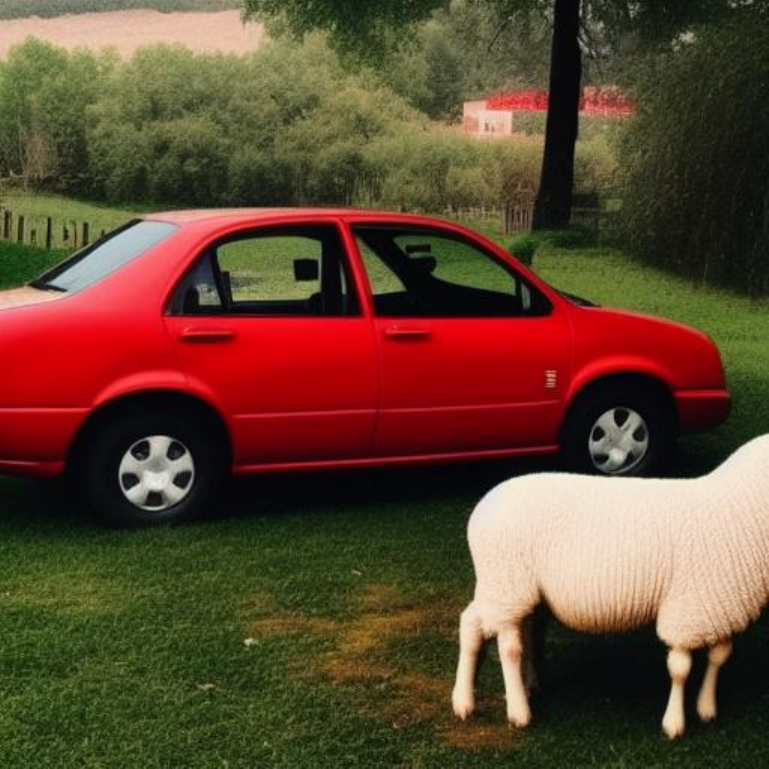} &
        \includegraphics[width=\linewidth]{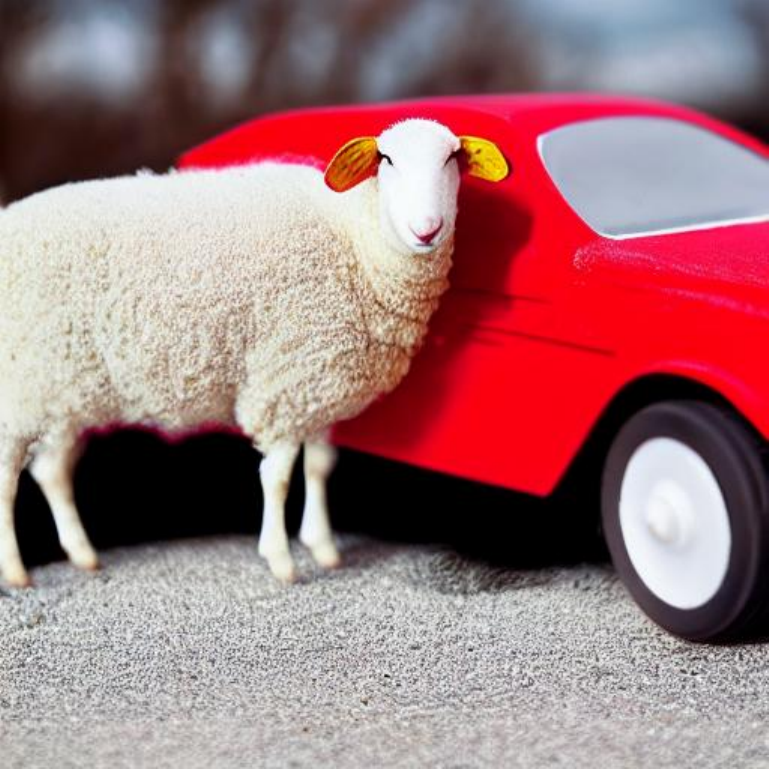} &
        \includegraphics[width=\linewidth]{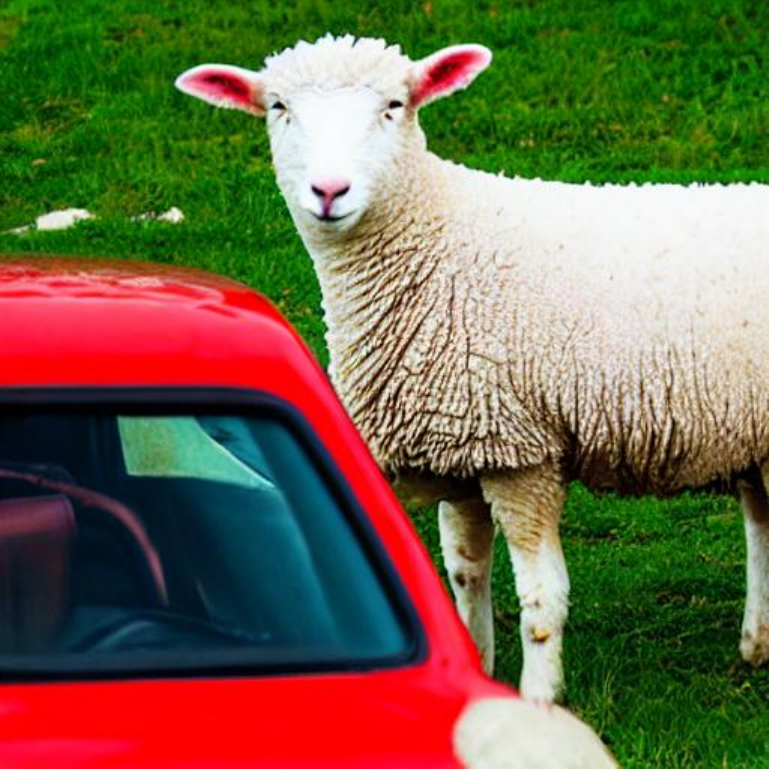} \\

        \multicolumn{8}{@{}p{\linewidth}@{}}{\centering \small \textit{Prompt: A red car and a white sheep.}} \\
		\midrule

        \includegraphics[width=\linewidth]{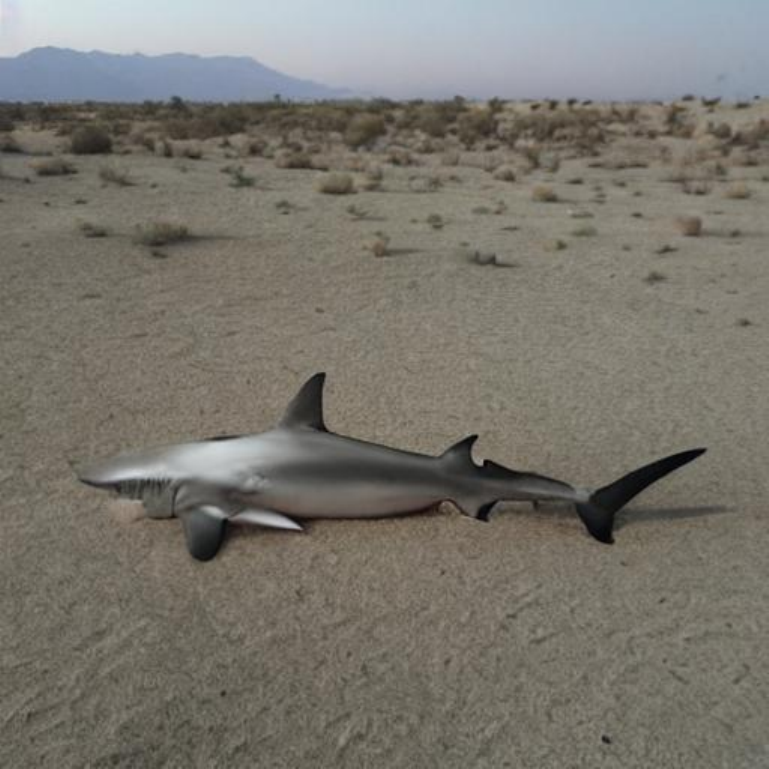} &
        \includegraphics[width=\linewidth]{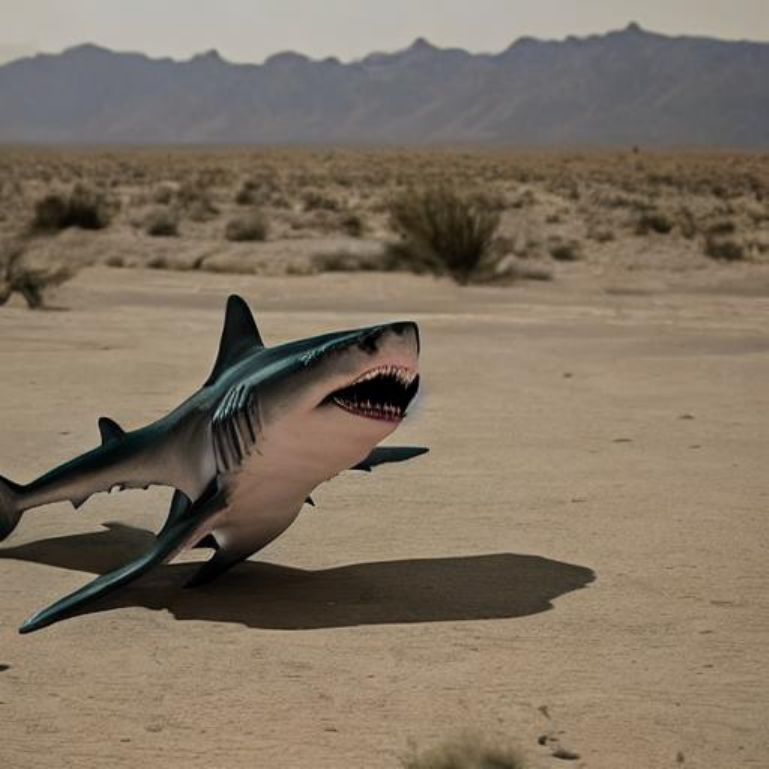} &
        \includegraphics[width=\linewidth]{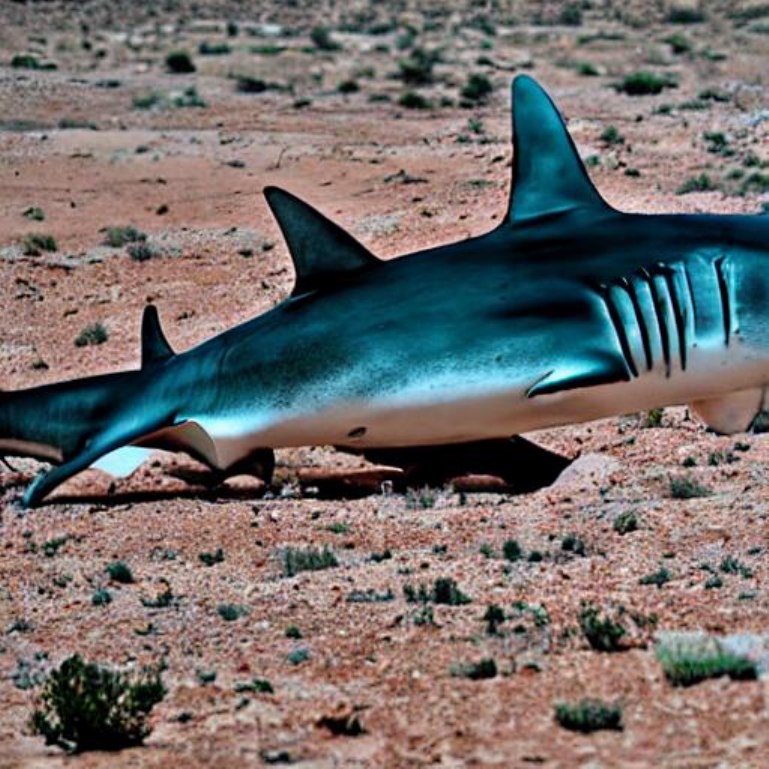} &
        \includegraphics[width=\linewidth]{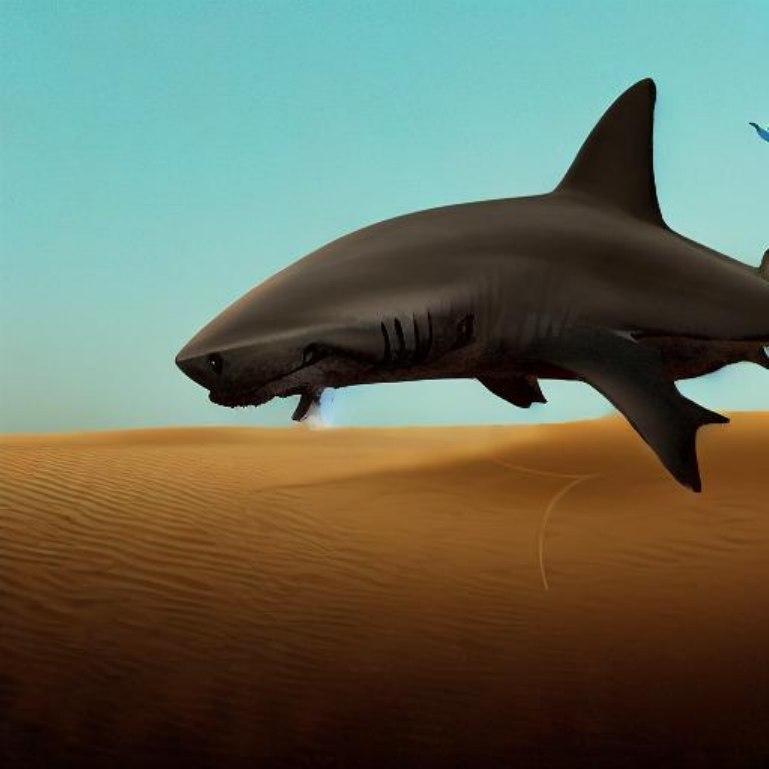} &
        \includegraphics[width=\linewidth]{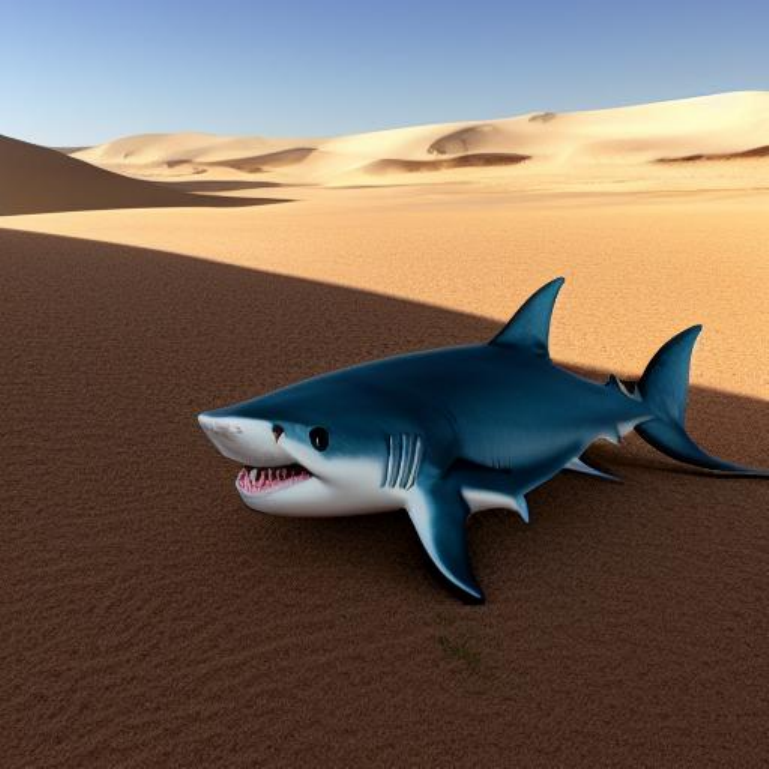} &
        \includegraphics[width=\linewidth]{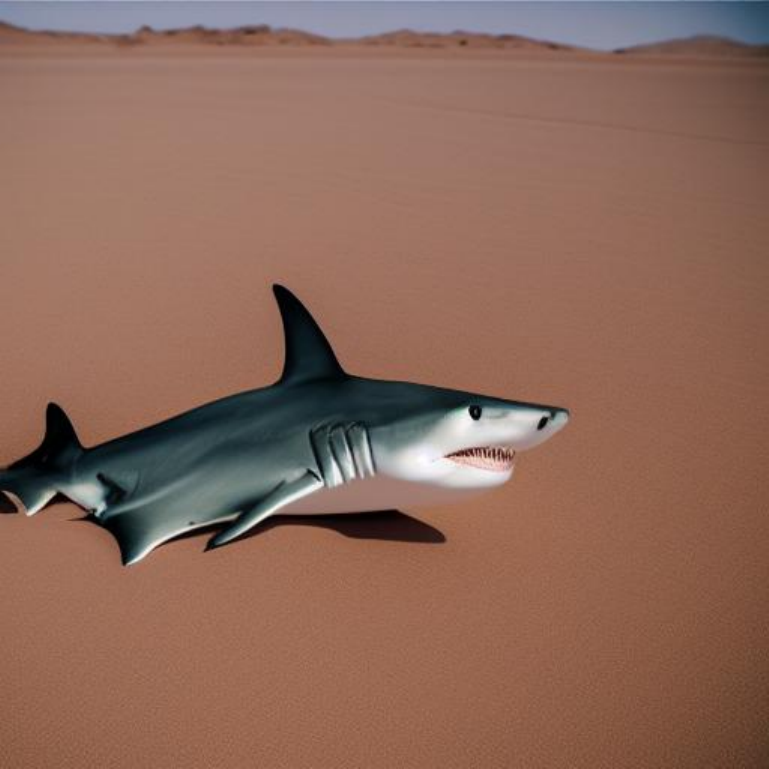} &
        \includegraphics[width=\linewidth]{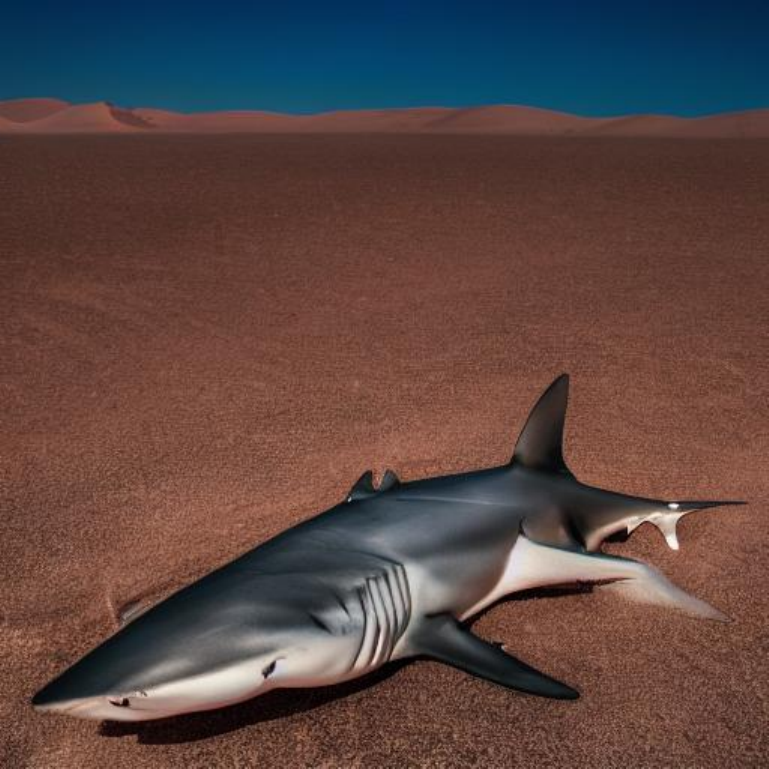} &
        \includegraphics[width=\linewidth]{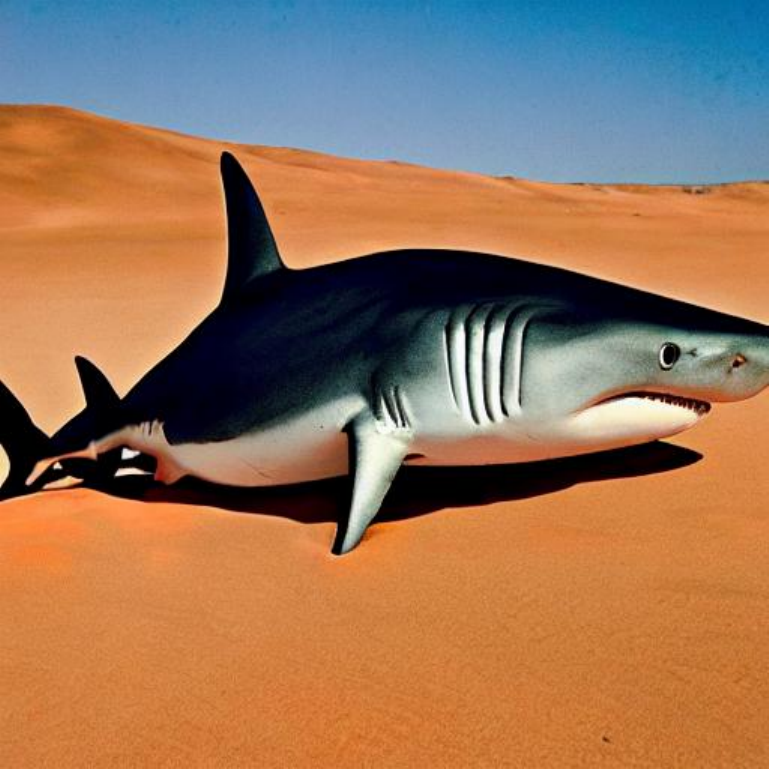} \\

        \multicolumn{8}{@{}p{\linewidth}@{}}{\centering \small \textit{Prompt: A shark in the desert.}} \\
		\midrule

        \includegraphics[width=\linewidth]{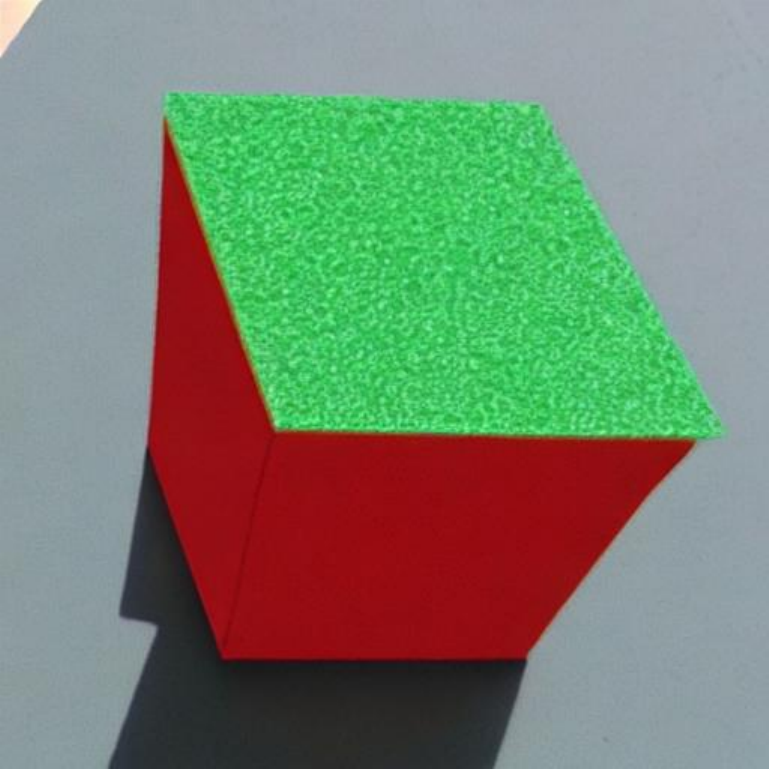} &
        \includegraphics[width=\linewidth]{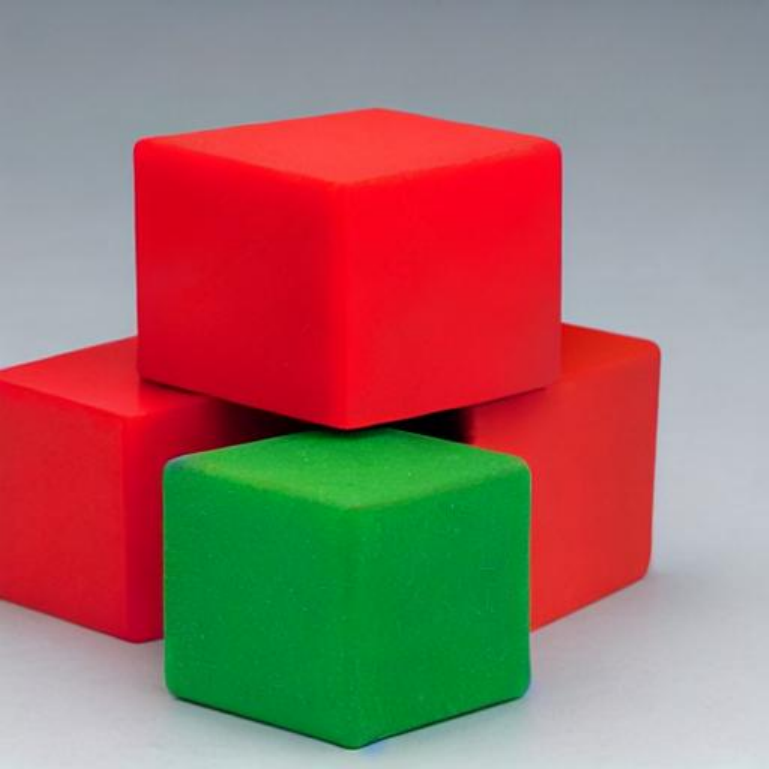} &
        \includegraphics[width=\linewidth]{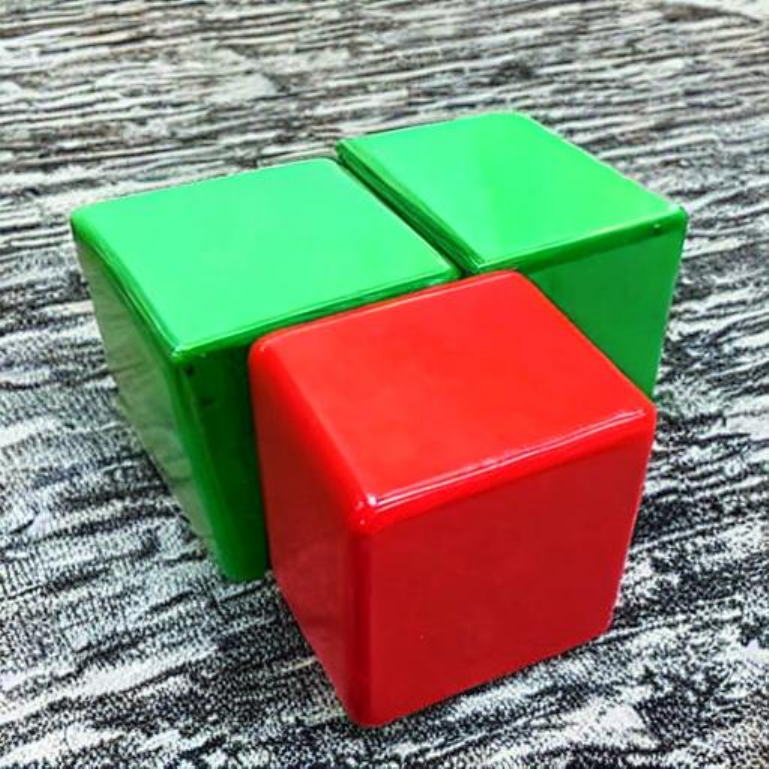} &
        \includegraphics[width=\linewidth]{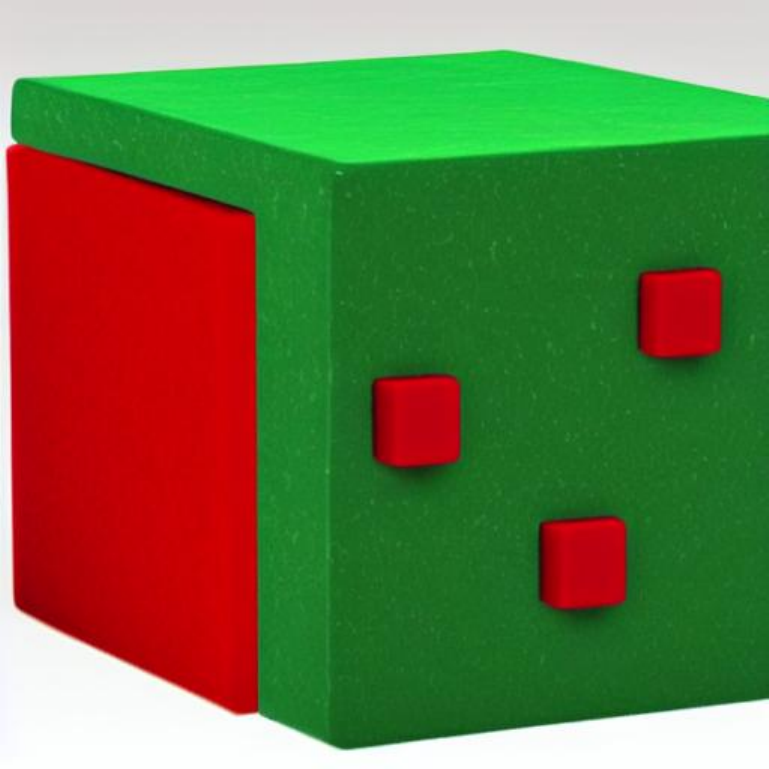} &
        \includegraphics[width=\linewidth]{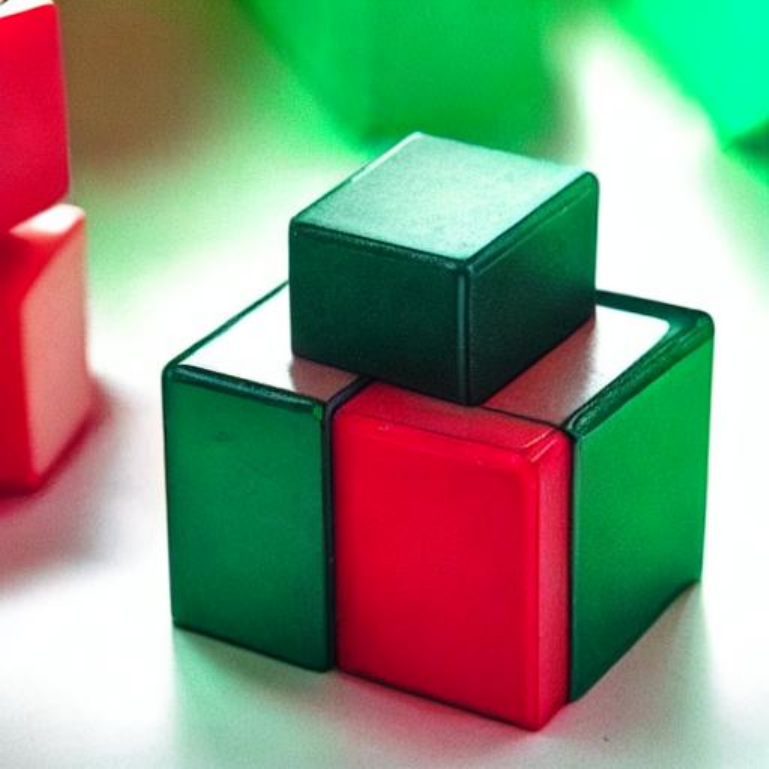} &
        \includegraphics[width=\linewidth]{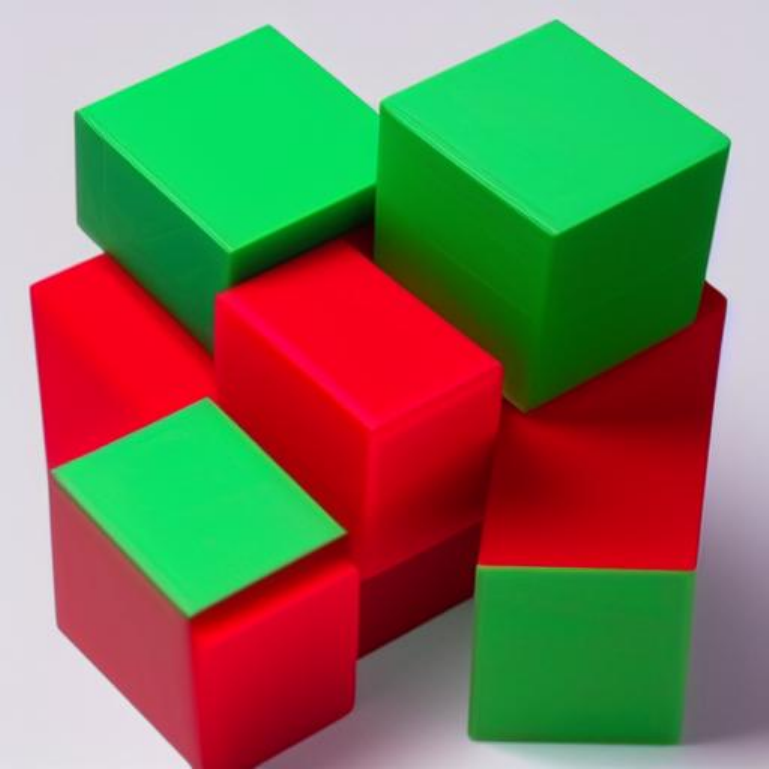} &
        \includegraphics[width=\linewidth]{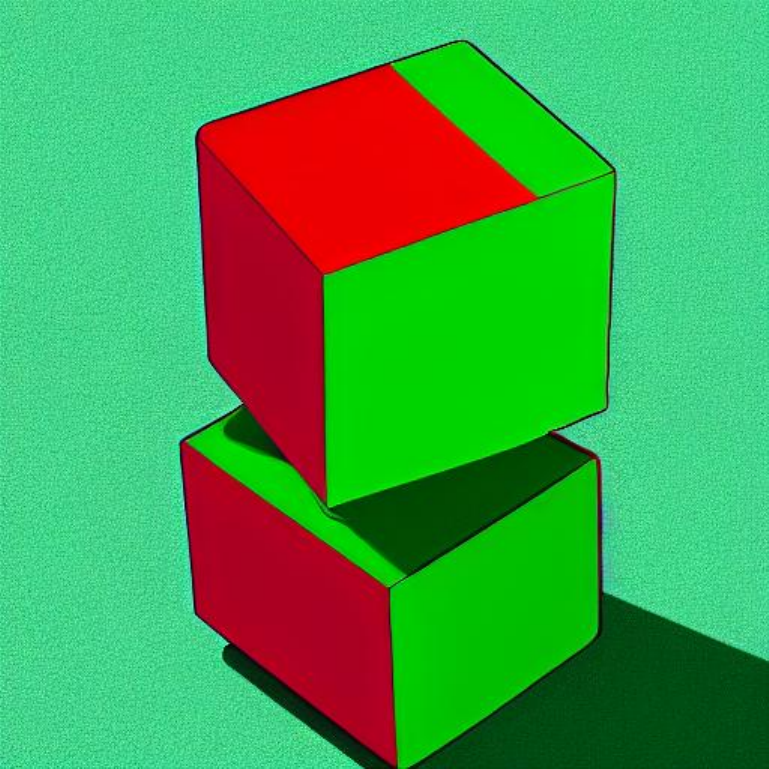} &
        \includegraphics[width=\linewidth]{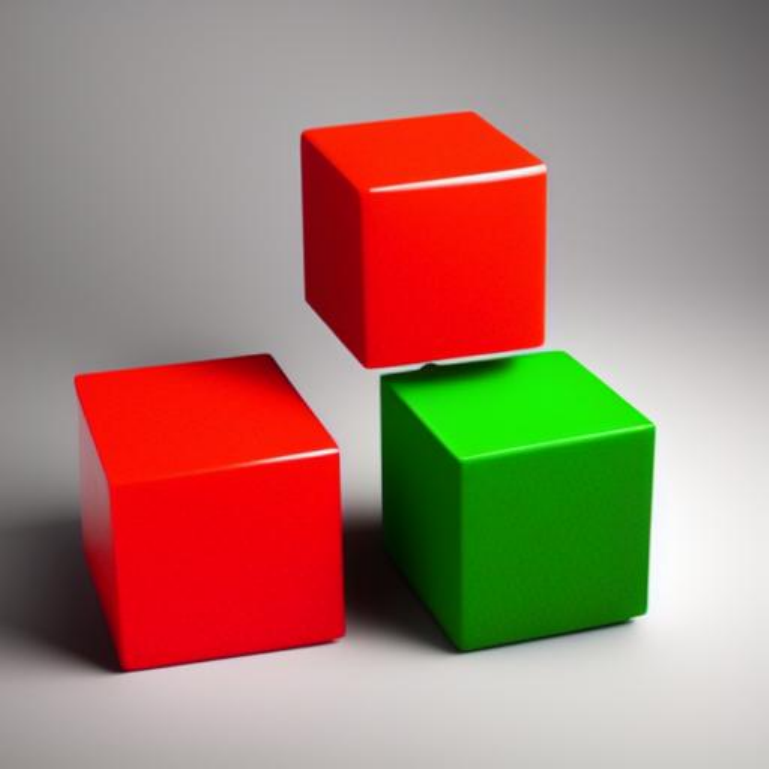} \\

        \multicolumn{8}{@{}p{\linewidth}@{}}{\centering \small \textit{Prompt: A stack of 3 cubes. A red cube is on the top, sitting on a red cube. The red cube is in the middle, sitting on a green cube. The green cube is on the bottom.}} \\
		\midrule

        \includegraphics[width=\linewidth]{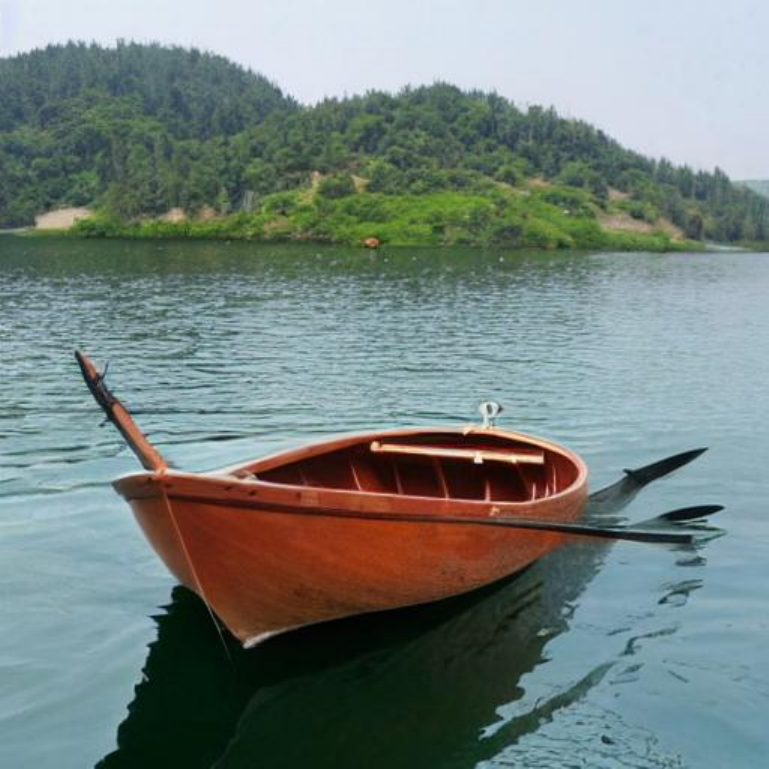} &
        \includegraphics[width=\linewidth]{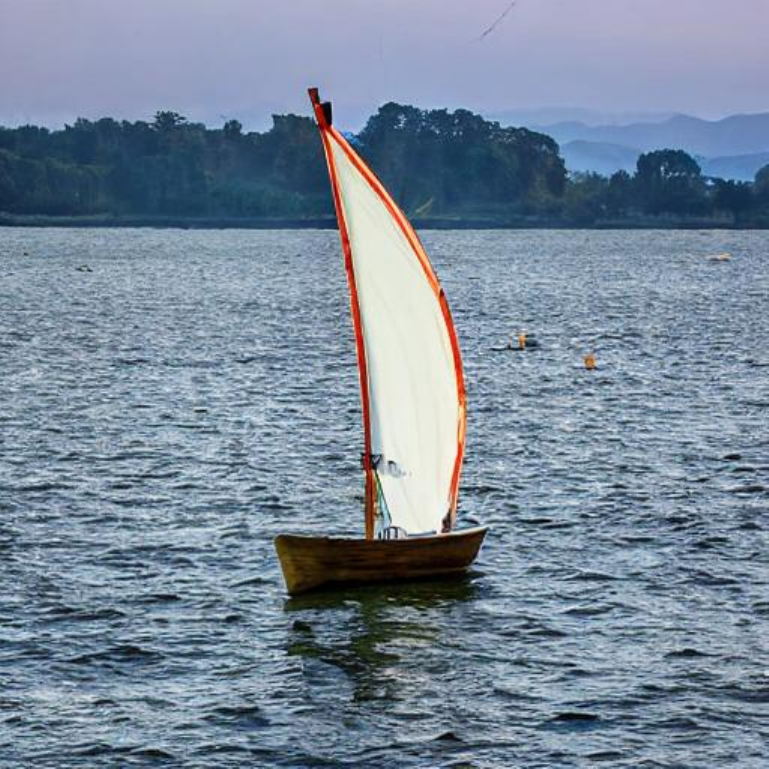} &
        \includegraphics[width=\linewidth]{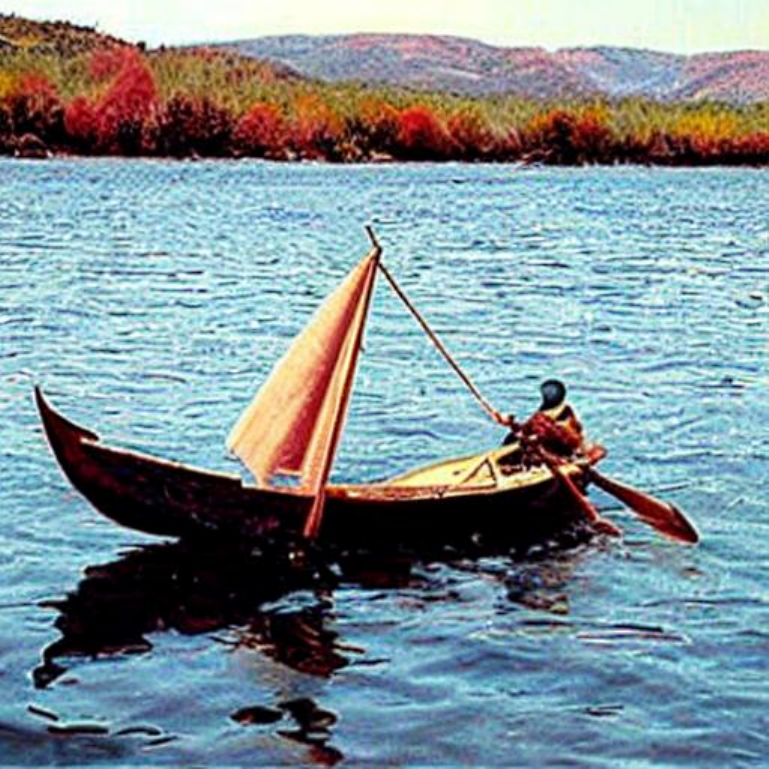} &
        \includegraphics[width=\linewidth]{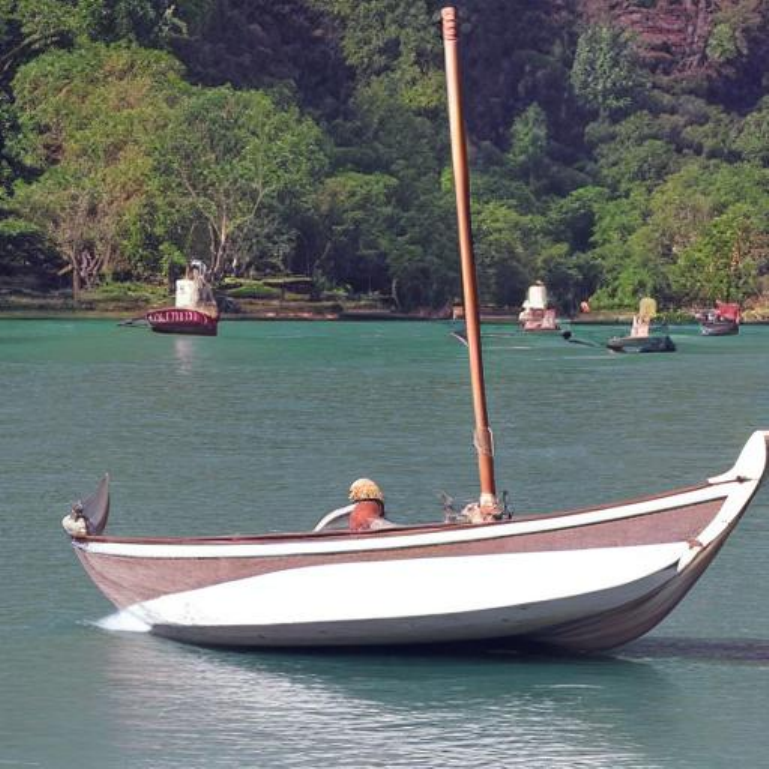} &
        \includegraphics[width=\linewidth]{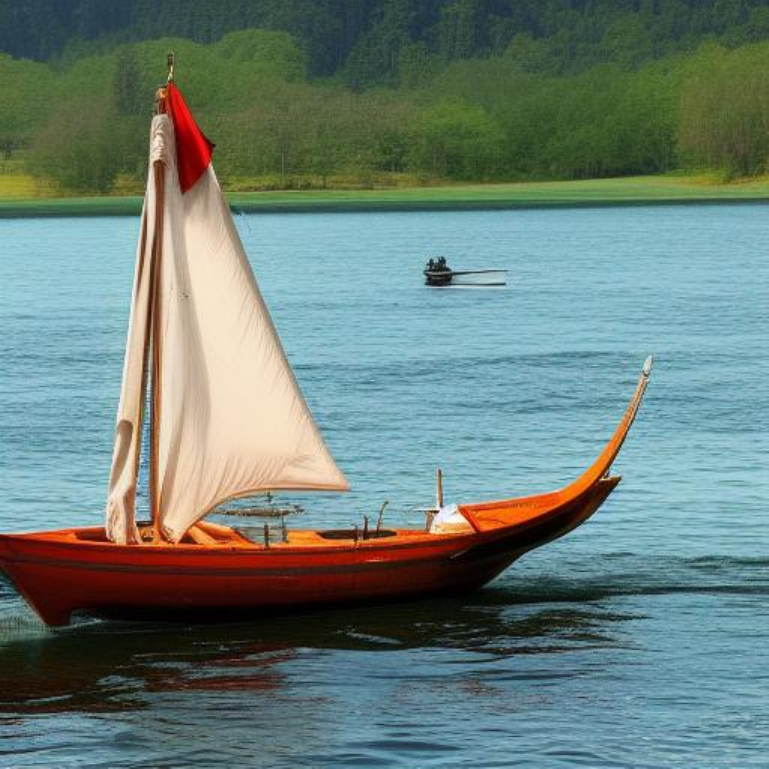} &
        \includegraphics[width=\linewidth]{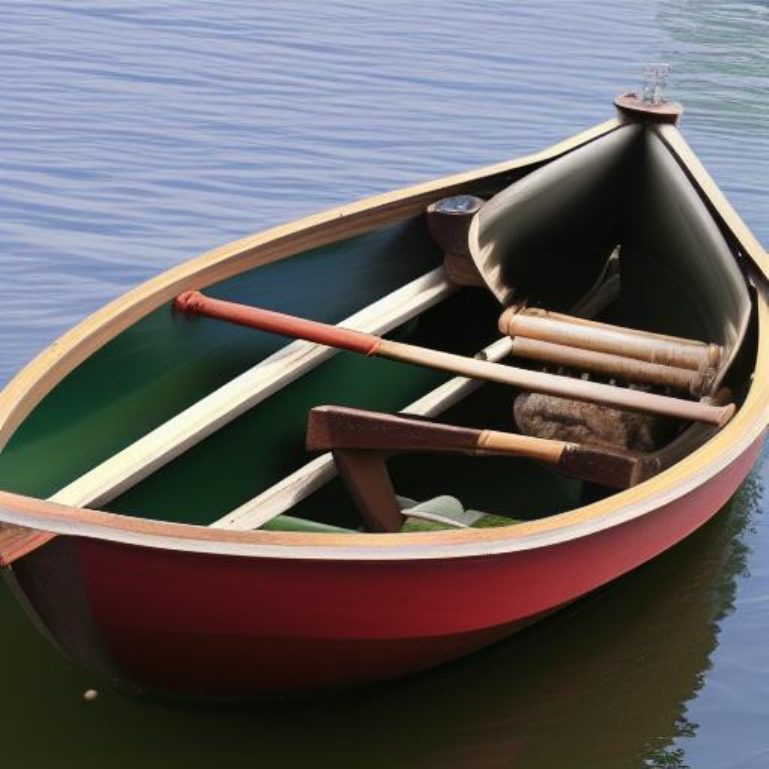} &
        \includegraphics[width=\linewidth]{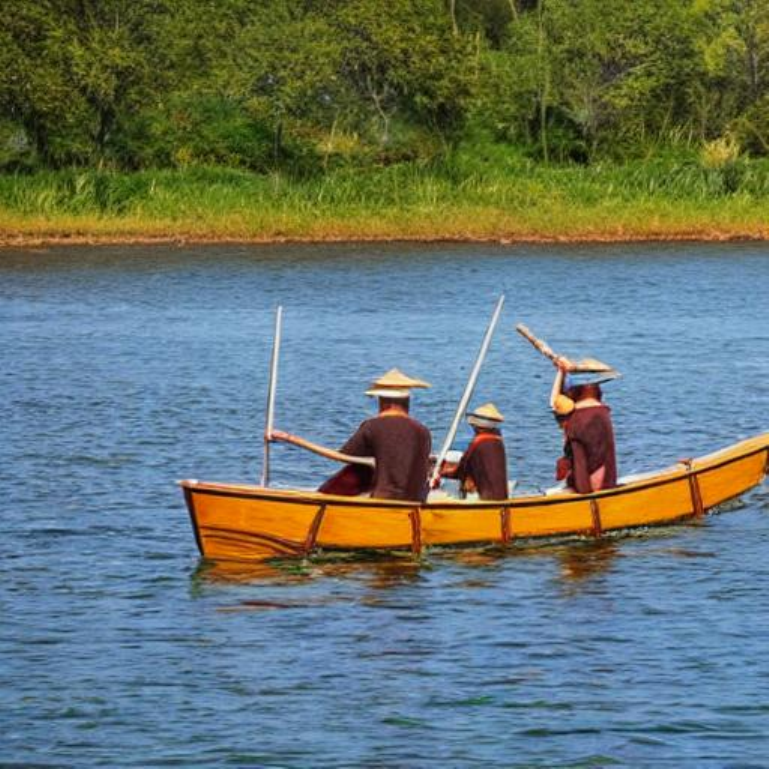} &
        \includegraphics[width=\linewidth]{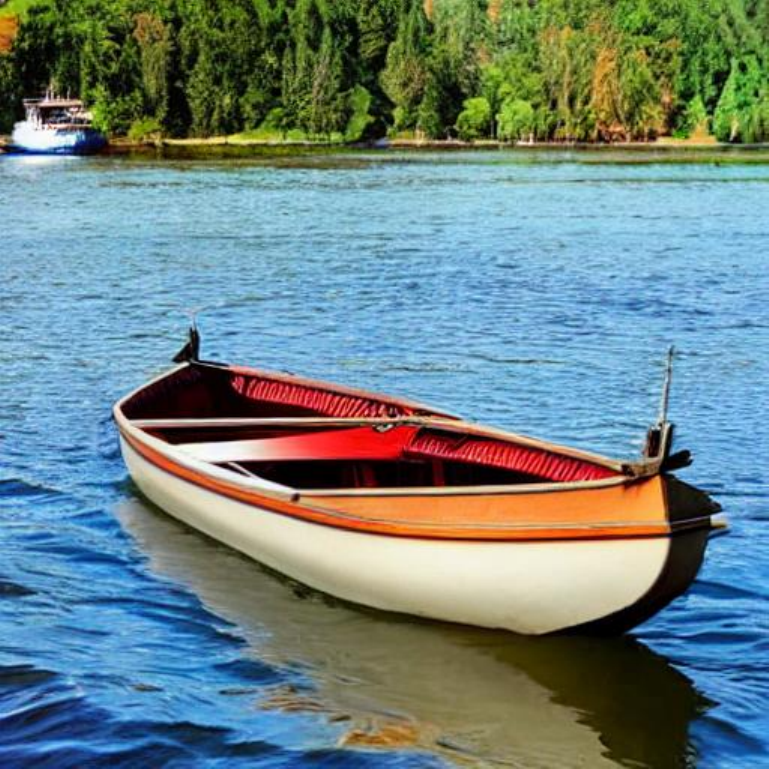} \\

        \multicolumn{8}{@{}p{\linewidth}@{}}{\centering \small \textit{Prompt: A small vessel propelled on water by oars, sails, or an engine.}} \\
		\midrule

        \includegraphics[width=\linewidth]{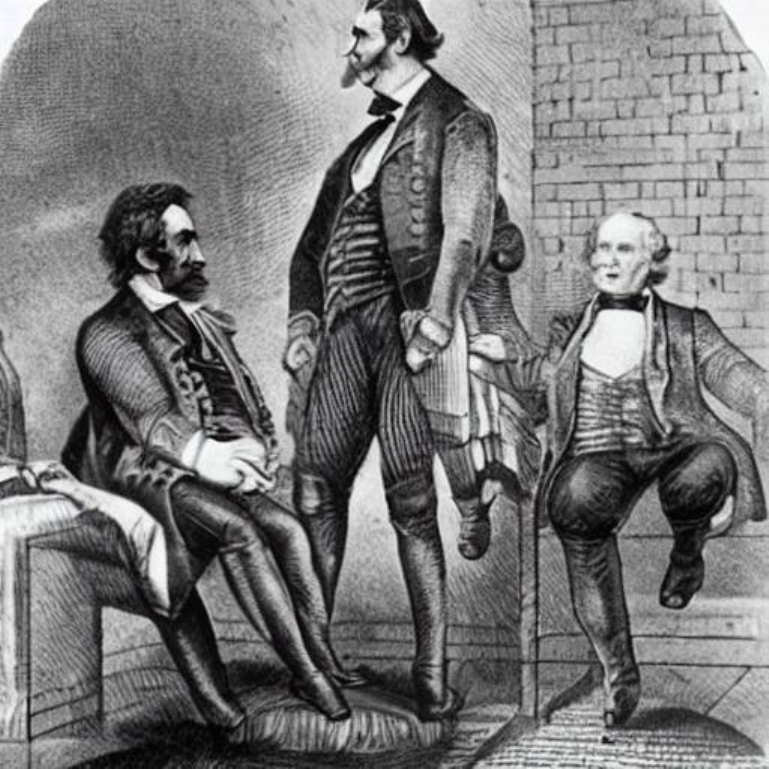} &
        \includegraphics[width=\linewidth]{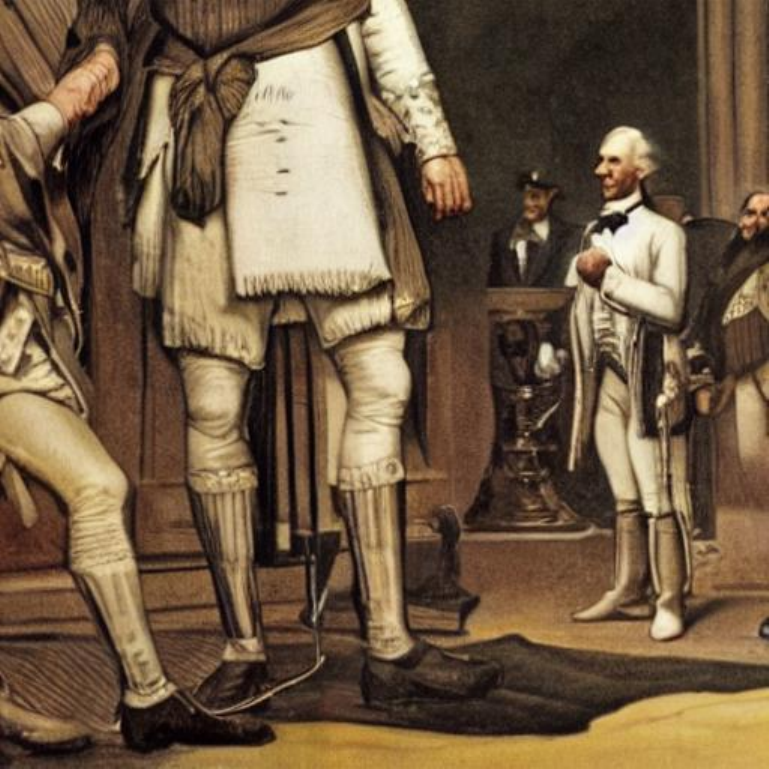} &
        \includegraphics[width=\linewidth]{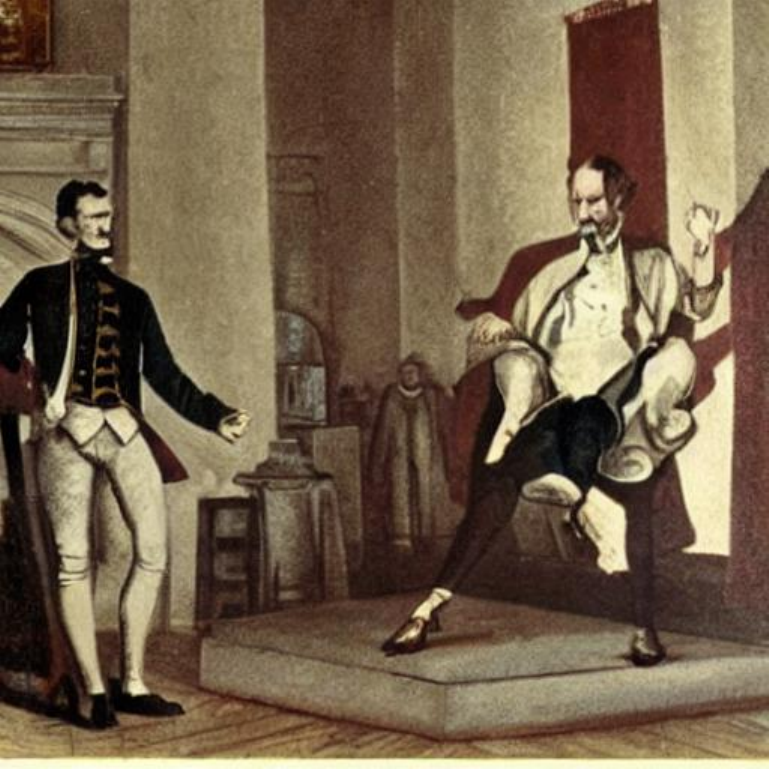} &
        \includegraphics[width=\linewidth]{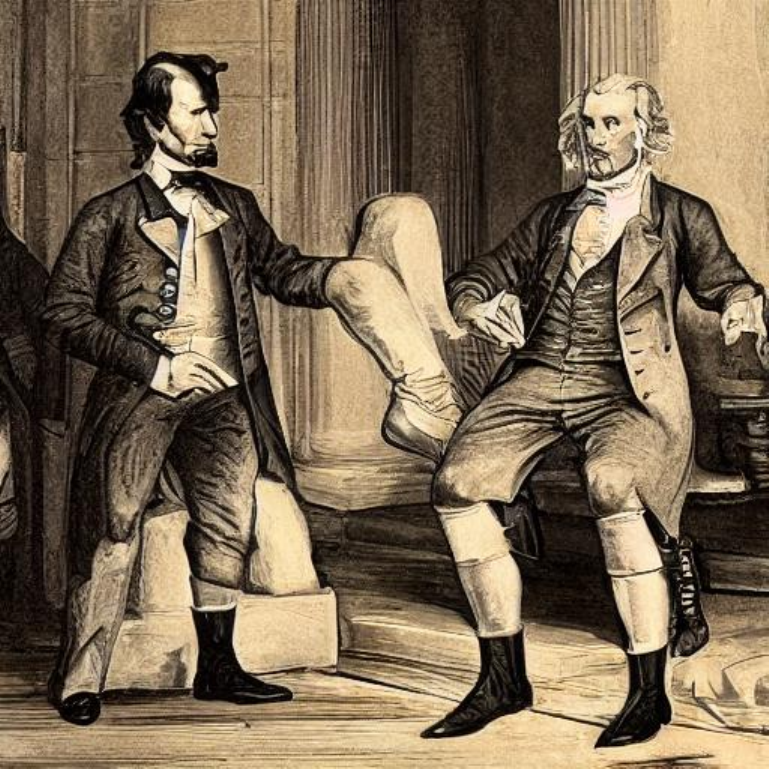} &
        \includegraphics[width=\linewidth]{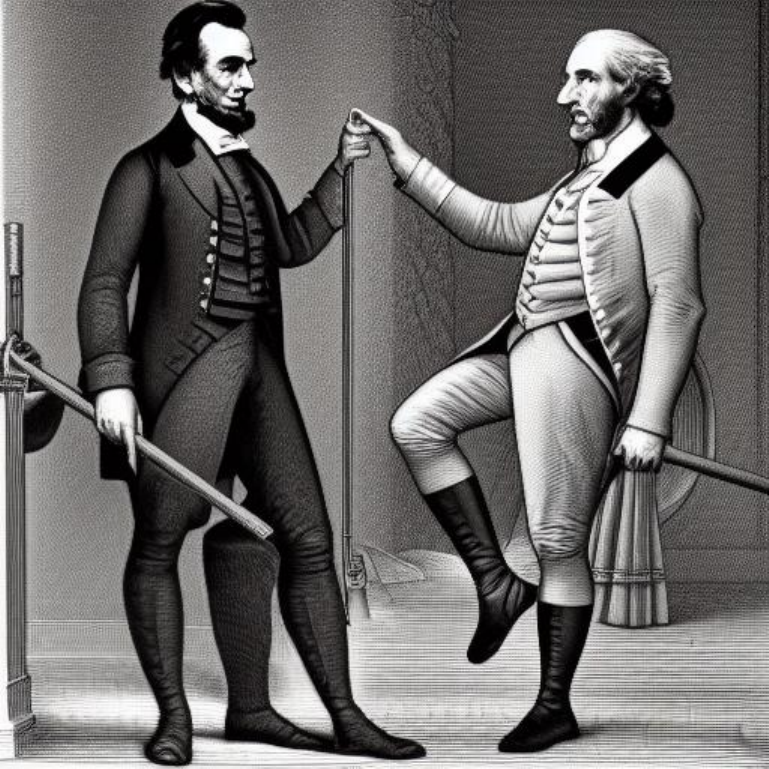} &
        \includegraphics[width=\linewidth]{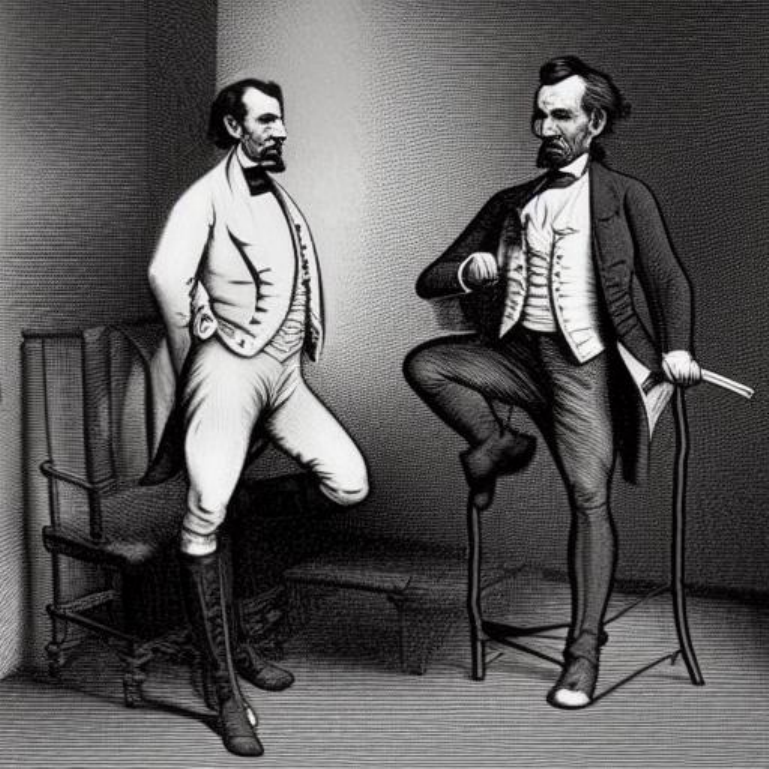} &
        \includegraphics[width=\linewidth]{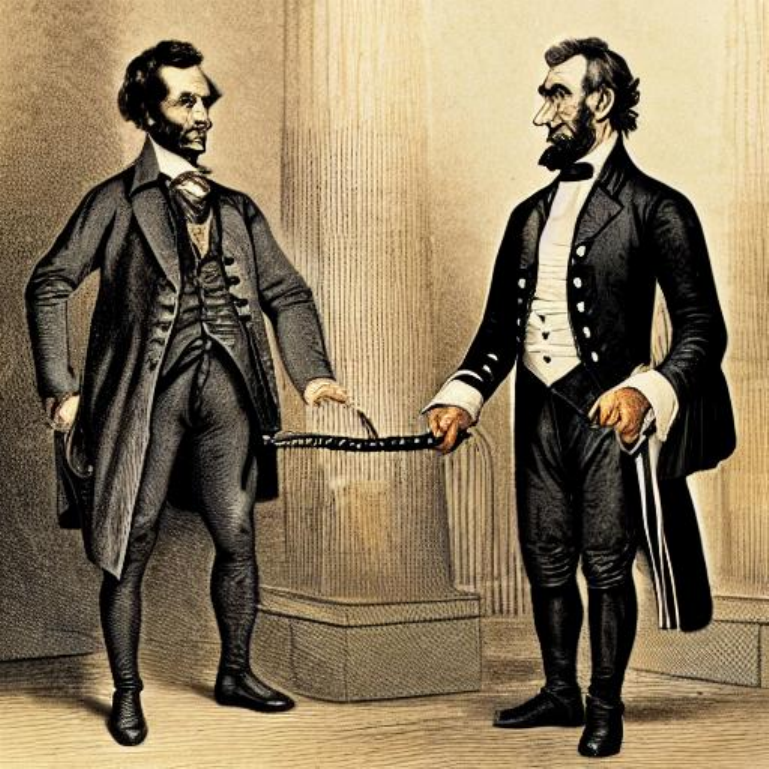} &
        \includegraphics[width=\linewidth]{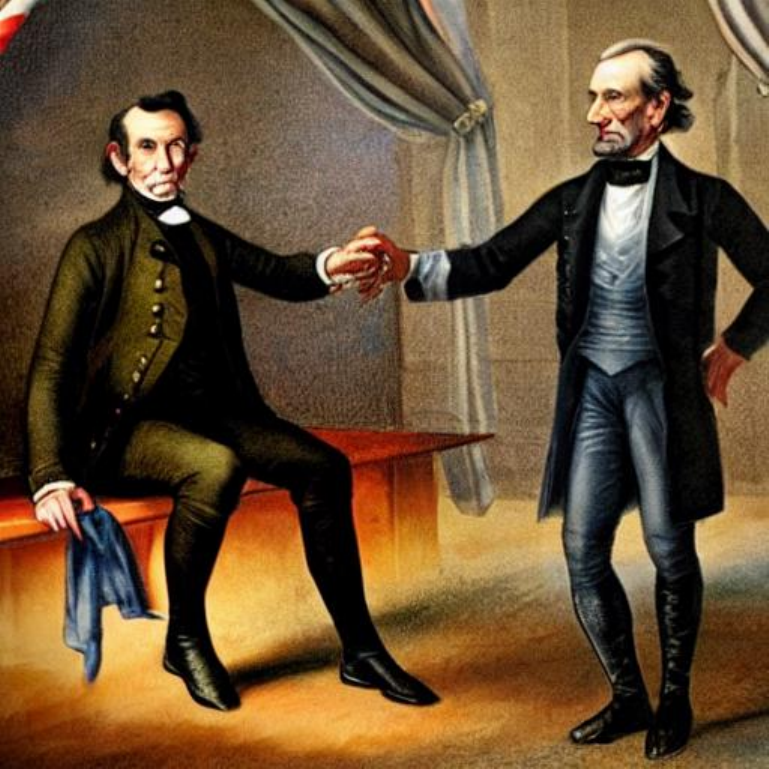} \\

        \multicolumn{8}{@{}p{\linewidth}@{}}{\centering \small \textit{Prompt: Abraham Lincoln touches his toes while George Washington does chin-ups. Lincoln is barefoot. Washington is wearing boots.}} \\
		\midrule

        \includegraphics[width=\linewidth]{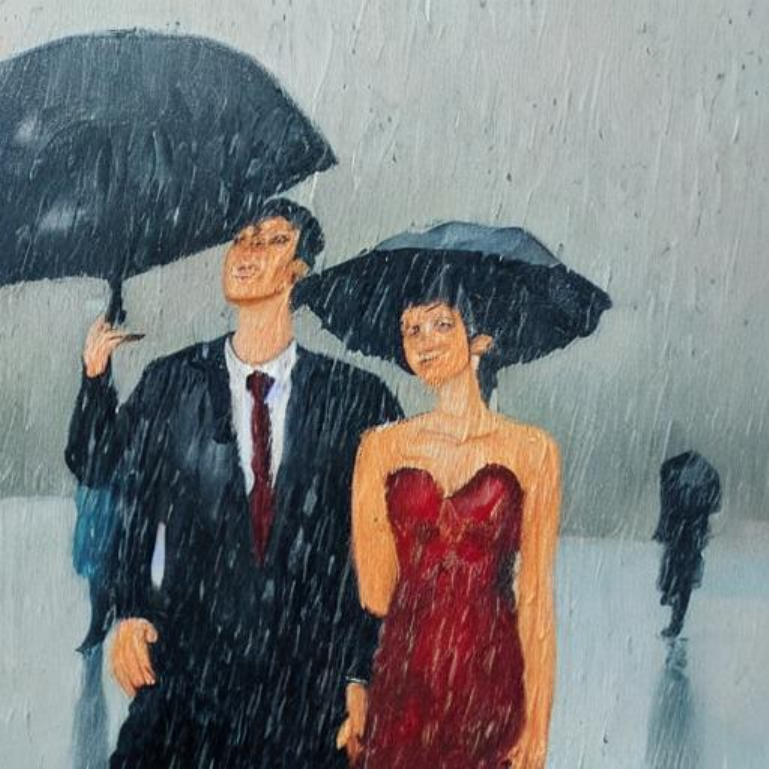} &
        \includegraphics[width=\linewidth]{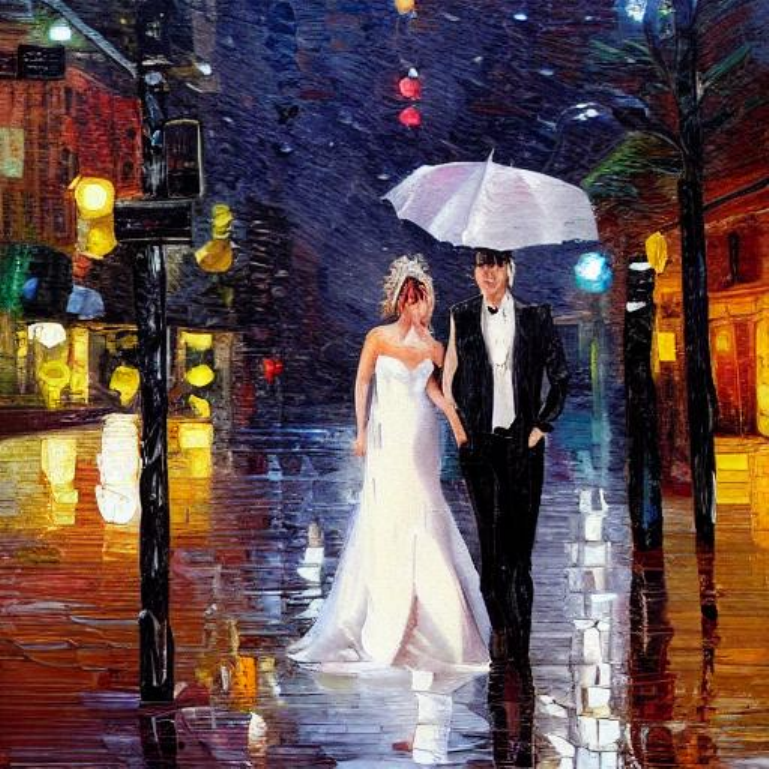} &
        \includegraphics[width=\linewidth]{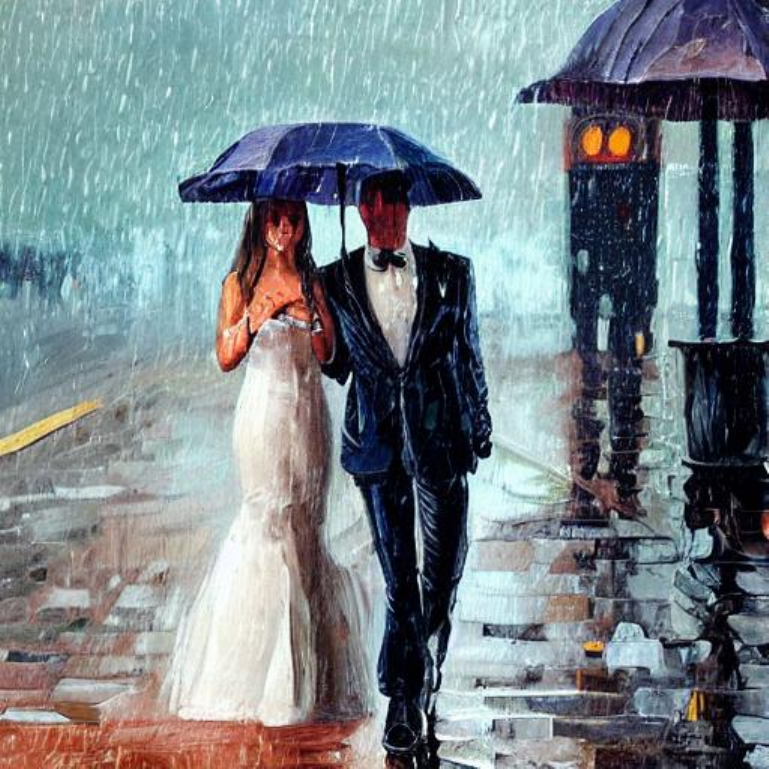} &
        \includegraphics[width=\linewidth]{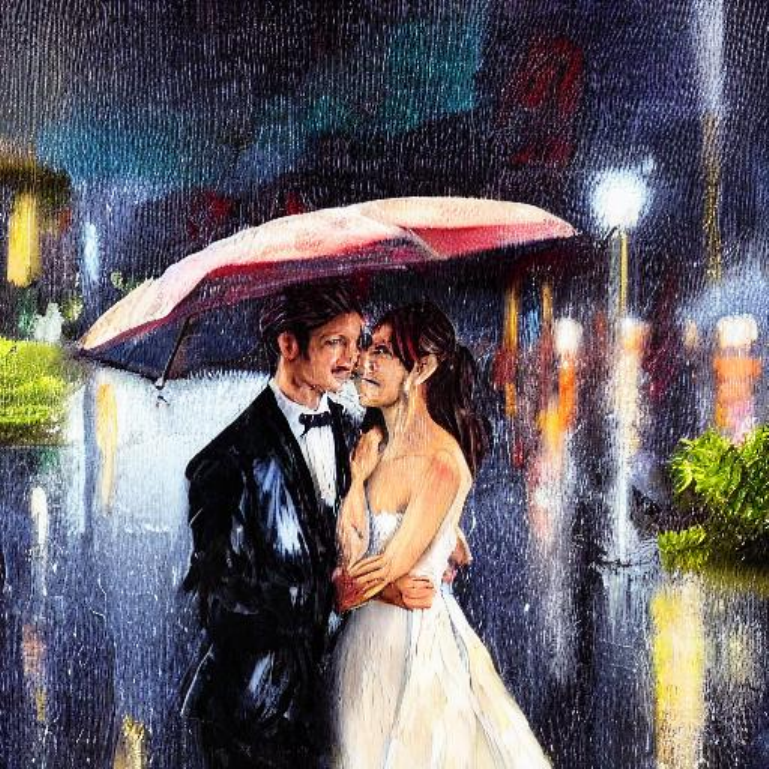} &
        \includegraphics[width=\linewidth]{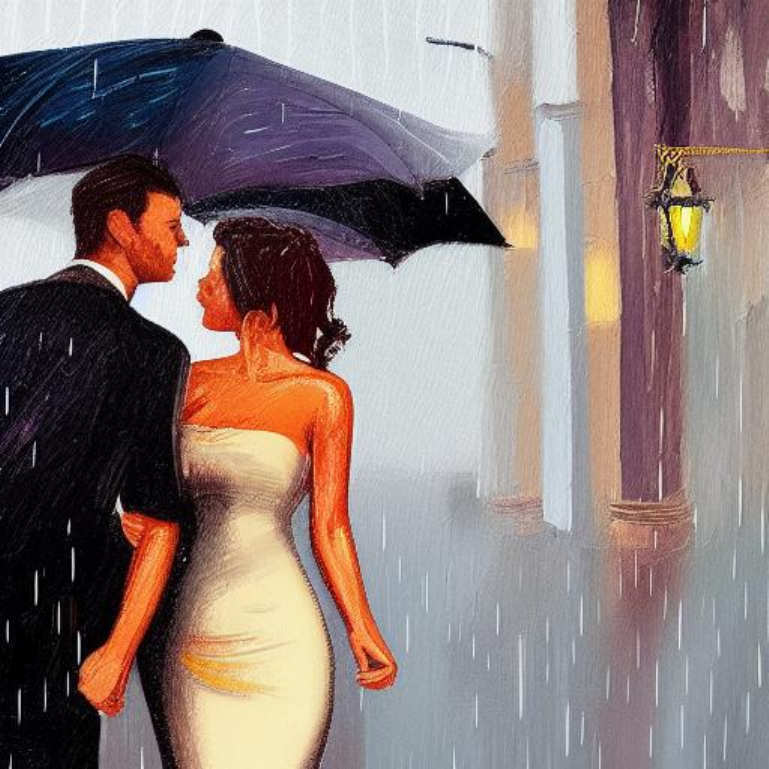} &
        \includegraphics[width=\linewidth]{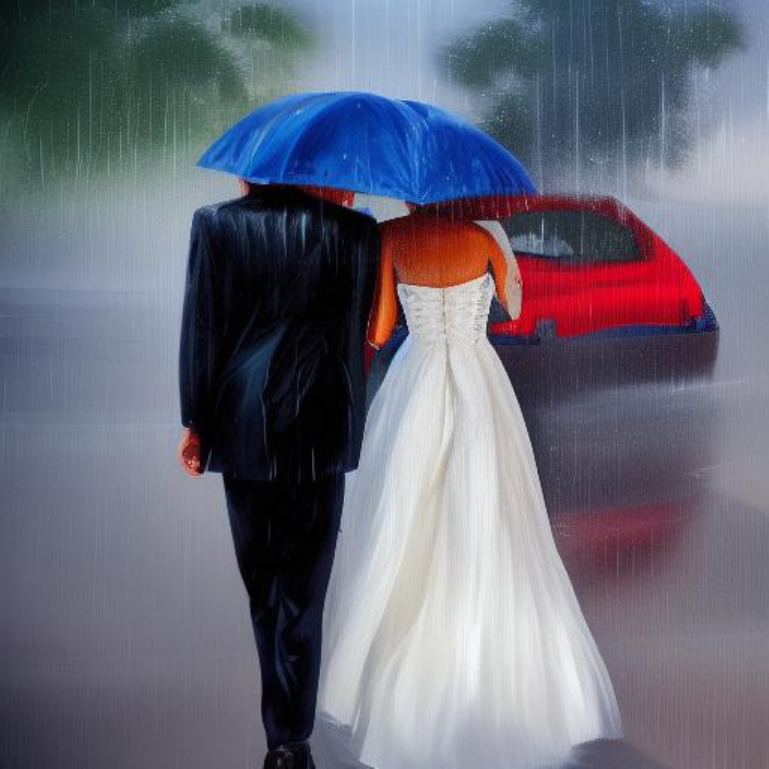} &
        \includegraphics[width=\linewidth]{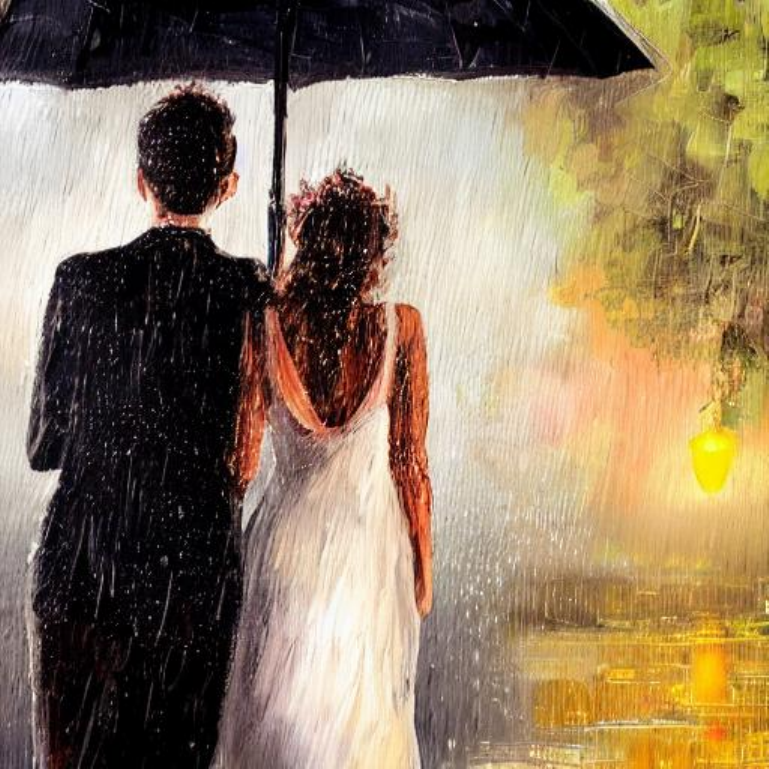} &
        \includegraphics[width=\linewidth]{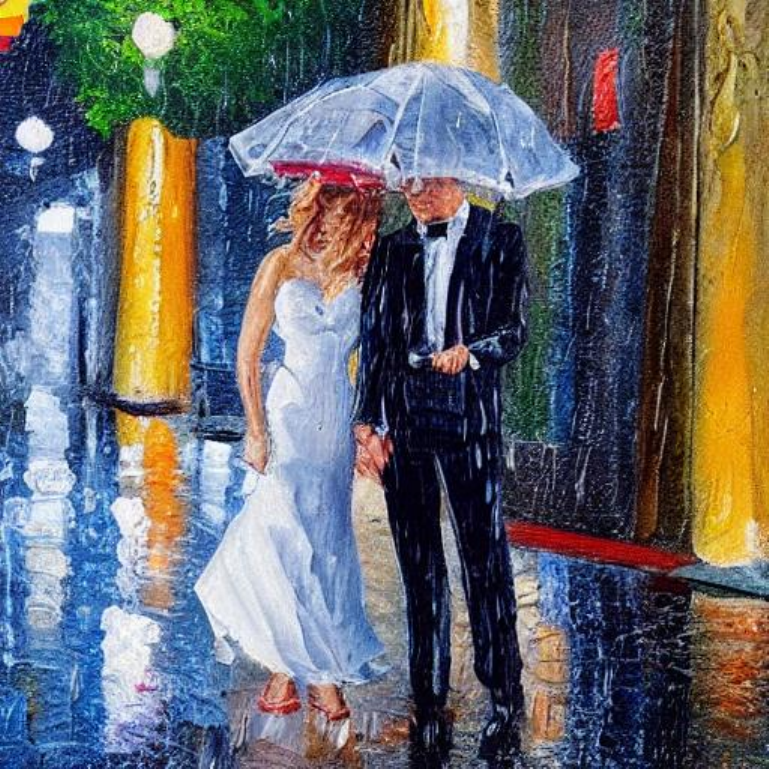} \\

        \multicolumn{8}{@{}p{\linewidth}@{}}{\centering \small \textit{Prompt: An oil painting of a couple in formal evening wear going home get caught in a heavy downpour with no umbrellas.}} \\

        \bottomrule
        
    \end{tabularx}
    \label{fig:qualitative_sdv1} 
    \caption{Visual comparison of SD v1.5 with $\text{NRE}=200$, targeting the HPSv2 reward model.} 
\end{figure}

\begin{figure}[t] 
    \centering
    \setlength{\tabcolsep}{1pt} 
    
    \begin{tabularx}{\linewidth}{YYYYYYYY}
        
        \toprule
        \centering \textbf{SDXL} & 
        \centering \textbf{BoN} & 
        \centering \textbf{ZO-N} & 
        \centering \textbf{SoP} & 
        \centering \textbf{SMC} & 
        \centering \textbf{SVDD} & 
        \centering \textbf{Demon} & 
        \centering \textbf{SES} \tabularnewline
        \midrule
        
        \includegraphics[width=\linewidth]{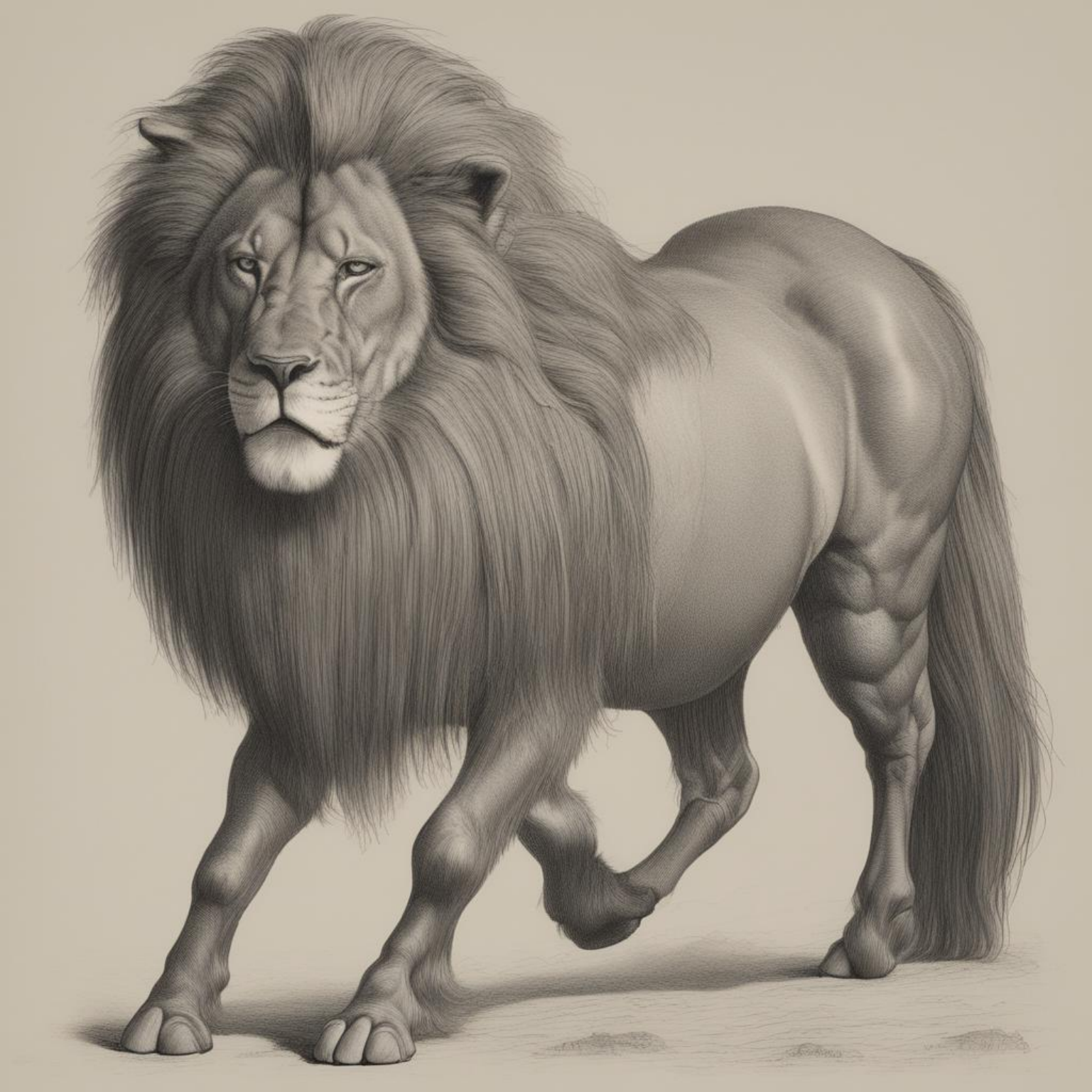} &
        \includegraphics[width=\linewidth]{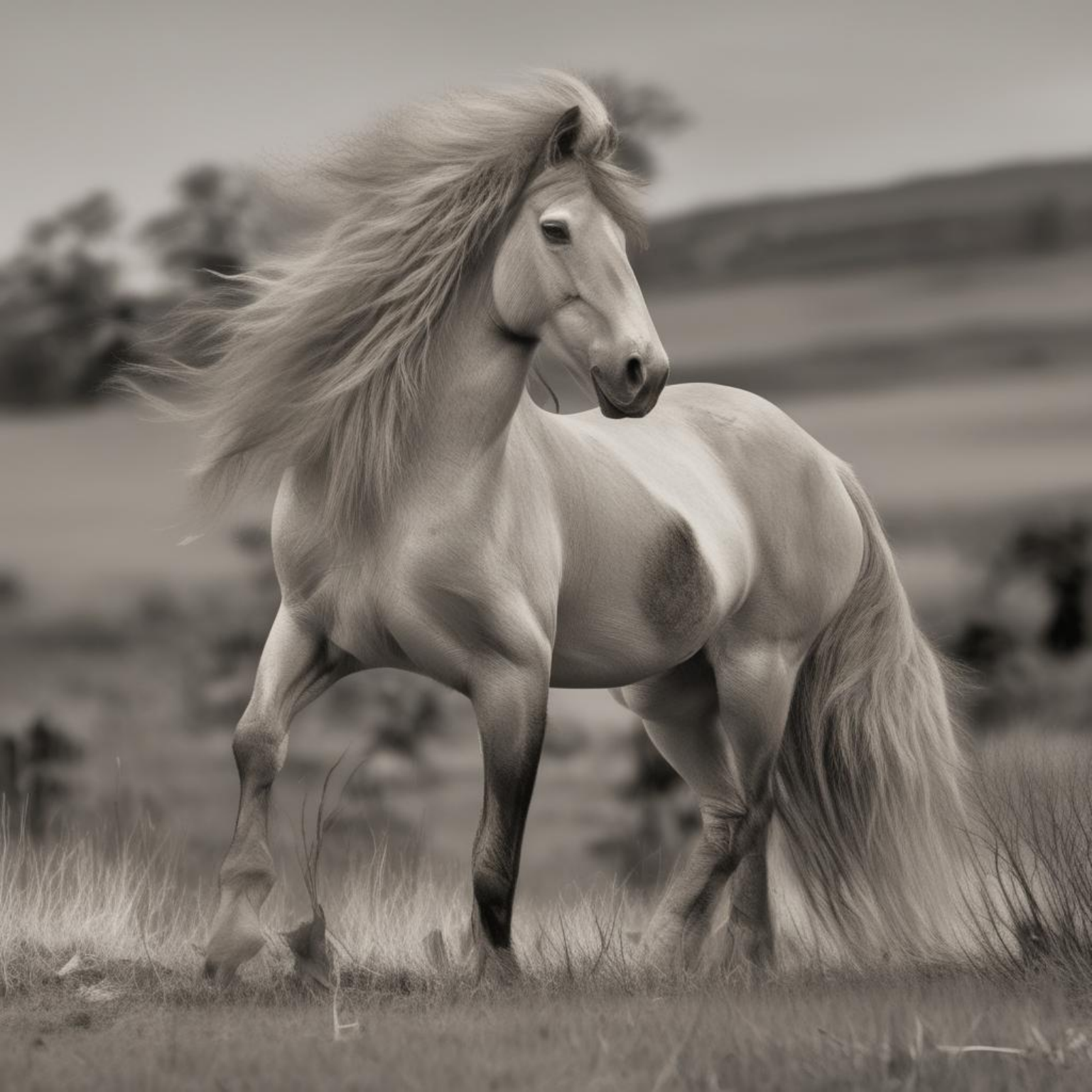} &
        \includegraphics[width=\linewidth]{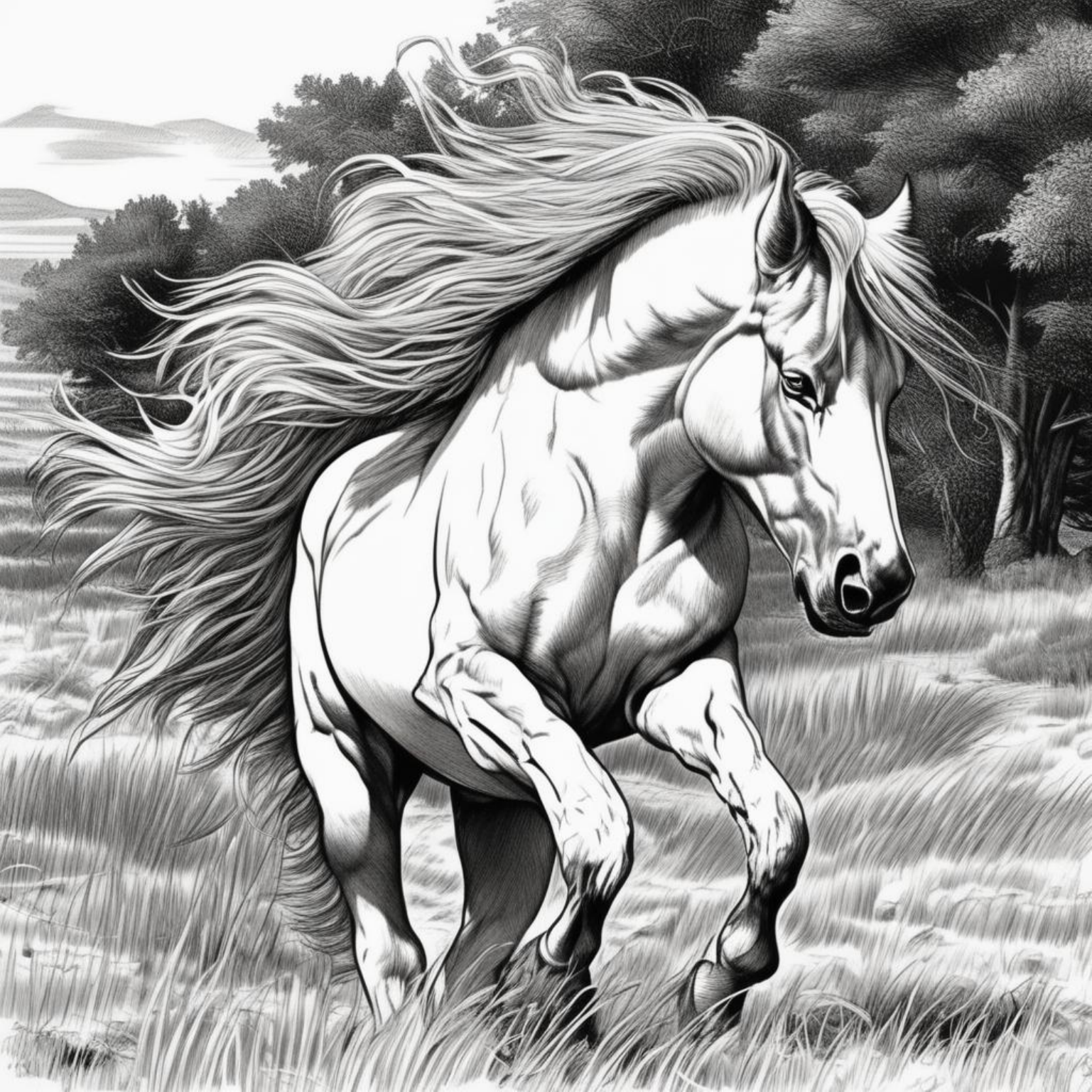} &
        \includegraphics[width=\linewidth]{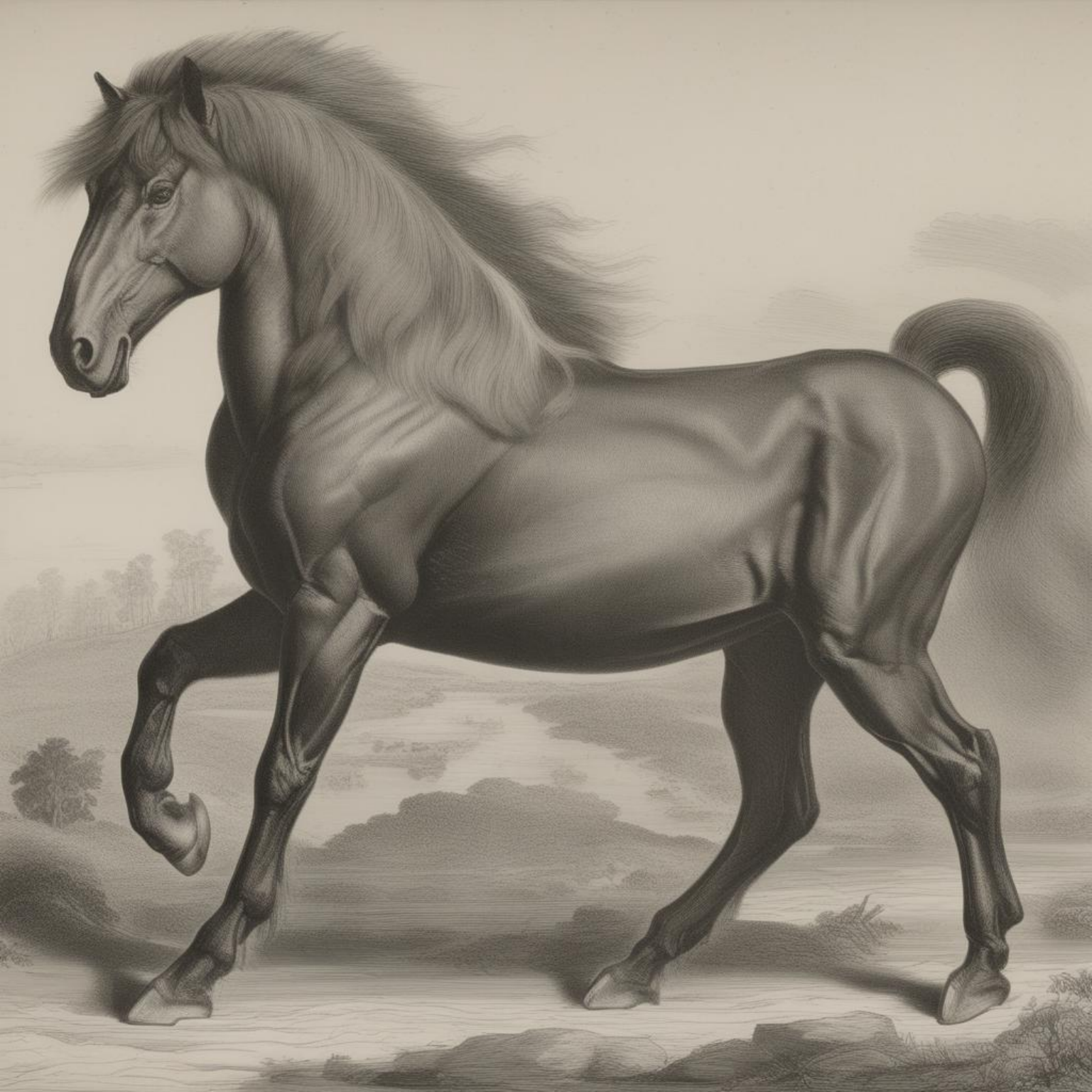} &
        \includegraphics[width=\linewidth]{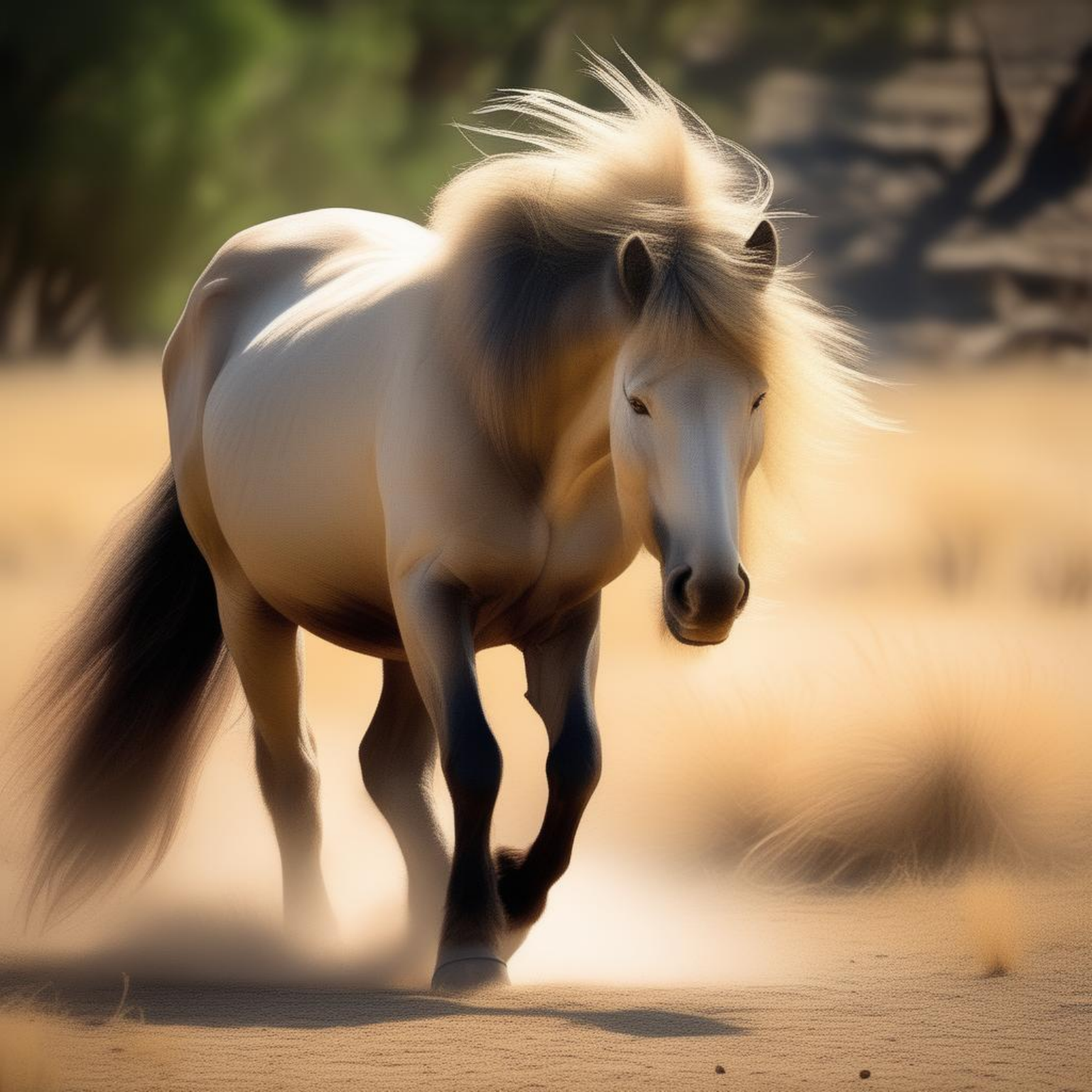} &
        \includegraphics[width=\linewidth]{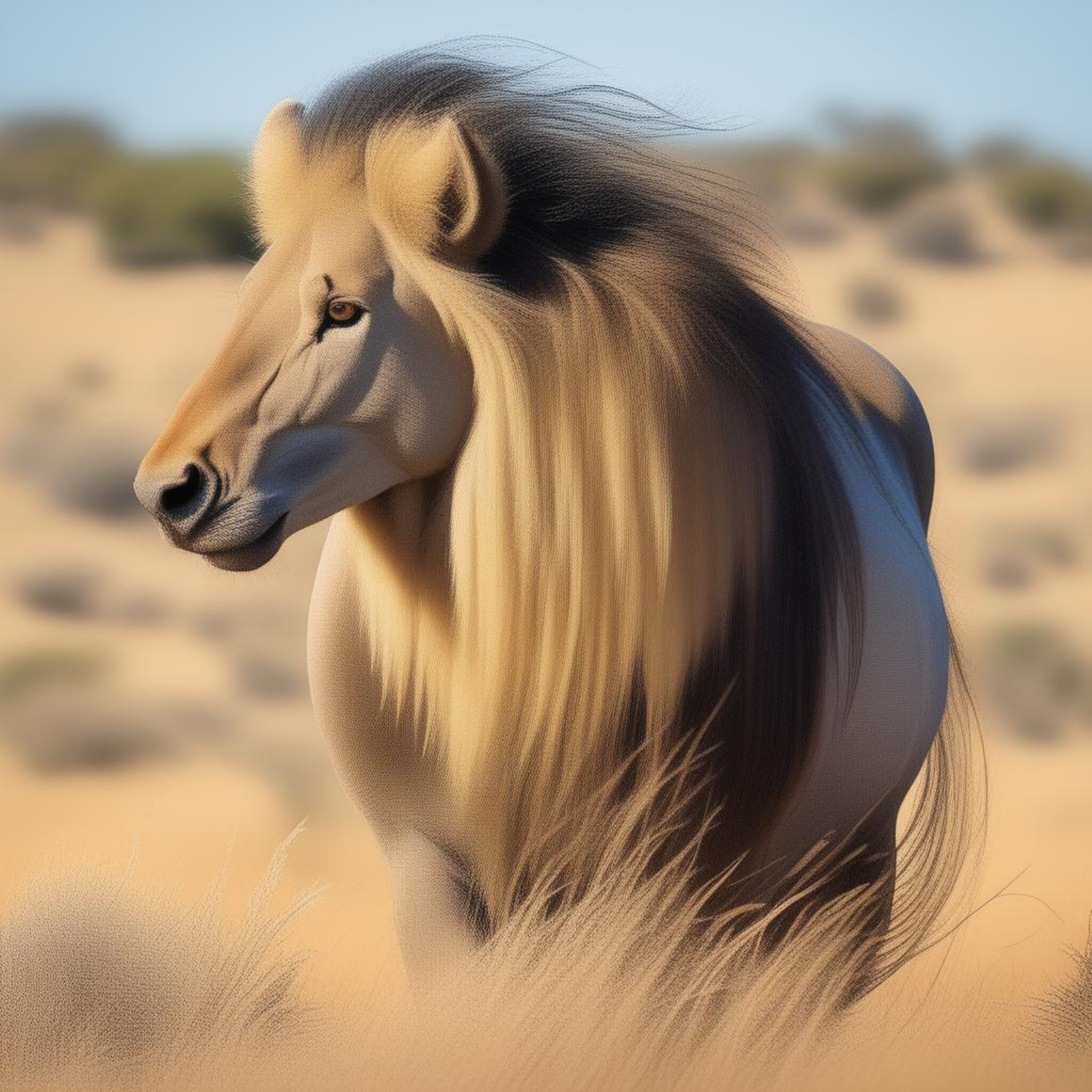} &
        \includegraphics[width=\linewidth]{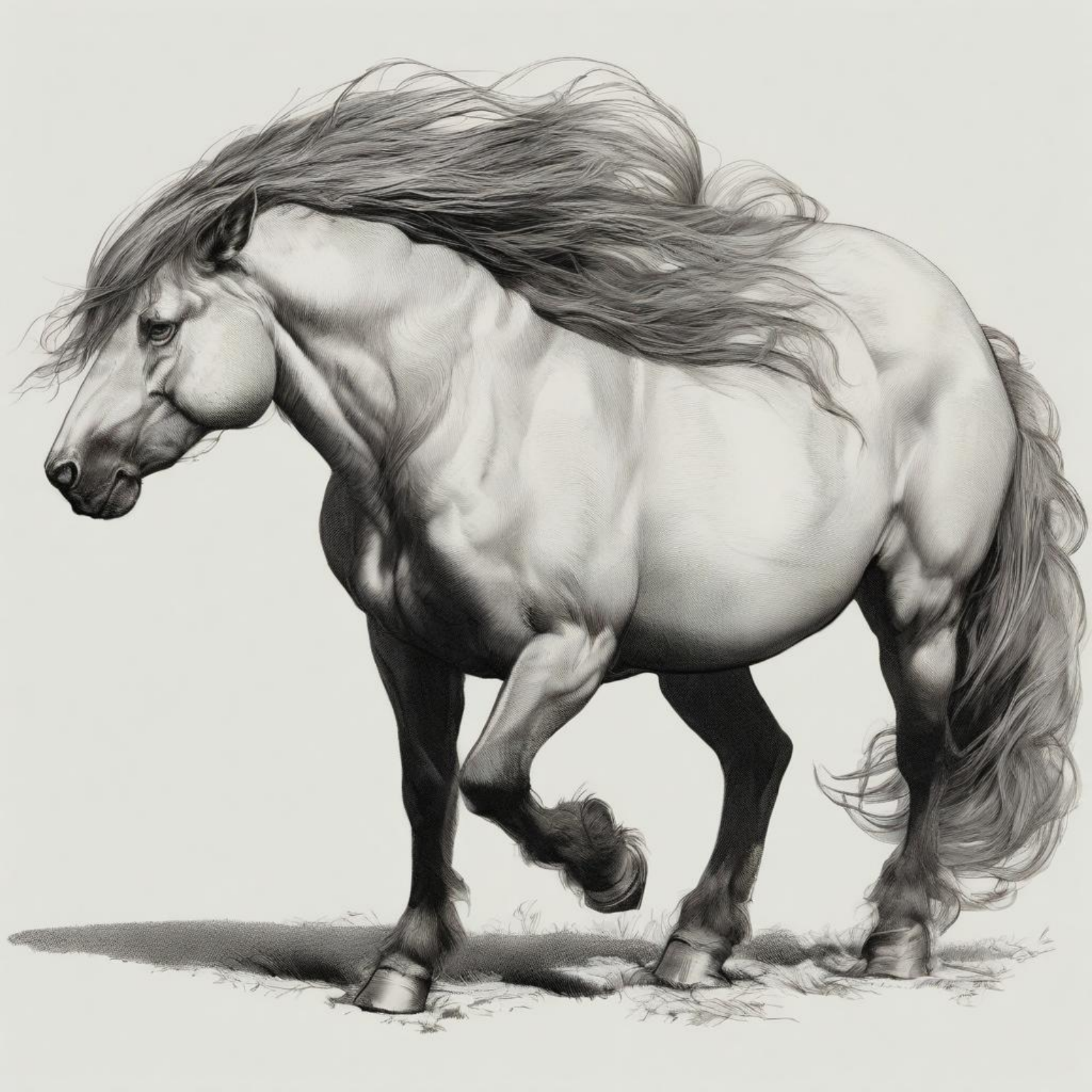} &
        \includegraphics[width=\linewidth]{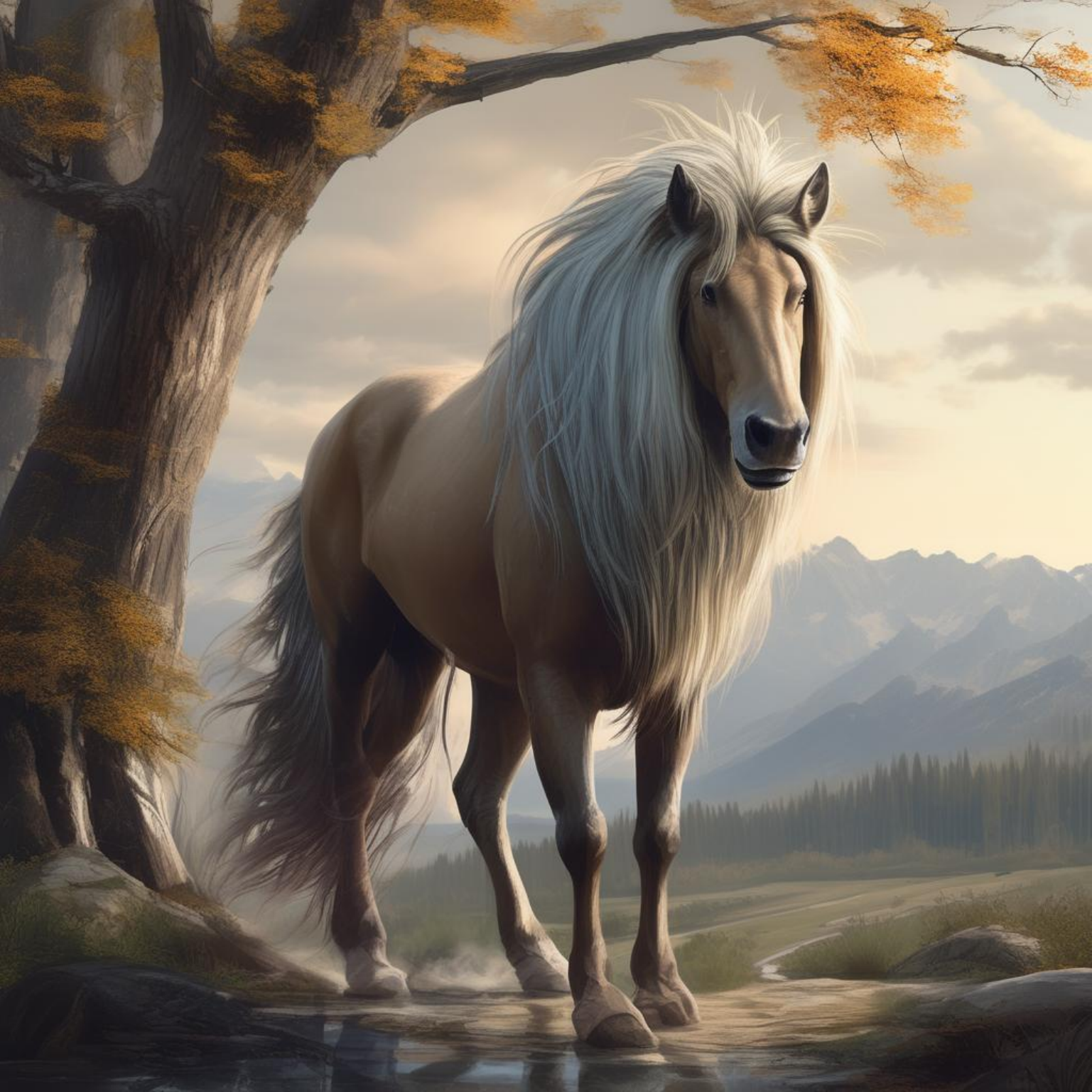} \\

        \multicolumn{8}{@{}p{\linewidth}@{}}{\centering \small \textit{Prompt: A large plant-eating domesticated mammal with solid hoofs and a flowing mane and tail, used for riding, racing, and to carry and pull loads.}} \\
		\midrule

        \includegraphics[width=\linewidth]{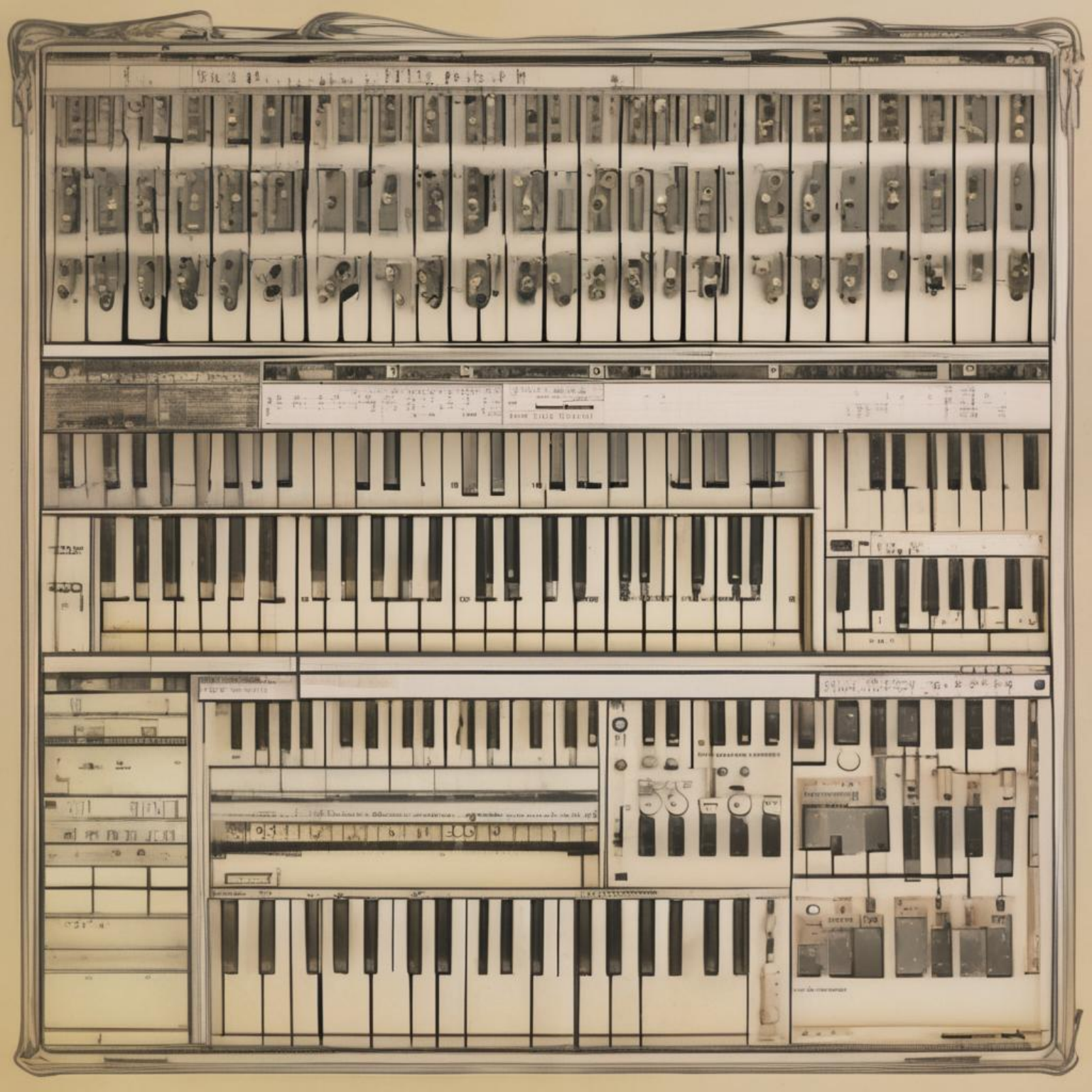} &
        \includegraphics[width=\linewidth]{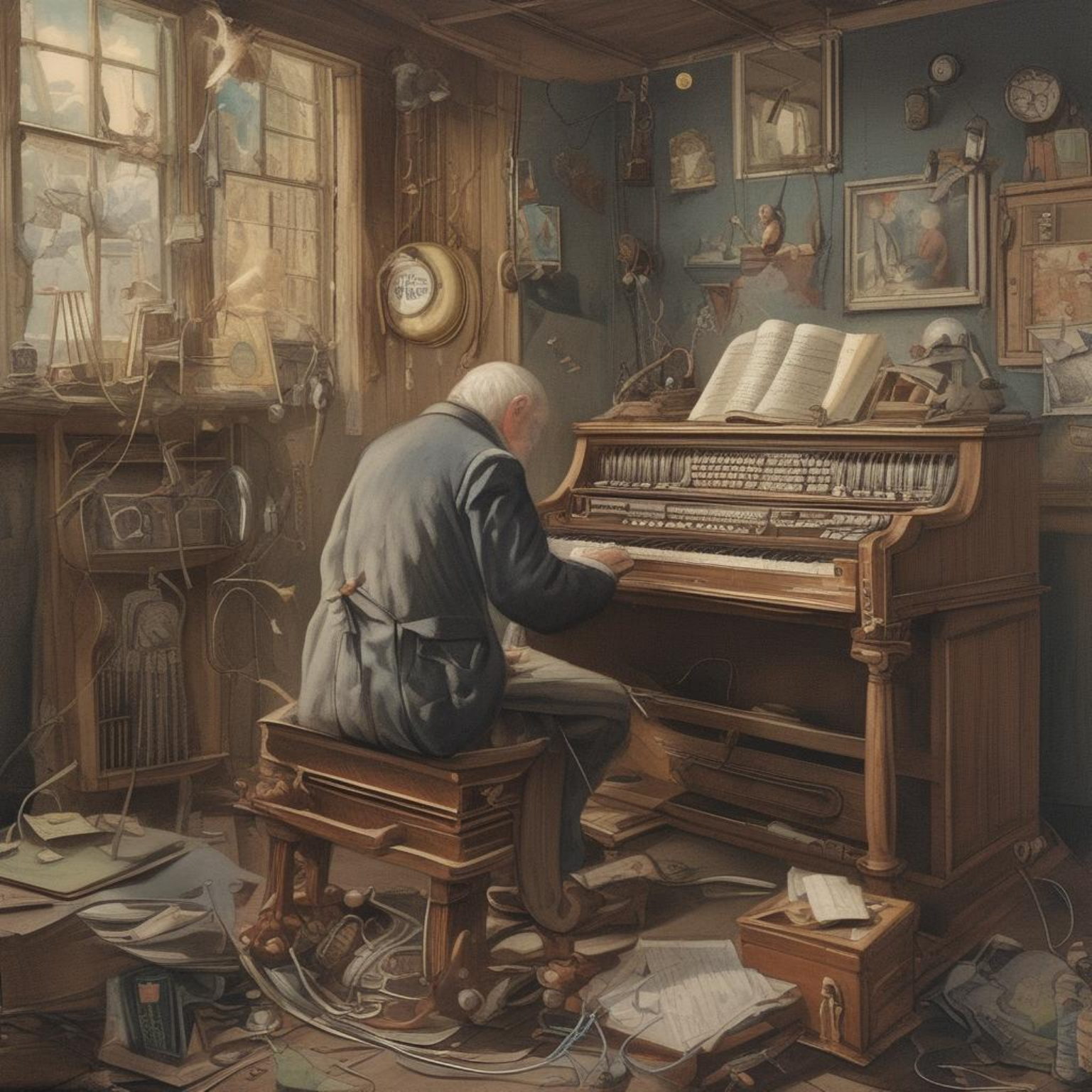} &
        \includegraphics[width=\linewidth]{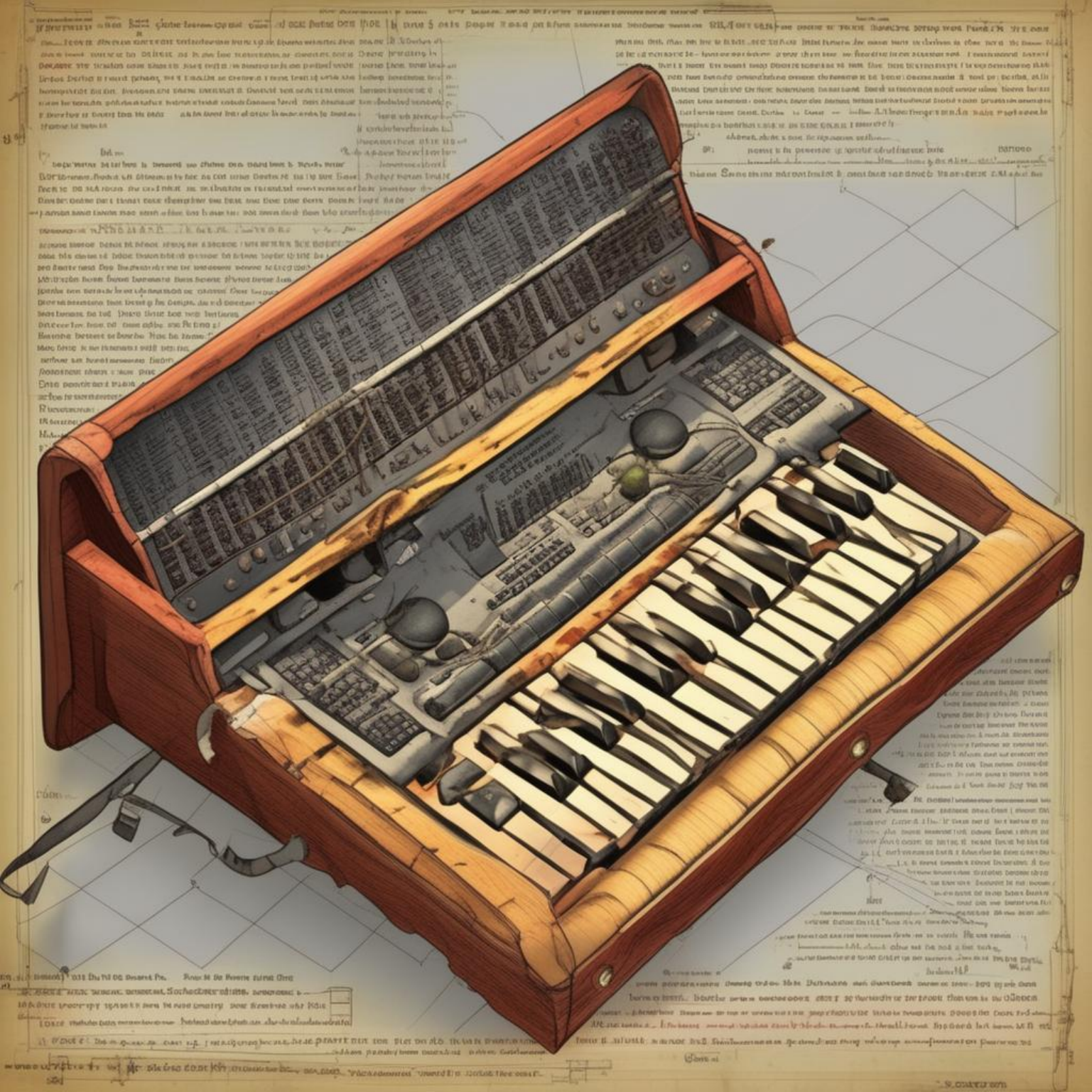} &
        \includegraphics[width=\linewidth]{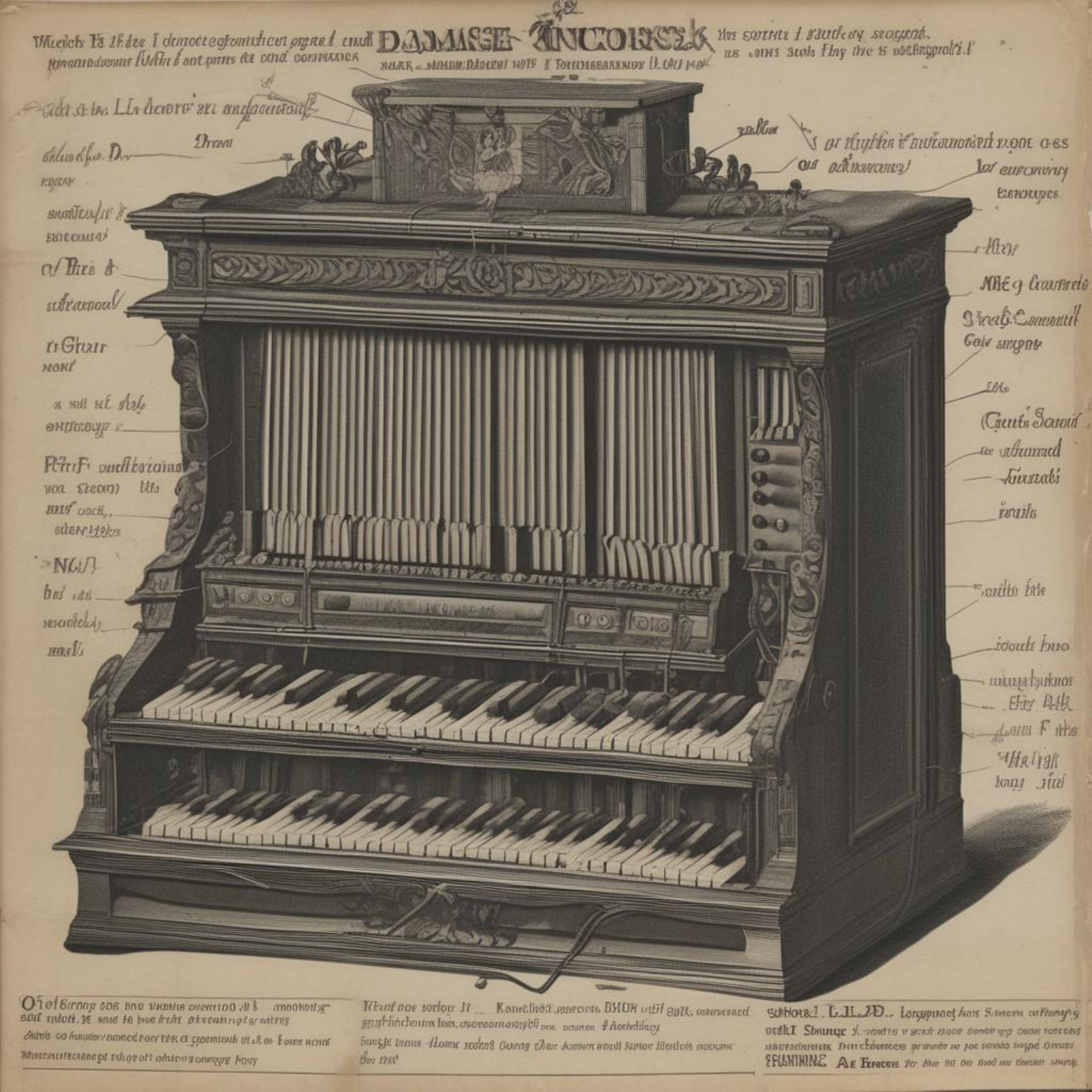} &
        \includegraphics[width=\linewidth]{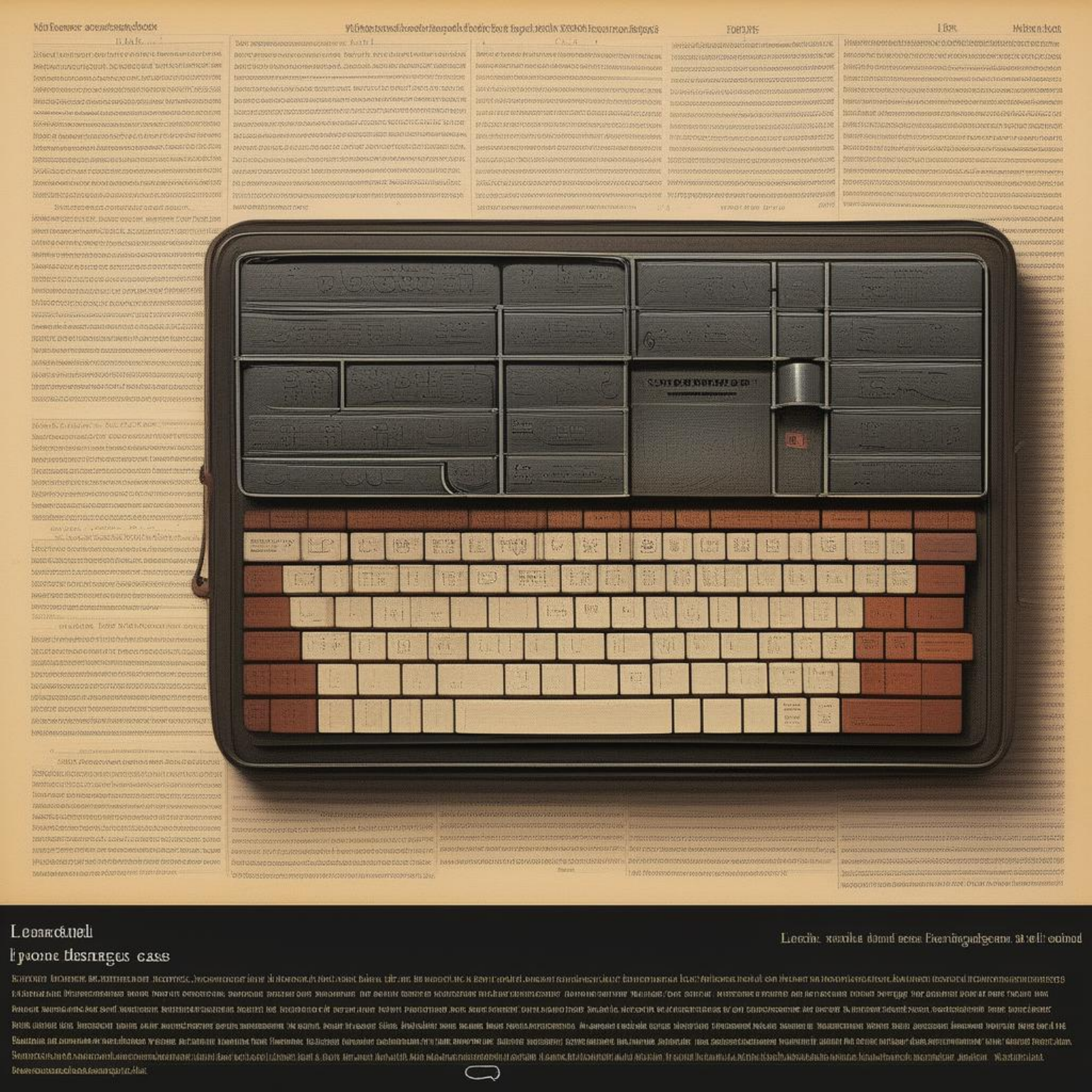} &
        \includegraphics[width=\linewidth]{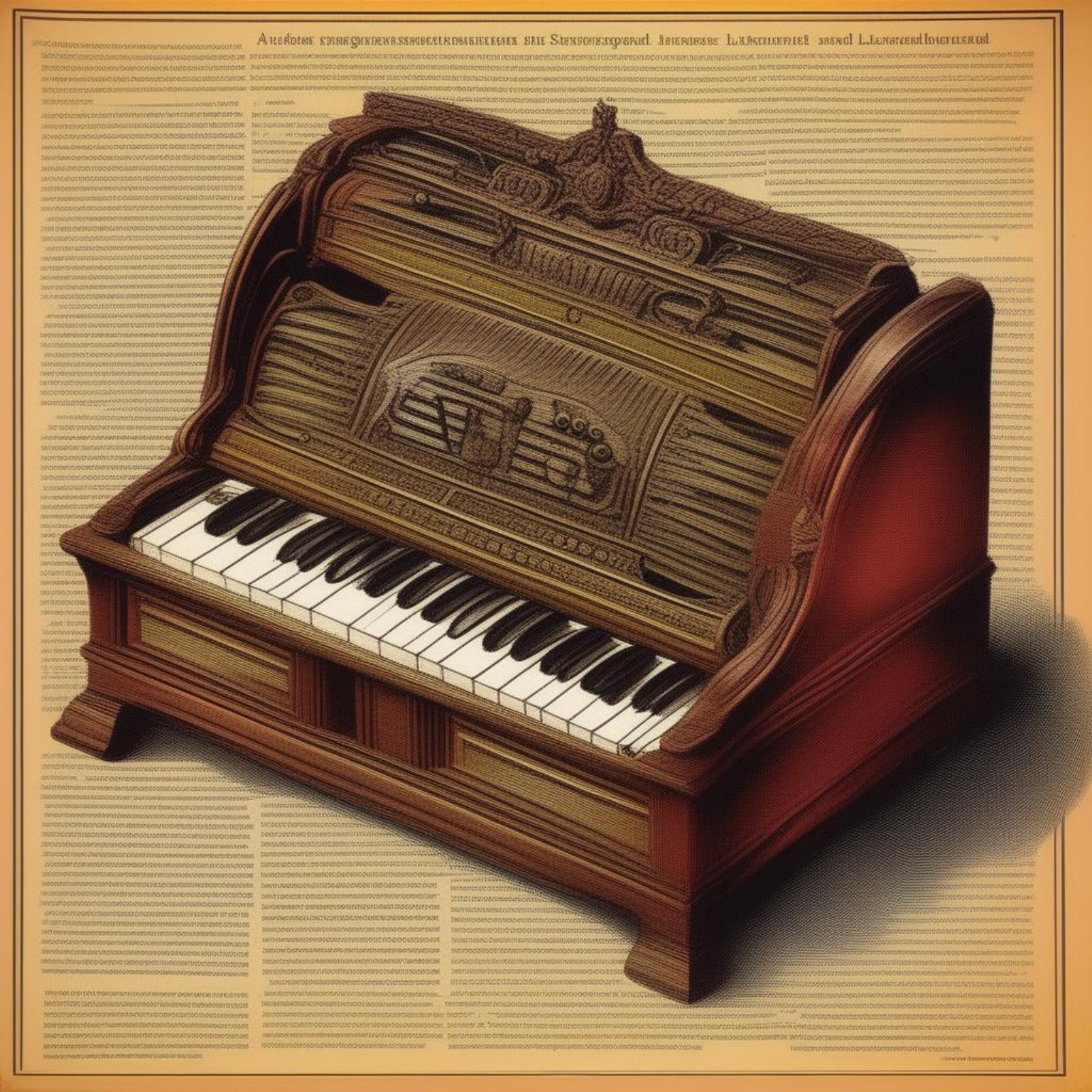} &
        \includegraphics[width=\linewidth]{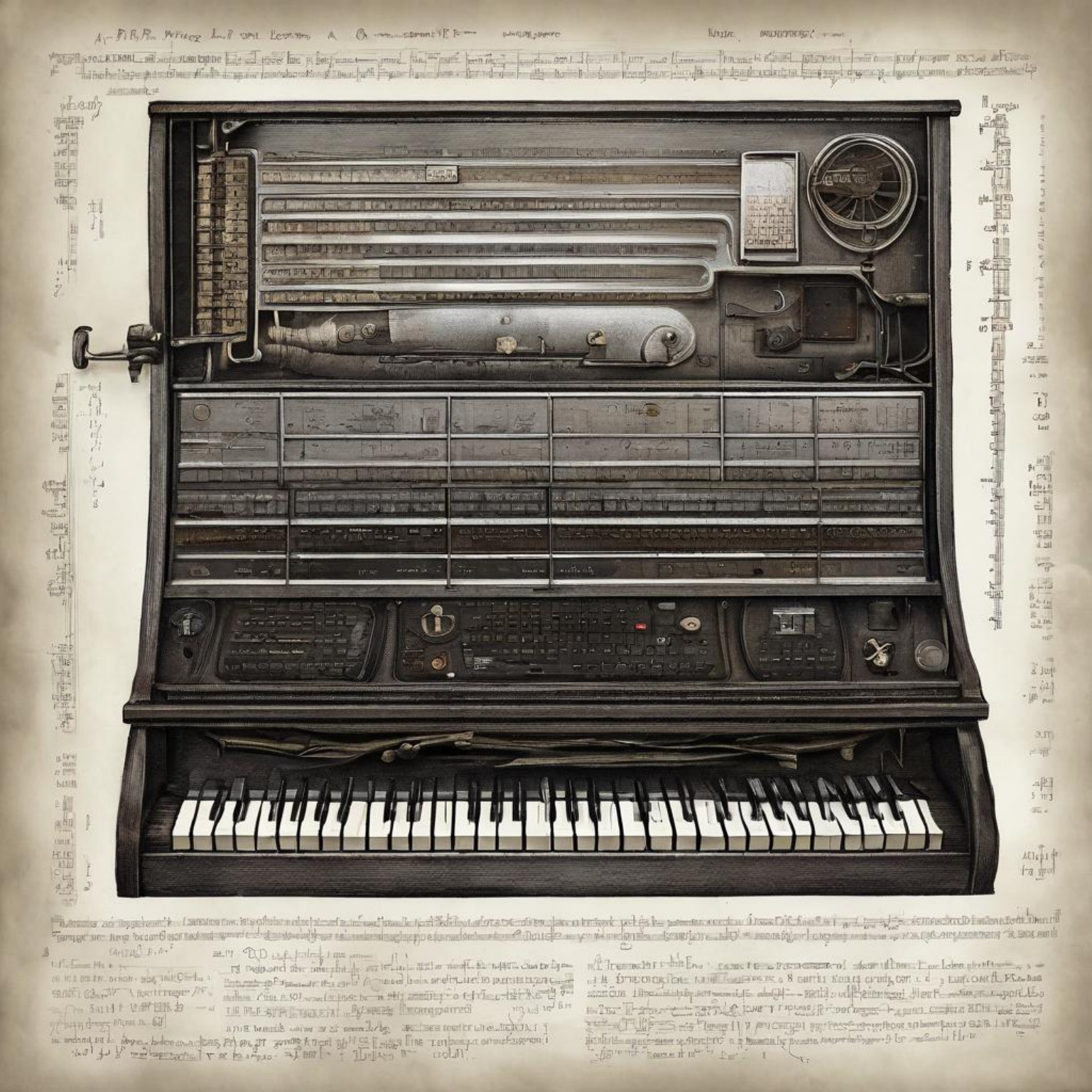} &
        \includegraphics[width=\linewidth]{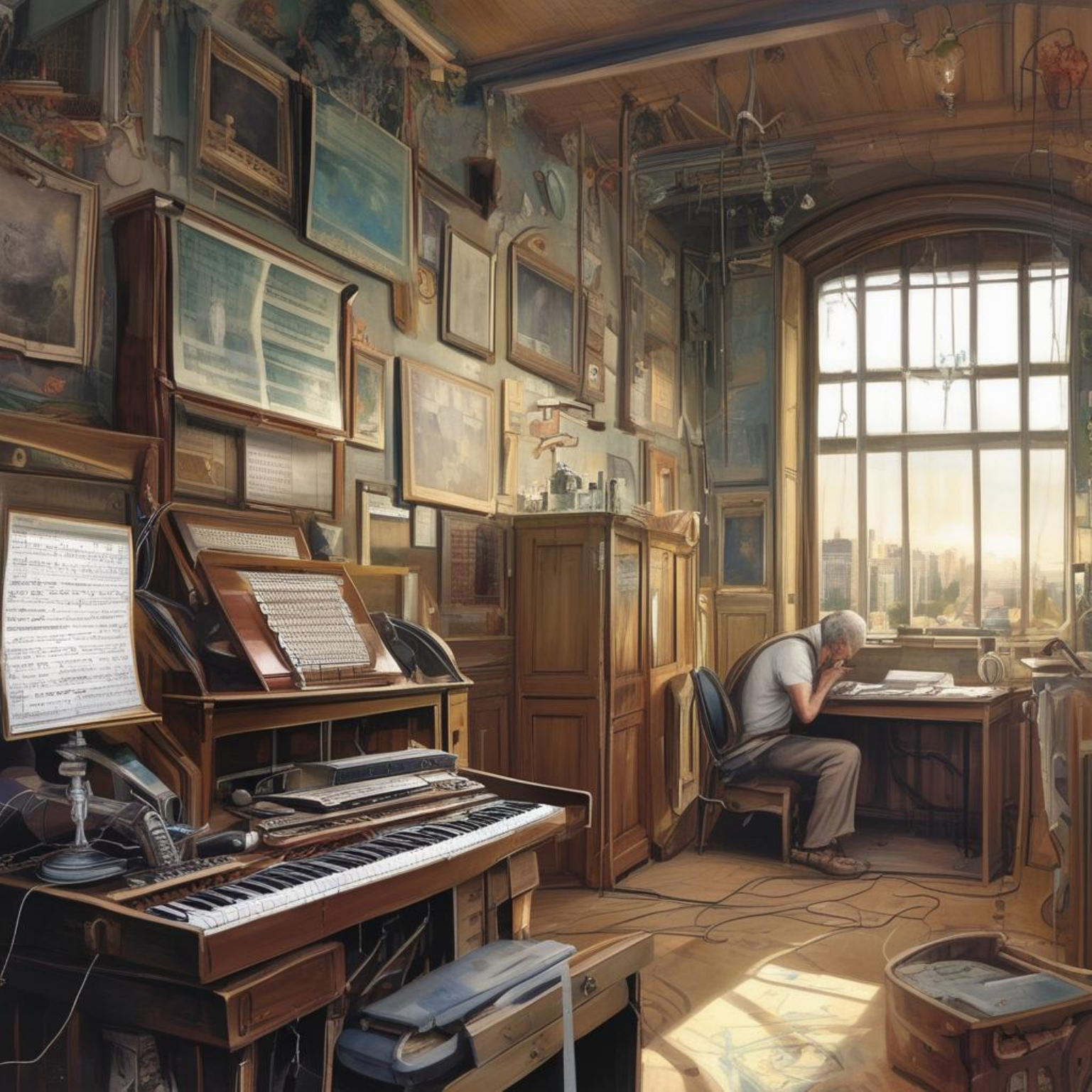} \\

        \multicolumn{8}{@{}p{\linewidth}@{}}{\centering \small \textit{Prompt: A ldarge keybord msical instroument lwith a woden case enmclosig a qsouvnkboajrd and mfgtal strivgf, which are strucrk b hammrs when the nels are depresdsmed.f lhe strsingsj' vibration ie stopped by damperds when the keys re released and can bce regulavewdd for lengh and vnolume y two or three pedalvs.}} \\
		\midrule

        \includegraphics[width=\linewidth]{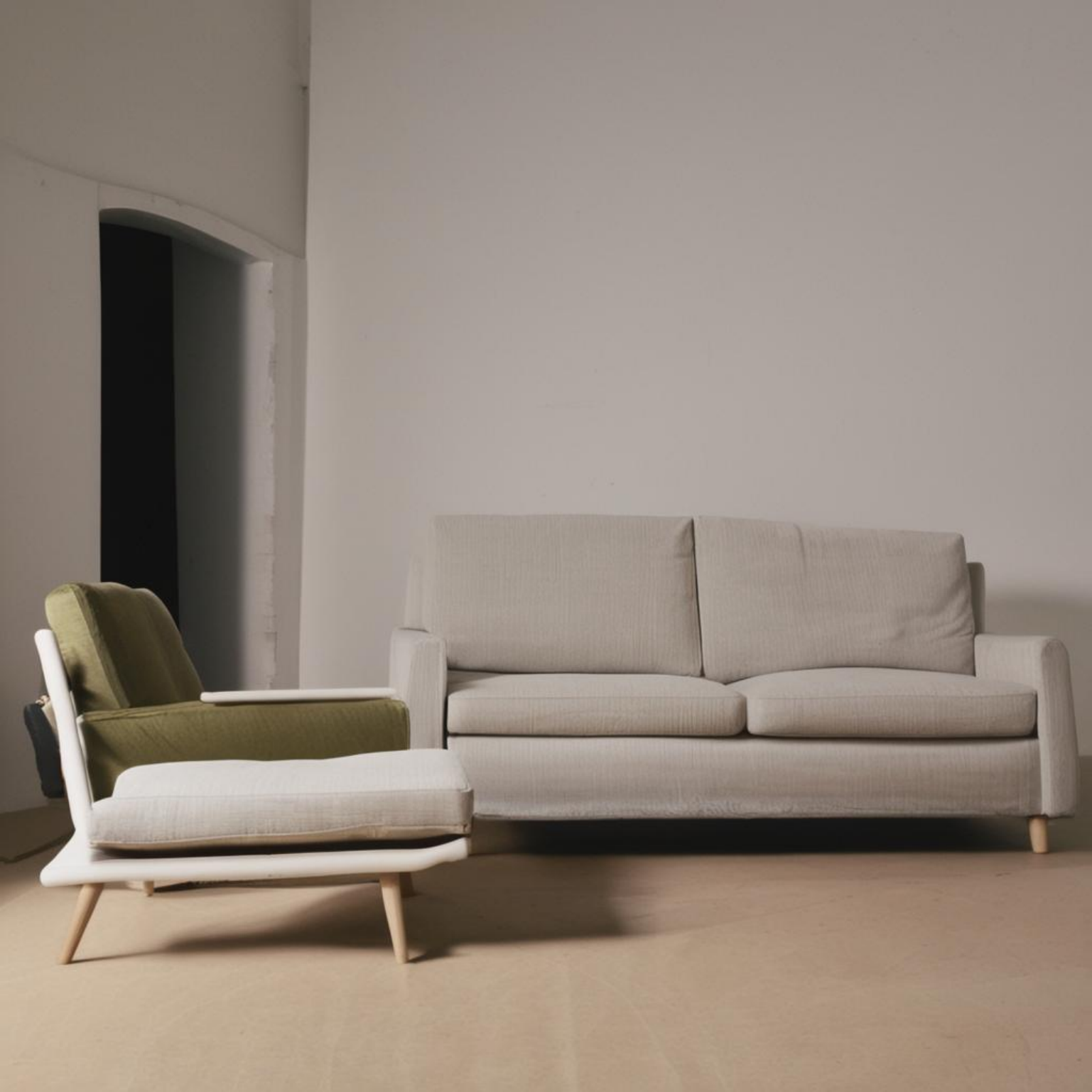} &
        \includegraphics[width=\linewidth]{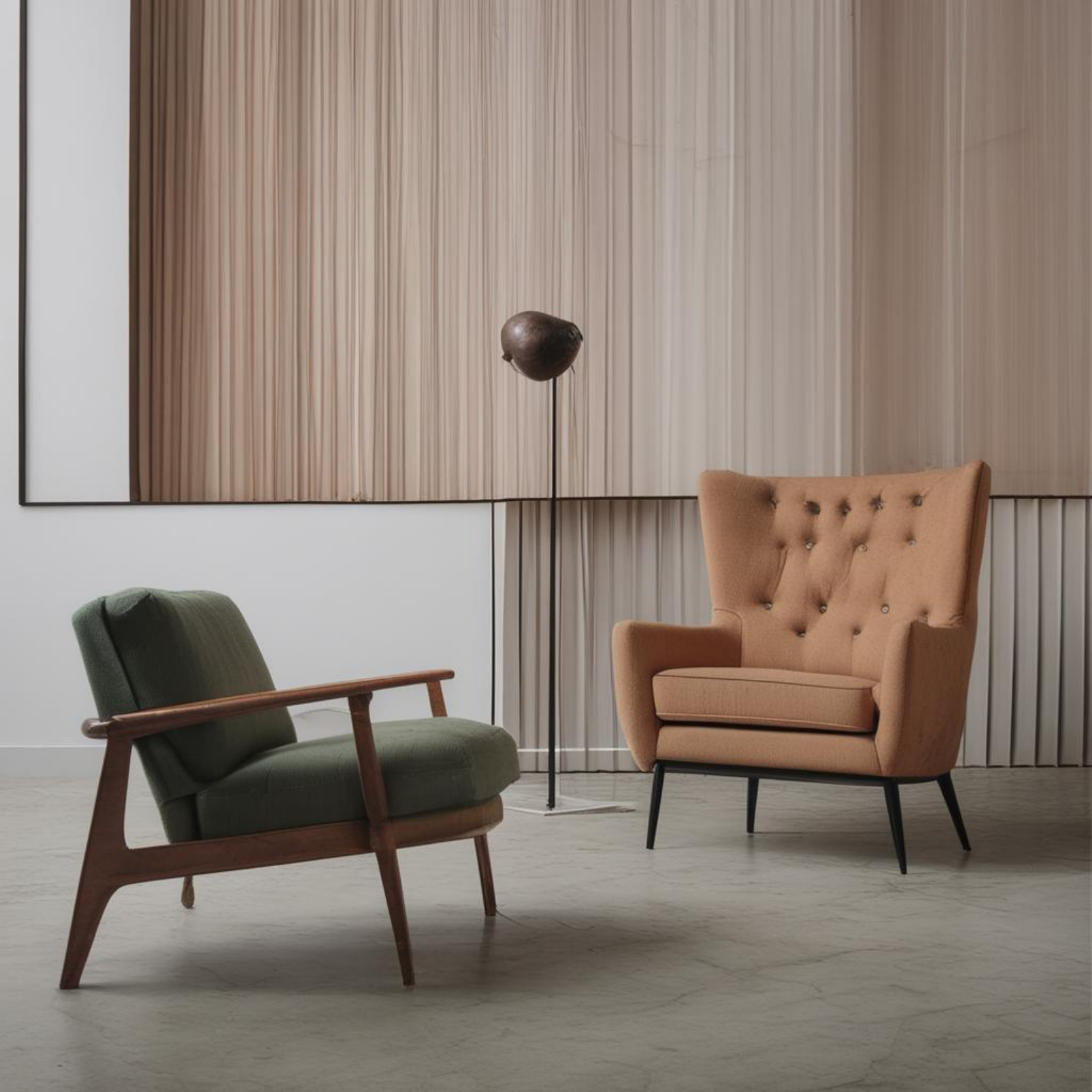} &
        \includegraphics[width=\linewidth]{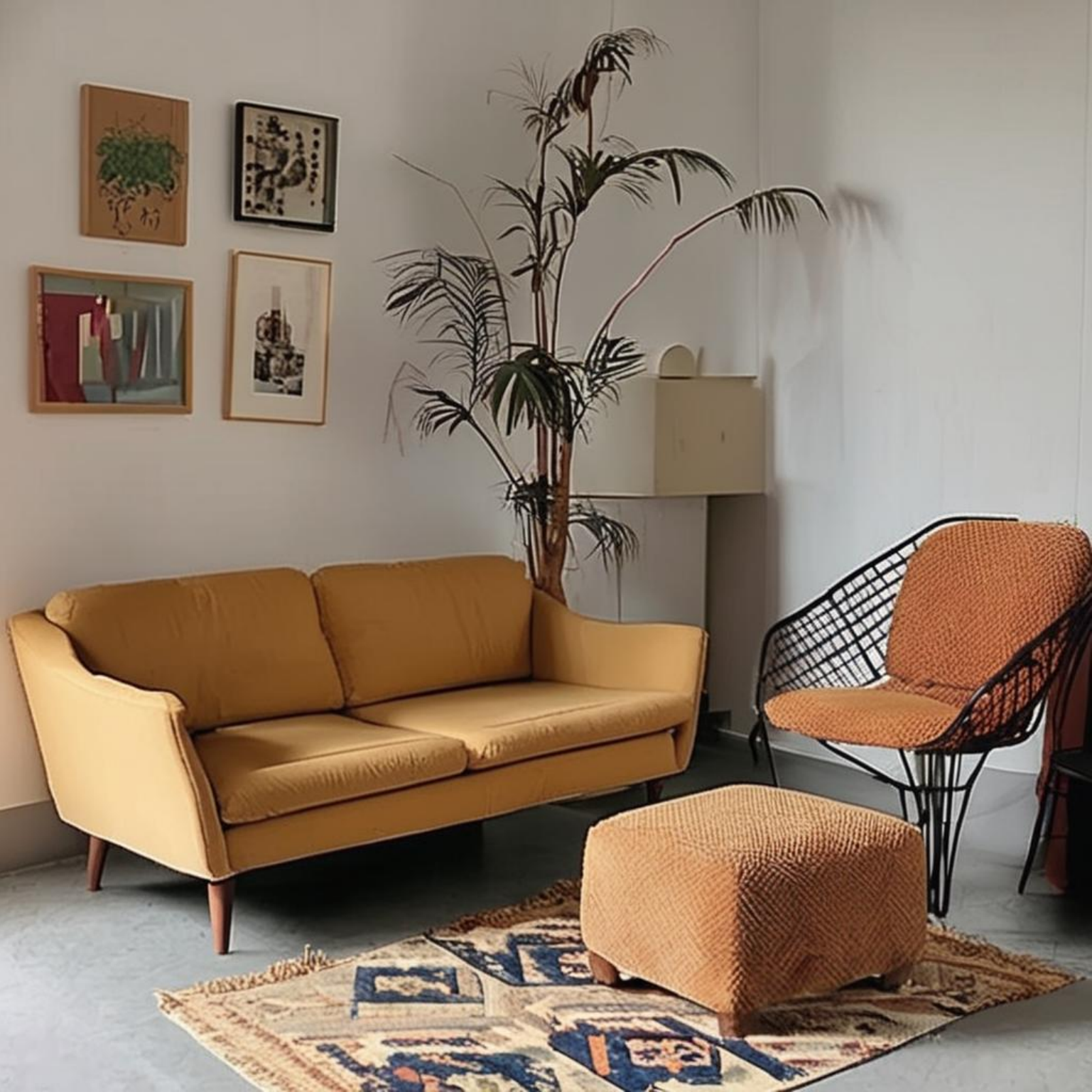} &
        \includegraphics[width=\linewidth]{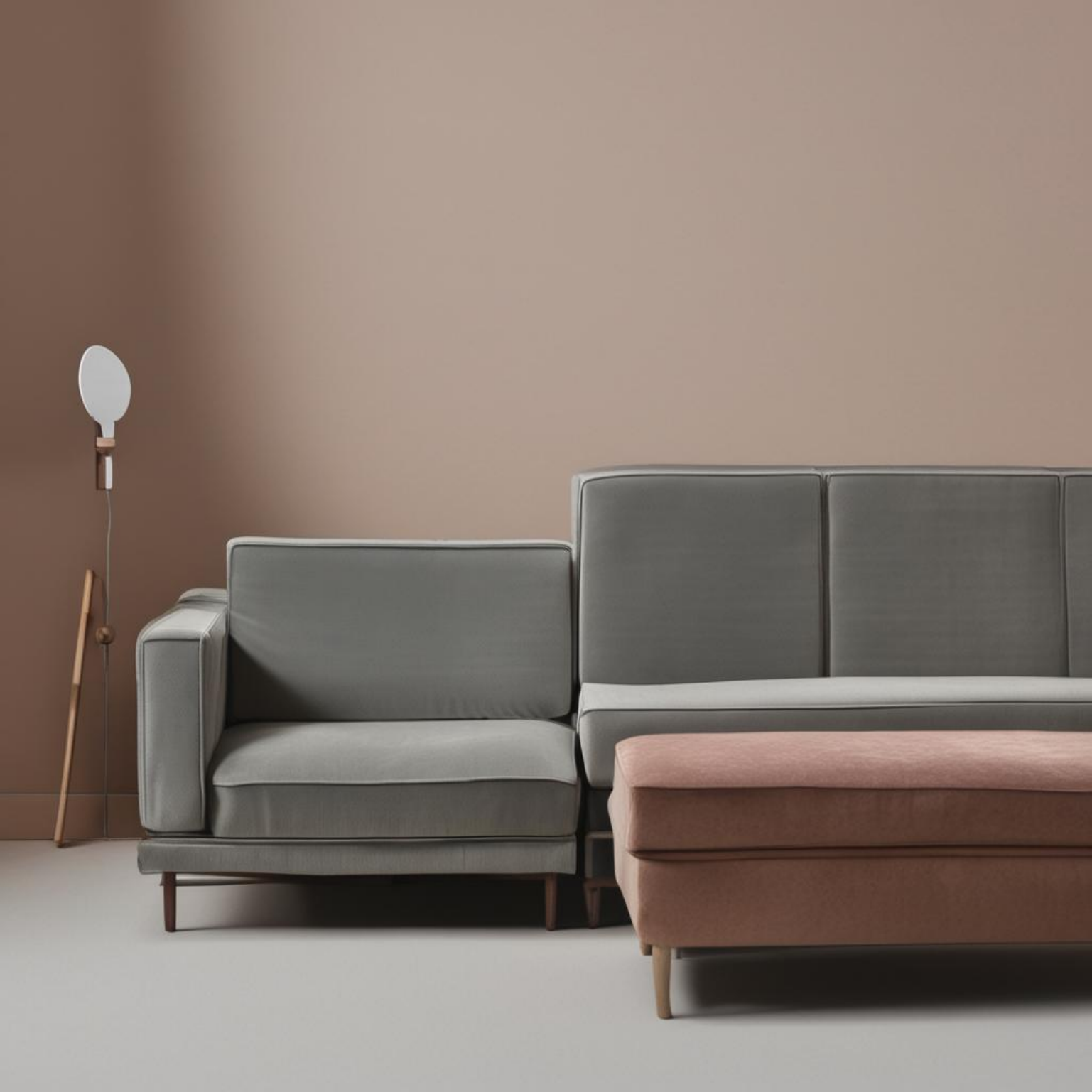} &
        \includegraphics[width=\linewidth]{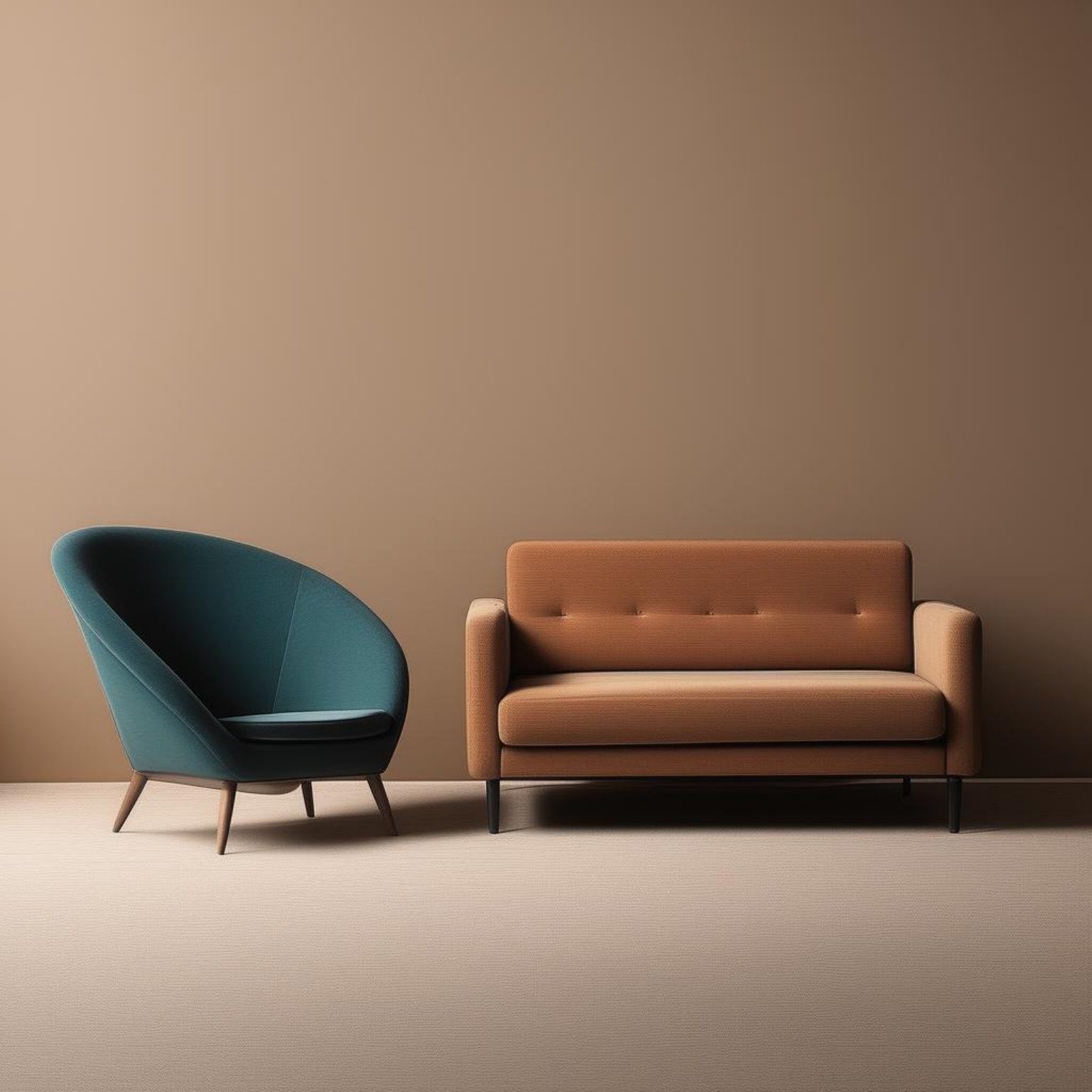} &
        \includegraphics[width=\linewidth]{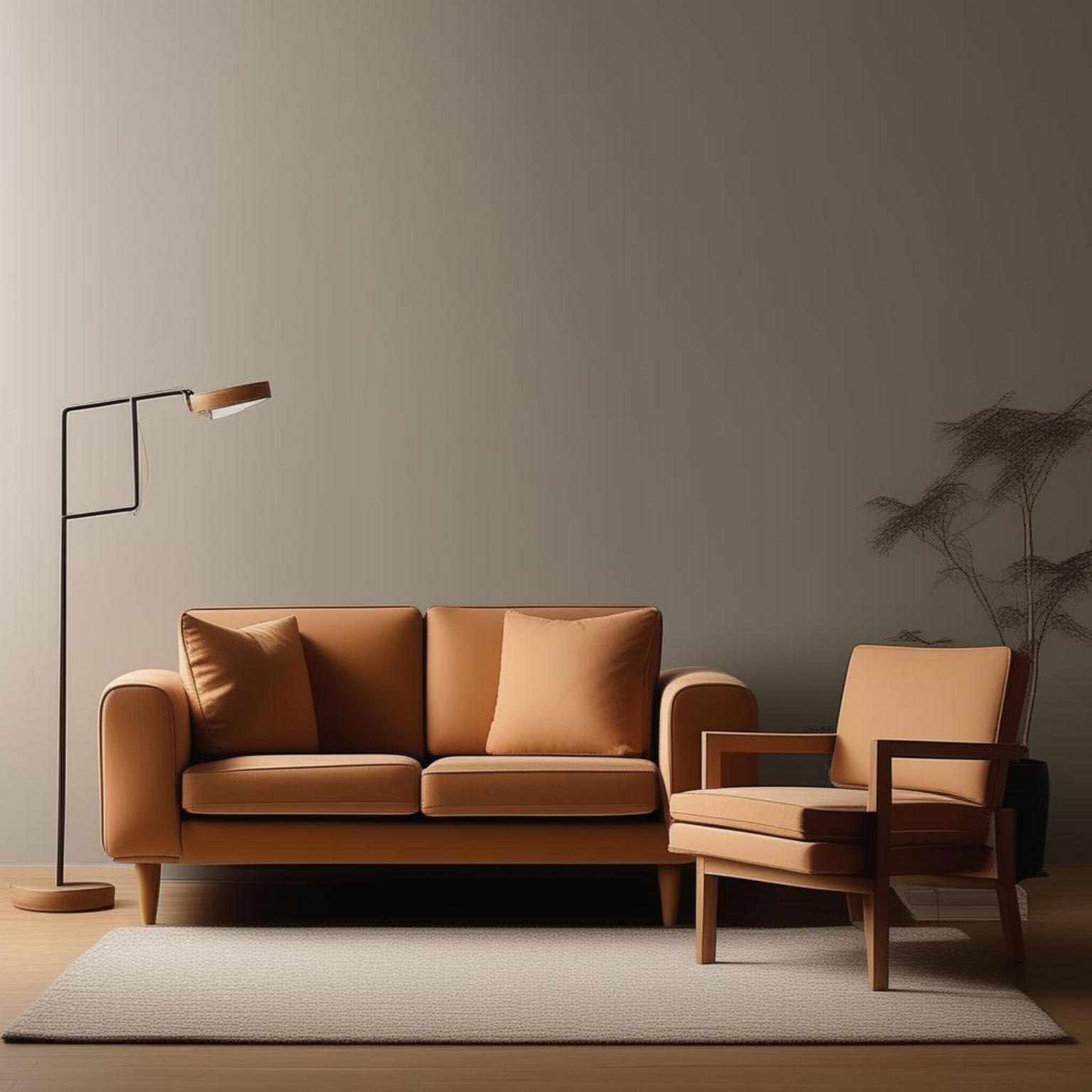} &
        \includegraphics[width=\linewidth]{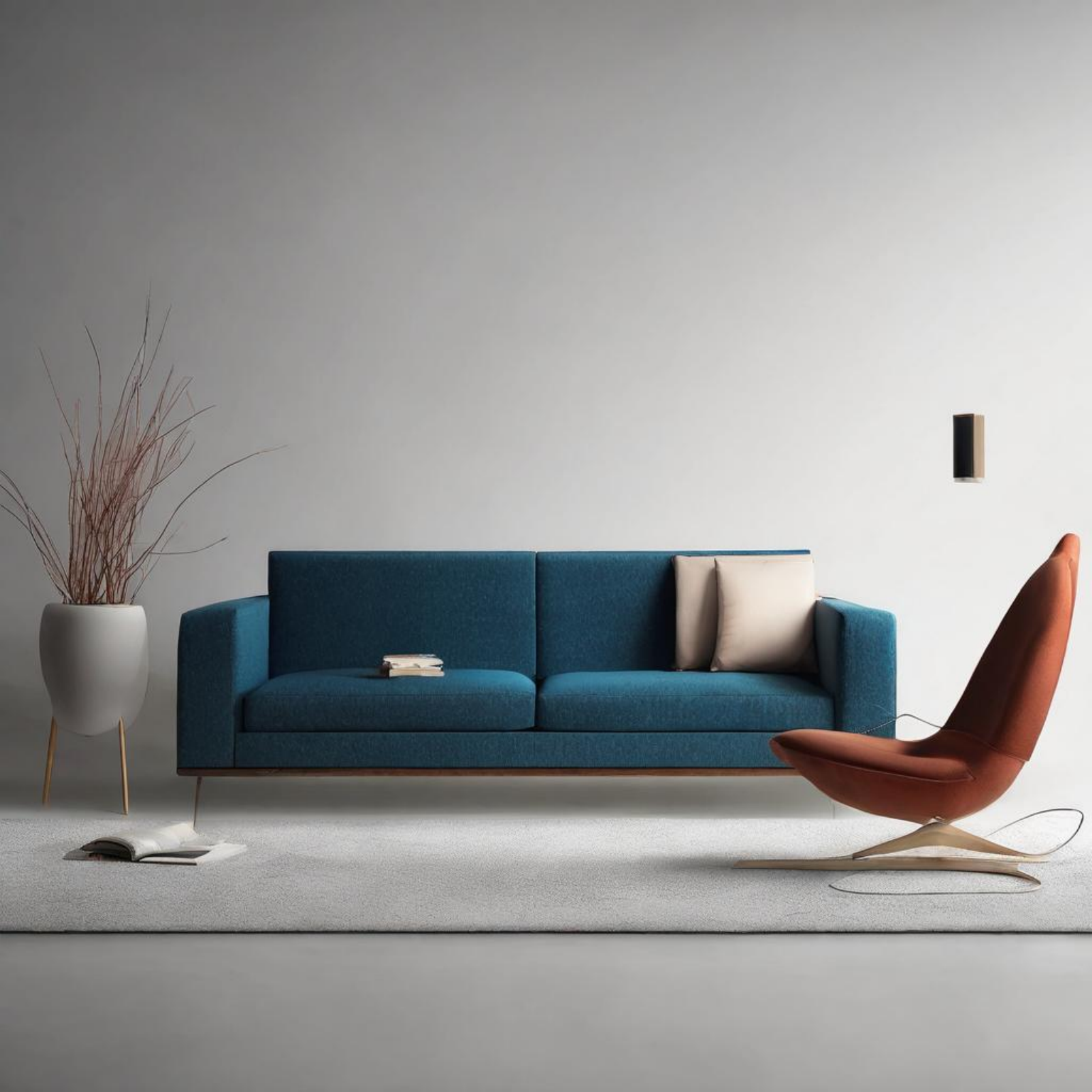} &
        \includegraphics[width=\linewidth]{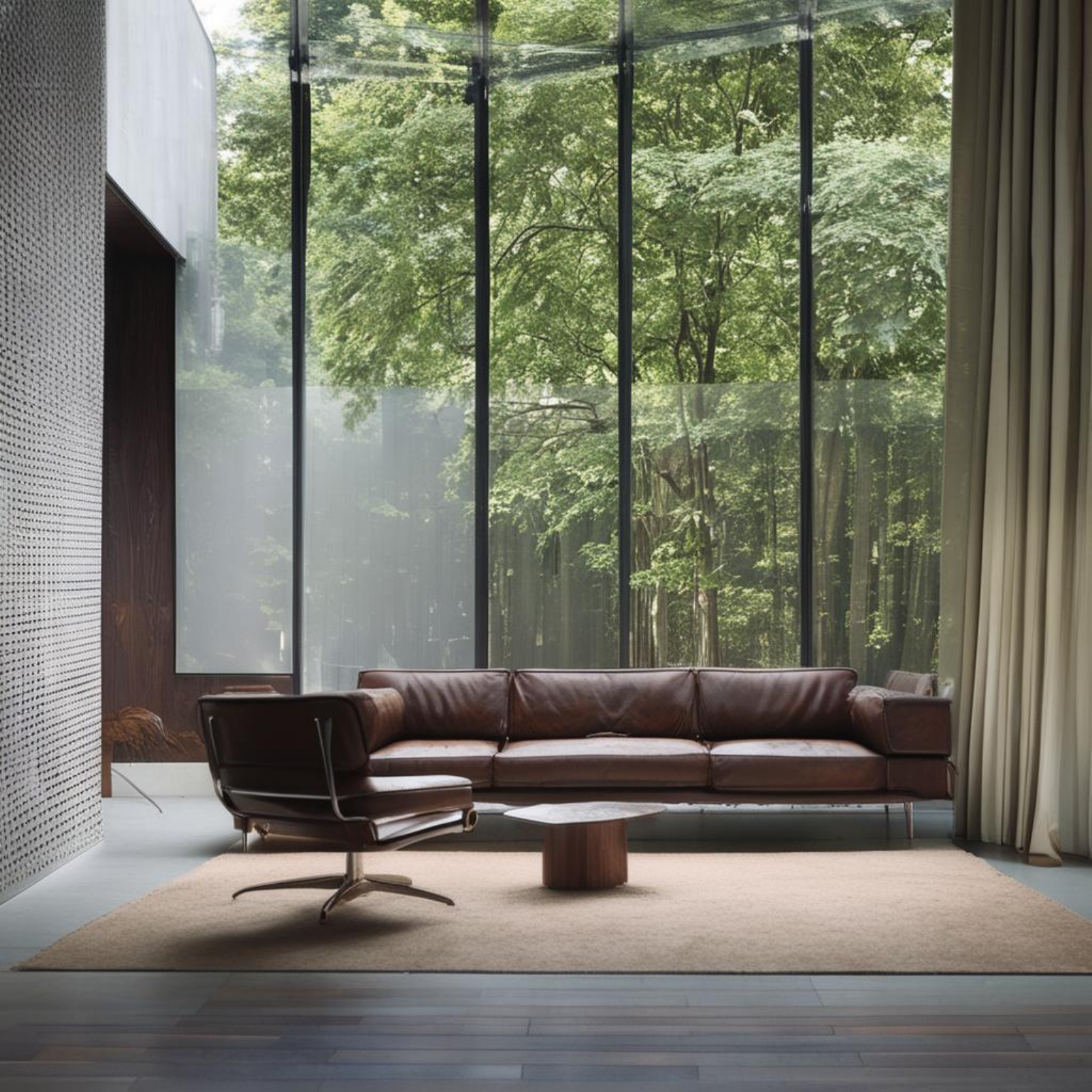} \\

        \multicolumn{8}{@{}p{\linewidth}@{}}{\centering \small \textit{Prompt: A couch on the right of a chair.}} \\
		\midrule

        \includegraphics[width=\linewidth]{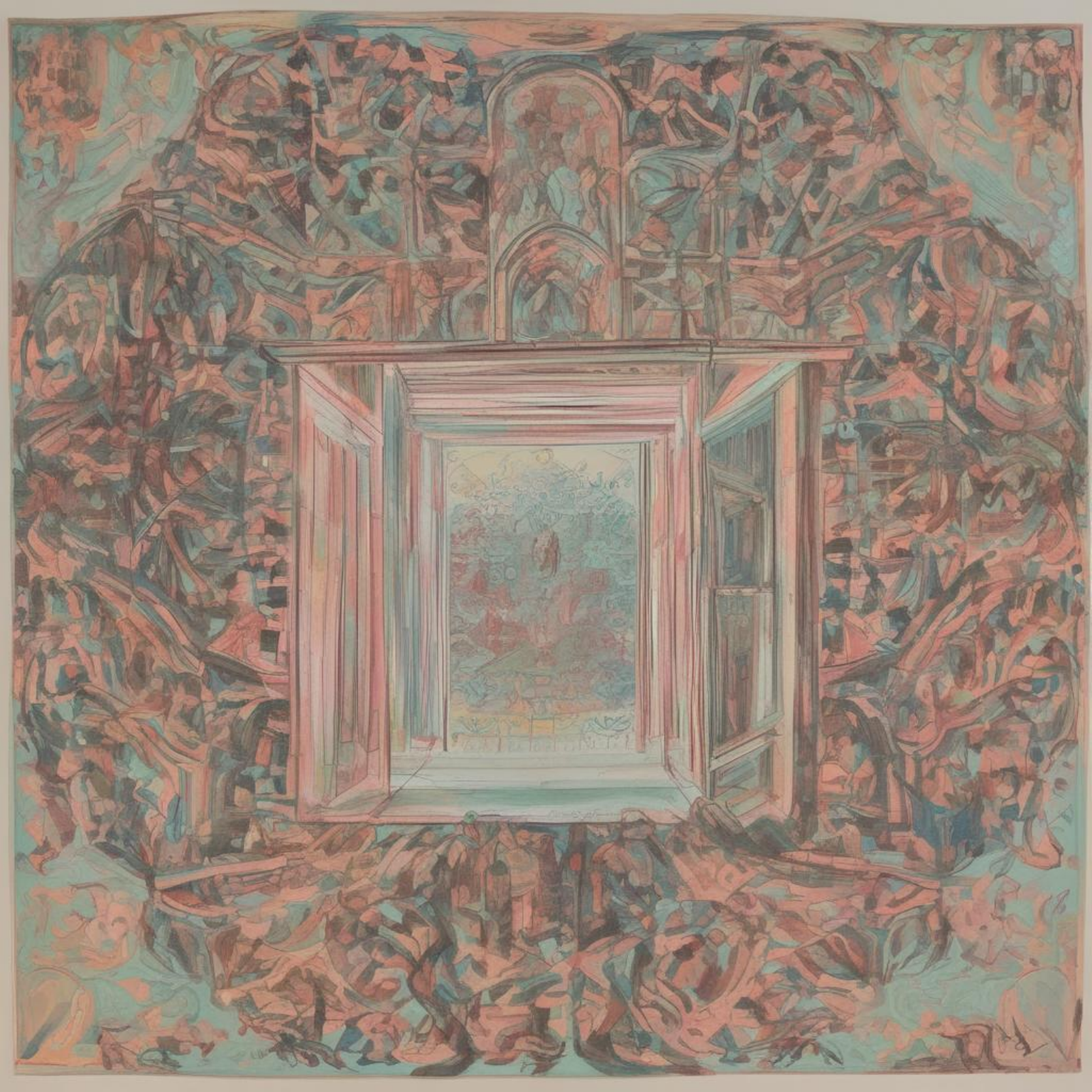} &
        \includegraphics[width=\linewidth]{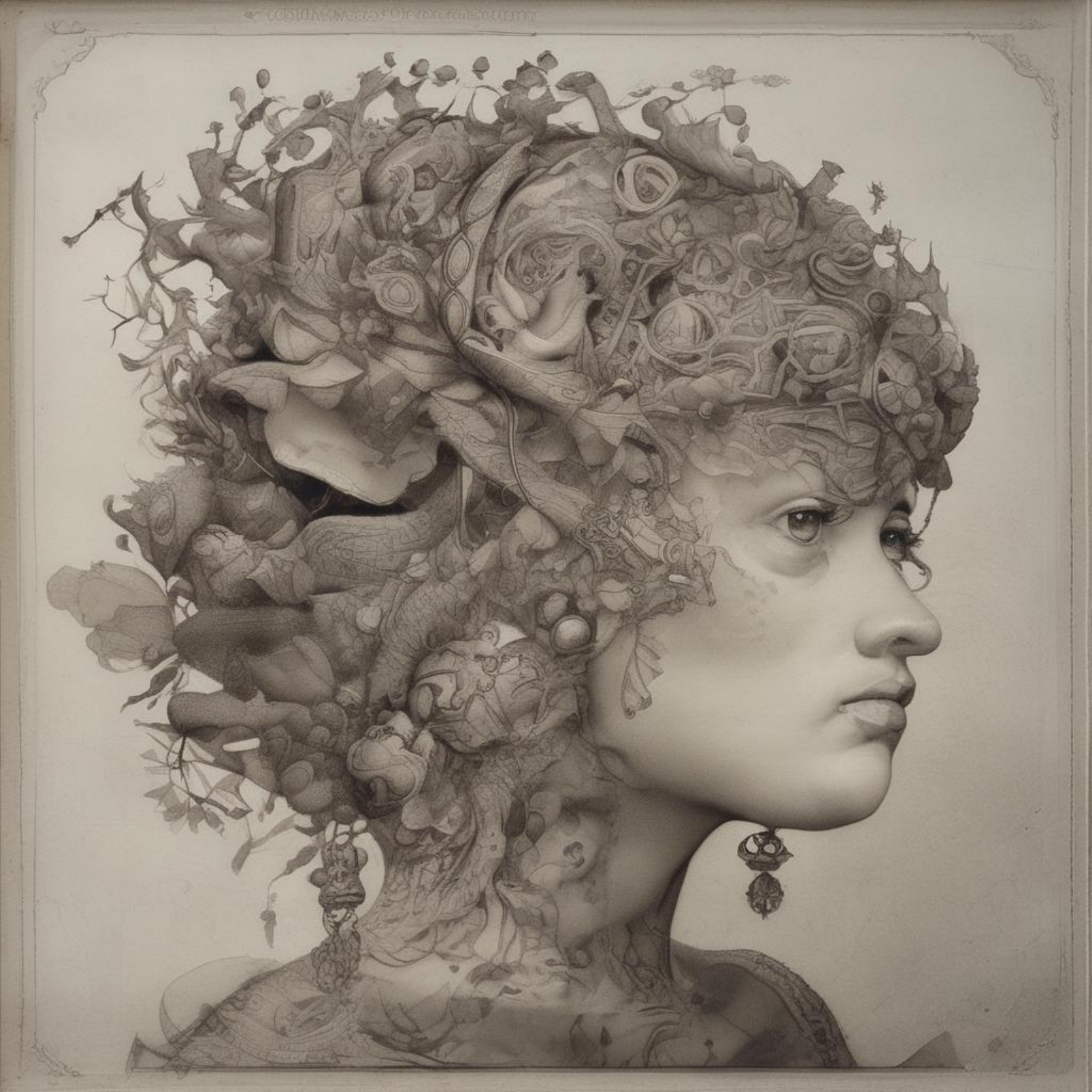} &
        \includegraphics[width=\linewidth]{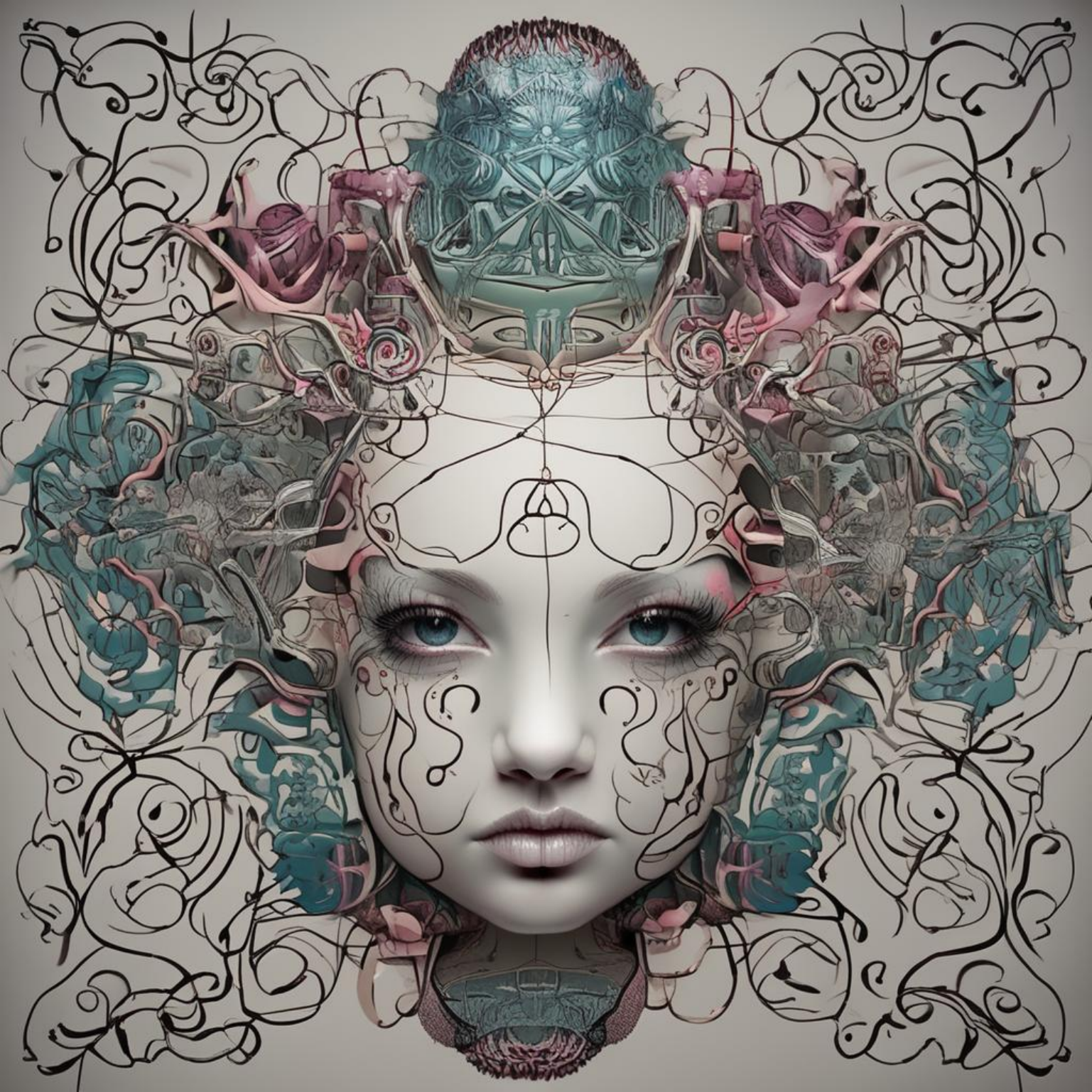} &
        \includegraphics[width=\linewidth]{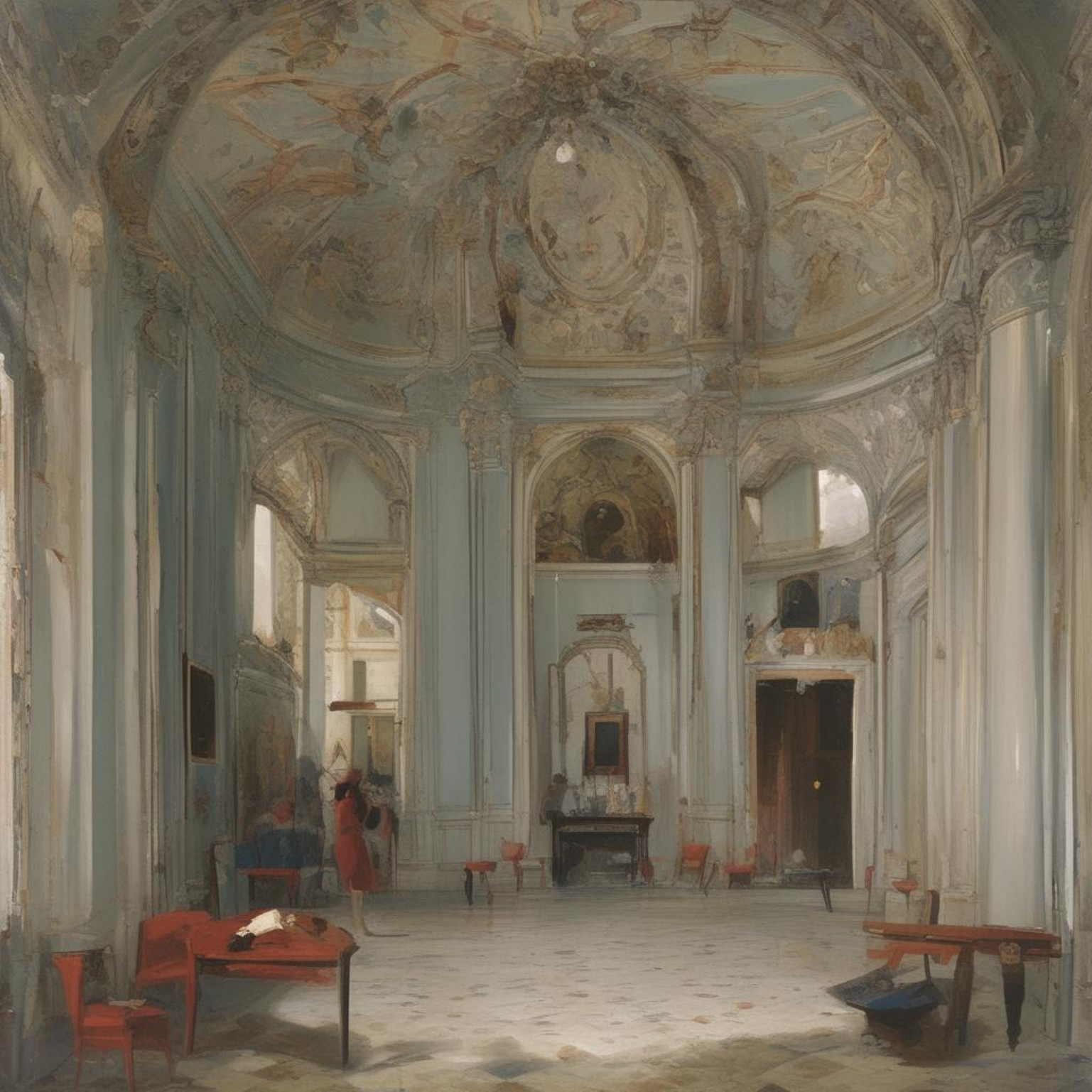} &
        \includegraphics[width=\linewidth]{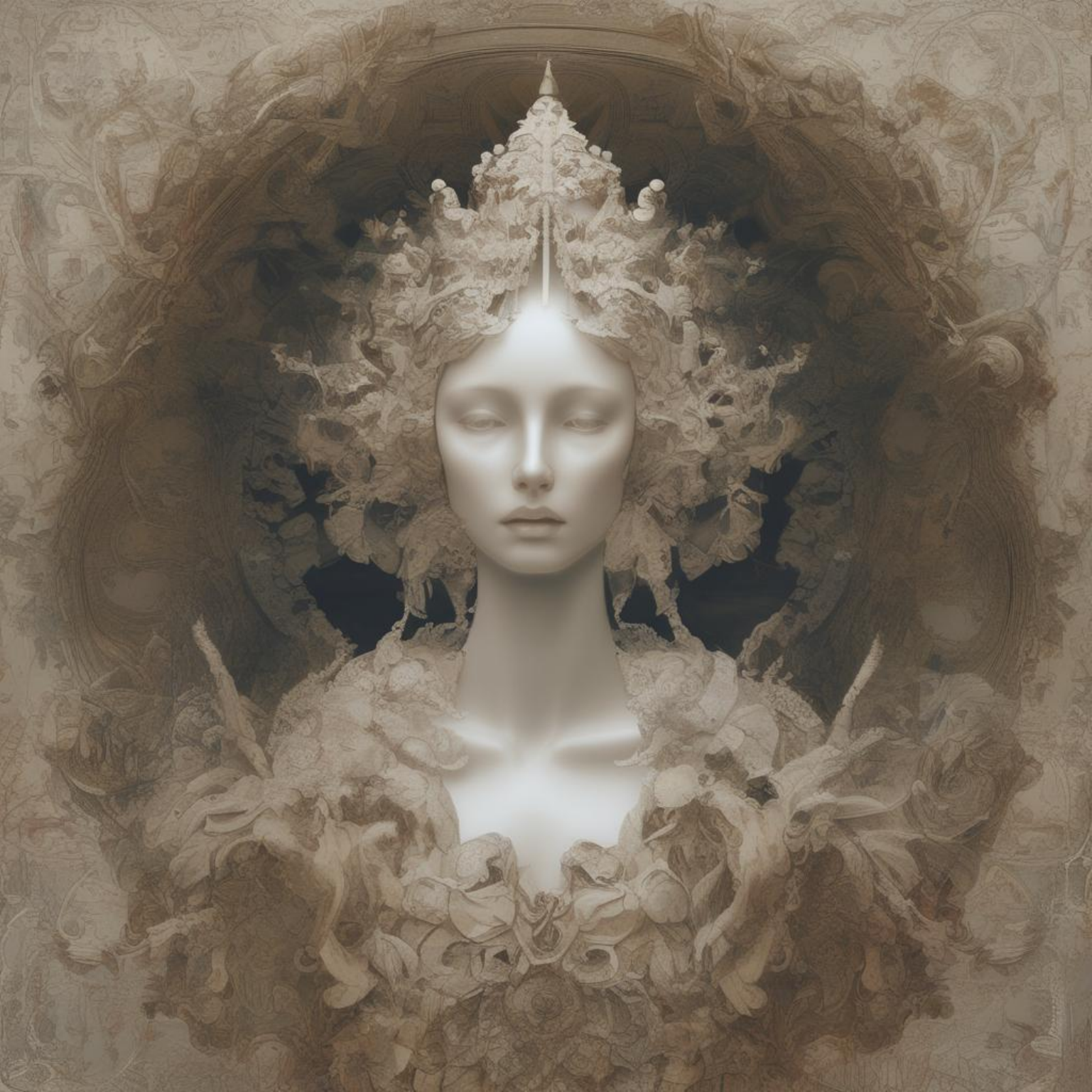} &
        \includegraphics[width=\linewidth]{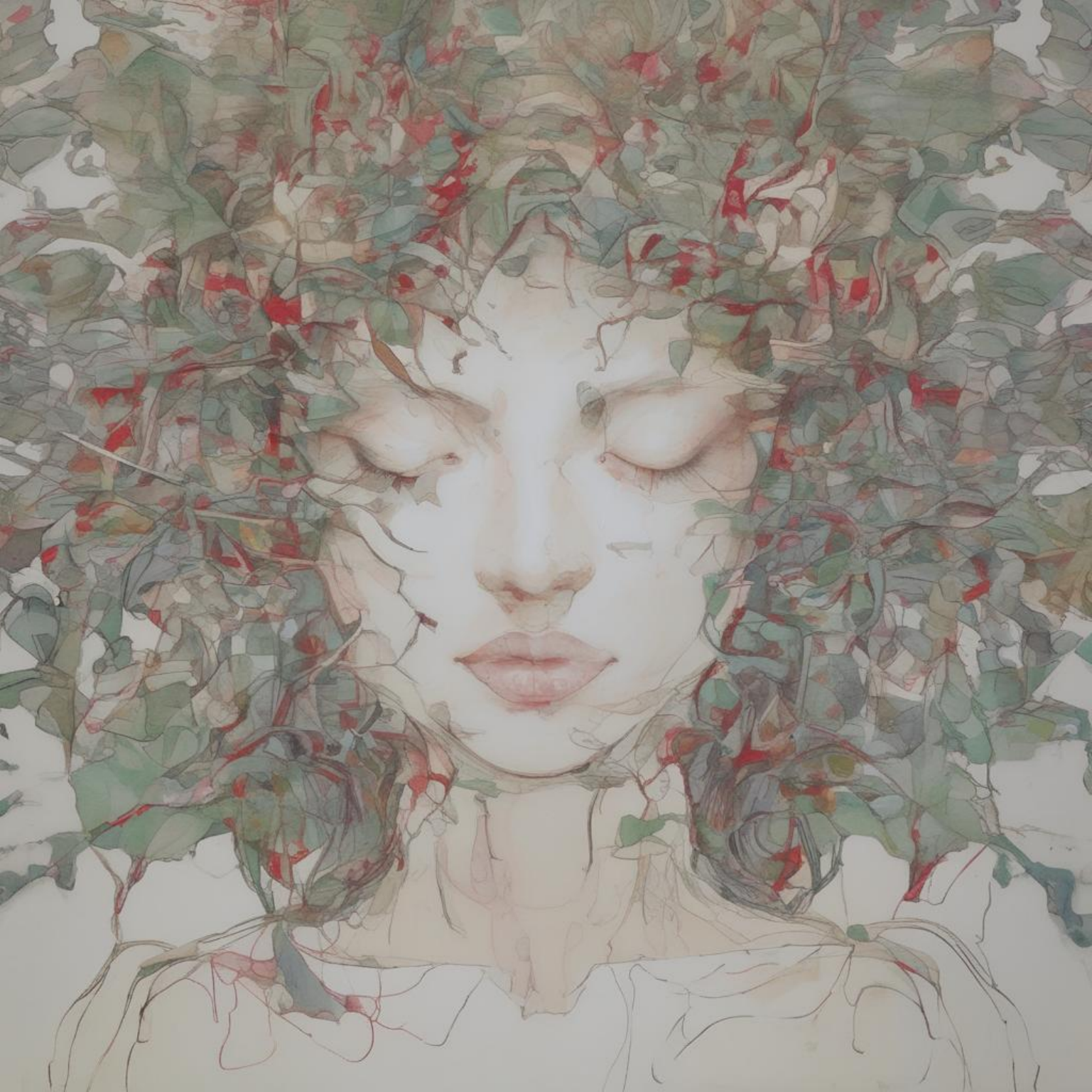} &
        \includegraphics[width=\linewidth]{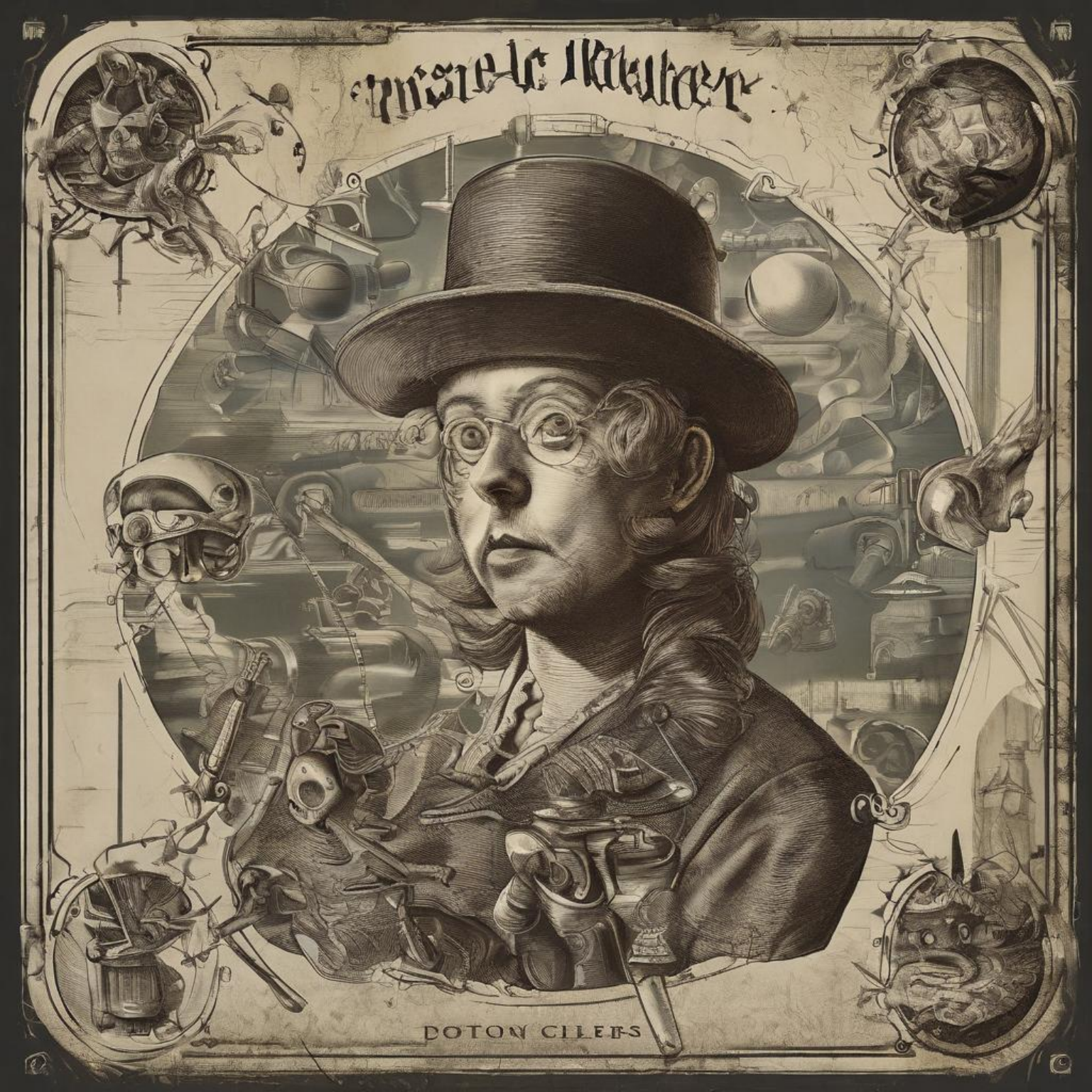} &
        \includegraphics[width=\linewidth]{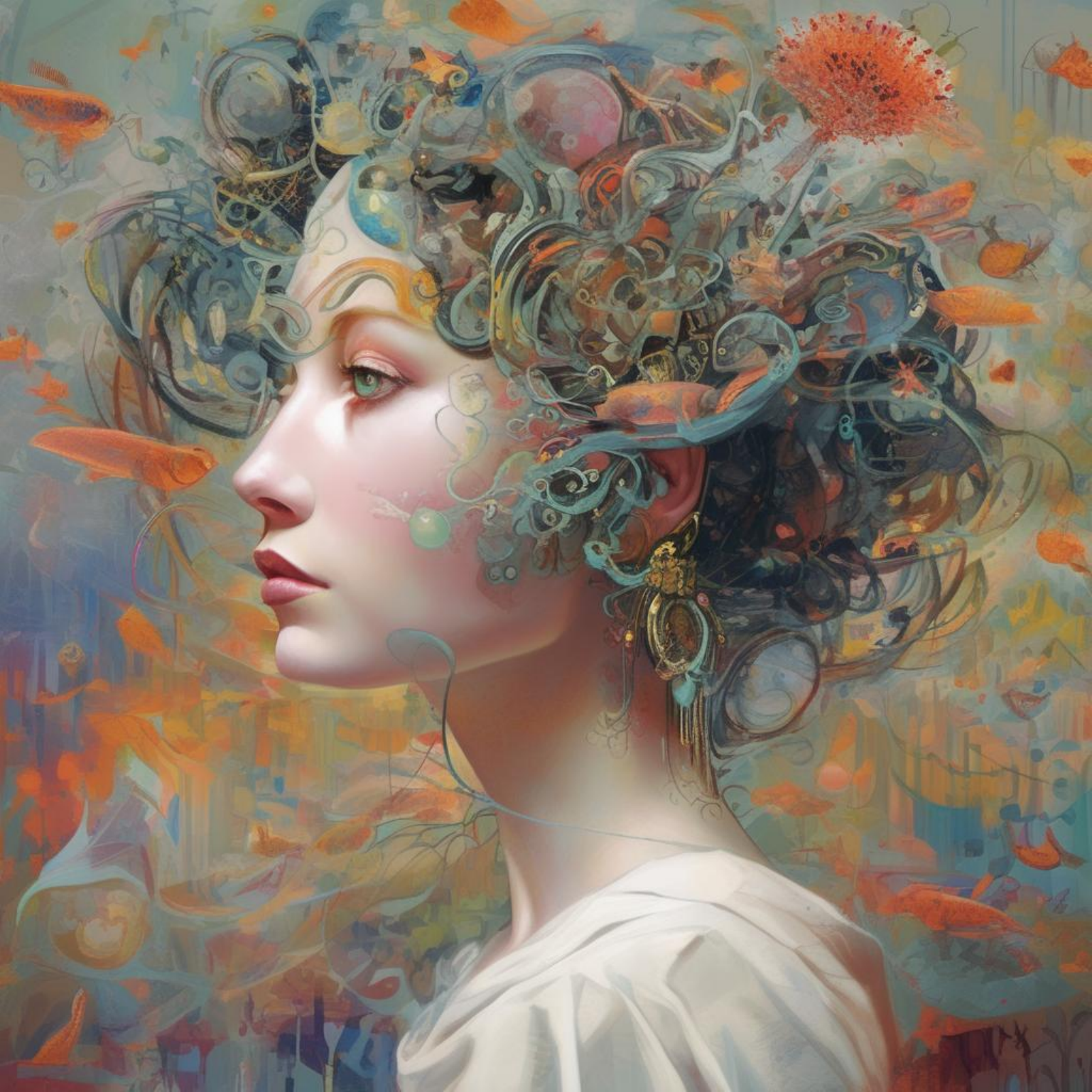} \\

        \multicolumn{8}{@{}p{\linewidth}@{}}{\centering \small \textit{Prompt: Artophagous.}} \\
		\midrule

        \includegraphics[width=\linewidth]{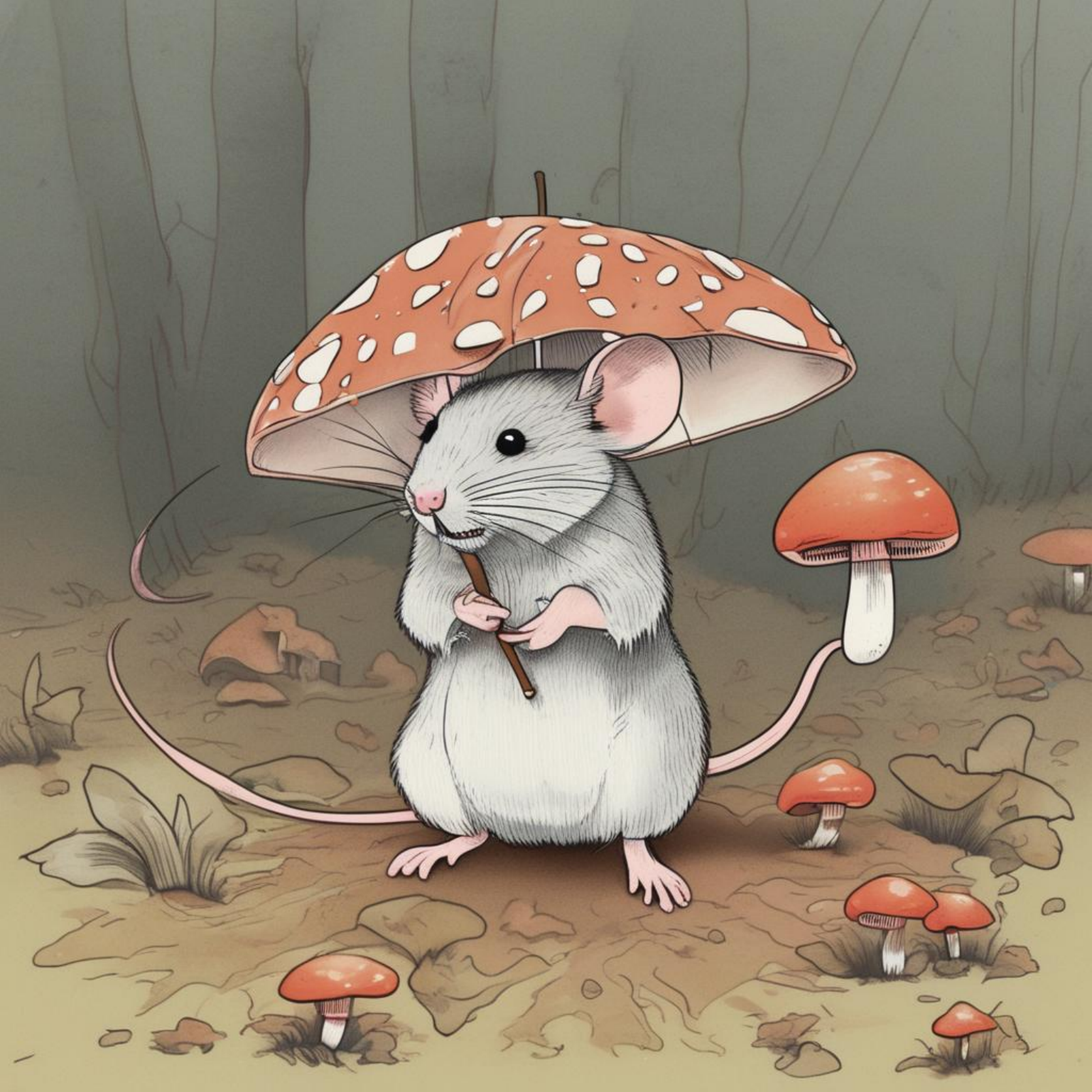} &
        \includegraphics[width=\linewidth]{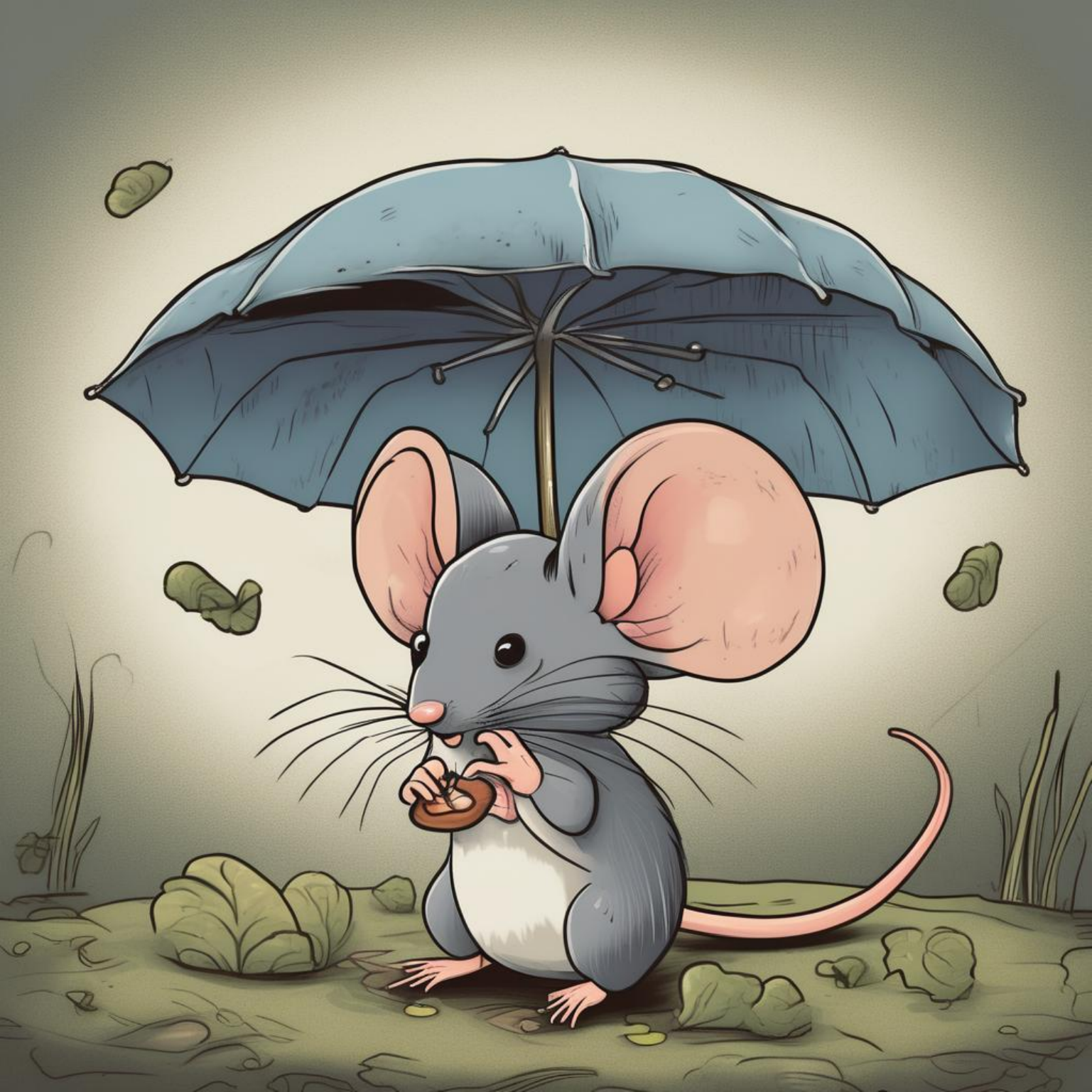} &
        \includegraphics[width=\linewidth]{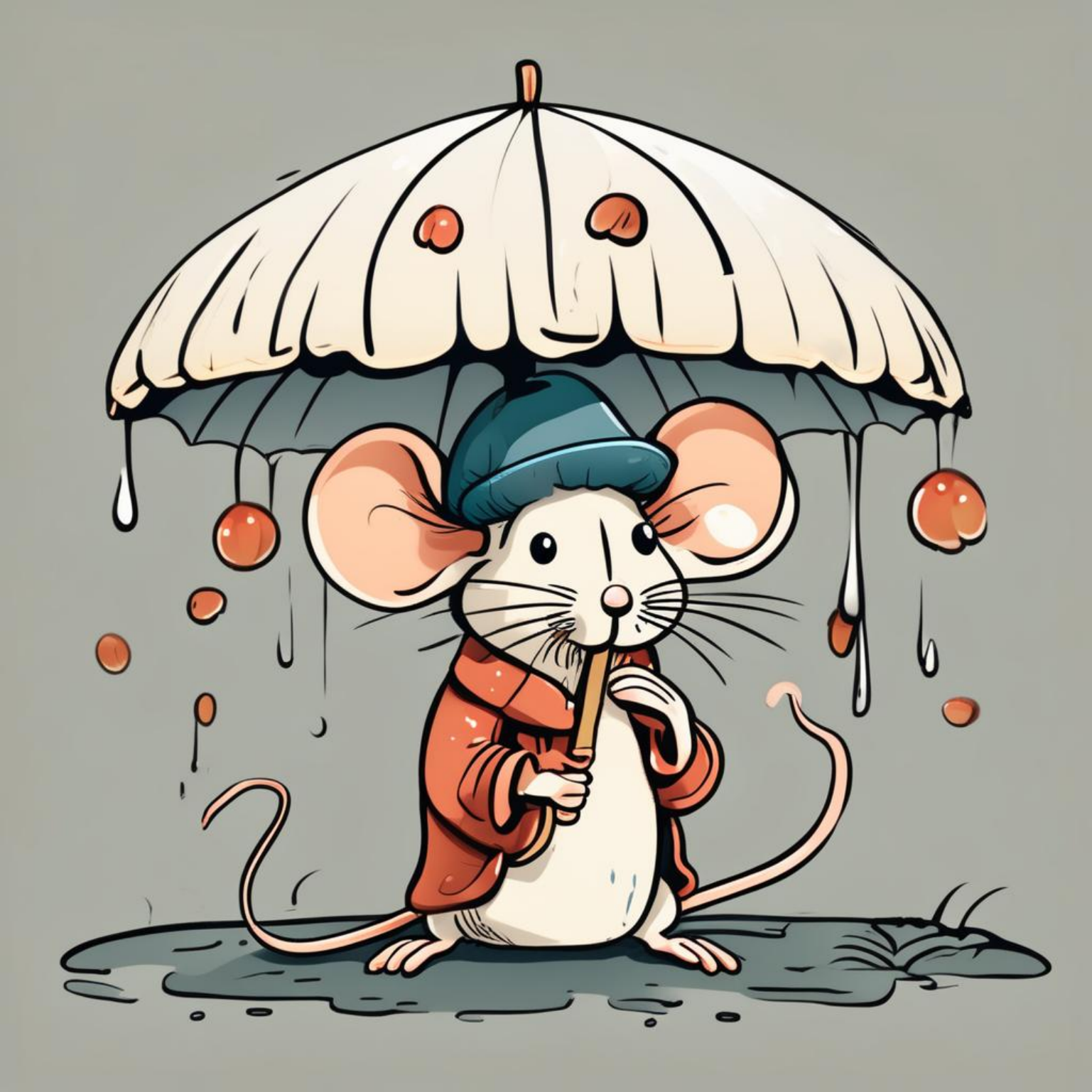} &
        \includegraphics[width=\linewidth]{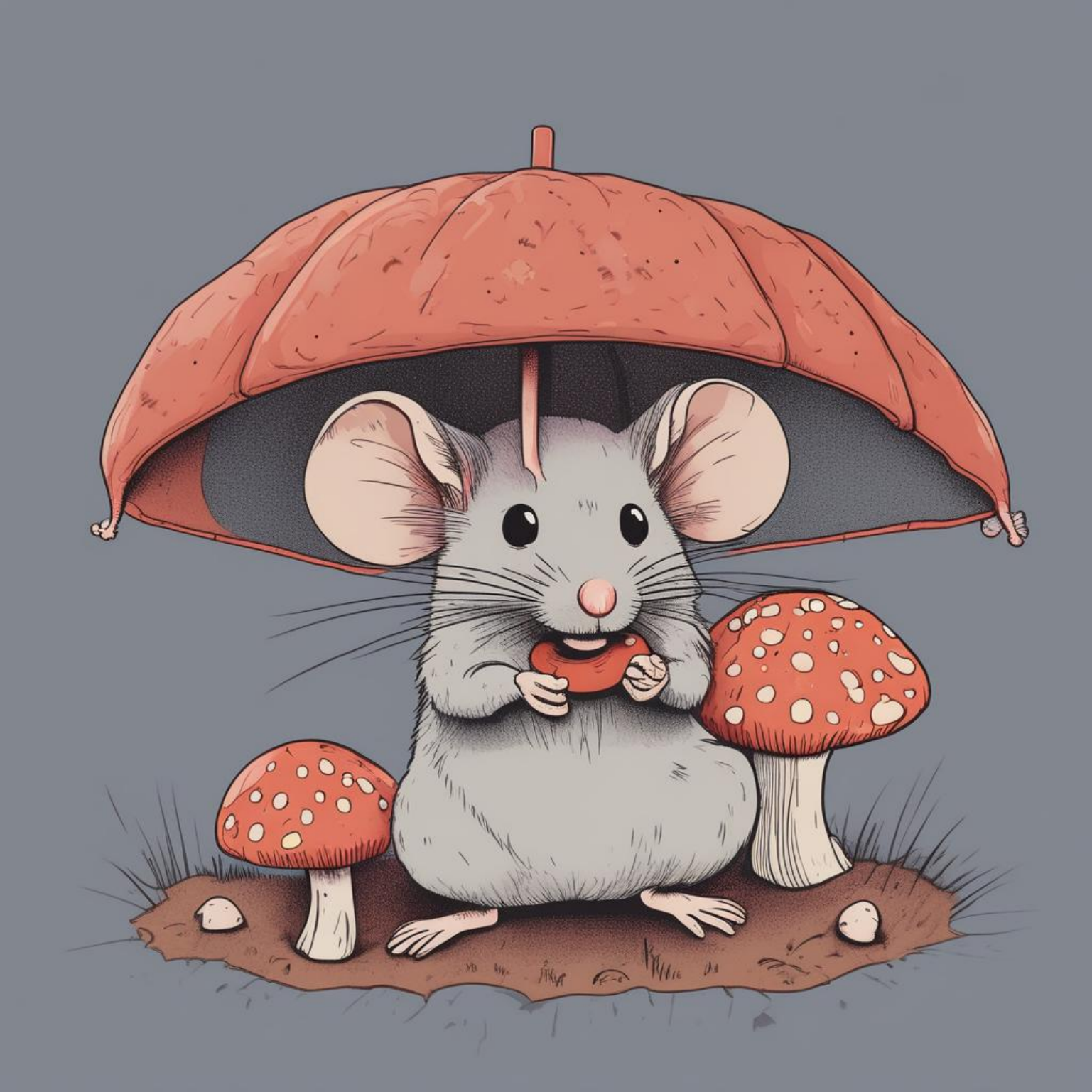} &
        \includegraphics[width=\linewidth]{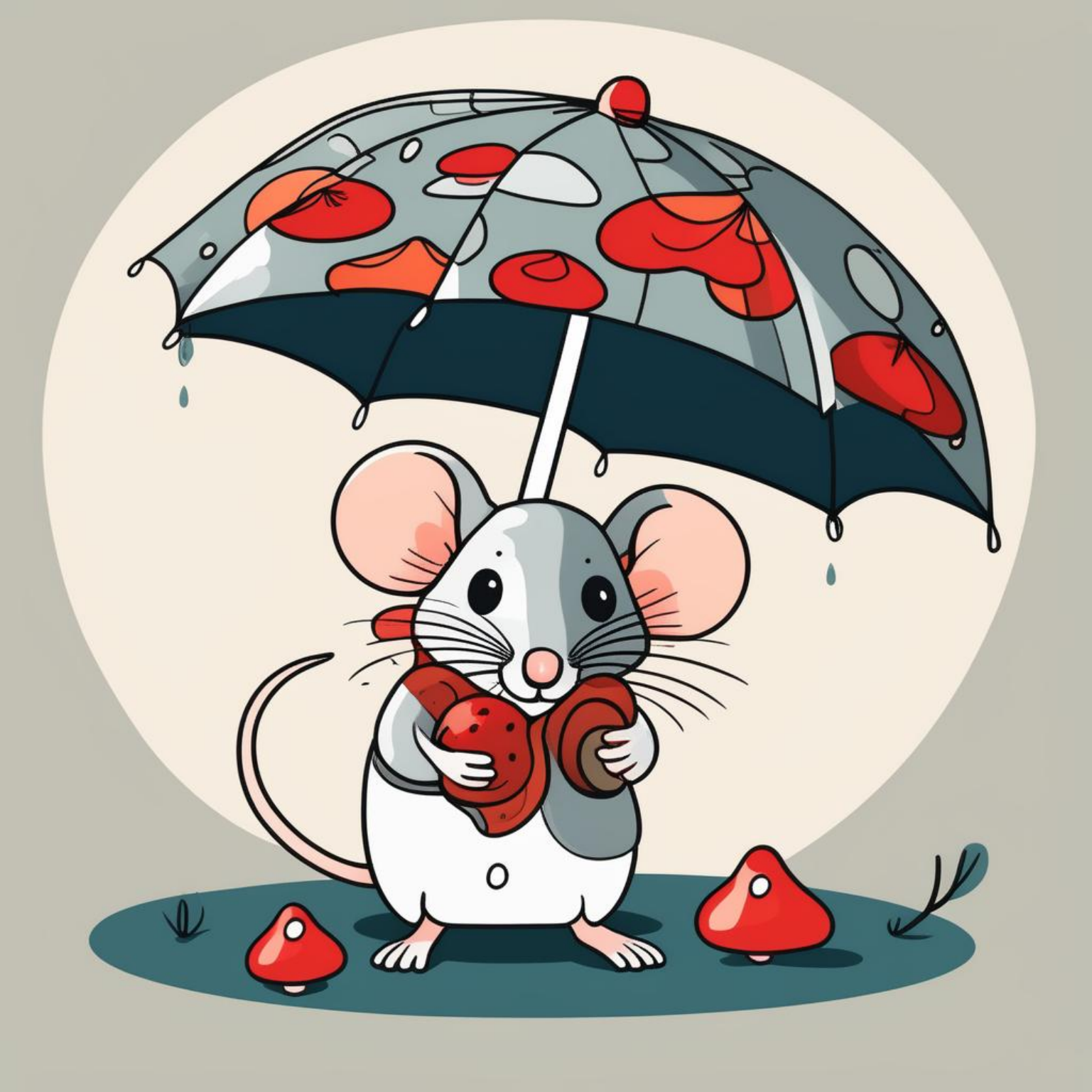} &
        \includegraphics[width=\linewidth]{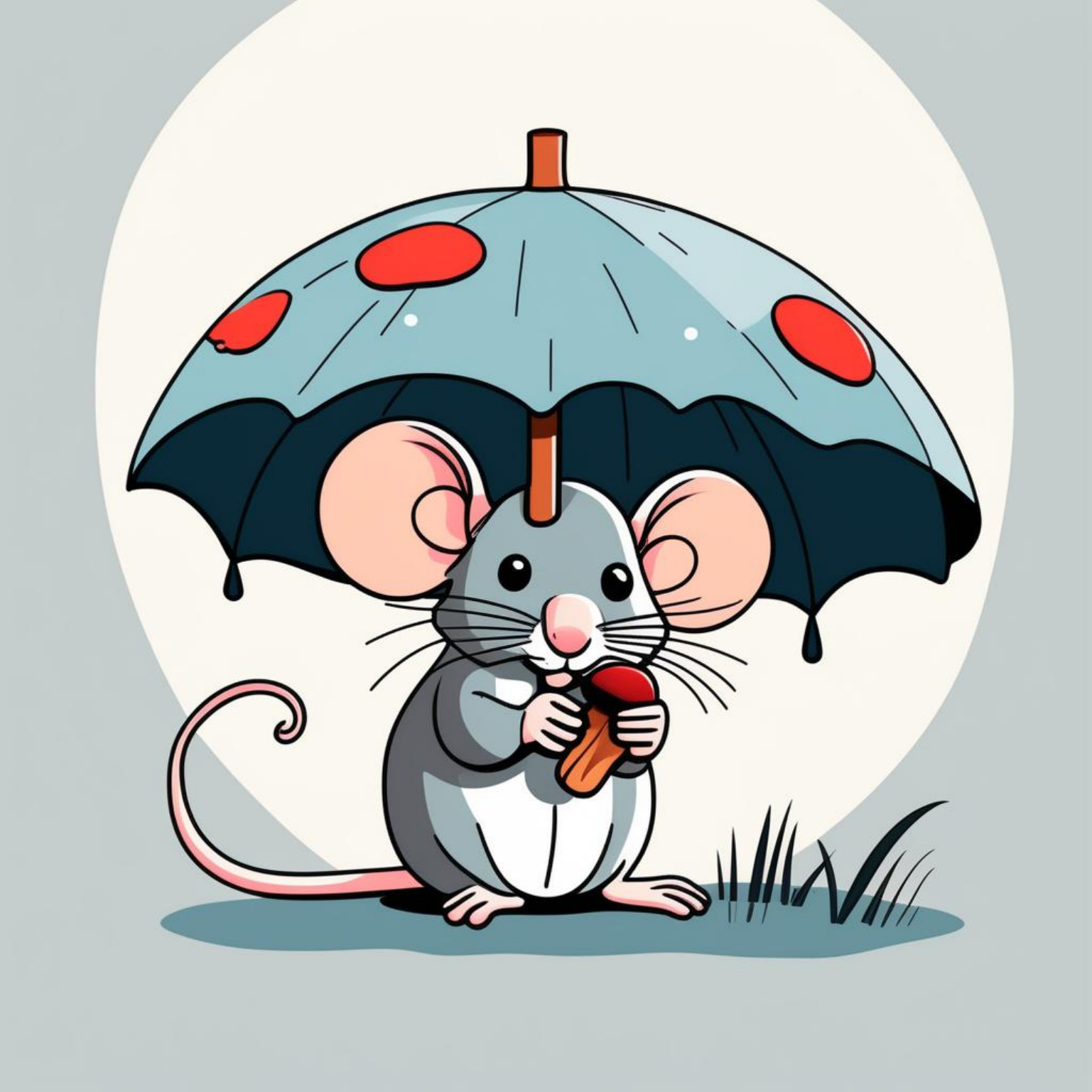} &
        \includegraphics[width=\linewidth]{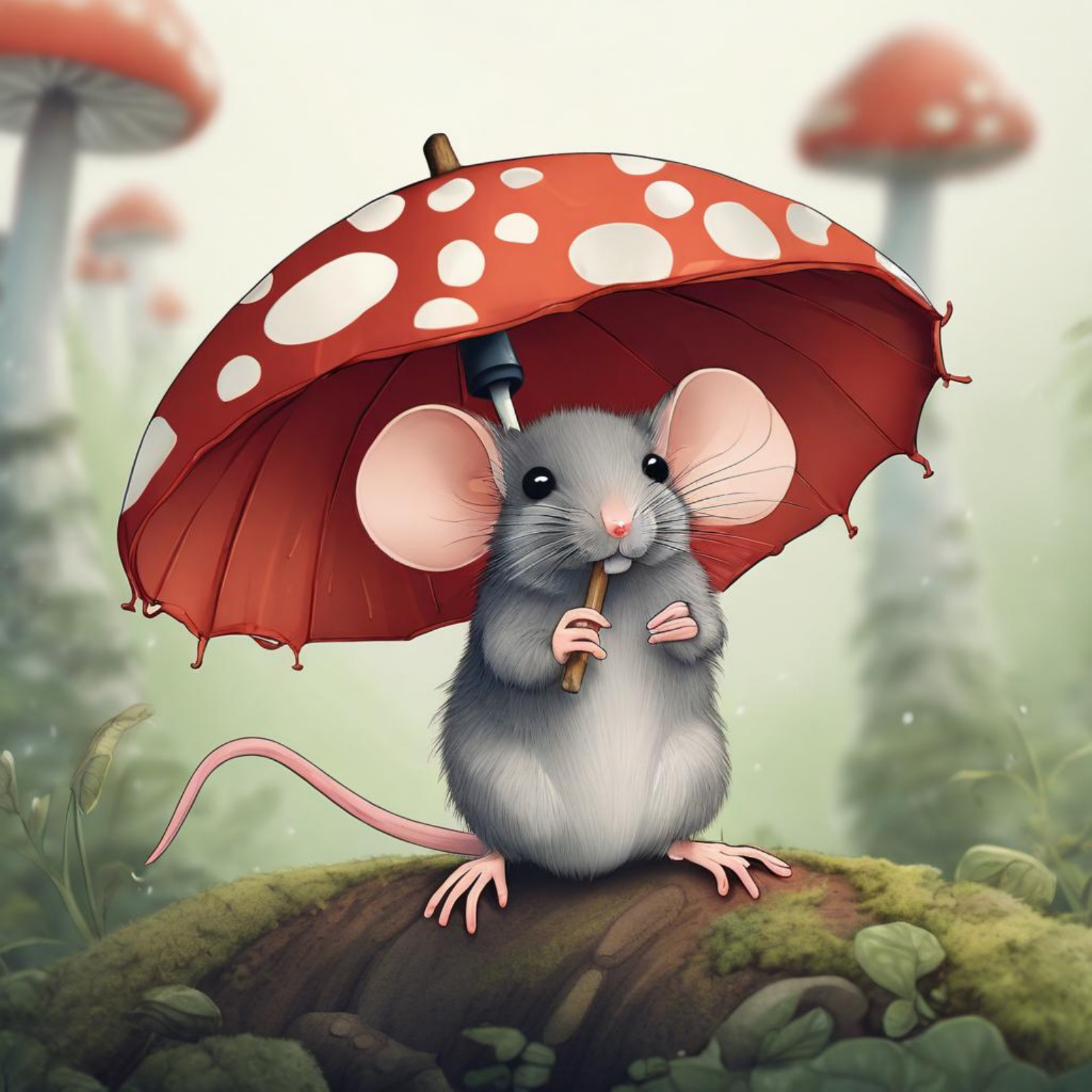} &
        \includegraphics[width=\linewidth]{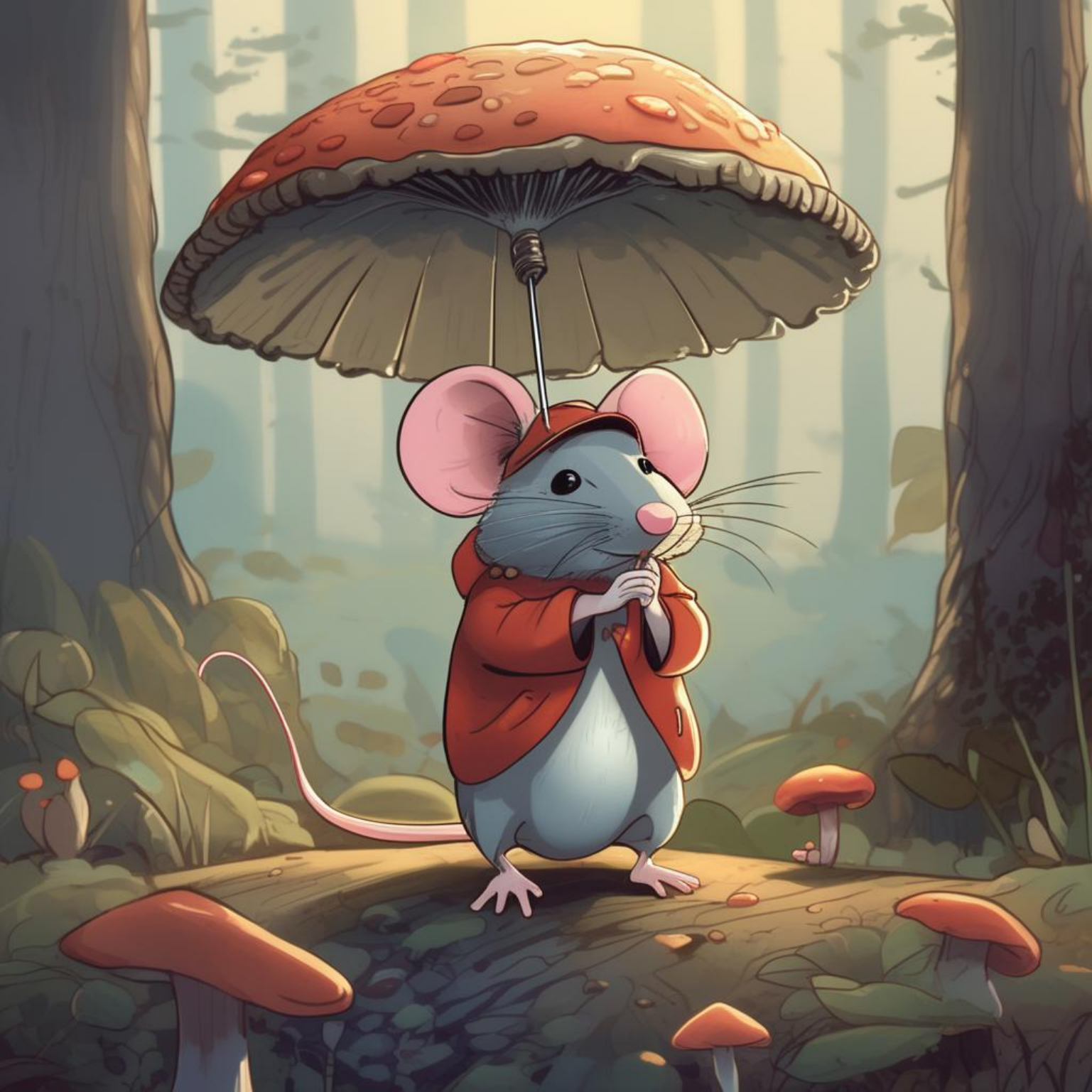} \\

        \multicolumn{8}{@{}p{\linewidth}@{}}{\centering \small \textit{Prompt: Illustration of a mouse using a mushroom as an umbrella.}} \\
		\midrule

        \includegraphics[width=\linewidth]{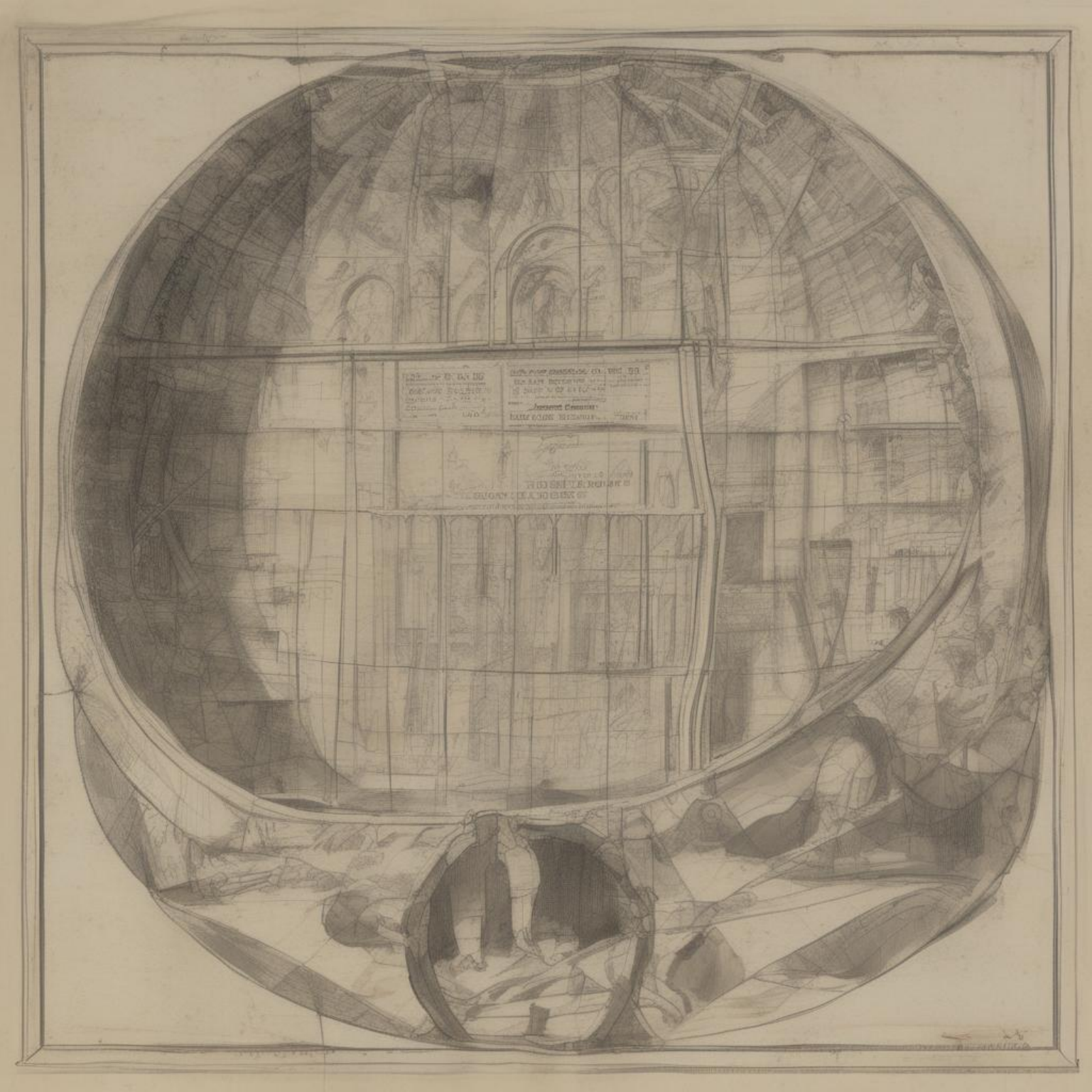} &
        \includegraphics[width=\linewidth]{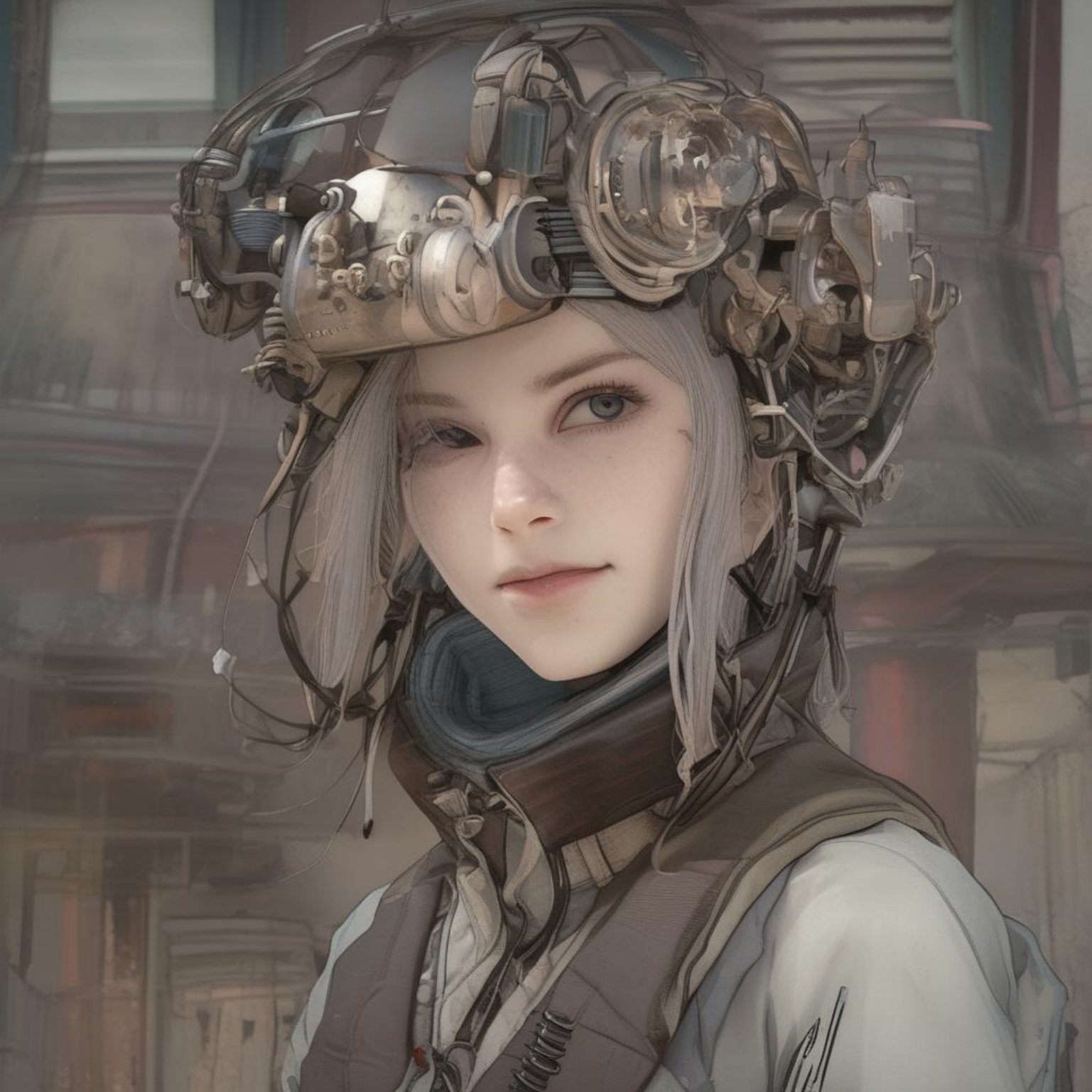} &
        \includegraphics[width=\linewidth]{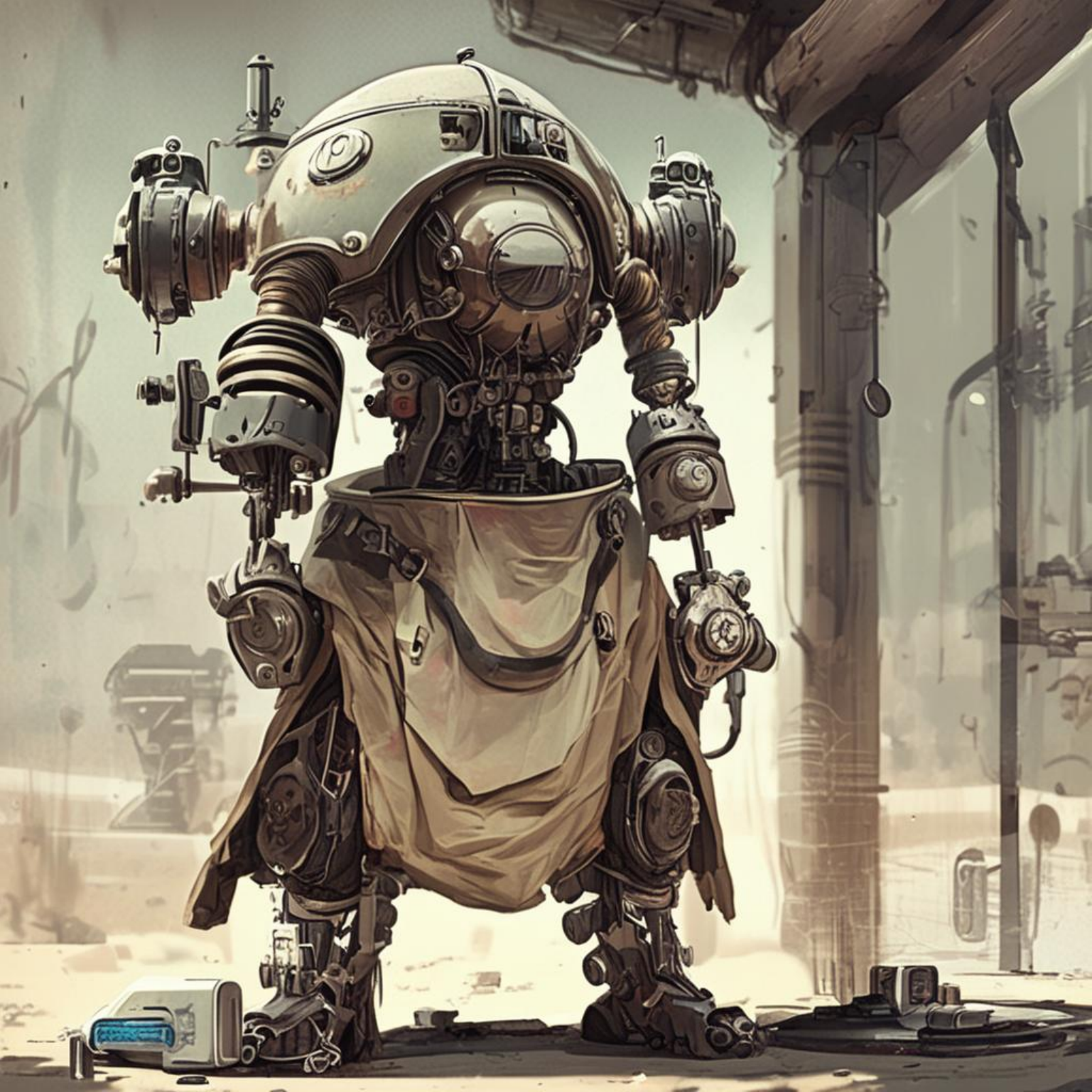} &
        \includegraphics[width=\linewidth]{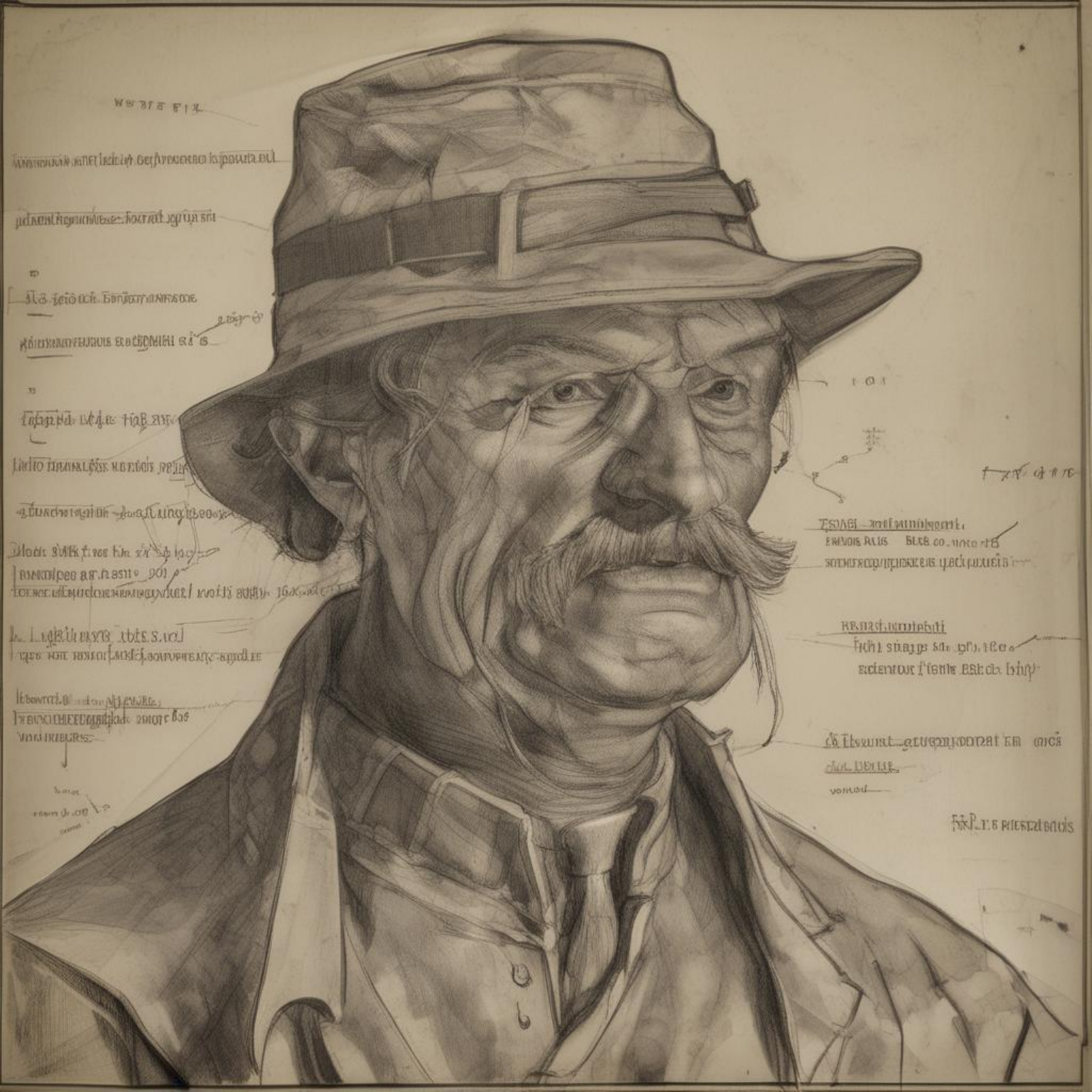} &
        \includegraphics[width=\linewidth]{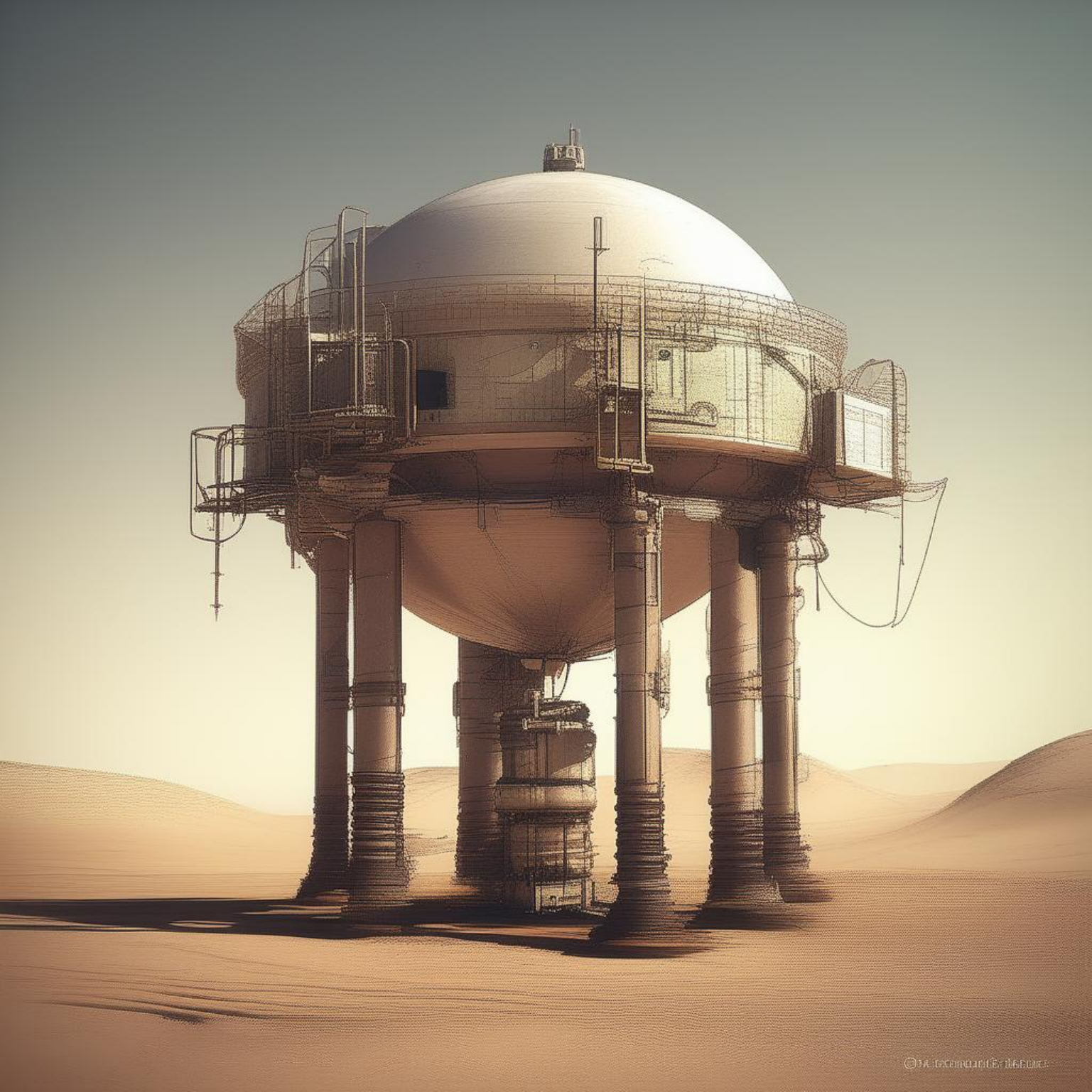} &
        \includegraphics[width=\linewidth]{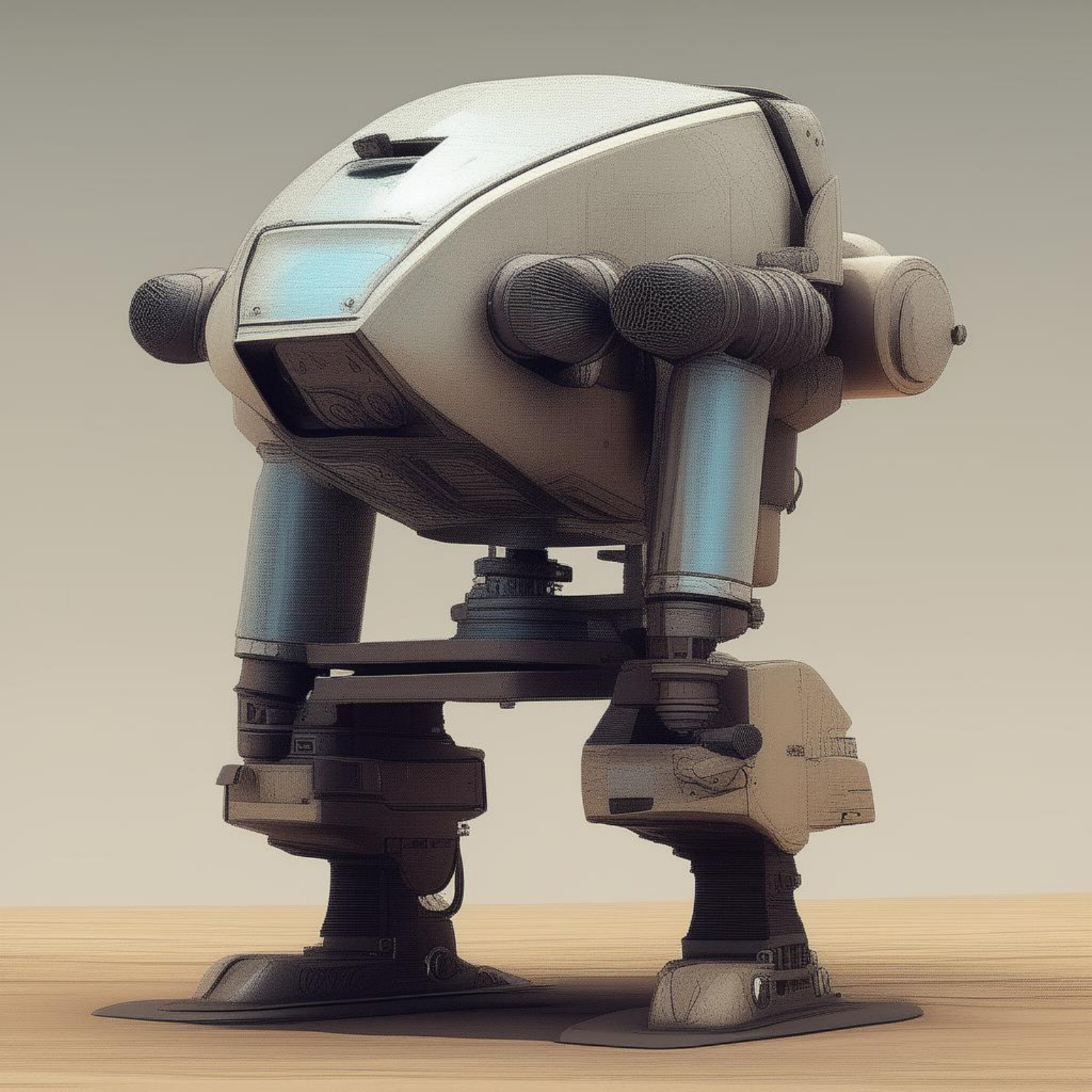} &
        \includegraphics[width=\linewidth]{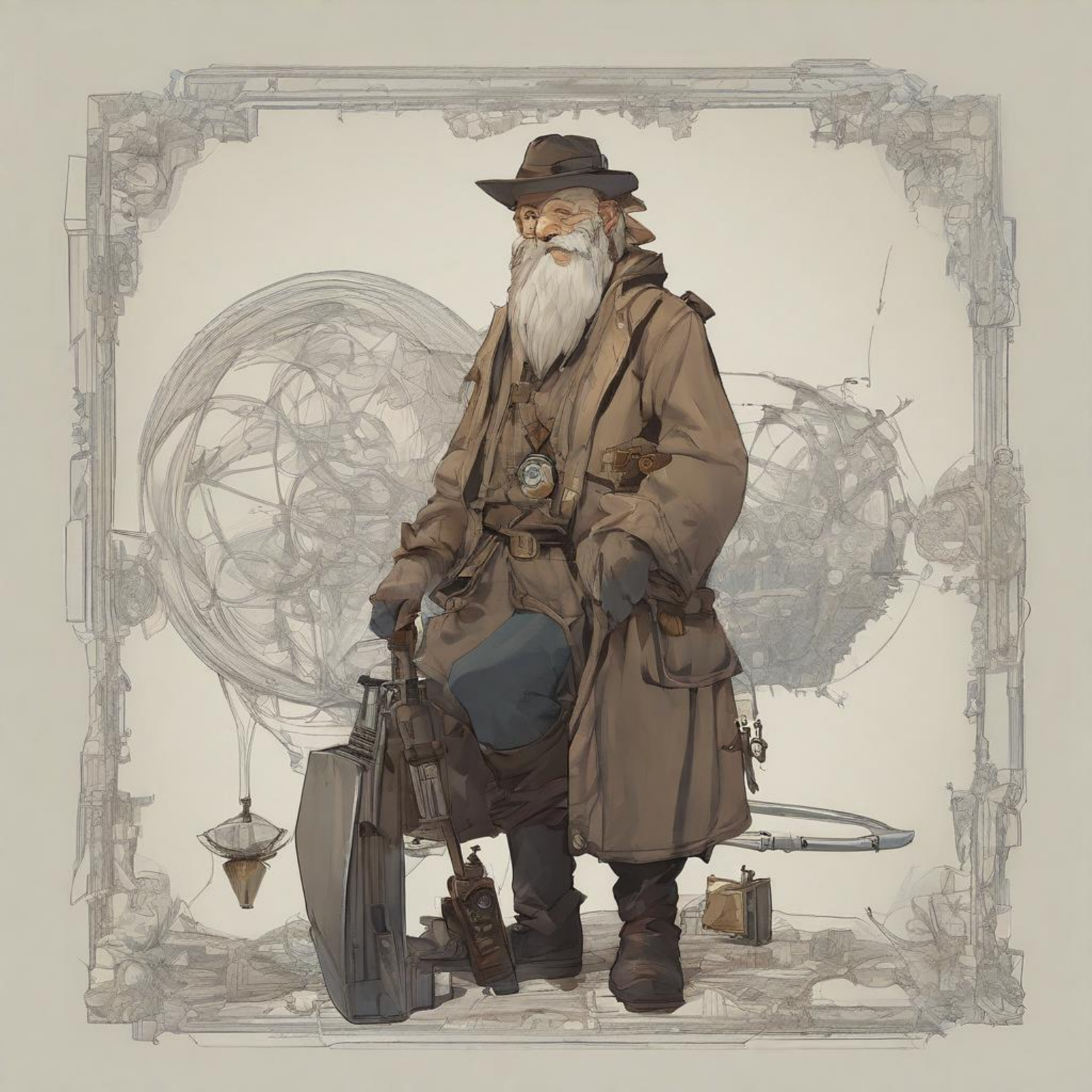} &
        \includegraphics[width=\linewidth]{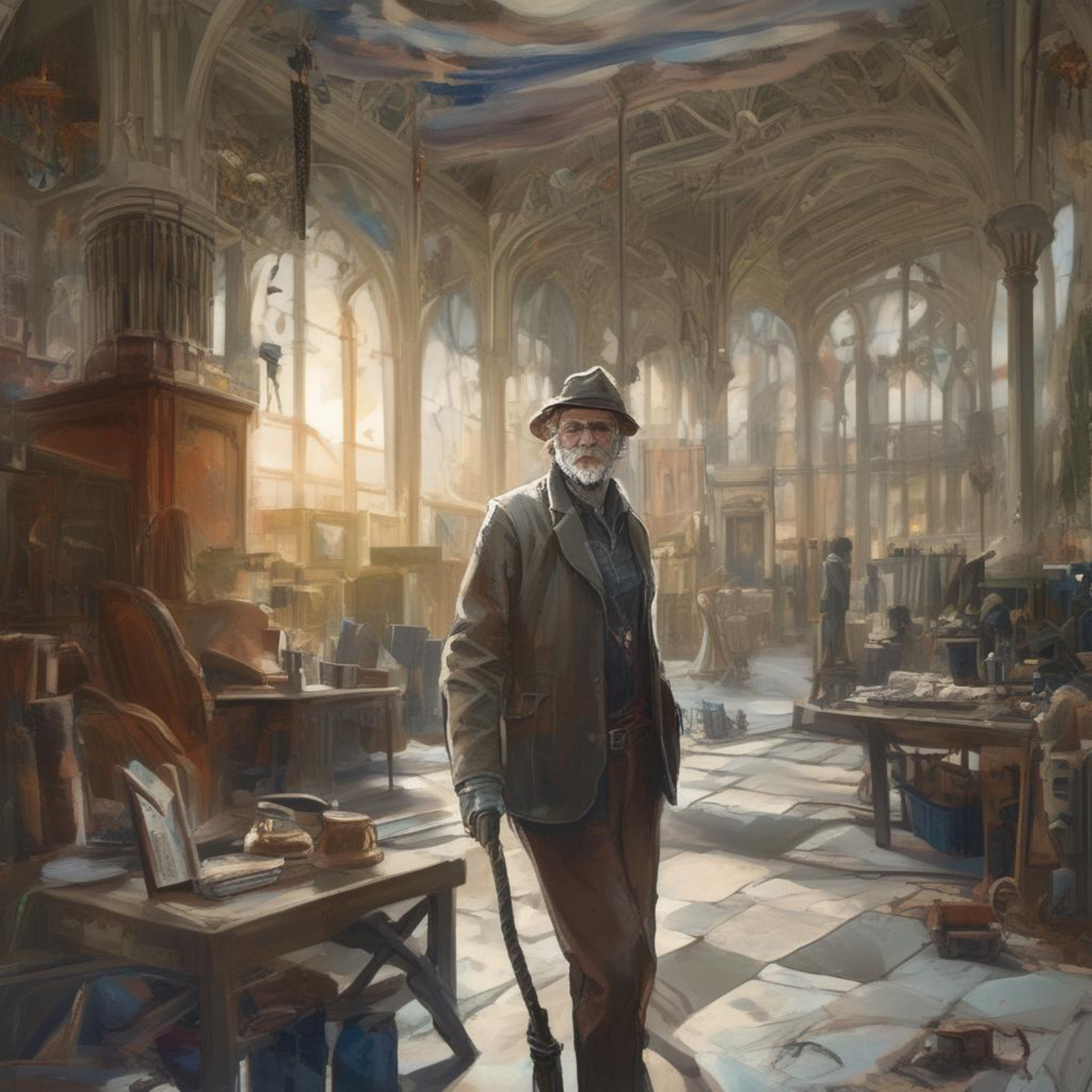} \\

        \multicolumn{8}{@{}p{\linewidth}@{}}{\centering \small \textit{Prompt: A smafml vessef epropoeilled on watvewr by ors, sauls, or han engie.}} \\
		\midrule

        \includegraphics[width=\linewidth]{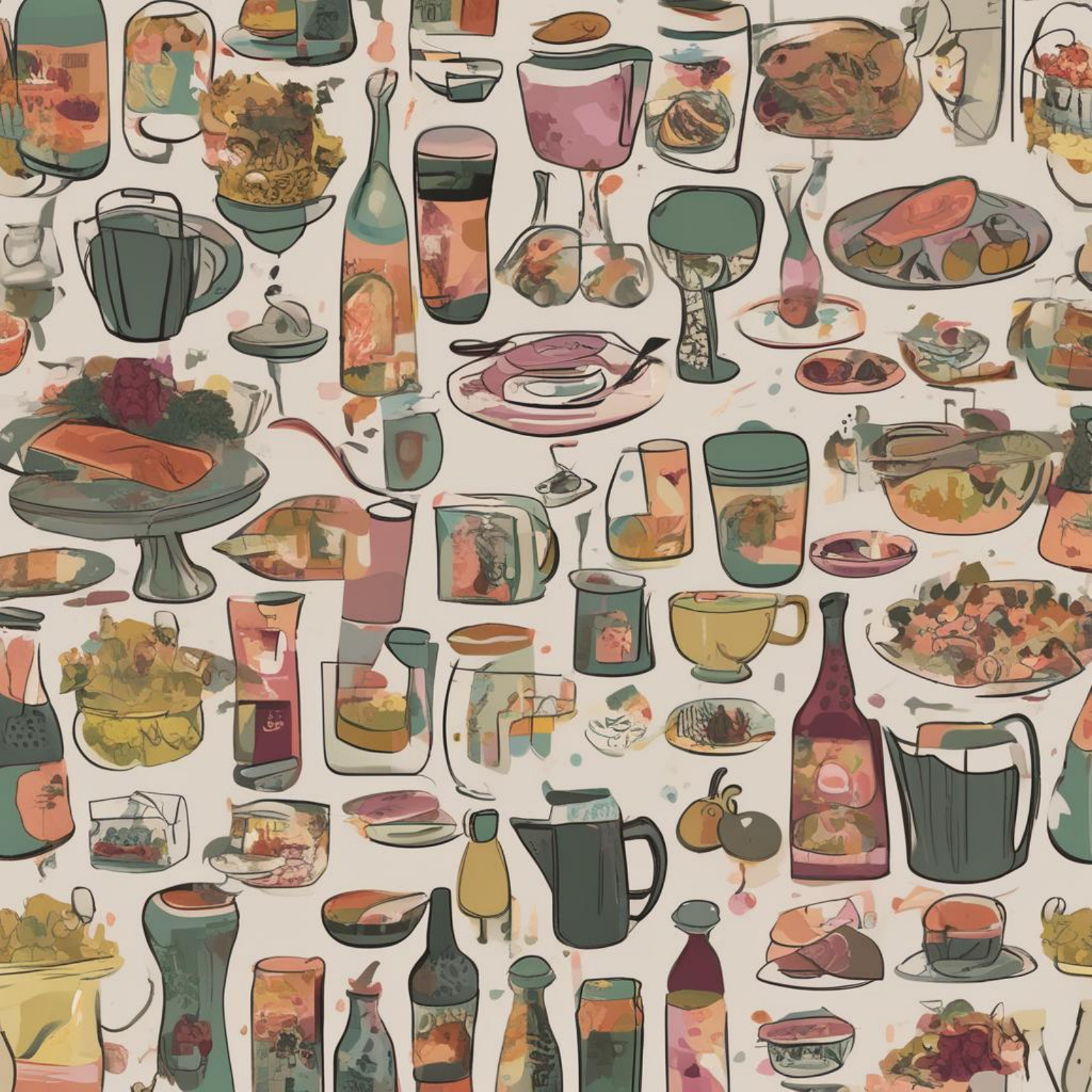} &
        \includegraphics[width=\linewidth]{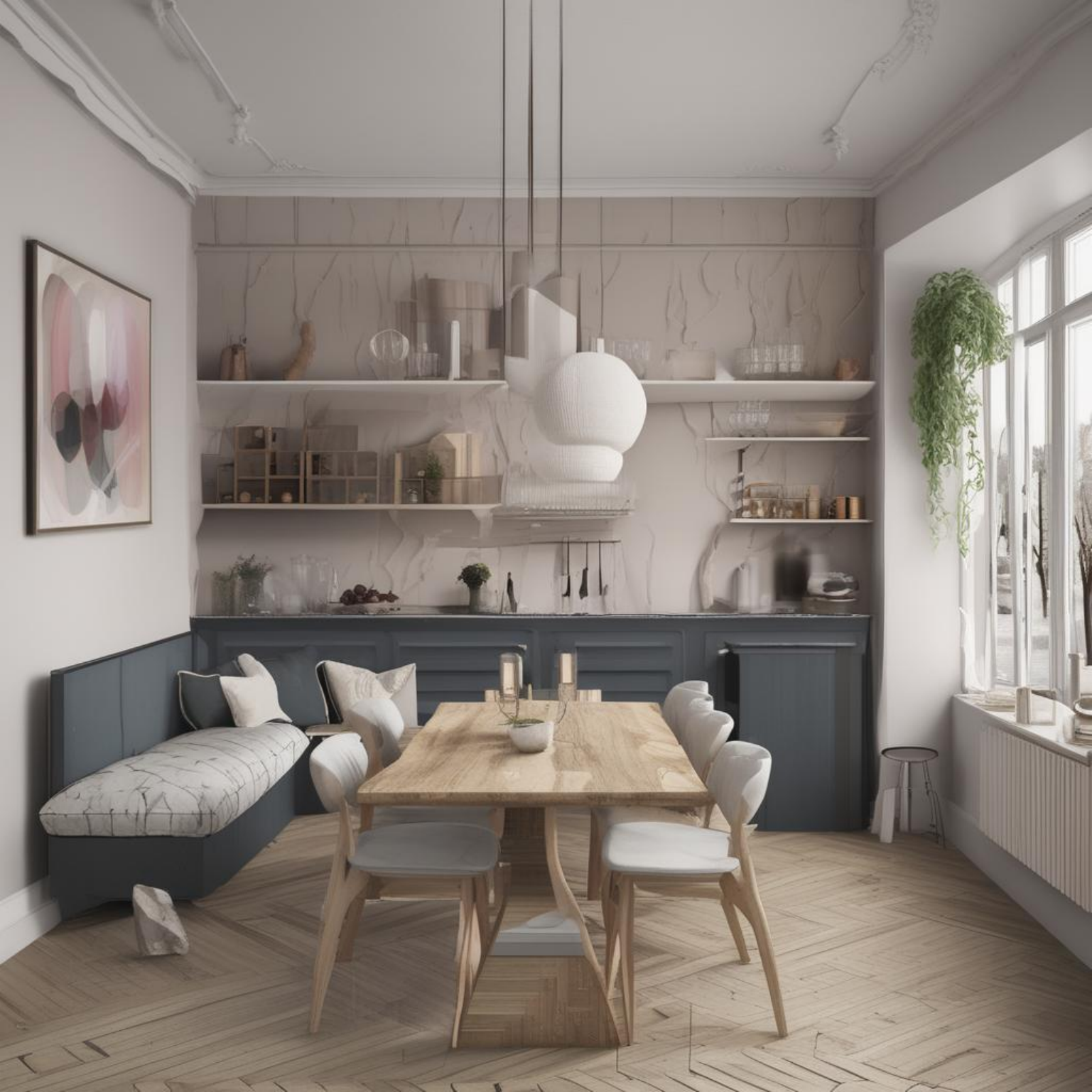} &
        \includegraphics[width=\linewidth]{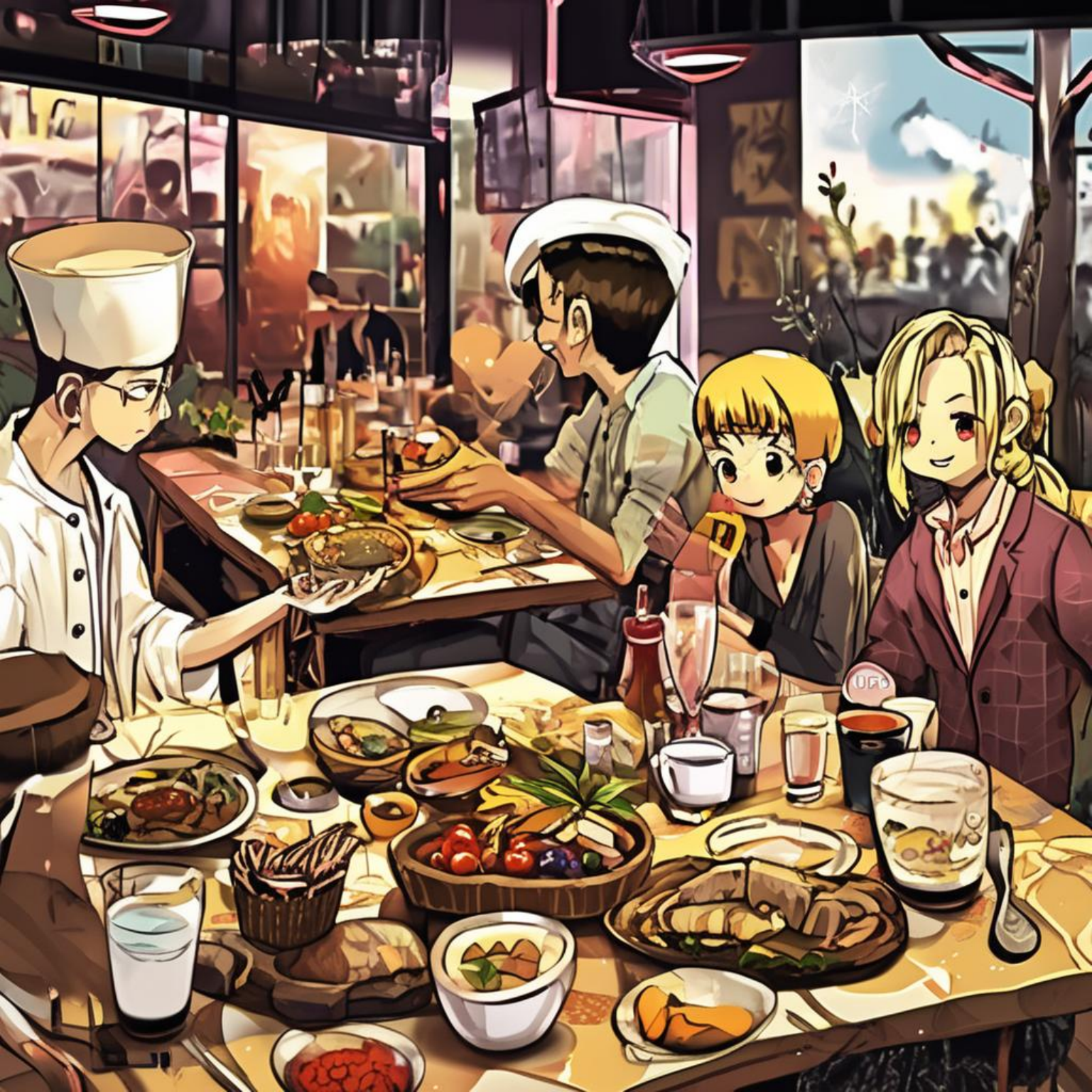} &
        \includegraphics[width=\linewidth]{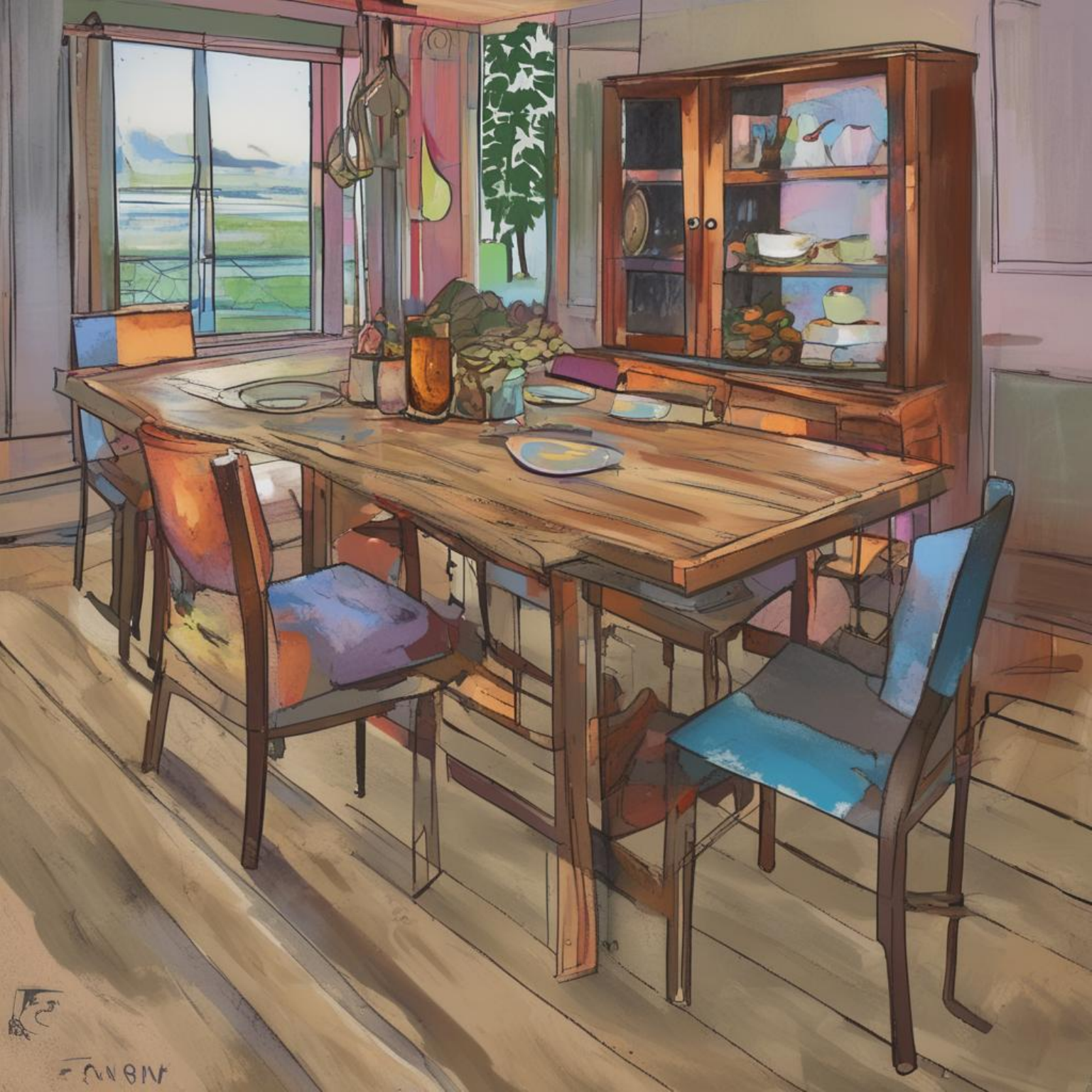} &
        \includegraphics[width=\linewidth]{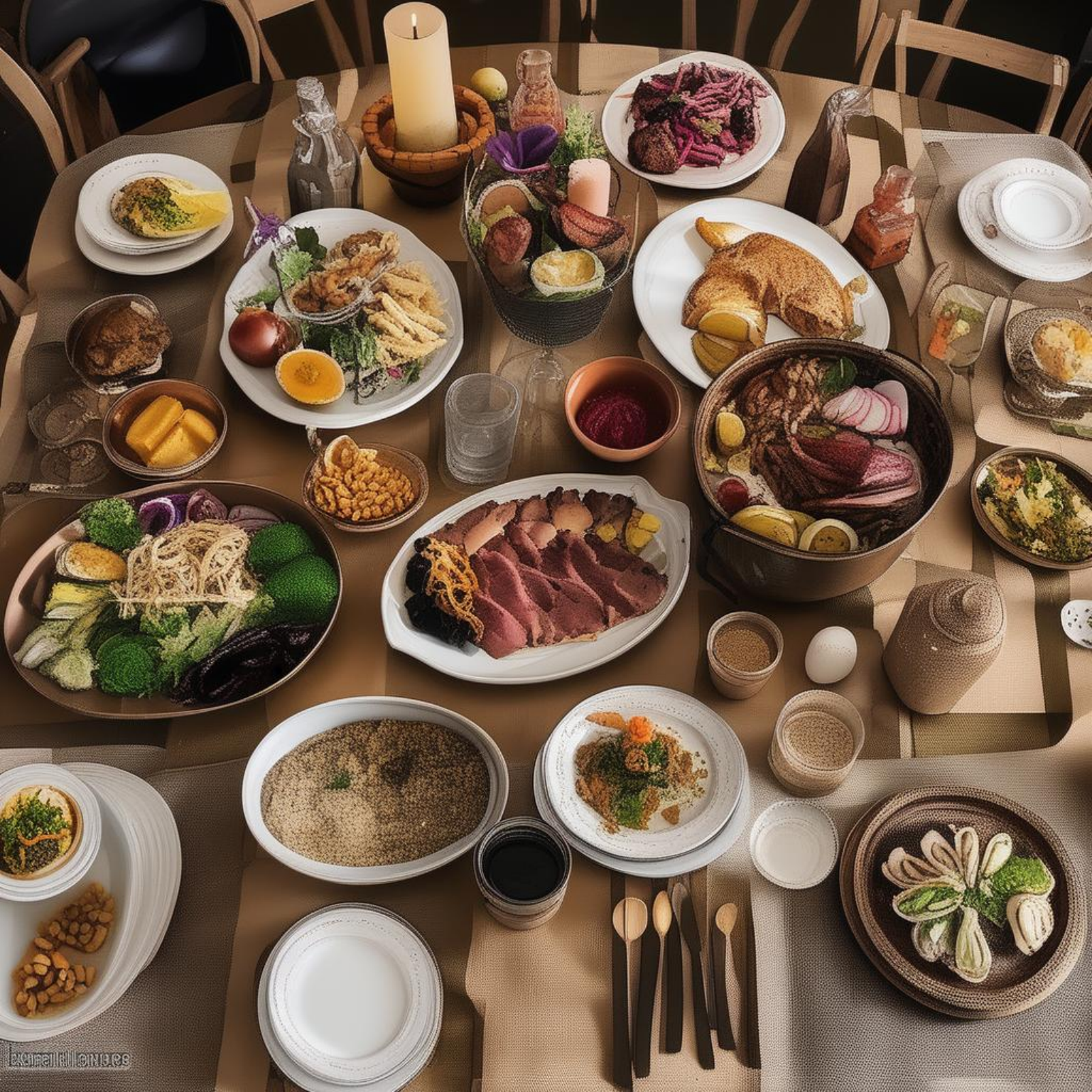} &
        \includegraphics[width=\linewidth]{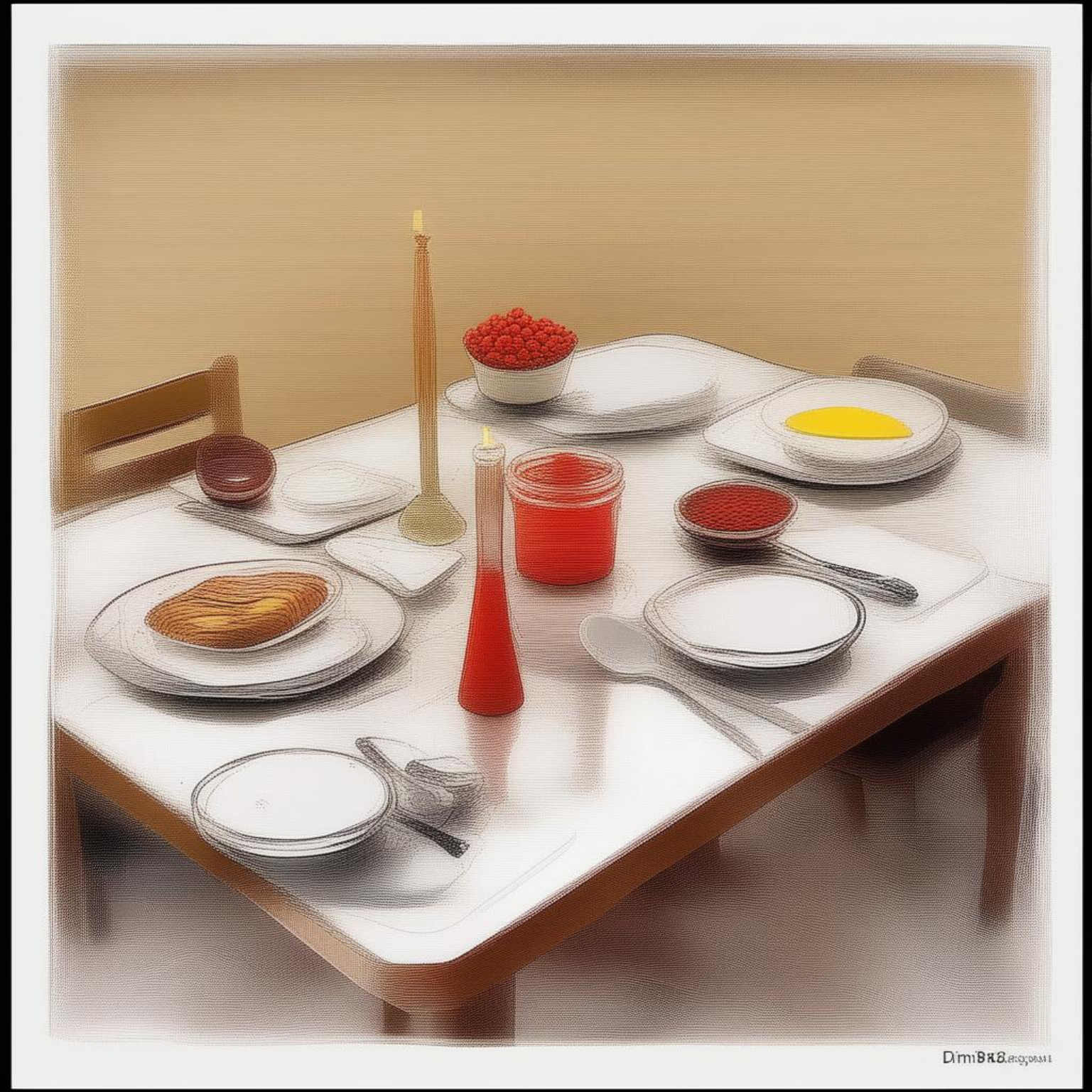} &
        \includegraphics[width=\linewidth]{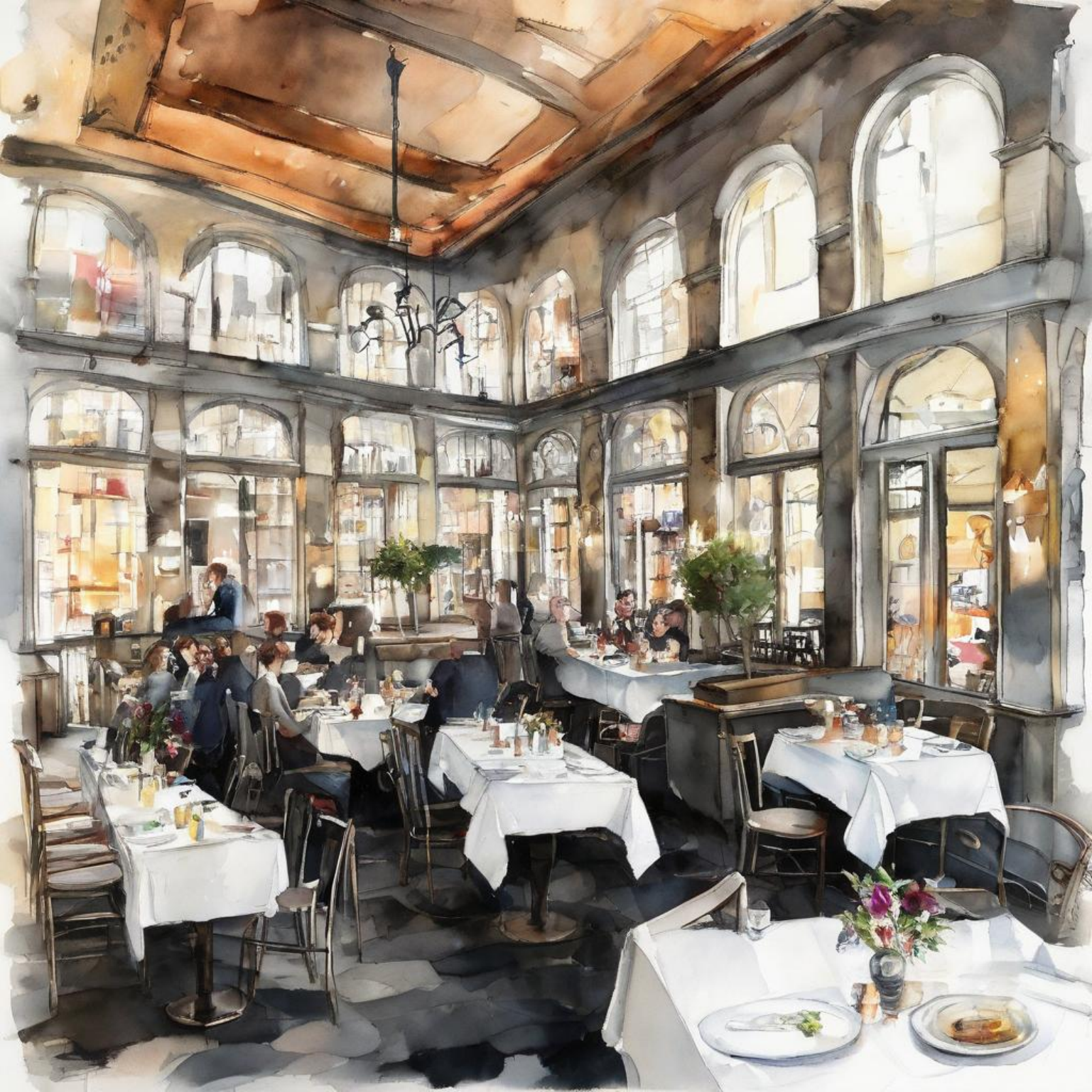} &
        \includegraphics[width=\linewidth]{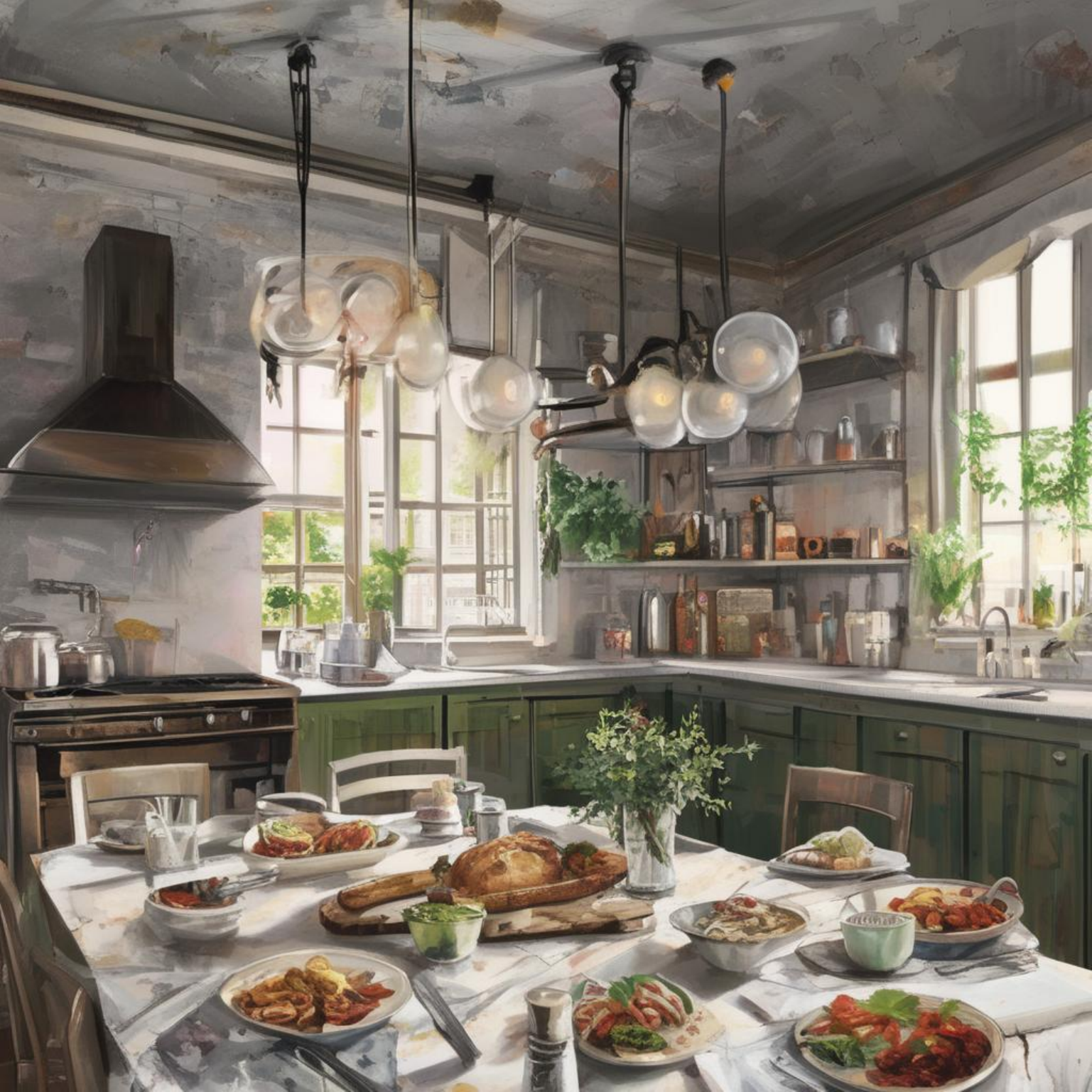} \\

        \multicolumn{8}{@{}p{\linewidth}@{}}{\centering \small \textit{Prompt: Dininrg tablez.}} \\

        \bottomrule
        
    \end{tabularx}
    \label{fig:qualitative_sdxl} 
    \caption{Visual comparison of SDXL with $\text{NRE}=1000$, targeting the Aesthetic reward model.} 
\end{figure}

\begin{figure}[t] 
    \centering
    \setlength{\tabcolsep}{1pt} 
    
    \begin{tabularx}{\textwidth}{YYYYY}
        
        \toprule
        \textbf{FLUX.1-dev} & \textbf{BoN} & \textbf{ZO-N} & \textbf{SoP} & \textbf{SES} \tabularnewline
        \midrule
        
        \includegraphics[width=\linewidth]{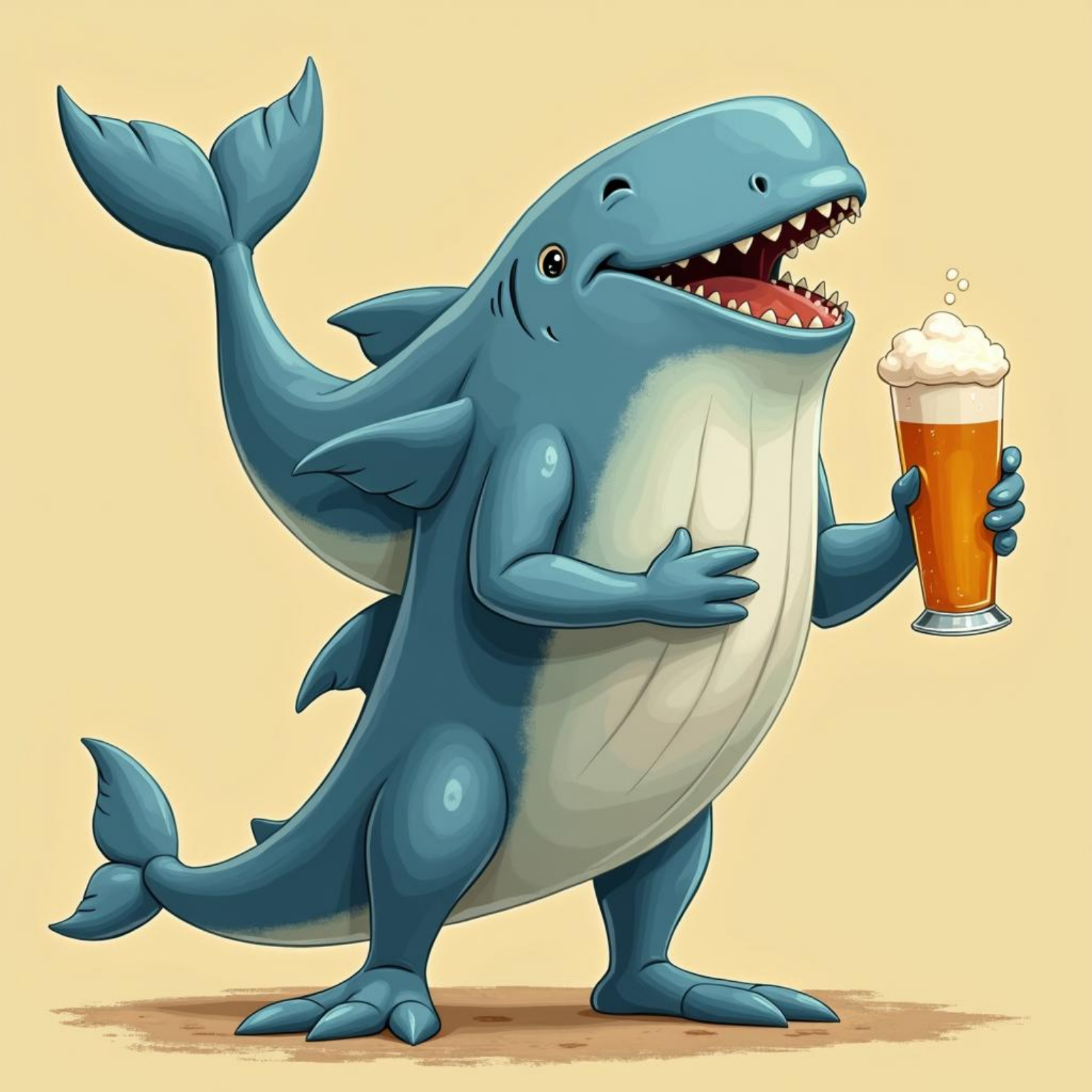} &
        \includegraphics[width=\linewidth]{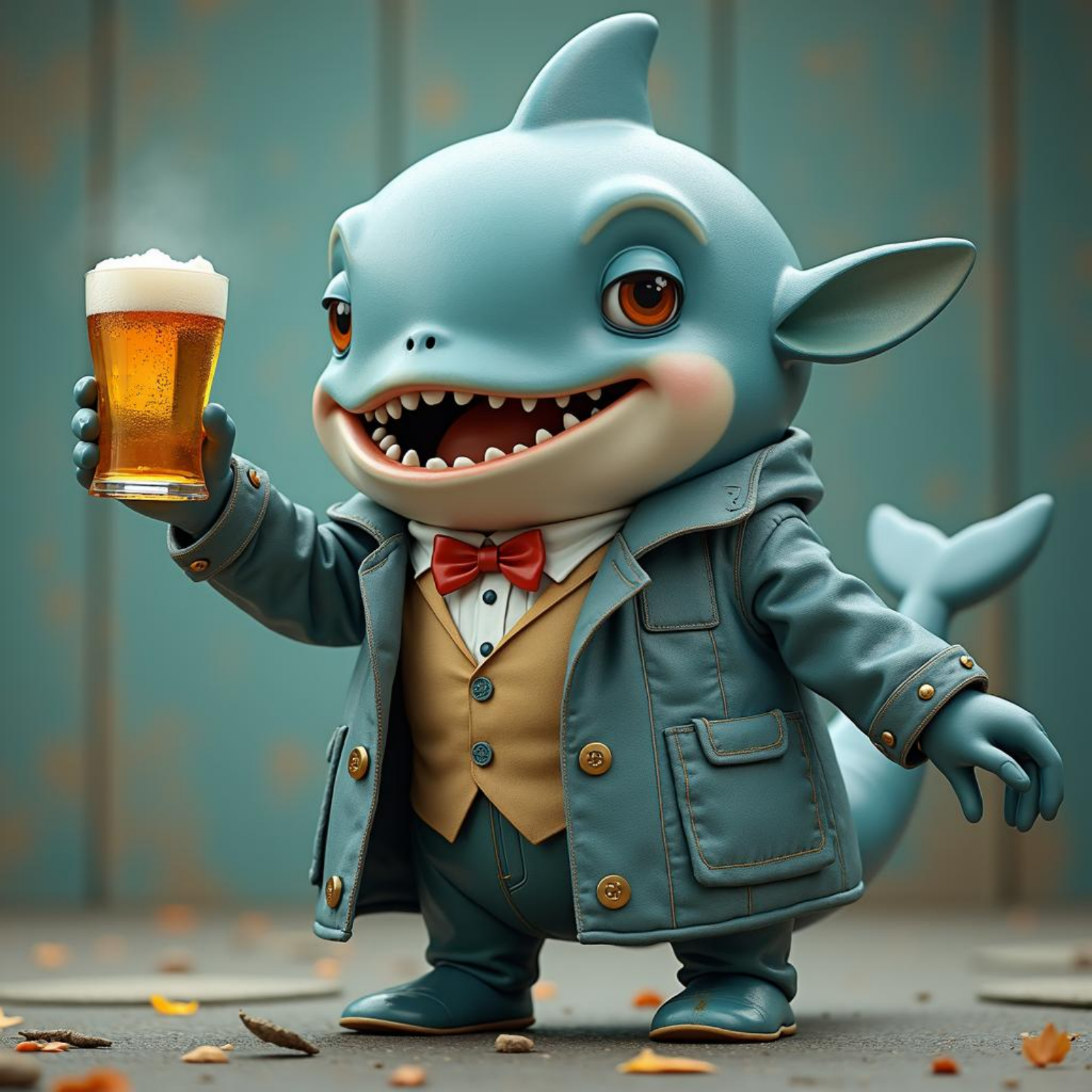} &
        \includegraphics[width=\linewidth]{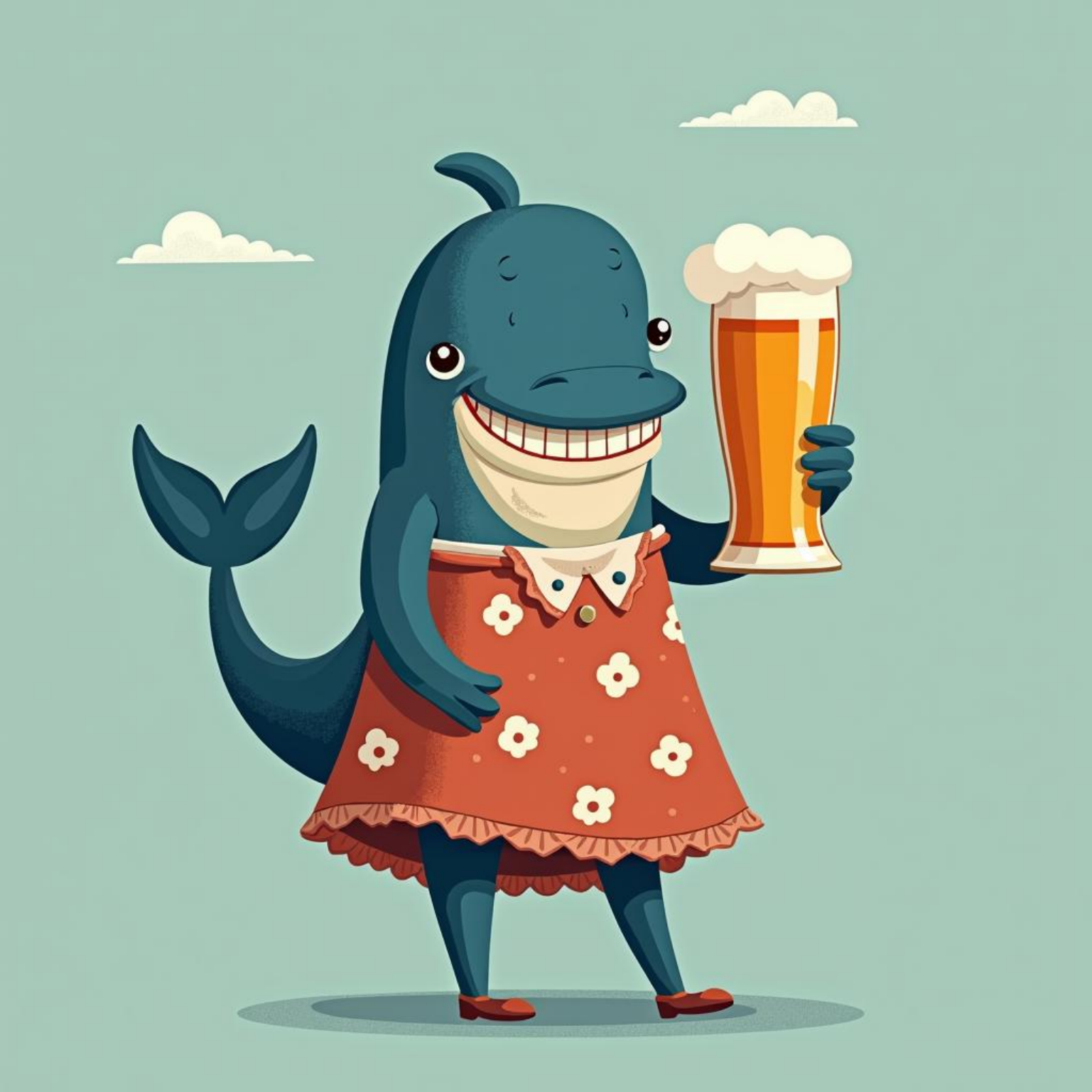} &
        \includegraphics[width=\linewidth]{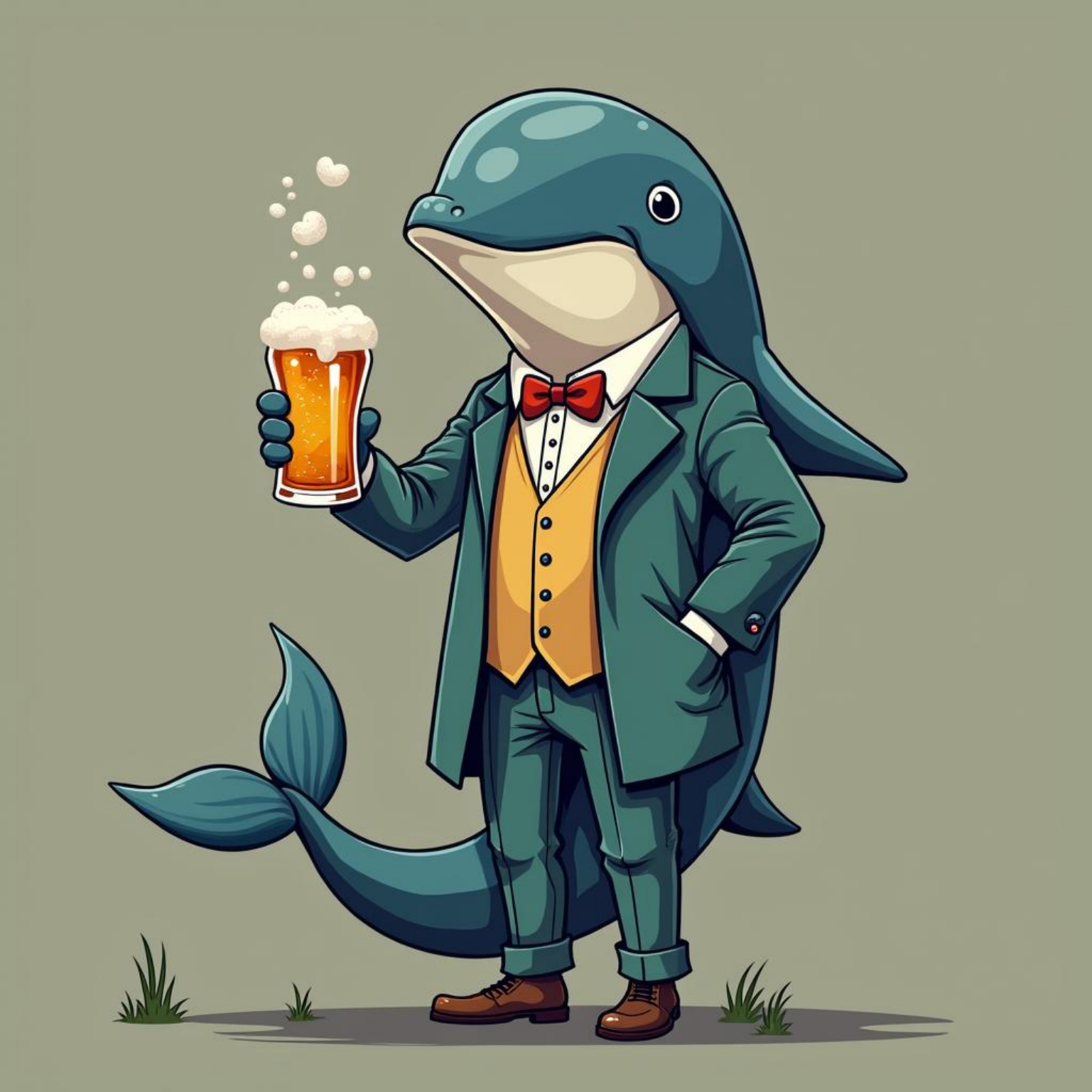} &
        \includegraphics[width=\linewidth]{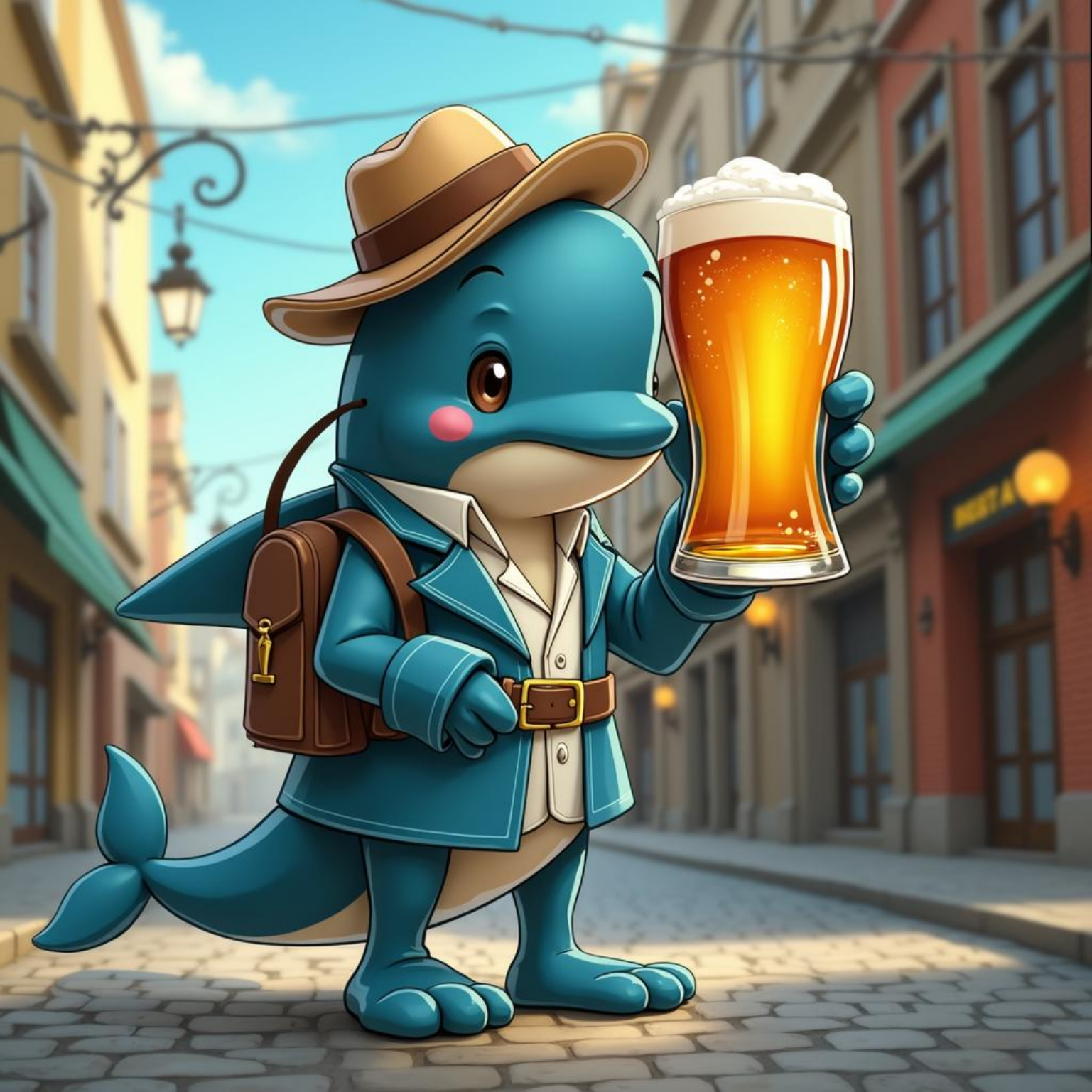} \\

        \multicolumn{5}{>{\centering\arraybackslash}p{\dimexpr\textwidth-2\tabcolsep}}{
            \small \textit{Prompt: An Image of animated whale with a beer in hand in tenue de ville dress}
        } \\
        \midrule

        \includegraphics[width=\linewidth]{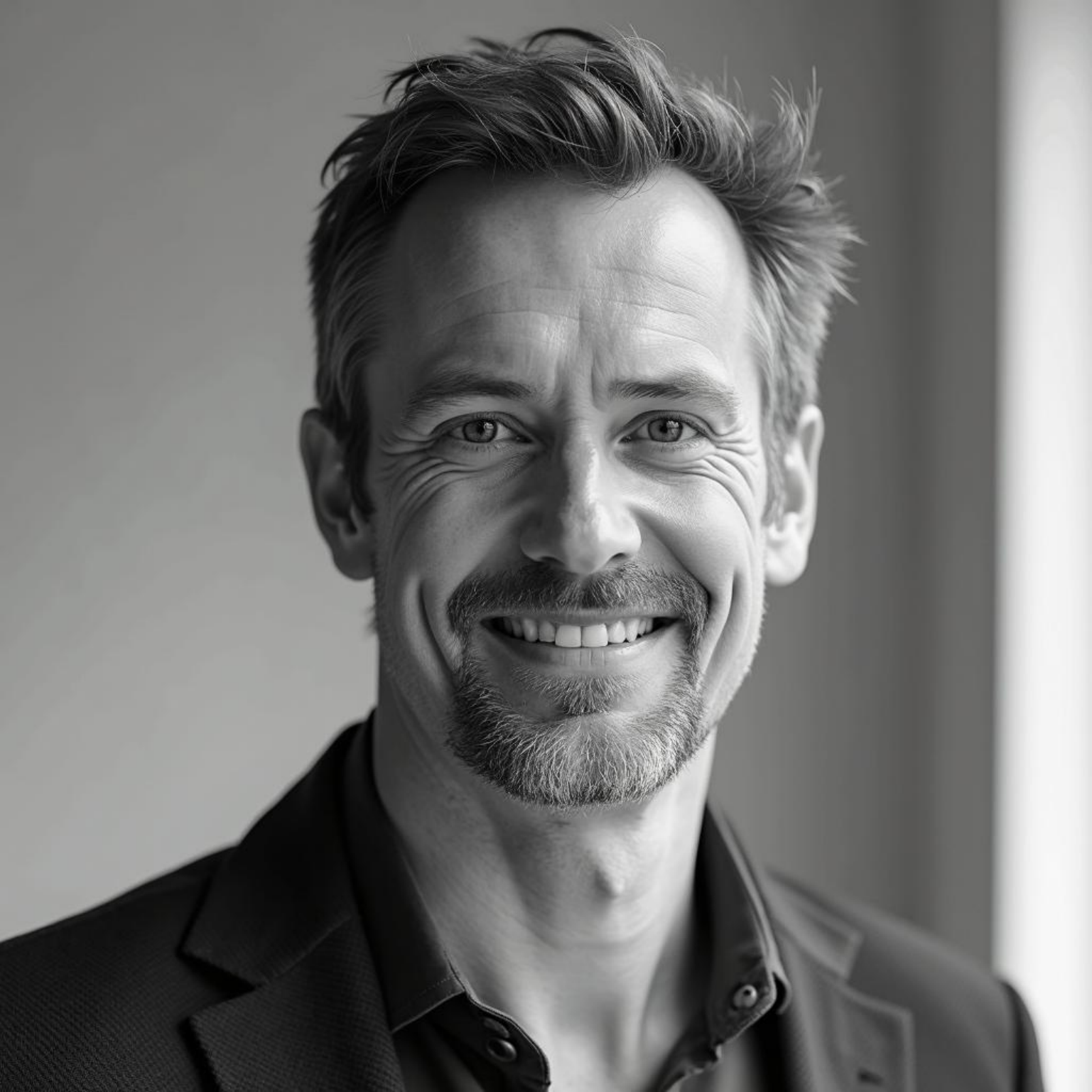} &
        \includegraphics[width=\linewidth]{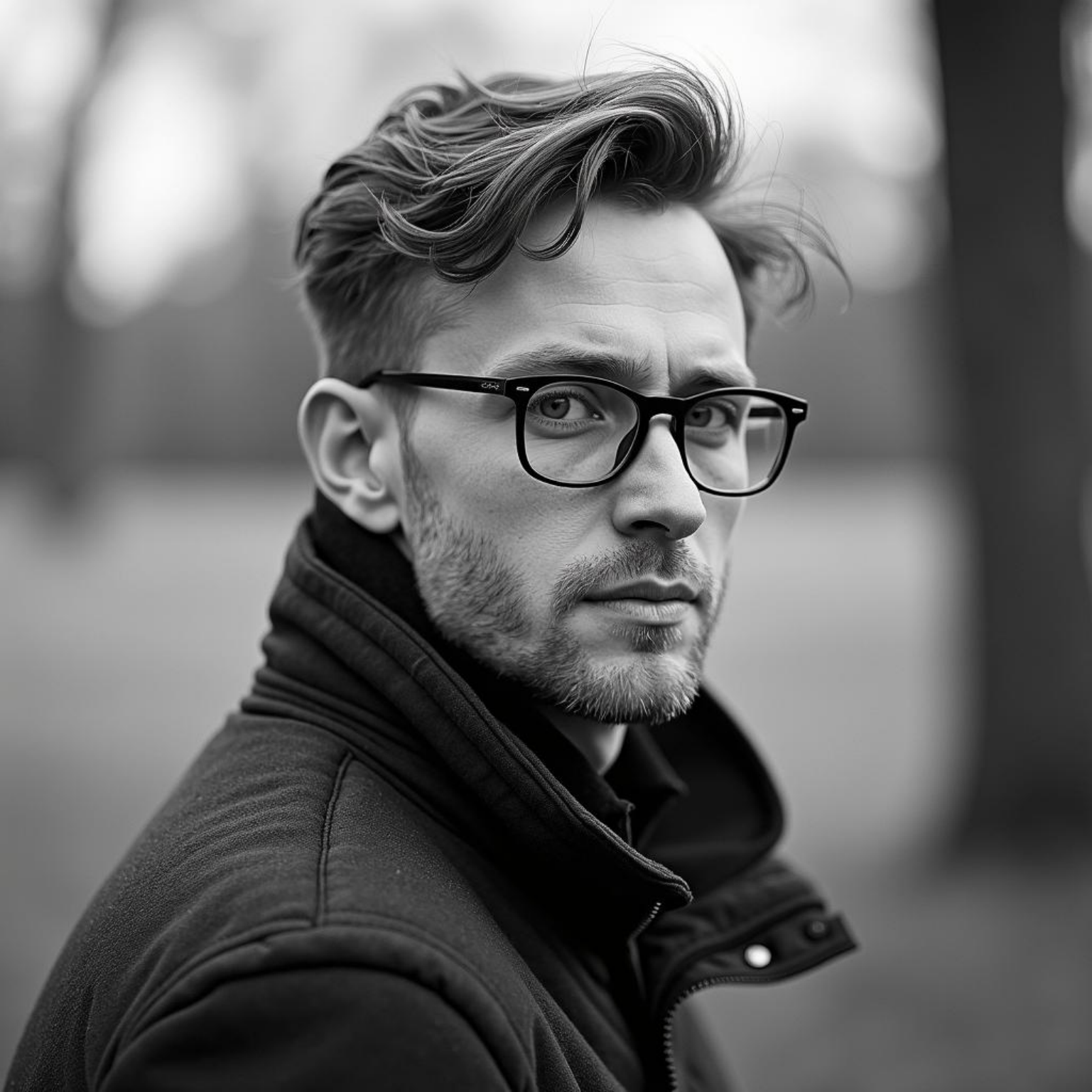} &
        \includegraphics[width=\linewidth]{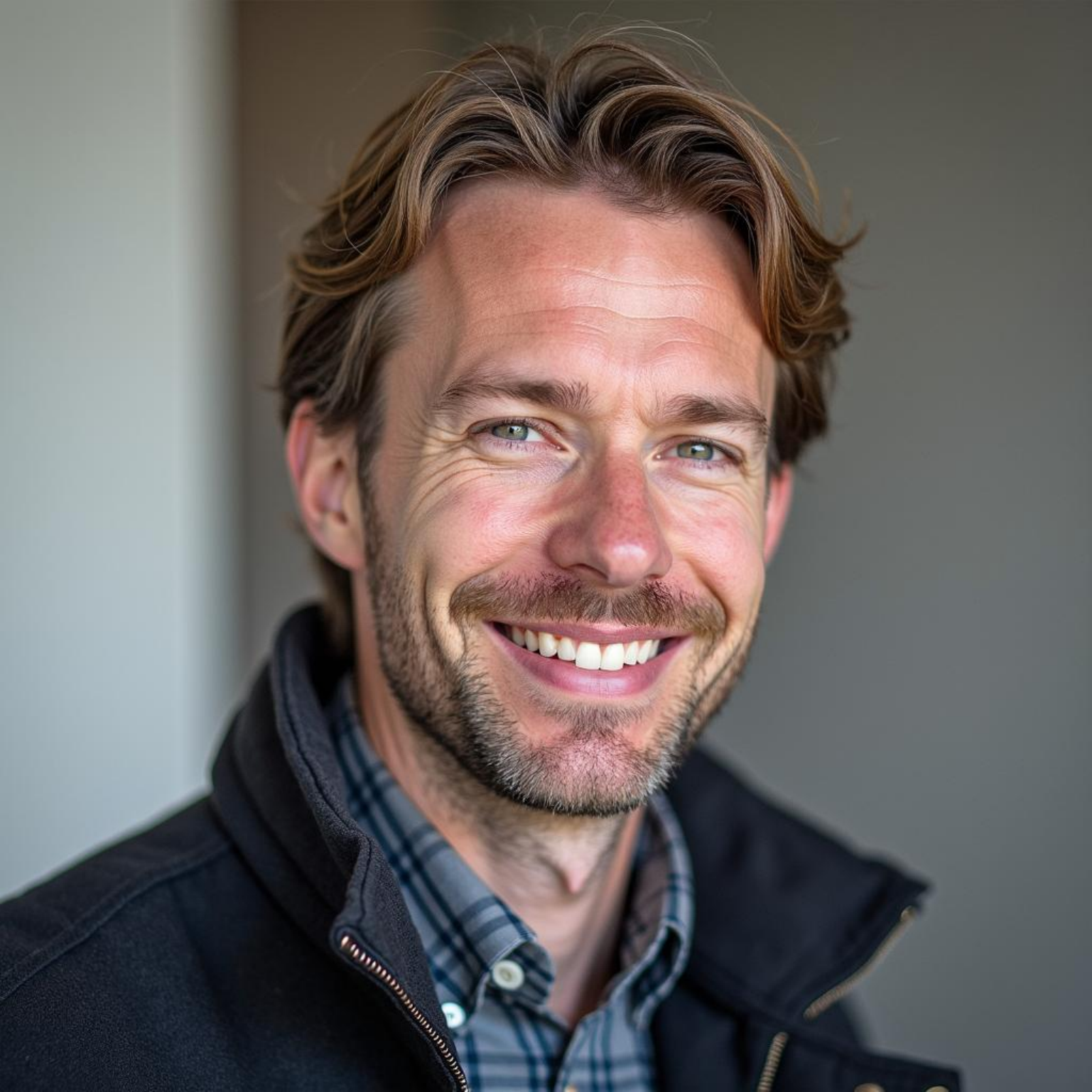} &
        \includegraphics[width=\linewidth]{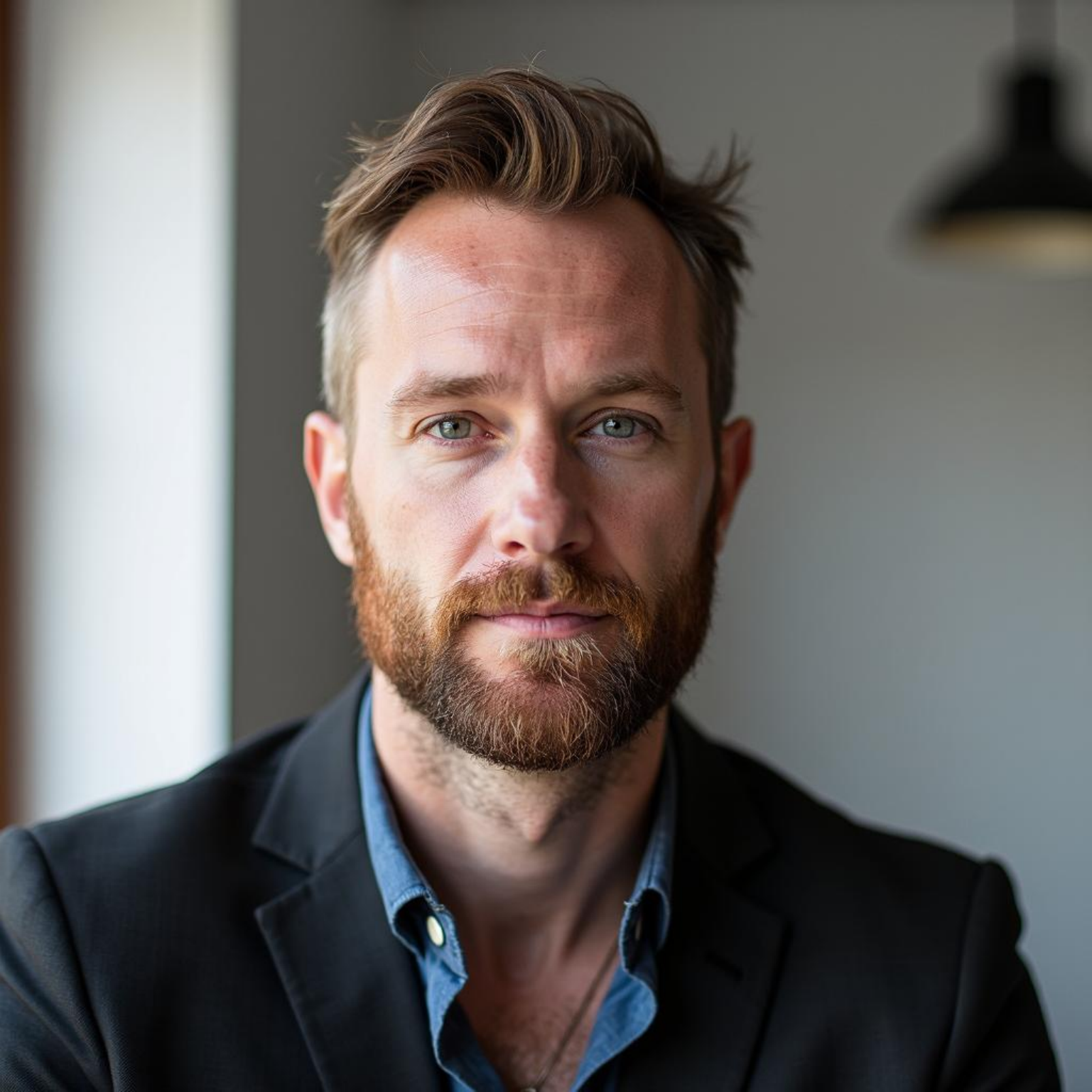} &
        \includegraphics[width=\linewidth]{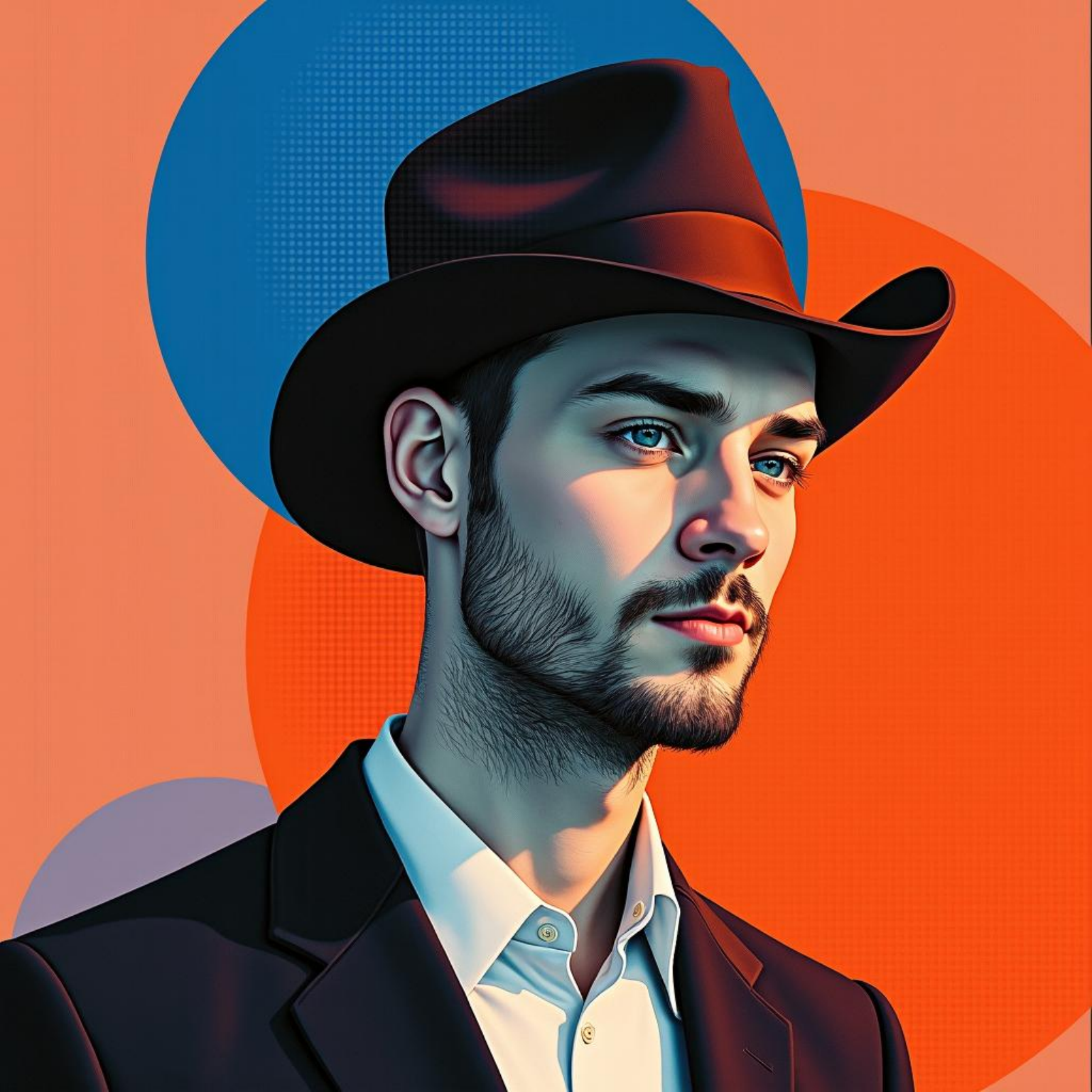} \\

        \multicolumn{5}{>{\centering\arraybackslash}p{\dimexpr\textwidth-2\tabcolsep}}{
            \small \textit{Prompt: Torben Ægidius Mogensen}
        } \\
        \midrule

        \includegraphics[width=\linewidth]{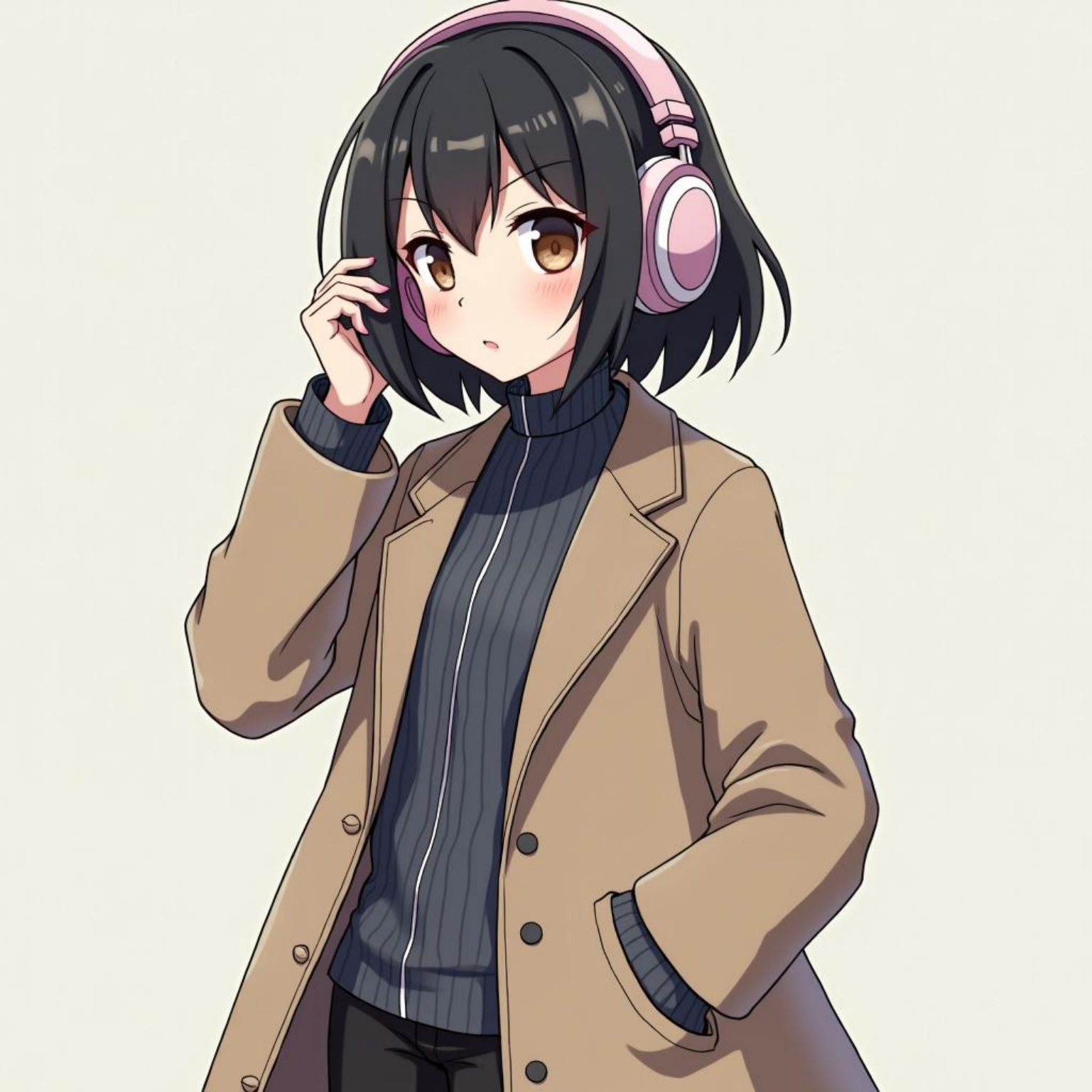} &
        \includegraphics[width=\linewidth]{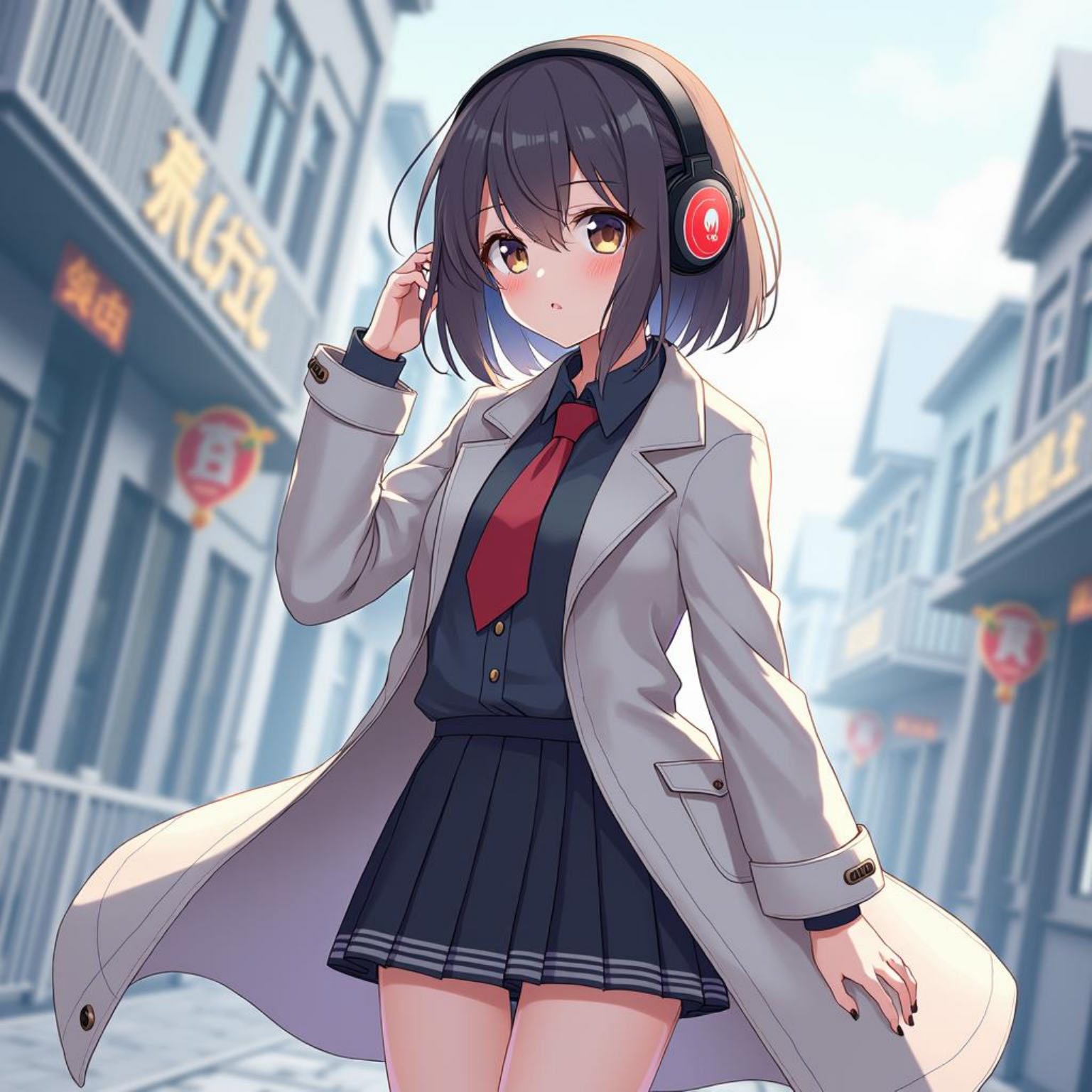} &
        \includegraphics[width=\linewidth]{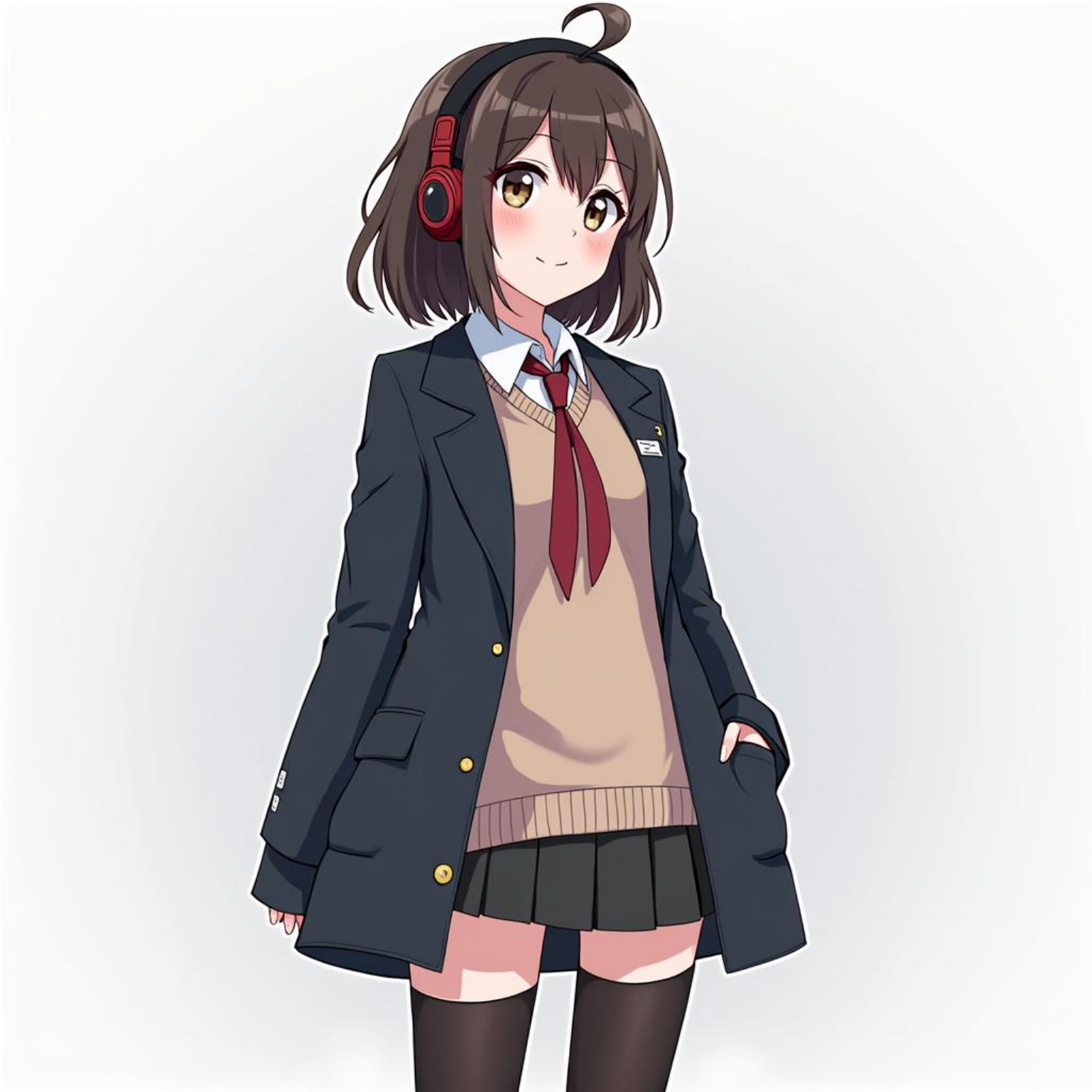} &
        \includegraphics[width=\linewidth]{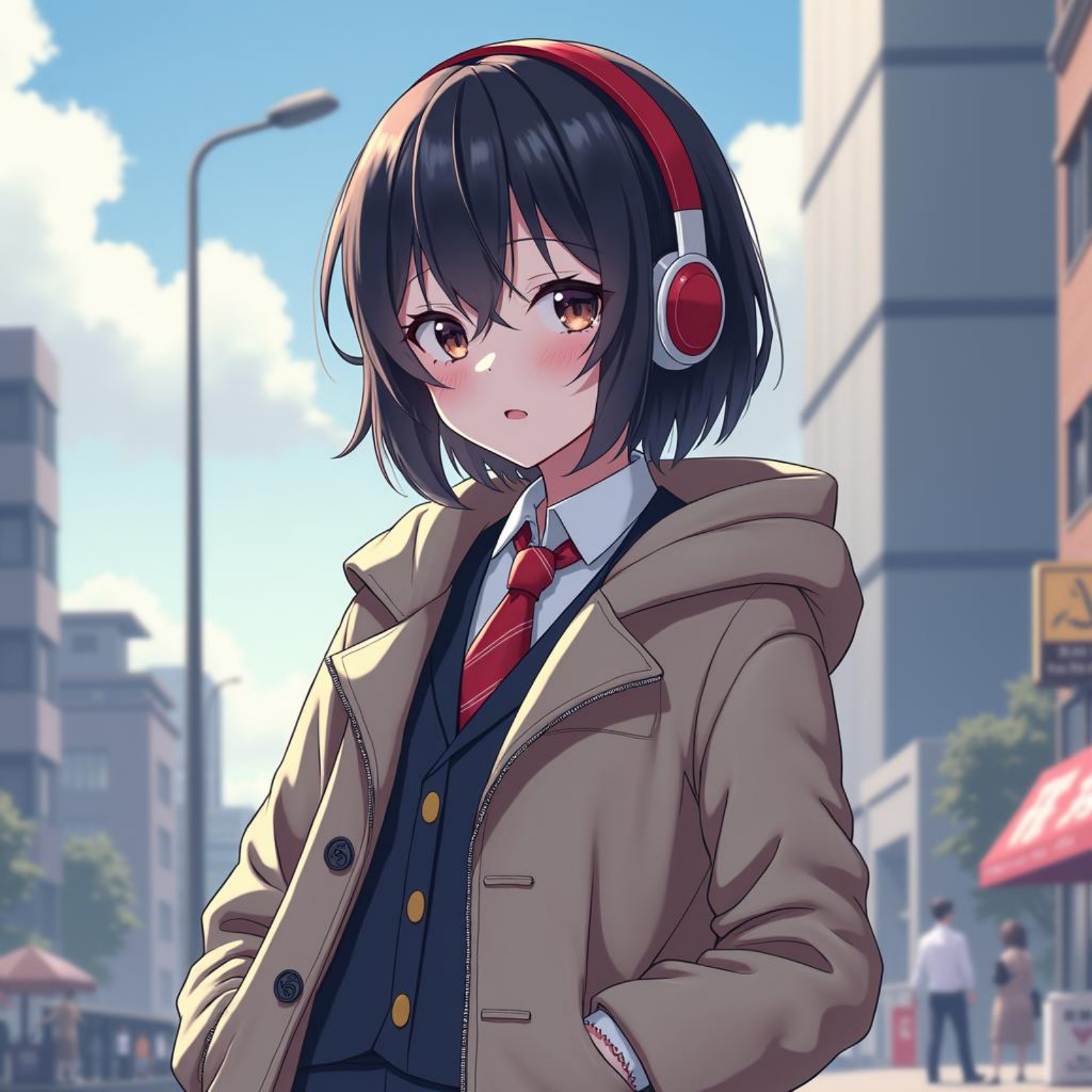} &
        \includegraphics[width=\linewidth]{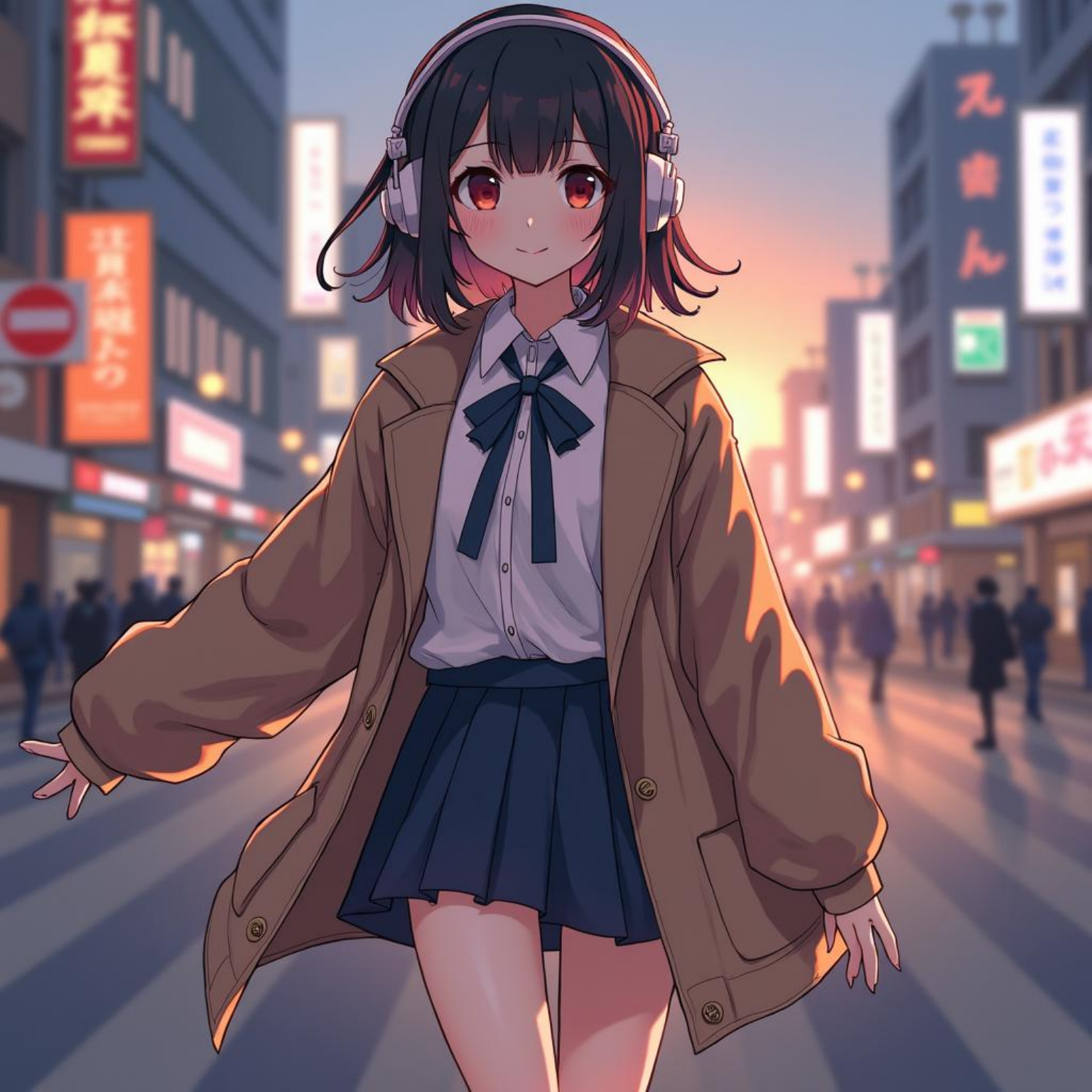} \\

        \multicolumn{5}{>{\centering\arraybackslash}p{\dimexpr\textwidth-2\tabcolsep}}{
            \small \textit{Prompt: fullbody japanese short headphones long coat girl}
        } \\
        \midrule

        \includegraphics[width=\linewidth]{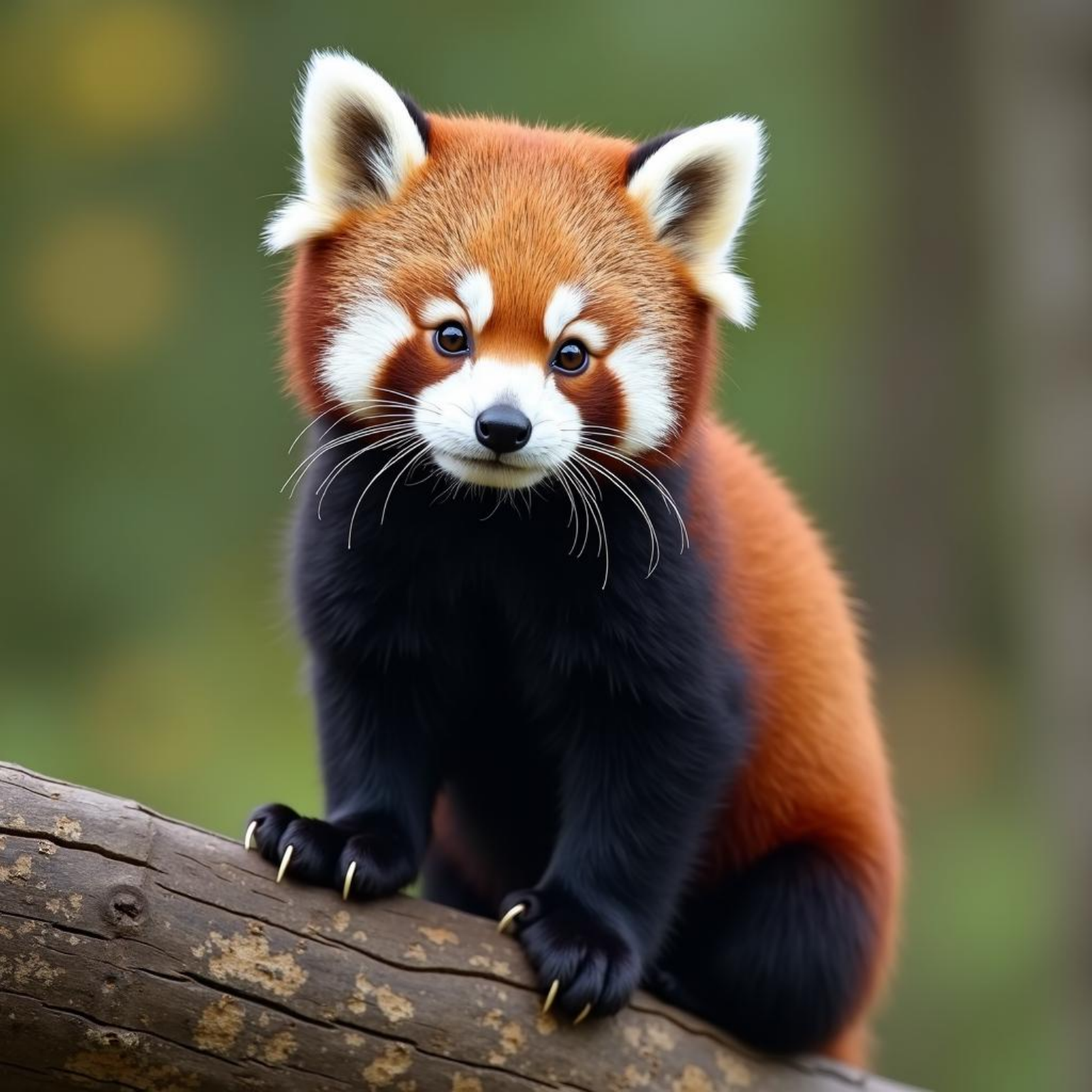} &
        \includegraphics[width=\linewidth]{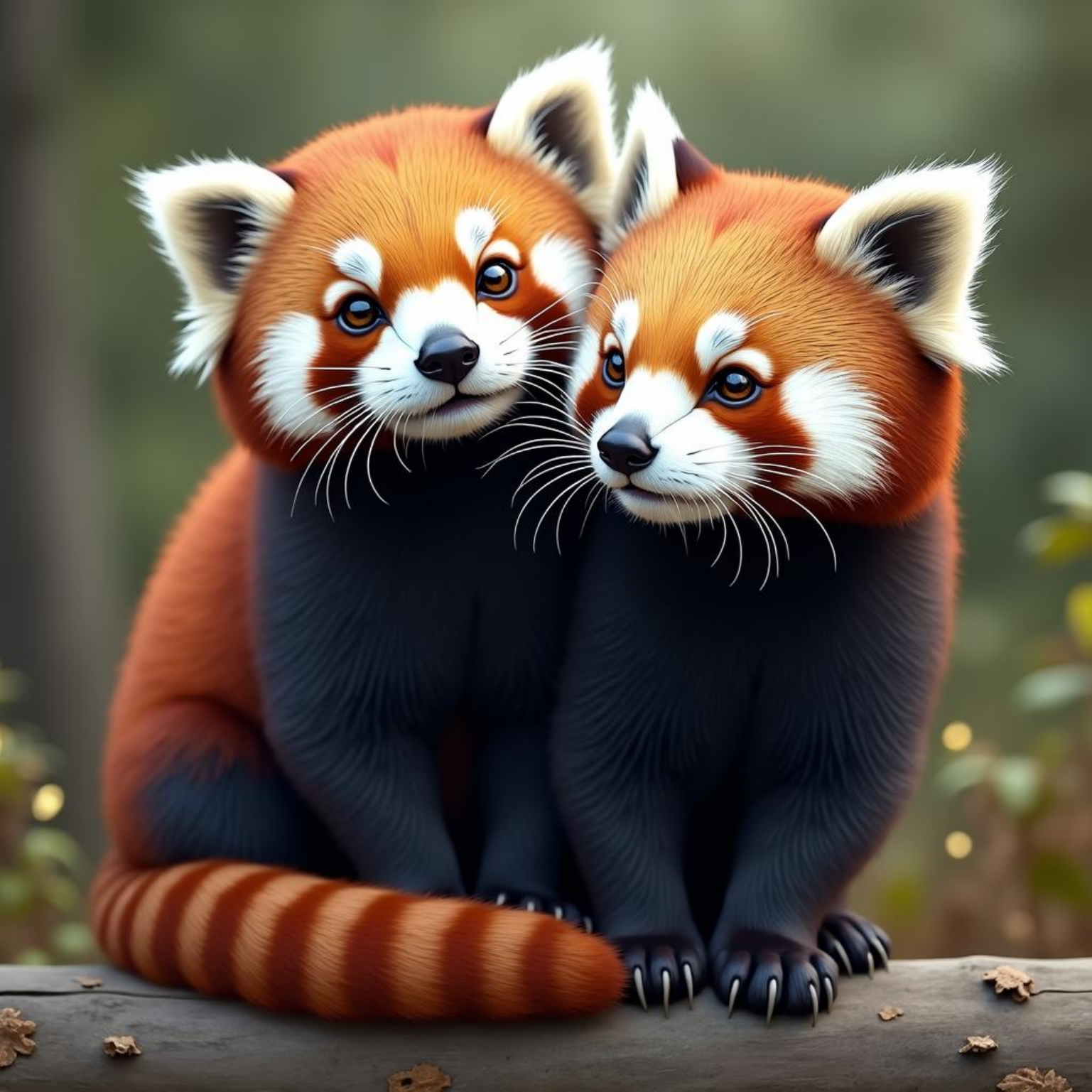} &
        \includegraphics[width=\linewidth]{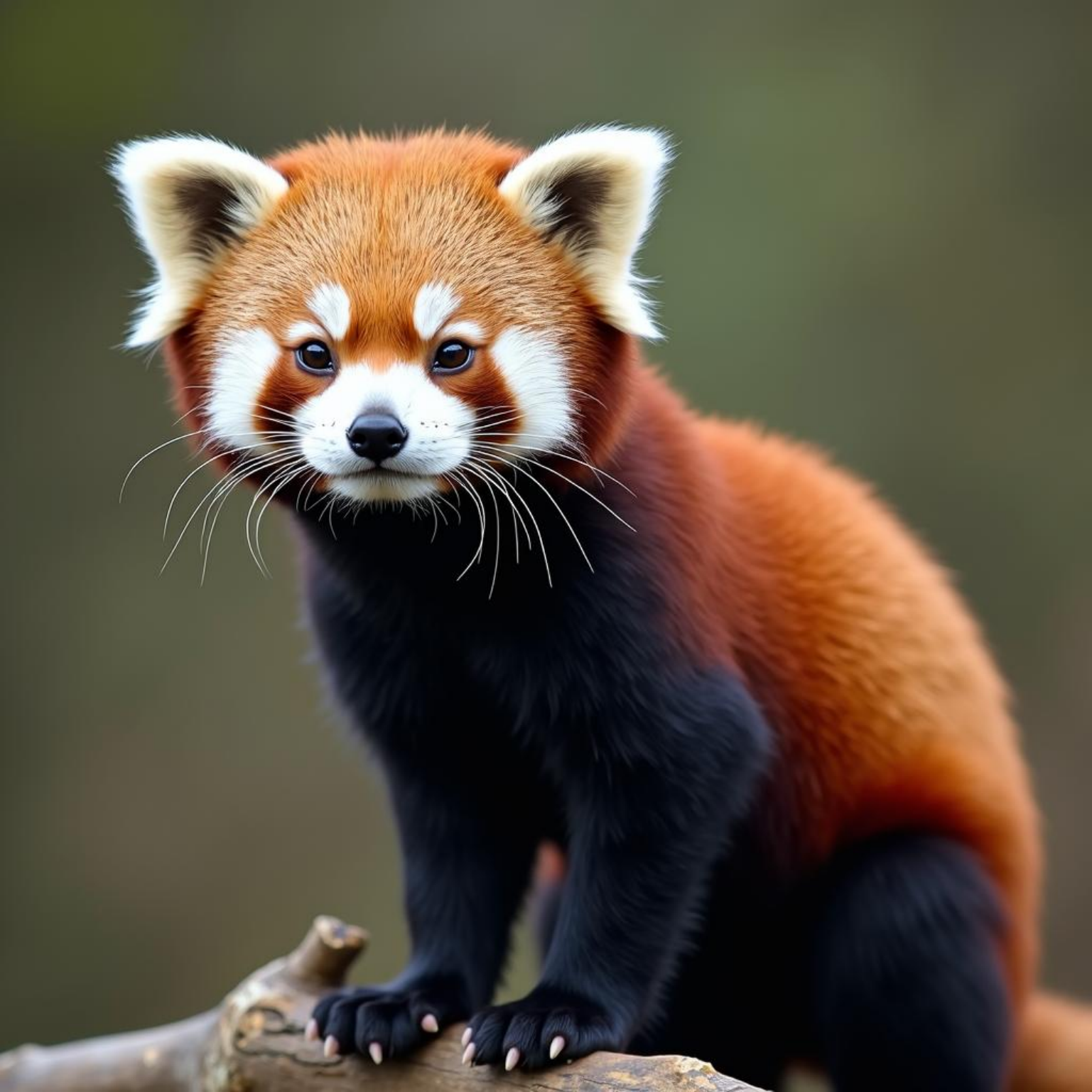} &
        \includegraphics[width=\linewidth]{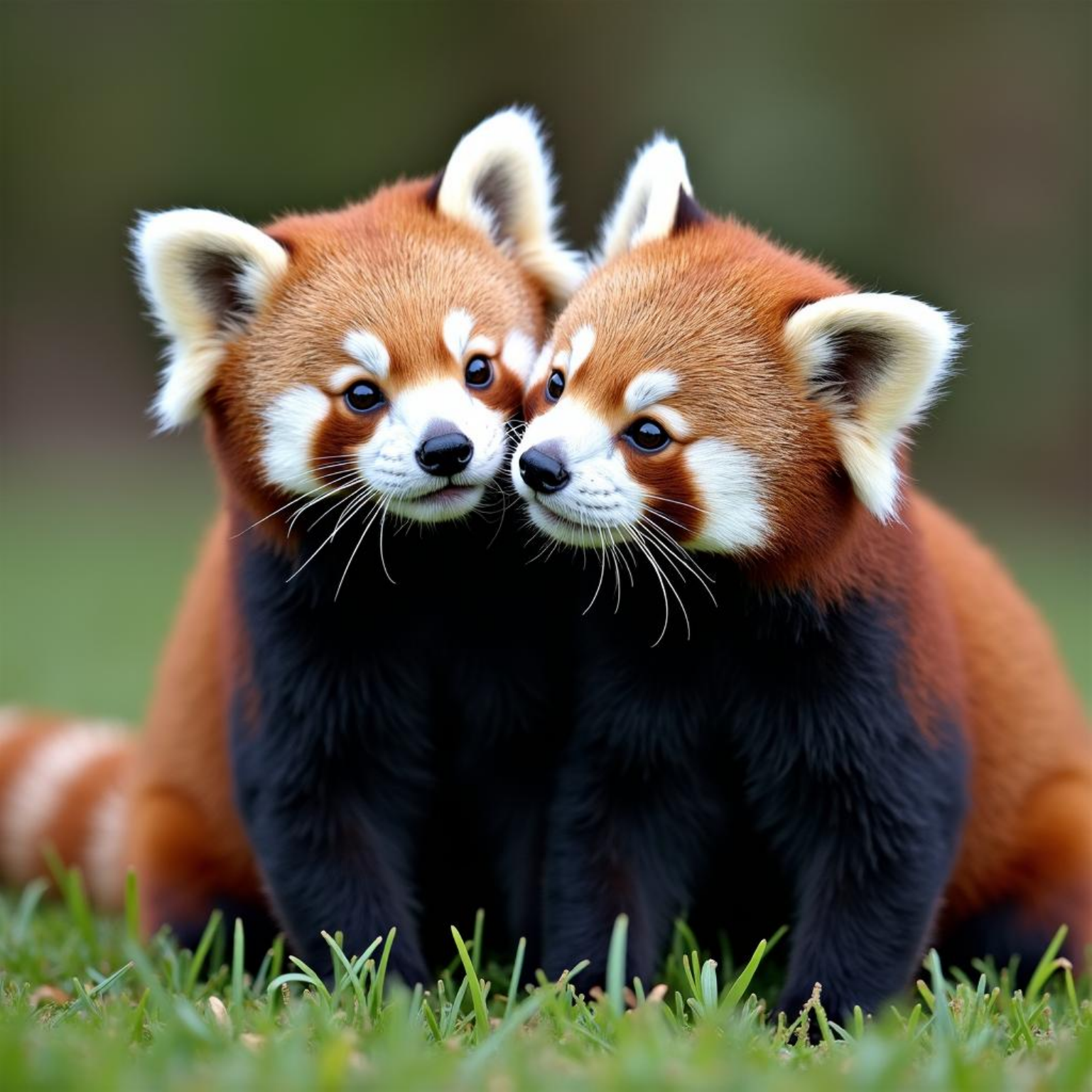} &
        \includegraphics[width=\linewidth]{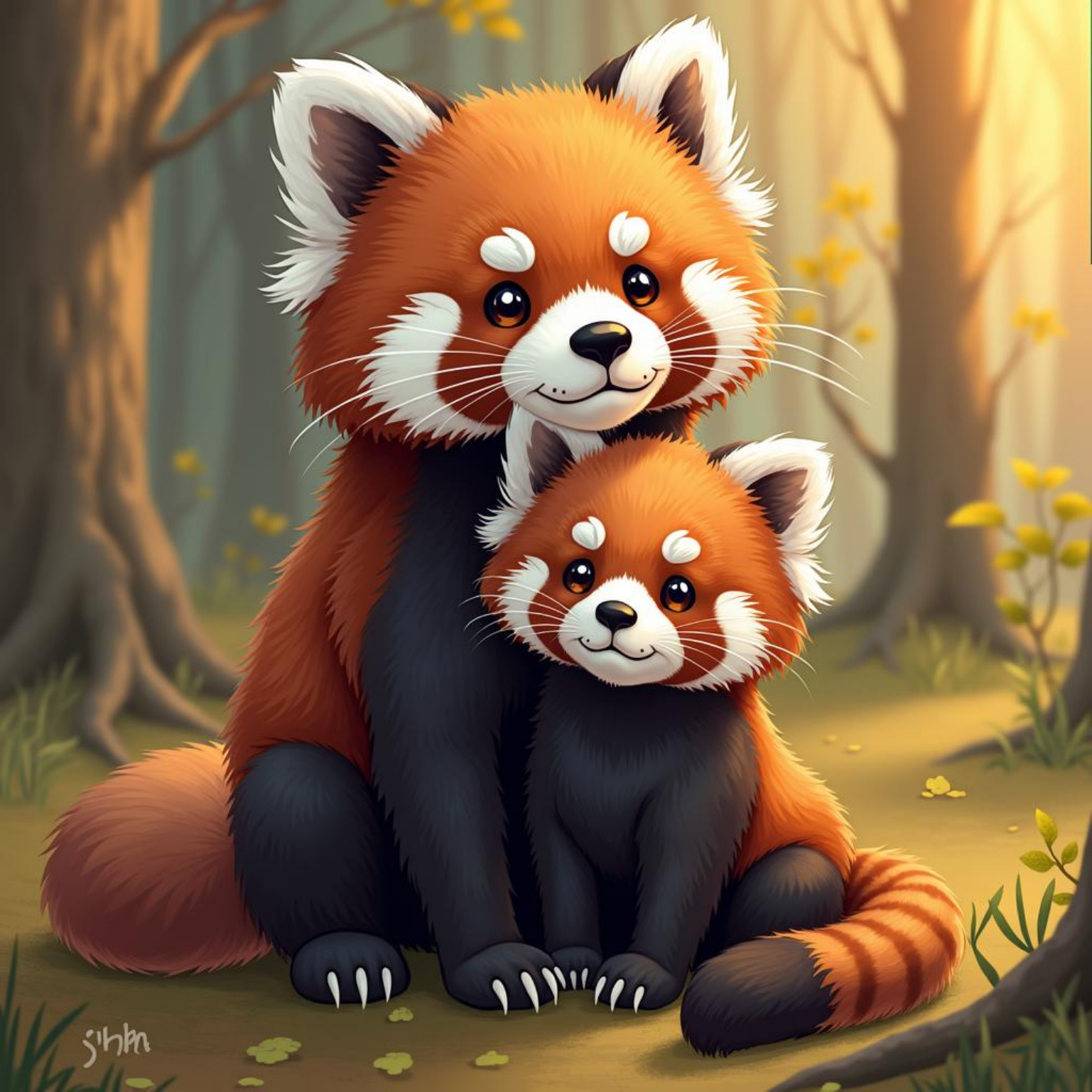} \\

        \multicolumn{5}{>{\centering\arraybackslash}p{\dimexpr\textwidth-2\tabcolsep}}{
            \small \textit{Prompt: Otter and red panda hybrid}
        } \\
        \midrule

        \includegraphics[width=\linewidth]{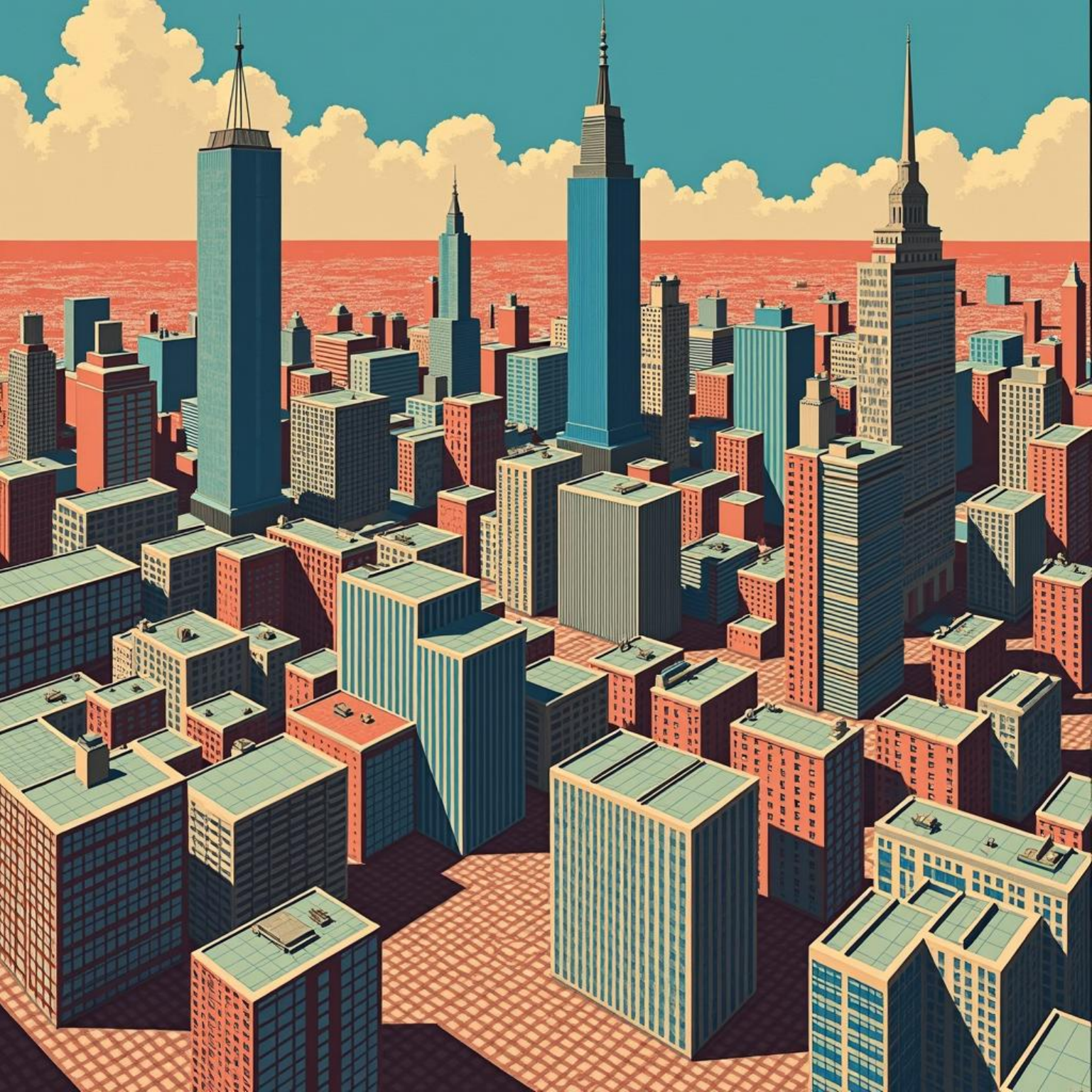} &
        \includegraphics[width=\linewidth]{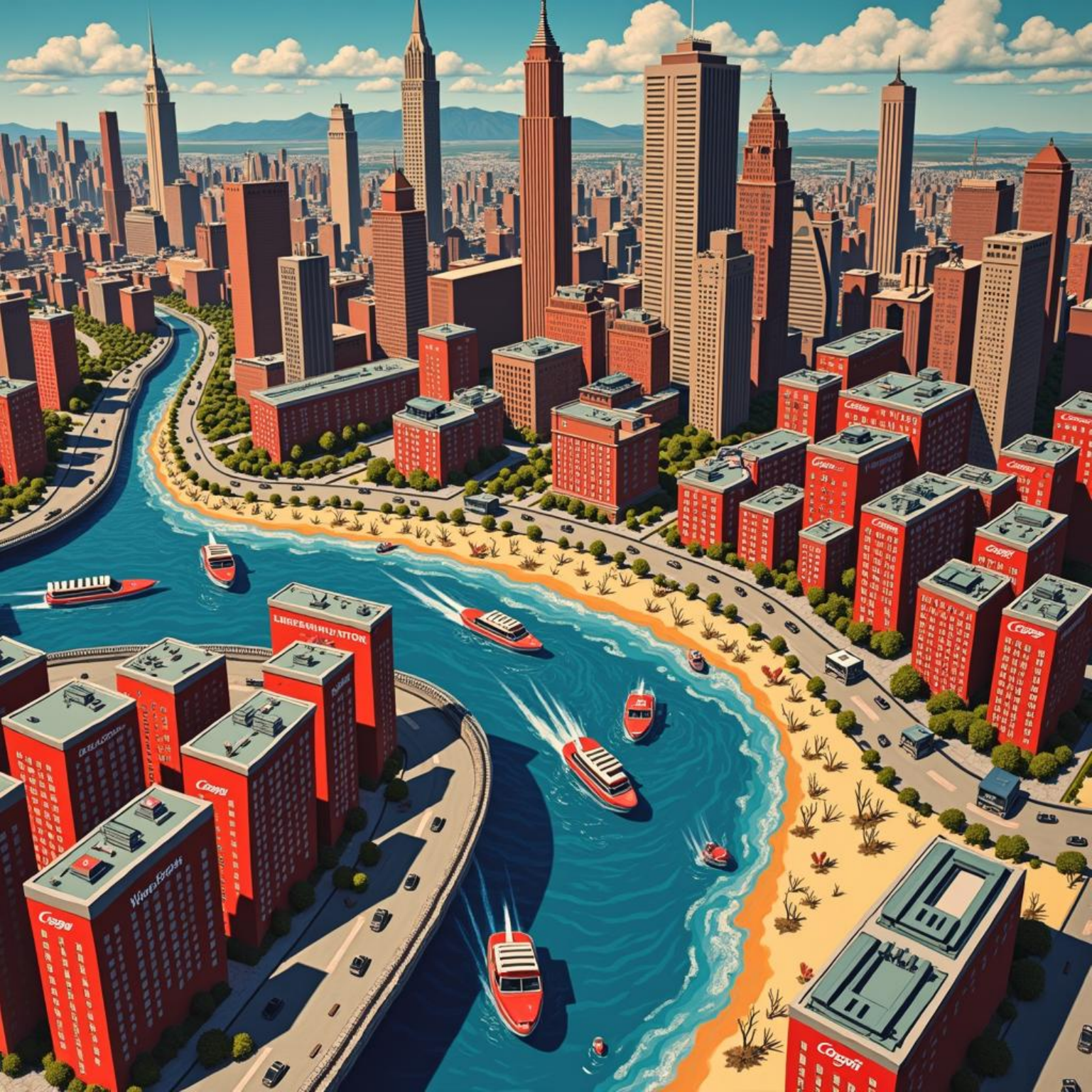} &
        \includegraphics[width=\linewidth]{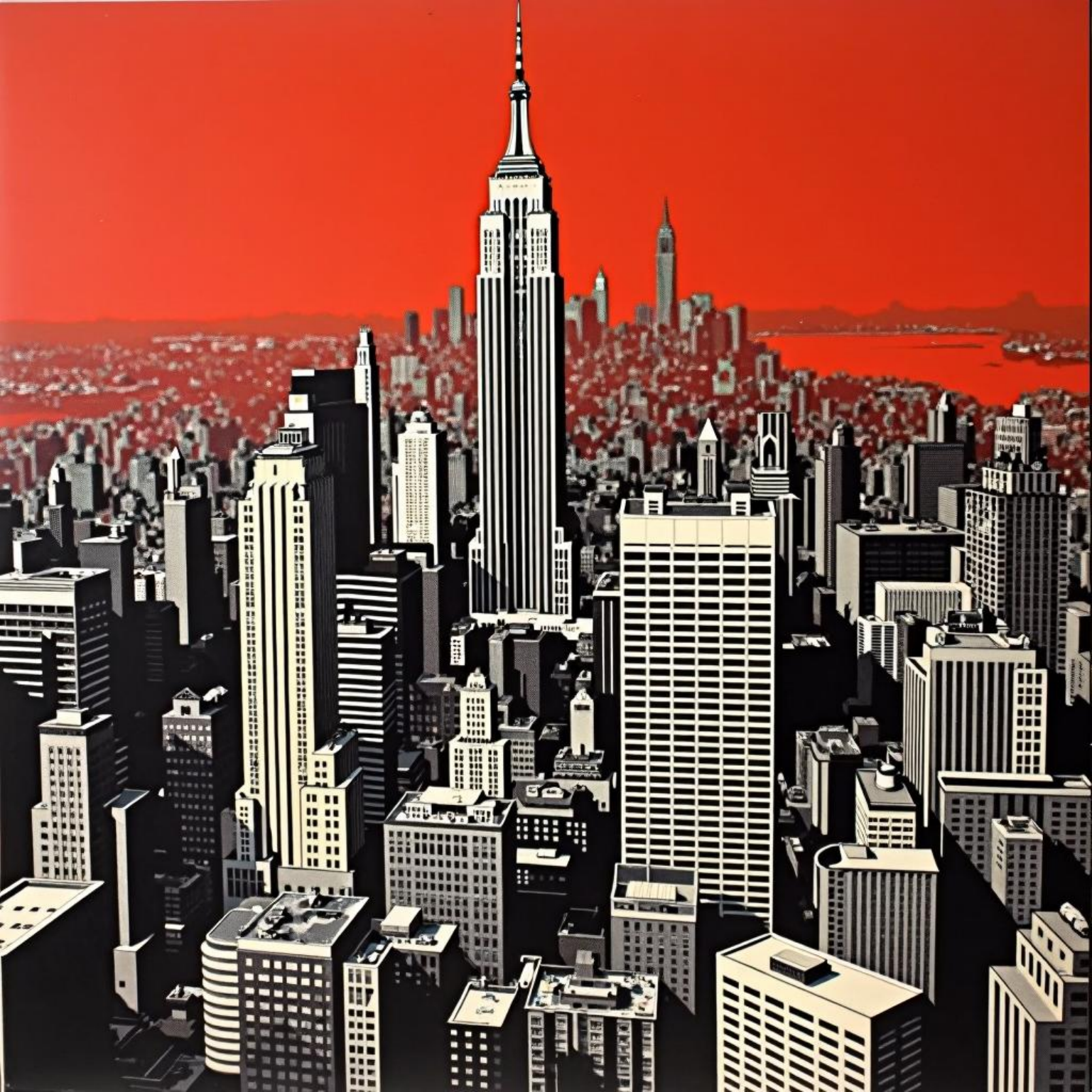} &
        \includegraphics[width=\linewidth]{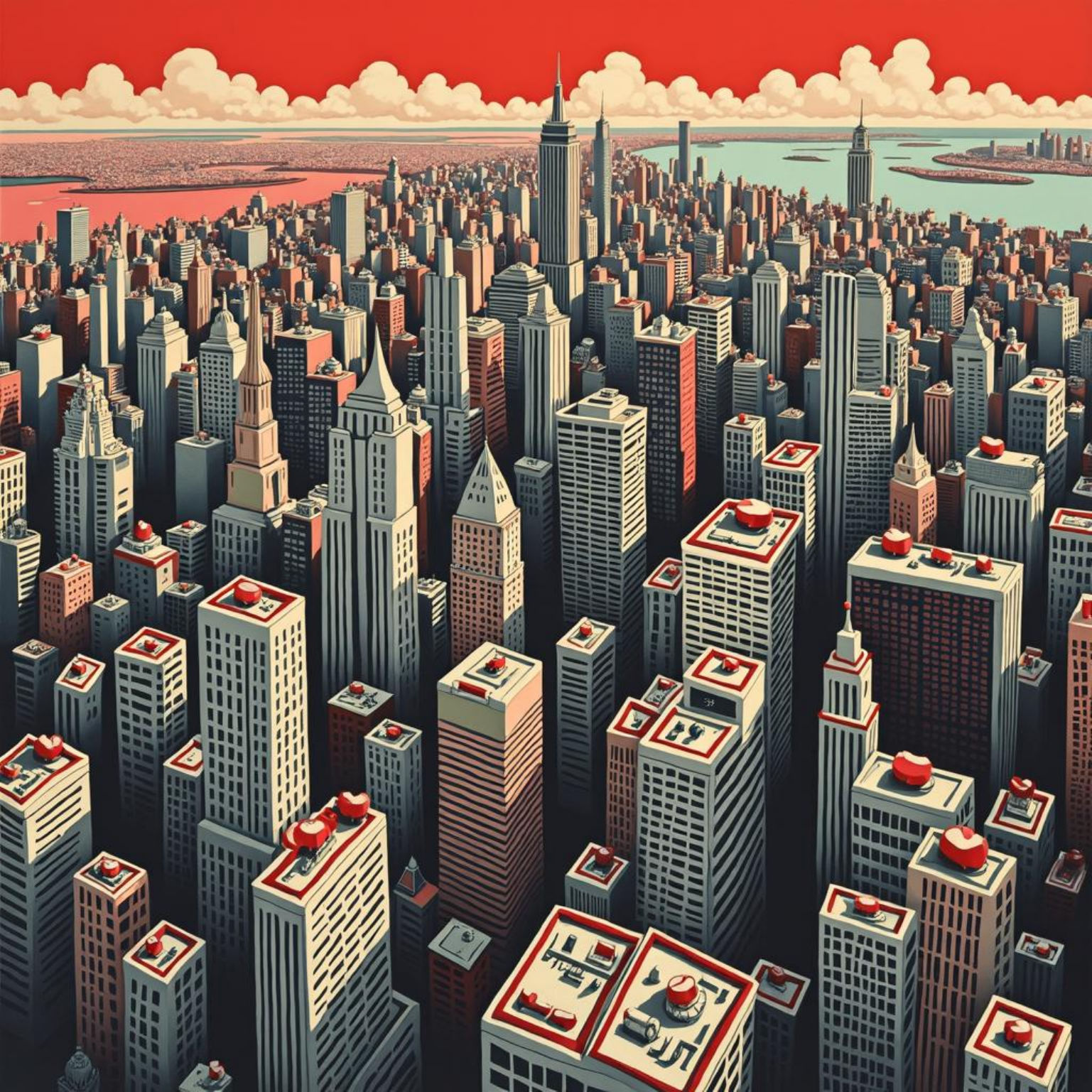} &
        \includegraphics[width=\linewidth]{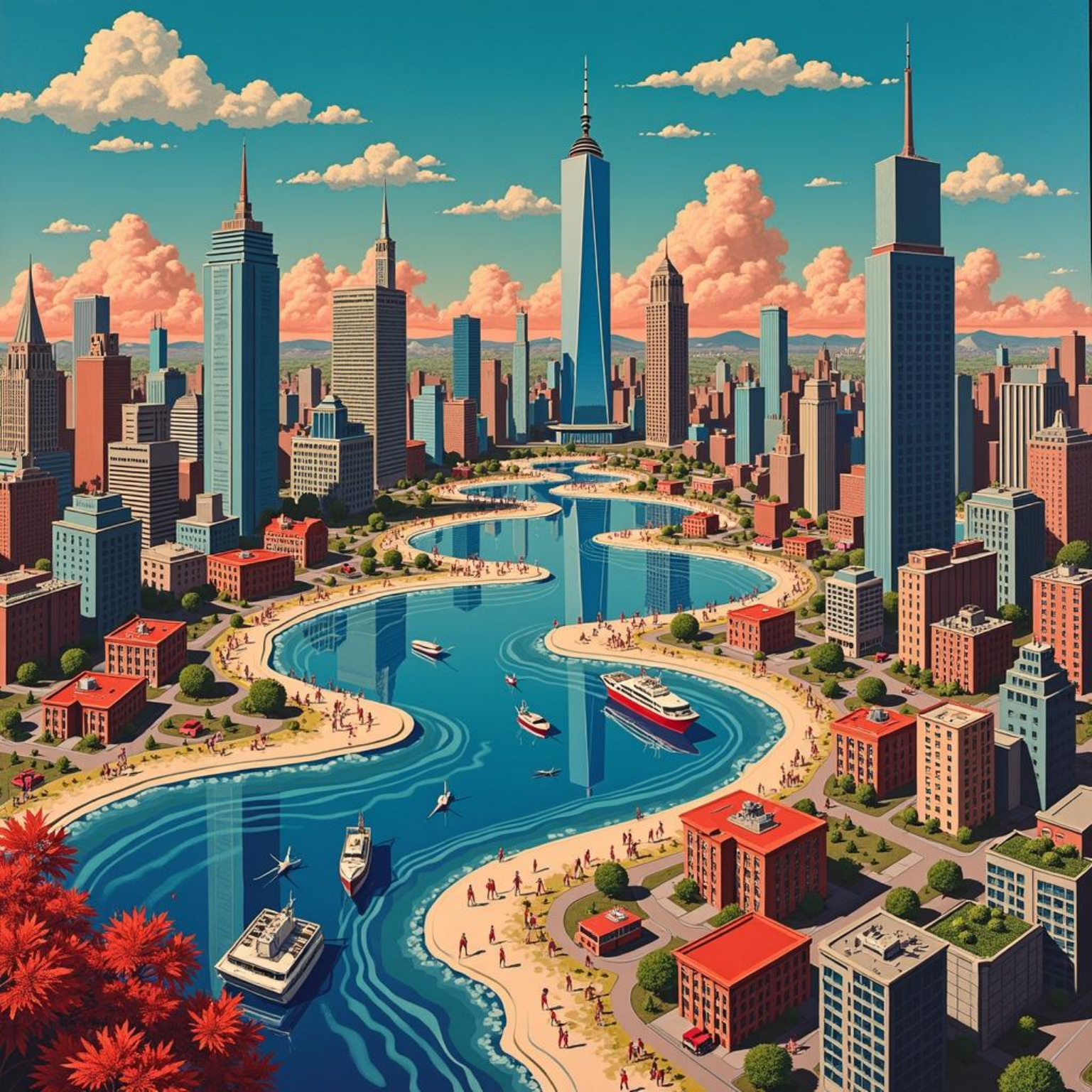} \\

        \multicolumn{5}{>{\centering\arraybackslash}p{\dimexpr\textwidth-2\tabcolsep}}{
            \small \textit{Prompt: in the style of andy warhol, depict Colgate-Palmolive Company's sprawling corporate conglomerate spreading across the world}
        } \\

        \bottomrule
        
    \end{tabularx}
    \label{fig:qualitative_flux} 
    \caption{Visual comparison of FLUX.1-dev with $\text{NRE}=200$, targeting the Aesthetic reward model.} 
\end{figure}

\begin{figure}[t] 
    \centering
    \setlength{\tabcolsep}{1pt} 
    
    \begin{tabularx}{\textwidth}{YYYYY}
        
        \toprule
        \textbf{Qwen-Image} & \textbf{BoN} & \textbf{ZO-N} & \textbf{SoP} & \textbf{SES} \tabularnewline
        \midrule
        
        \includegraphics[width=\linewidth]{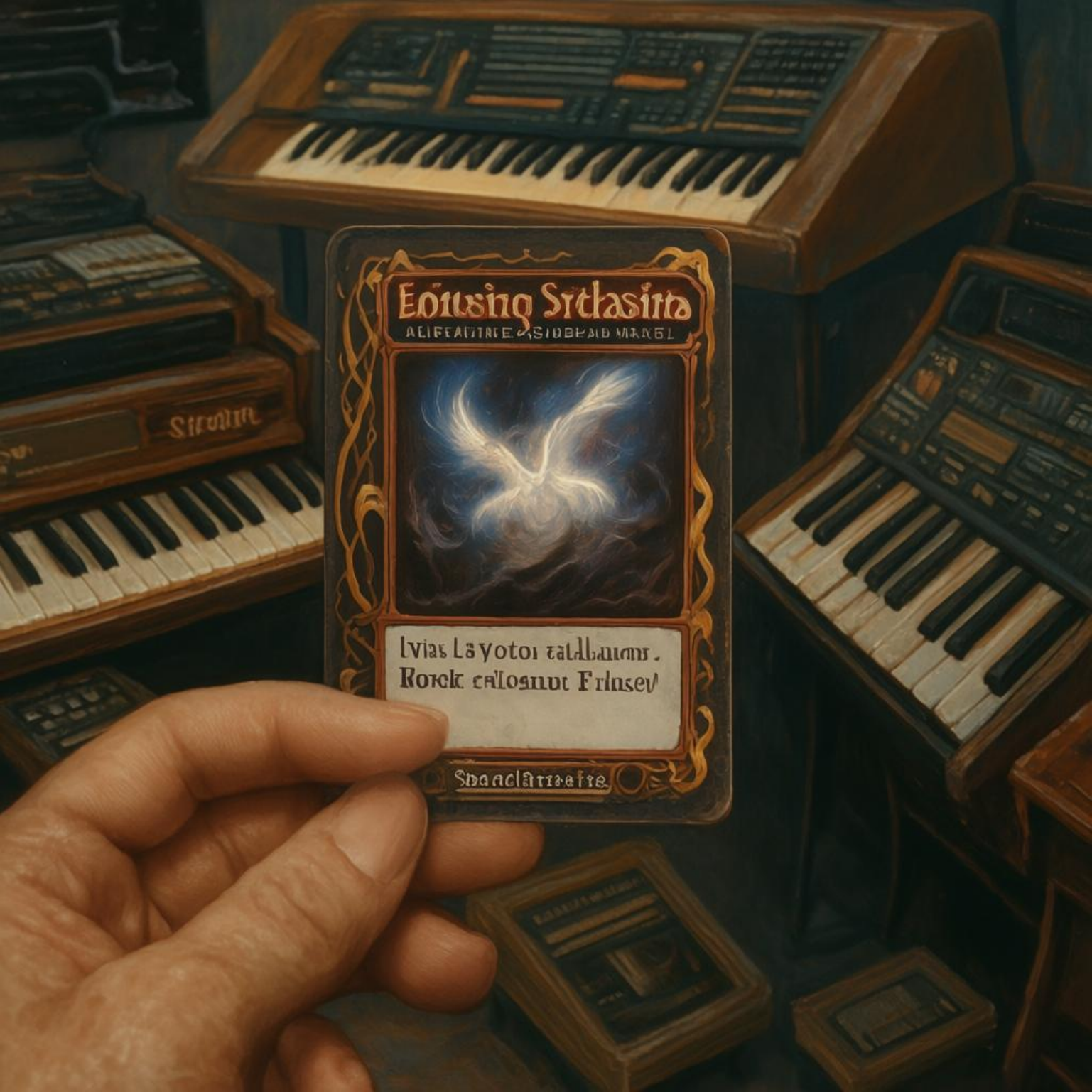} &
        \includegraphics[width=\linewidth]{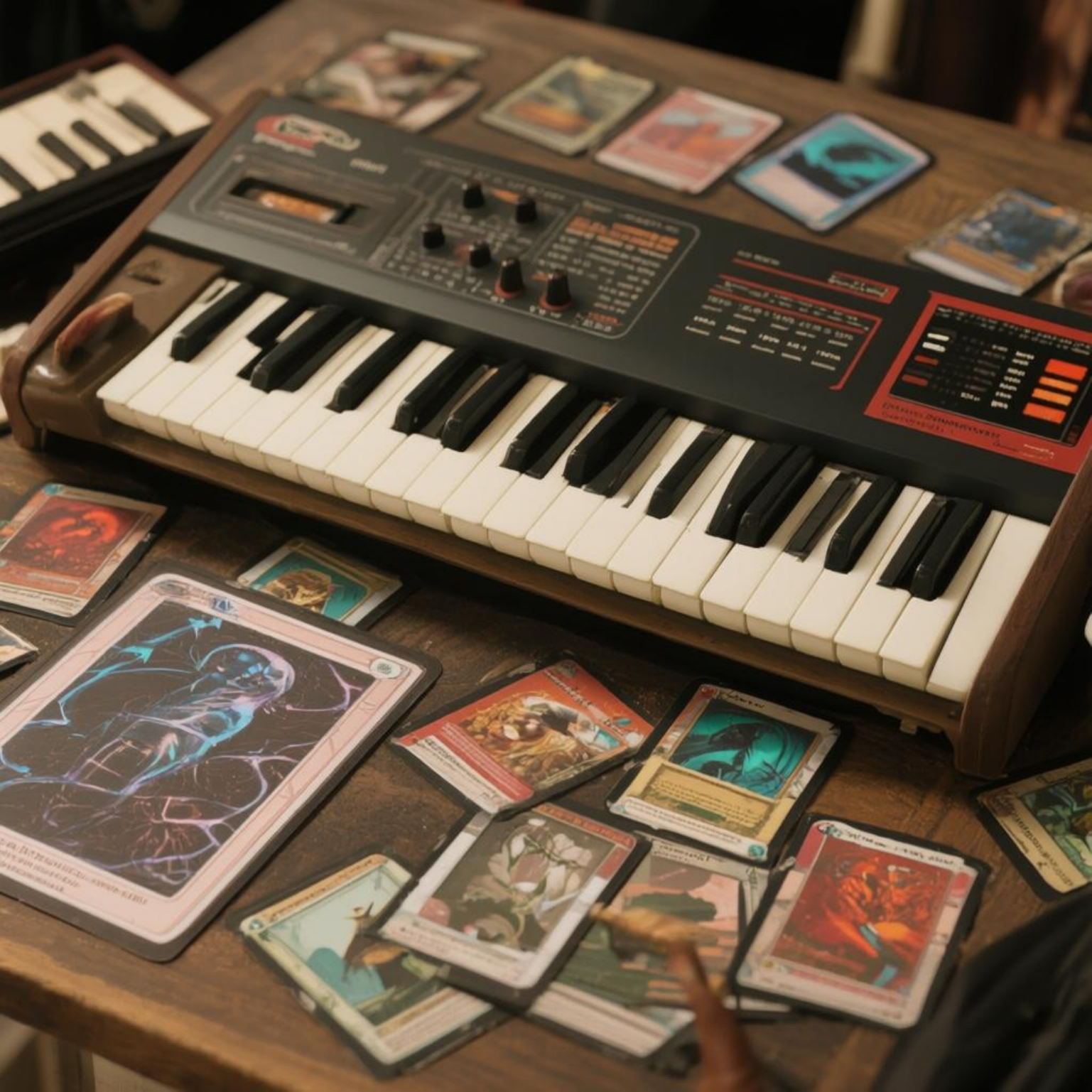} &
        \includegraphics[width=\linewidth]{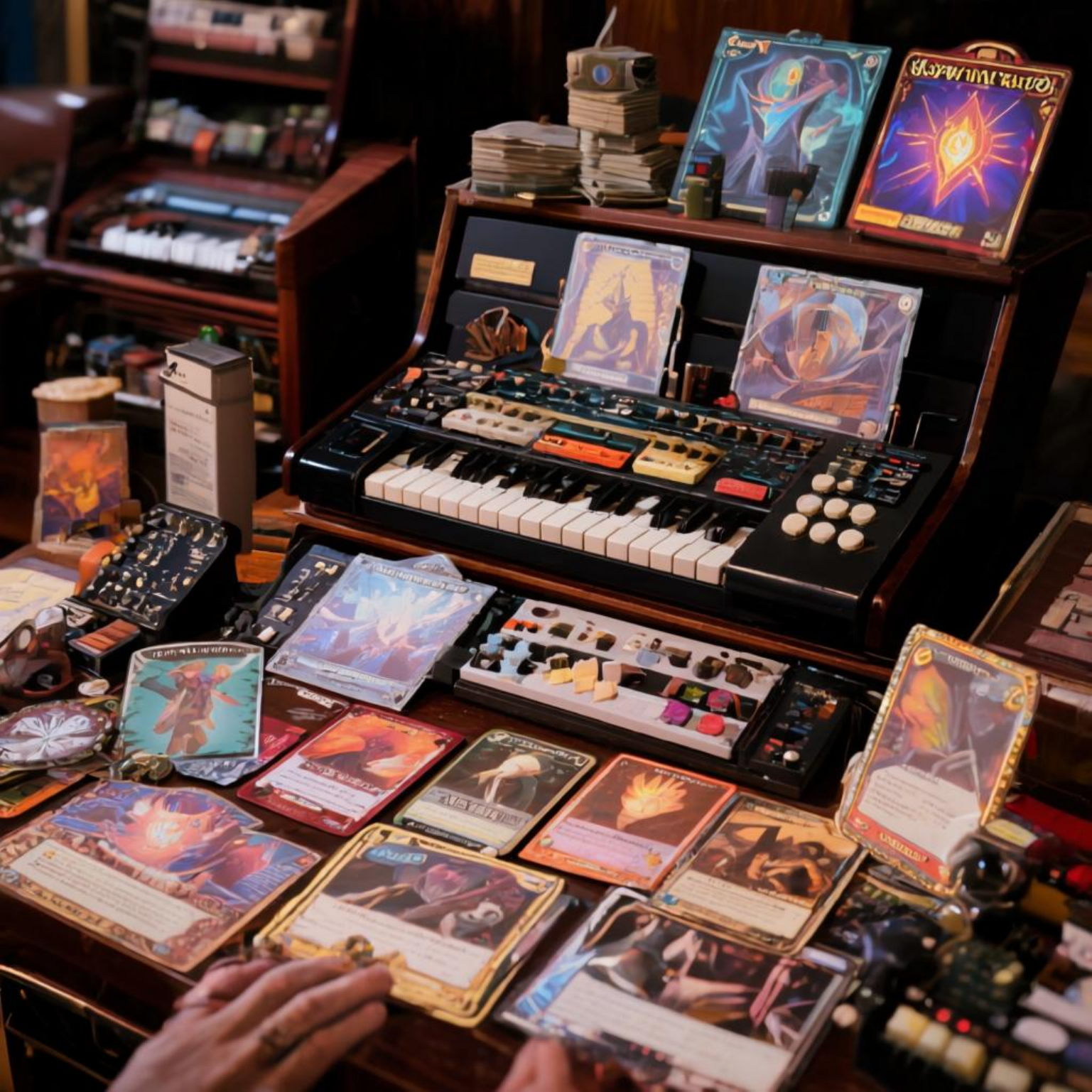} &
        \includegraphics[width=\linewidth]{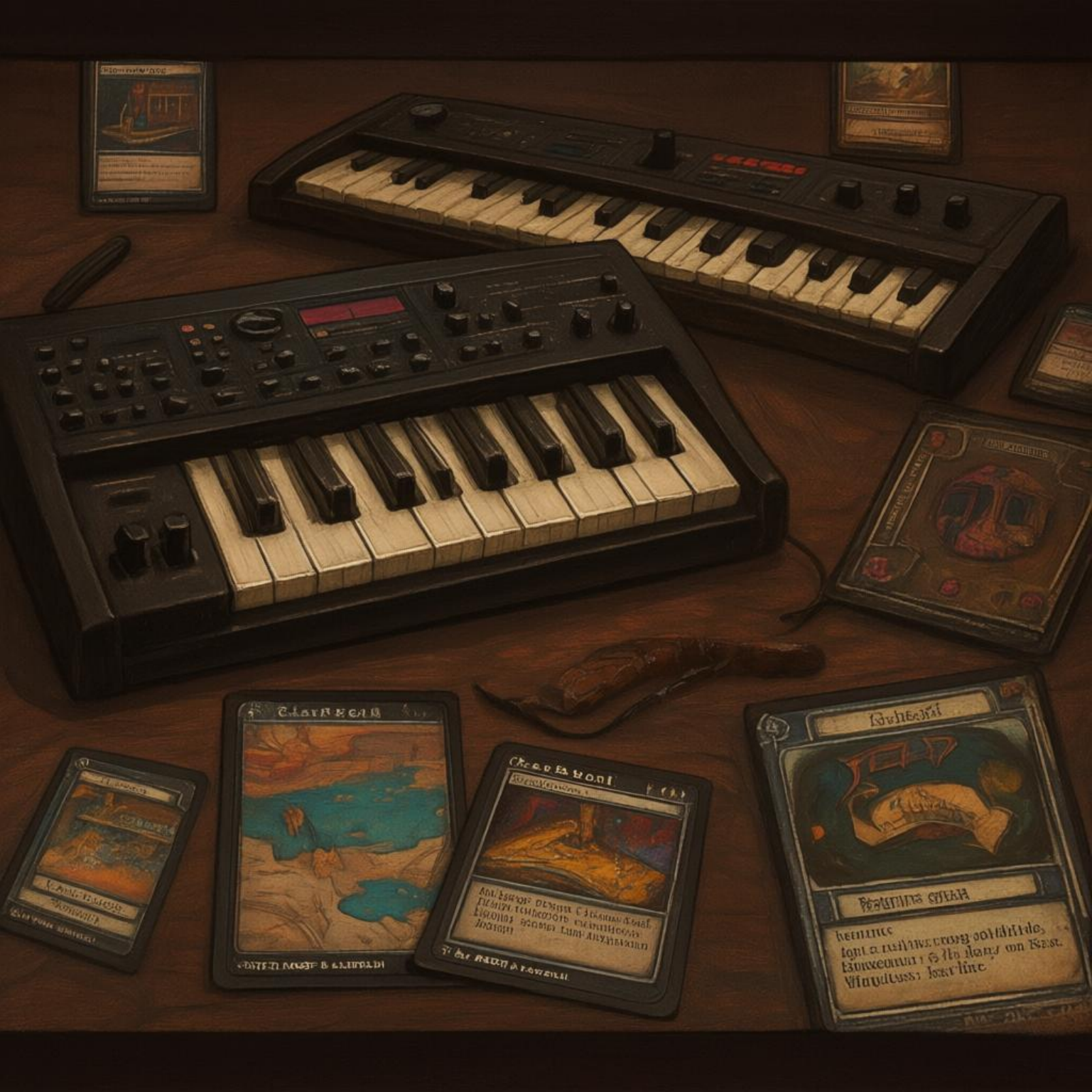} &
        \includegraphics[width=\linewidth]{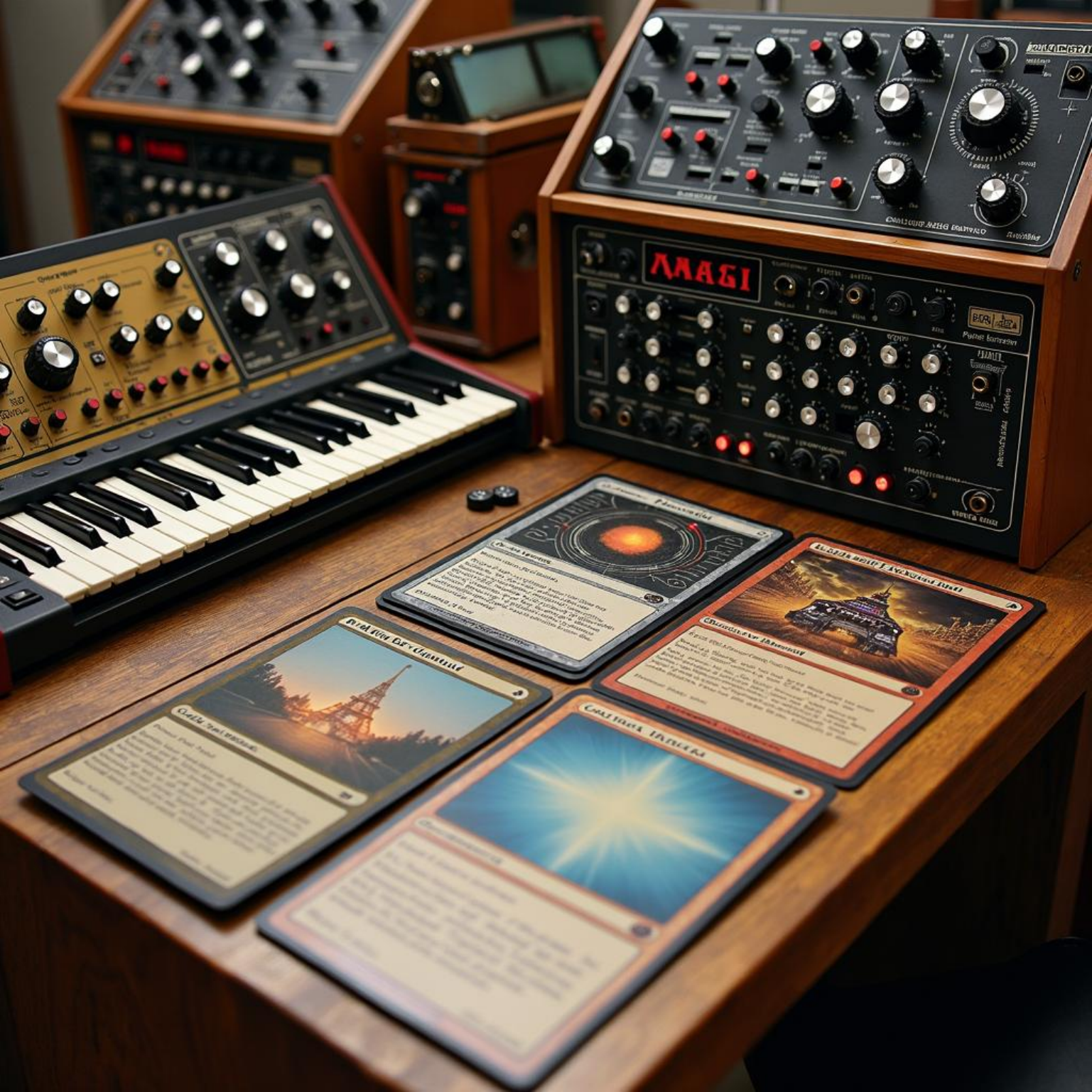} \\

        \multicolumn{5}{>{\centering\arraybackslash}p{\dimexpr\textwidth-2\tabcolsep}}{
            \small \textit{Prompt: magic the gathering cards of vintage analogue synthesizers}
        } \\
        \midrule

        \includegraphics[width=\linewidth]{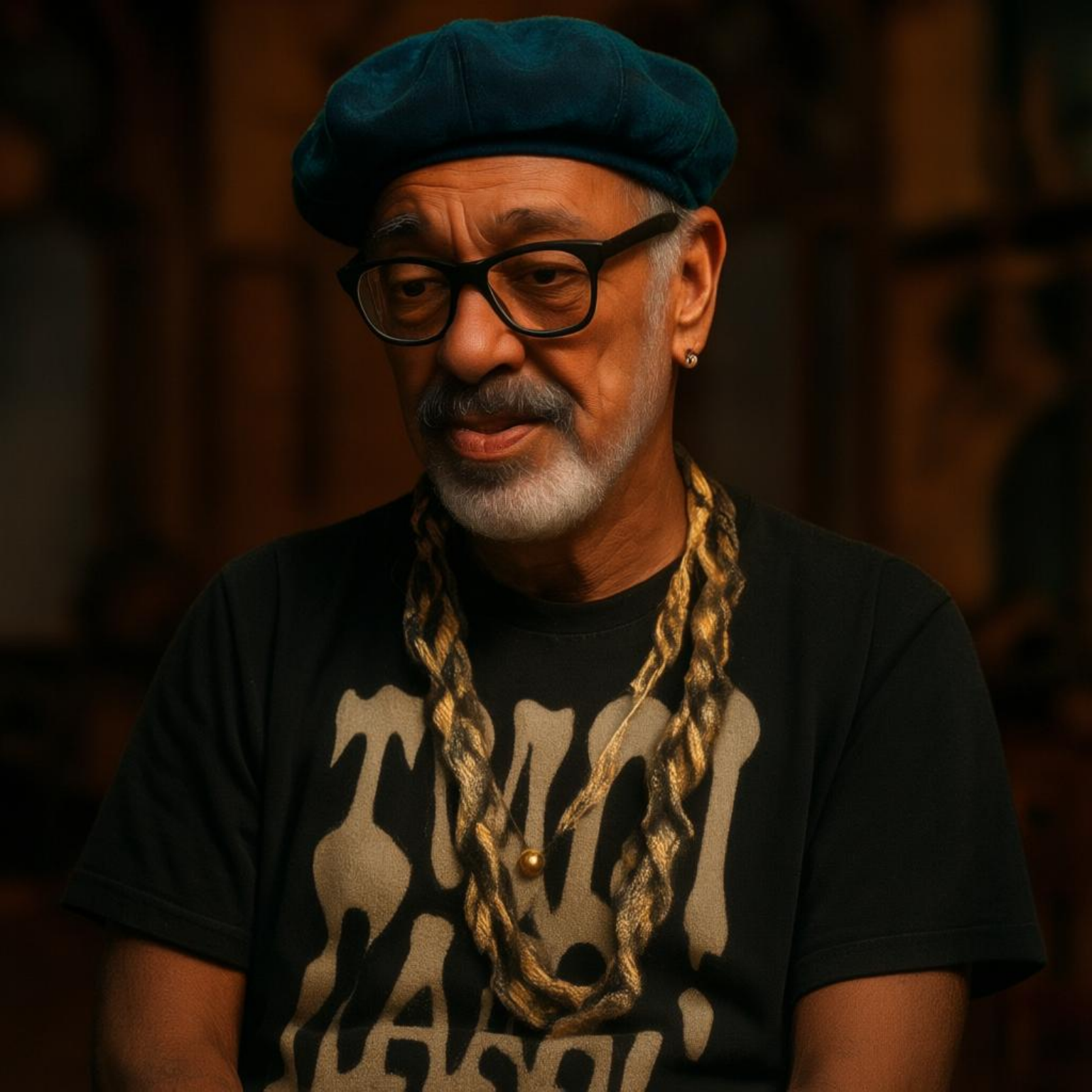} &
        \includegraphics[width=\linewidth]{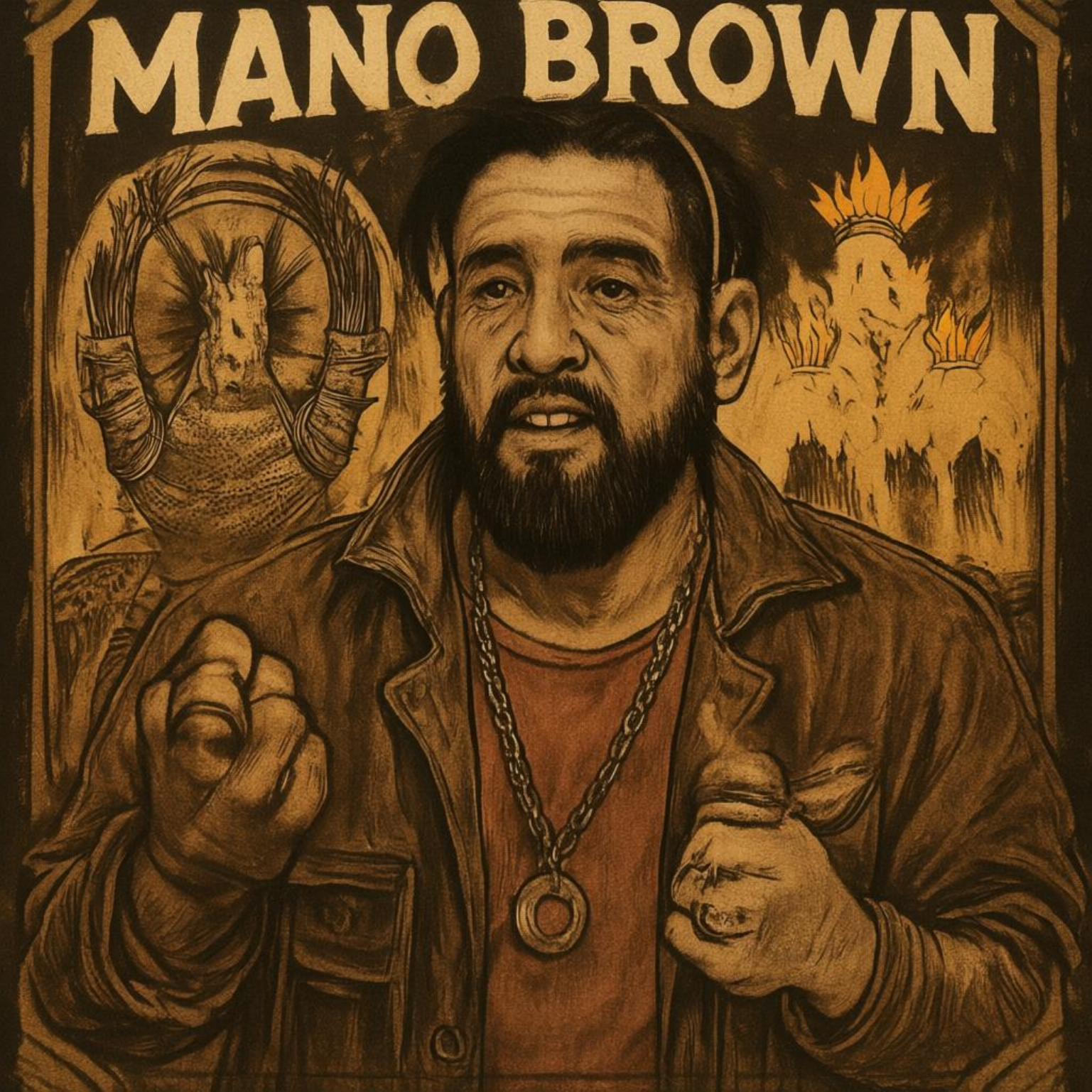} &
        \includegraphics[width=\linewidth]{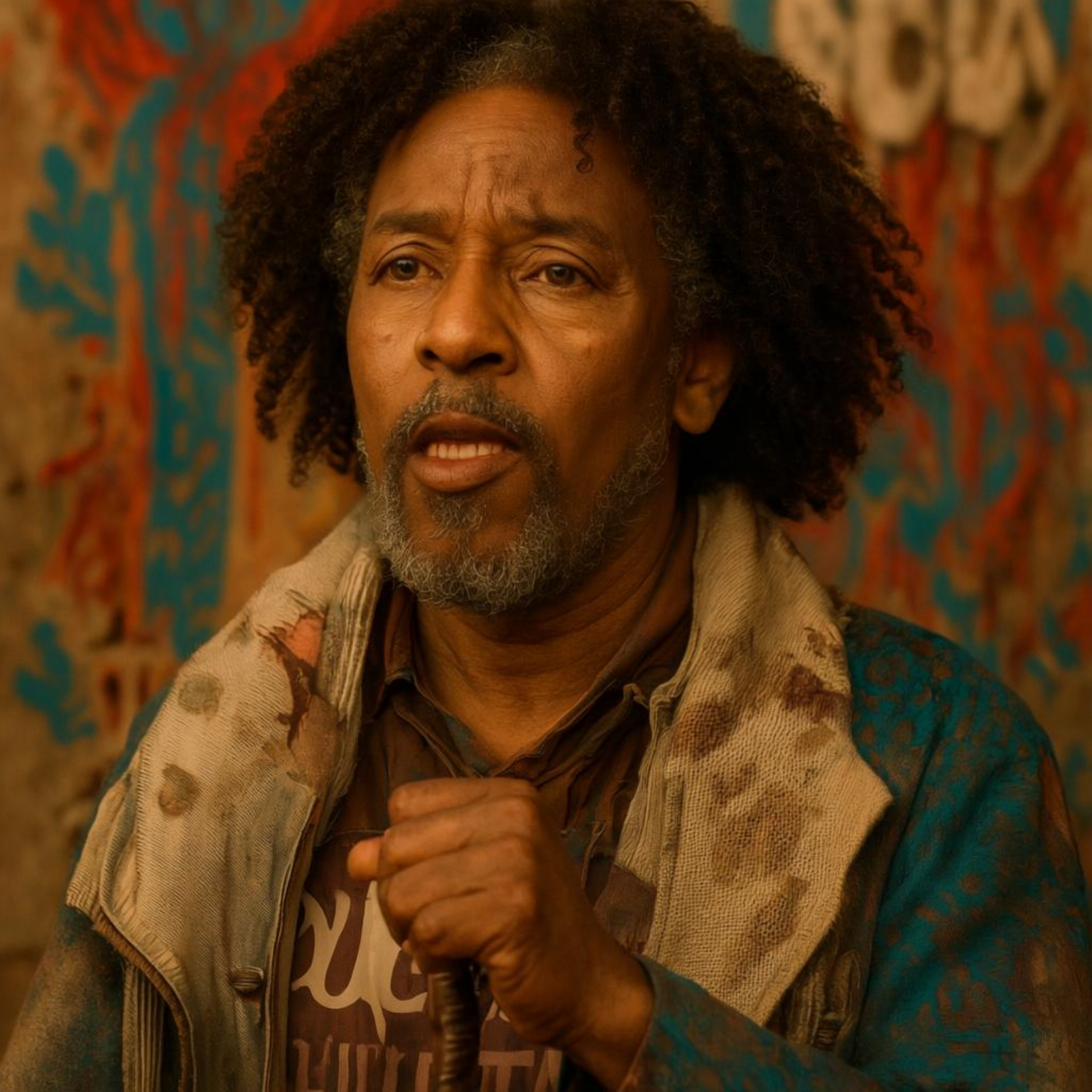} &
        \includegraphics[width=\linewidth]{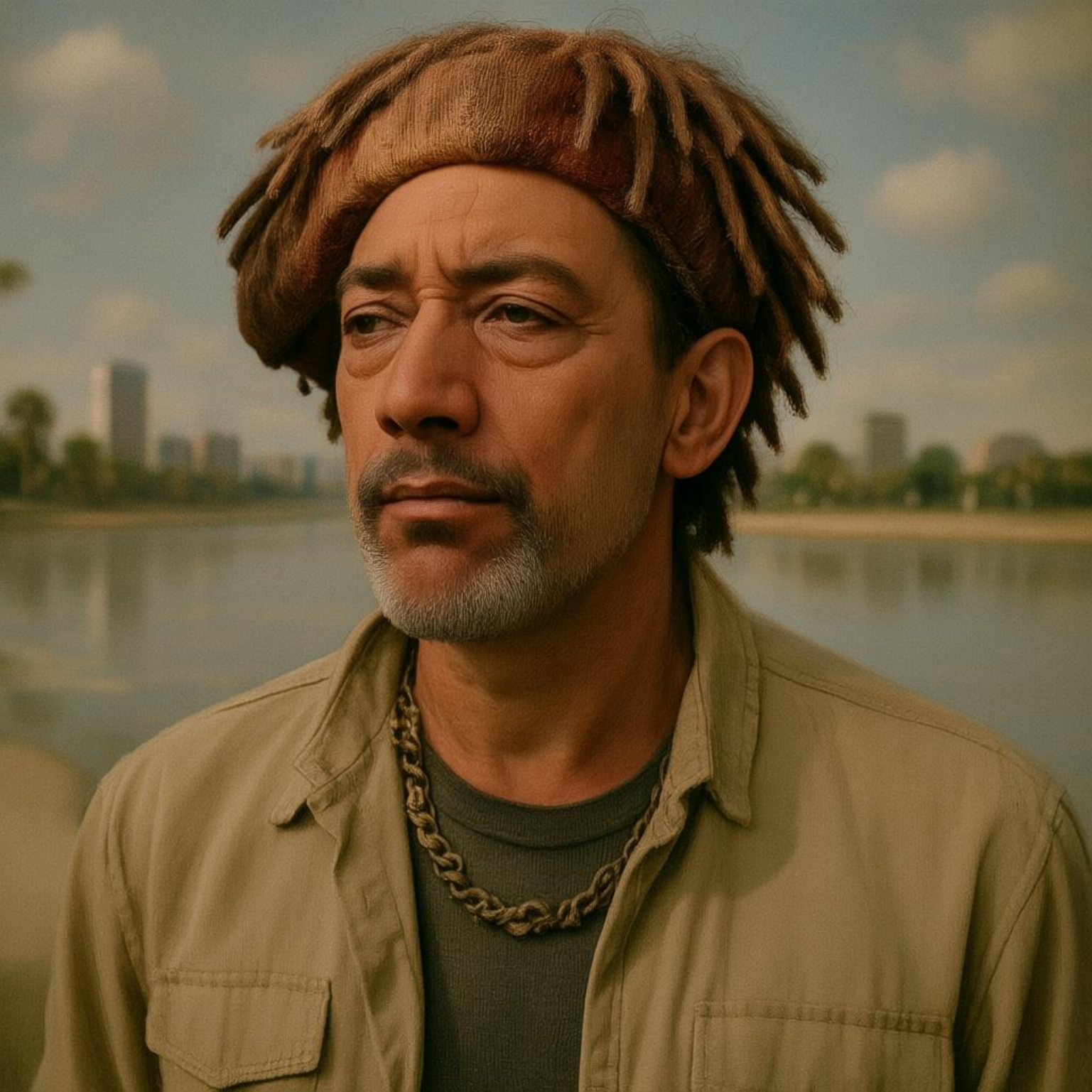} &
        \includegraphics[width=\linewidth]{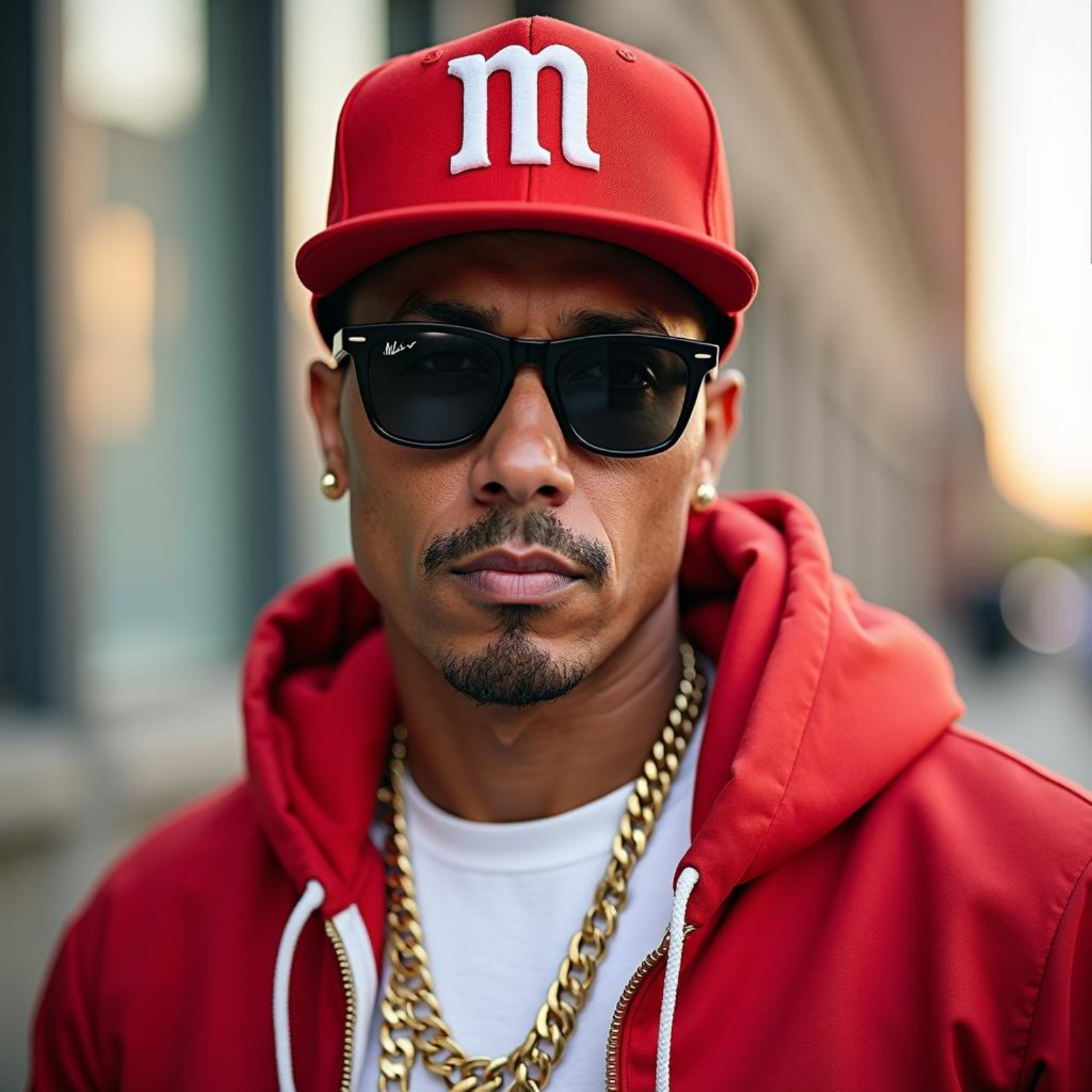} \\

        \multicolumn{5}{>{\centering\arraybackslash}p{\dimexpr\textwidth-2\tabcolsep}}{
            \small \textit{Prompt: Rapper Mano Brown}
        } \\
        \midrule

        \includegraphics[width=\linewidth]{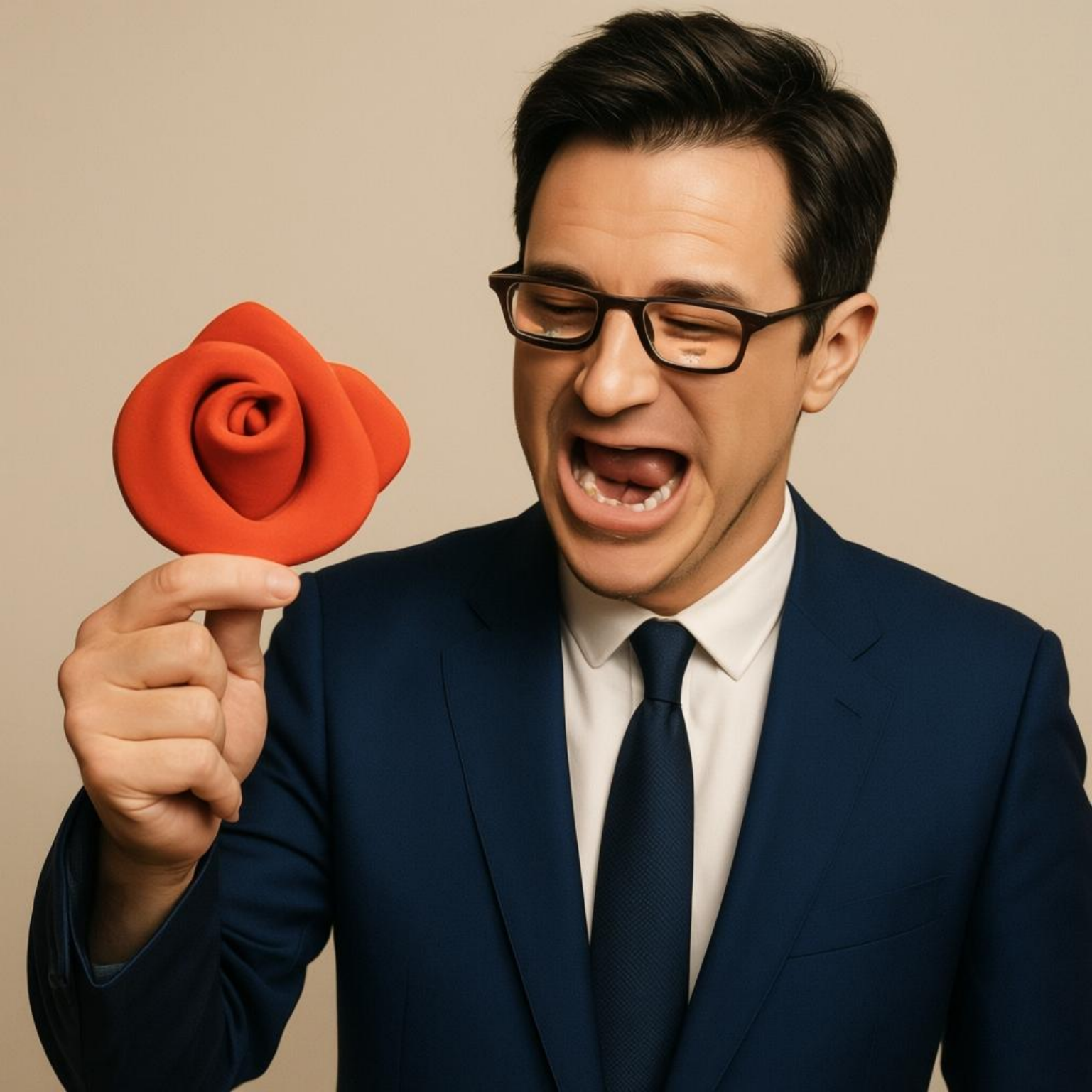} &
        \includegraphics[width=\linewidth]{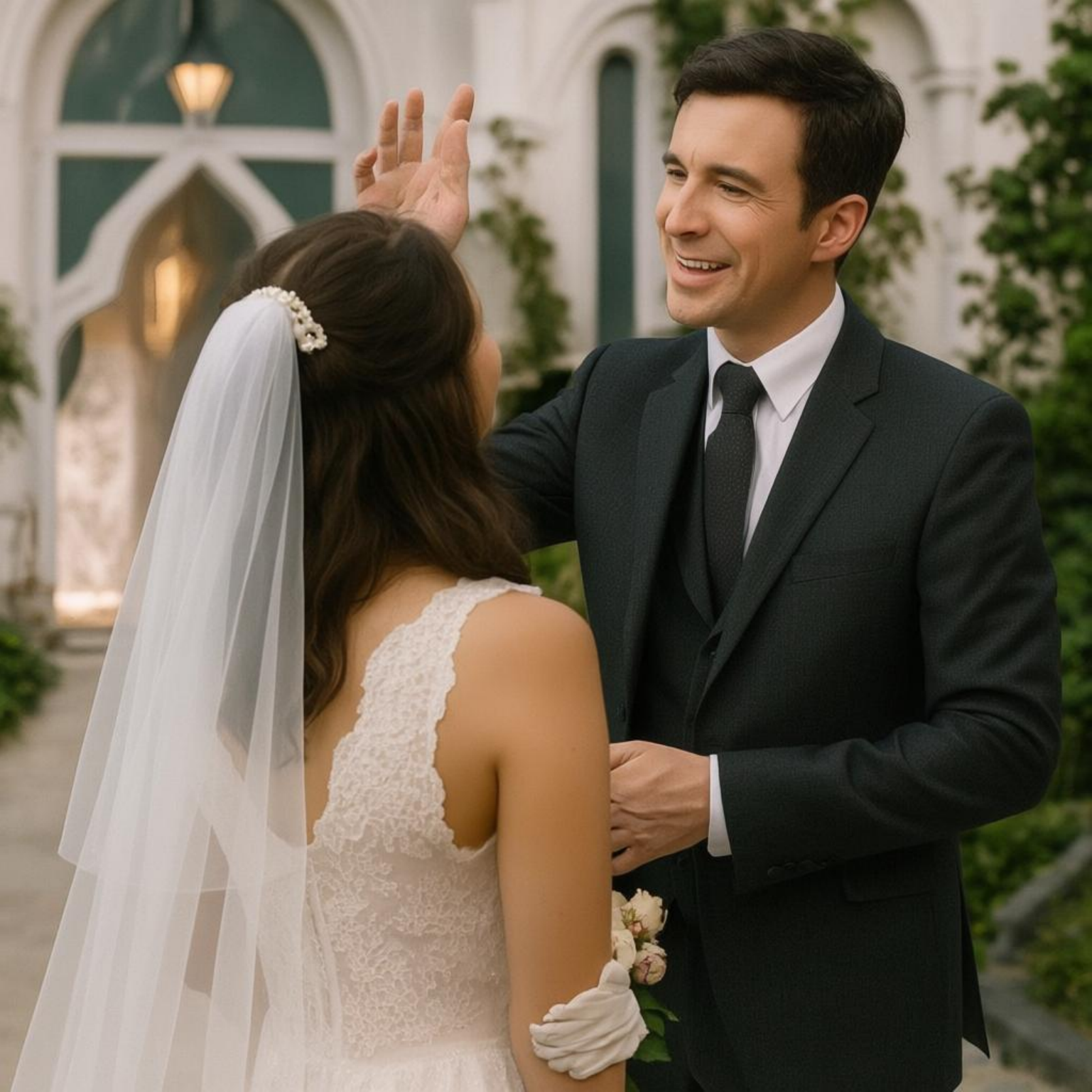} &
        \includegraphics[width=\linewidth]{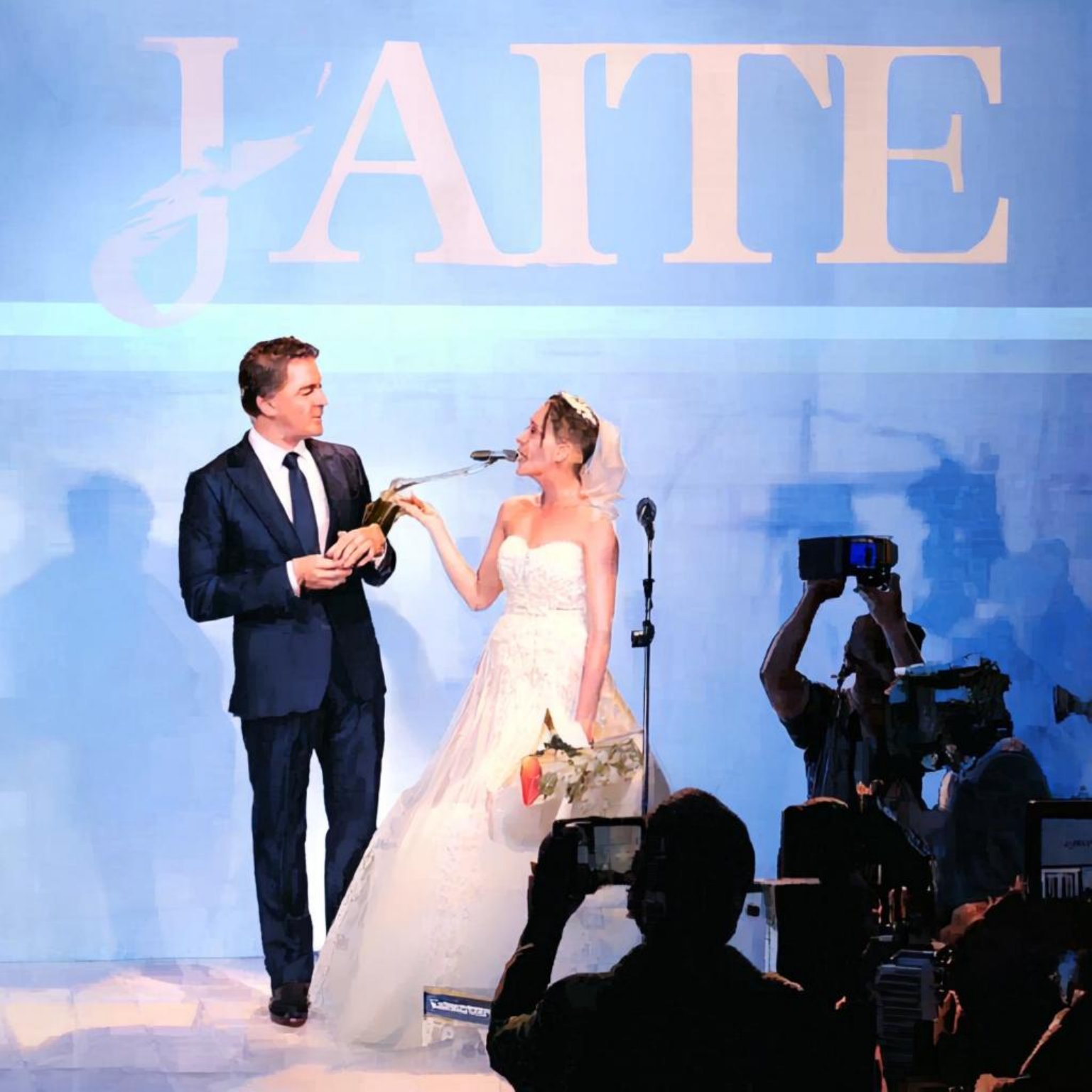} &
        \includegraphics[width=\linewidth]{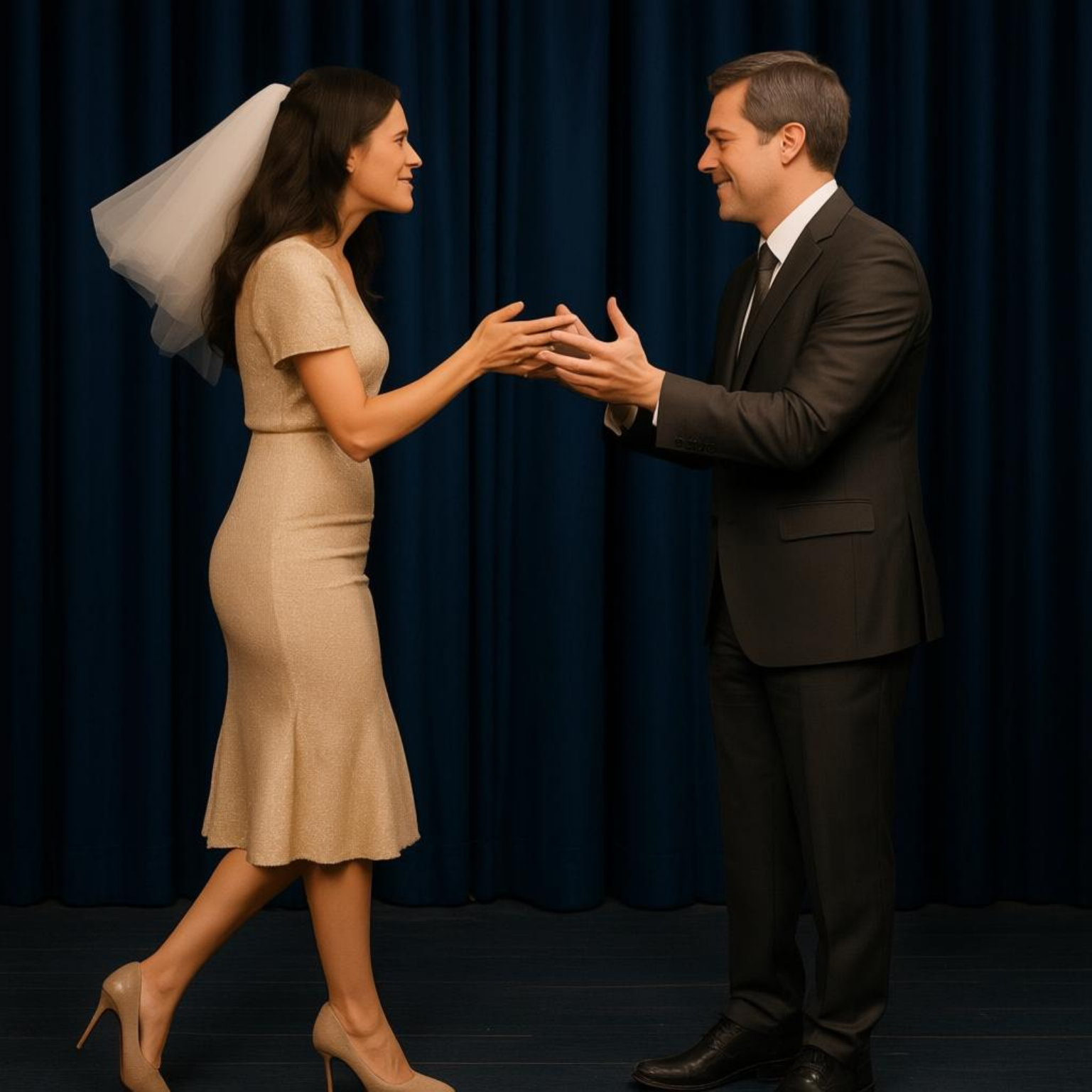} &
        \includegraphics[width=\linewidth]{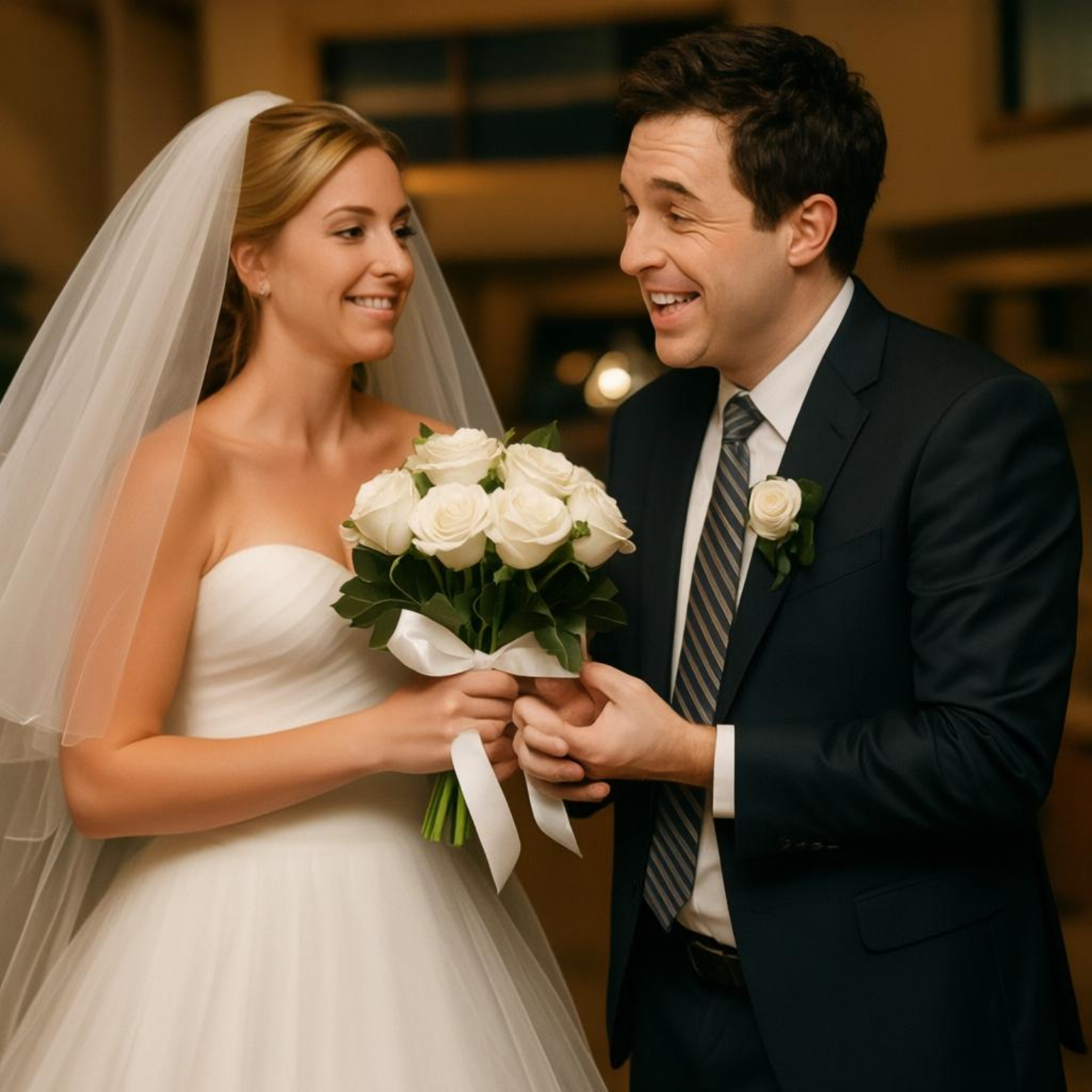} \\

        \multicolumn{5}{>{\centering\arraybackslash}p{\dimexpr\textwidth-2\tabcolsep}}{
            \small \textit{Prompt: John Oliver proposing marriage}
        } \\
        \midrule

        \includegraphics[width=\linewidth]{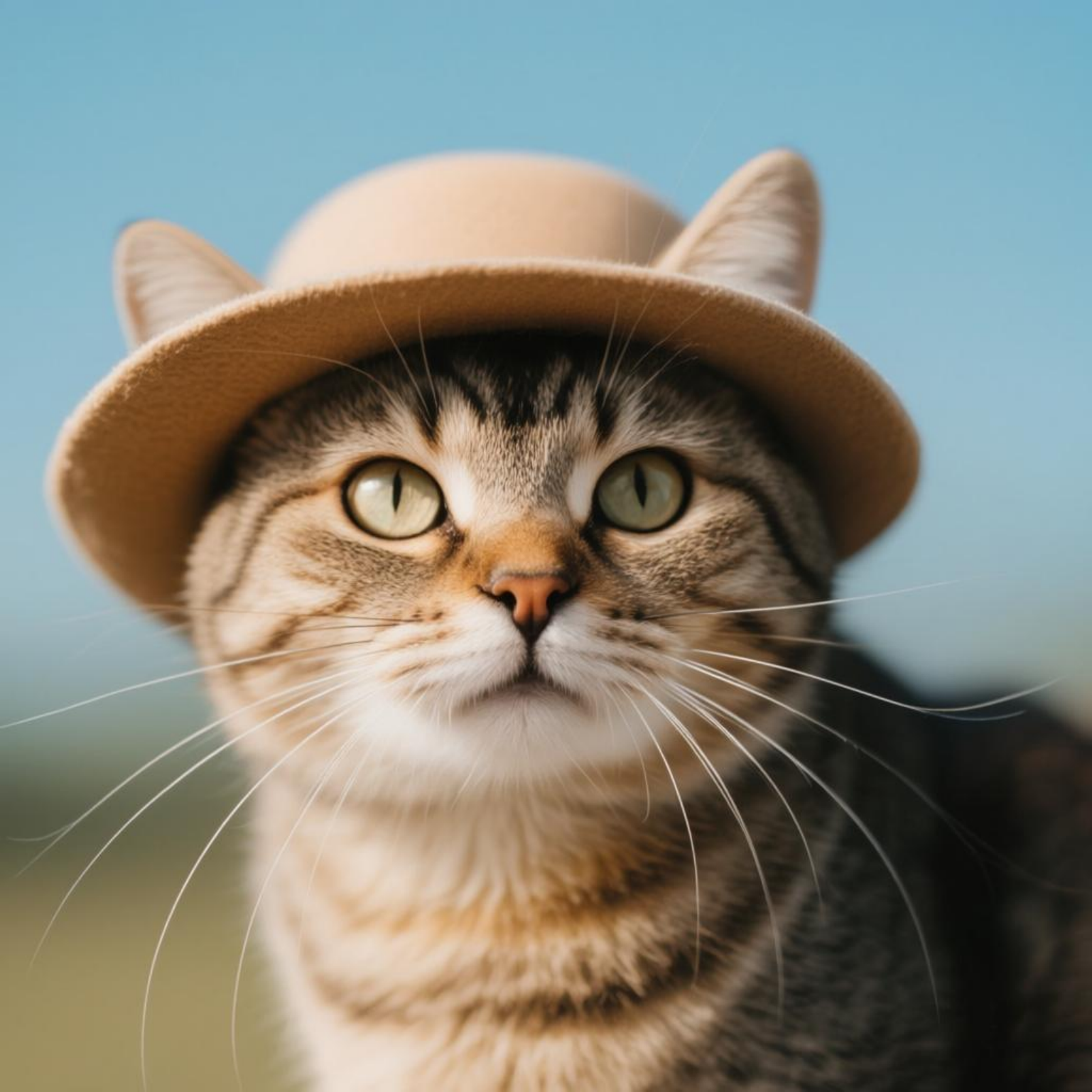} &
        \includegraphics[width=\linewidth]{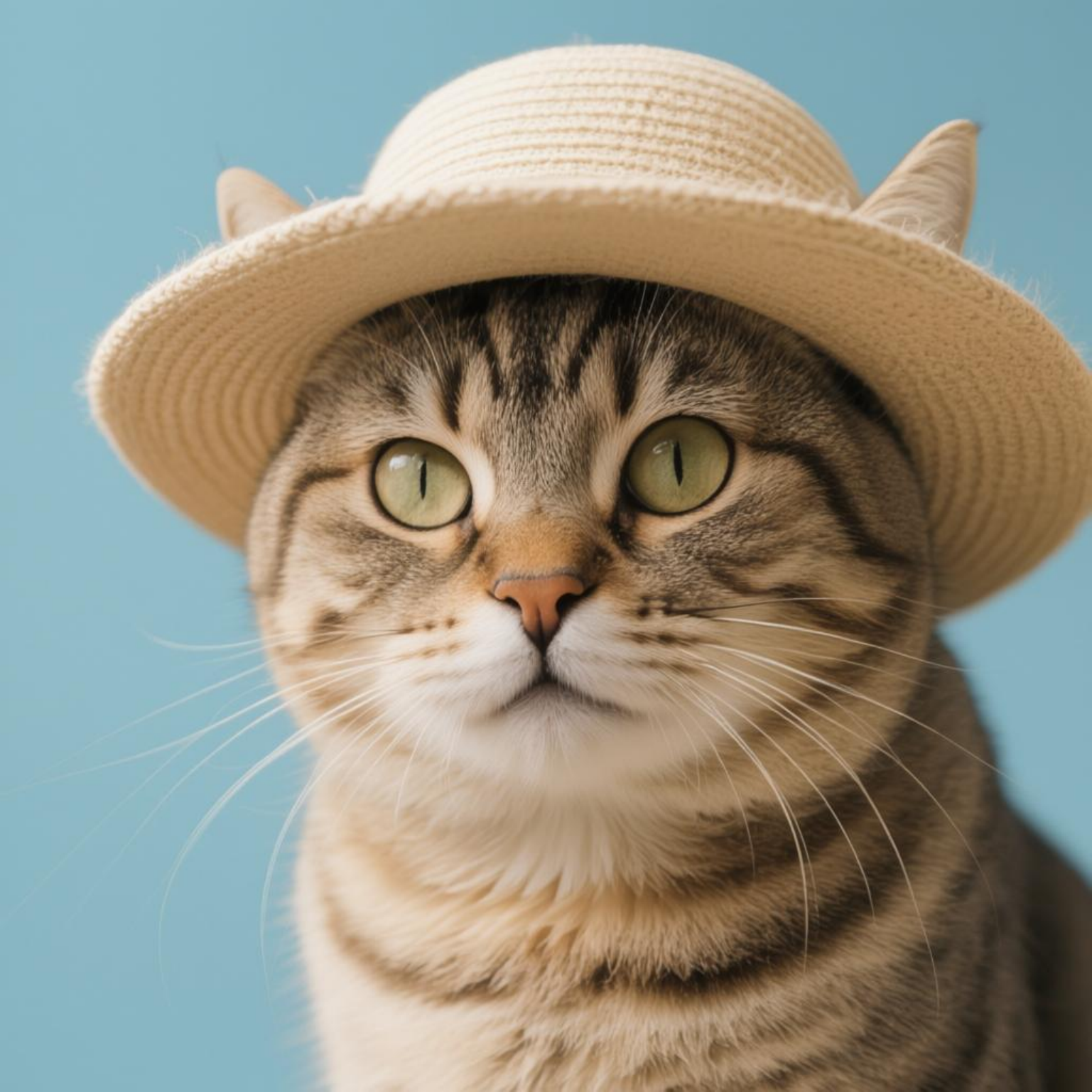} &
        \includegraphics[width=\linewidth]{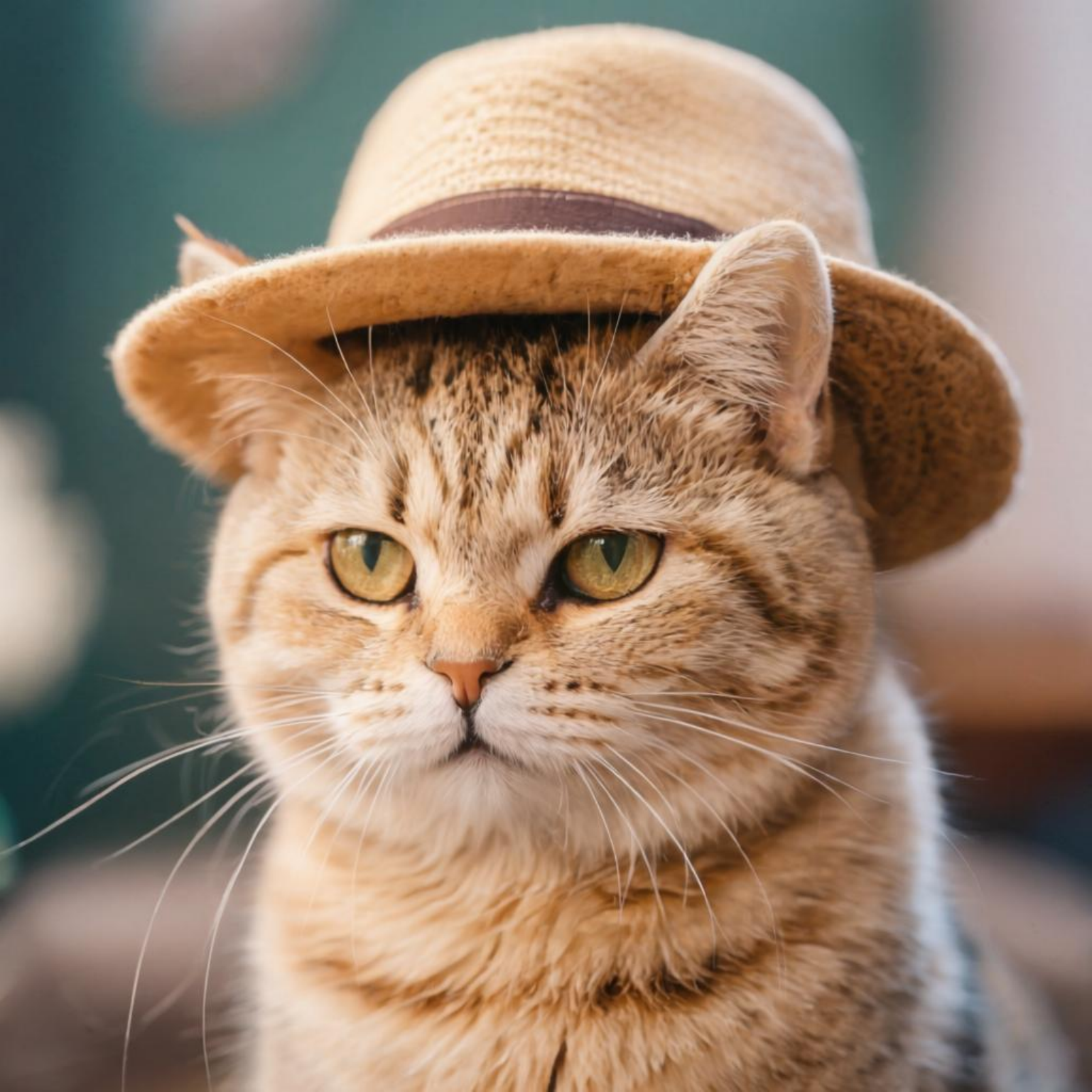} &
        \includegraphics[width=\linewidth]{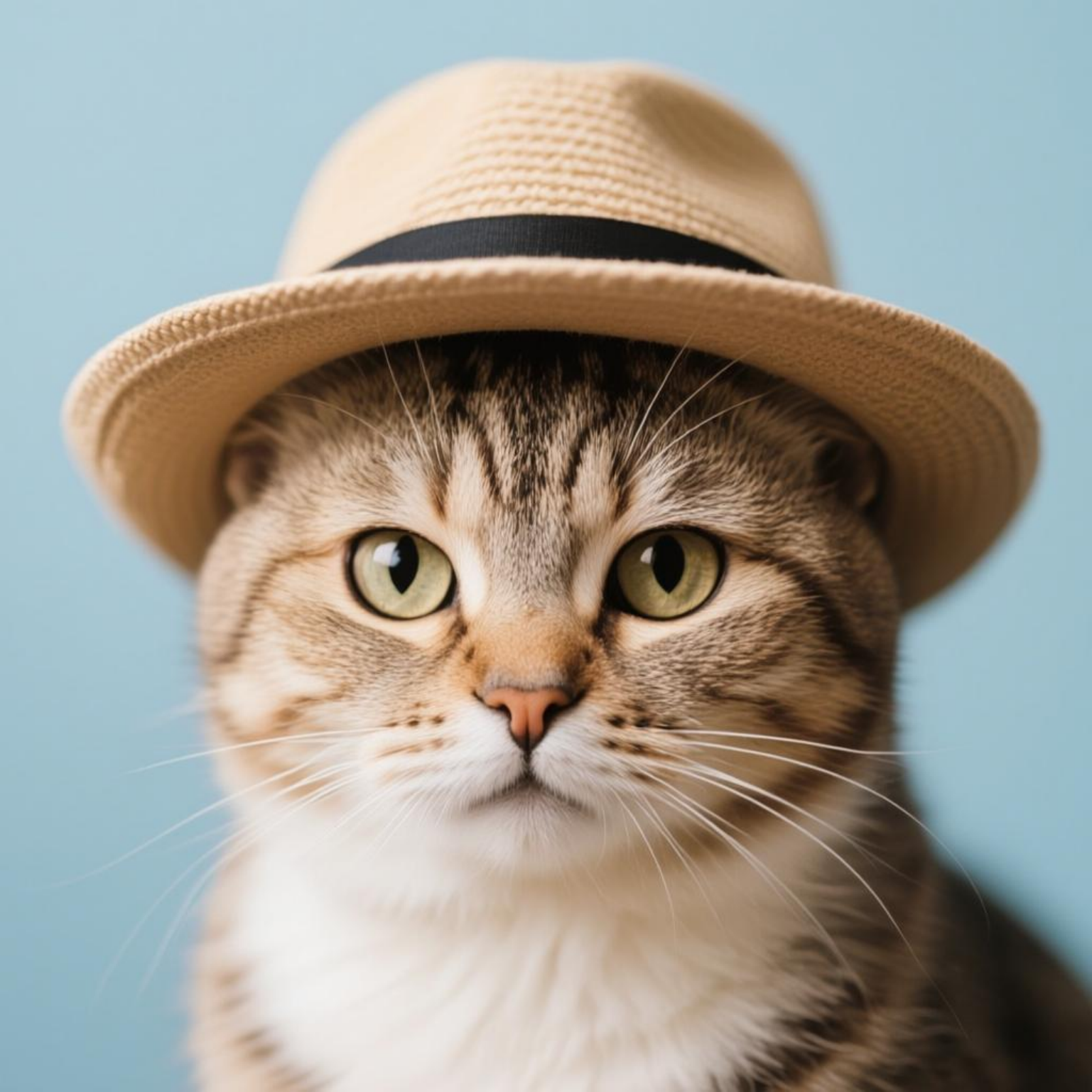} &
        \includegraphics[width=\linewidth]{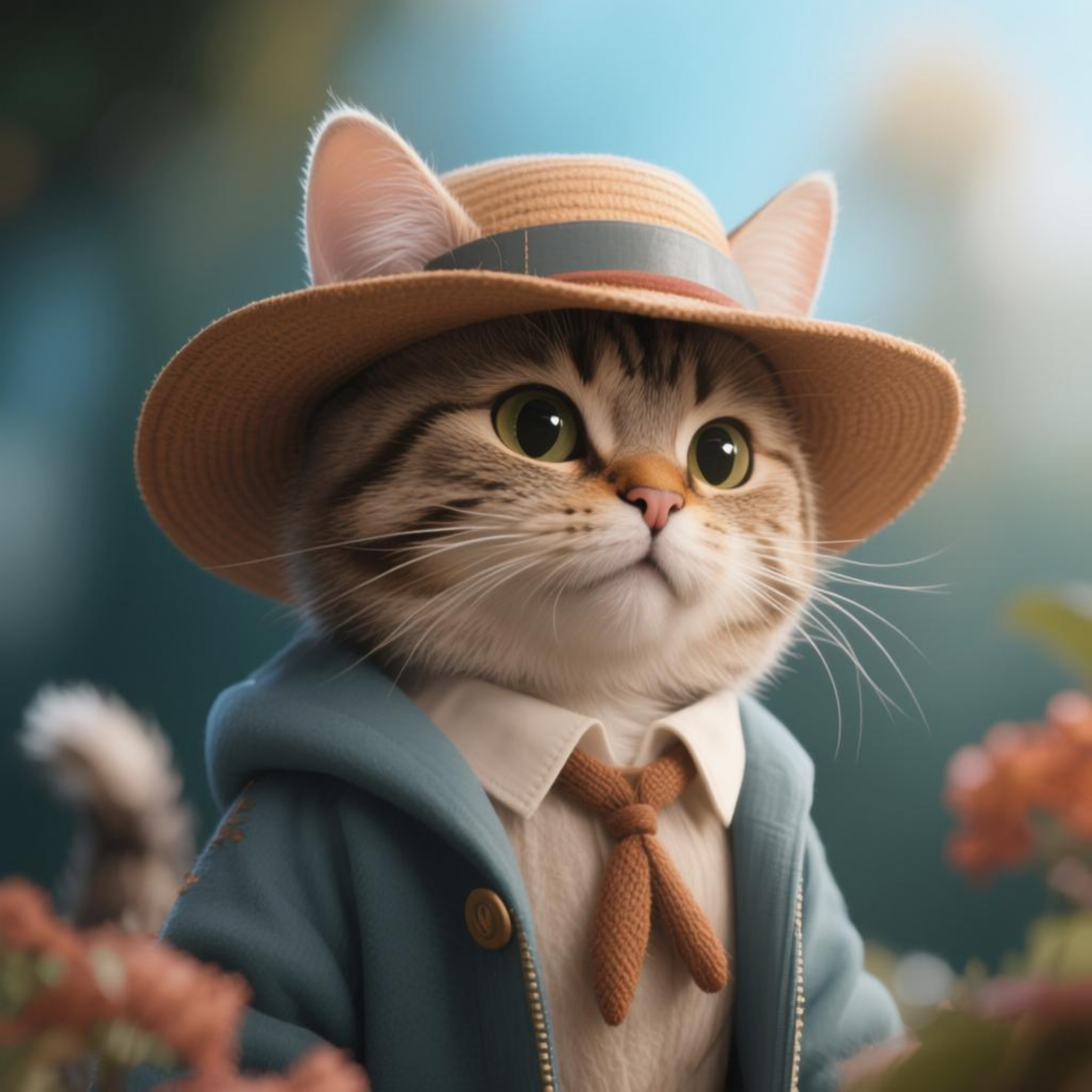} \\

        \multicolumn{5}{>{\centering\arraybackslash}p{\dimexpr\textwidth-2\tabcolsep}}{
            \small \textit{Prompt: cat with a hat}
        } \\
        \midrule

        \includegraphics[width=\linewidth]{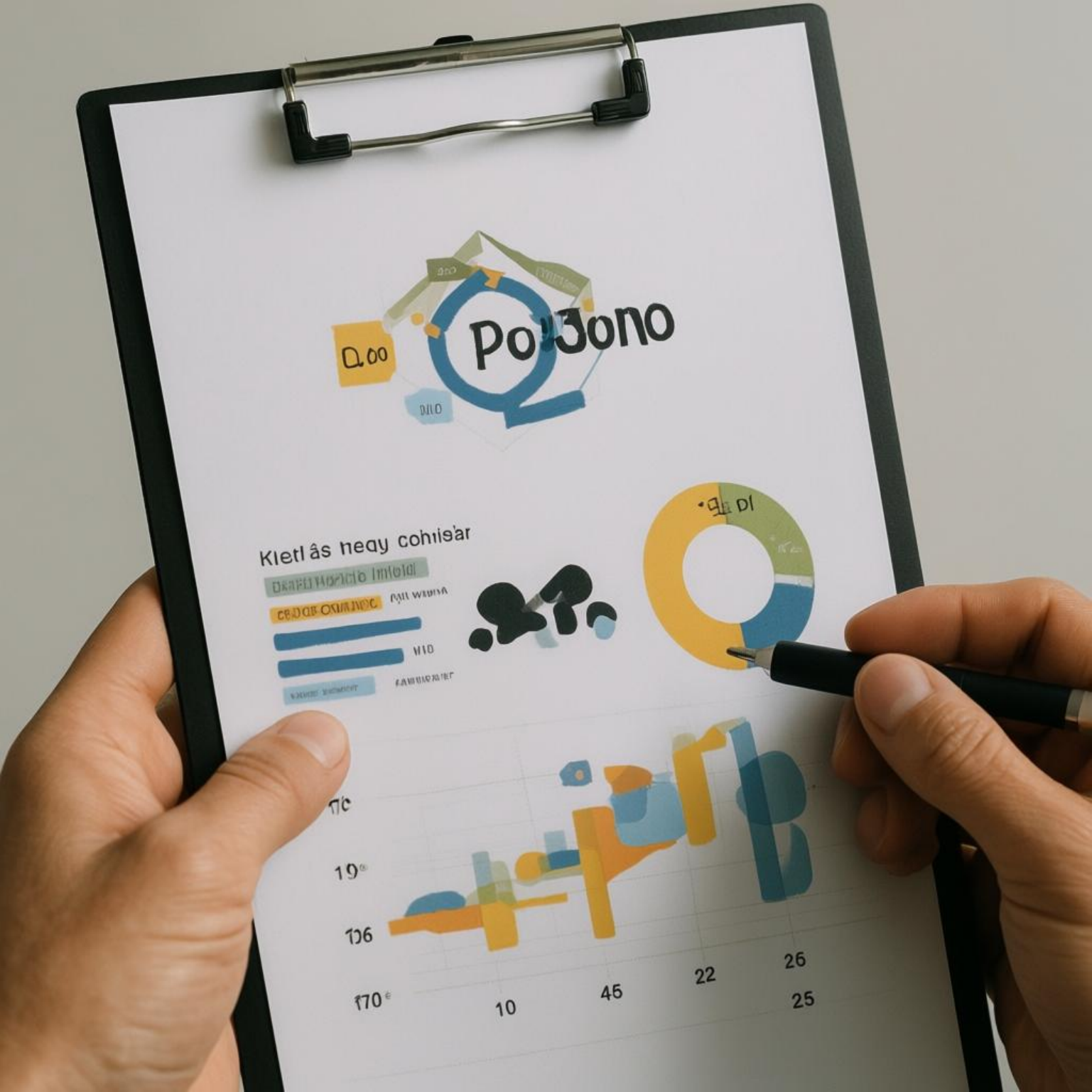} &
        \includegraphics[width=\linewidth]{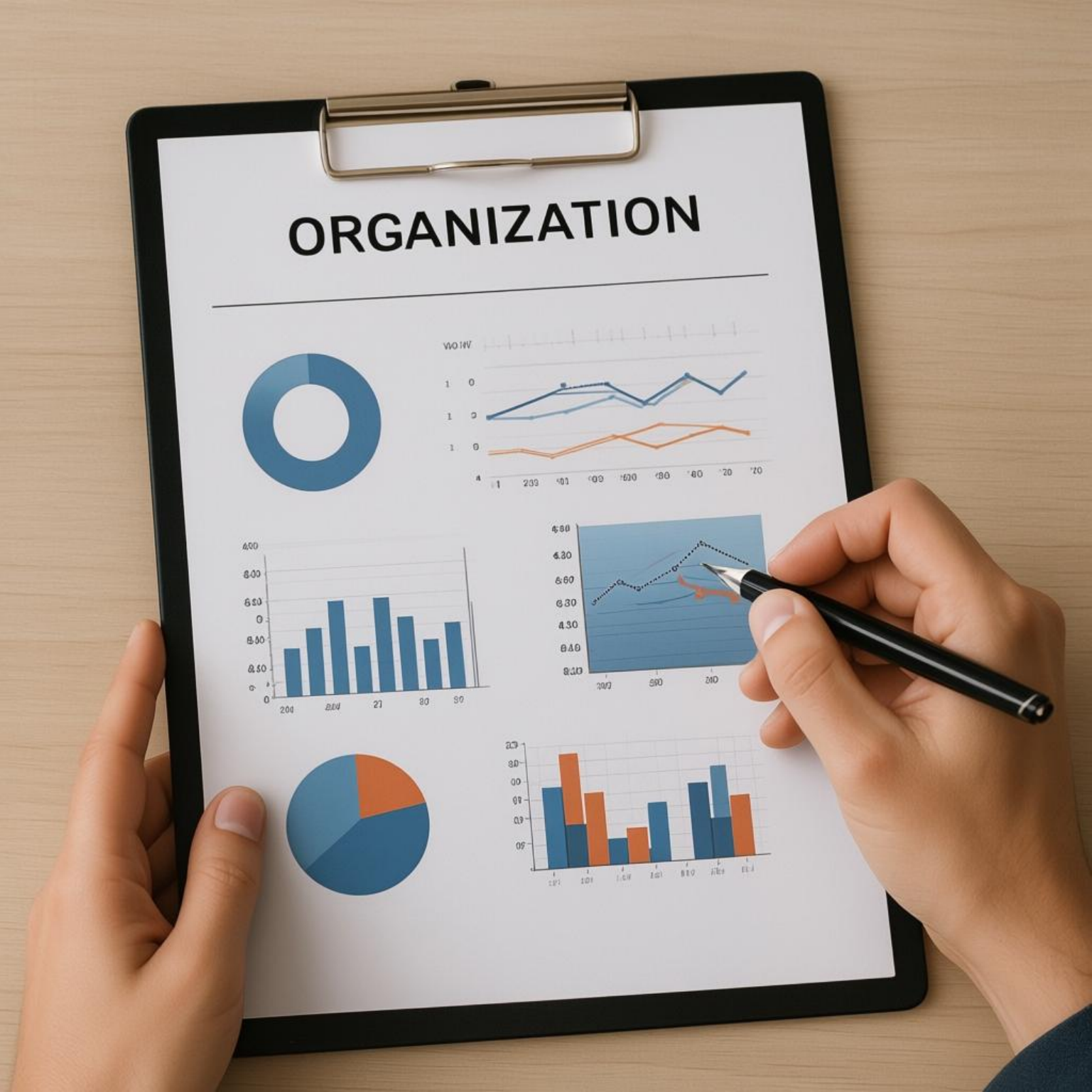} &
        \includegraphics[width=\linewidth]{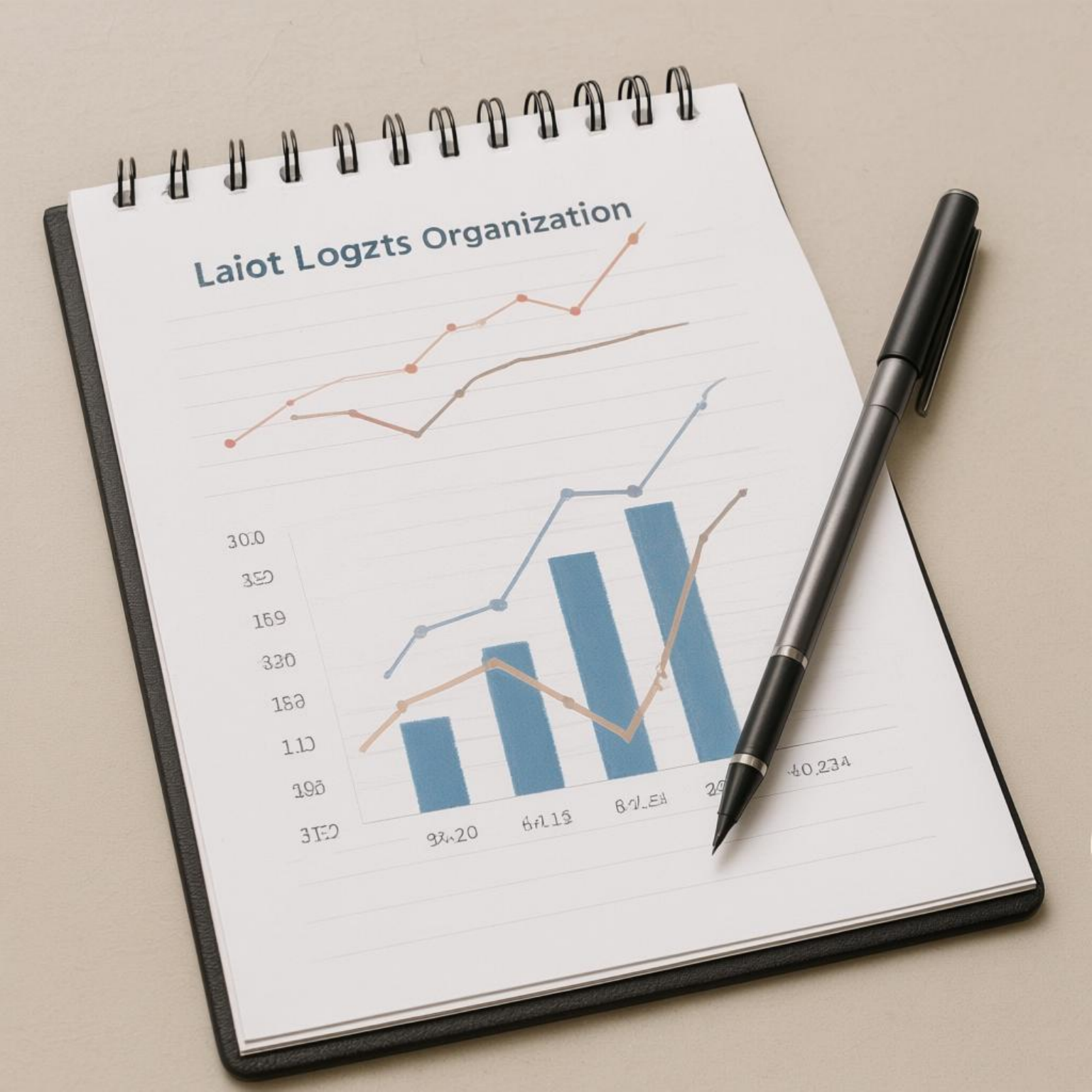} &
        \includegraphics[width=\linewidth]{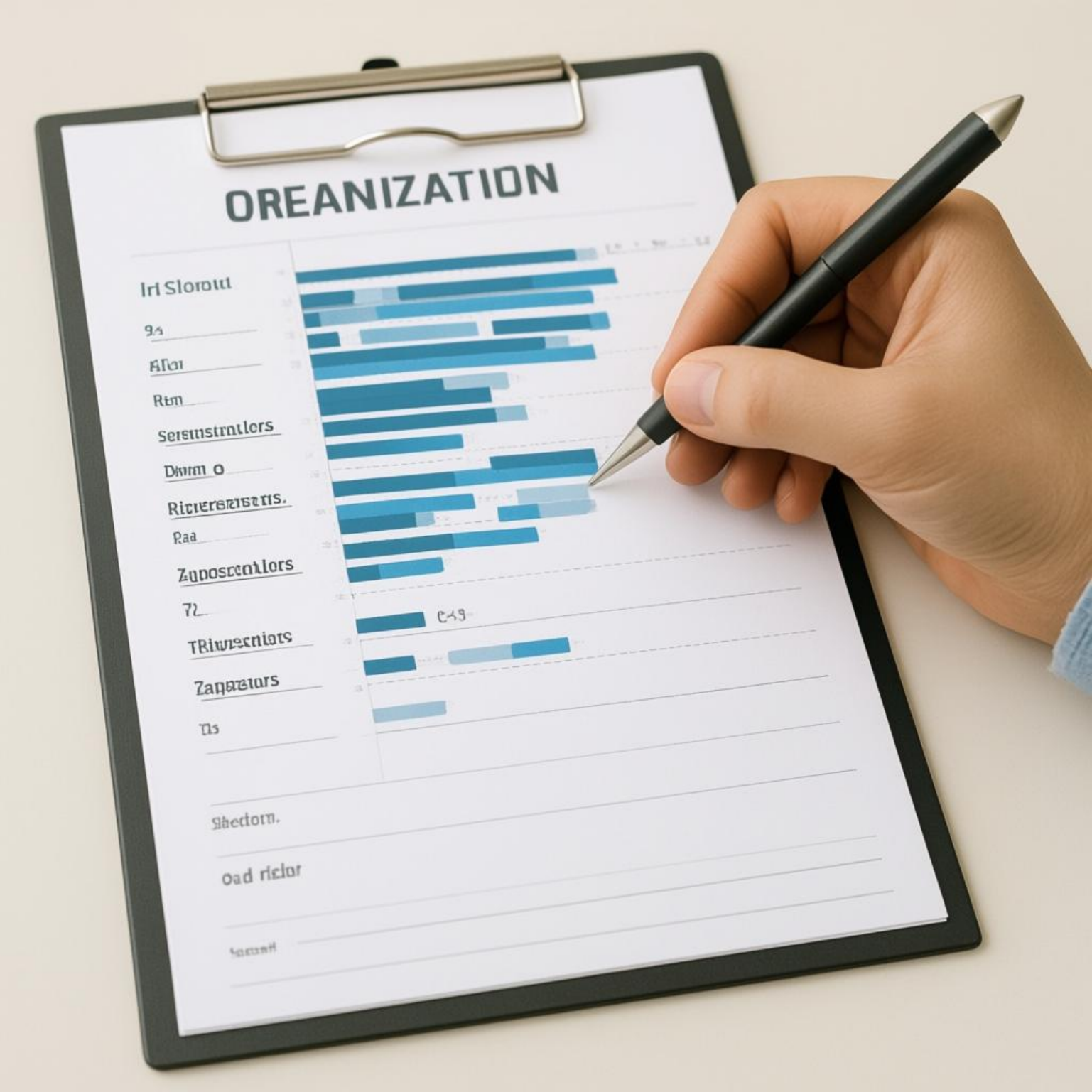} &
        \includegraphics[width=\linewidth]{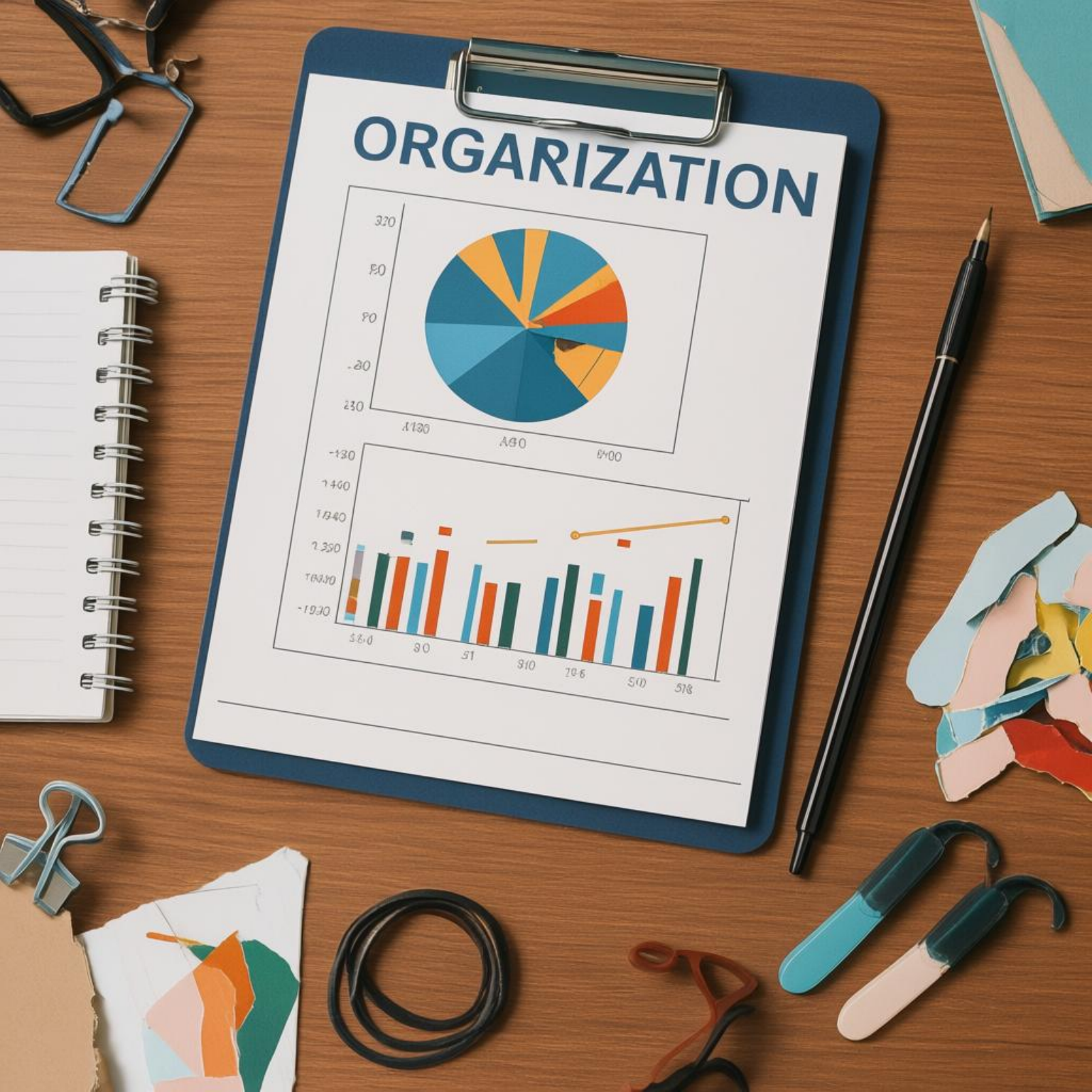} \\

        \multicolumn{5}{>{\centering\arraybackslash}p{\dimexpr\textwidth-2\tabcolsep}}{
            \small \textit{Prompt: Create a log for a statistical organization}
        } \\

        \bottomrule
        
    \end{tabularx}
    \label{fig:qualitative_qwen} 
    \caption{Visual comparison of Qwen-Image with $\text{NRE}=200$, targeting the PickScore reward model.} 
\end{figure}

\end{document}